\documentclass[10pt,twocolumn,letterpaper]{article}

\usepackage[pagenumbers]{iccv} %

\newcommand{\chapquote}[2]{%
  \vspace{0.5em} %
  \begin{flushleft}
    \textit{#1} %
    \\[-0.2em] %
    \hfill -- #2 %
  \end{flushleft}
}

\def\etal{\emph{et al}\onedot}

\def\etal{\emph{et al}\onedot}

\definecolor{iccvblue}{rgb}{0.21,0.49,0.74}
\usepackage[pagebackref,breaklinks,colorlinks,allcolors=iccvblue]{hyperref}
\usepackage{multirow}
\usepackage{appendix}

\def\paperID{10086} %
\def\confName{ICCV}
\def\confYear{2025}

\title{Piece it Together: Part-Based Concepting with IP-Priors \\[-0.1cm]}

\author{Elad Richardson\textsuperscript{1}
\qquad
Kfir Goldberg\textsuperscript{1,2}
\qquad
Yuval Alaluf\textsuperscript{1}
\qquad
Daniel Cohen-Or\textsuperscript{1}\\ \\
\textsuperscript{1}Tel Aviv University \qquad \textsuperscript{2}Bria AI
}

\begin{document}

\twocolumn[{
	\renewcommand\twocolumn[1][]{#1}
	\maketitle
 \vspace{-30pt}
	\begin{center}
        \setlength{\tabcolsep}{2pt}
        \includegraphics[width=0.975\textwidth] {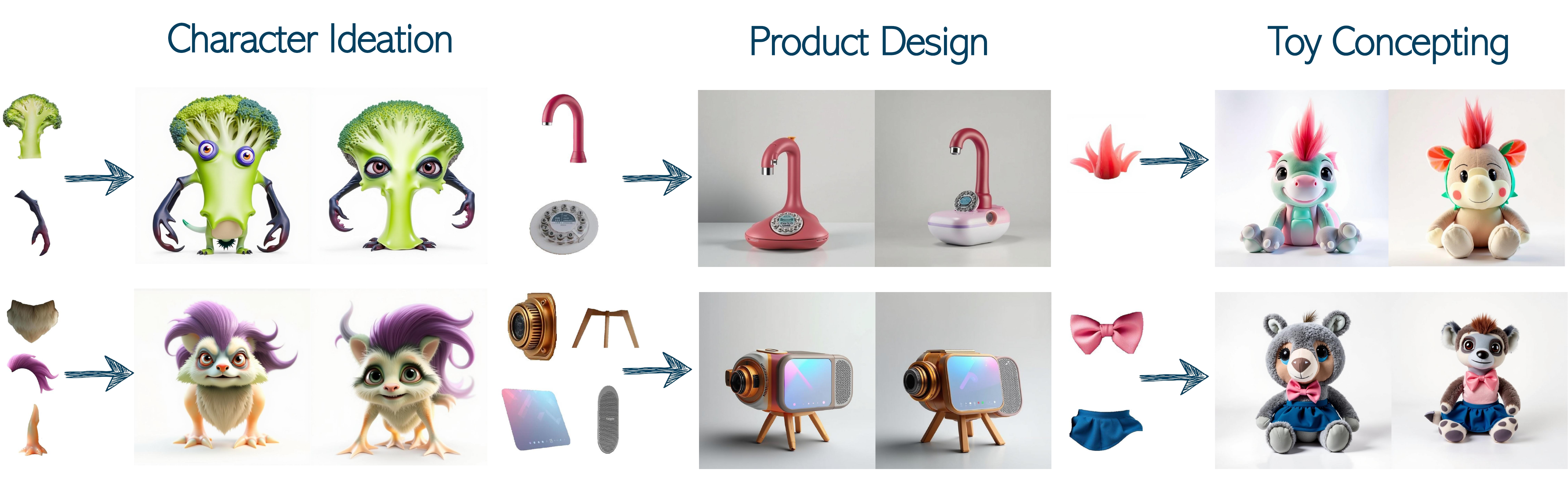}
        \vspace{-0.35cm}
        \captionof{figure}{
        Using a dedicated prior for the target domain, our method, Piece it Together (PiT), effectively completes missing information by seamlessly integrating given elements into a coherent composition while adding the necessary missing pieces needed for the complete concept to reside in the prior domain.
        \vspace{-0.2cm}
        }
    \label{fig:teaser}
	\end{center}
}]

\maketitle
\begin{abstract}
Advanced generative models excel at synthesizing images but often rely on text-based conditioning. Visual designers, however, often work beyond language, directly drawing inspiration from existing visual elements.  In many cases, these elements represent only fragments of a potential concept—such as an uniquely structured wing, or a specific hairstyle—serving as inspiration for the artist to explore how they can come together creatively into a coherent whole. Recognizing this need, we introduce a generative framework that seamlessly integrates a partial set of user-provided visual components into a coherent composition while simultaneously sampling the missing parts needed to generate a plausible and complete concept. Our approach builds on a strong and underexplored representation space, extracted from IP-Adapter+, on which we train IP-Prior, a lightweight flow-matching model that synthesizes coherent compositions based on domain-specific priors, enabling diverse and context-aware generations. Additionally, we present a LoRA-based fine-tuning strategy that significantly improves prompt adherence in IP-Adapter+ for a given task, addressing its common trade-off between reconstruction quality and prompt adherence.  Project page can be found at \url{https://eladrich.github.io/PiT/} 
\end{abstract}
    
\section{Introduction}\label{sec:intro}

\chapquote{``To create is to recombine''}{Fran\c{c}ois Jacob}

Recent advancements in image generation have significantly enhanced our ability to prototype and visualize new ideas. However, while modern generative models achieve impressive results, they are still typically conditioned on text, assuming that concepts can be fully articulated through language. In practice, however, artists and designers often work visually --- drawing from references, reconfiguring elements, and refining compositions in ways that cannot always be expressed through text alone~\cite{gonccalves2014inspires}. 
This has led to the development of techniques for conditioning generative models directly on image conditions~\cite{ye2023ip, ramesh2022hierarchical, razzhigaev2023kandinsky}. However, these methods alone offer limited control on how the visual concepts influence the generated results. 
Recognizing this, recent works have explored more advanced manipulation of visual concepts using generative models~\cite{vinker2023concept,dorfman2025ip,richardson2024pops, huang2023composer, rowles2024ipadapter}, offering a more interactive and intuitive creative process.

Expanding on this line of work, we ask: Can we simultaneously \textit{assemble} given visual components and \textit{sample} plausible completions for missing parts to create a coherent whole? To this end, we propose a model that dynamically adapts to user inputs, assembling provided elements into a coherent structure while inferring missing components in a manner consistent with the provided context.

\begin{figure}
    \centering
    \setlength{\tabcolsep}{0.5pt}
    \addtolength{\belowcaptionskip}{-5pt}
    \renewcommand{\arraystretch}{0.5}
    {\small
    \begin{tabular}{c  c c c c c @{\hspace{0.1cm}} c}

        \raisebox{-.5\height}[0pt][0pt]{\includegraphics[width=0.075\textwidth]{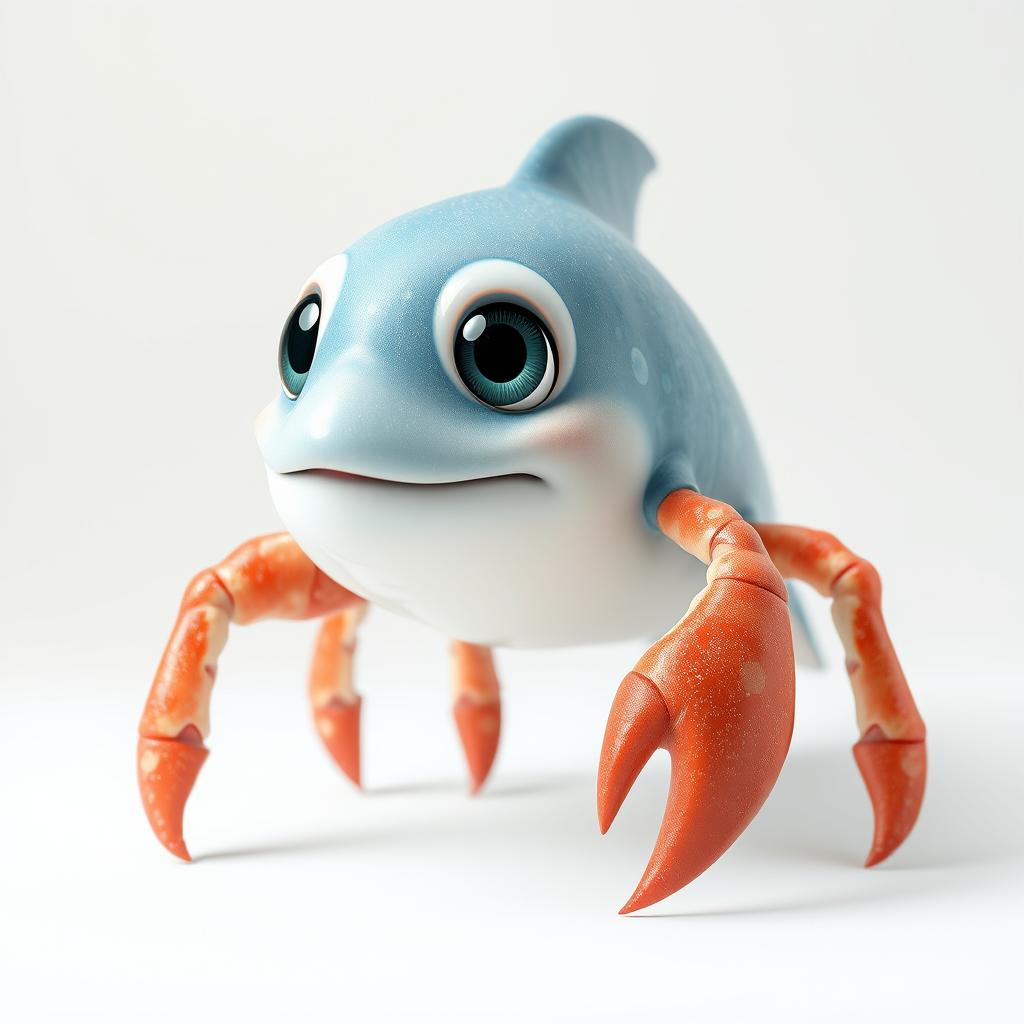}} &

        \includegraphics[width=0.075\textwidth]{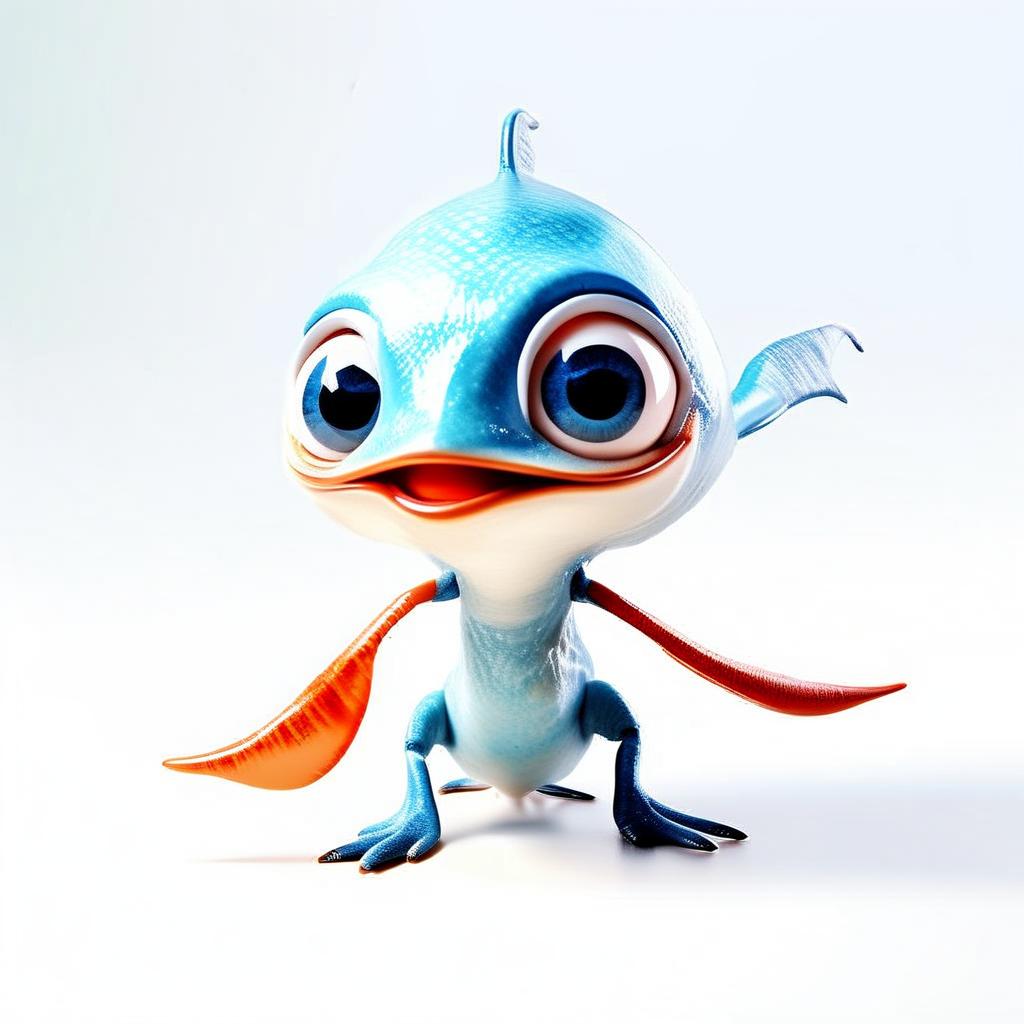} &
        \includegraphics[width=0.075\textwidth]{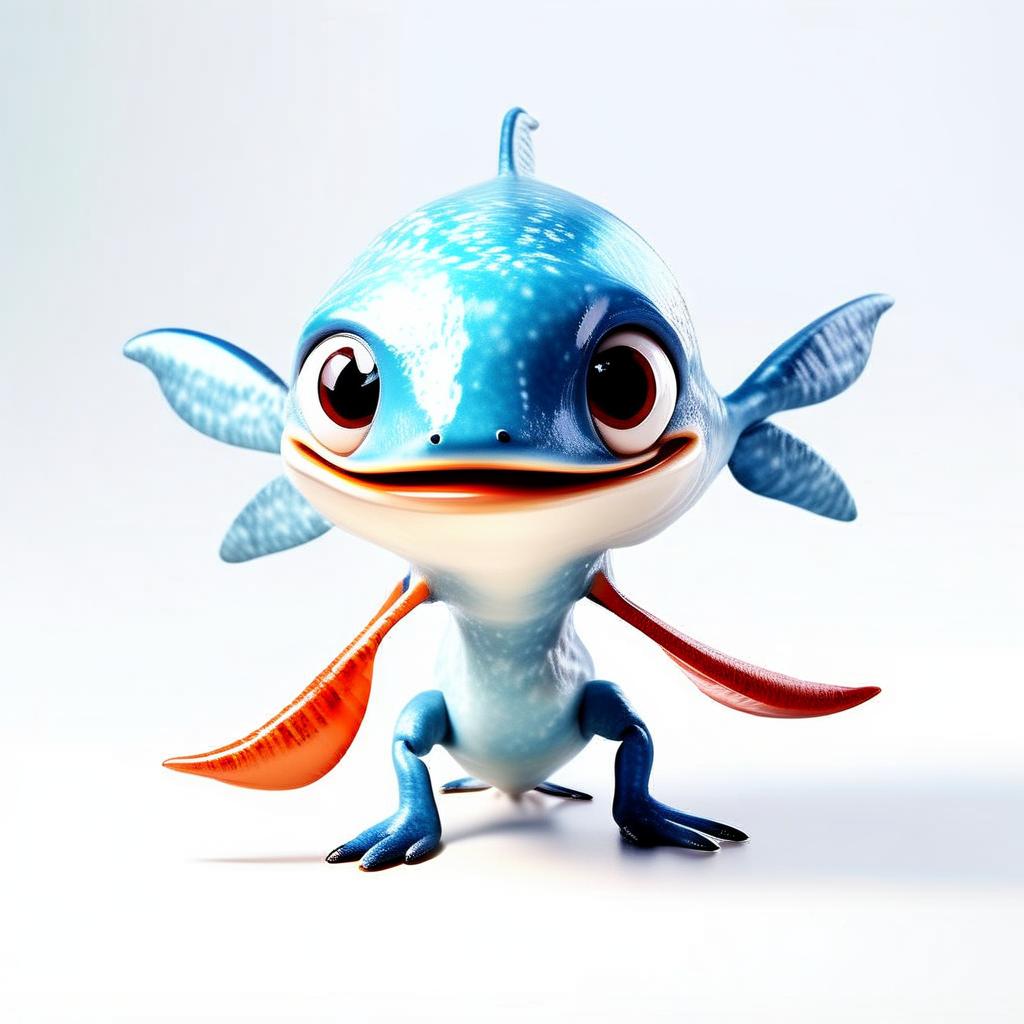} &  \includegraphics[width=0.075\textwidth]{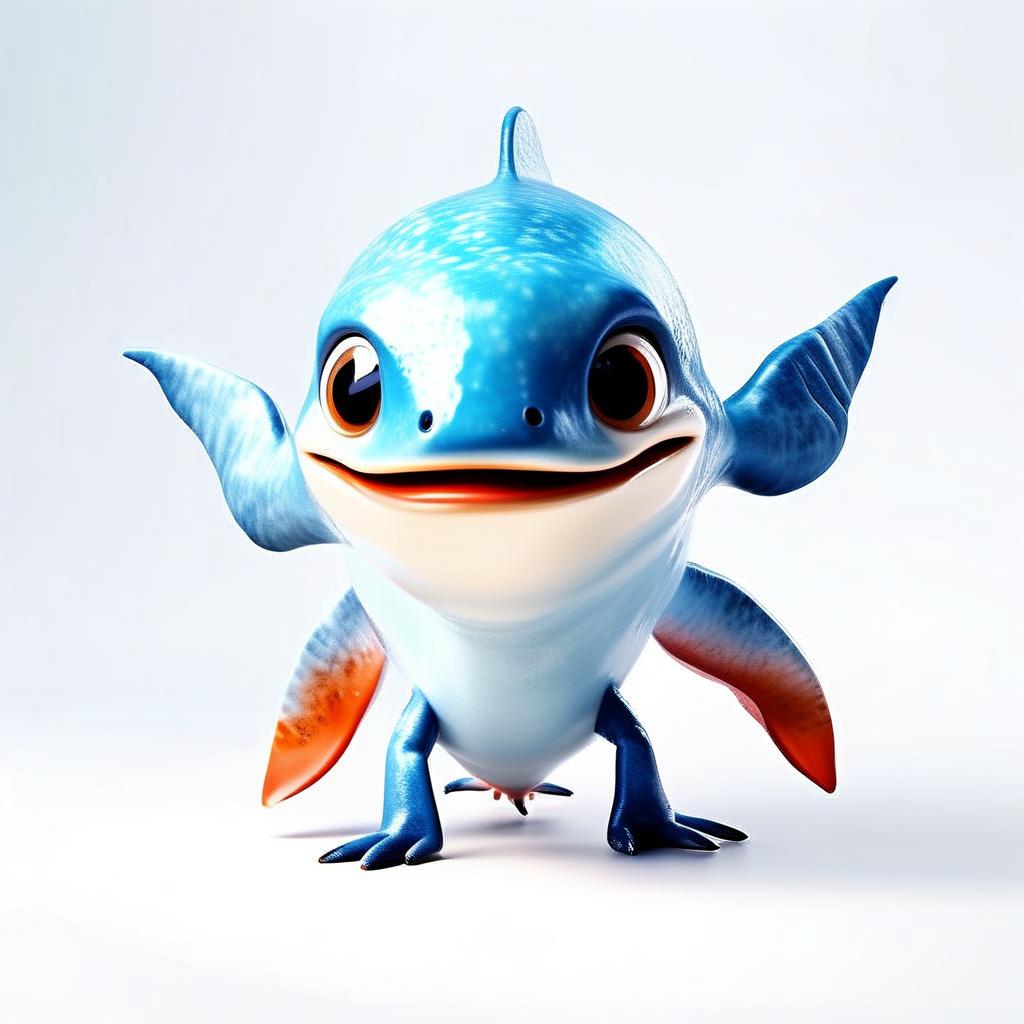} &
        \includegraphics[width=0.075\textwidth]{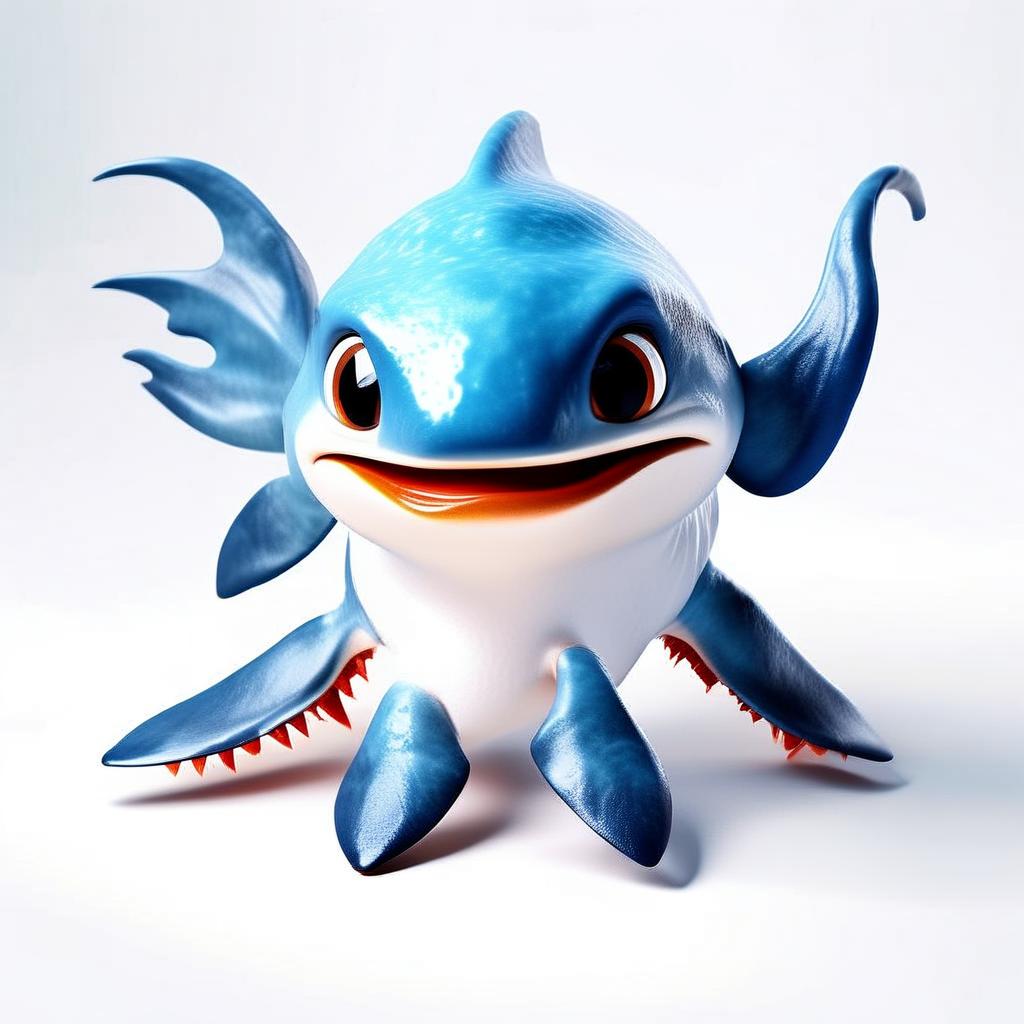} &
        \includegraphics[width=0.075\textwidth]{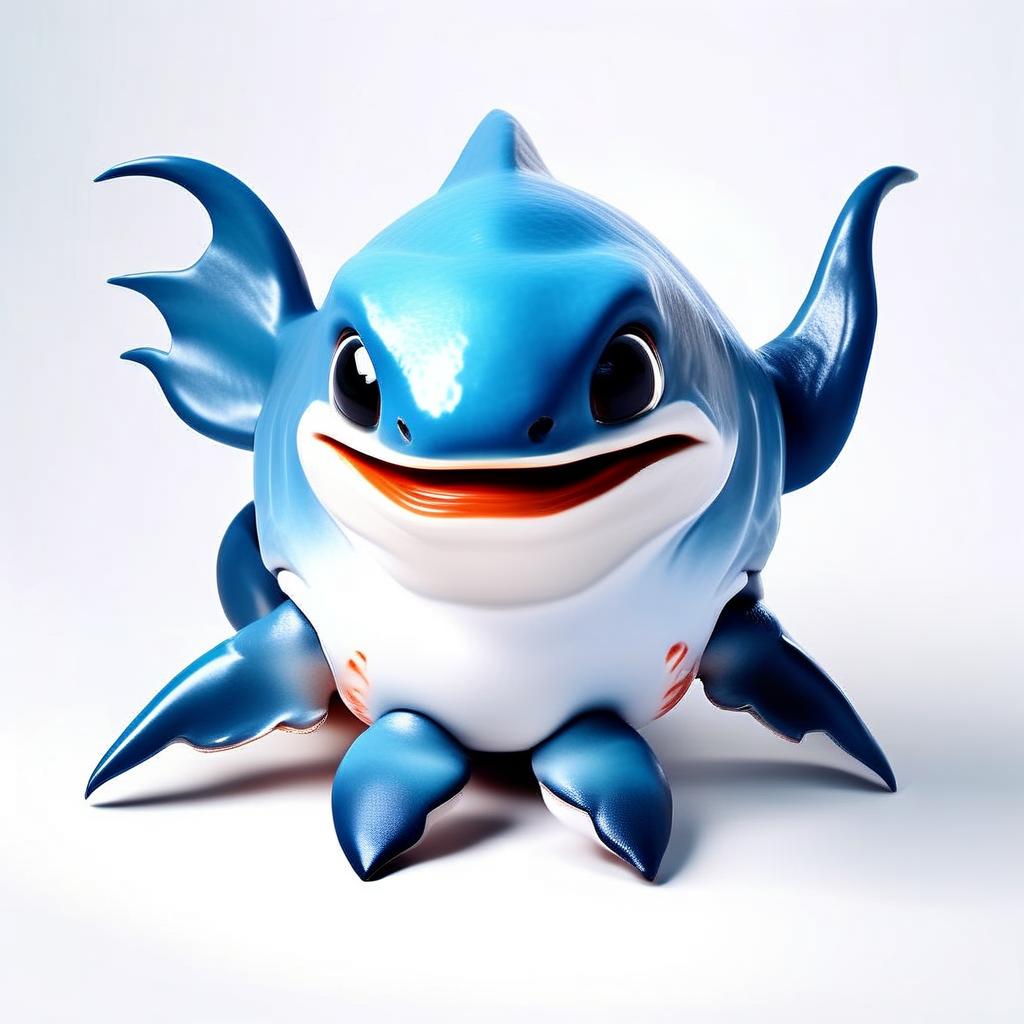}  &
        \raisebox{0.035\linewidth}{\rotatebox[origin=t]{90}{\begin{tabular}{c@{}c@{}c@{}c@{}} CLIP \end{tabular}}} 

        \\

        &
        \includegraphics[width=0.075\textwidth]{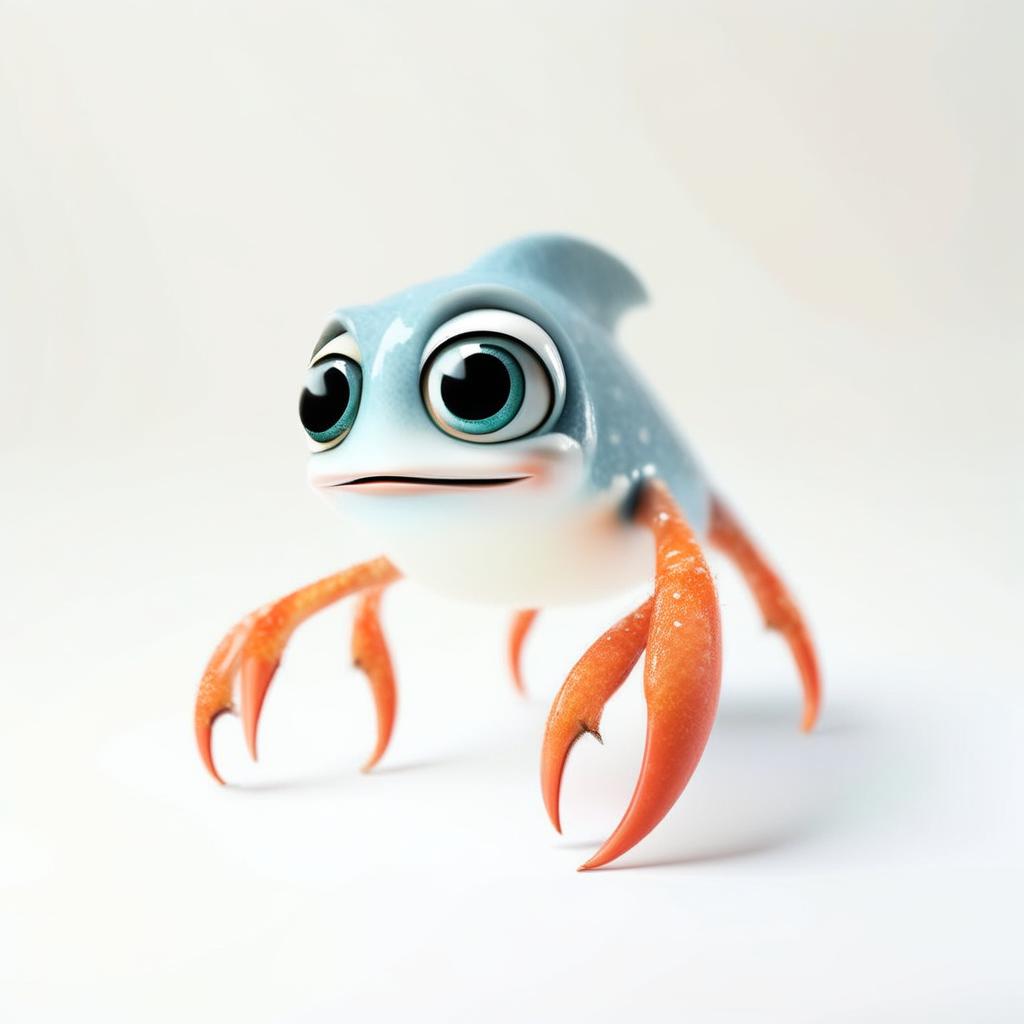} &
        \includegraphics[width=0.075\textwidth]{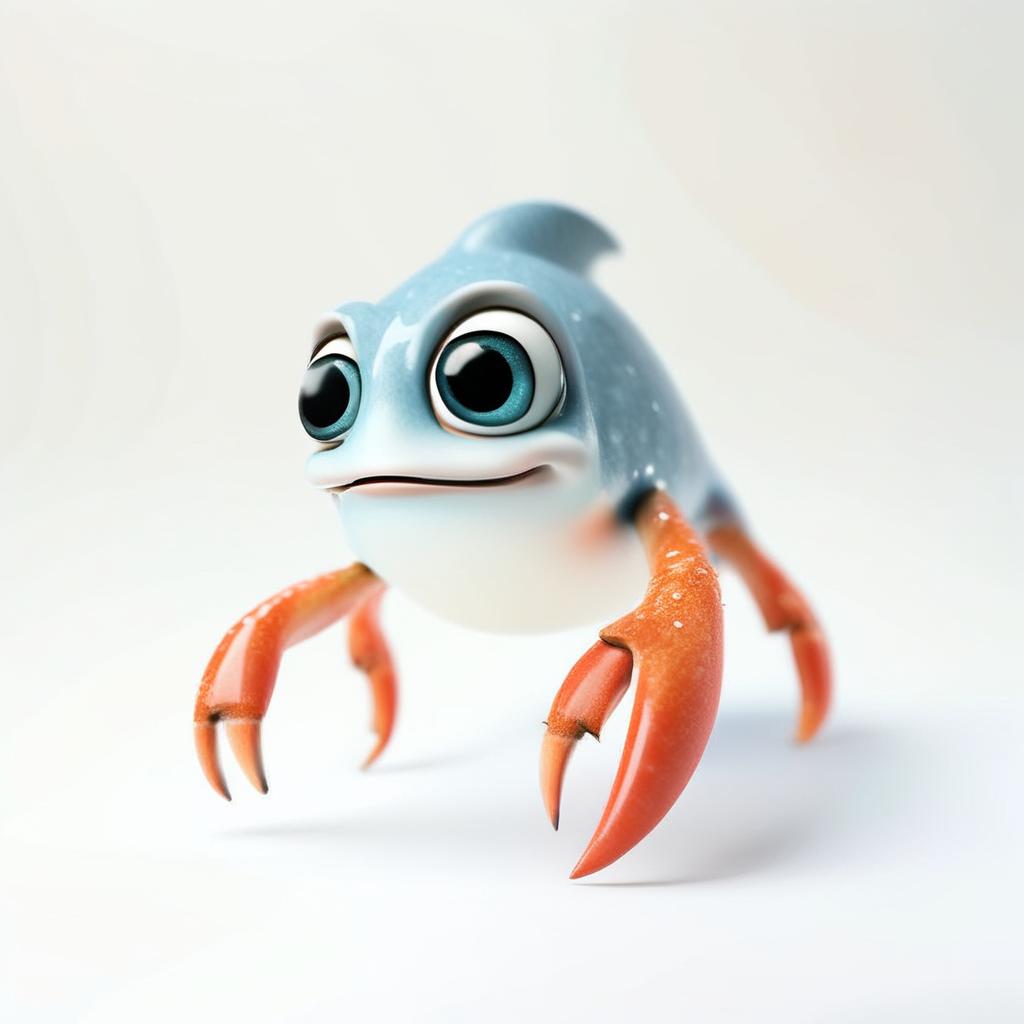} &  \includegraphics[width=0.075\textwidth]{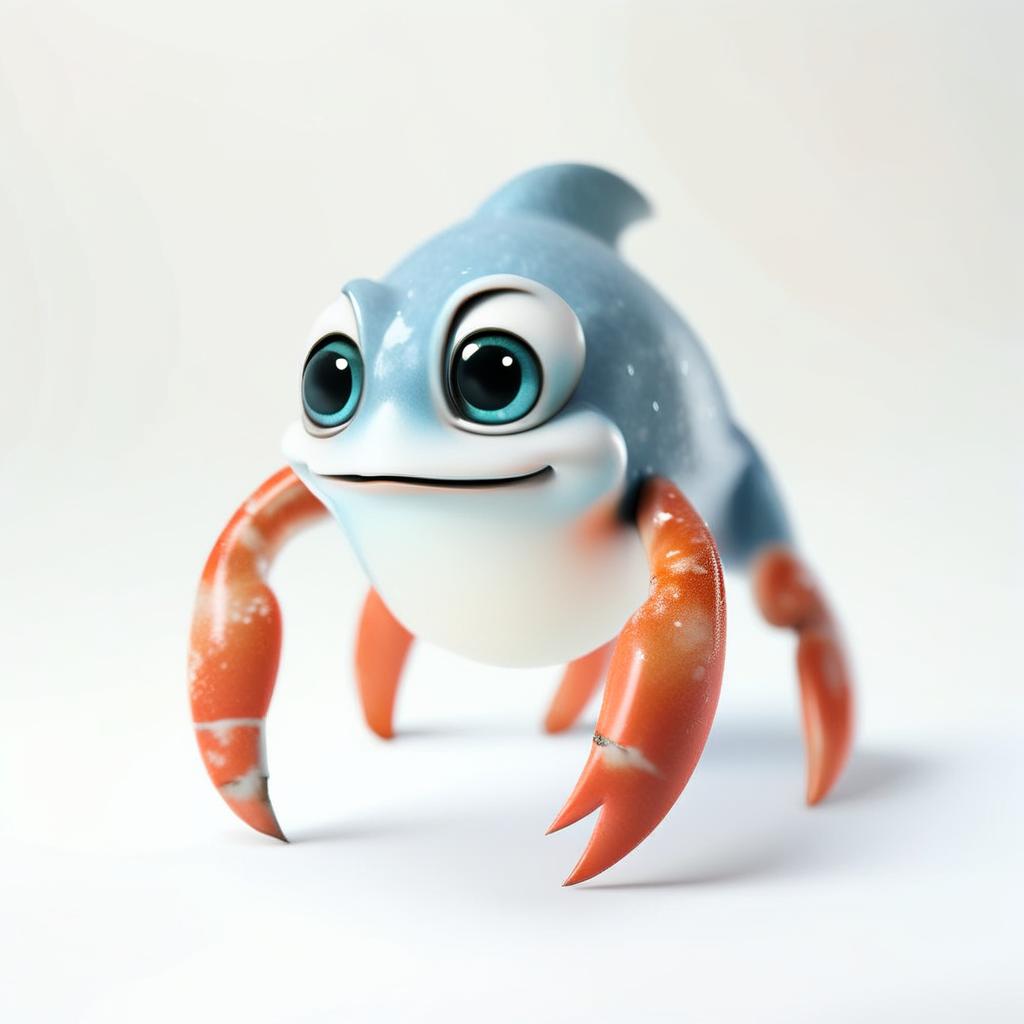} &
        \includegraphics[width=0.075\textwidth]{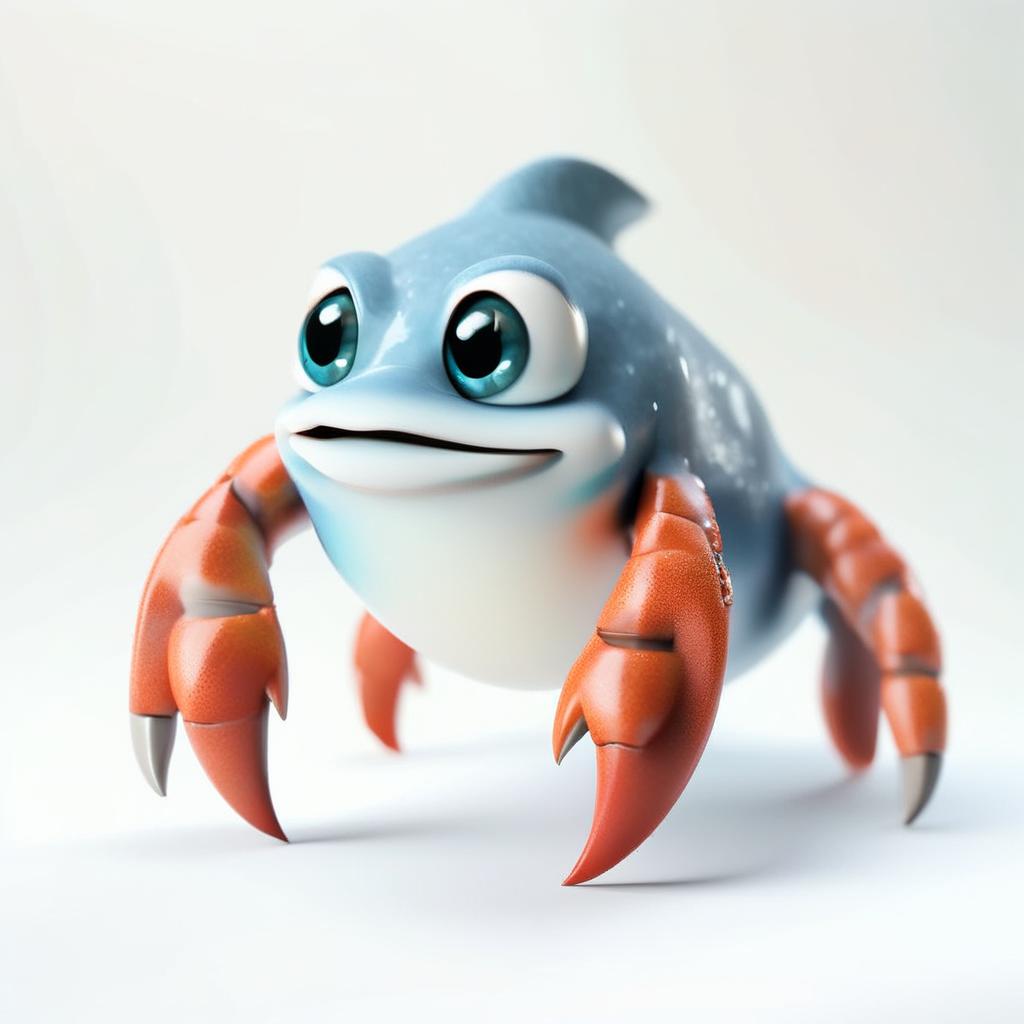} &
        \includegraphics[width=0.075\textwidth]{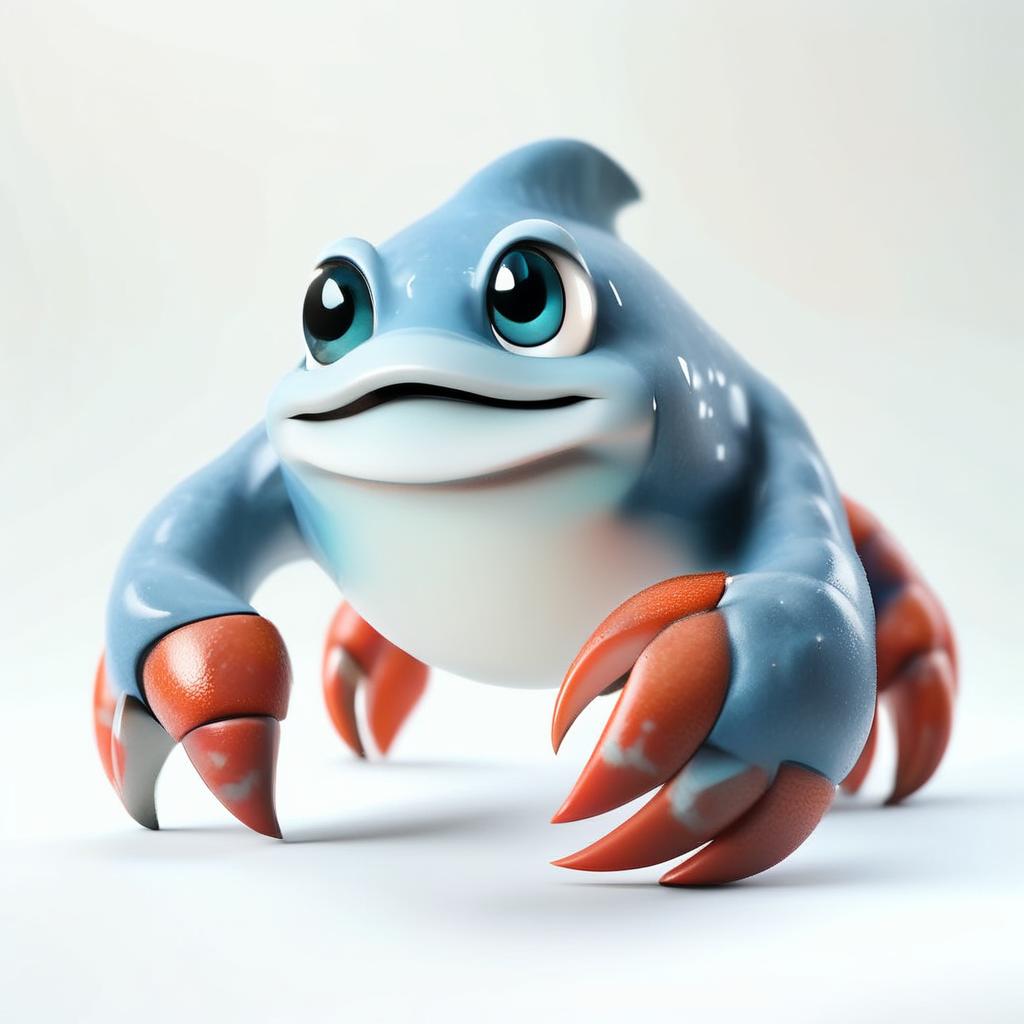}  &
        \raisebox{0.035\linewidth}{\rotatebox[origin=t]{90}{\begin{tabular}{c@{}c@{}c@{}c@{}} $\mathcal{IP}^+$  \end{tabular}}} 
        \\
        Input & \multicolumn{5}{c}{Scrawny $\rightarrow$ Muscular} 

    \end{tabular}
    }
    \vspace{-0.2cm}
    \caption{\textbf{Semantic Manipulation in CLIP Space vs. $\mathcal{IP}+$ Space.}  We encode the input image (left) into two different embedding spaces, modify its latent representation by traversing each space, and render the edited image using SDXL~\cite{podell2023sdxl}. As shown, CLIP struggles to both reconstruct the concept and follow the desired edit, whereas in $\mathcal{IP}+$ space, the rendered images are faithful both to the concept and the desired edit across the entire range.
    }
    \label{fig:semantic_intro}
\end{figure}

The choice of representation space for our model inputs is crucial. Inspiration Tree~\cite{vinker2023concept} and ConceptLab~\cite{richardson2024conceptlab} use learnable tokens within the text encoder's input space~\cite{gal2022image}. While effective for their respective tasks, this optimization-based approach would hinder our ability to use our method efficiently at inference time.
Alternatively, pOps~\cite{richardson2024pops} and IP-Composer~\cite{dorfman2025ip} operate in CLIP space.
This allows them to efficiently encode visual concepts via a pretrained CLIP encoder and is well suited for semantic manipulations. However, the CLIP space is limited in its ability to preserve complex concepts, resulting in a loss of details~\cite{richardson2024pops,ramesh2022hierarchical}. 
This intuitively stems from the fact that CLIP was never trained to reconstruct images but rather to learn a joint representation space for text and images~\cite{radford2021learning}. While this encourages a semantic representation, it does not require the representation to encode visual details that cannot be easily described through text.
To improve on this, we explore alternative spaces and ultimately converge on the internal representation of IP-Adapter+. This adapter extends the popular model from~\cite{ye2023ip}. While not formally described in a paper, it has gained popularity for its improved reconstruction quality. 
We show that using this $\mathcal{IP}^+$ space not only results in improved reconstructions but also retains the ability to perform semantic manipulations and thus can serve as a new representation for visual concepts, see~\Cref{fig:semantic_intro}.

With our chosen representation in place, we turn to training a part-conditioned model on a set of generated samples from a given target domain. We train our model to sample both conditionally and unconditionally from that domain while interpreting any given input within that context. For instance, a model tuned on monsters and creatures would always generate a creature from that domain. Importantly, this strong prior allows the artist to reinterpret everyday objects as potential parts within the learned domain (e.g., the broccoli input in~\Cref{fig:teaser}) and sample multiple plausible results for the same set of inputs.
In essence, the trained model captures a prior distribution over the target domain in the $\mathcal{IP}^+$ space and is therefore dubbed an \textit{IP-Prior} model.

Finally, once a complete visual concept has been generated using an IP-Prior, it can be rendered as an image by passing it to a pretrained image generation model~\cite{podell2023sdxl}.
Ideally, at this stage, we would additionally like to introduce additional conditions --- for example, incorporating a text prompt to place the generated concept within a specific scene.
Unfortunately, a key limitation of IP-Adapter+ is its inherent trade-off between reconstruction quality and prompt adherence. We hypothesize this arises from the expressiveness of the $\mathcal{IP}^+$ space, which makes the text conditioning redundant during fine-tuning, and propose a simple LoRA-based mechanism for re-enabling text conditioning on the generated concepts. Together, this results in PiT, a flexible pipeline that first facilitates concept ideation and then renders those concepts as high-quality images.

\section{Related Work}\label{sec:related}

\paragraph{Image-Conditioned Generation.}
While text has become the de facto interface for generating visual content~\cite{po2023state,yin2024survey,ramesh2022hierarchical,nichol2021glide,balaji2023ediffi,kandinsky2,ding2022cogview2,saharia2022photorealistic,rombach2021highresolution}, its ability to describe specific visual concepts can be limited. 
As a result, various approaches have been proposed to incorporate image inputs into generative models. These techniques can be roughly grouped based on the mechanism they use to incorporate the visual information into the generative network. 

In optimization-based personalization methods, a single visual concept is encoded into the text embedding inputs~\cite{gal2023image,voynov2023p,alaluf2023neural} or directly into the network weights~\cite{ruiz2022dreambooth,kumari2022customdiffusion,tewel2023keylocked}. The model can then generate images of the given concept under different prompts. 
Numerous efforts have been made to turn these personalization methods optimization-free. These methods train an encoder to directly map a concept into some form of compact representation that is then passed to the model ~\cite{gal2023encoderbased,ruiz2024hyperdreambooth,arar2023domain,wei2023elite,ye2023ip-adapter,zeng2024jedi,shi2023instantbooth,patashnik2025nested}. Specifically, in~\cite{ye2023ip} the input image is encoded through CLIP and passed to a set of decoupled cross-attention layers. This has proven to be an effective conditioning mechanism that has gained popularity in the community.

\vspace{-0.25cm}
\paragraph{Multi-Concept Generation.}
Successfully depicting multiple concepts in a single image remains a challenging task, even in text-based generation ~\cite{chefer2023attendandexcite,phung2024grounded,dahary2024yourself,li2023gligen,gu2023mix,parmar2025object}. This challenge becomes even more pronounced in image-conditioned generation. Consequently, a substantial body of work has focused on improving conditioning techniques to support multiple inputs. In essence, these methods build upon existing image-conditioning approaches, extending them to better support multiple visual constraints.

Some approaches aim to solve an optimization problem resulting in a disentangled representation for each visual concept~\cite{avrahami2023break,garibi2025tokenverse,kumari2022customdiffusion}. Another line of work leverages LoRA~\cite{hu2022lora} to learn new concepts and aims to apply multiple LoRA modules in conjunction~\cite{yang2024lora,kong2024omg,po2024orthogonal,gu2023mixofshow}, often through spatial maps or localized prompts. Others follow the encoder-based approaches and inject a representation of the visual information into the model while accounting for the different objects~\cite{xiao2024fastcomposer,hao2024conceptexpress,patel2024lambdaeclipse, parmar2025object}. 
Alternatively, numerous works utilize the advancement in the research on vision-language models ~\cite{li2023blip,liu2023visual, yang2024qwen2} to define multimodal prompts, interleaving multiple reference images with an input text prompt~\cite{xiao2024omnigen,pan2023kosmos,zhou2024transfusion,xie2024showo}. 

Crucially, the aforementioned works primarily focus on embedding complete objects into a scene where all elements are predefined, eliminating the need for the model to generate novel content. In contrast, our approach generates new concepts from partial user-provided parts, requiring the model to both integrate the given inputs and infer missing details to form a single, coherent concept.

\vspace{-0.25cm}
\paragraph{Visually Inspired and Creative Generation.}
The exploration of human creativity within computer graphics has been a significant area of research, with numerous works investigating how computational tools can enhance the creative design process~\cite{hertzmann2018can,elhoseiny2019creativity,kantosalo2014isolation,wang2024can,Oppenlaender_2022,esling2020creativity,xu2011photo}. 
Interestingly, a fundamental aspect of creativity is the ability to leverage prior knowledge to generate novel ideas~\cite{bonnardel2005towards,wilkenfeld2001similarity,eckert2000sources}. Aligned with these observations, recent works have started to explore how strong generative models can serve to inspire creative generation and exploration.

More specifically, Inspiration Tree~\cite{vinker2023concept} showed that one can decompose a visual concept into distinct attributes, organizing them hierarchically in a tree structure by building on the literature of concept personalization. This was extended in~\cite{lee2024languageinformed}, which learns disentangled concept representations along language-informed axes. 
In the context of creative generation, ConceptLab~\cite{richardson2024conceptlab} uses guidance from a VLM to learn a new token embedding representing novel concepts within a given broad category. 
However, while their approach leverages VLMs for the creative generation process, they do not offer control over the visual attributes of the learned concept.

In the context of generating images inspired by multiple visual concepts, it has been shown that generative models that are conditioned on the CLIP image embeddings, such as IP-Adapter~\cite{ye2023ip} or Kandinsky~\cite{kandinsky2}, allow one to use the CLIP embedding space to manipulate and compose different concepts together~\cite{richardson2024pops,dorfman2025ip,patel2024lambdaeclipse}. 
More specifically, in pOps~\cite{richardson2024pops}, a Diffusion Prior model leveraging the CLIP space is used to learn semantic operators (e.g., union, texturing) over given input concepts. Similarly, IP-Composer~\cite{dorfman2025ip} enables compositional image generation by extracting and integrating visual concepts from multiple reference images. However, while CLIP provides a semantically meaningful space, it often struggles to accurately reconstruct desired concepts, limiting its usability. We show that the proposed usage of the $\mathcal{IP}^+$ space overcomes this limitation while still allowing for semantically meaningful manipulations of image embeddings.

\section{Preliminaries}

\paragraph{Diffusion Prior.}
Diffusion models are typically trained with a conditioning vector, \( c \), directly derived from a given text prompt, \( y \). Ramesh~\etal~\cite{ramesh2022hierarchical} introduce a two-stage approach that decomposes the text-to-image generation process into two steps. 

First, a diffusion model is trained to generate an image conditioned on an image embedding, \( c \), using the standard diffusion loss:
\begin{equation}~\label{eq:ldm}
    \mathcal{L}_{diffusion} = \mathbb{E}_{z,y,\varepsilon,t} \left [ || \varepsilon - \varepsilon_\theta(z_t, t, c) ||_2^2 \right ].
\end{equation}
Here, the denoising network, \( \varepsilon_\theta \), learns to remove noise \( \varepsilon \) added to the latent code \( z_t \) at time step \( t \), given the conditioning image embedding \( c \).

Next, the Diffusion Prior model, \( P_\theta \), is trained to recover a clean image embedding, \( e \), from its noisy version, \( e_t \), conditioned on the corresponding text prompt, \( y \). The objective function is given by:
\begin{equation}\label{eq:prior}
    \mathcal{L}_{prior} = \mathbb{E}_{e,y,t} \left [ || e - P_\theta (e_t, t, y) ||_2^2 \right ] .
\end{equation}

Once both models are trained separately, they are combined into a full text-to-image generation pipeline. In this work, we extend the Diffusion Prior beyond its conventional role of predicting image embeddings from text. Instead, we adapt it to operate over multiple image parts, enabling finer-grained control over the synthesis process. 

\section{Method}
We begin by laying the foundation of Piece-it-Together (PiT), detailing the representation space in which our IP-Prior operates. Next, we describe the design and training process of our generative model within this chosen representation space. Finally, we demonstrate how the generated concepts can be integrated with pretrained image generation models alongside existing conditions.

\begin{figure*}
    \centering
    \includegraphics[width=0.875\textwidth]{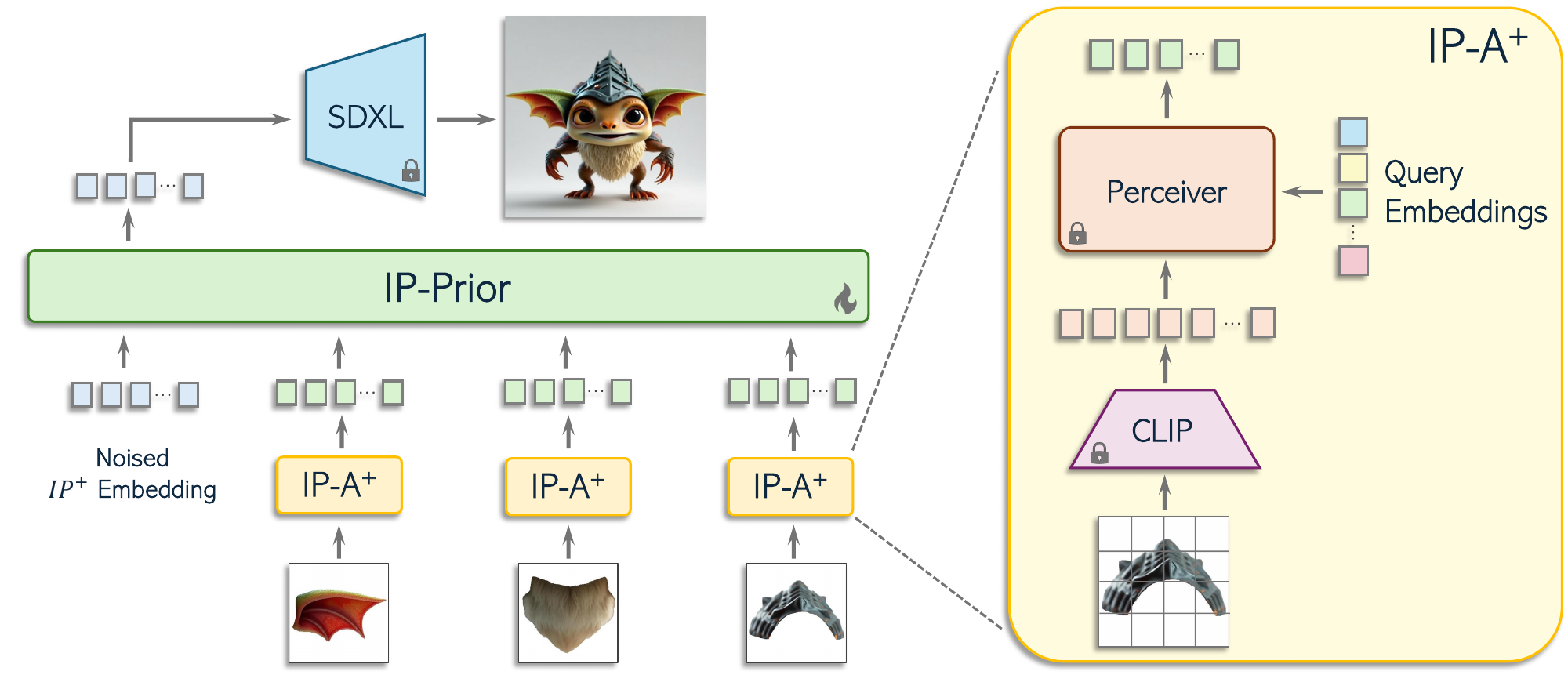}
    \\[-0.2cm]
    \caption{
    \textbf{Piece-it-Together Overview.} Given an input image, we extract its semantic components (e.g., using SAM~\cite{kirillov2023segment}) and encode each image patch into the $\mathcal{IP}+$ space using frozen IP-Adapter+ (IP-A+) blocks (shown in \textcolor{Dandelion}{yellow}). The resulting set of compact image embeddings are then passed together through our IP-Prior model (\textcolor{ForestGreen}{green}), which also receives a noised image embedding representing our desired complete concept. The IP-Prior model outputs a cleaned image embedding that captures the intended concept, which is subsequently used to generate the final concept image using SDXL~\cite{podell2023sdxl} (\textcolor{blue}{blue}). At inference time, users can provide a varying number of object-part images to generate a new concept that aligns with the learned distribution.
    }
    \vspace{-0.425cm}
    \label{fig:method}
\end{figure*}

\subsection{In Search of a Space}
The CLIP space is widely used as an image representation for image-conditioned generation and manipulation tasks~\cite{ye2023ip-adapter,kandinsky2,richardson2024pops,dorfman2025ip}. Beyond its alignment with the corresponding text encoding, CLIP has several compelling advantages, including an efficient image encoder and a semantically rich representation.
However, as originally highlighted in~\cite{ramesh2022hierarchical}, the CLIP space struggles to encode distinct visual patterns and is prone to attribute leakage. This limitation stems from CLIP’s training objective, which encourages the model to learn a joint text-image representation but does not explicitly require it to capture fine-grained visual details that cannot be easily described through text. While this design choice is reasonable for CLIP’s intended purpose, it poses clear challenges for image-conditioned tasks.

Recognizing that the performance of our method is constrained by the reconstruction capabilities of its visual representation, we must first search for an alternative representation that remains compact and inherently semantic while also offering improved reconstructions. To this end, we find that the internal representation of IP-Adapter+ meets these criteria and offers a compelling set of properties.
This model was released as an extension of the original IP-Adapter~\cite{ye2023ip-adapter} and gained popularity in the community for image-conditioned generation. 
While the original IP-Adapter utilizes the already compact CLIP embedding as input, as illustrated in~\Cref{fig:method}, IP-Adapter+ employs a Perceiver-like architecture~\cite{jaegle2021perceiver} operating over the full internal representation of the CLIP model. This approach produces a  representation of $16\times2048$ vectors, explicitly optimized with reconstruction in mind. The resulting representation is then fed into a pretrained generative model via a set of trainable attention layers. 

While IP-Adapter+ is usually used end-to-end as a method for image conditioning, we propose to conceptually decouple this process into two parts: first, encoding images into a compact representation, and second, conditioning the generative model on this representation. 
We then demonstrate that this intermediate encoding, which we refer to as the $\mathcal{IP}^+$ space, is a semantic representation space with significantly improved reconstructions compared to the standard CLIP image embedding and can serve as an effective representation for our generative model.
As we will show, operating in a compact latent space rather than directly on pixels makes training efficient and allows for easy manipulation of the generated concepts.

\subsection{How to Train Your IP-Prior}
A key requirement for our model is the ability to generate multiple plausible outputs for the same set of input parts. This enables designers to explore a range of variations, a crucial property for ideation.
To achieve this, we follow pOps~\cite{richardson2024pops} and design our model as a generative operator acting on image embeddings. During training, the model takes a sequence of visual concepts representing object parts, each encoded as $\mathcal{IP}^+$ vectors, and learns to produce a representation of the complete object, see~\Cref{fig:method}.

Deviating from~\cite{richardson2024pops}, which fine-tuned a pretrained prior model for the desired generative operator, our approach cannot rely on a pretrained prior, as no existing model has been trained on $\mathcal{IP}^+$ vectors. `This required us to build and train our network from scratch. Through empirical exploration, we chose to train a 4-block Diffusion Transformer (DiT)~\cite{peebles2023dit} using rectified flow~\cite{lipman2022flow} instead of the standard denoising loss used in~\cite{richardson2024pops,kandinsky2}. This approach provides an efficient training on a single GPU while keeping the model lightweight for both training and inference.

\begin{figure}
    \centering
    \setlength{\tabcolsep}{0pt}
    \addtolength{\belowcaptionskip}{-5pt}
    \renewcommand{\arraystretch}{0}
    {
    \begin{tabular}{c c c c c}

        \includegraphics[height=0.0825\textheight]{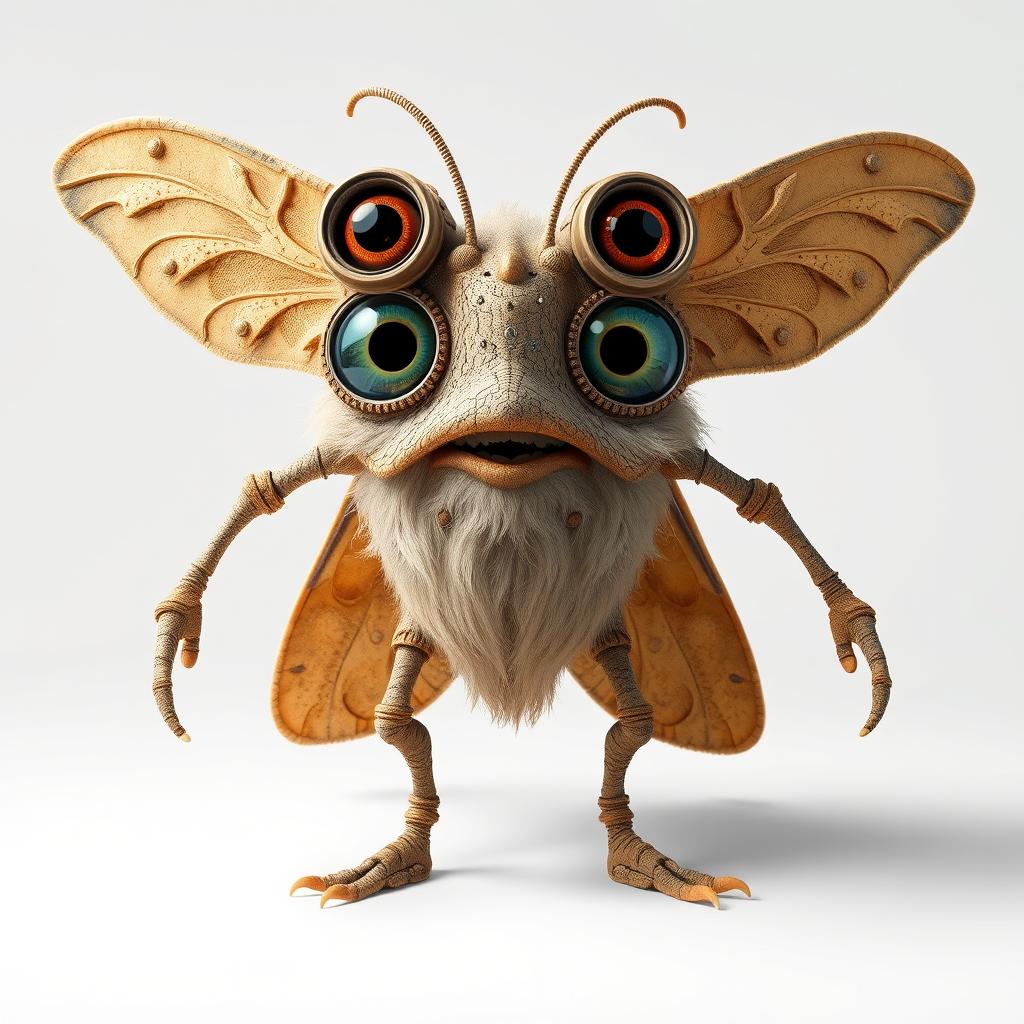} &
        \includegraphics[height=0.0825\textheight]{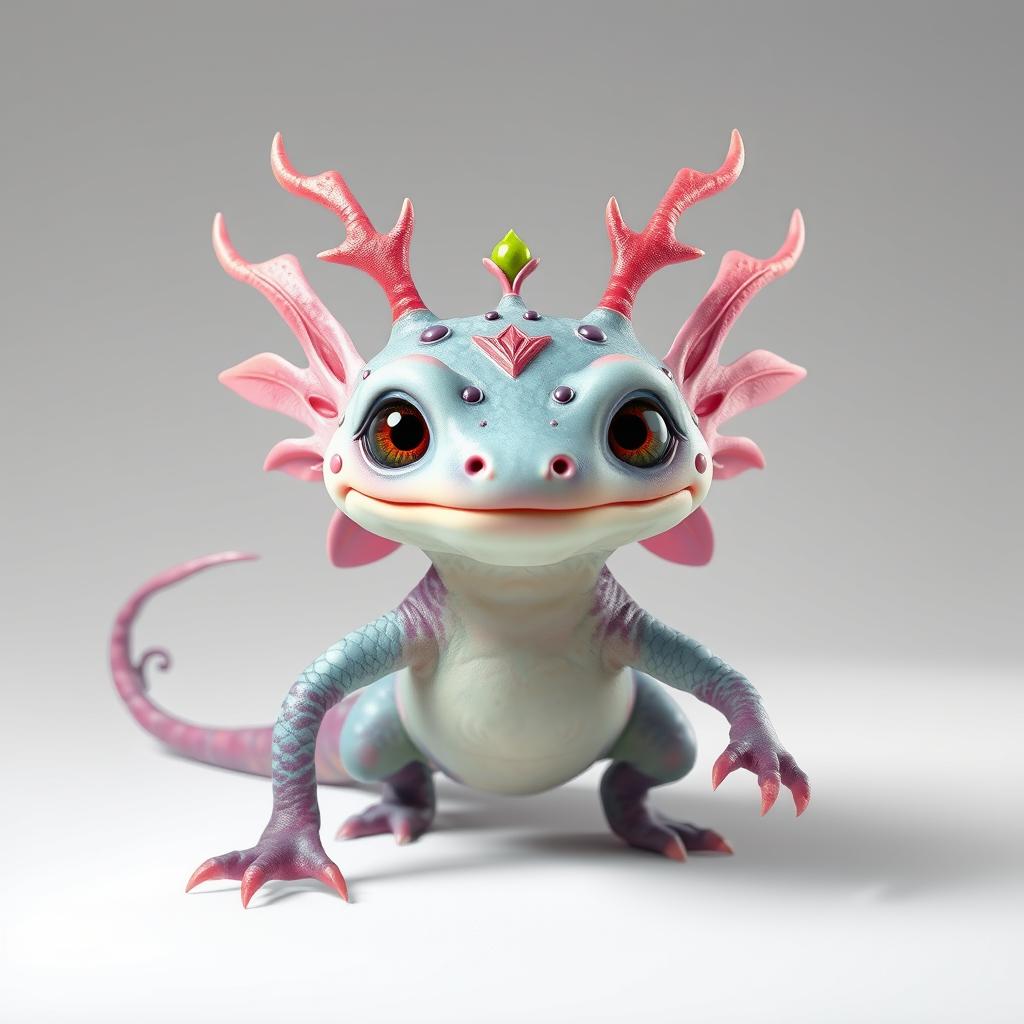} &
        \includegraphics[height=0.0825\textheight]{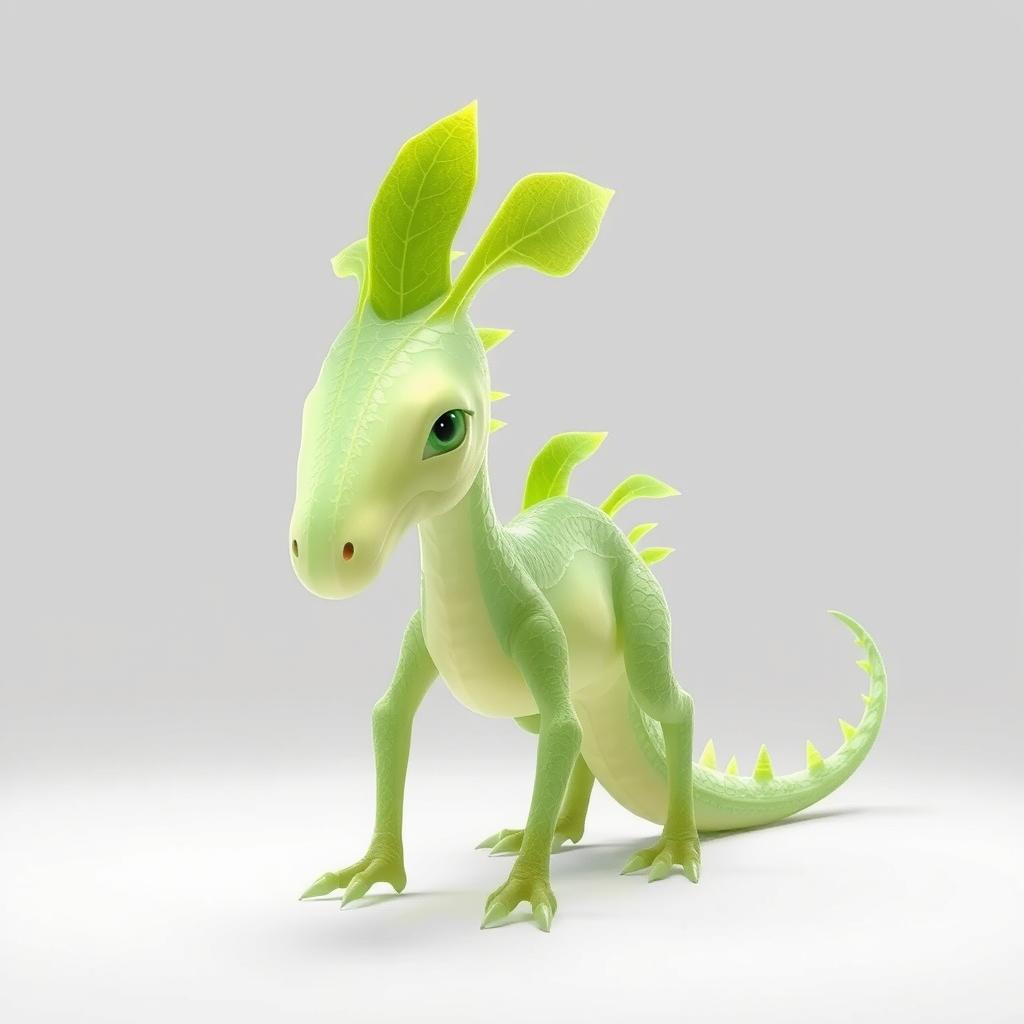} &
        \includegraphics[height=0.0825\textheight]{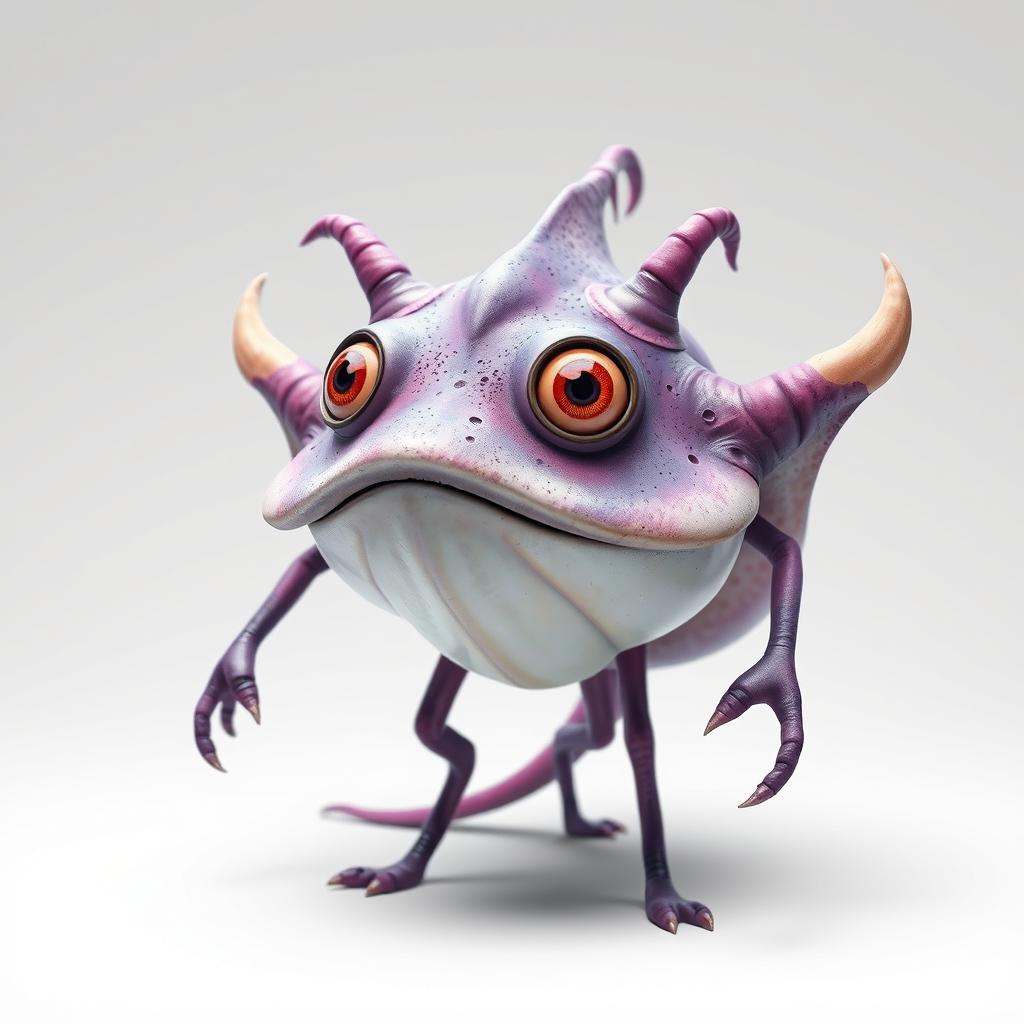} \\

        \includegraphics[height=0.0825\textheight]{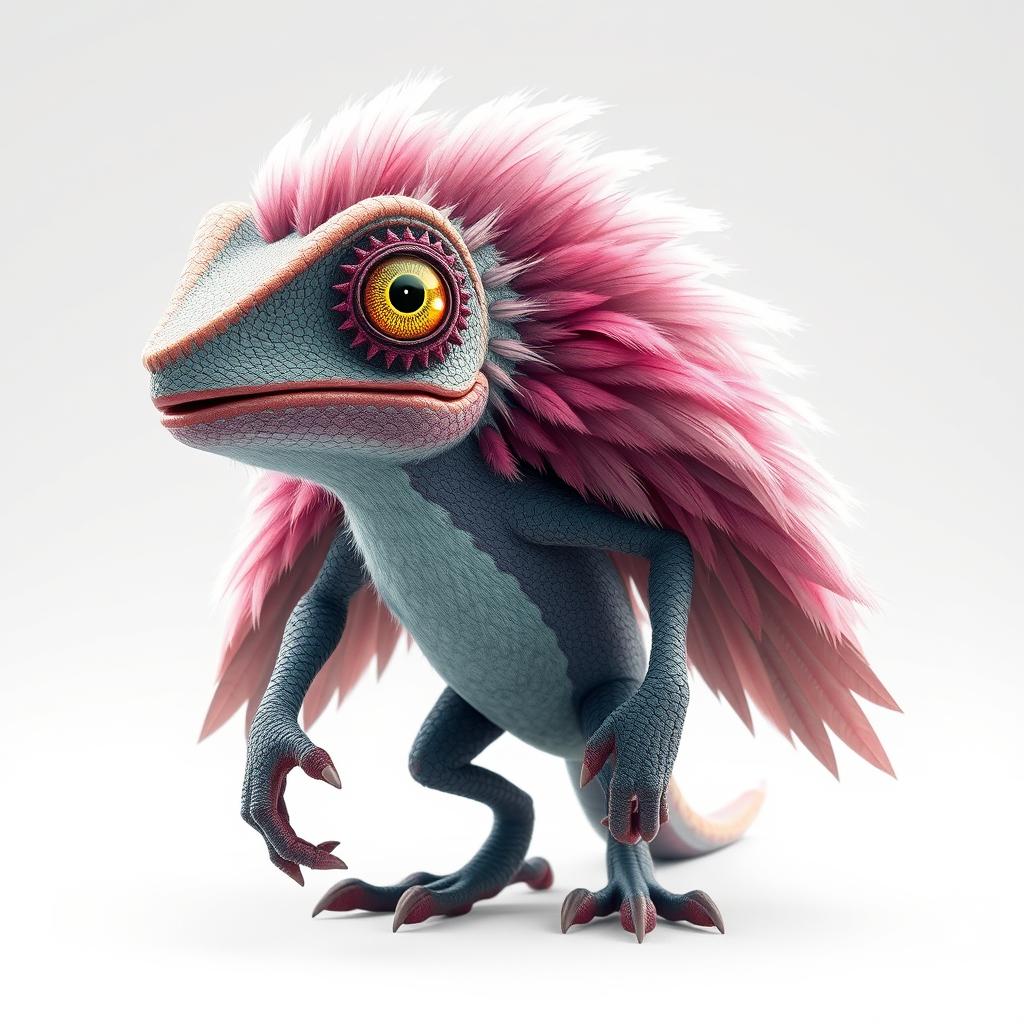} &
        \includegraphics[height=0.0825\textheight]{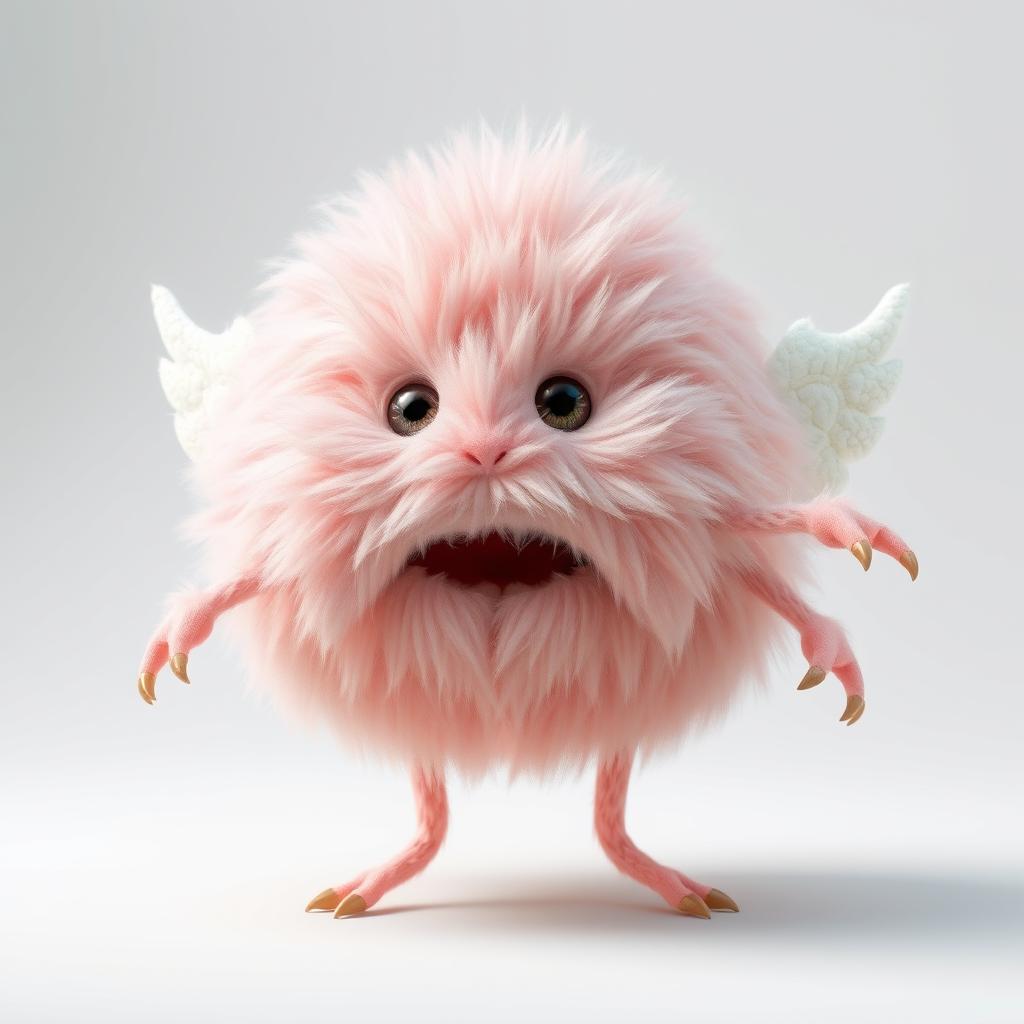} &
        \includegraphics[height=0.0825\textheight]{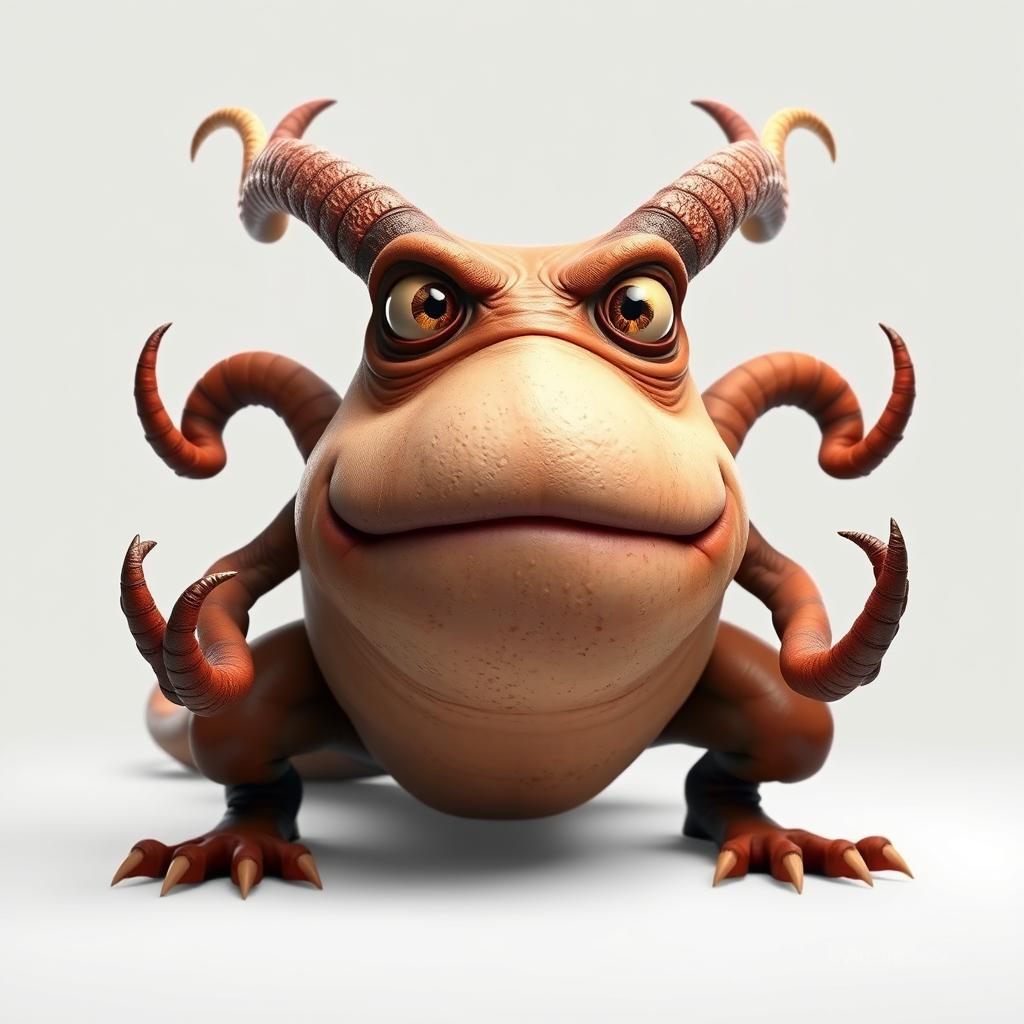} &
        \includegraphics[height=0.0825\textheight]{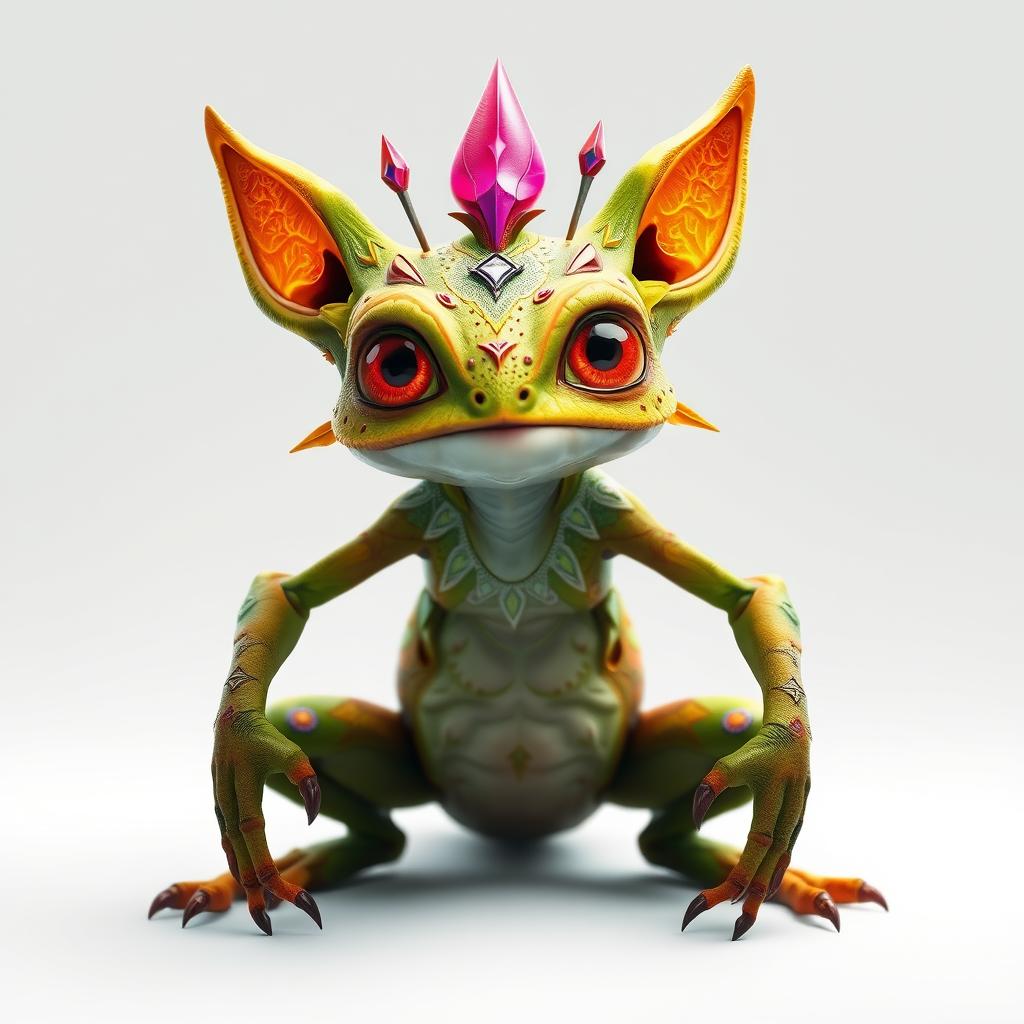}

    \end{tabular}
    }
    \vspace{-0.2cm}
    \caption{\textbf{Generated Data Samples.} We present sample images generated using FLUX-Schnell~\cite{blackforest2024flux}, which are used to train our IP-Prior model for the ``creatures'' domain.
    }
    \label{fig:grid}
\end{figure}

\subsection{Defining a Prior via Data}
Having demonstrated how to train a prior given a corresponding dataset, we now turn to the data-gathering process. While any visual dataset could be used to train an IP-Prior, we find that leveraging generated data offers a more flexible and efficient approach.
Our data generation process is based on the assumption that modern text-based generative models can effectively sample from specific domains. For instance, in a character ideation workflow, one could easily use the prompt ``Concept art of an imaginary creature, white background'' to generate diverse character concepts.
That is, the domain is often well-within the text-based generative model. However, text offers a very specific form of control over the sampled results, making the exploration process tedious. 

Building on this observation, we first generate a large set of images using a pretrained text-to-image model, specifically Flux-Schnell~\cite{blackforest2024flux}. To enhance the diversity of the generated dataset, we randomly append relevant adjectives to the base prompt (e.g., ``shadowy'', ``cryptic'', ``magnificent''). A sample training domain generated using this technique is shown in~\Cref{fig:grid}.
To create the input pairs, we extract semantic parts from the target images using a general-purpose segmentation method~\cite{kirillov2023segment} and randomly sample a subset of them. This process encourages the model to better solve the target task --- spatially assembling the provided visual concepts while generating the missing parts --- all within the context of the learned prior.
Interestingly, since we operate in an embedding space, we do not encounter the overfitting issues reported in relevant literature for other tasks when using a simple segmentation-based data generation process~\cite{tan2024ominicontrol}.

\subsection{Unleashing Text Adherence with IP-LoRA}
Once a complete visual concept has been generated using an IP-Prior, it can be rendered as an image by passing it to a pretrained image generation model~\cite{podell2023sdxl} via IP-Adapter+. 
When rendered as-is, the generated concept closely matches its representation in the training data, which in our case often features a simple, clean background. Ideally, however, at this point, one would want to explore how the concept fits into different scenarios and scenes. A natural approach to achieve this is to leverage the text-conditioning of the pretrained image model alongside our outputted IP vector.
In practice, however, when using IP-Adapter+, the model tends to exhibit low prompt adherence, see~\Cref{sec:ip_lora}. A common way to address this is by adjusting the scaling factor of the adapter outputs, but this comes at the cost of reconstruction quality, leading to a suboptimal trade-off between fidelity and controllability. 
We hypothesize this arises from the expressiveness of the $\mathcal{IP}^+$ space, which makes the text conditioning redundant during fine-tuning, causing the model to ignore it in inference. Despite this, we argue that since the model functions as an adapter, its text understanding capabilities are still ``hidden'' inside the model and simply need to be ``reactivated''. 

\begin{figure}
    \centering
    \includegraphics[width=0.425\textwidth]{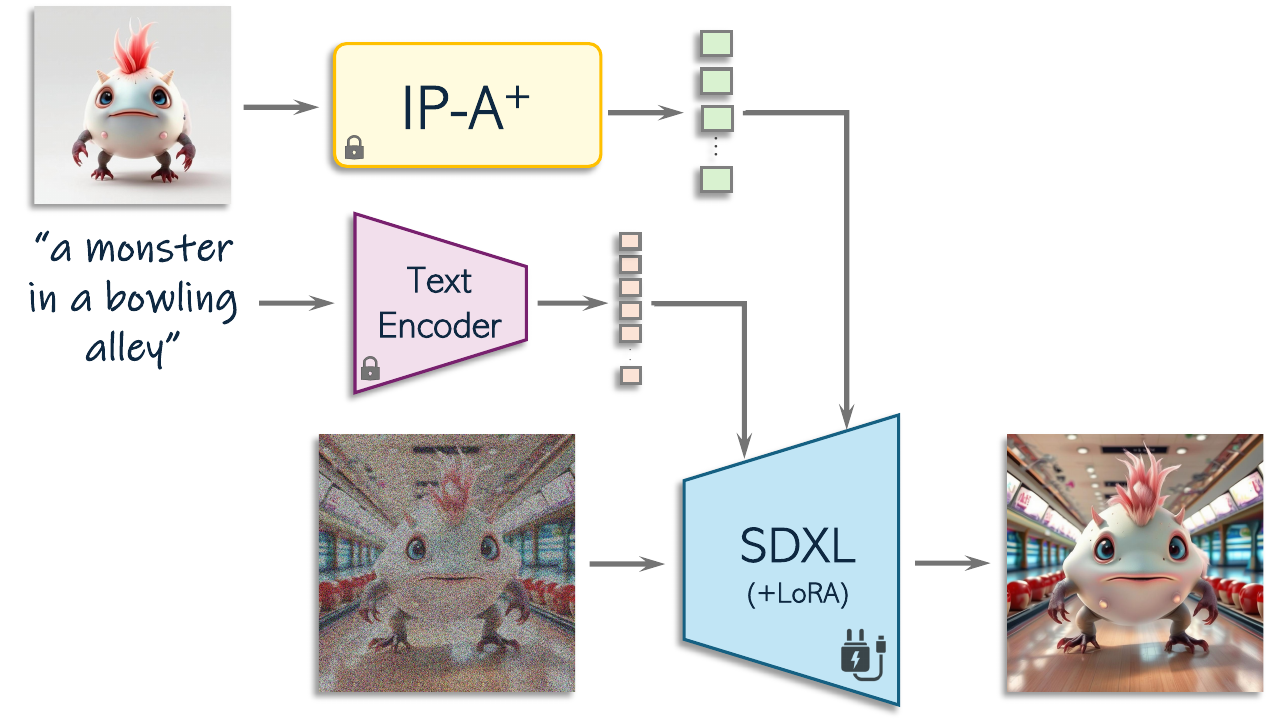}
    \\[-0.15cm]
    \caption{
    \textbf{Recovering the Text Adherence via IP-LoRA.} 
    IP-Adapter+ enables rendering generated concepts via SDXL~\cite{podell2023sdxl} but often struggles with text adherence. To address this, we fine-tune a LoRA adapter over paired examples, where the conditioning image has a clean background and the target image places the object in a scene described using a text prompt. This lightweight training (using just $50$ prompts) effectively restores text control while maintaining visual fidelity.
    }
    \vspace{-0.4cm}
    \label{fig:text_lora}
\end{figure}

To re-enable text conditioning, we show that training a LoRA adapter~\cite{hu2022lora} on a small set of examples can effectively restore this capability. In these examples, the conditioning and target images are not identical. Instead, the conditioning image depicts the object on a clean background, while the target image places the object in a new scene described by a text prompt. We refer to this training setup as \textit{IP-LoRA} and illustrate it in~\Cref{fig:text_lora}.
This setup forces the model to incorporate the text prompt while preserving the visual characteristics of the target concept. Remarkably, even when trained on just $50$ text prompts, our model generalizes well to unseen prompts describing different backgrounds, showing that the LoRA adapter enables the model to reuse its existing text comprehension abilities.
Furthermore, we show that this mechanism can also be leveraged to personalize the model for specific subdomains based on the given image embedding. For example, it can generate an image depicting a full character sheet from a single generated concept.

\begin{figure*}
    \centering
    \setlength{\tabcolsep}{0.5pt}
    \addtolength{\belowcaptionskip}{-7.5pt}
    {\small

    \begin{tabular}{c @{\hspace{0.2cm}} c c c c @{\hspace{0.5cm}} c c c c}
        \raisebox{0.045\linewidth}{\rotatebox[origin=t]{90}{\begin{tabular}{c@{}c@{}c@{}c@{}} Characters \end{tabular}}} &

        \includegraphics[height=0.126\textwidth]{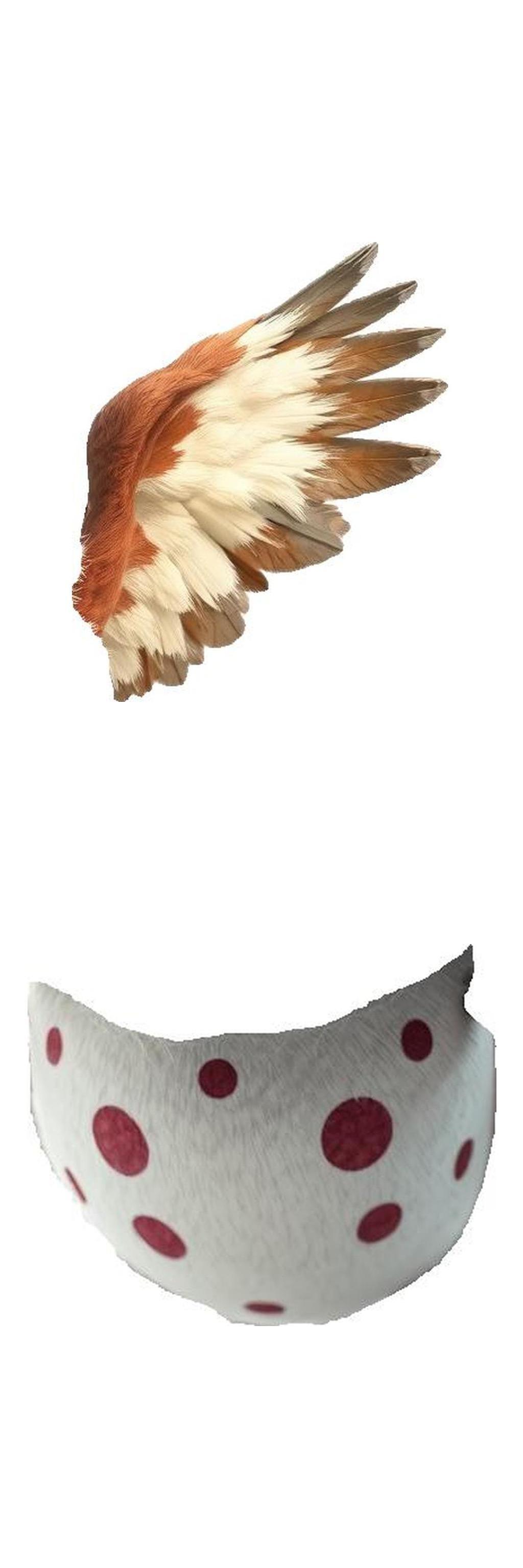} &
        \includegraphics[height=0.126\textwidth]{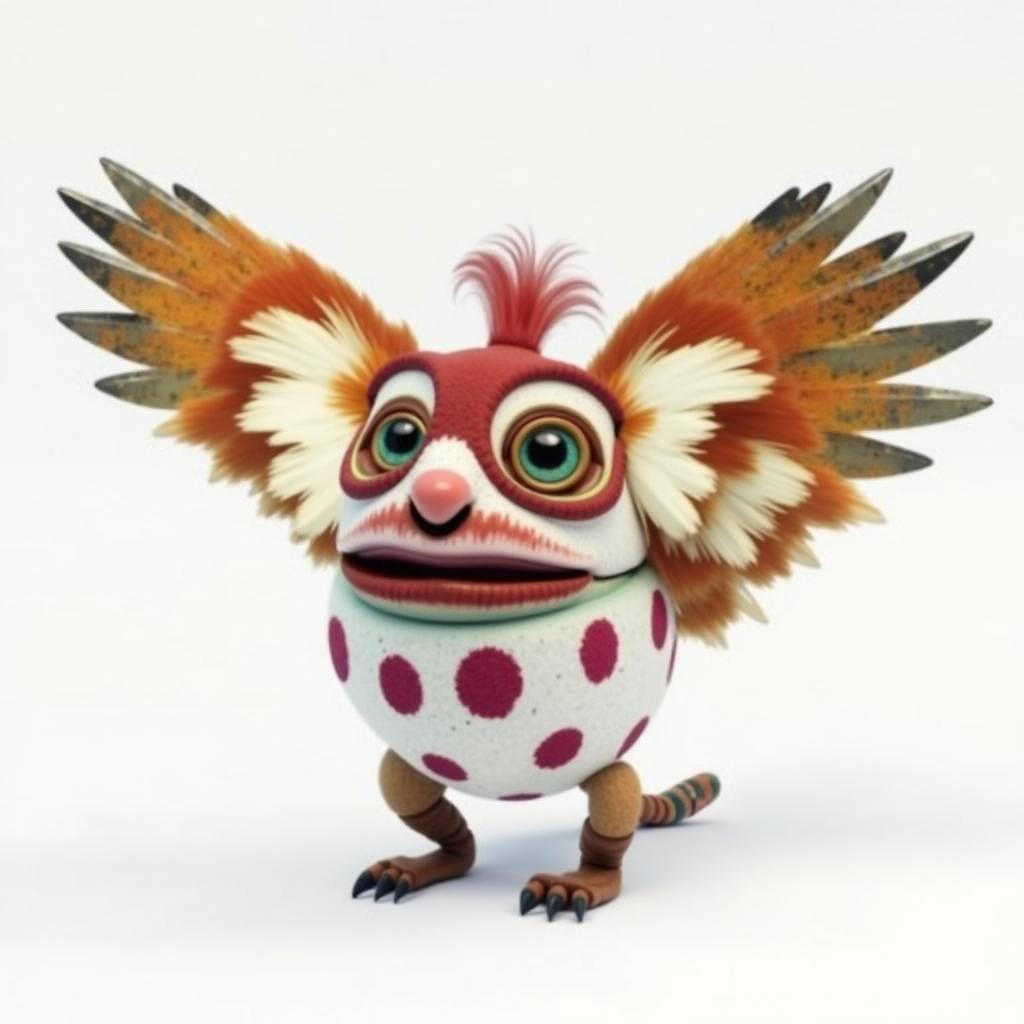} &
        \includegraphics[height=0.126\textwidth]{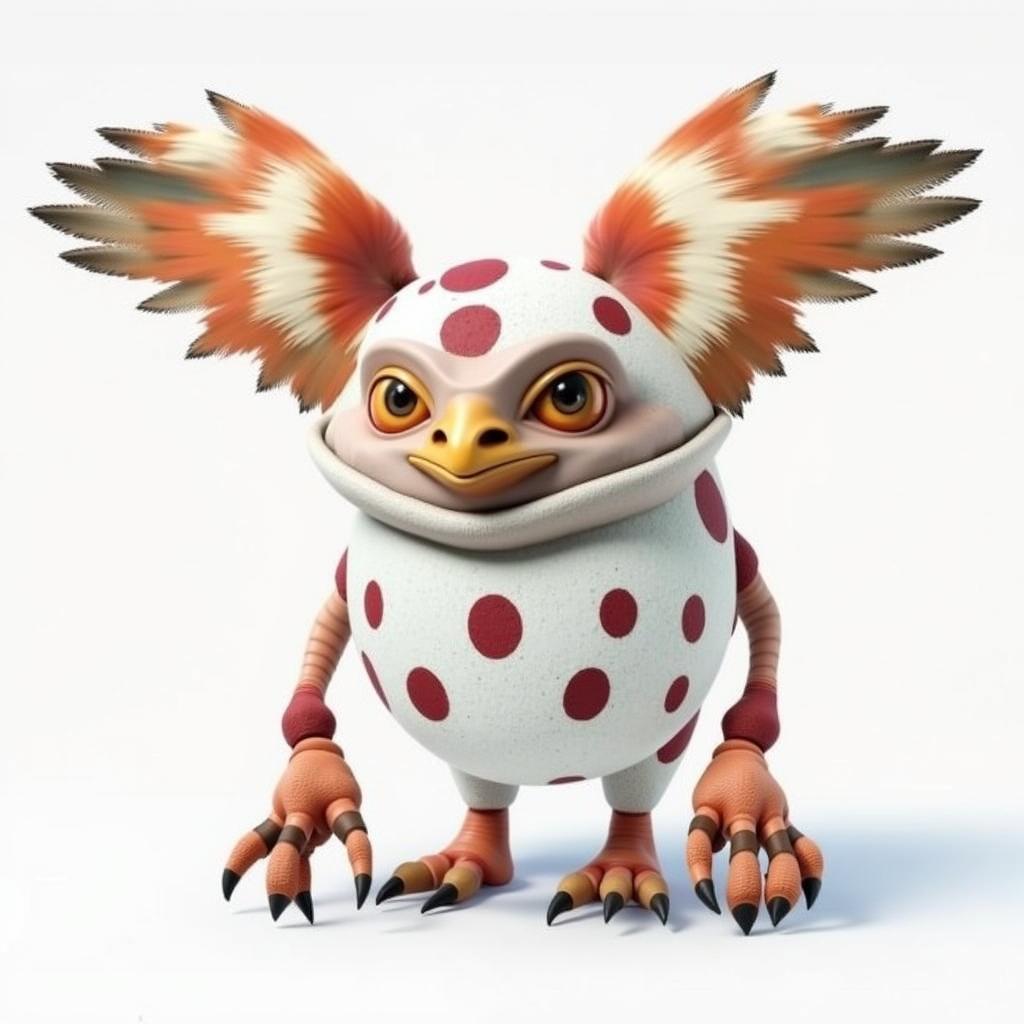} &

        \includegraphics[height=0.126\textwidth]{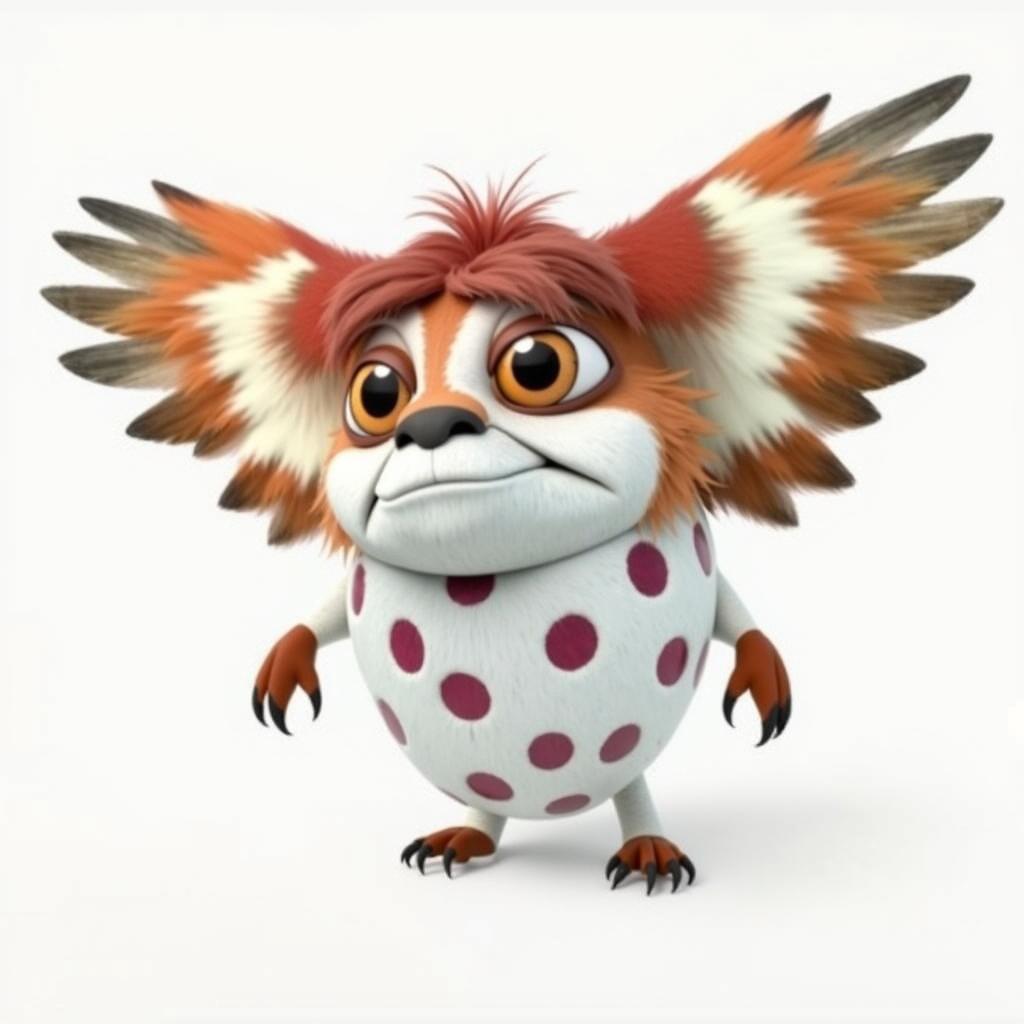} &
        
        \includegraphics[height=0.126\textwidth]{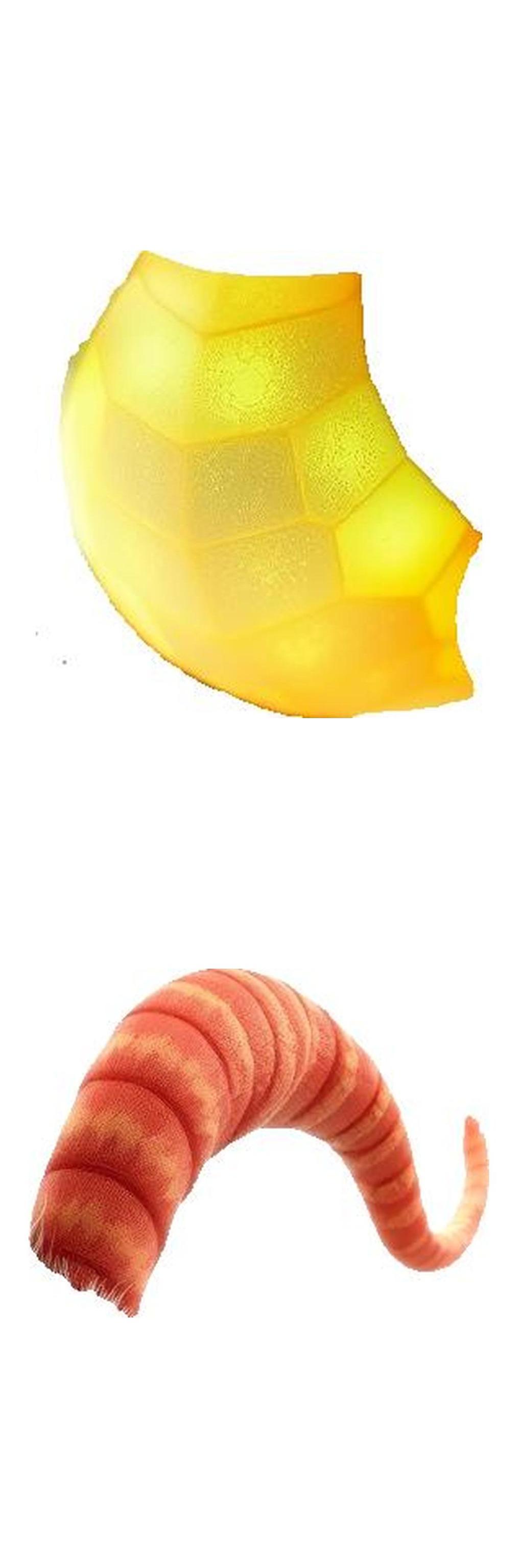} &

        \includegraphics[height=0.126\textwidth]{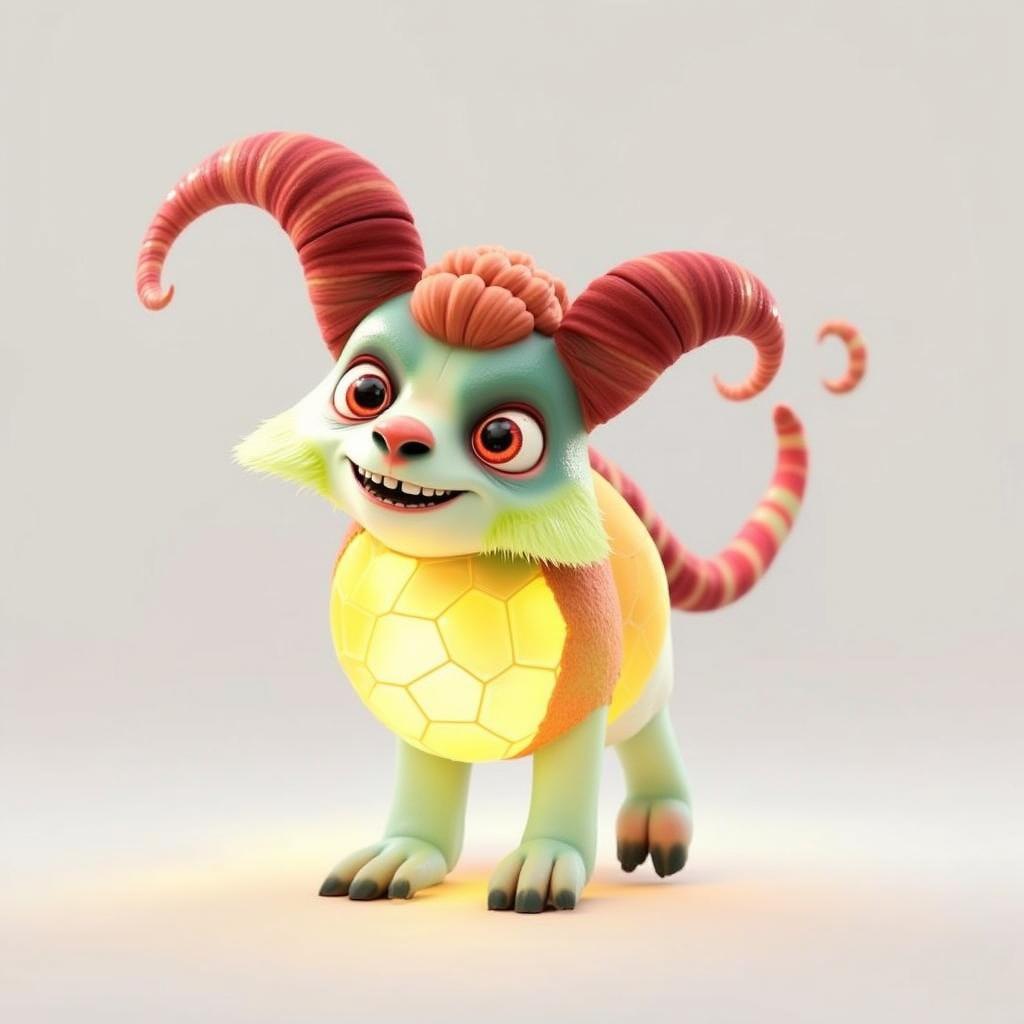} &

        \includegraphics[height=0.126\textwidth]{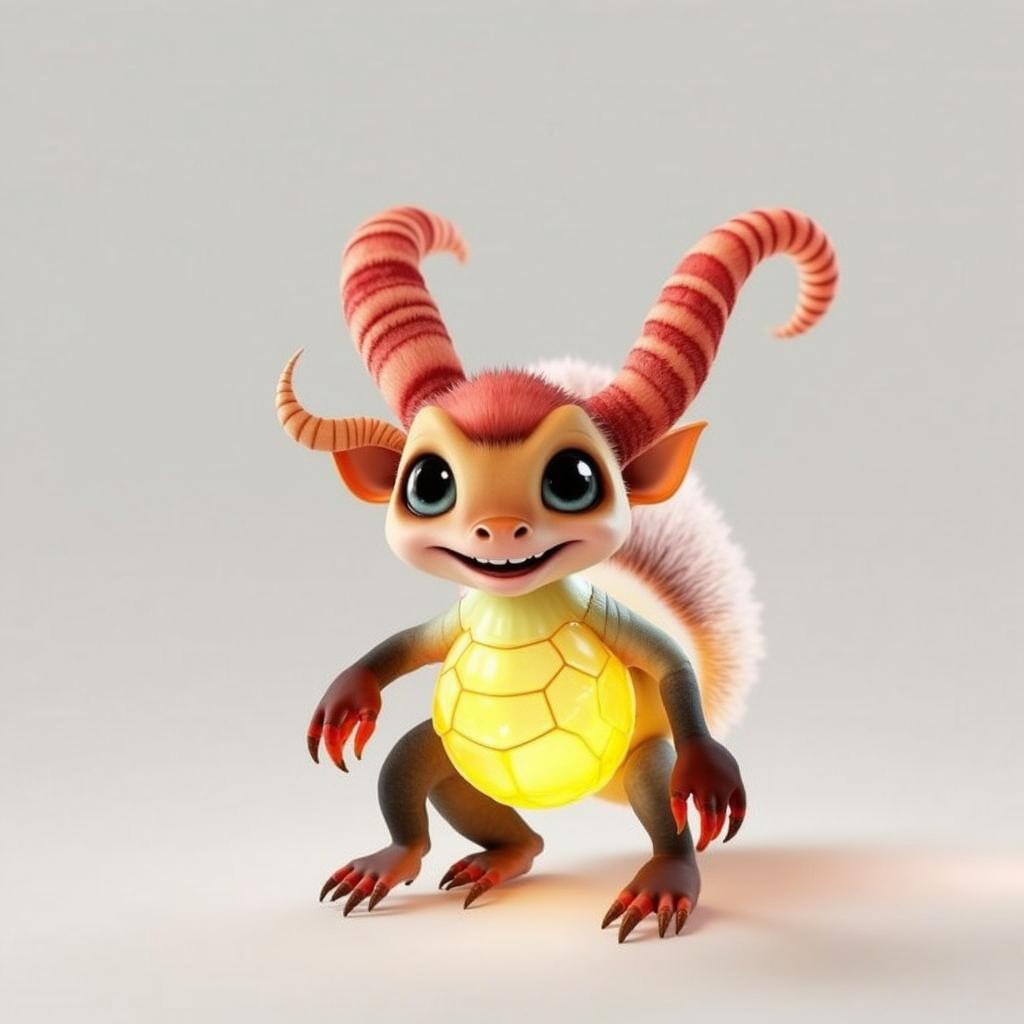} &

        \includegraphics[height=0.126\textwidth]{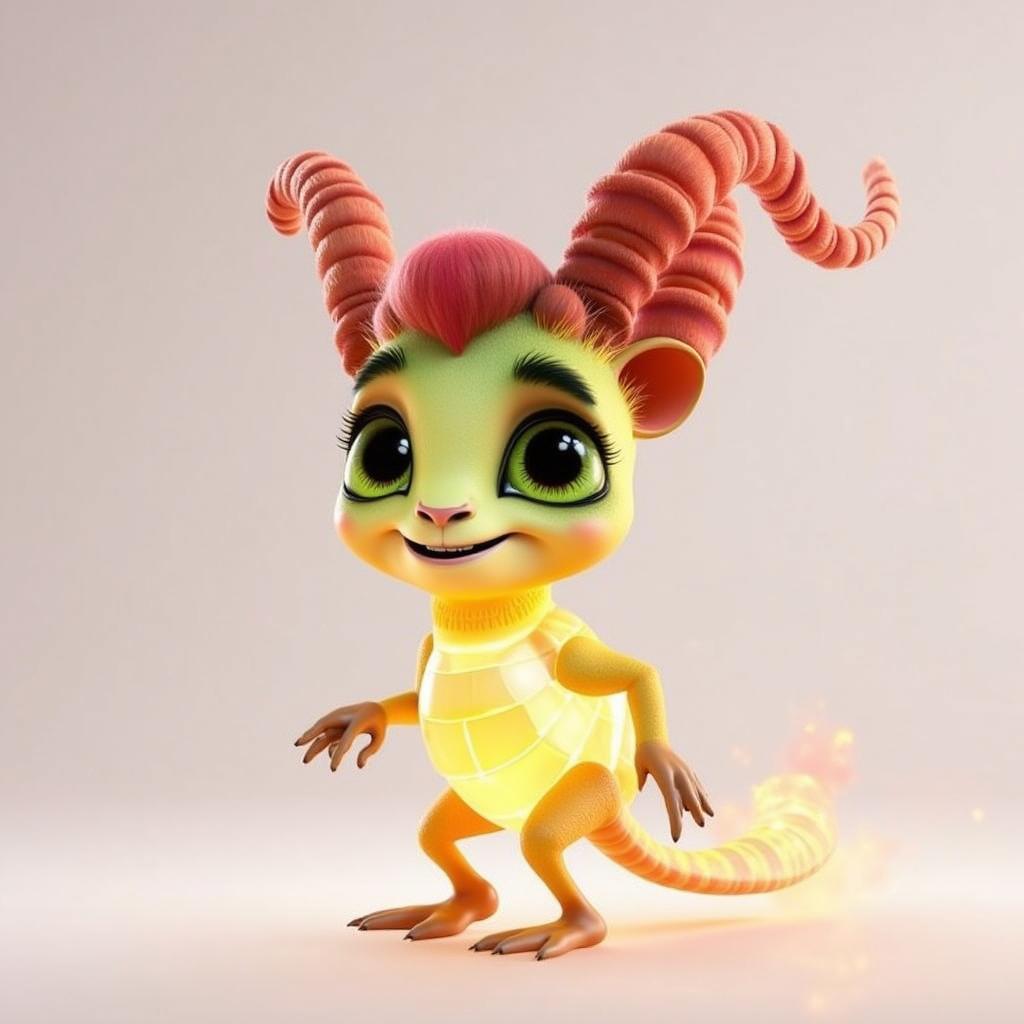} 
        
        \\

         \multicolumn{2}{c}{ 
        \includegraphics[height=0.126\textwidth]{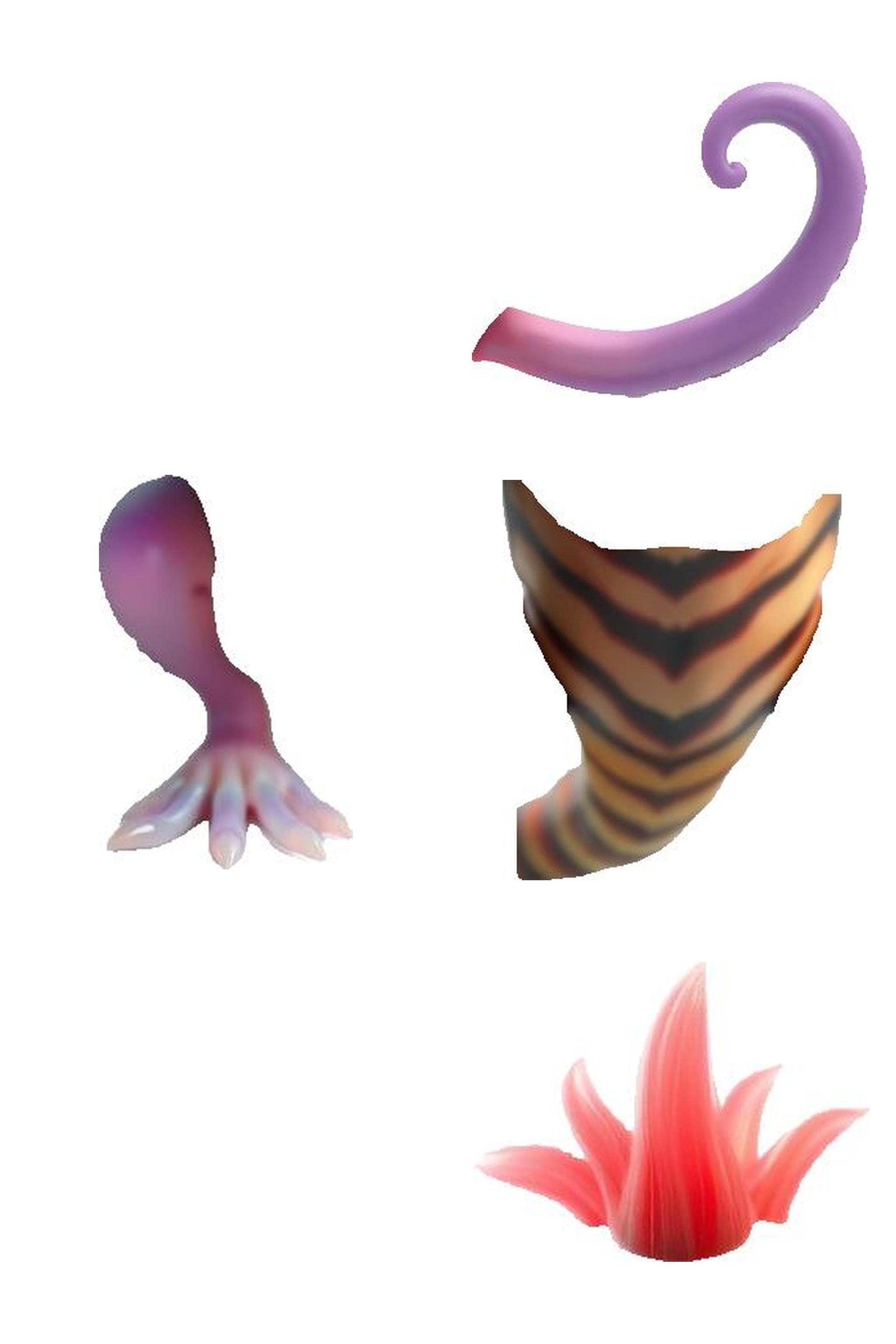}} &
        \includegraphics[height=0.126\textwidth]{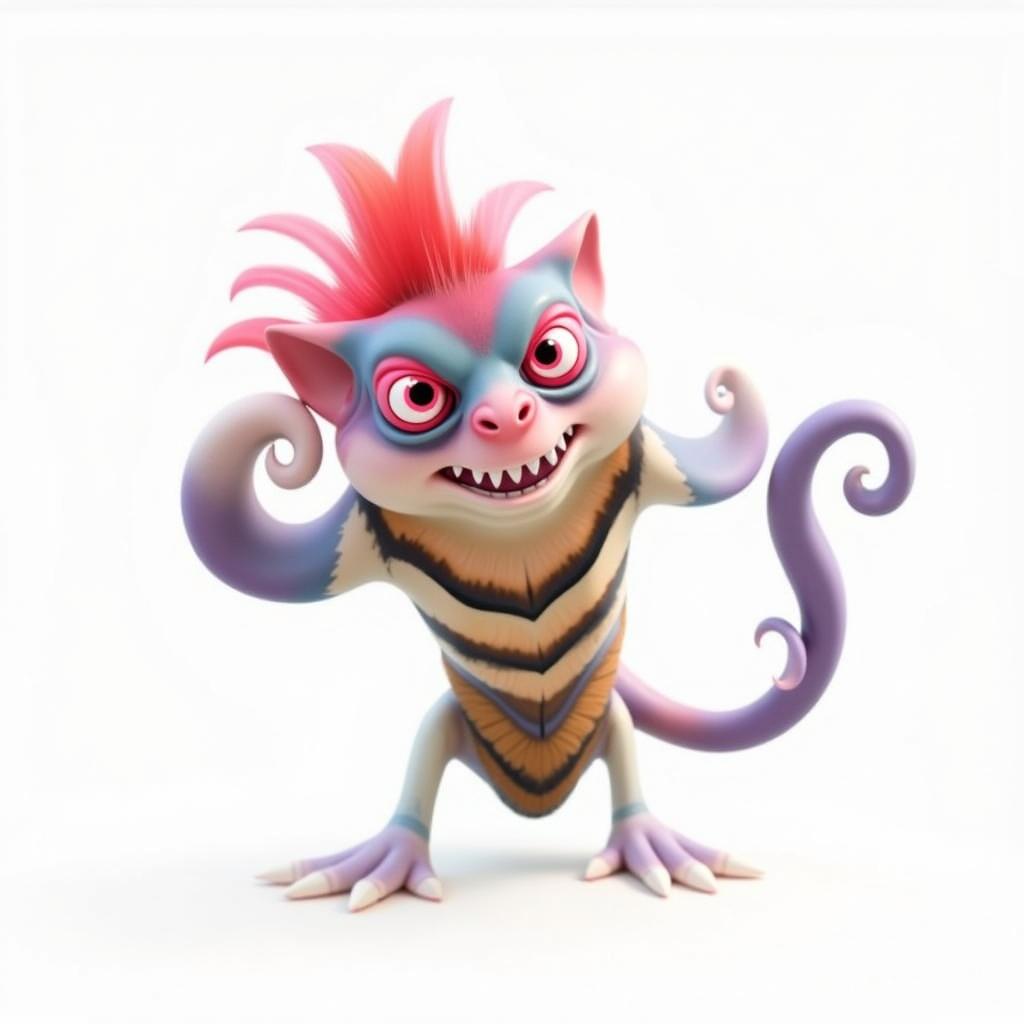} &
        \includegraphics[height=0.126\textwidth]{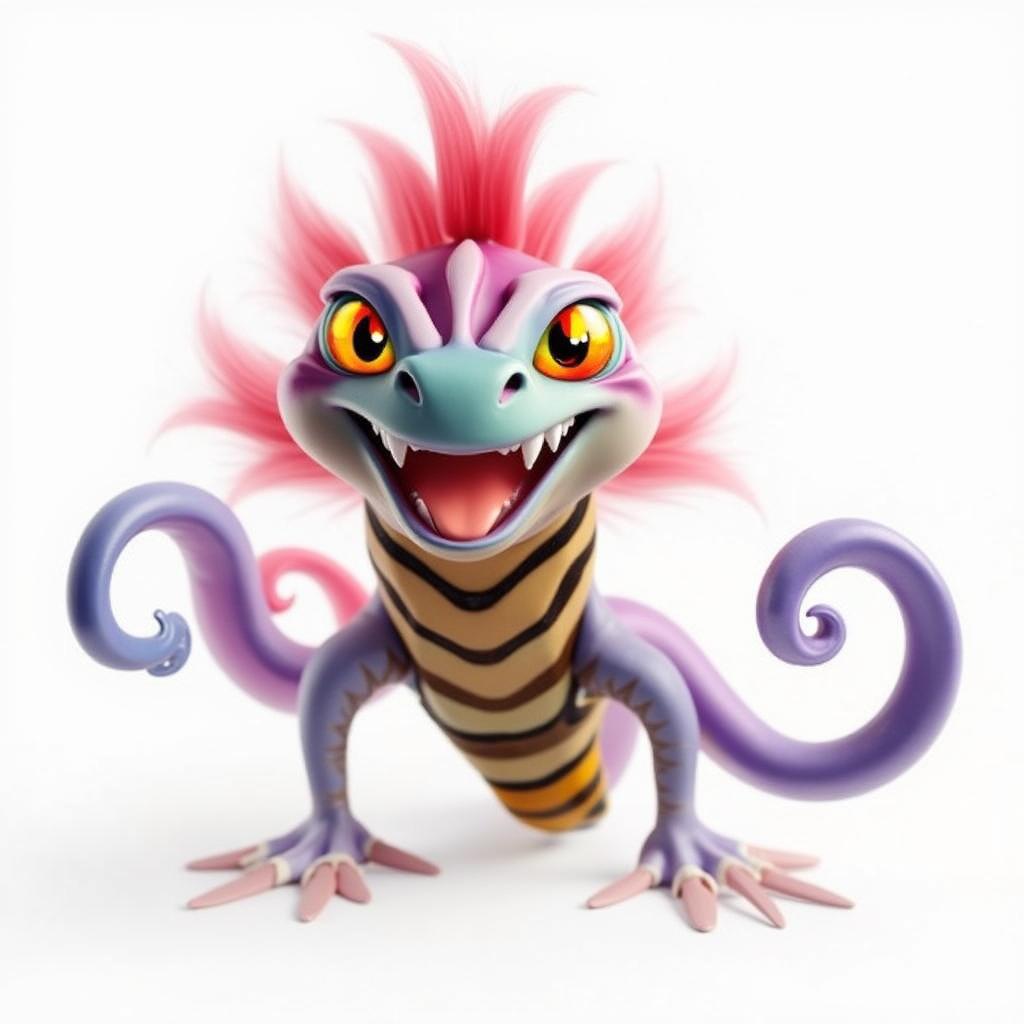} &

        \includegraphics[height=0.126\textwidth]{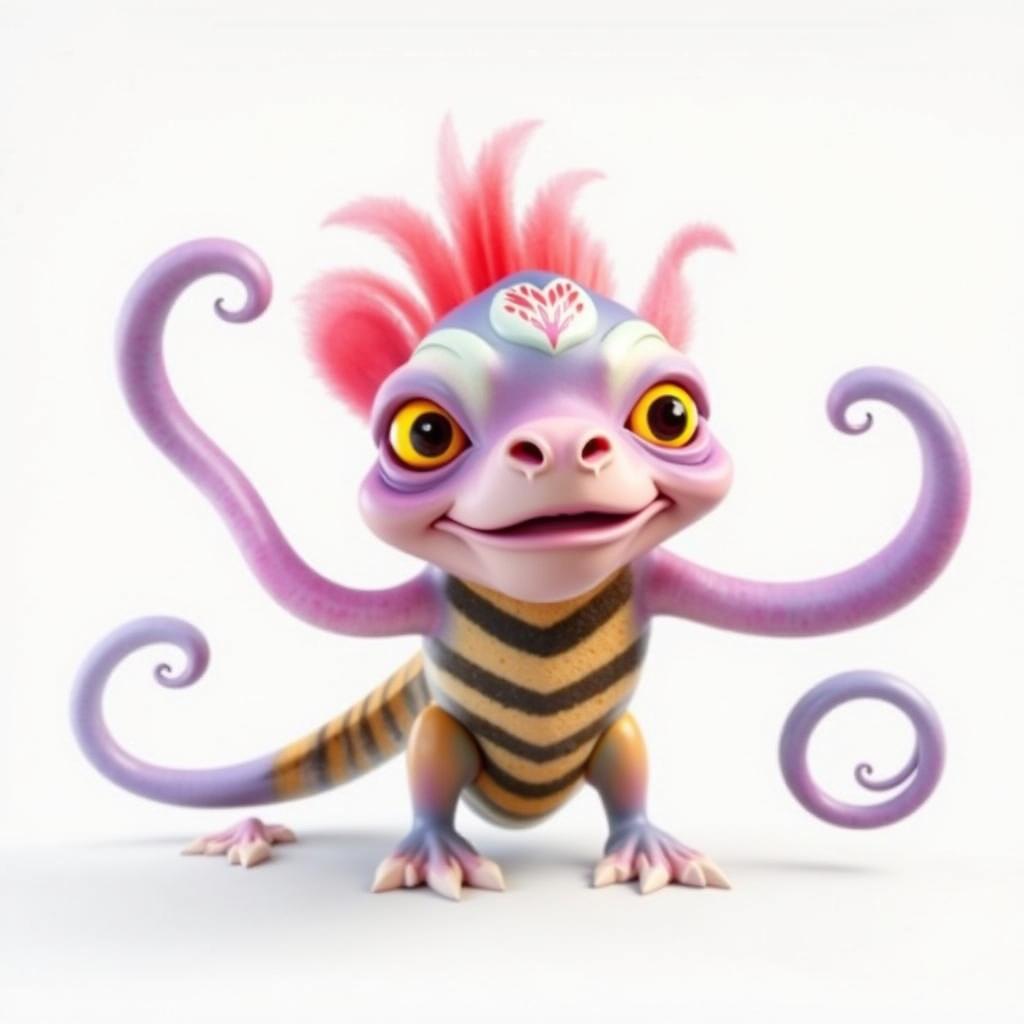} &

        \includegraphics[height=0.126\textwidth]{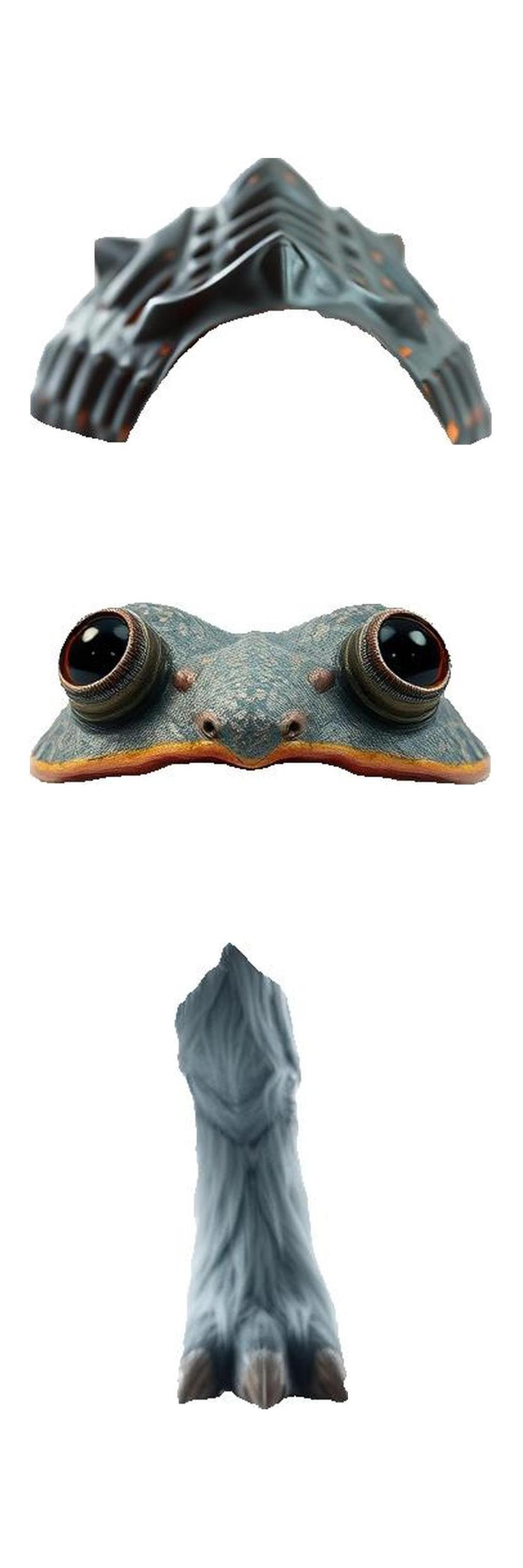} &

        \includegraphics[height=0.126\textwidth]{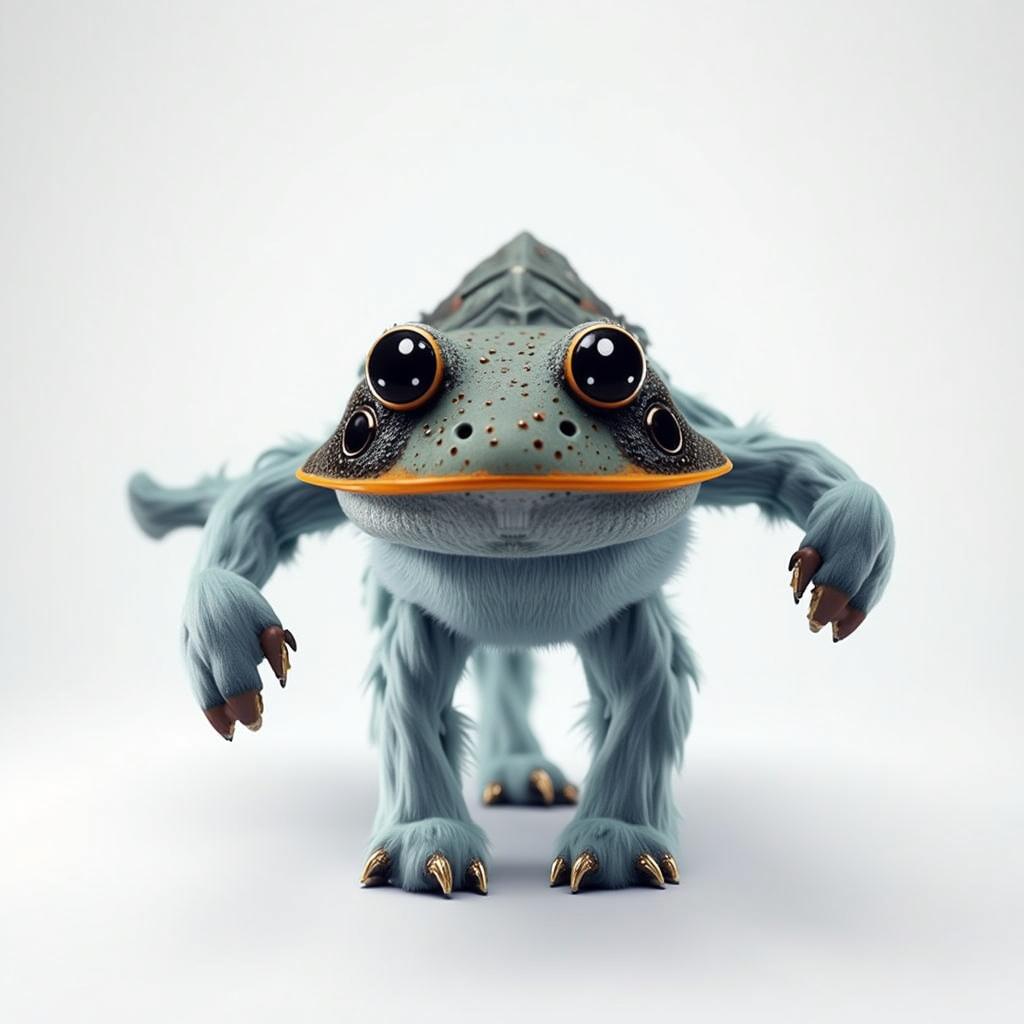} &

        \includegraphics[height=0.126\textwidth]{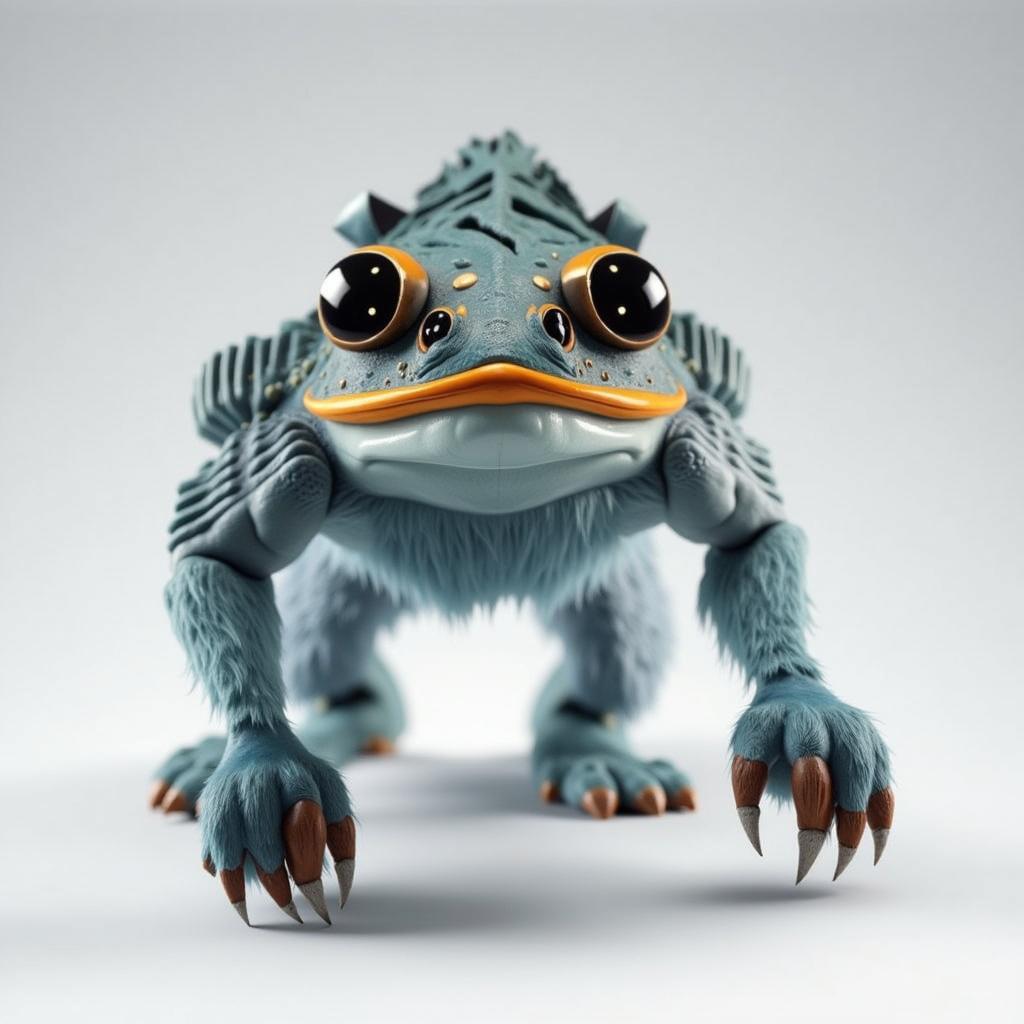} &

        \includegraphics[height=0.126\textwidth]{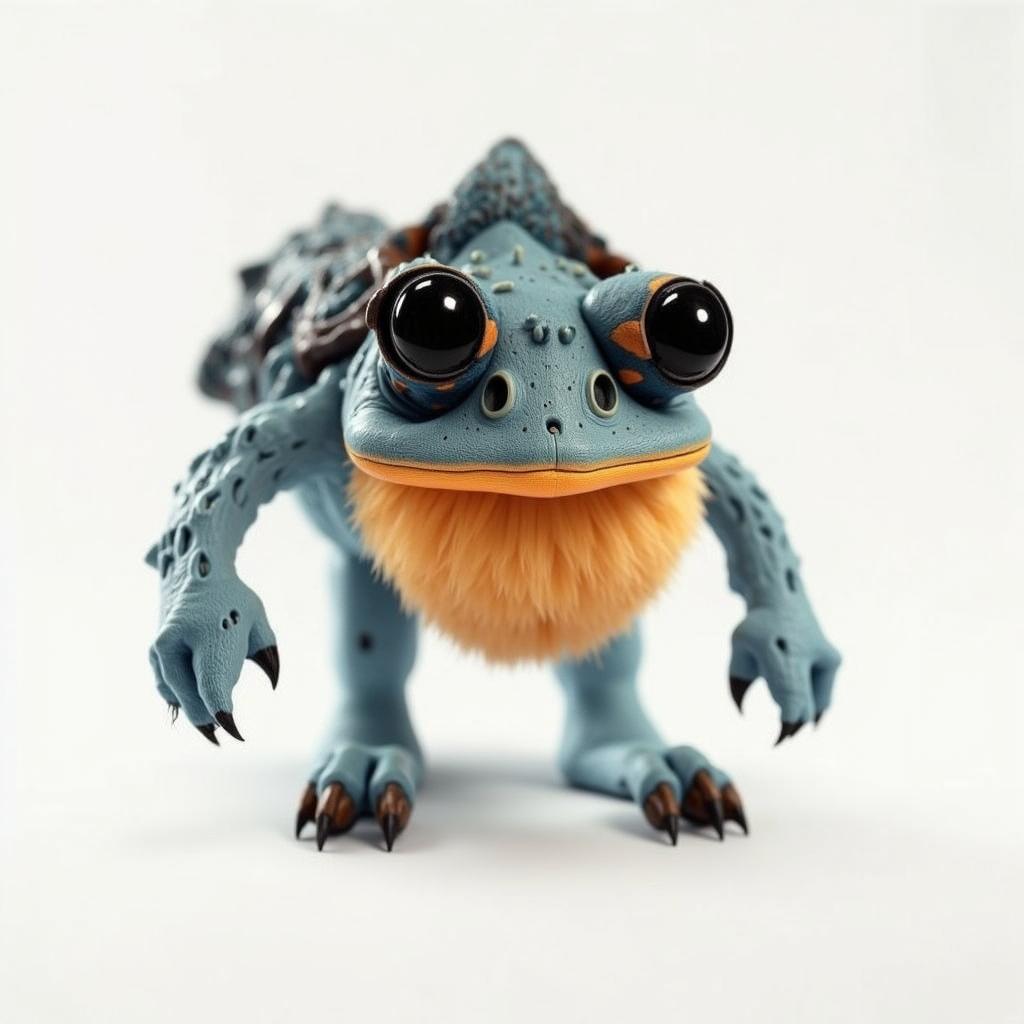}

        \\

        \raisebox{0.045\linewidth}{\rotatebox[origin=t]{90}{\begin{tabular}{c@{}c@{}c@{}c@{}} Products \end{tabular}}} &

        \includegraphics[height=0.126\textwidth]{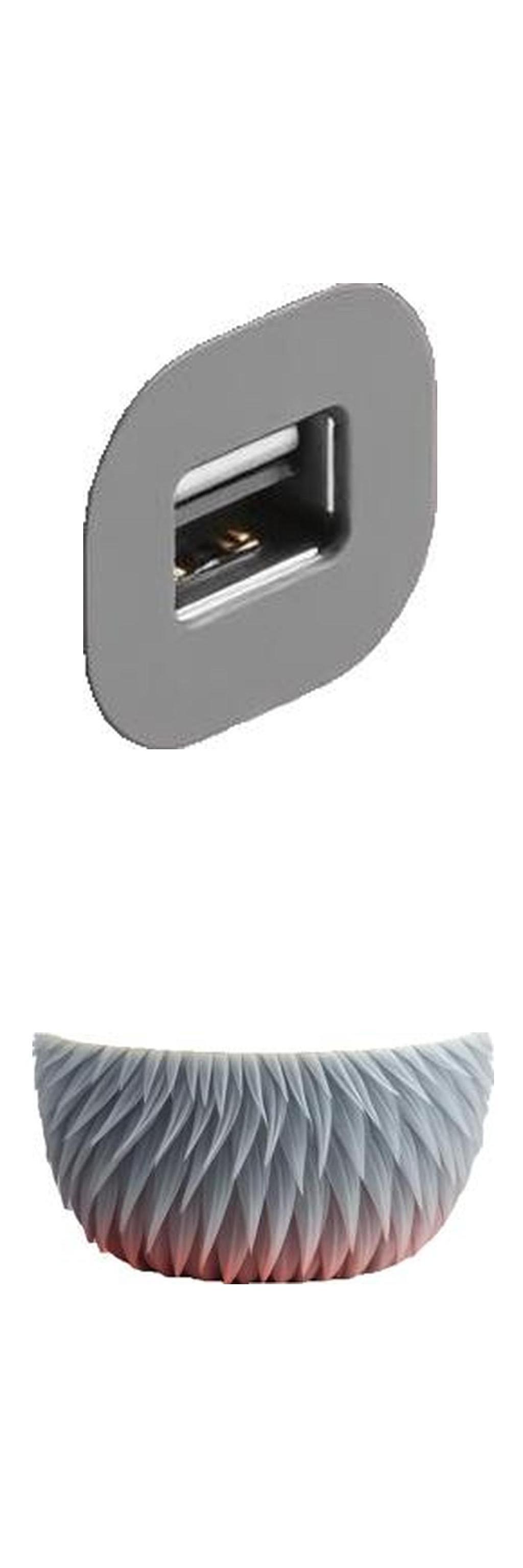} &
        \includegraphics[height=0.126\textwidth]{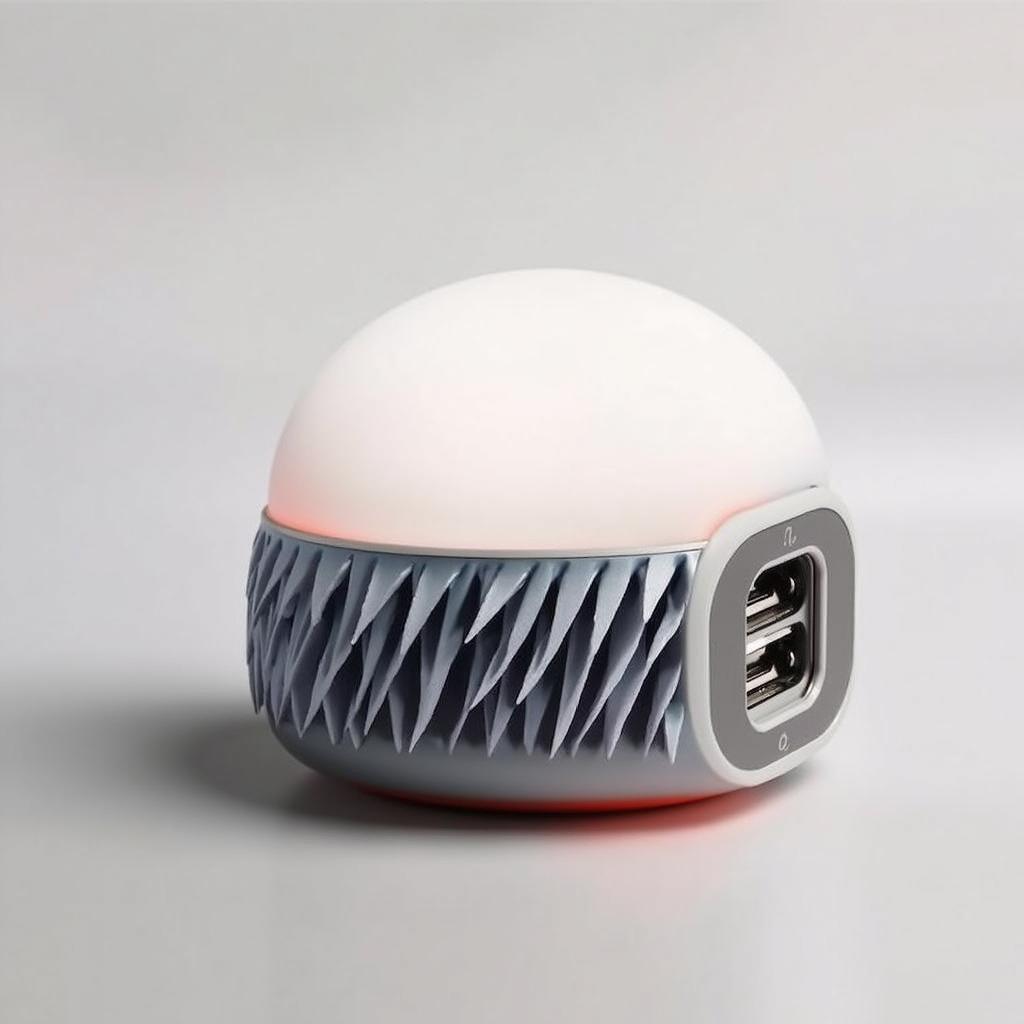} &
        \includegraphics[height=0.126\textwidth]{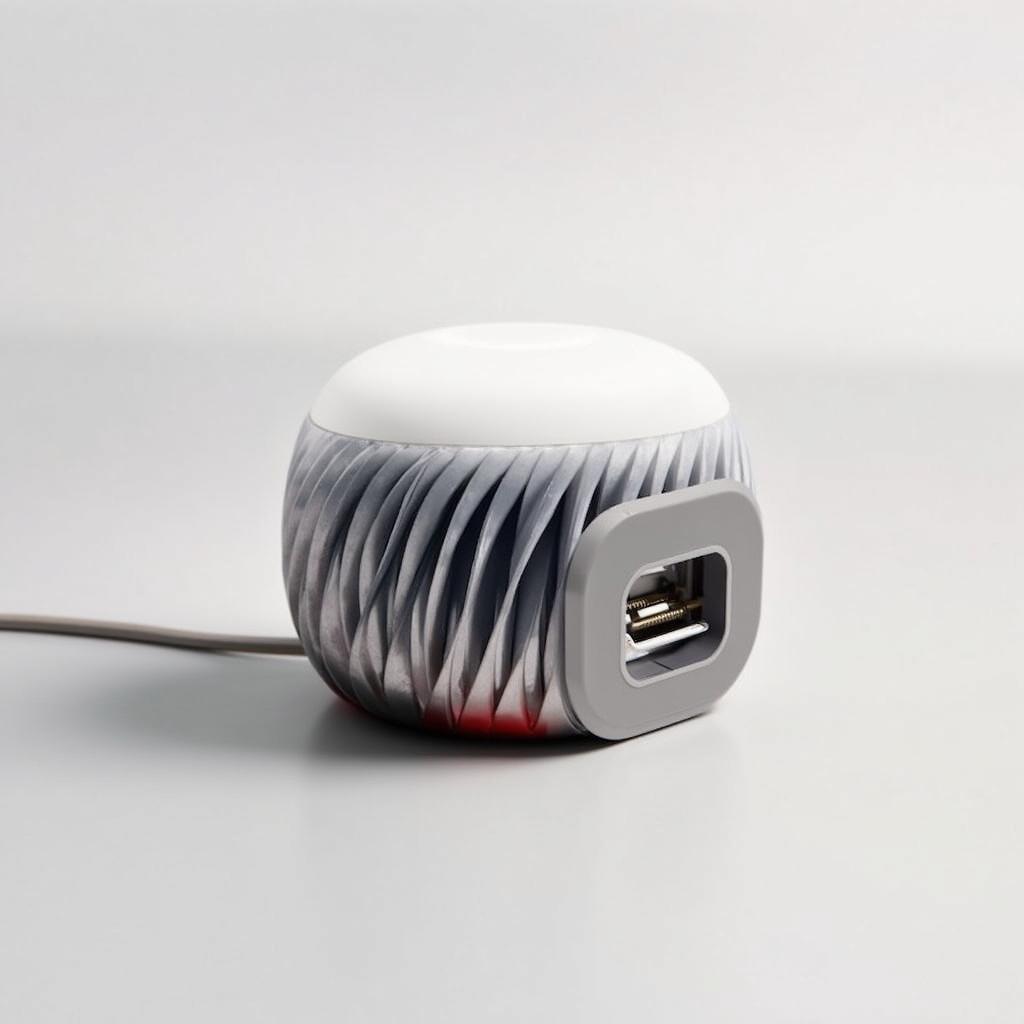} &

        \includegraphics[height=0.126\textwidth]{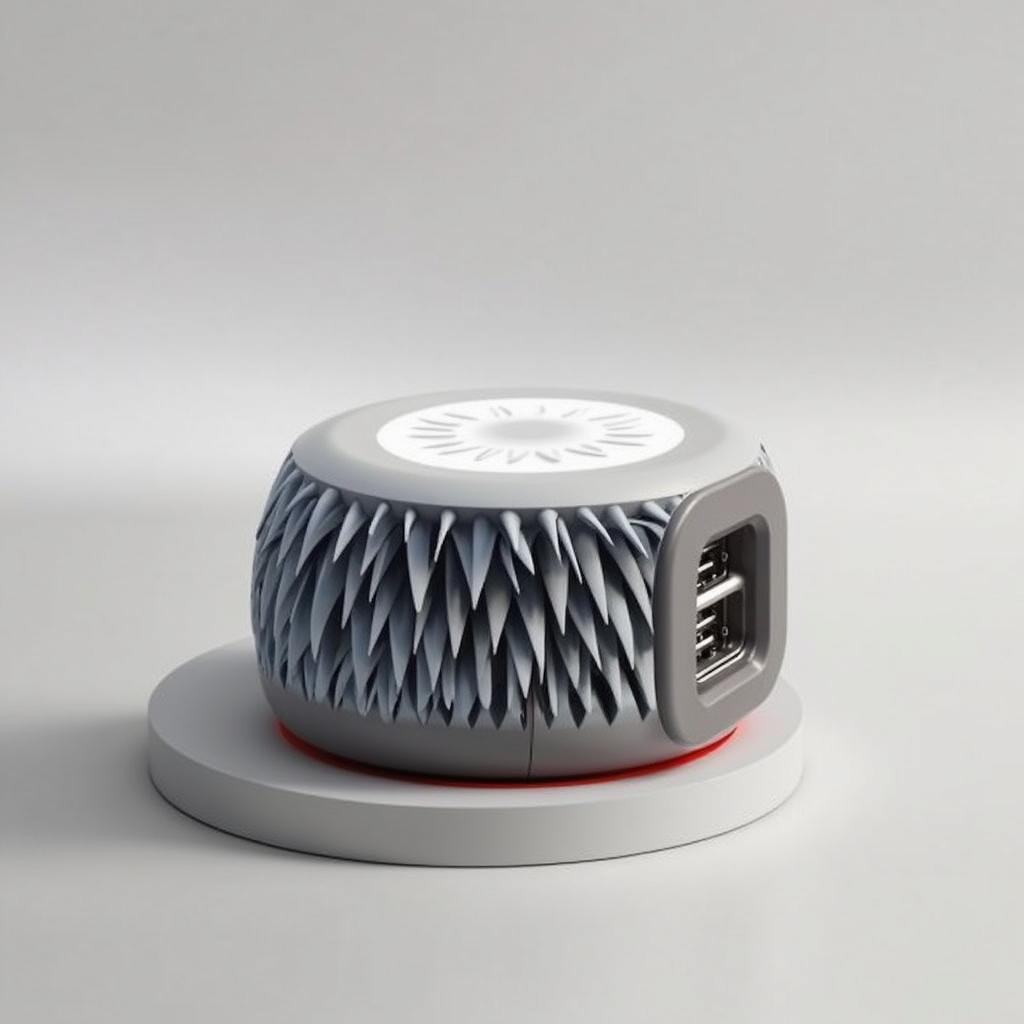} &

        \includegraphics[height=0.126\textwidth]{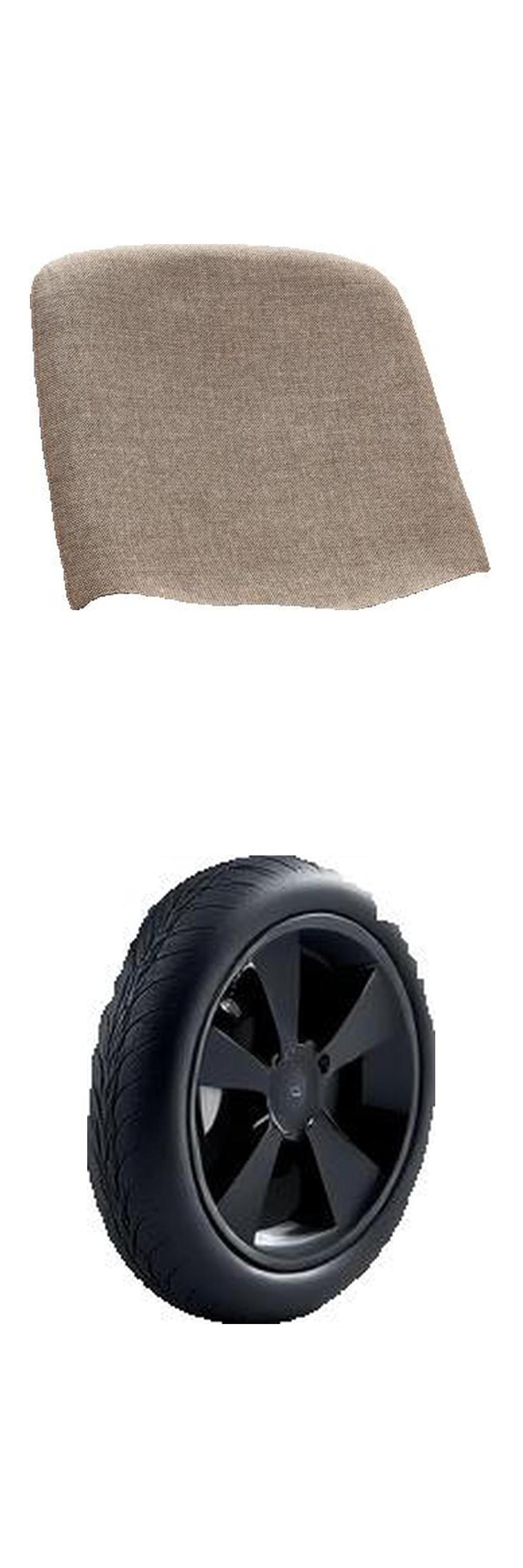} &

        \includegraphics[height=0.126\textwidth]{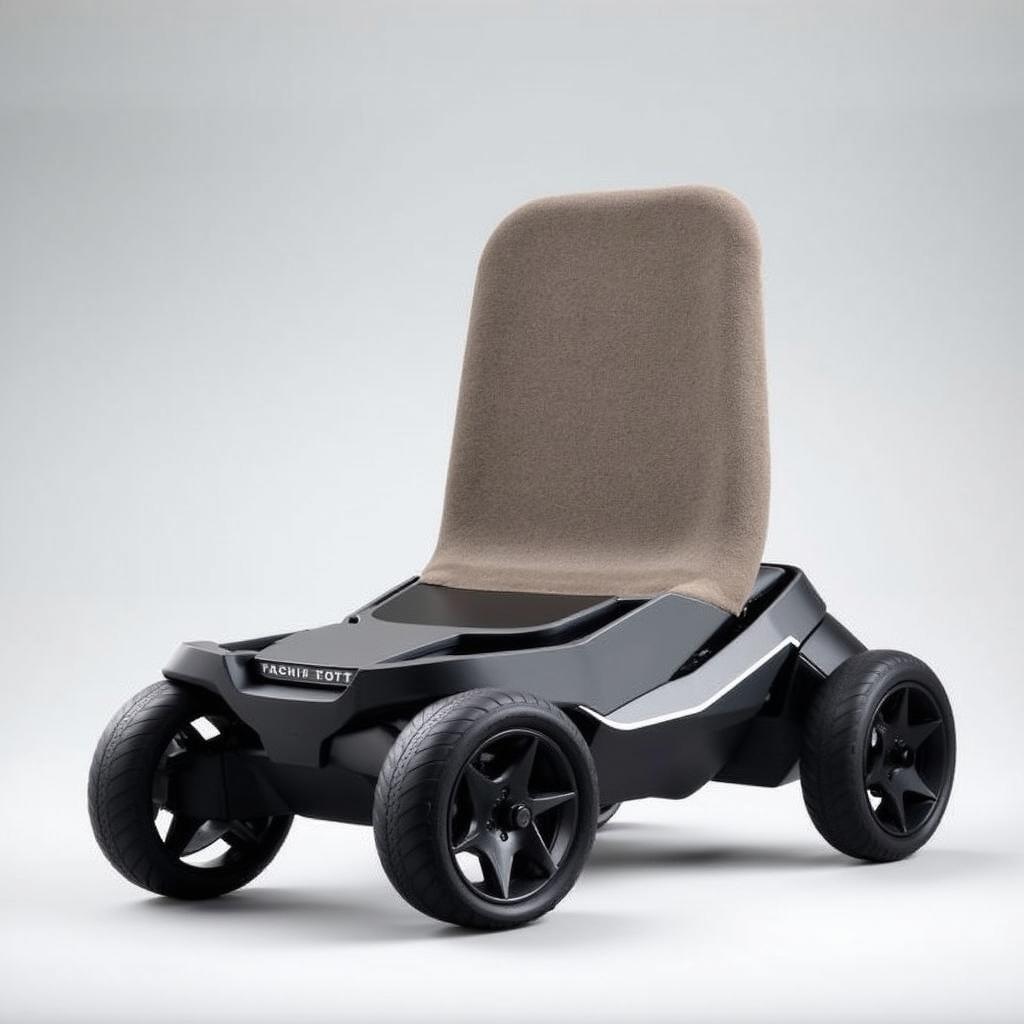} &

        \includegraphics[height=0.126\textwidth]{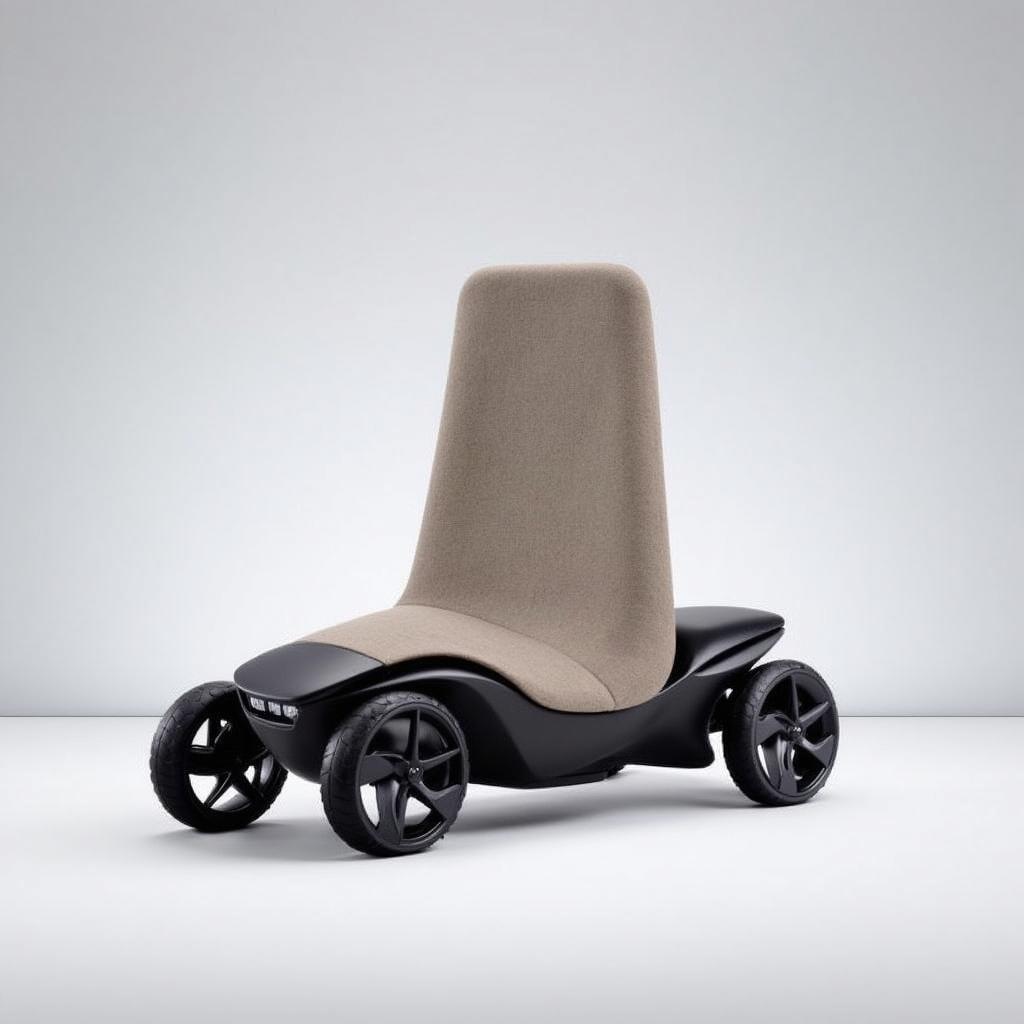} &

        \includegraphics[height=0.126\textwidth]{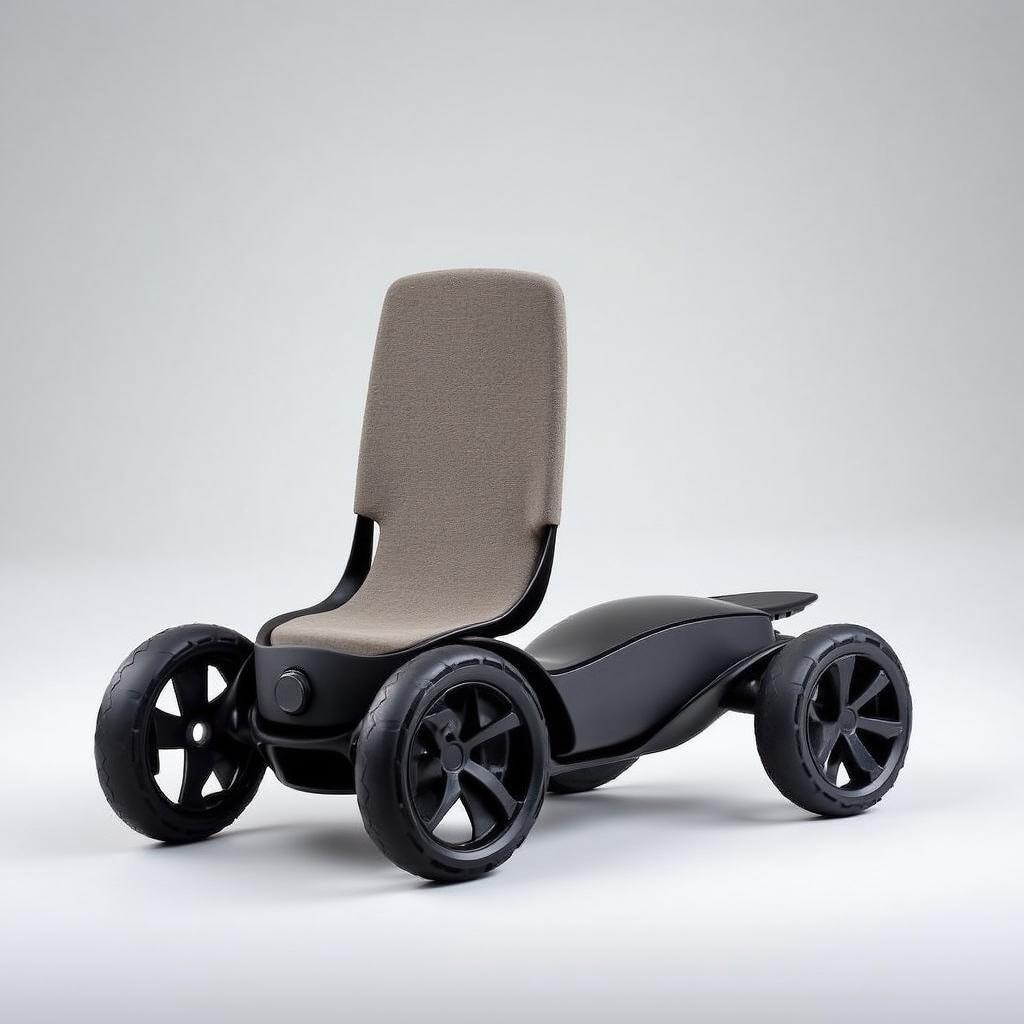}

        \\

        \raisebox{0.045\linewidth}{\rotatebox[origin=t]{90}{\begin{tabular}{c@{}c@{}c@{}c@{}} Toys \end{tabular}}} &

        \includegraphics[height=0.126\textwidth]{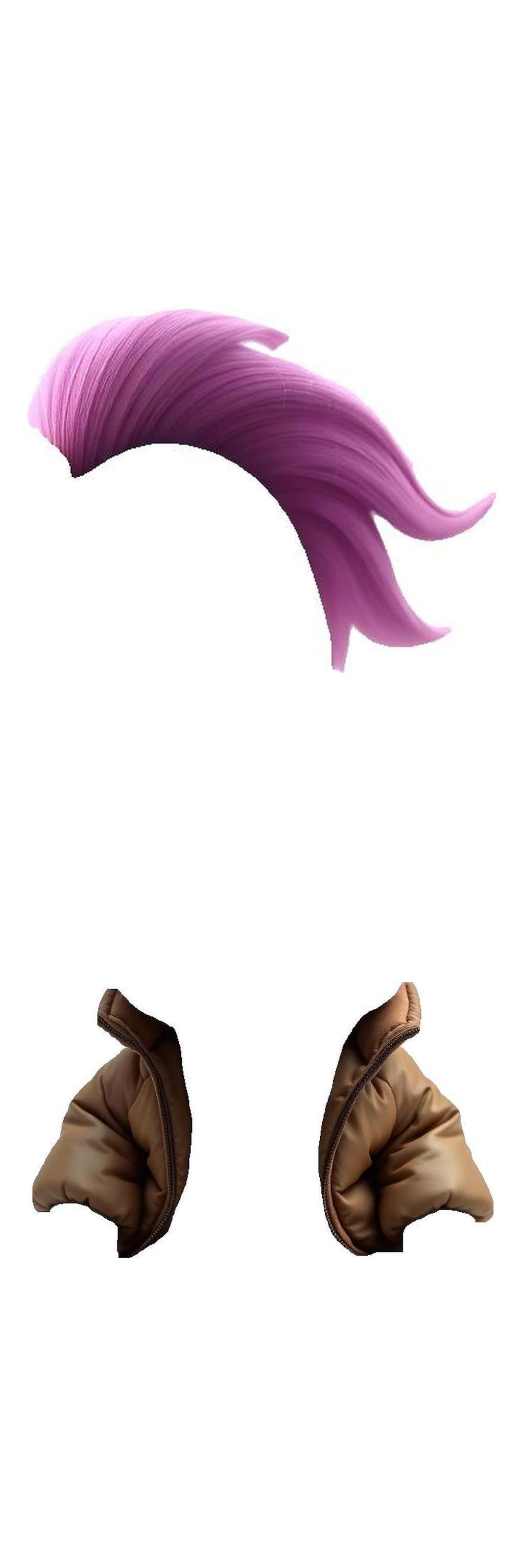} &
        \includegraphics[height=0.126\textwidth]{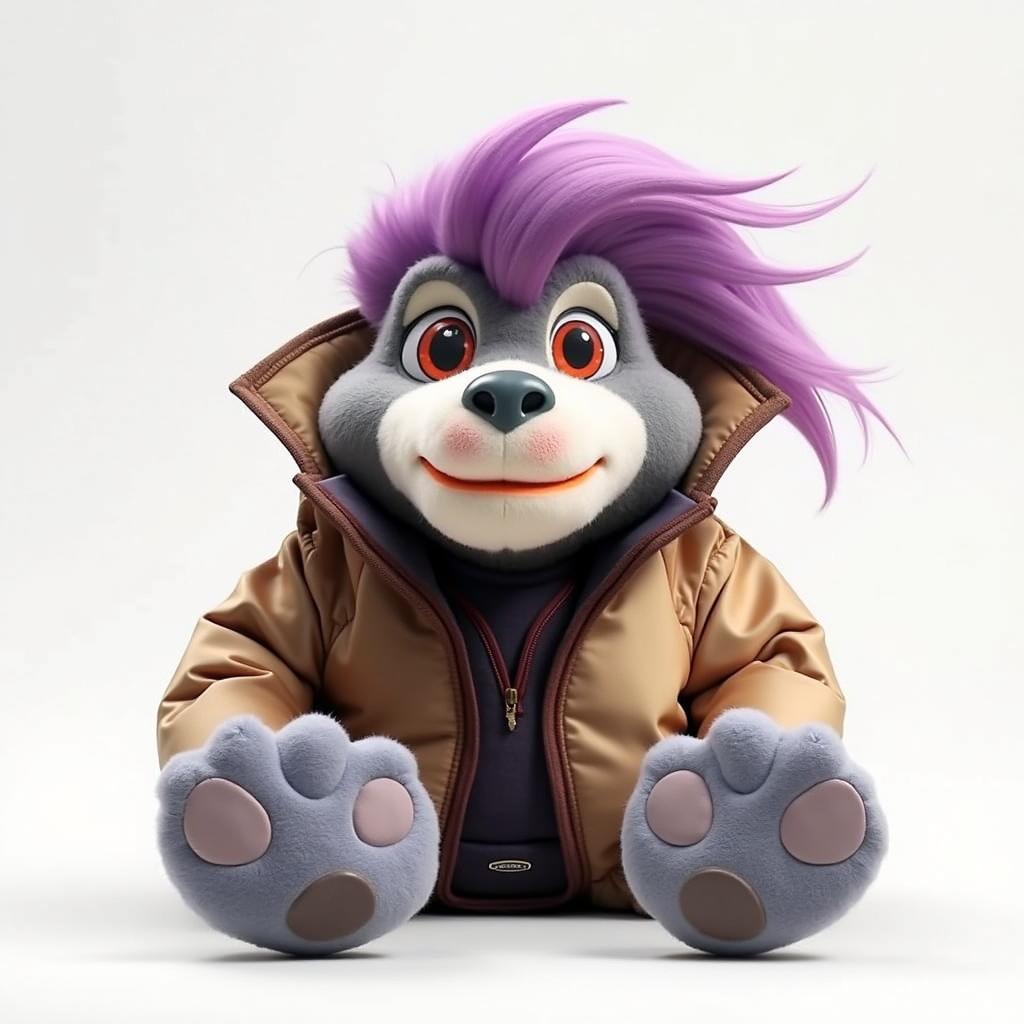} &
        \includegraphics[height=0.126\textwidth]{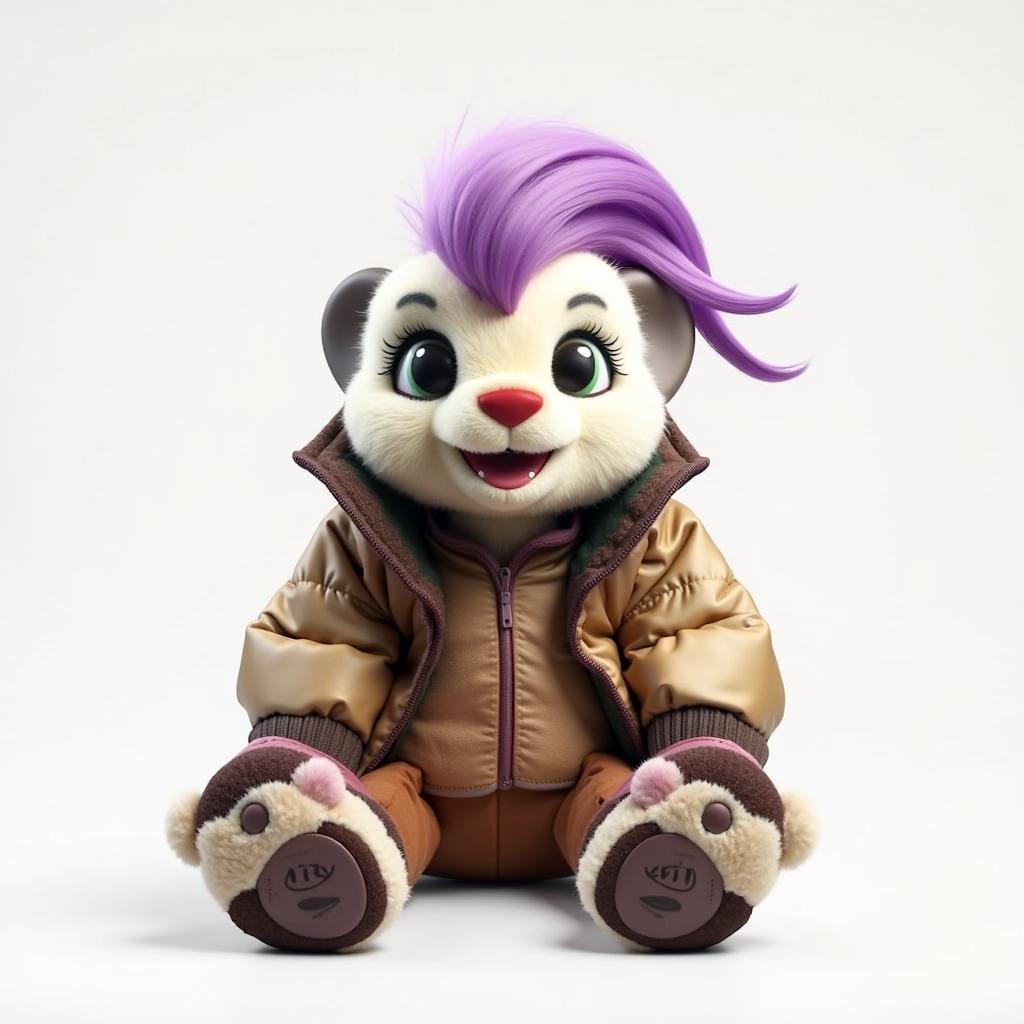} &

        \includegraphics[height=0.126\textwidth]{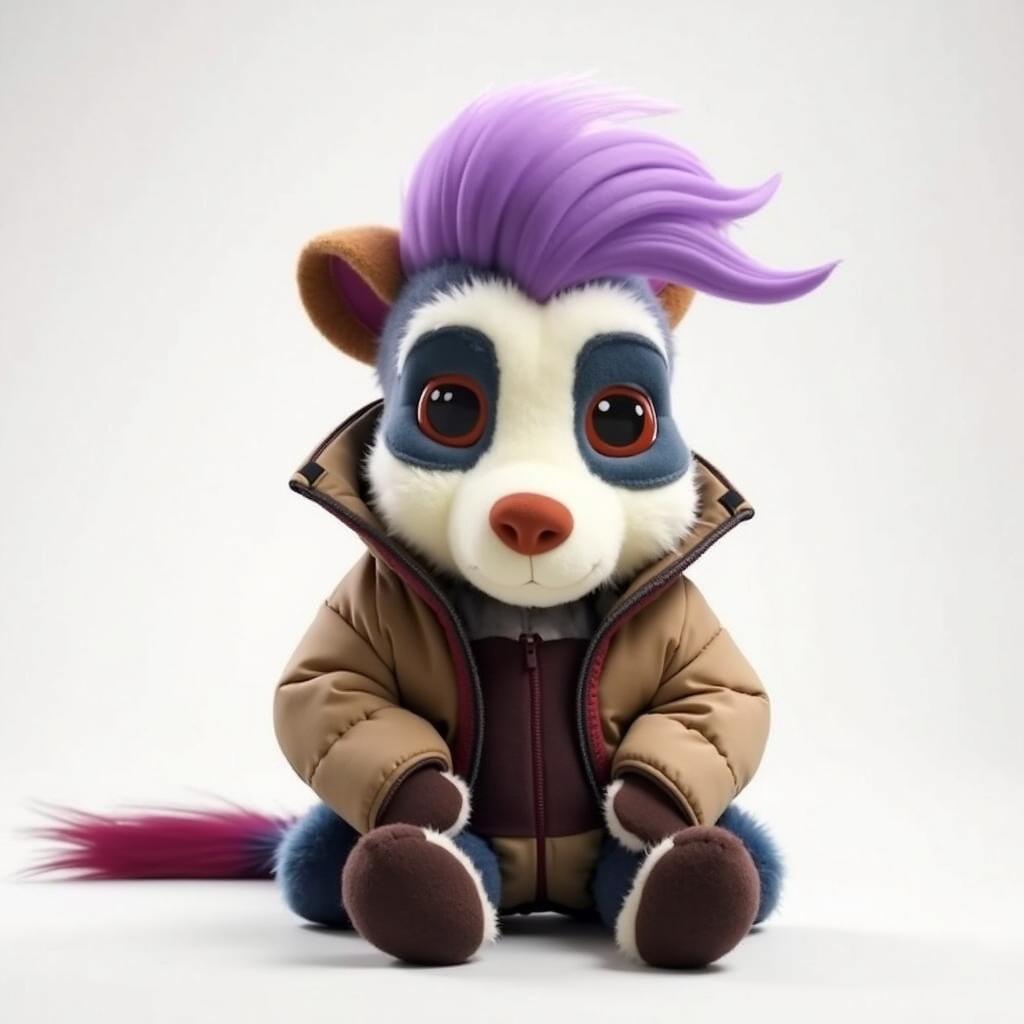} 

        &
        
        \includegraphics[height=0.126\textwidth]{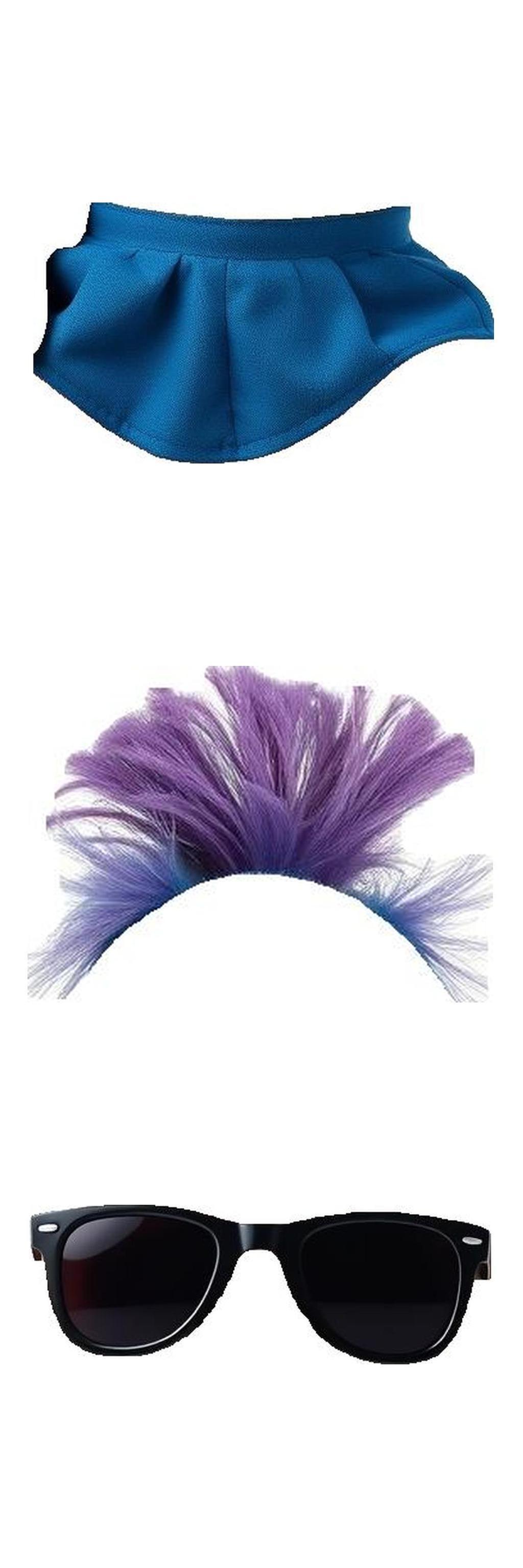} &
        \includegraphics[height=0.126\textwidth]{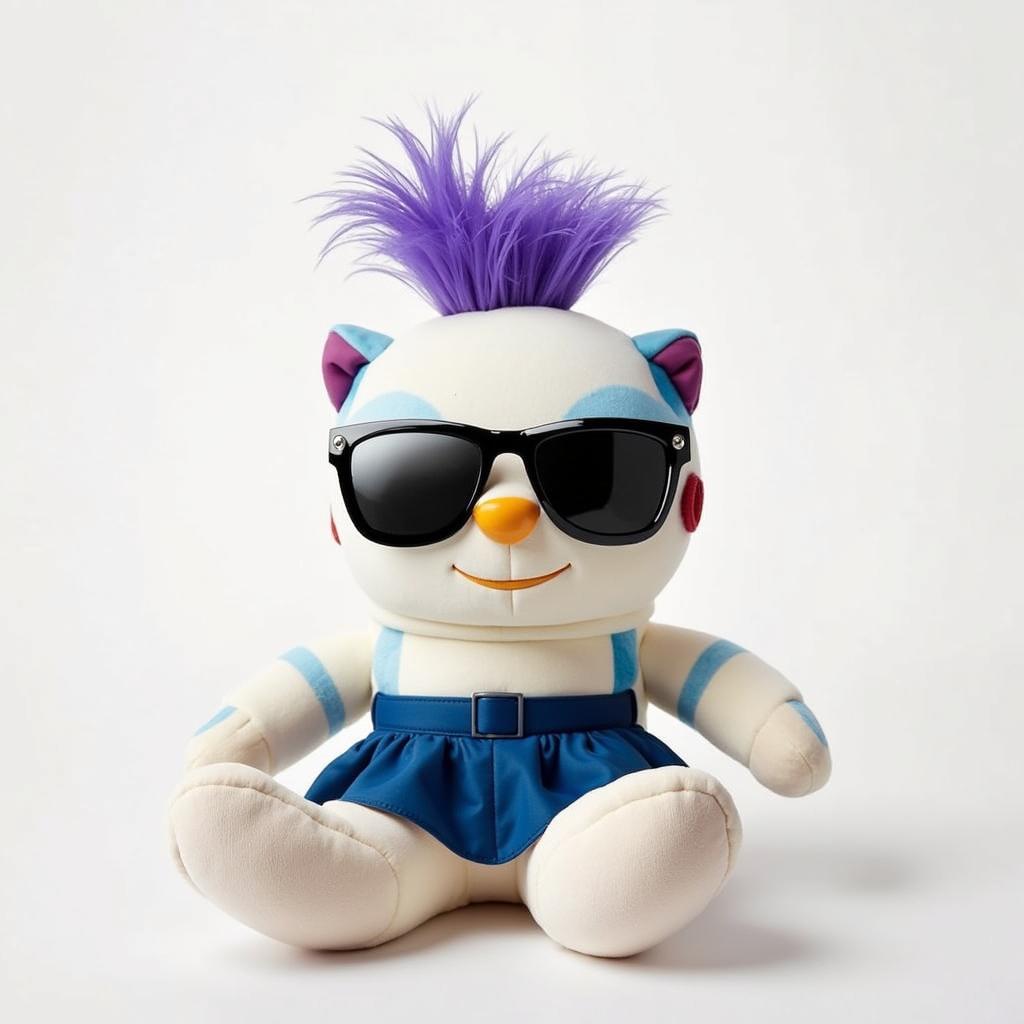} &
        \includegraphics[height=0.126\textwidth]{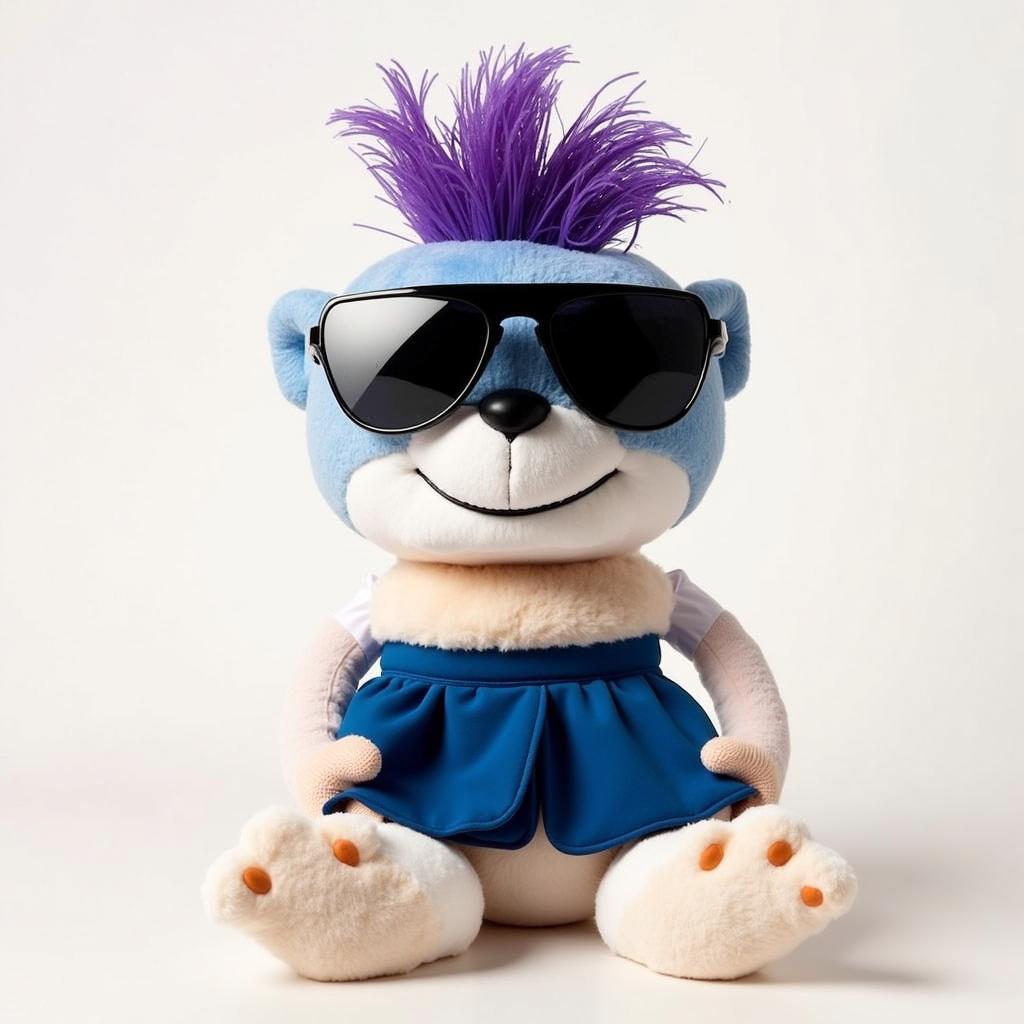} &

        \includegraphics[height=0.126\textwidth]{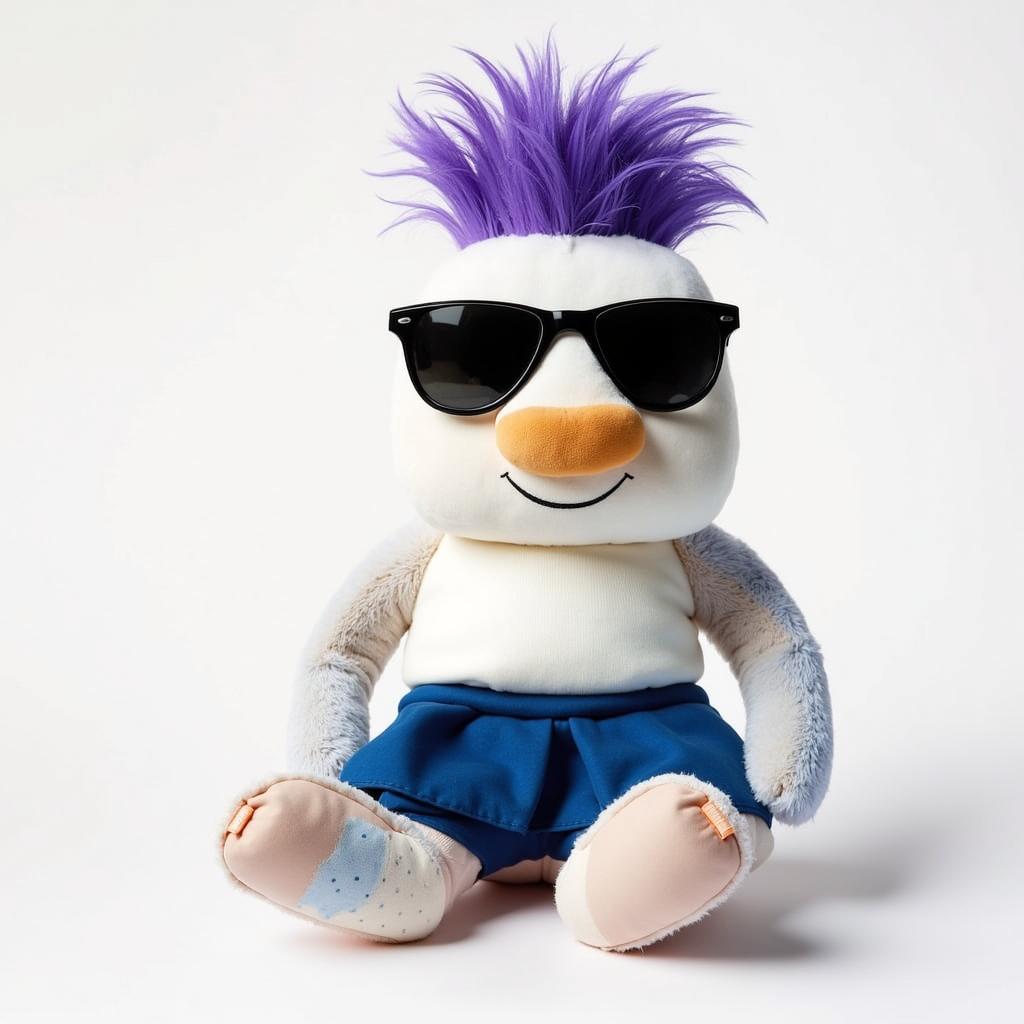} 

        \\

    \end{tabular}
    }
    \vspace{-0.3cm}
    \caption{\textbf{PiT Results for Different Priors.}
    We show results generated by our approach across three different domains. For each result, we use a varying number of input parts and generate multiple plausible outputs by altering the seed used for learning the representation. 
    }
    \vspace{-0.1cm}
    \label{fig:large_in_paper}
\end{figure*}

\subsection{Implementation Details}
Our IP-Prior consists of approximately 270M parameters. Training is performed with a batch size of 64 for 500K steps, requiring 30GB of RAM on a single GPU. 
The IP-LoRA training is conducted using the AdamW optimizer with a learning rate of $1\times10^{-4}$ for 10K steps. 
During inference, the IP-Prior model runs for 25 steps, while SDXL uses 50 steps. To enhance fine details, we apply SDEdit~\cite{meng2021sdedit} over Flux-Dev~\cite{blackforest2024flux} with a strength of 0.35.

\section{Experiments}
\subsection{IP-Prior Results}
\paragraph{Qualitative Results.}
In~\Cref{fig:large_in_paper}, we present the results of PiT across three different priors. During inference, our model receives a sequence of tokens of possibly varying lengths and generates a corresponding output for each seed. The parts themselves are gathered from online sources as well as from newly generated images.  As shown, our model successfully recognizes the semantic meaning of each given part and integrates it coherently into the generated result. Notably, this process is entirely free of any additional text supervision, requiring the user to provide the desired object. Furthermore, beyond merely integrating the given parts, the model effectively generates meaningful and coherent completions to the missing information, ultimately producing an in-domain result.

\begin{figure}
    \centering
    \setlength{\tabcolsep}{0.5pt}
    \addtolength{\belowcaptionskip}{-5pt}
    \renewcommand{\arraystretch}{0.5}
    {\small
    \begin{tabular}{c @{\hspace{0.2cm}} c c c c}

        \includegraphics[width=0.07\textwidth]{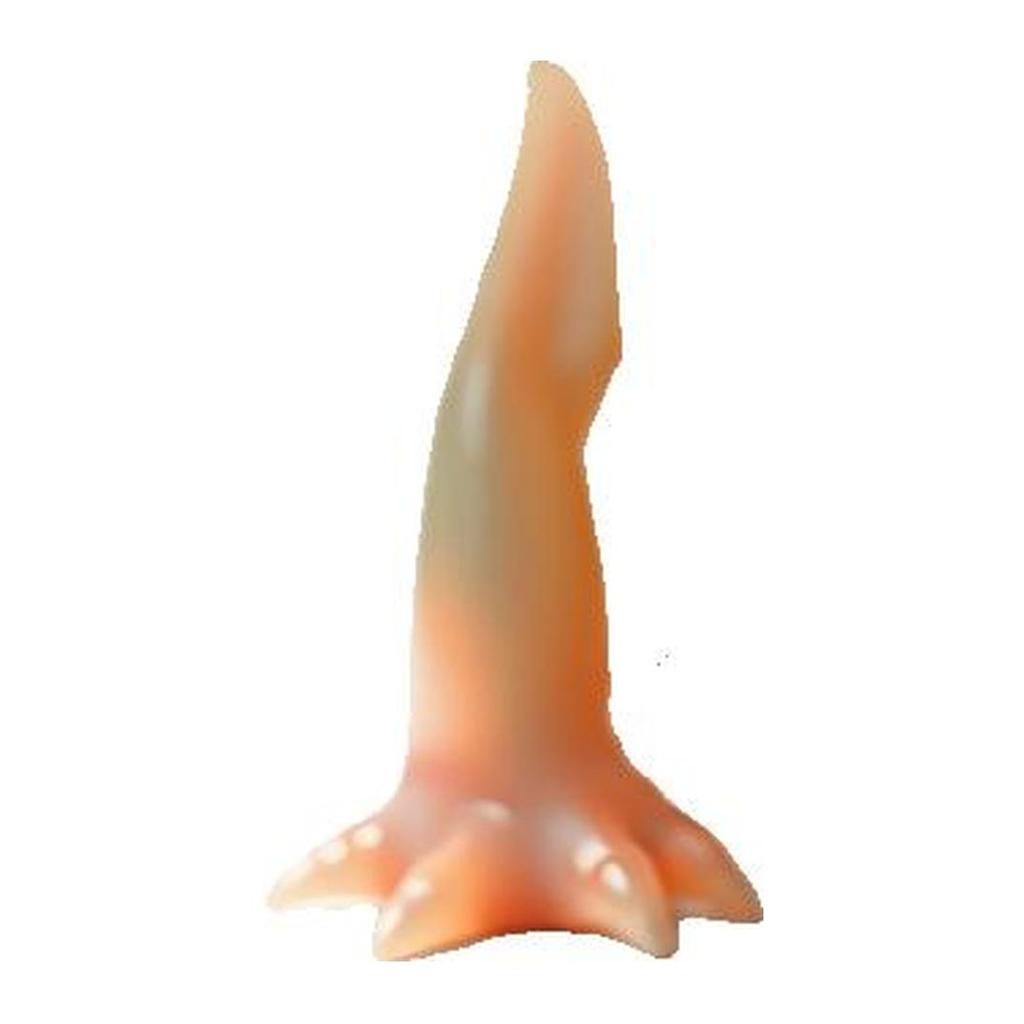} &
        \includegraphics[width=0.0875\textwidth]{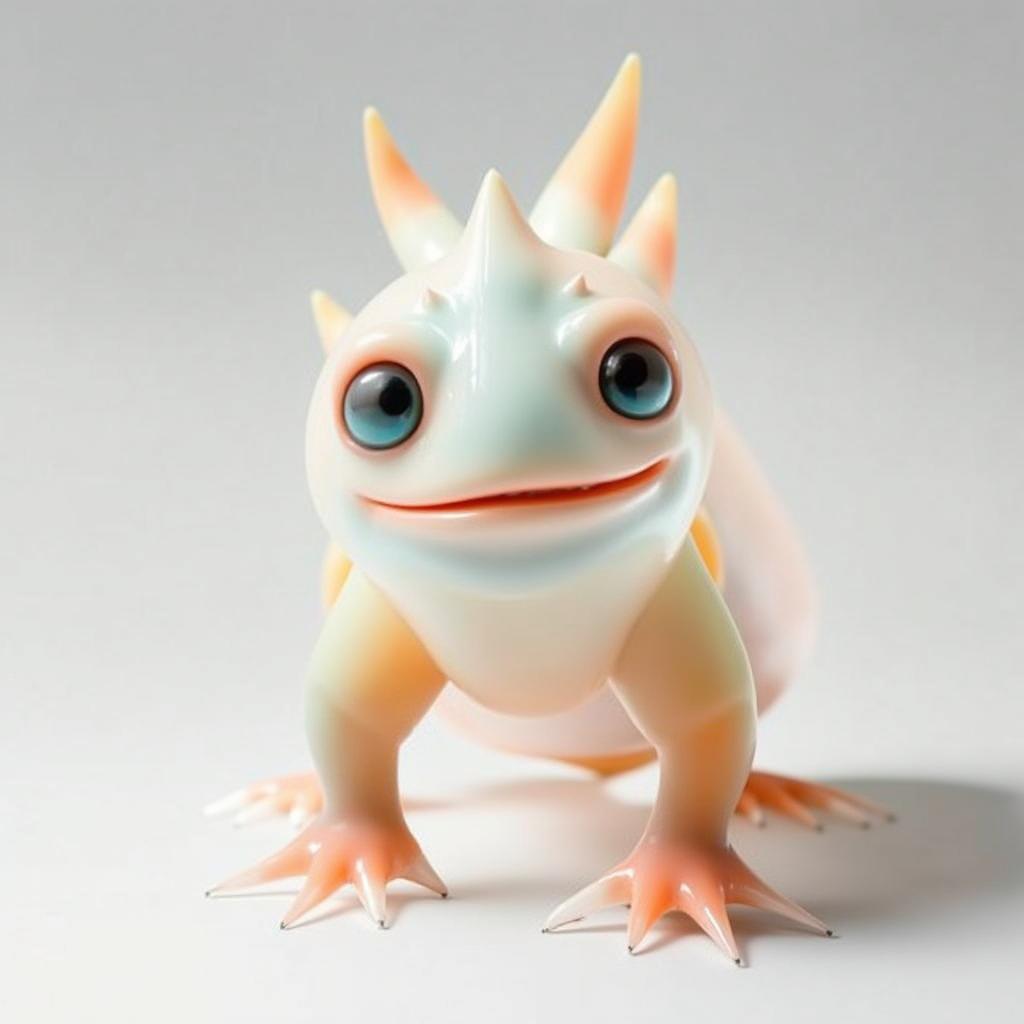} &
        \includegraphics[width=0.0875\textwidth]{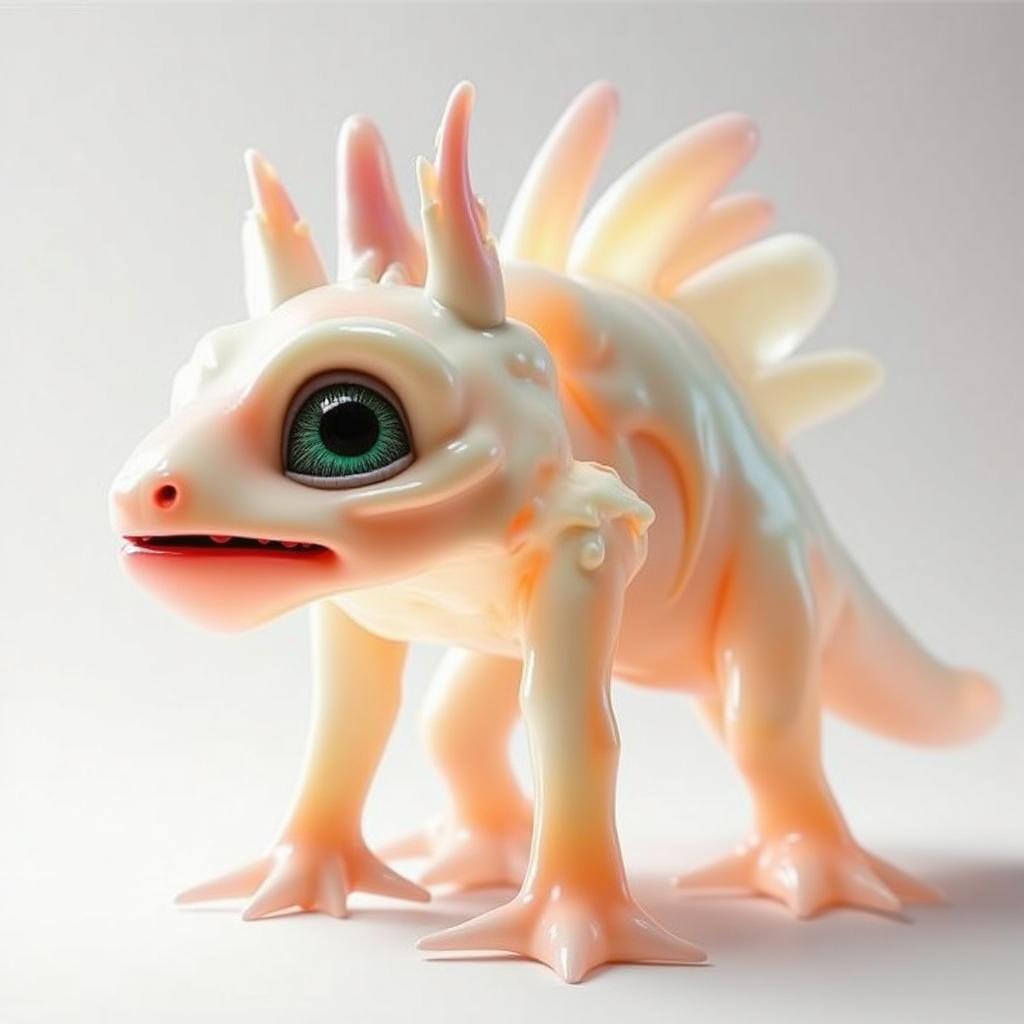} &

        \includegraphics[width=0.0875\textwidth]{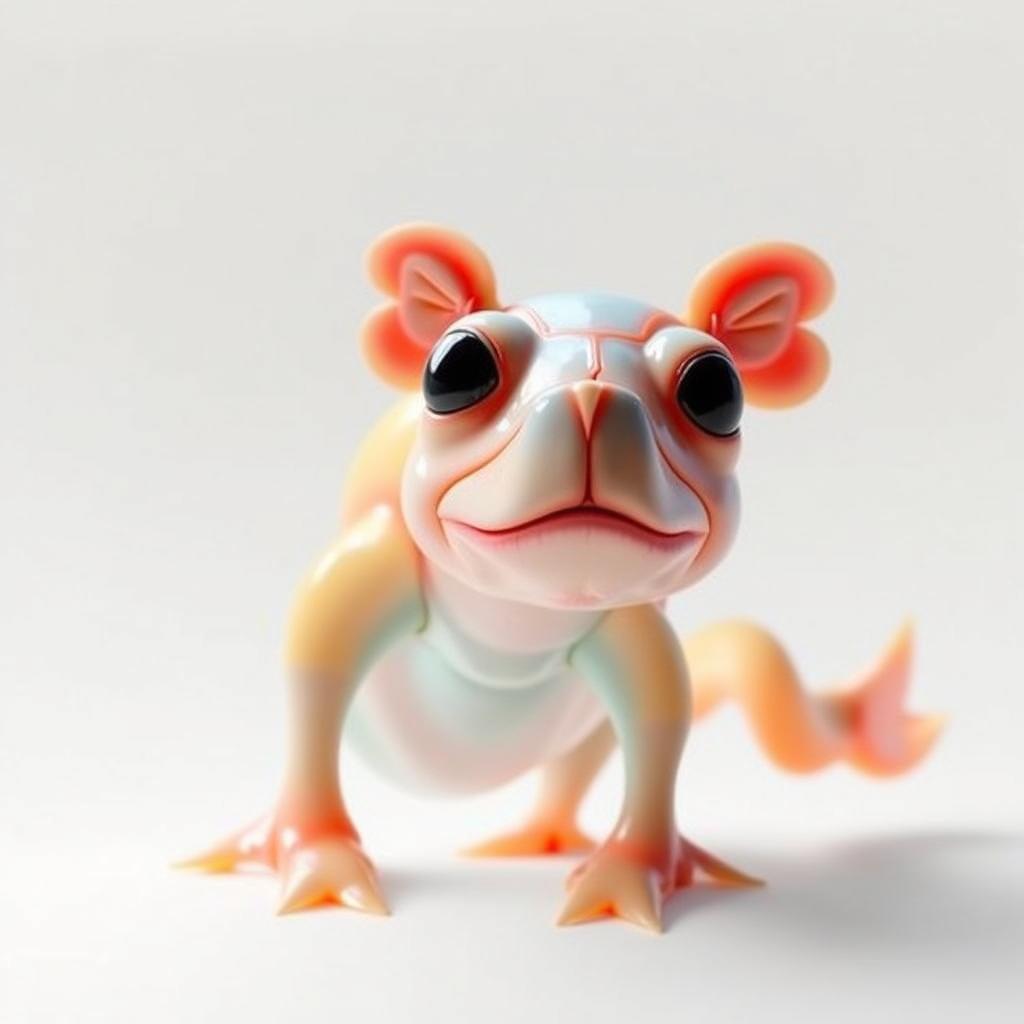} &

        \includegraphics[width=0.0875\textwidth]{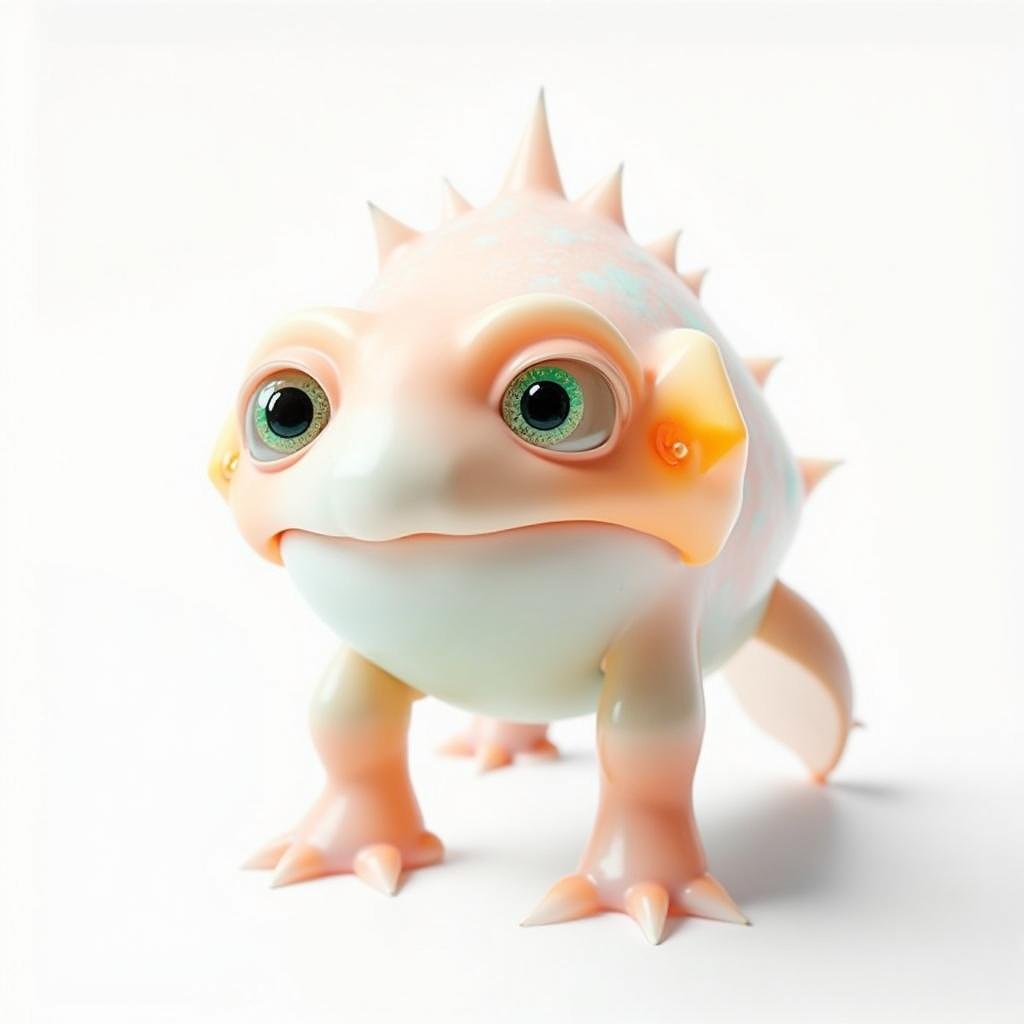} \\

         \includegraphics[width=0.07\textwidth]{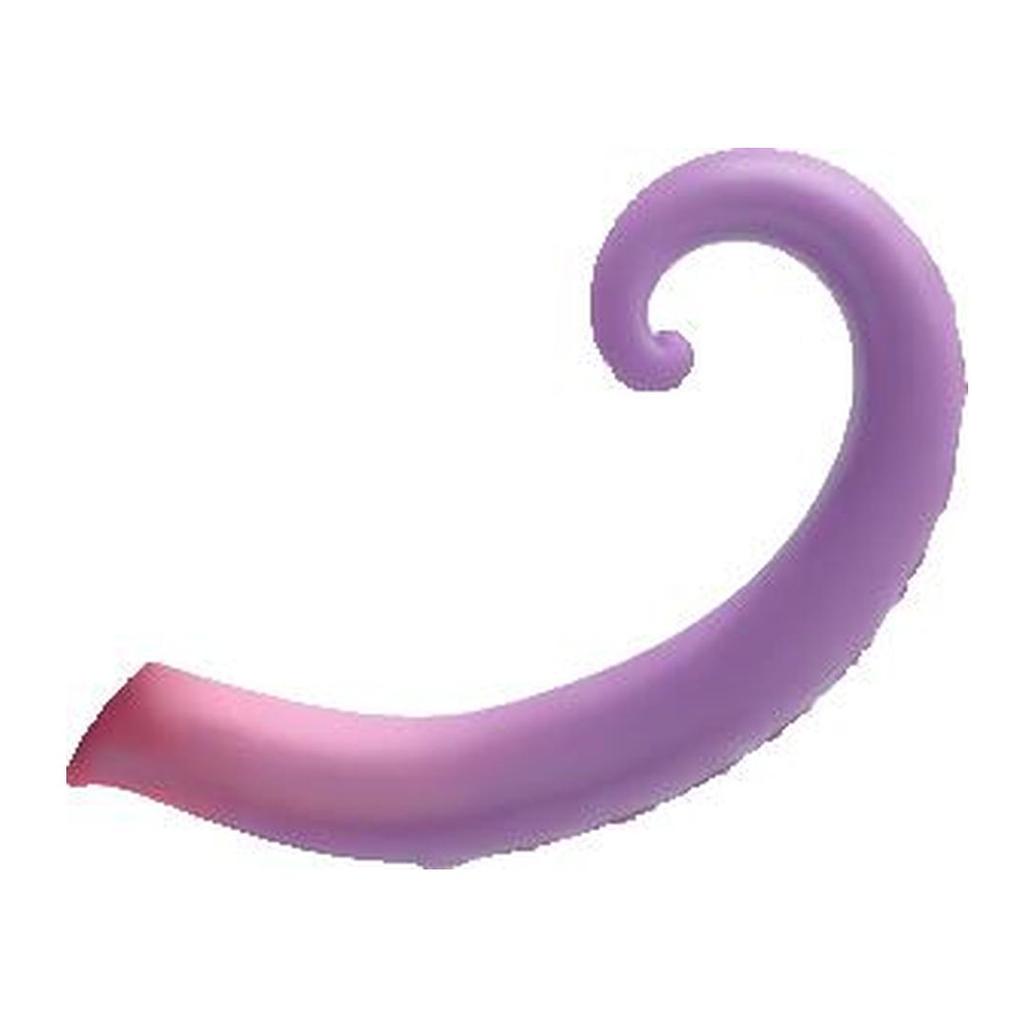} &
        \includegraphics[width=0.0875\textwidth]{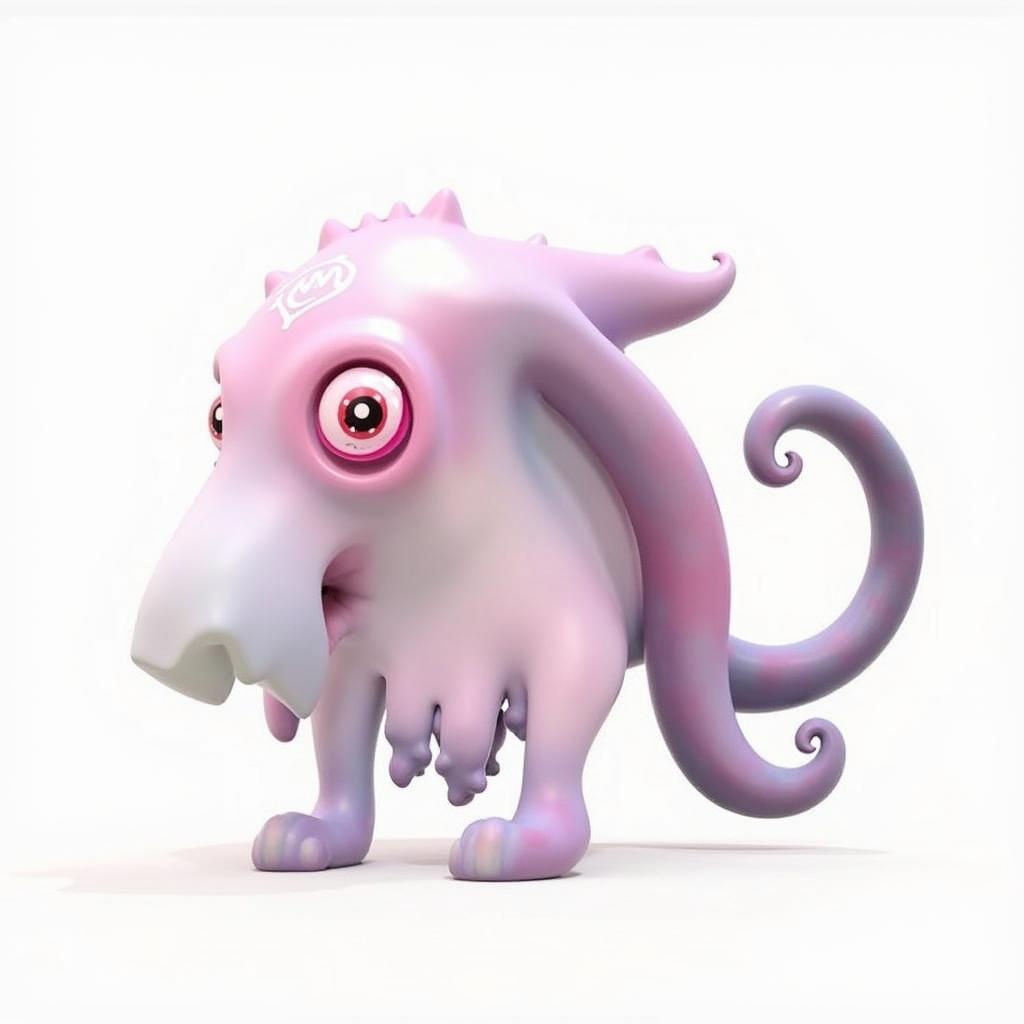} &
        \includegraphics[width=0.0875\textwidth]{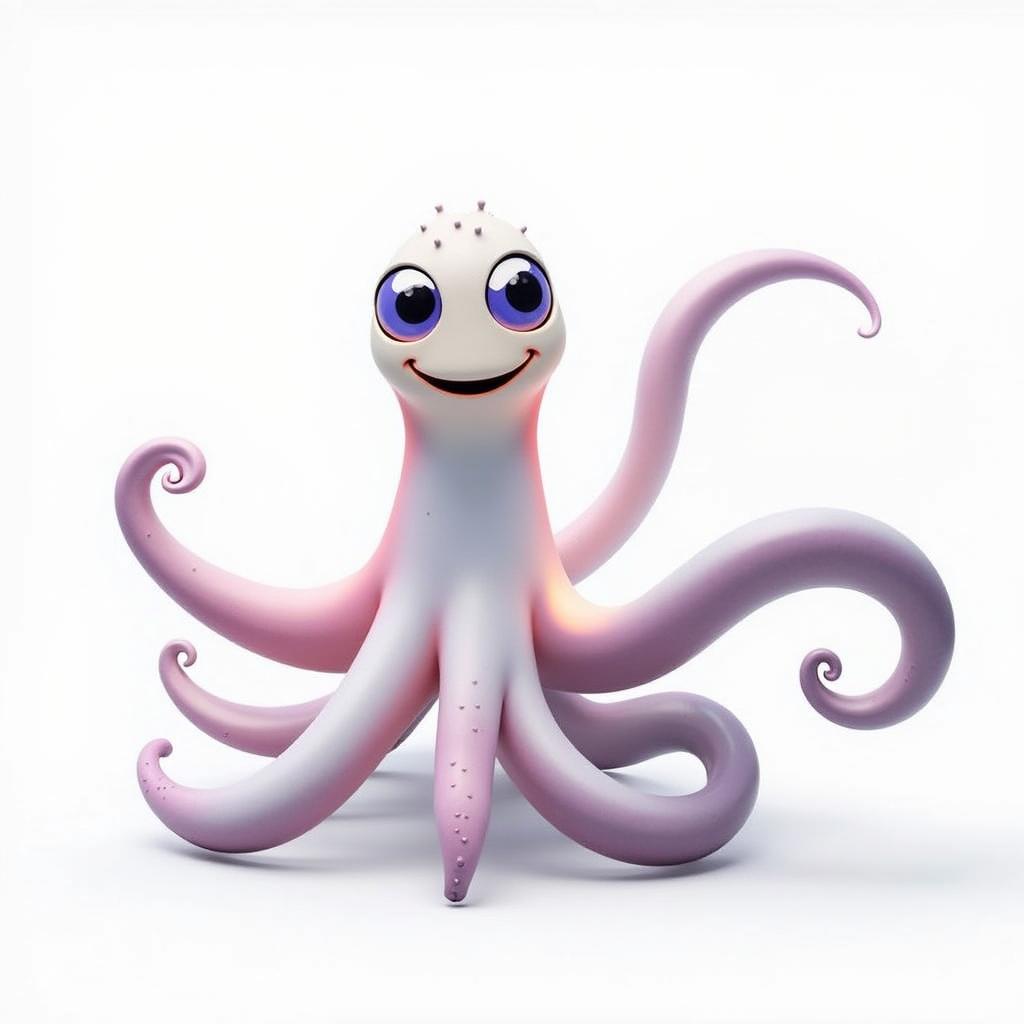} &

        \includegraphics[width=0.0875\textwidth]{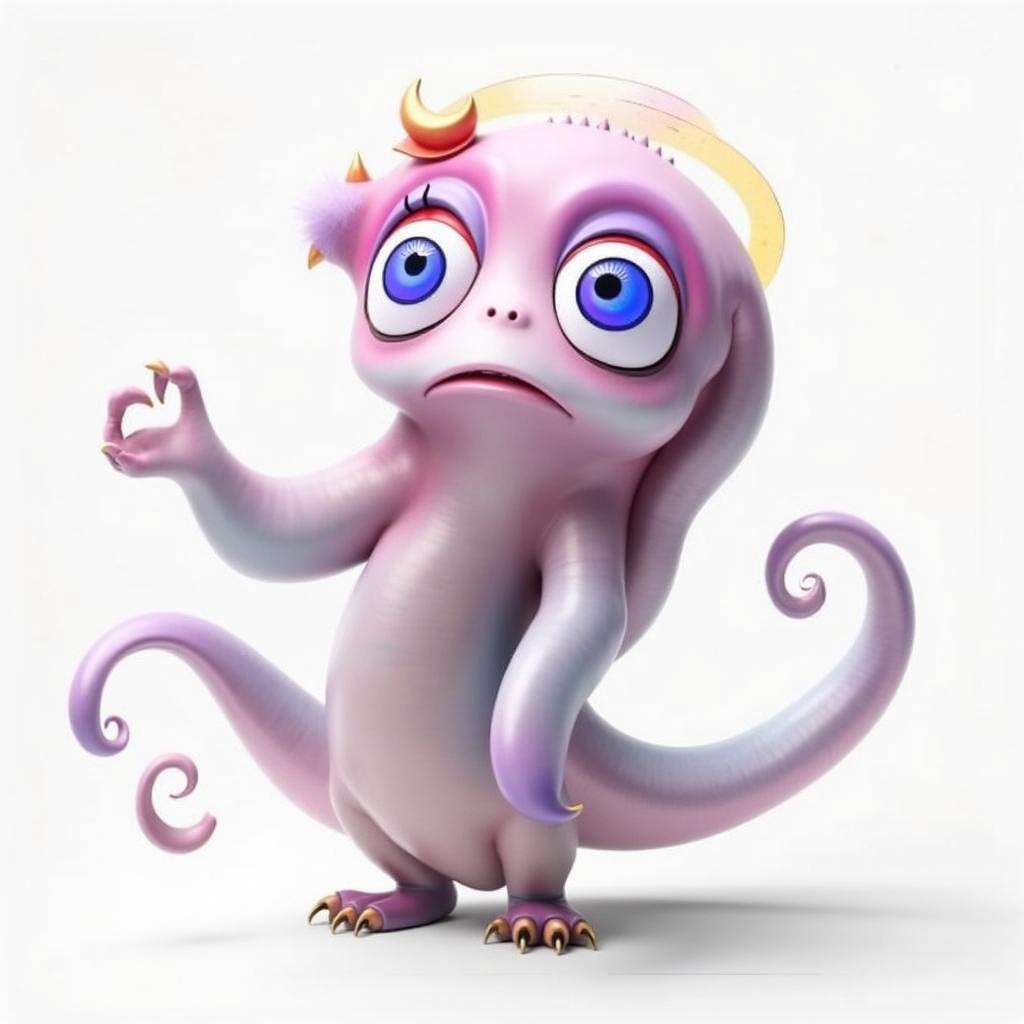} &

        \includegraphics[width=0.0875\textwidth]{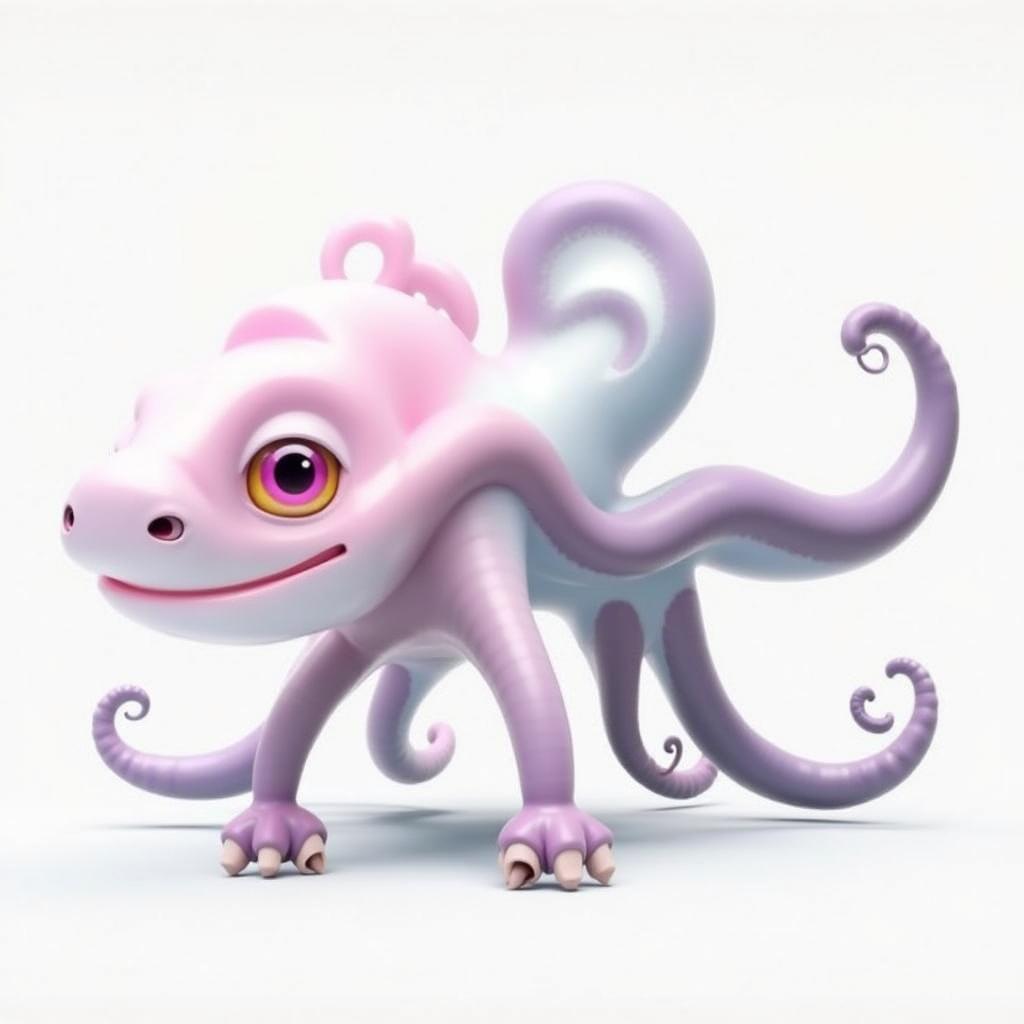} \\

         \includegraphics[width=0.07\textwidth]{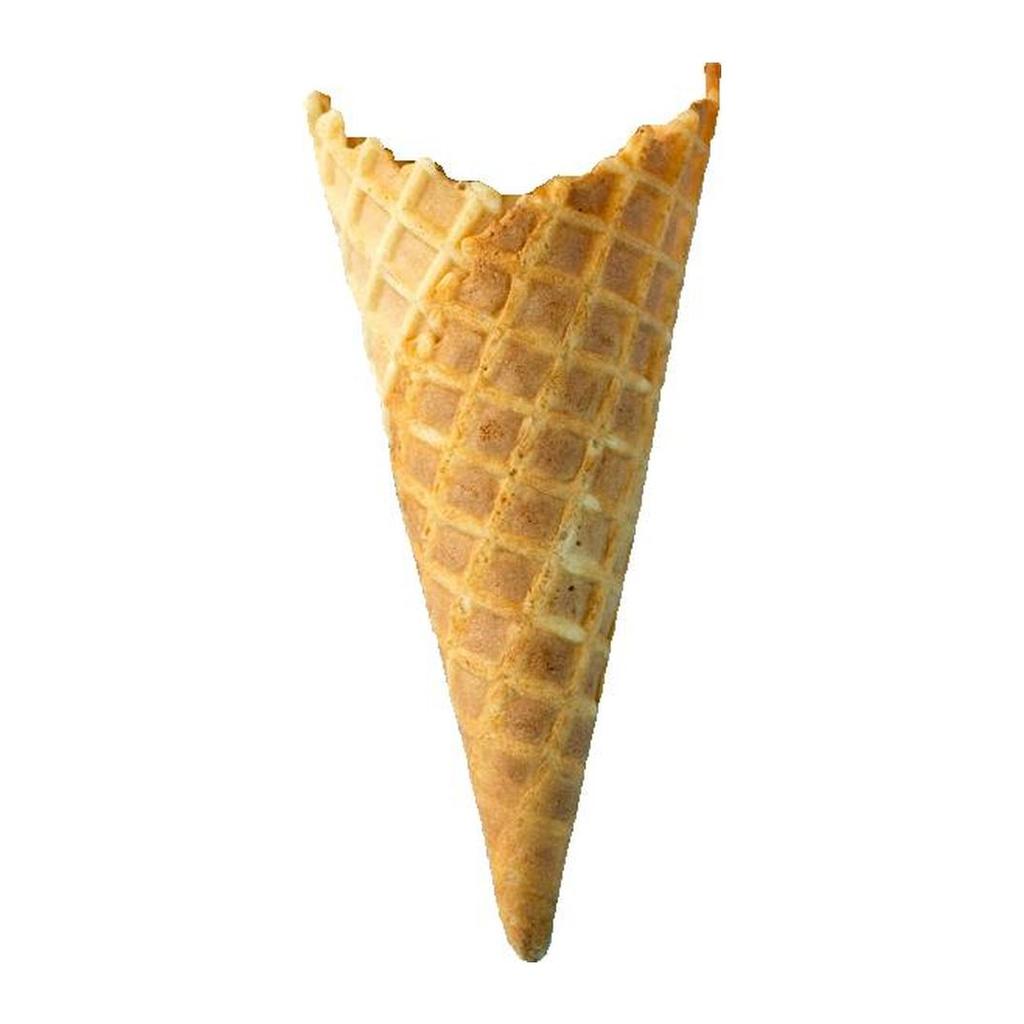} &
        \includegraphics[width=0.0875\textwidth]{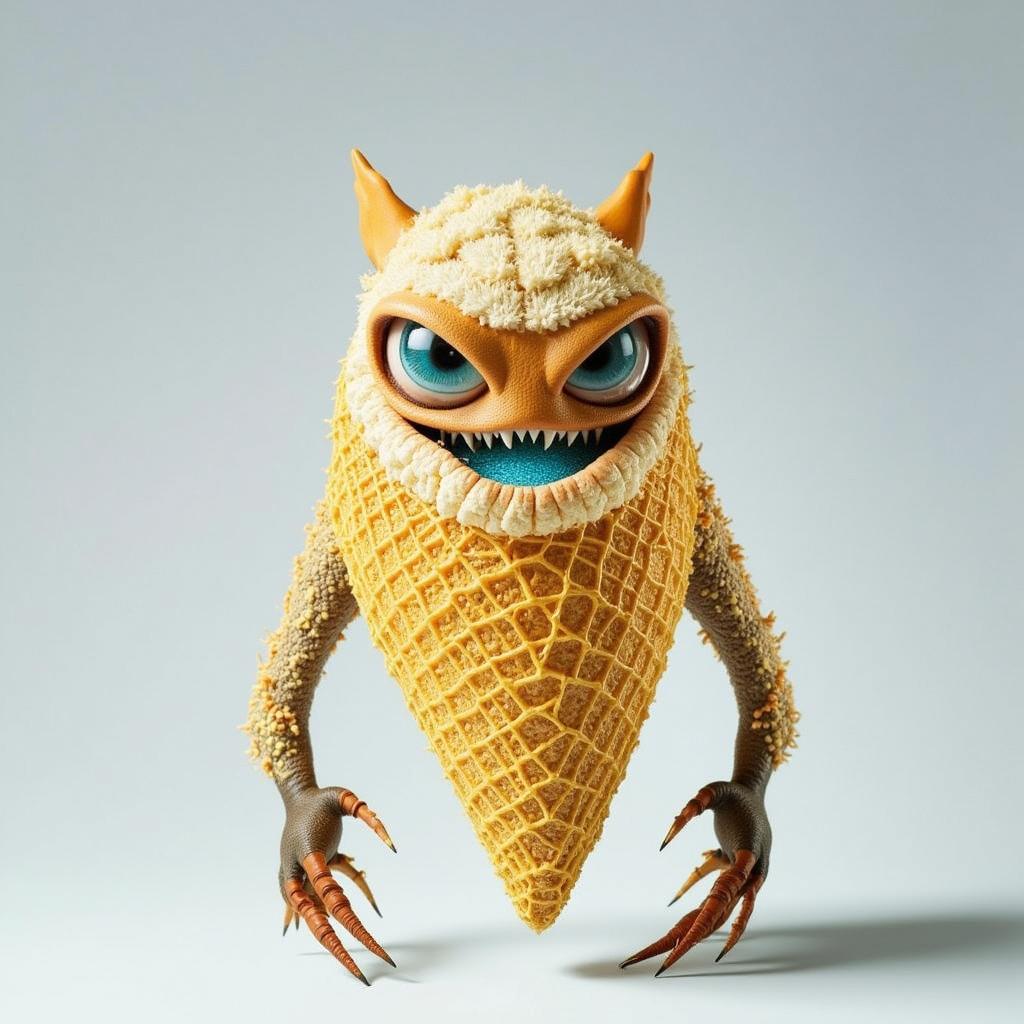} &
        \includegraphics[width=0.0875\textwidth]{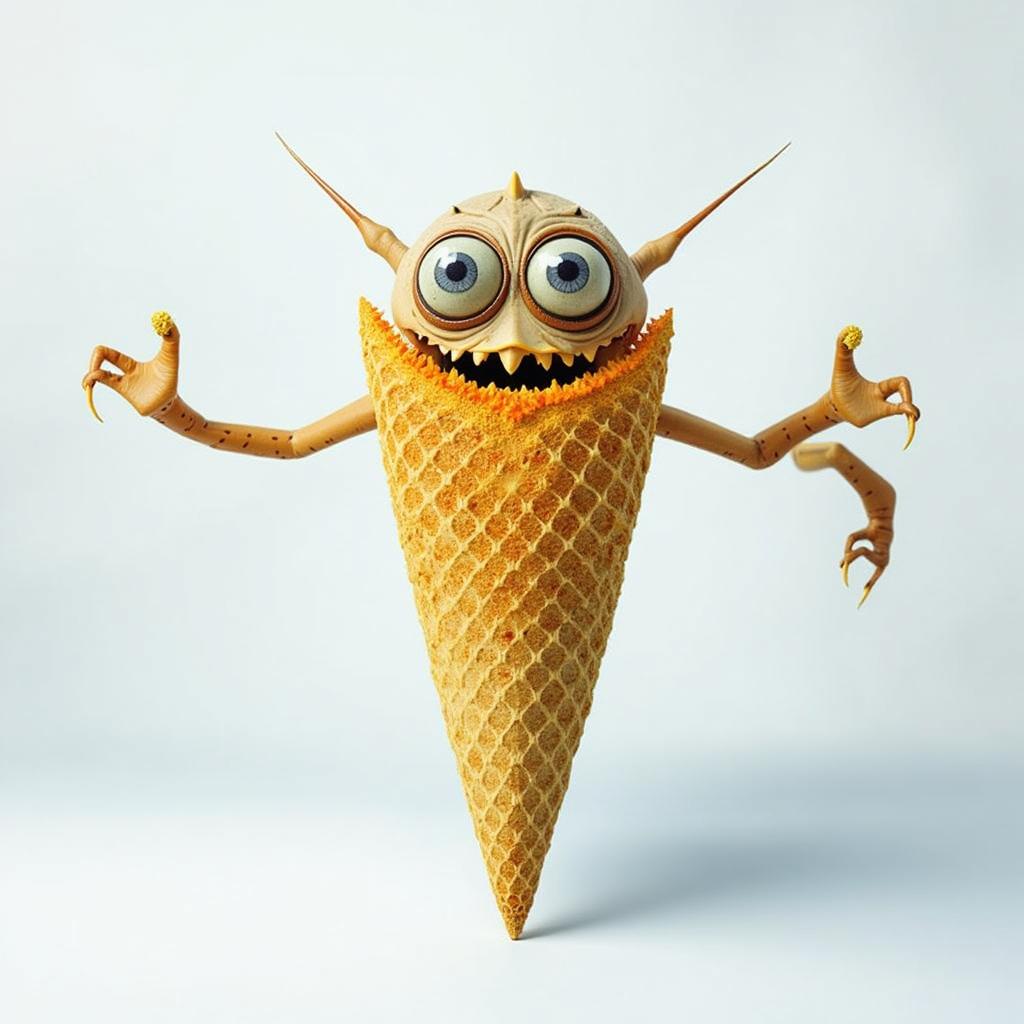} &

        \includegraphics[width=0.0875\textwidth]{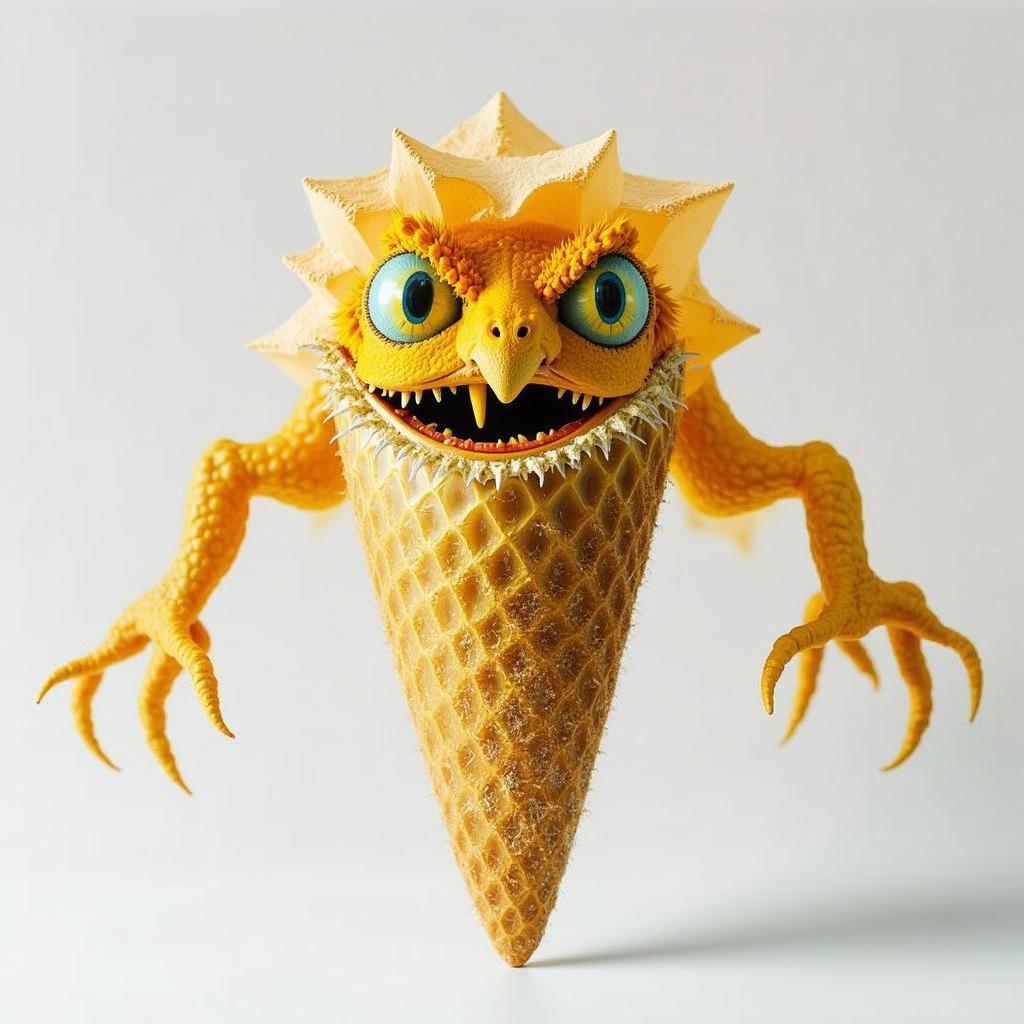} &

        \includegraphics[width=0.0875\textwidth]{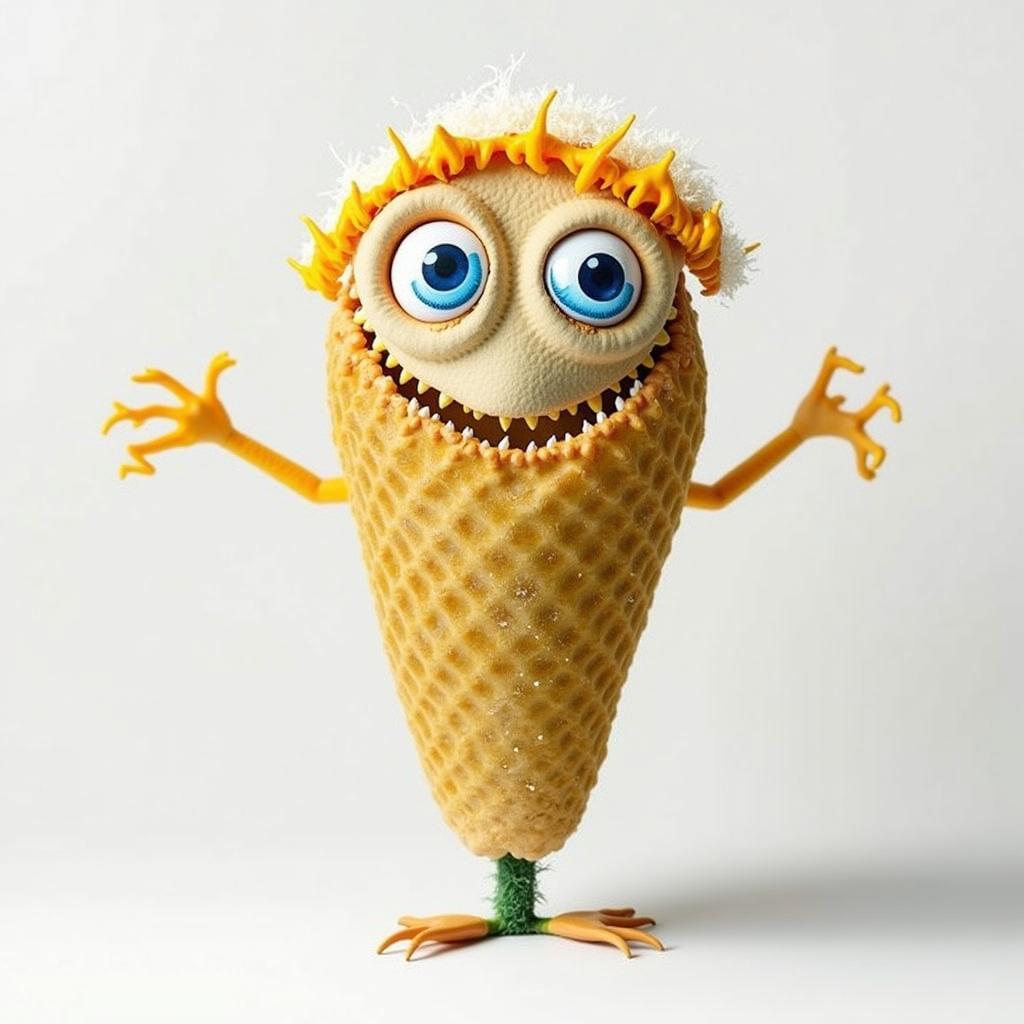} \\

        \includegraphics[width=0.07\textwidth]{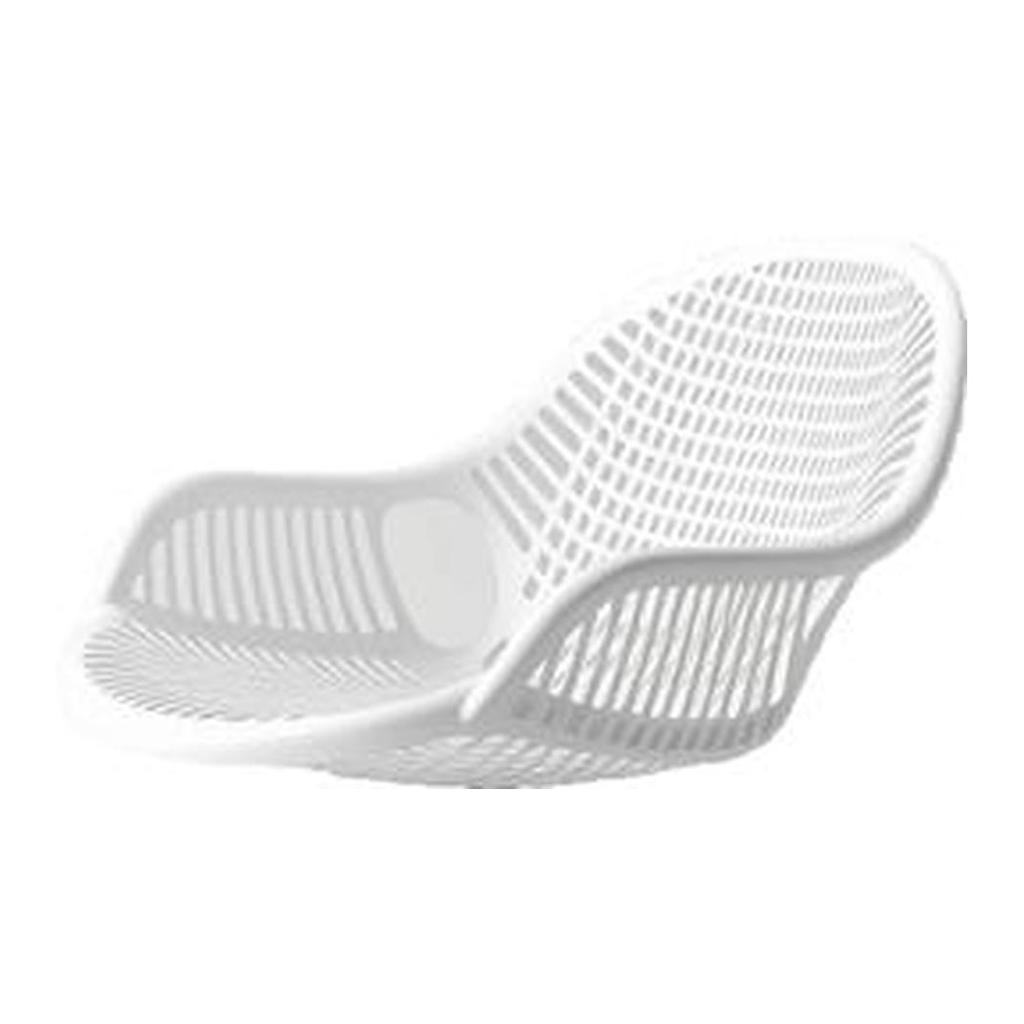} &
        \includegraphics[width=0.0875\textwidth]{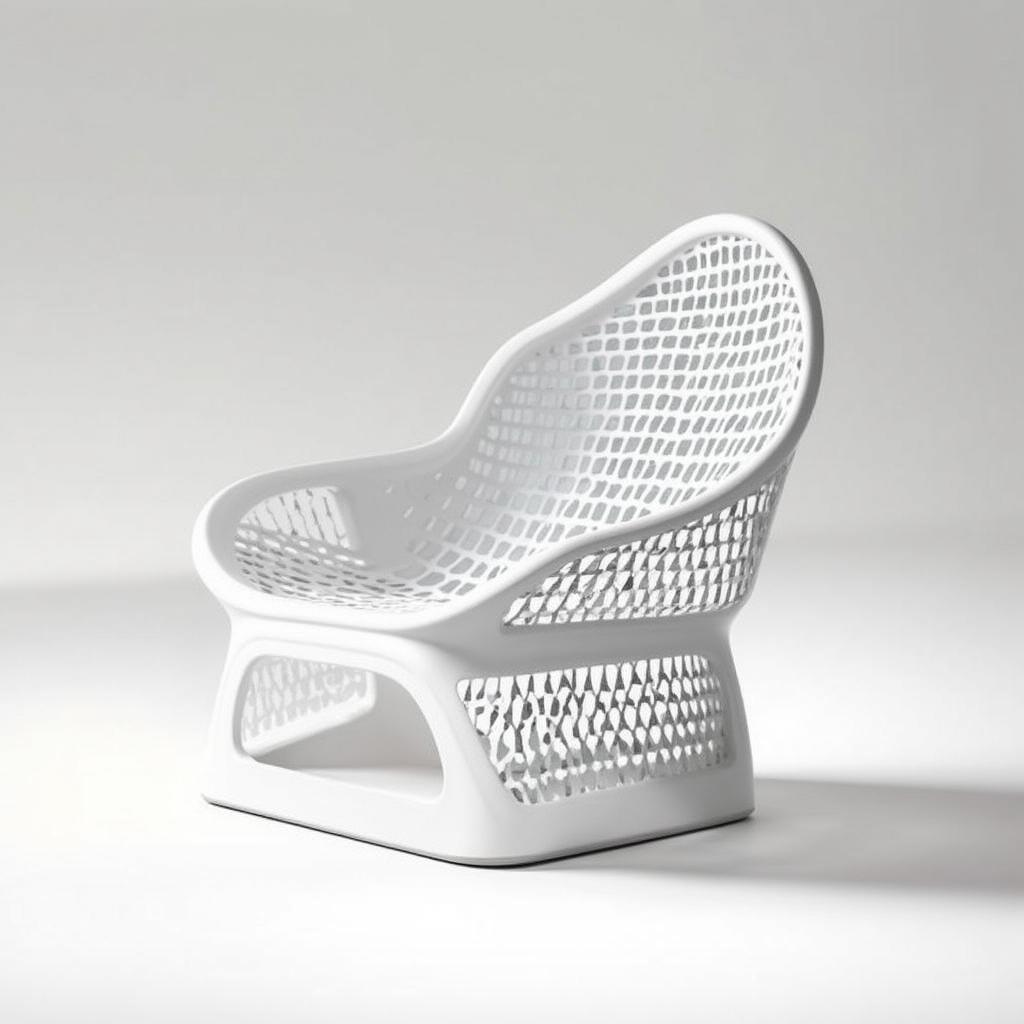} &
        \includegraphics[width=0.0875\textwidth]{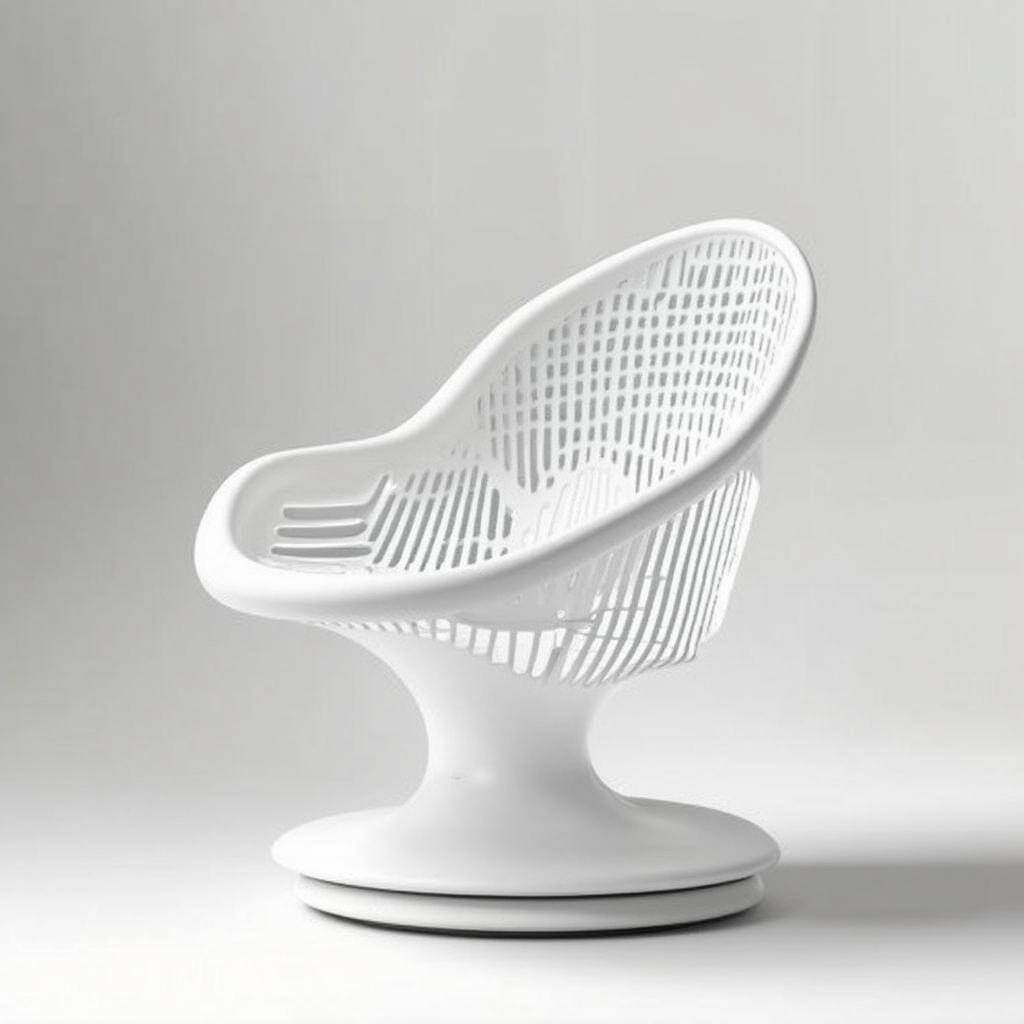} &

        \includegraphics[width=0.0875\textwidth]{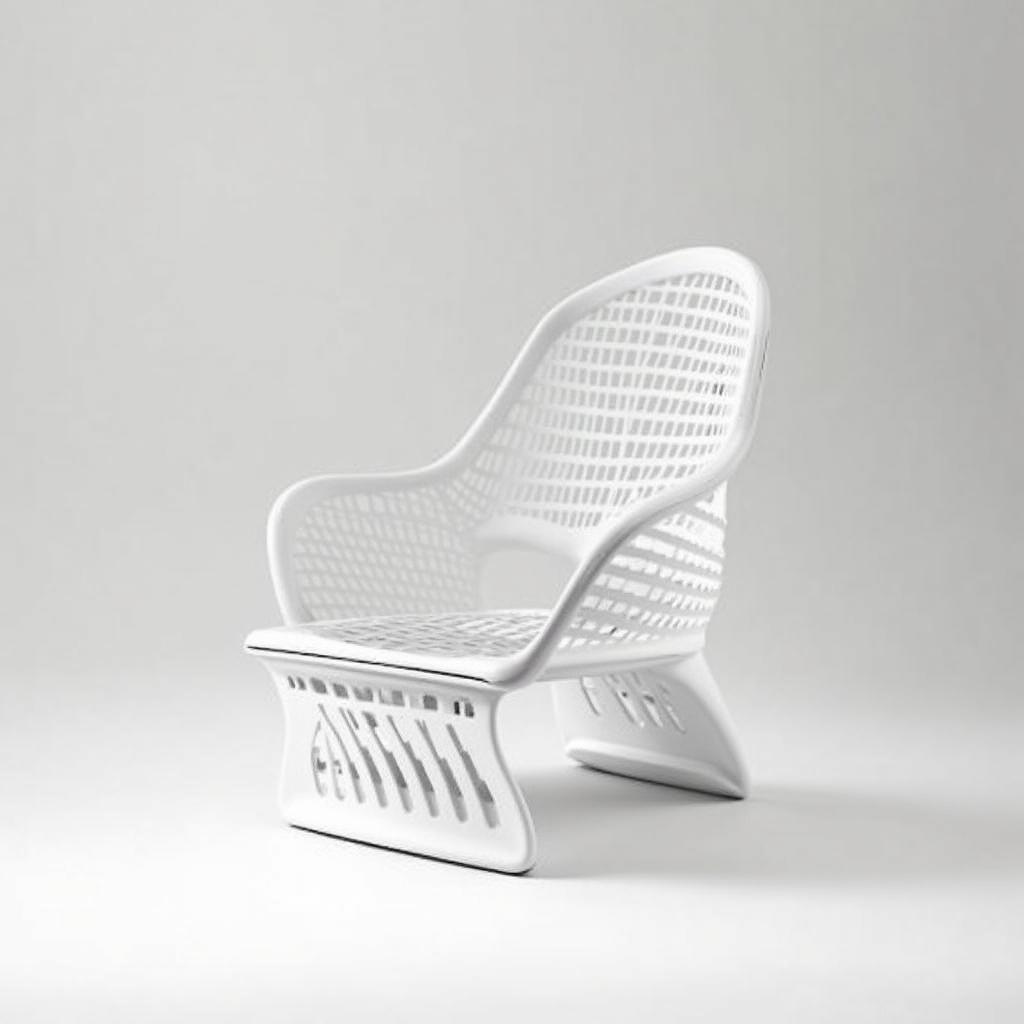} &

        \includegraphics[width=0.0875\textwidth]{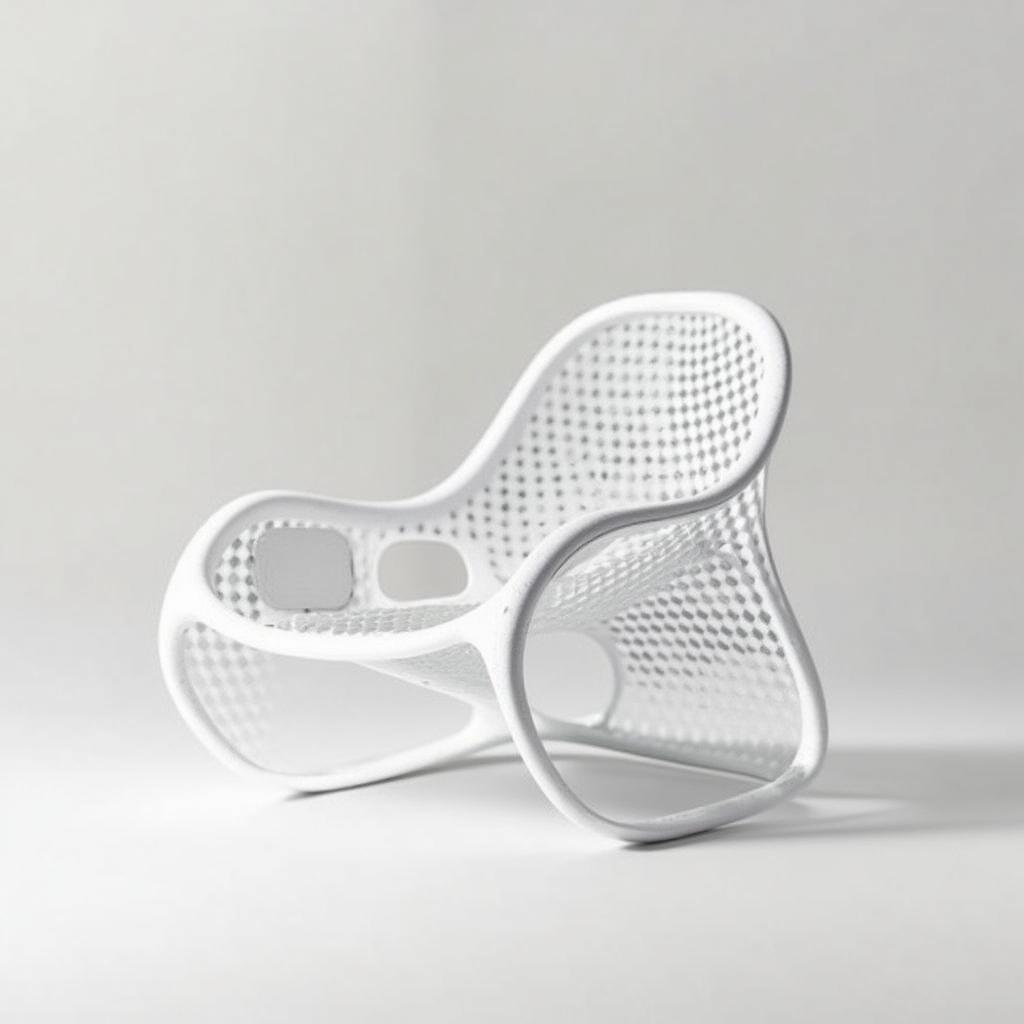}  \\

        \\[-0.075cm]

        Input & \multicolumn{4}{c}{Sampled Results}

    \end{tabular}
    }
    \vspace{-0.175cm}
    \caption{\textbf{Sampling From a Single Input.} 
    We present concepts generated by PiT using a single input part, which encourages greater variation across the generated results.
    }
    \vspace{-0.125cm}
    \label{fig:single_inputs_paper}
\end{figure}

Note that, as expected, the model exhibits greater variability when provided with fewer inputs, as there is more room for interpretation. This behavior is further highlighted in~\Cref{fig:single_inputs_paper}, where we present results generated from a single conditioning image. Here, the outputs vary significantly from one another, aligning with the intended use case in which an artist iteratively refines their target concept.

\begin{figure}
    \centering
    \setlength{\tabcolsep}{0.5pt}
    \addtolength{\belowcaptionskip}{-2.5pt}
    \renewcommand{\arraystretch}{0.5}
    {\small
    \begin{tabular}{c @{\hspace{0.2cm}} c c c c}

        \includegraphics[height=0.0875\textwidth]{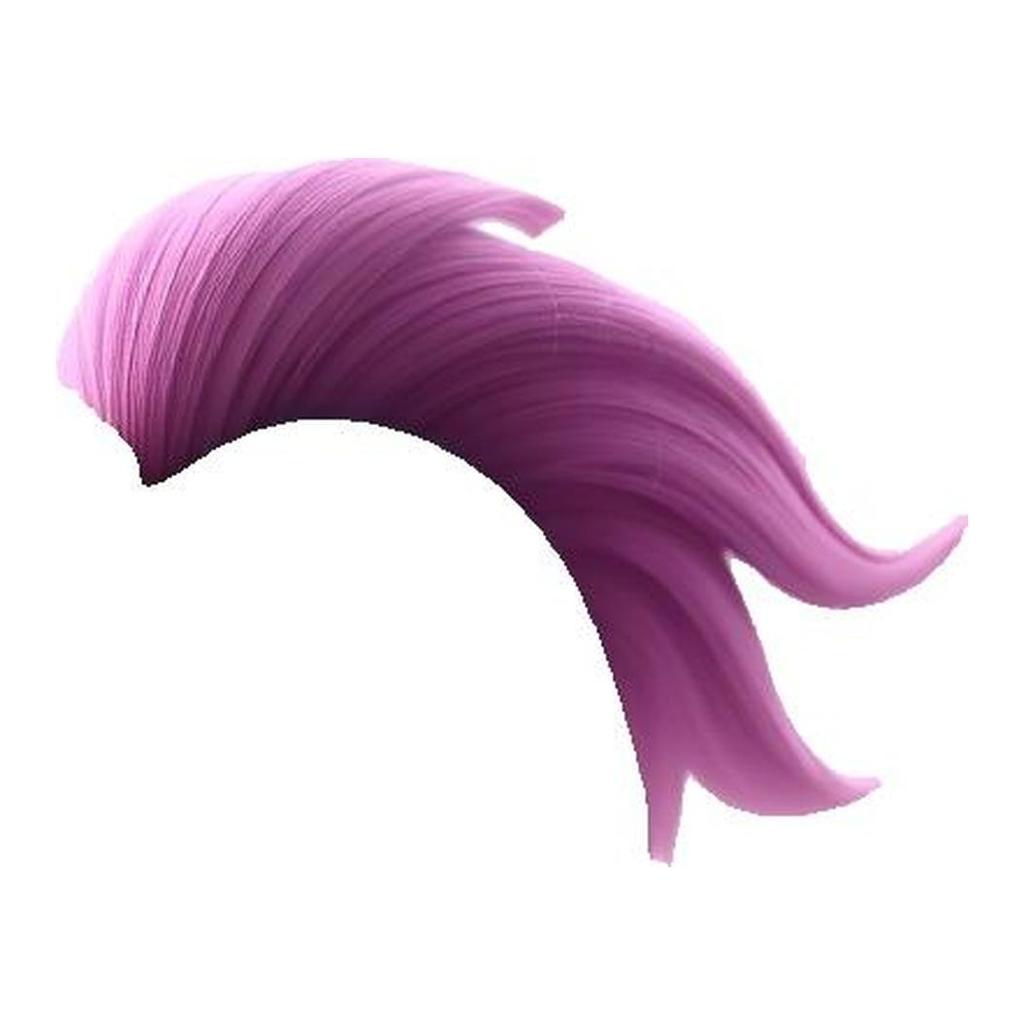} &
        \includegraphics[height=0.0875\textwidth]{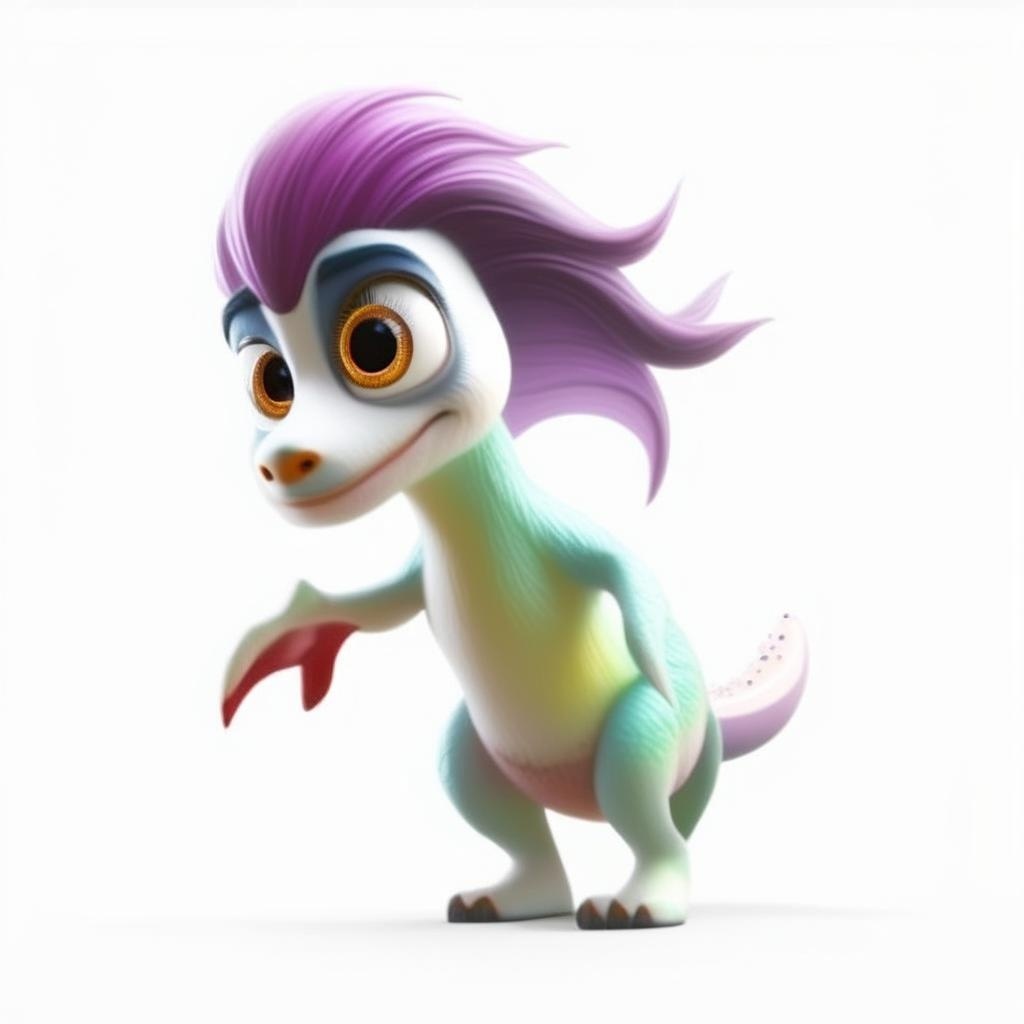} &
        \includegraphics[height=0.0875\textwidth]{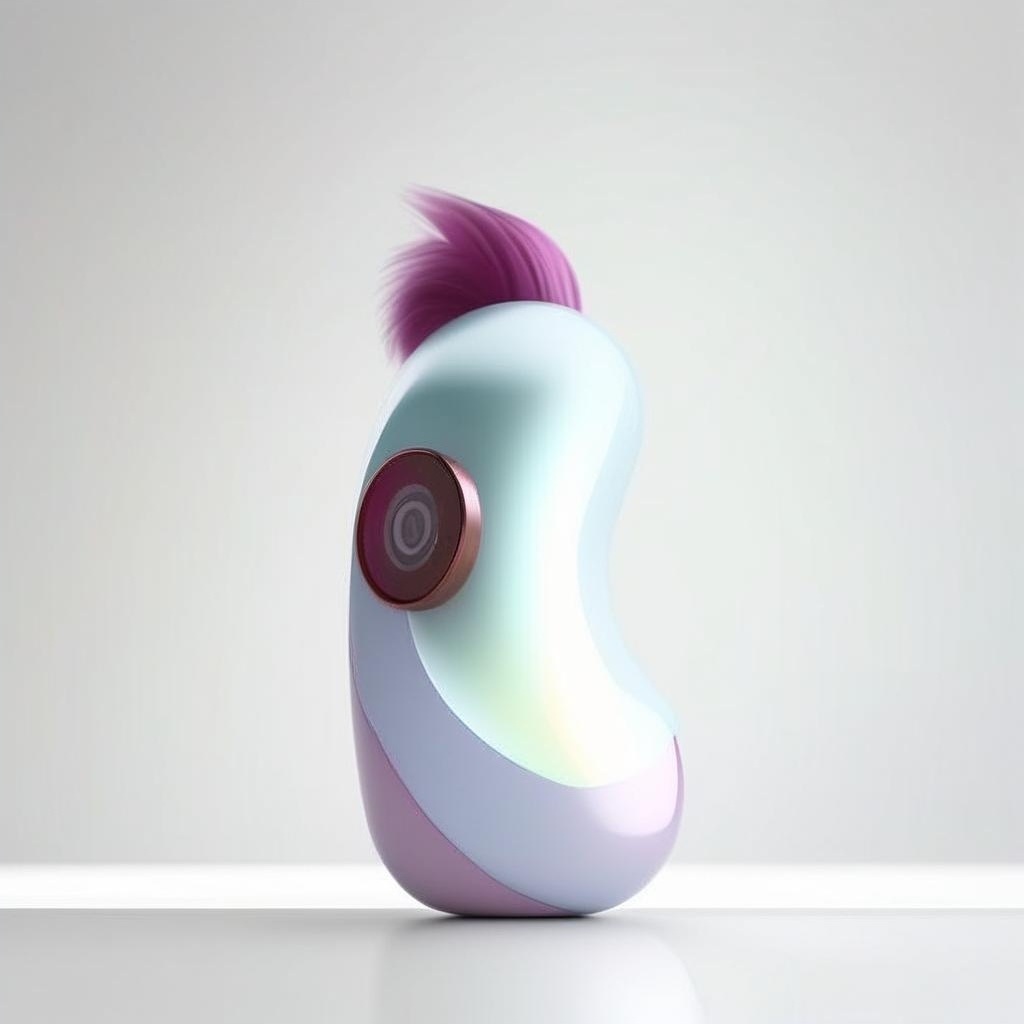} &
        \includegraphics[height=0.0875\textwidth]{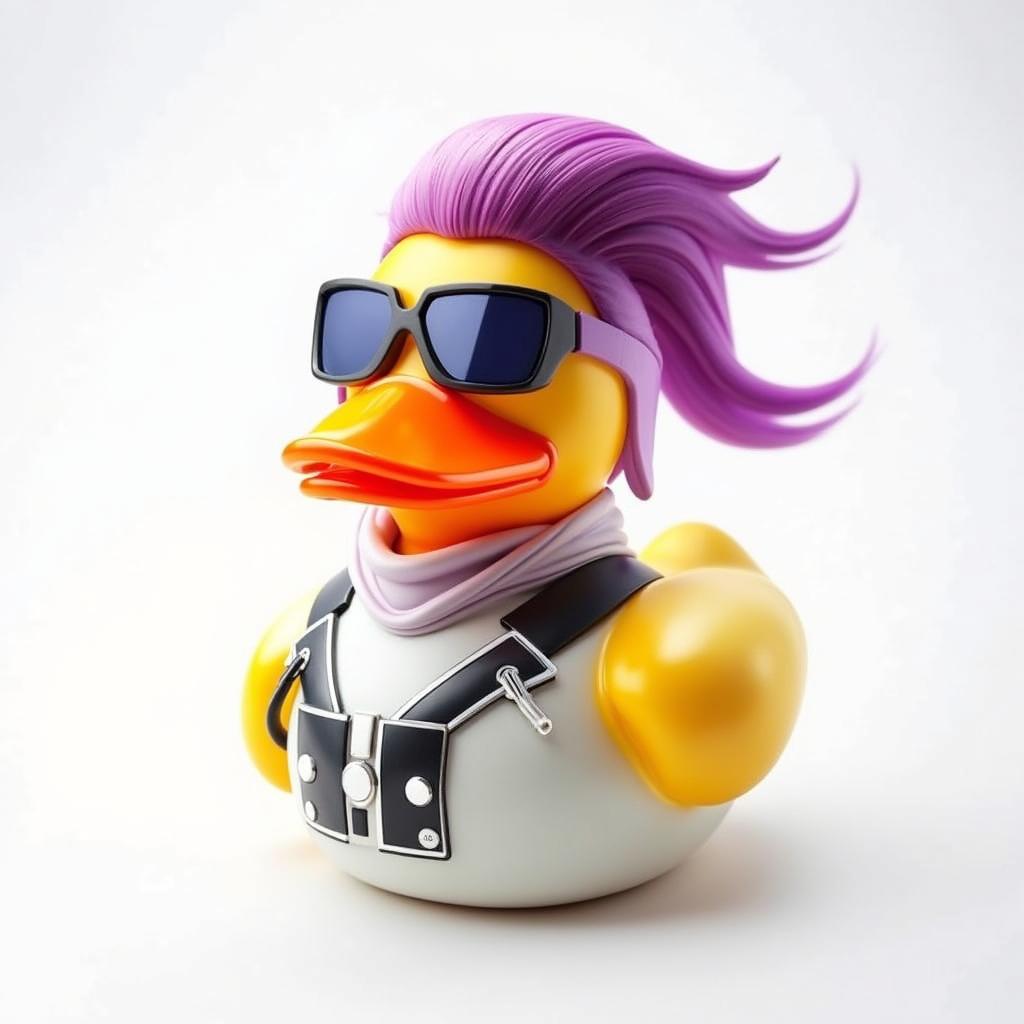} 
        & 

        \includegraphics[height=0.0875\textwidth]{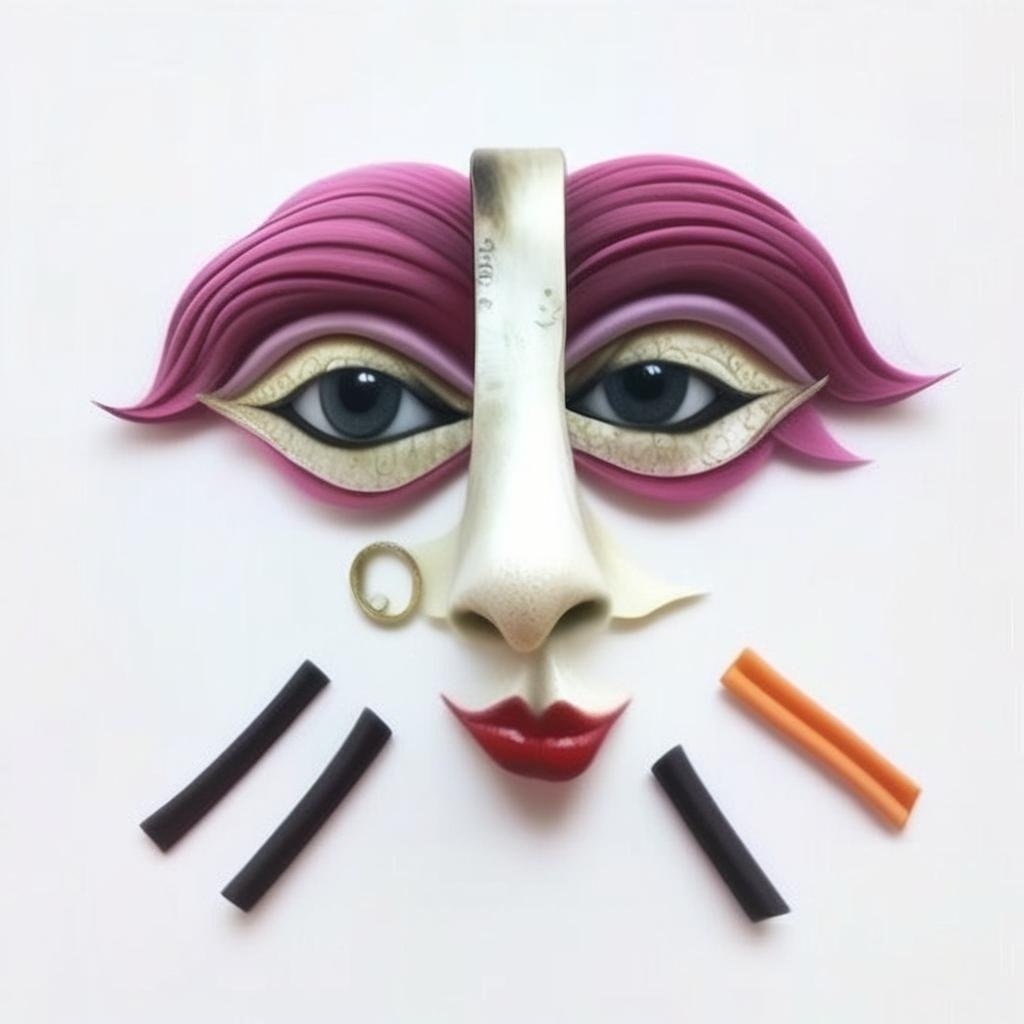} 
        \\

       \includegraphics[height=0.0875\textwidth]{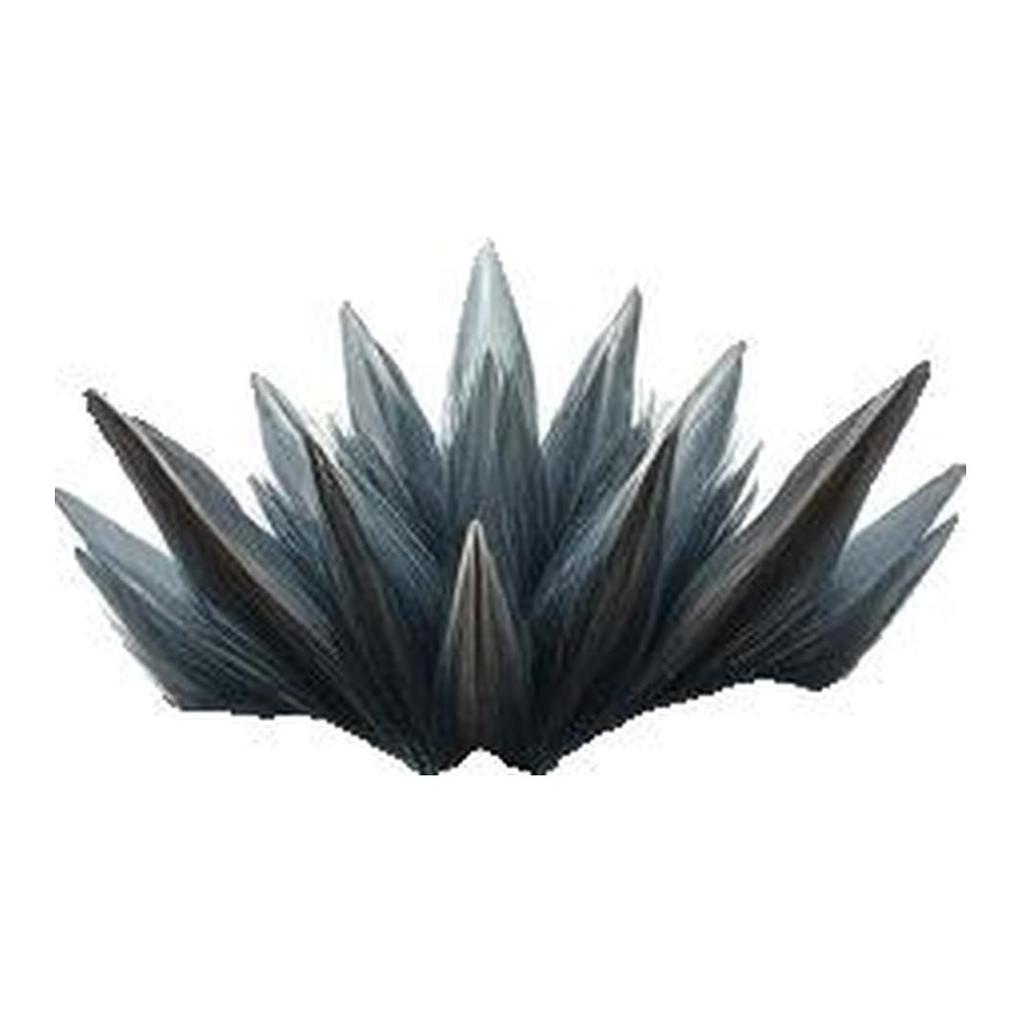} &
        \includegraphics[height=0.0875\textwidth]{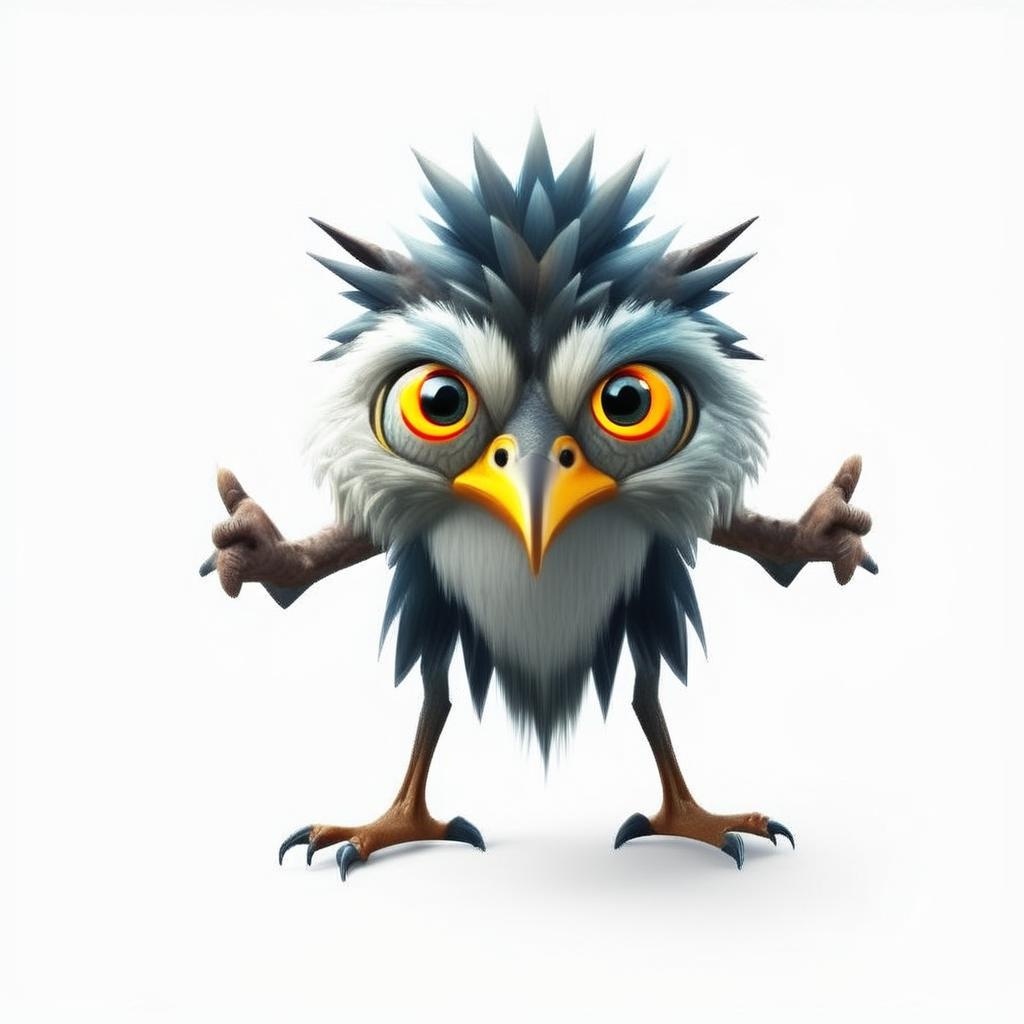} &
        \includegraphics[height=0.0875\textwidth]{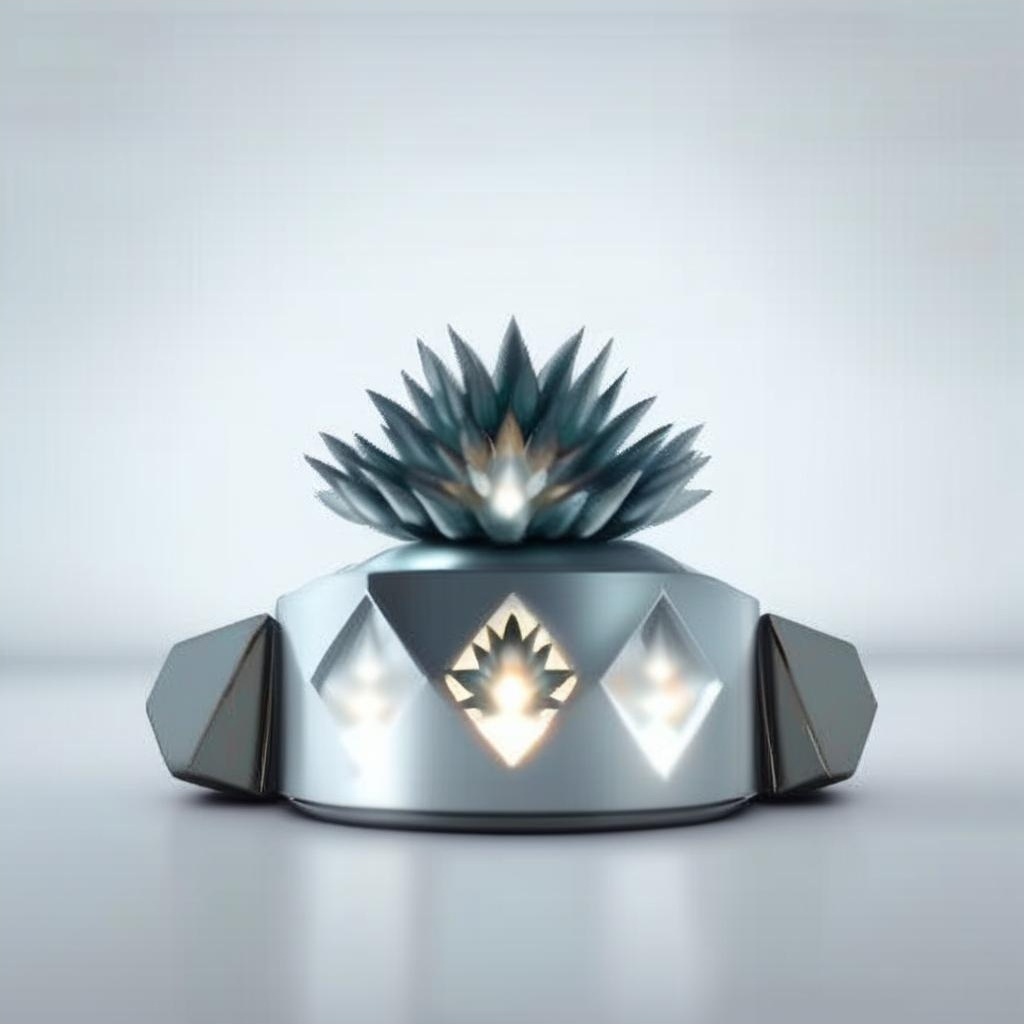} &

        \includegraphics[height=0.0875\textwidth]{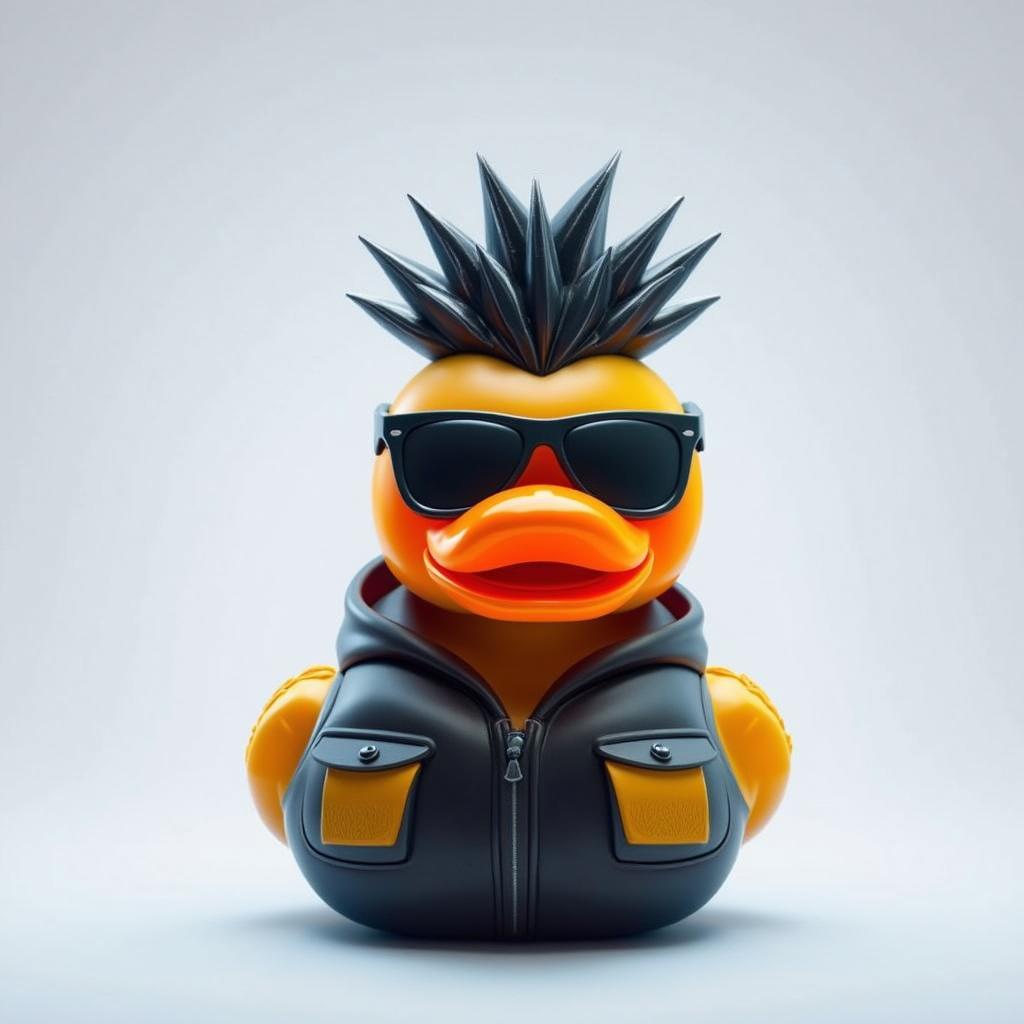} &
        \includegraphics[height=0.0875\textwidth]{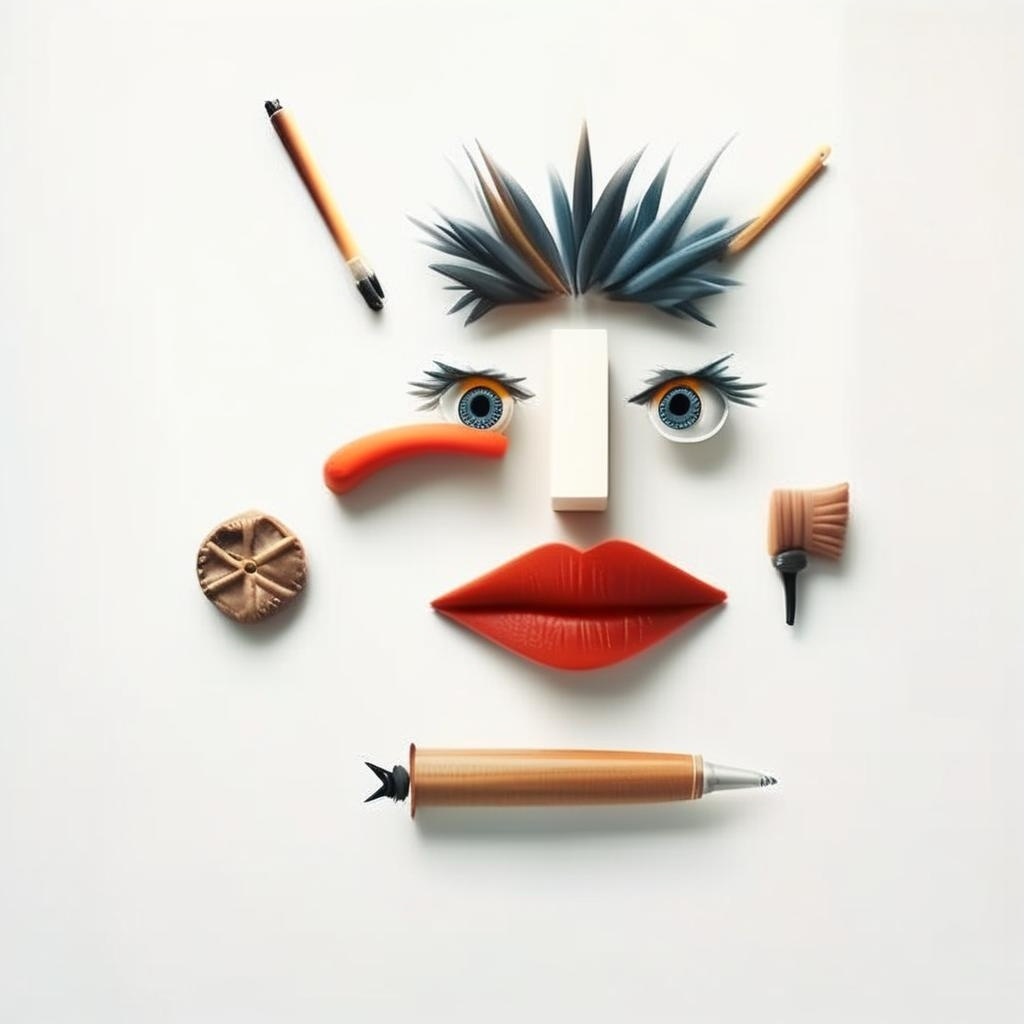} 
        \\

       \includegraphics[height=0.0875\textwidth]{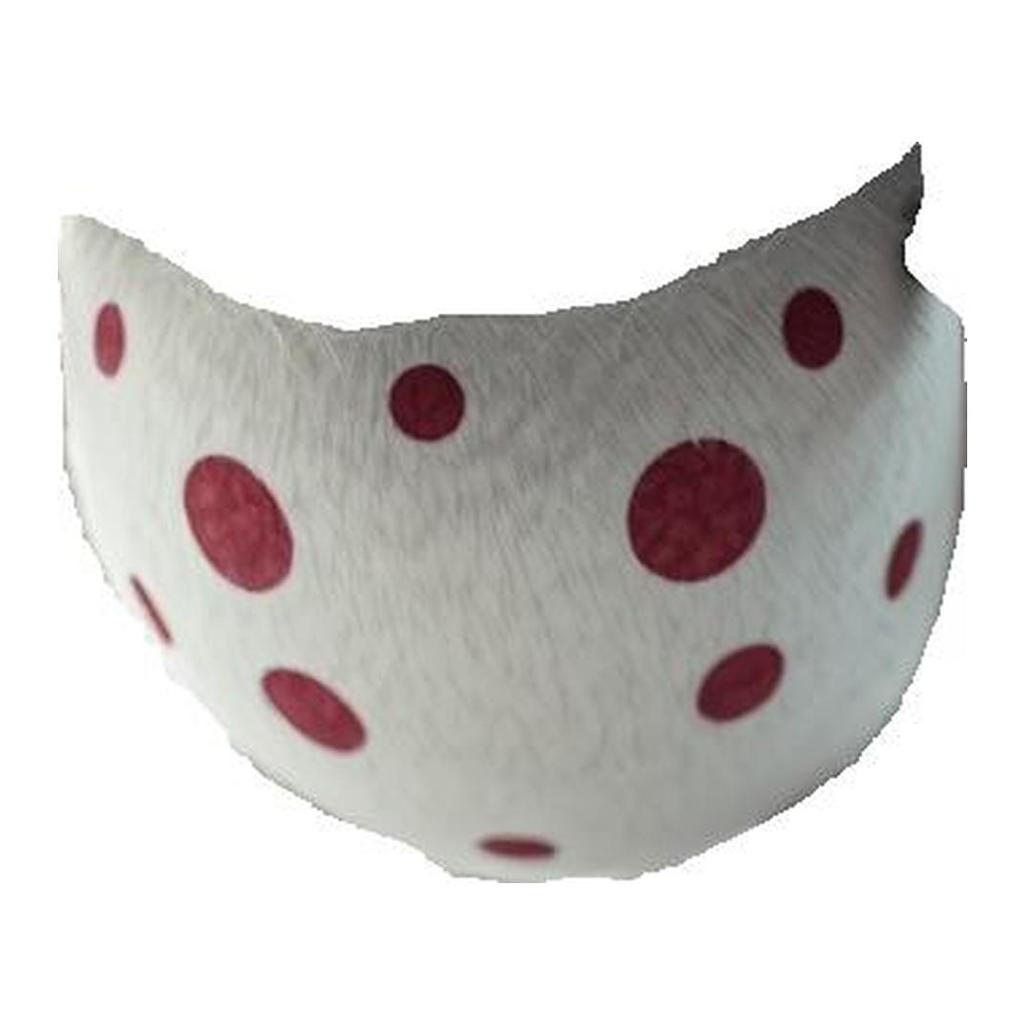} &
        \includegraphics[height=0.0875\textwidth]{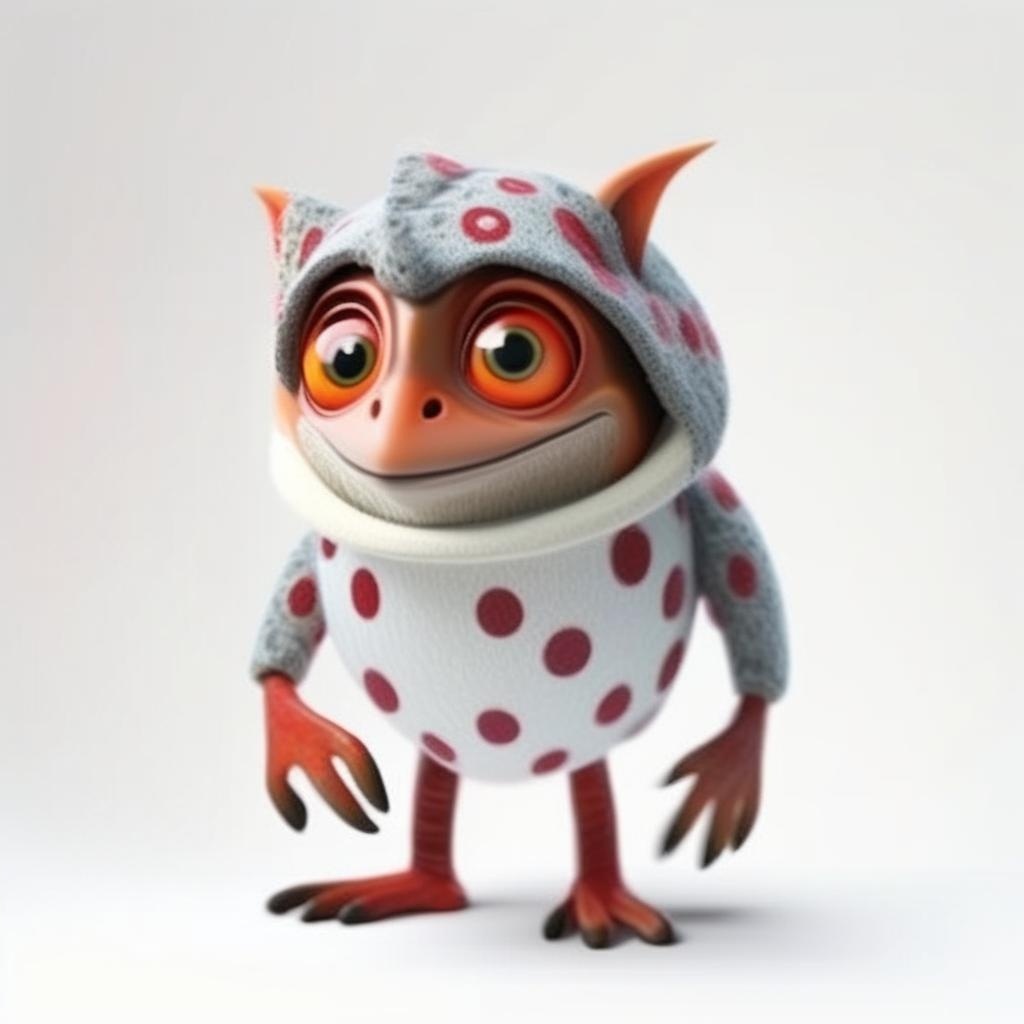} &
        \includegraphics[height=0.0875\textwidth]{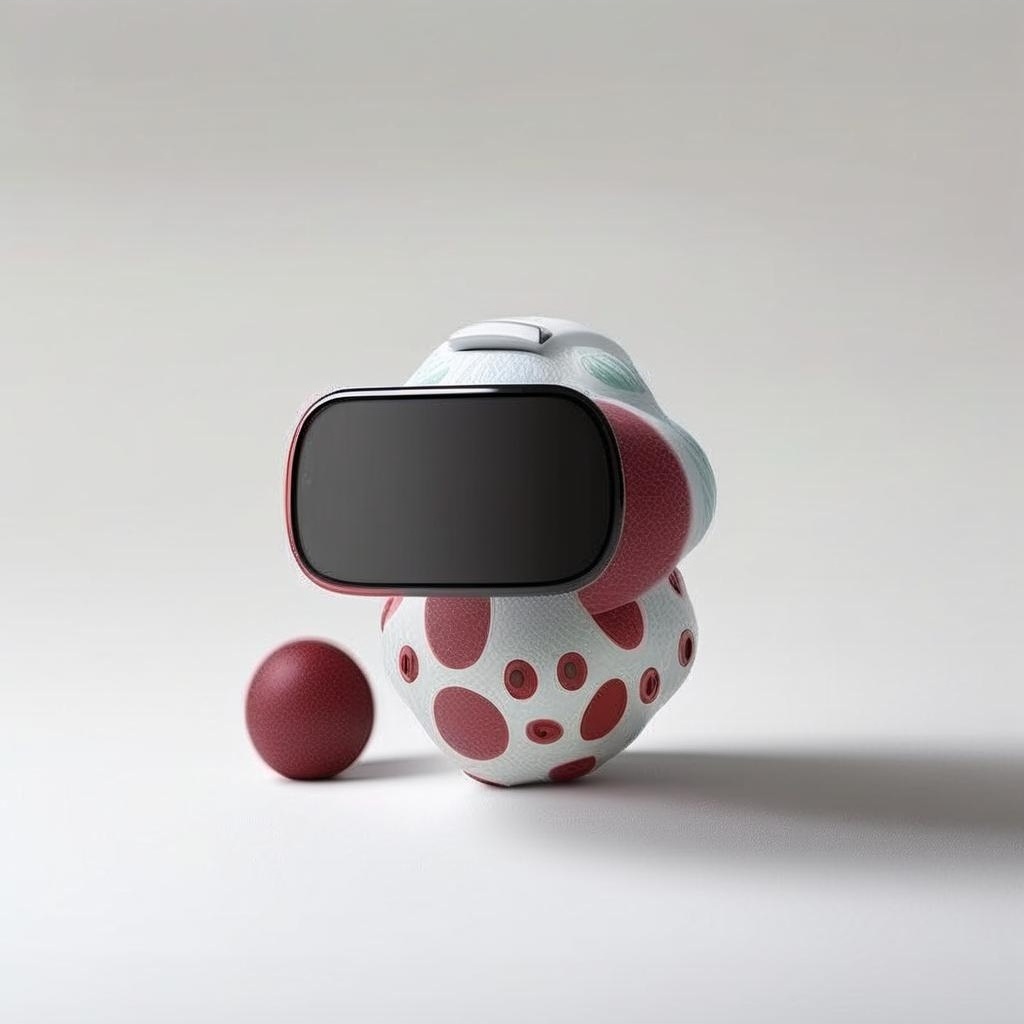} &
        \includegraphics[height=0.0875\textwidth]{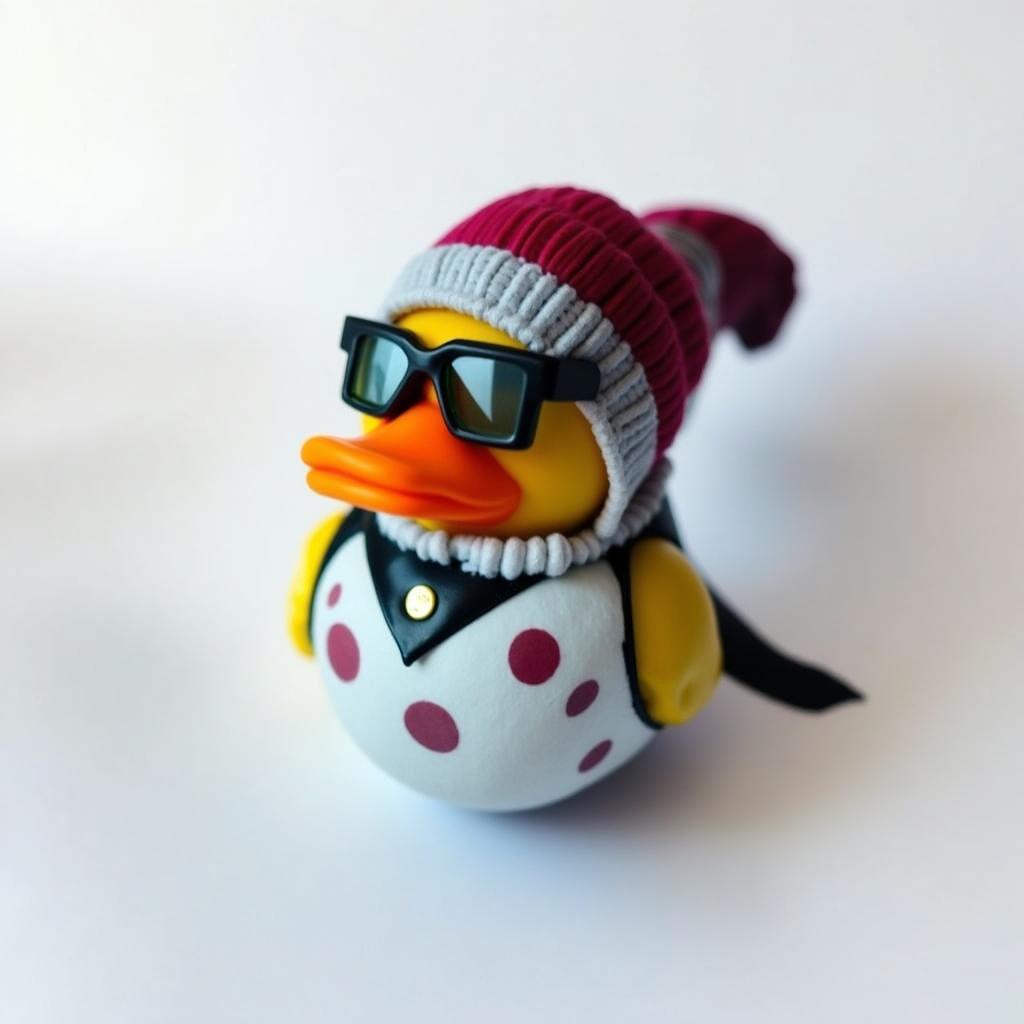} &
        \includegraphics[height=0.0875\textwidth]{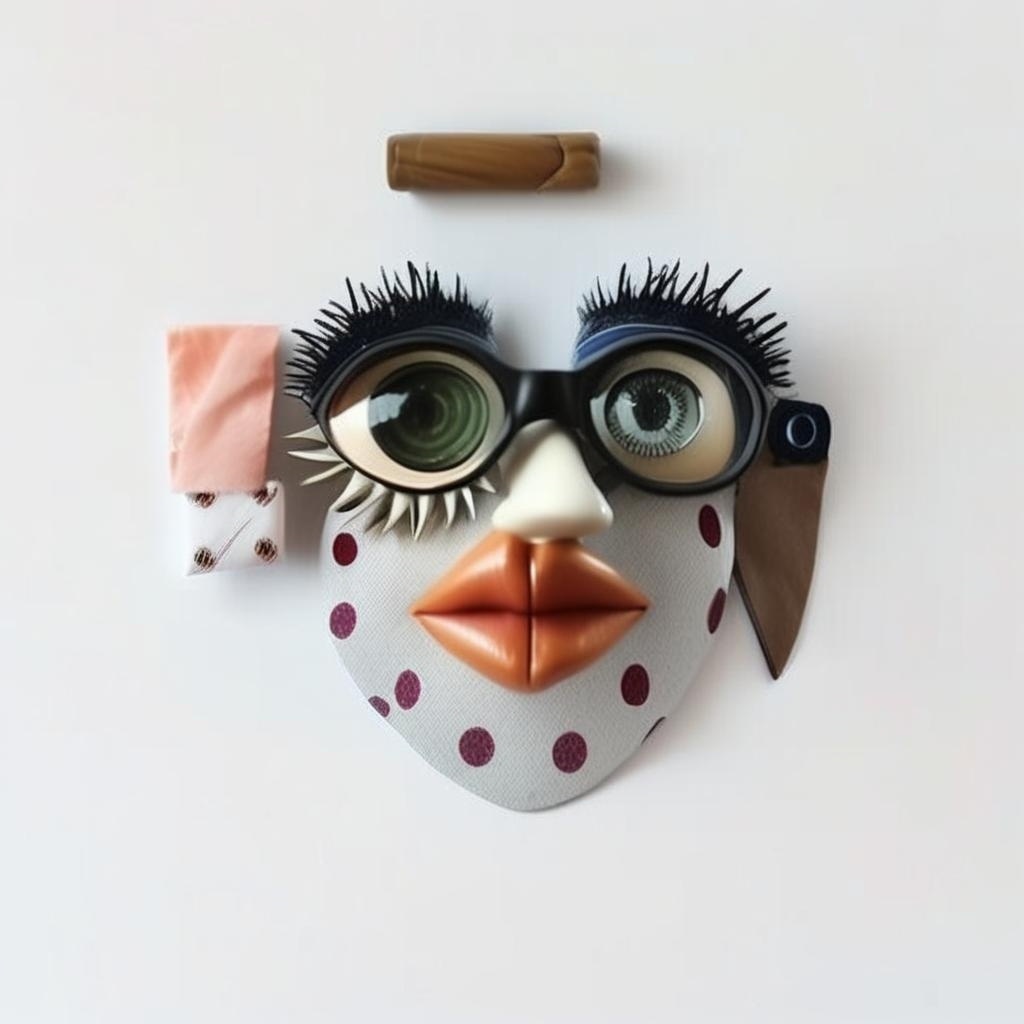} 
        \\

        Input & Character & Product & Ducks & Portraits

    \end{tabular}
    }
    \vspace{-0.2cm}
    \caption{\textbf{Sampling Across Different Priors} Given a single input part, we generate concepts across different learned IP-Prior models, highlighting how each model naturally interprets and adapts the part according to its learned distribution.
    }
    \vspace{-0.2cm}
    \label{fig:multiple_priors}
\end{figure}

To further highlight our model's ability to interpret the same elements in different ways based on the trained IP-Prior, we show results in~\Cref{fig:multiple_priors} using multiple different priors for the same set of inputs. 
As shown, each model correctly interprets the provided elements and seamlessly integrates missing components to produce a complete generation.
In the ``portrait'' domain, the model is trained on ``Found Object Portraits'' and in the ''Duck'' domain, the model is trained on customized rubber ducks. Notably, for the ``portrait'' domain, the model must interpret each given input as part of a facial structure. For instance, in the first row, the hair is interpreted as an eyebrow.

\begin{figure}
    \centering
    \setlength{\tabcolsep}{0.5pt}
    \renewcommand{\arraystretch}{0.8}
    \addtolength{\belowcaptionskip}{-5pt}
    {\footnotesize
    \begin{tabular}{c c c c c c}

        \raisebox{0.2cm}{\rotatebox{90}{\begin{tabular}{c} Cute $\rightarrow$ \\ Scary \end{tabular}}} &
        \includegraphics[width=0.085\textwidth]{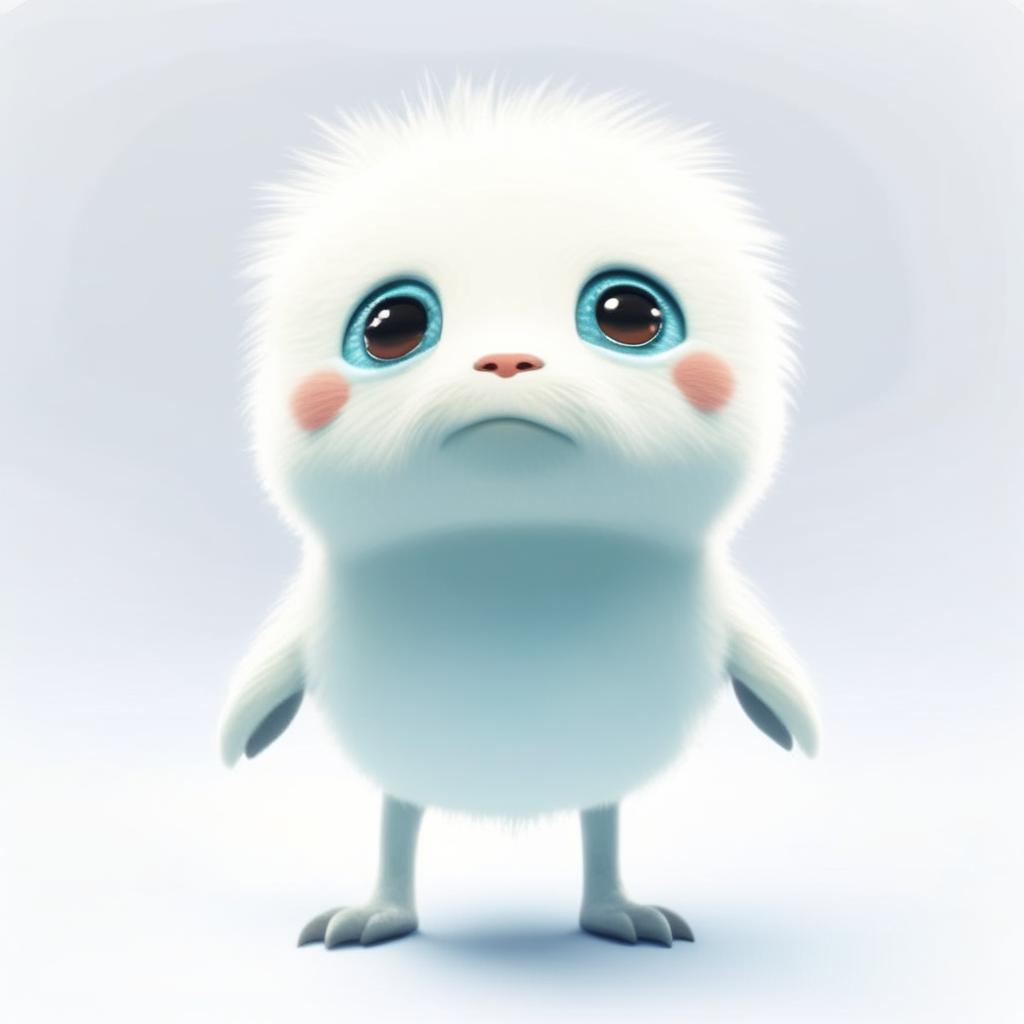} &
        \includegraphics[width=0.085\textwidth]{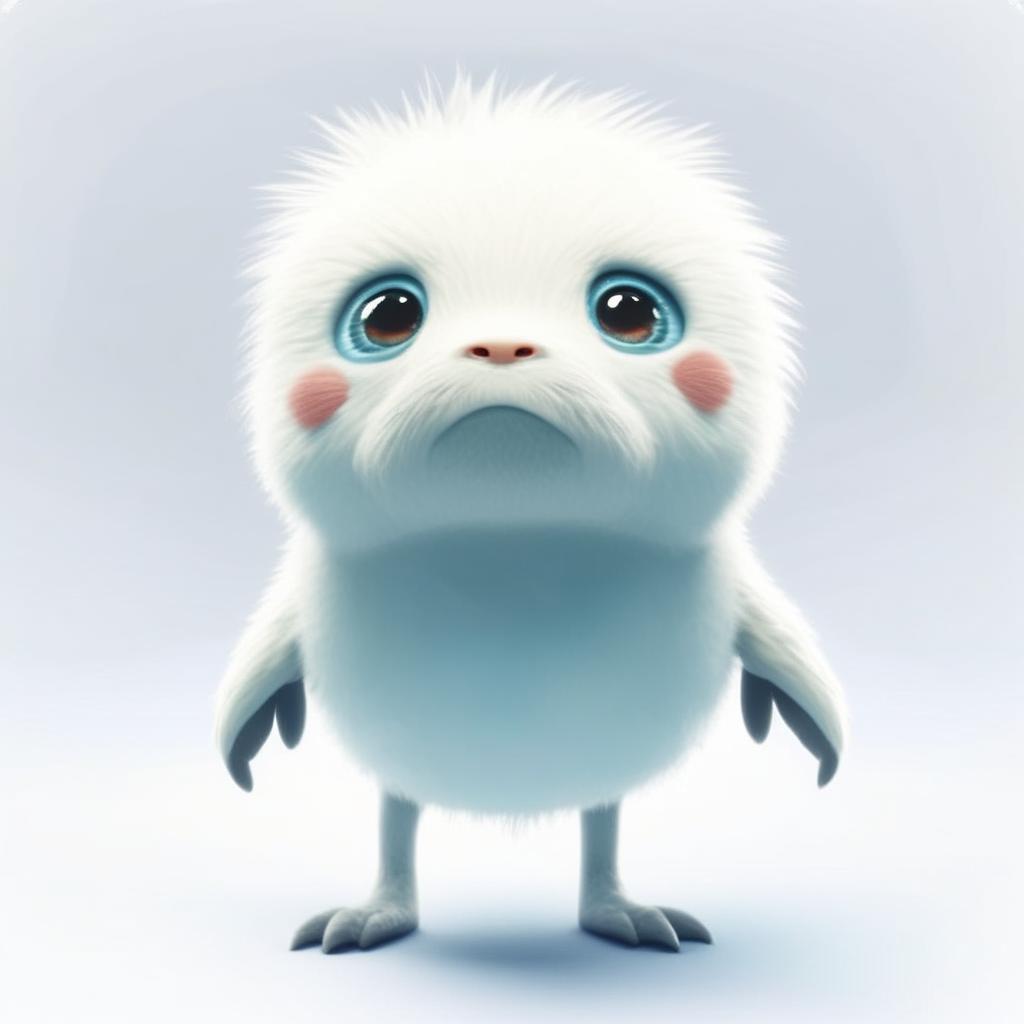} &  
        \includegraphics[width=0.085\textwidth]{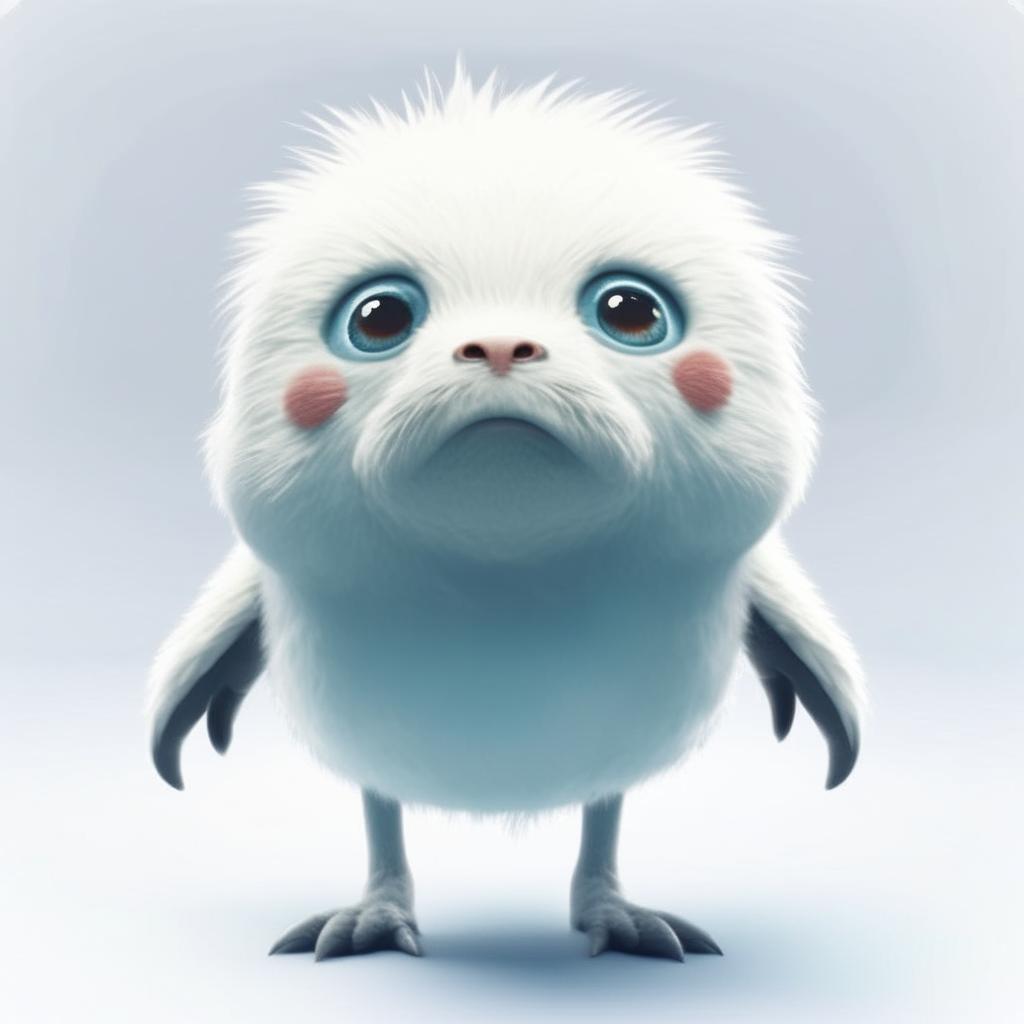} &
        \includegraphics[width=0.085\textwidth]{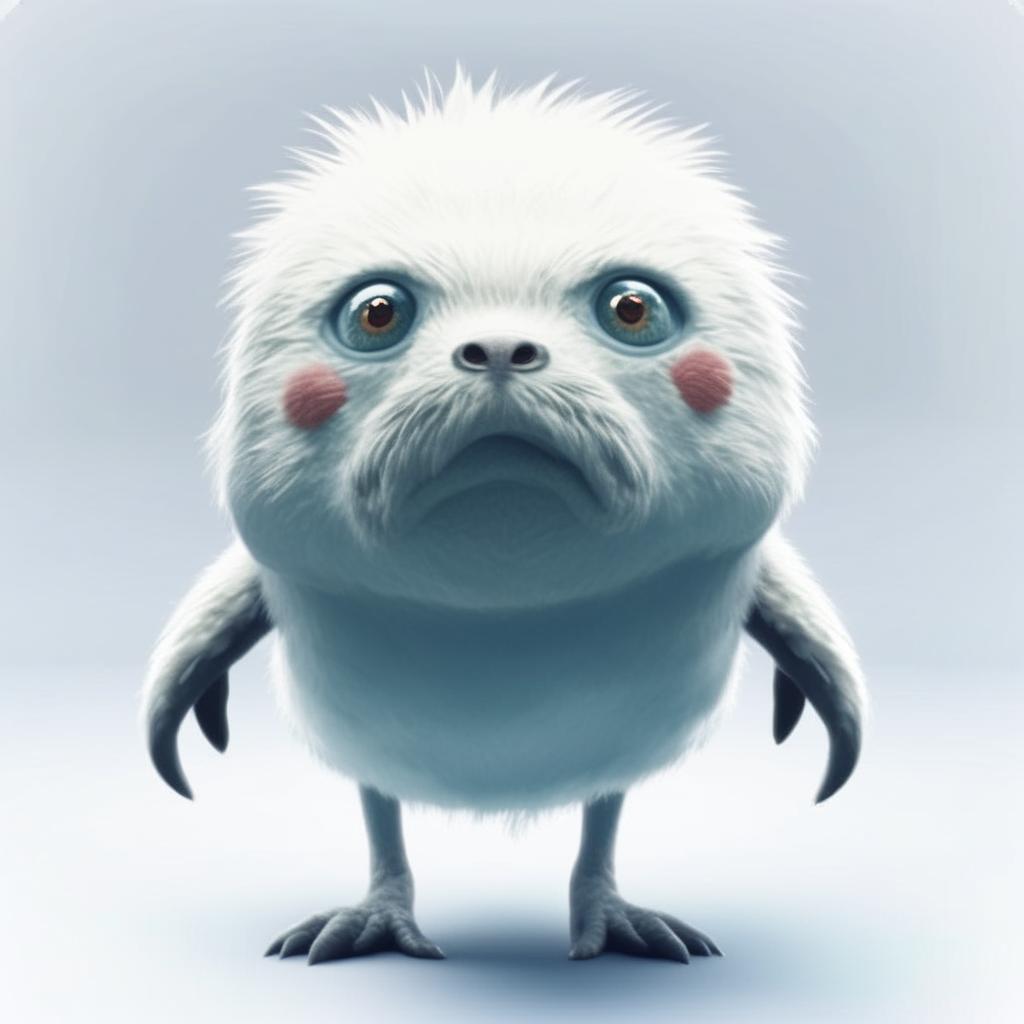} &
        \includegraphics[width=0.085\textwidth]{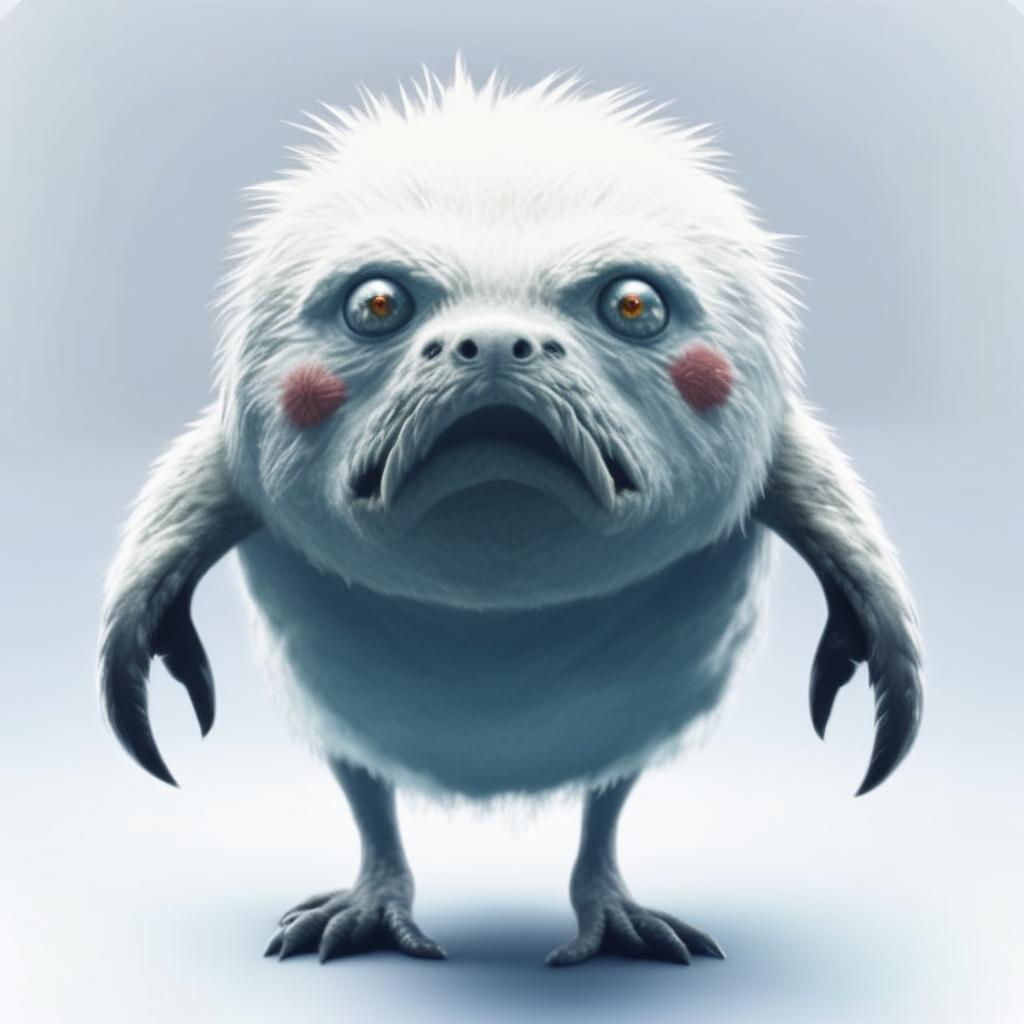} 
        \\ 

        \raisebox{0.2cm}{\rotatebox{90}{\begin{tabular}{c} Frosty $\rightarrow$ \\ Blazing \end{tabular}}} &
        \includegraphics[width=0.085\textwidth]{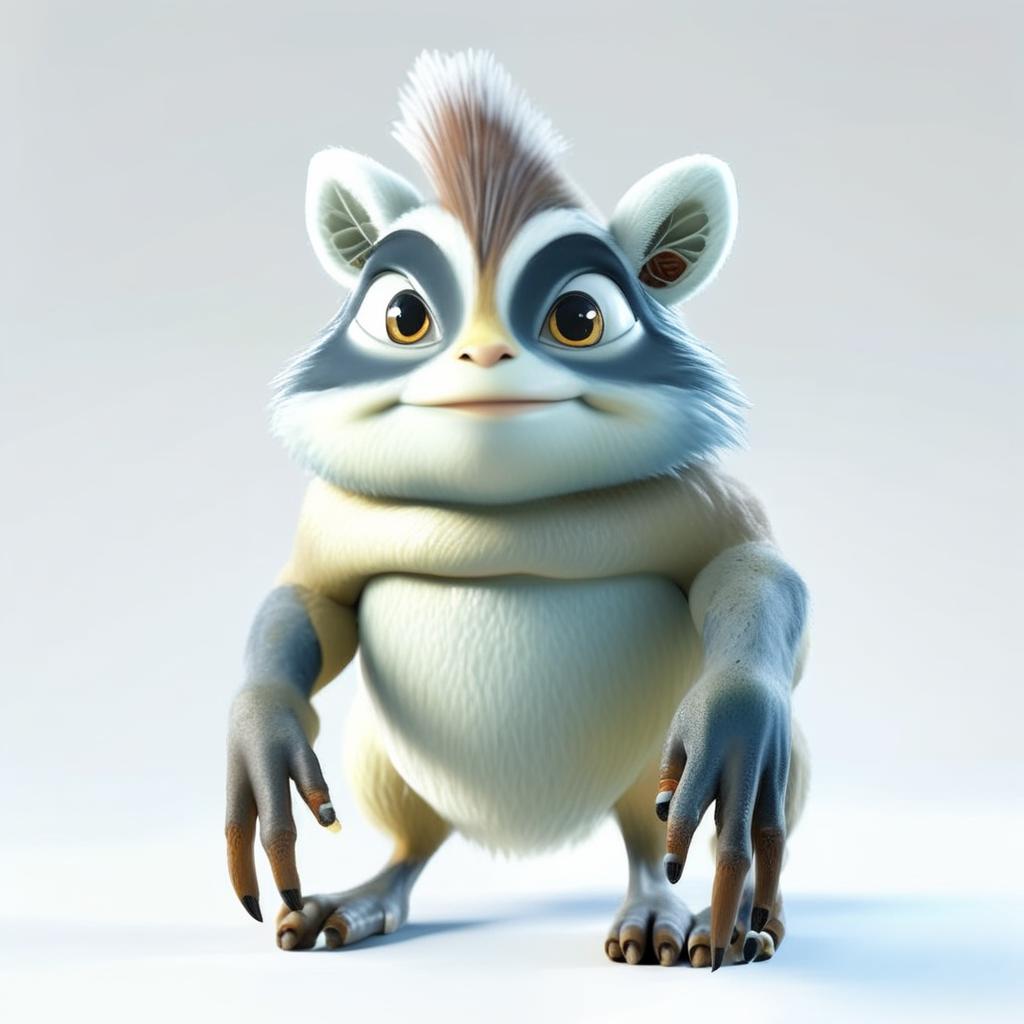} &
        \includegraphics[width=0.085\textwidth]{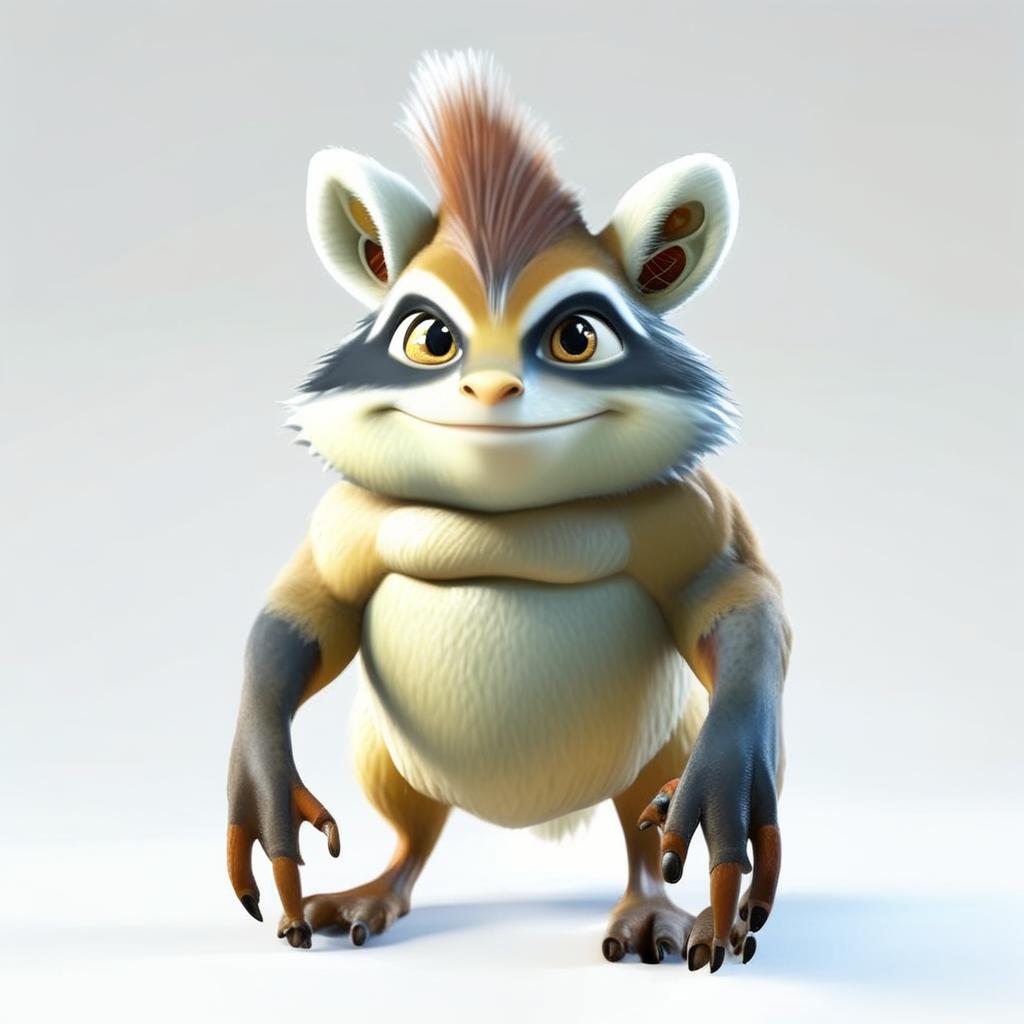} &  
        \includegraphics[width=0.085\textwidth]{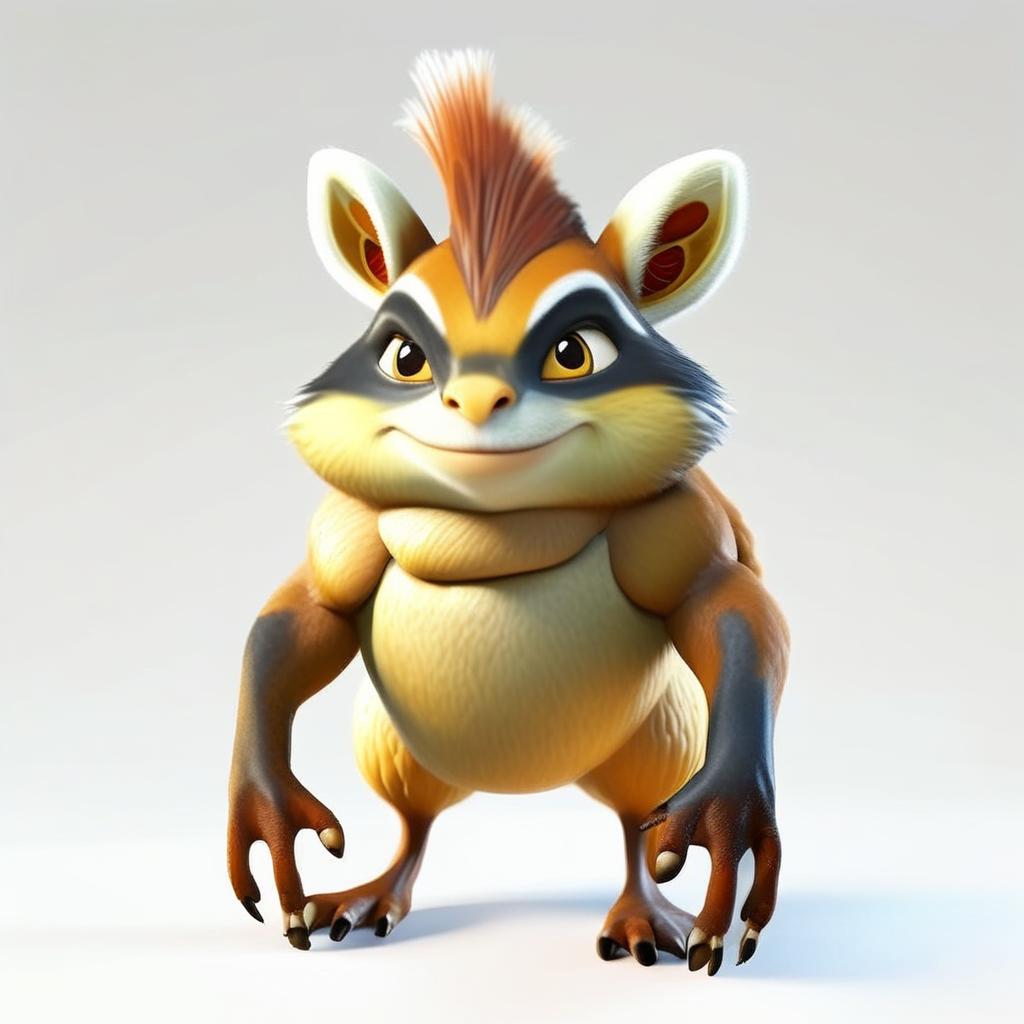} &
        \includegraphics[width=0.085\textwidth]{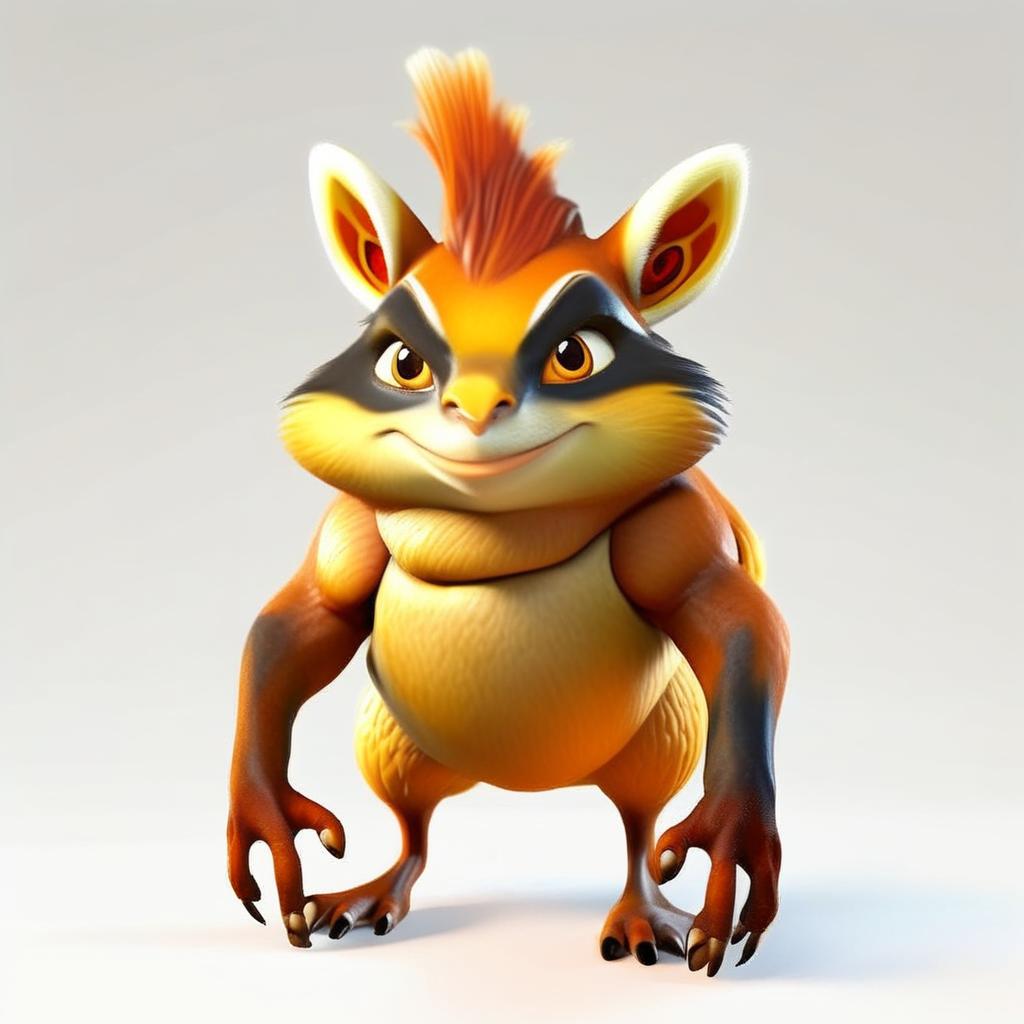} &
        \includegraphics[width=0.085\textwidth]{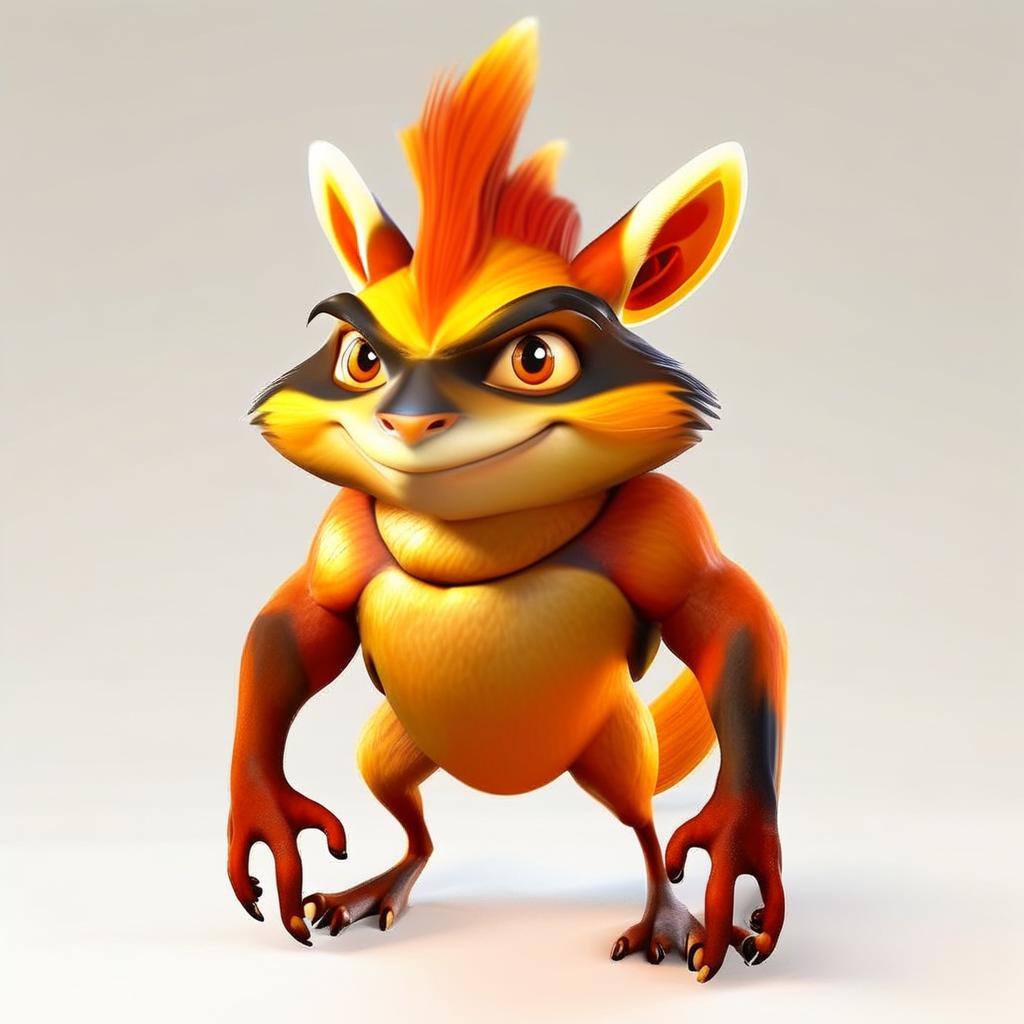}  \\

        \raisebox{0.075cm}{\rotatebox{90}{\begin{tabular}{c} Scrawny $\rightarrow$ \\ Muscular \end{tabular}}} &
        \includegraphics[width=0.085\textwidth]{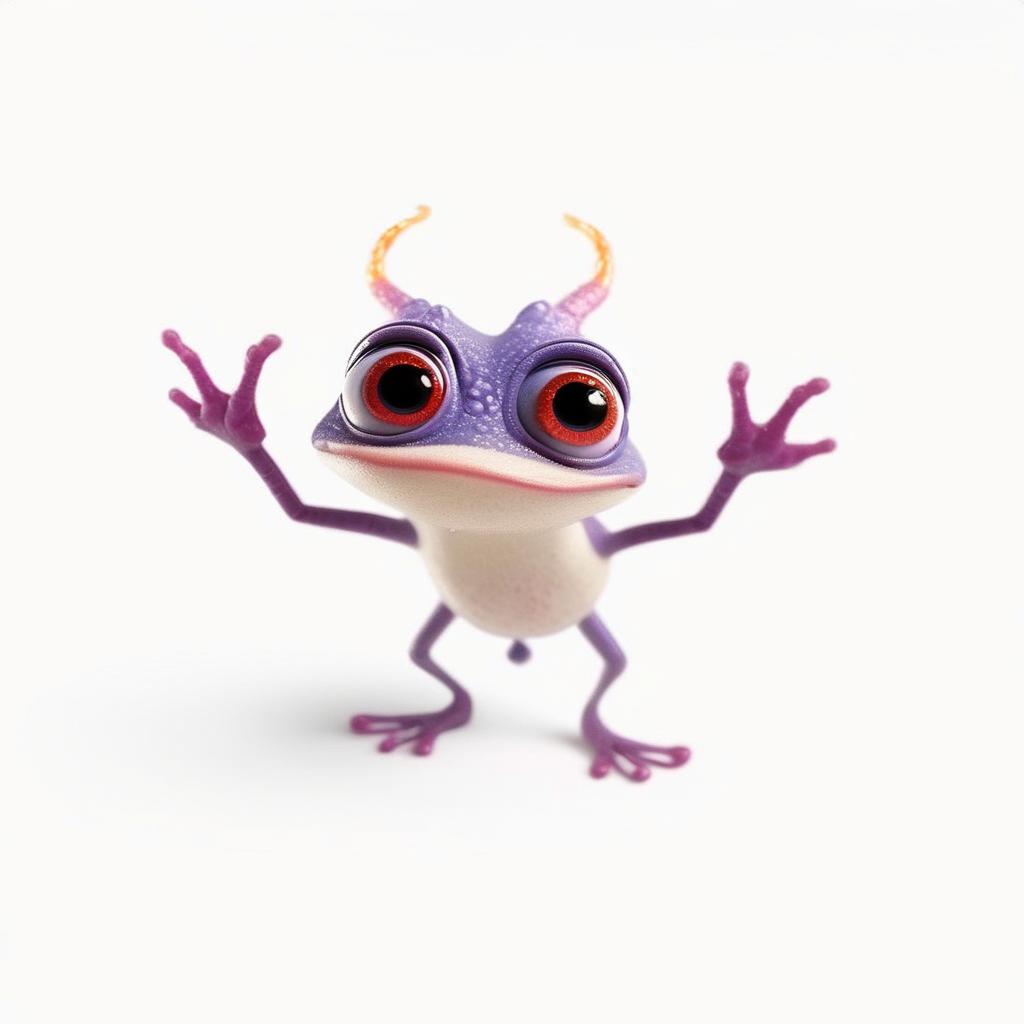} &
        \includegraphics[width=0.085\textwidth]{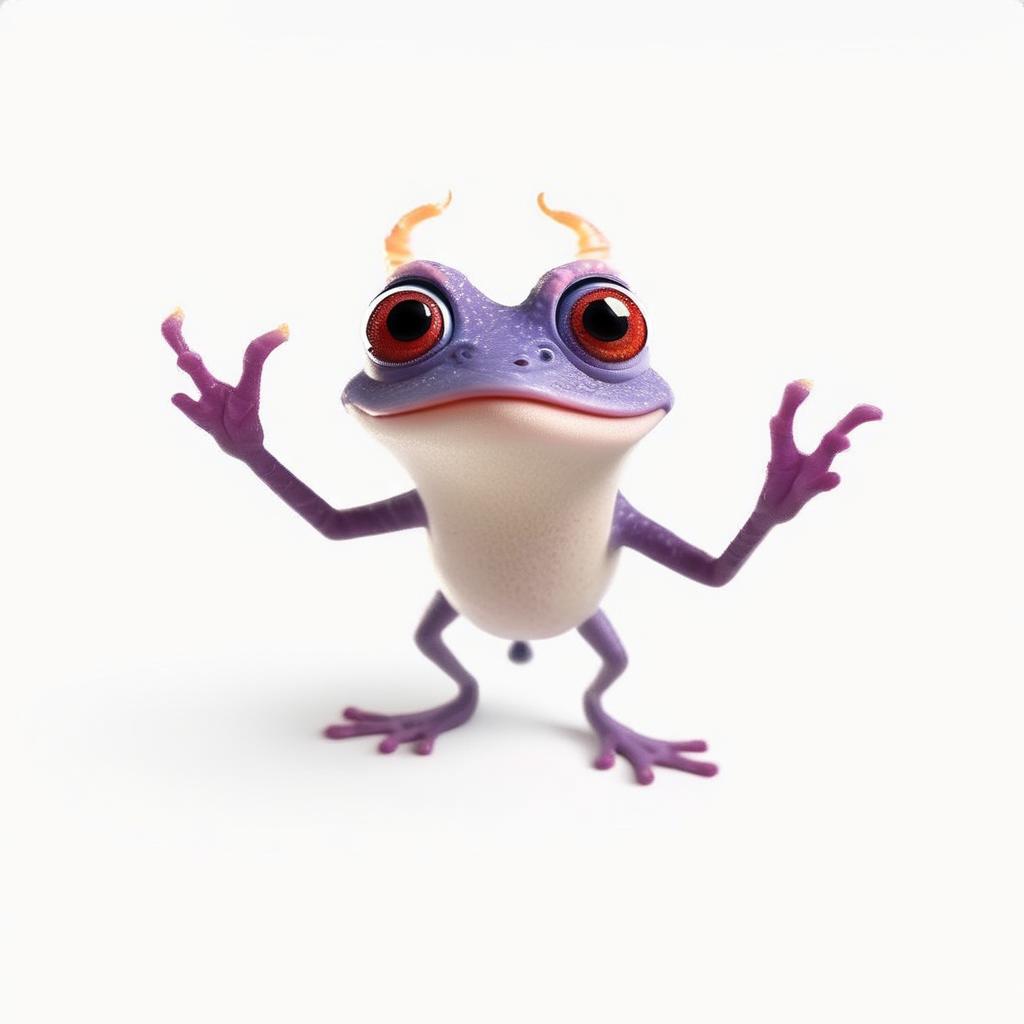} &  
        \includegraphics[width=0.085\textwidth]{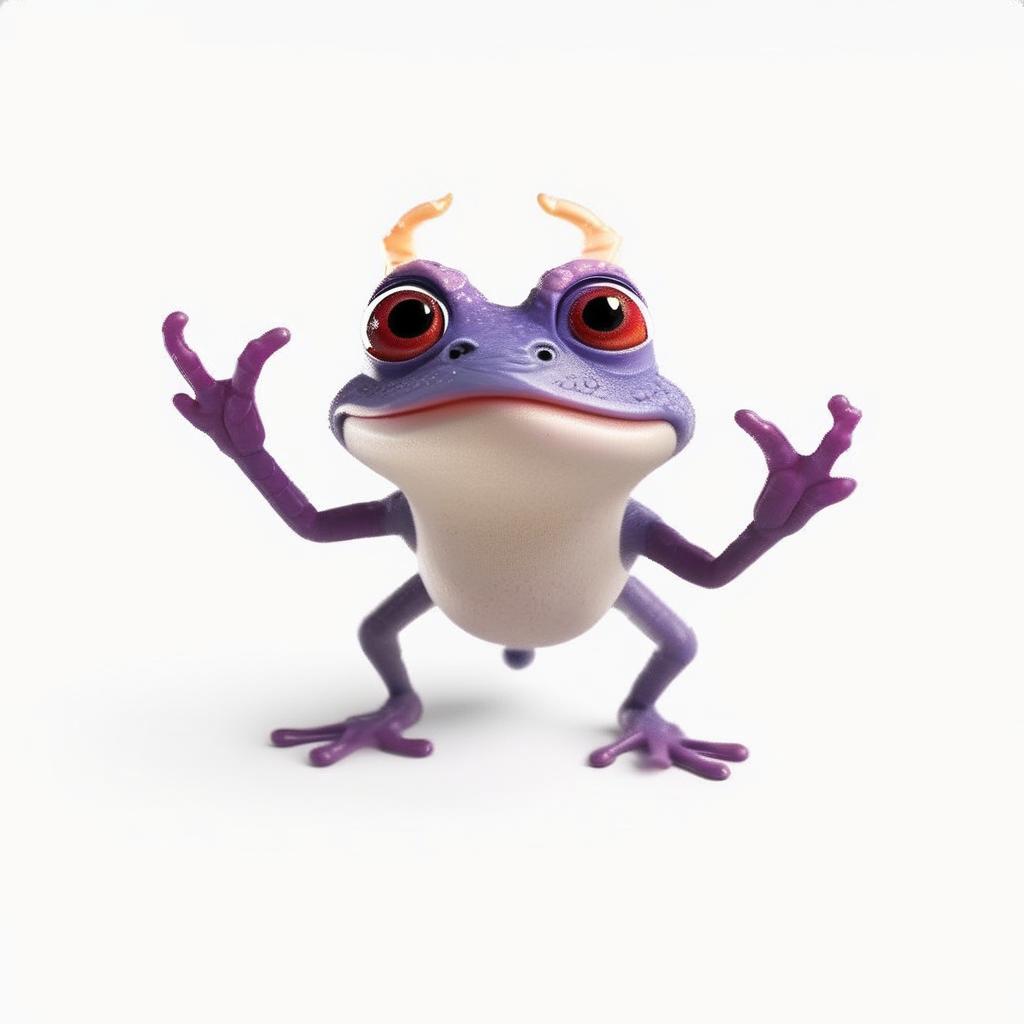} &
        \includegraphics[width=0.085\textwidth]{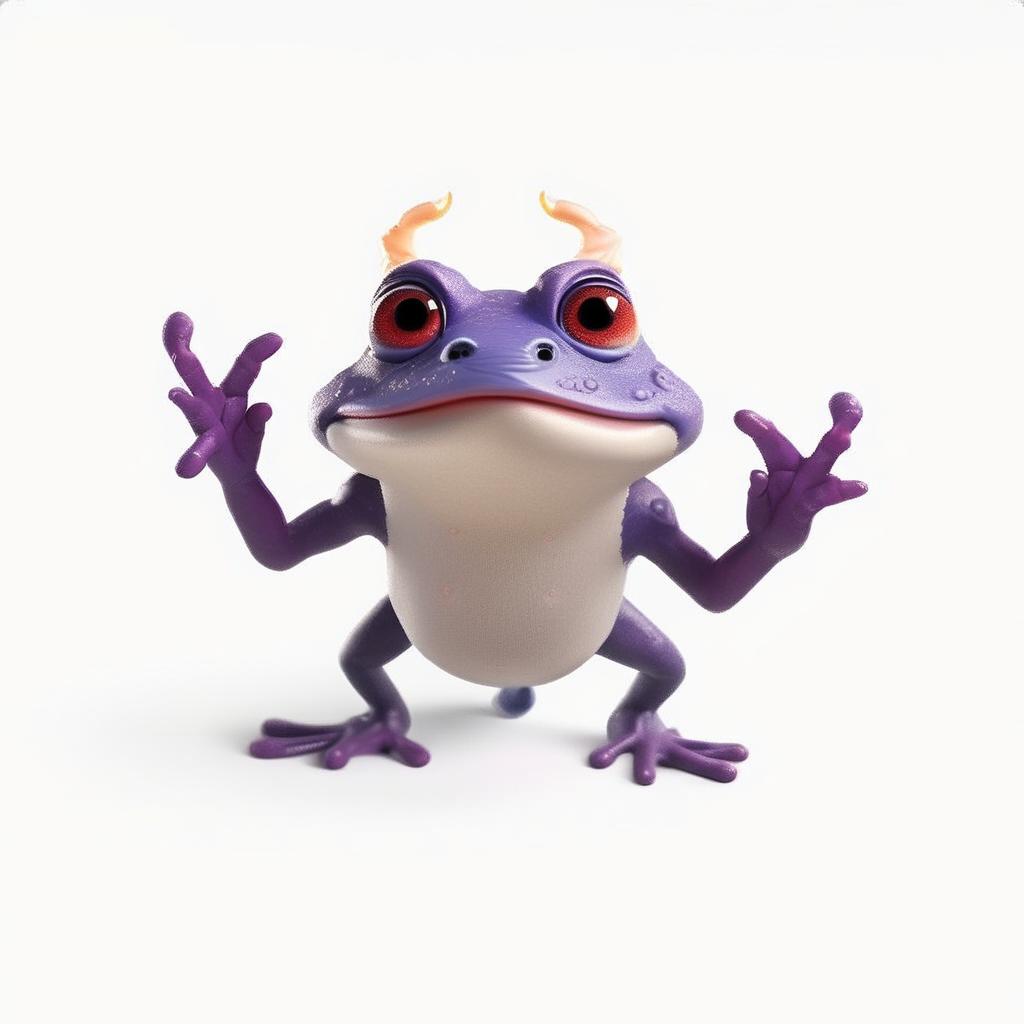} &
        \includegraphics[width=0.085\textwidth]{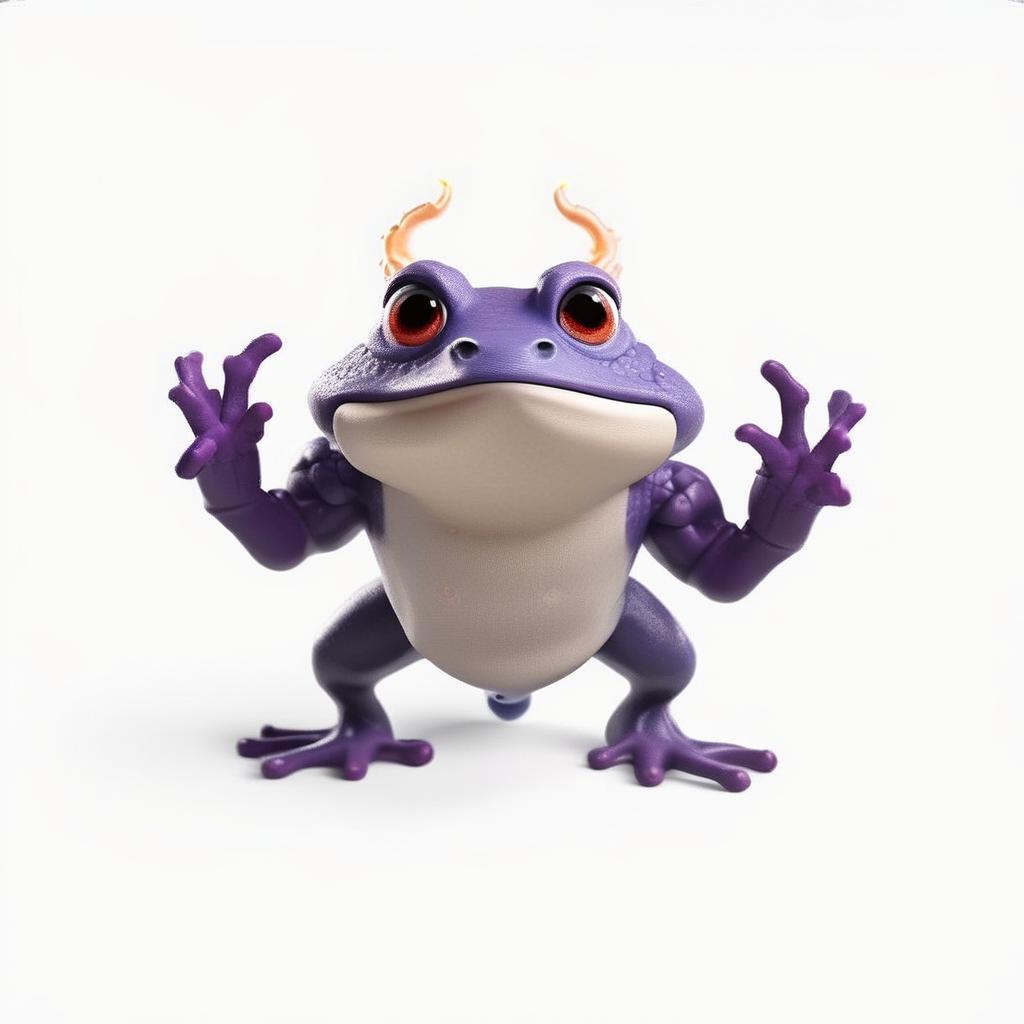}
        
    \end{tabular}
    }
    \vspace{-0.2cm}
    \caption{\textbf{Semantic Manipulations in $\mathcal{IP}+$ Space.} Using embeddings learned by our IP-Prior, we apply semantic manipulations to edit the concept before rendering the image with SDXL.}
    \label{fig:ip_edits}
\end{figure}

\paragraph{Semantic Manipulations in $\mathcal{IP}+$.}
One of the key advantages of working in embedding spaces for part-based generation is the ability to semantically edit and manipulate generated concepts, further enriching the ideation process. We demonstrate this in~\Cref{fig:ip_edits}, where we apply various semantic transformations within the $\mathcal{IP}^+$ space.
Each edit direction is found by generating $50$ samples corresponding to a pair of contrasting words (e.g., ``cute'' and ``scary''), embedding them into our space, and computing the vector direction between the mean embeddings of each set. This simple yet effective approach enables meaningful semantic edits, further highlighting both the potential of working in the $\mathcal{IP}^+$ space and the broader benefit of working with embeddings in PiT.

\begin{figure}
    \centering
    \setlength{\tabcolsep}{1pt}
    \addtolength{\belowcaptionskip}{-2.5pt}
    \renewcommand{\arraystretch}{0.5}
    {\small
    \begin{tabular}{ccccc}

        Input & PiT & IP-A (0.2) & IP-A (0.4) & IP-A (0.6) \\ \\[-0.1cm]
    
        {\includegraphics[width=0.19\columnwidth]{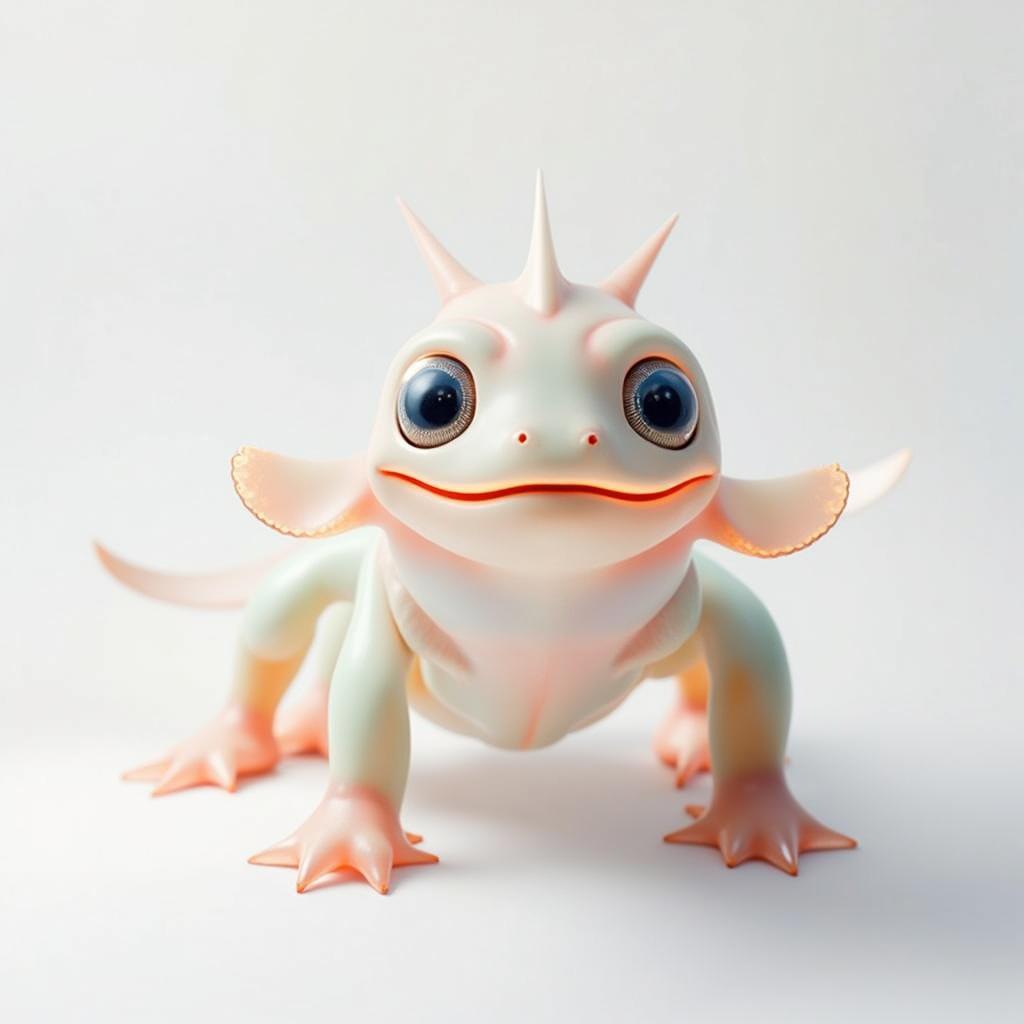}} &
        \includegraphics[width=0.19\columnwidth]{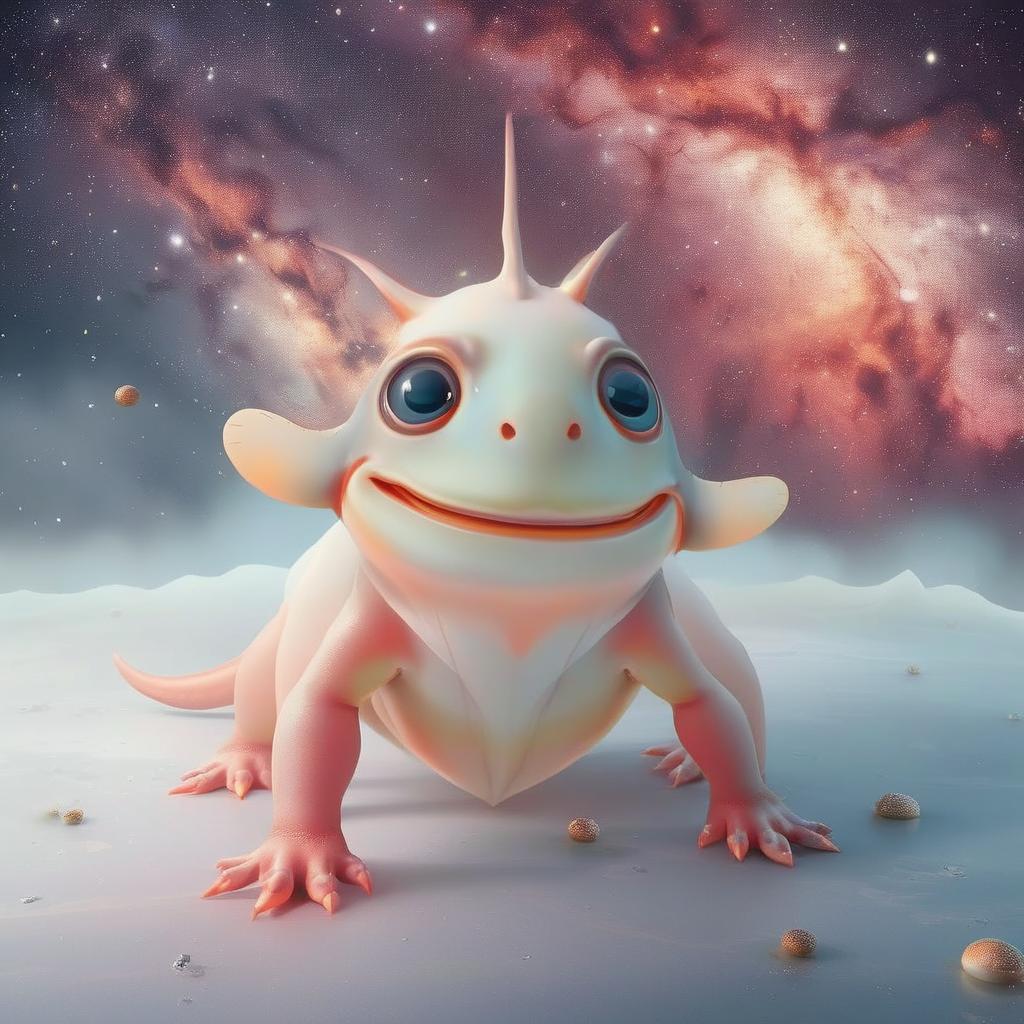} &
        \includegraphics[width=0.19\columnwidth]{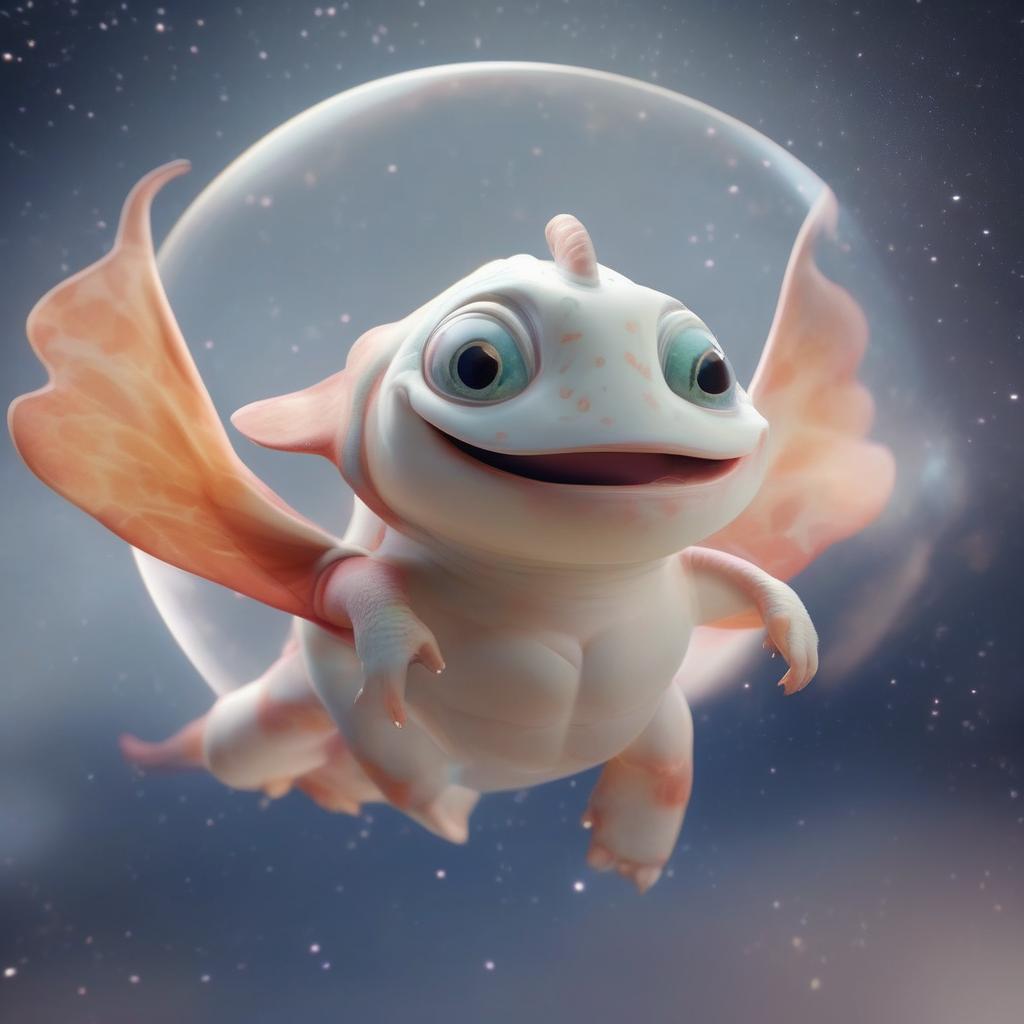} &  
        \includegraphics[width=0.19\columnwidth]{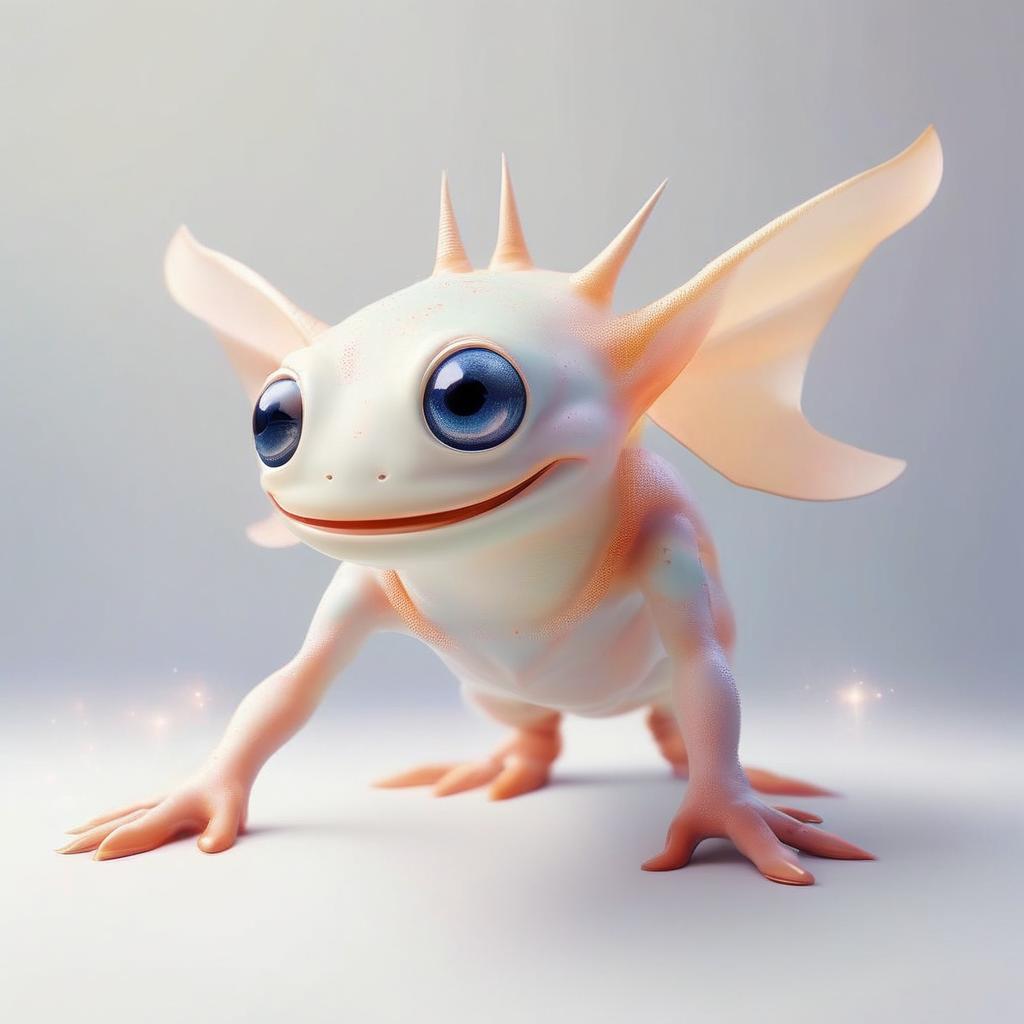} &
        \includegraphics[width=0.19\columnwidth]{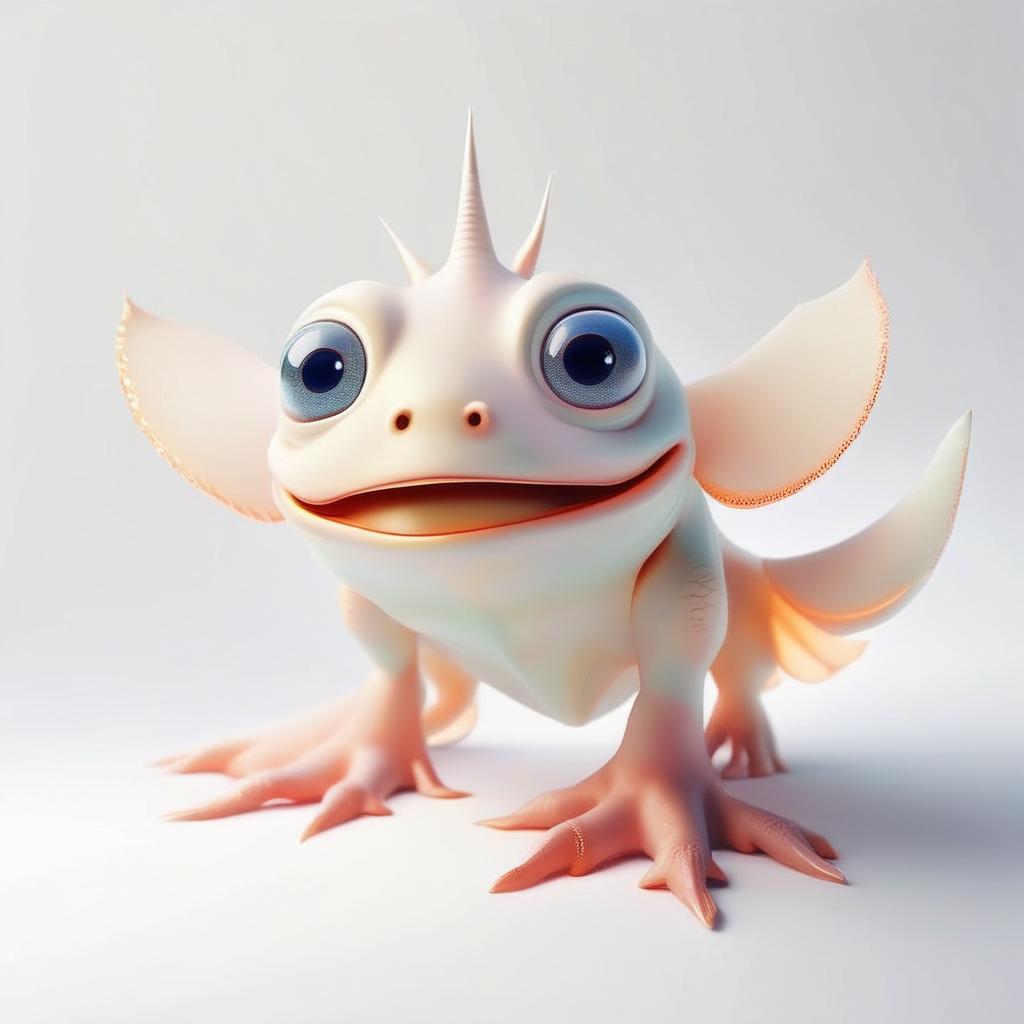} \\
        & \multicolumn{4}{c}{``...in space with the milky way behind him''} \\ \\[-0.075cm]
        
        {\includegraphics[width=0.19\columnwidth]{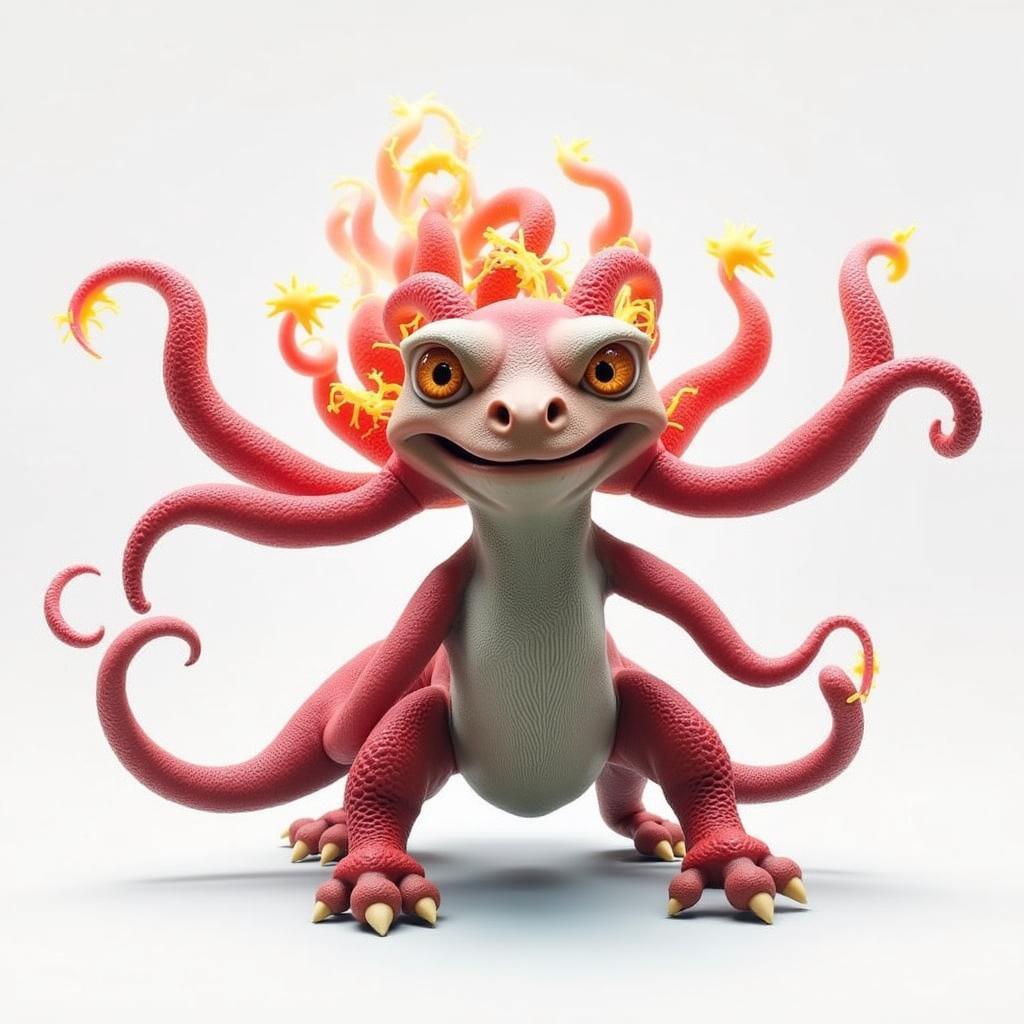}} &
        \includegraphics[width=0.19\columnwidth]{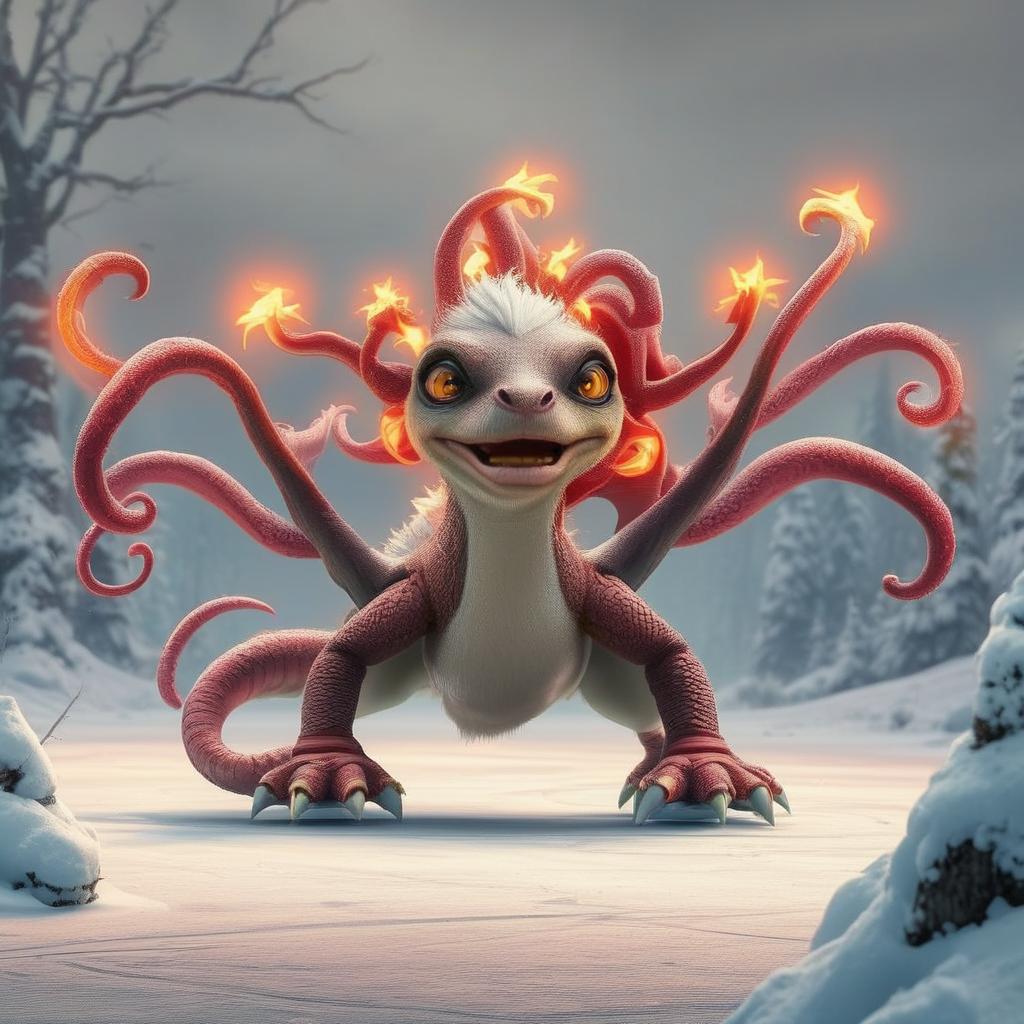} &
        \includegraphics[width=0.19\columnwidth]{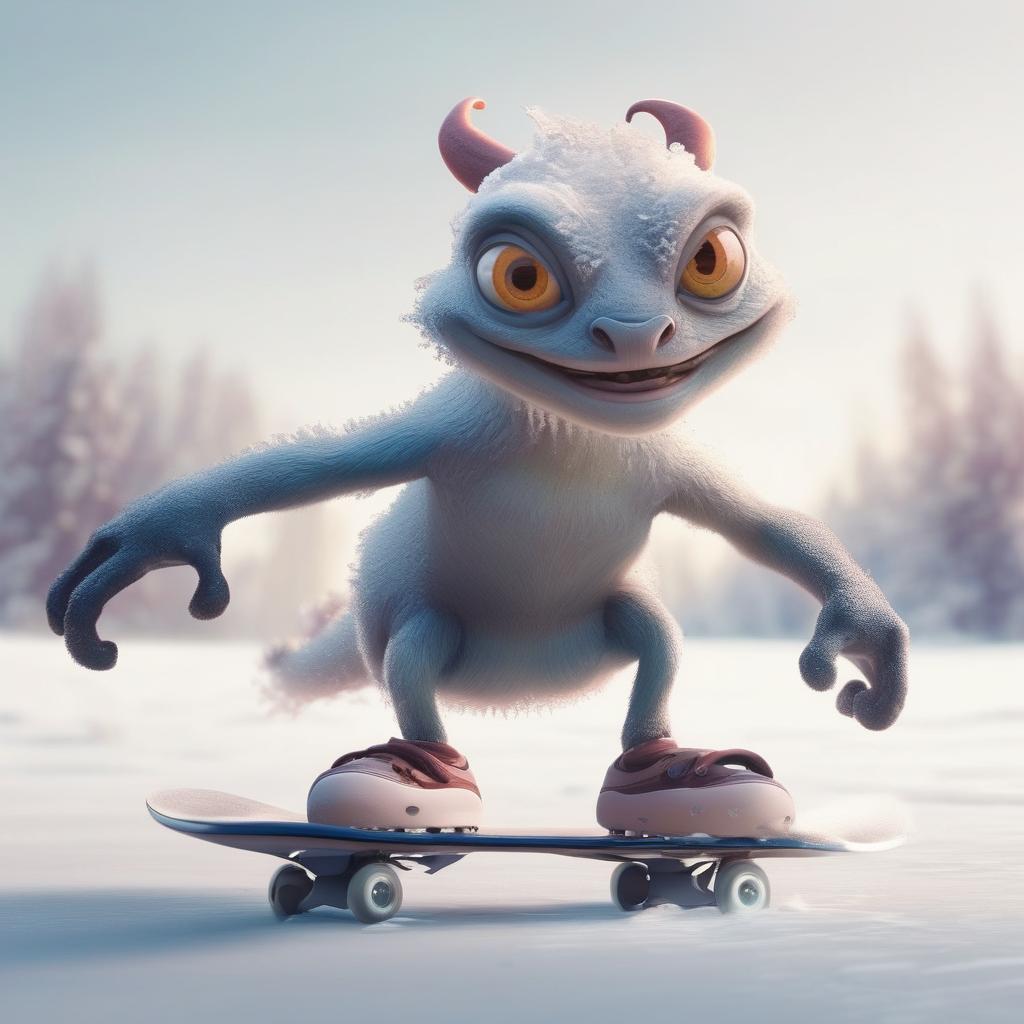} &  
        \includegraphics[width=0.19\columnwidth]{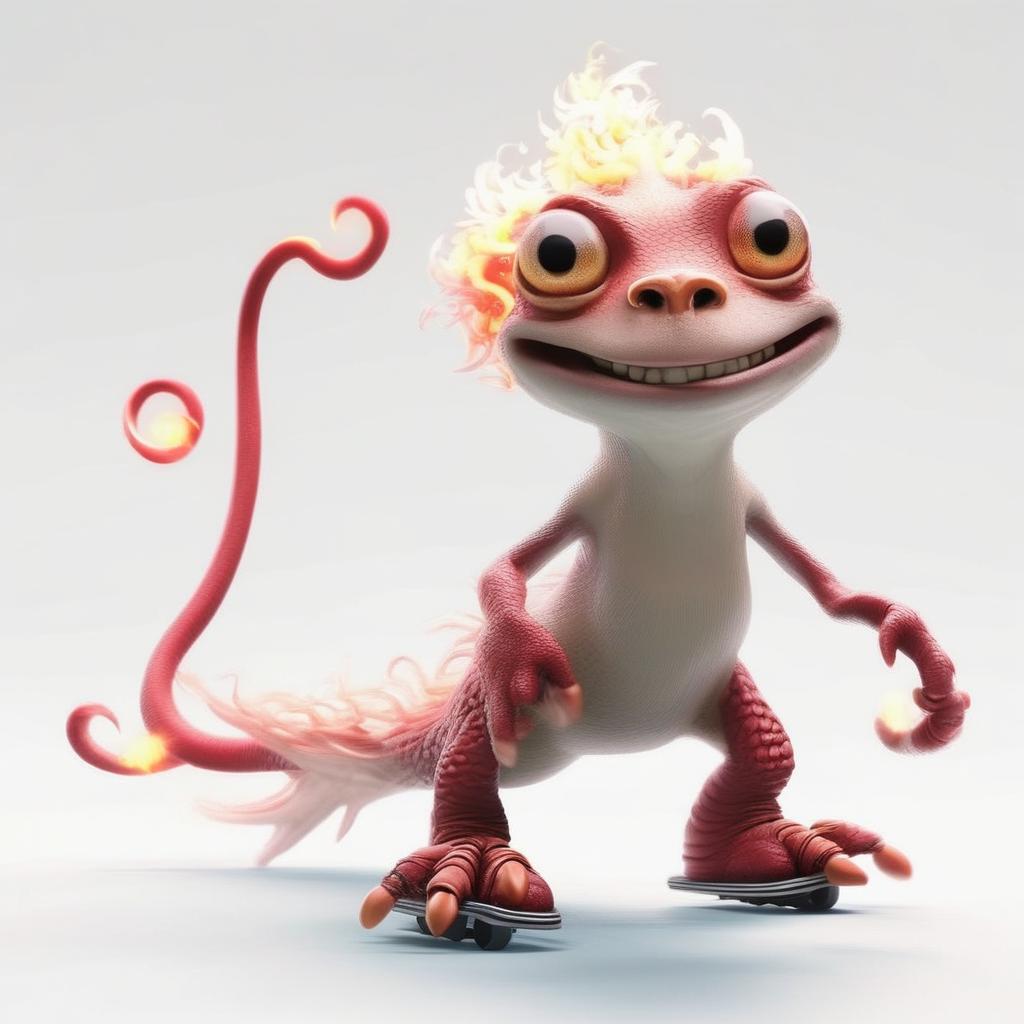} &
        \includegraphics[width=0.19\columnwidth]{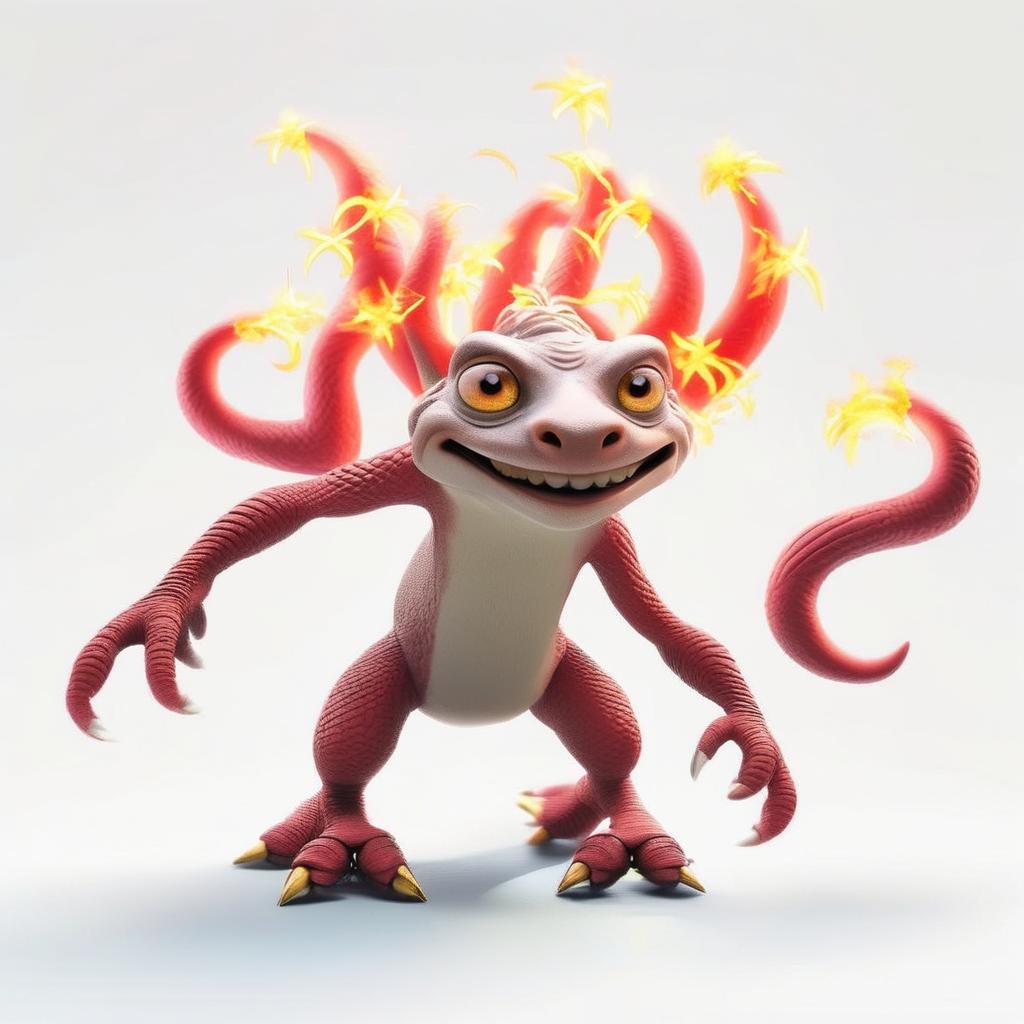} \\
        & \multicolumn{4}{c}{``...in the snow''} \\ \\[-0.075cm]

        {\includegraphics[width=0.19\columnwidth]{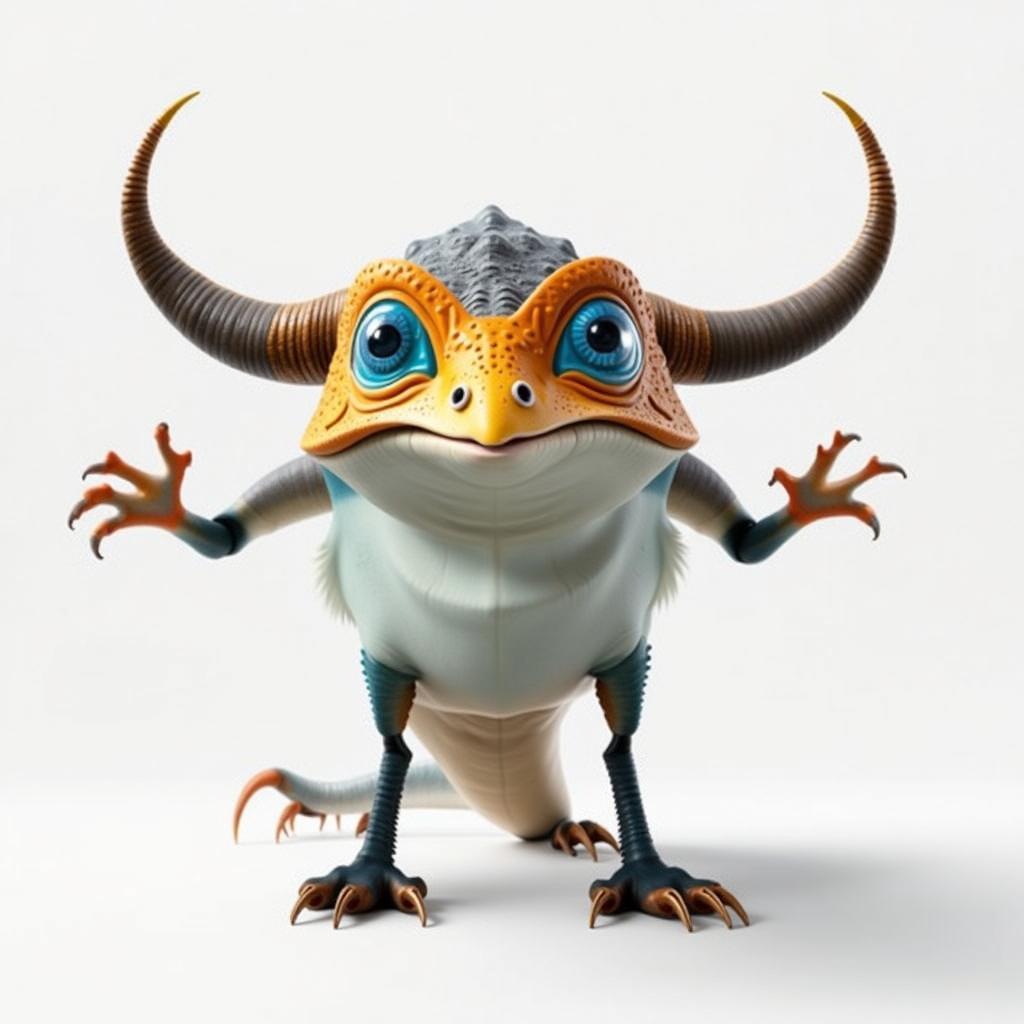}} &
        \includegraphics[width=0.19\columnwidth]{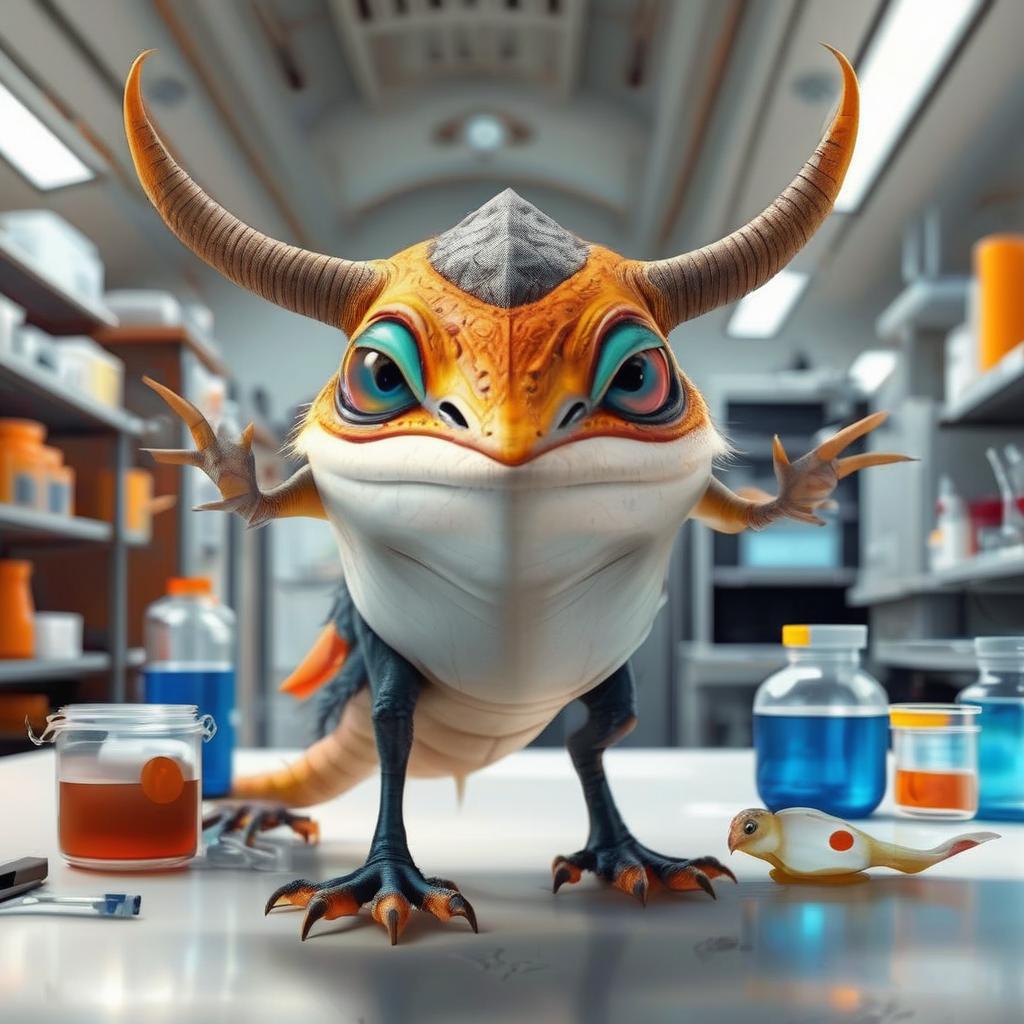} &
        \includegraphics[width=0.19\columnwidth]{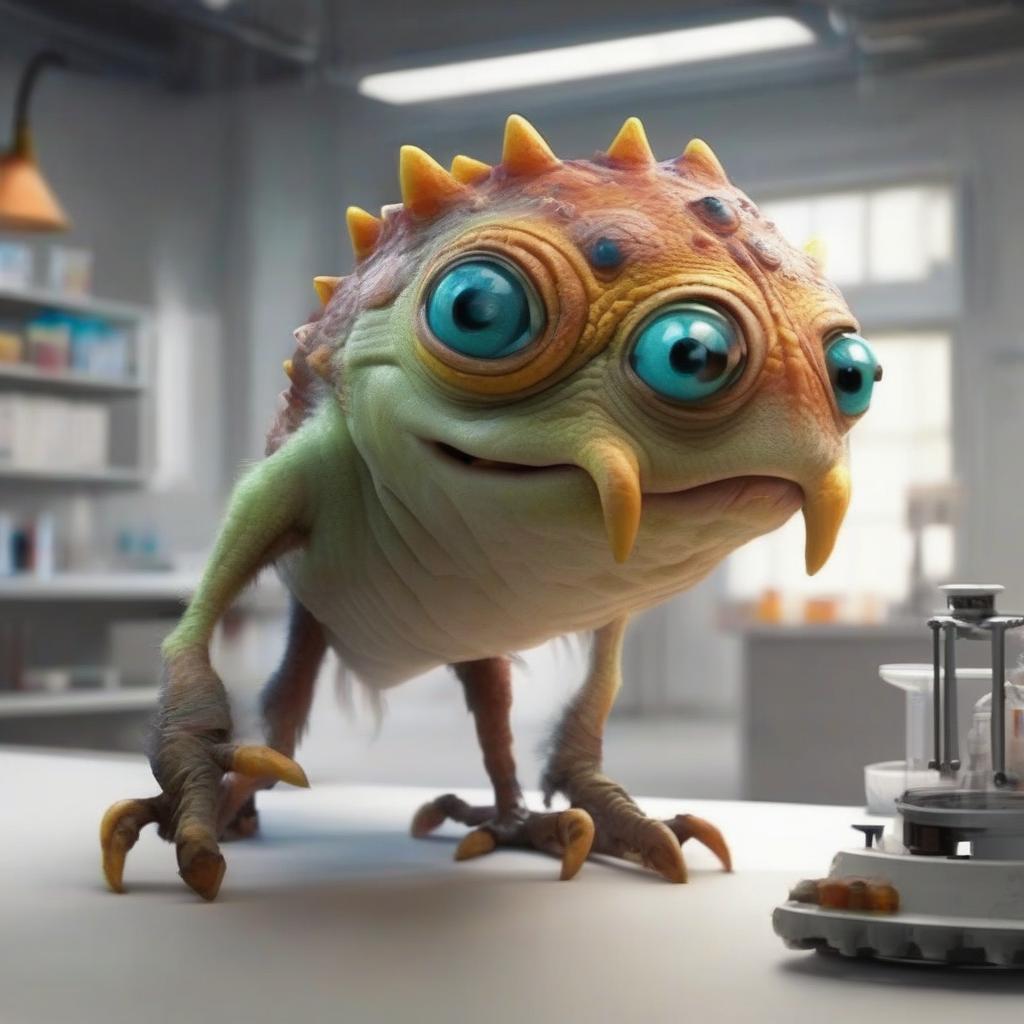} &  
        \includegraphics[width=0.19\columnwidth]{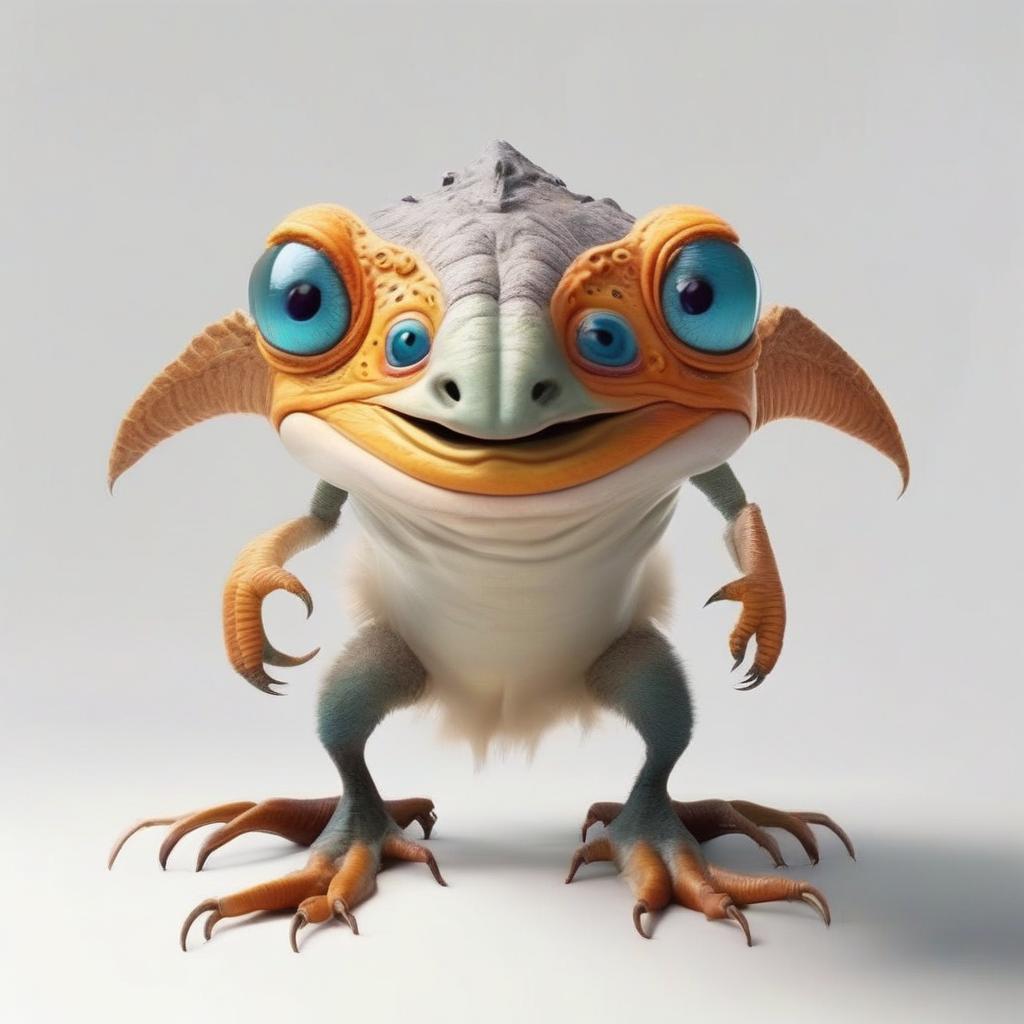} &
        \includegraphics[width=0.19\columnwidth]{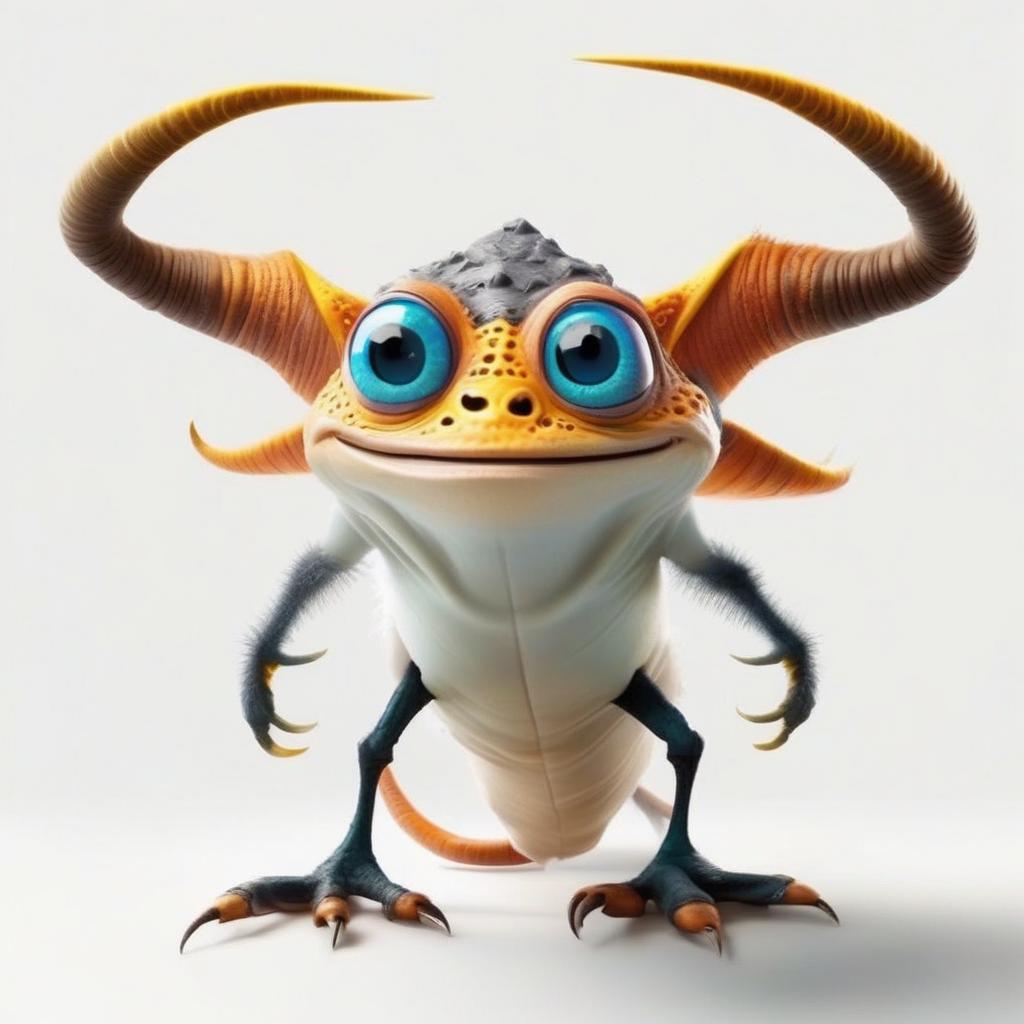} \\
        & \multicolumn{4}{c}{``...in a lab''} \\

        {\includegraphics[width=0.19\columnwidth]{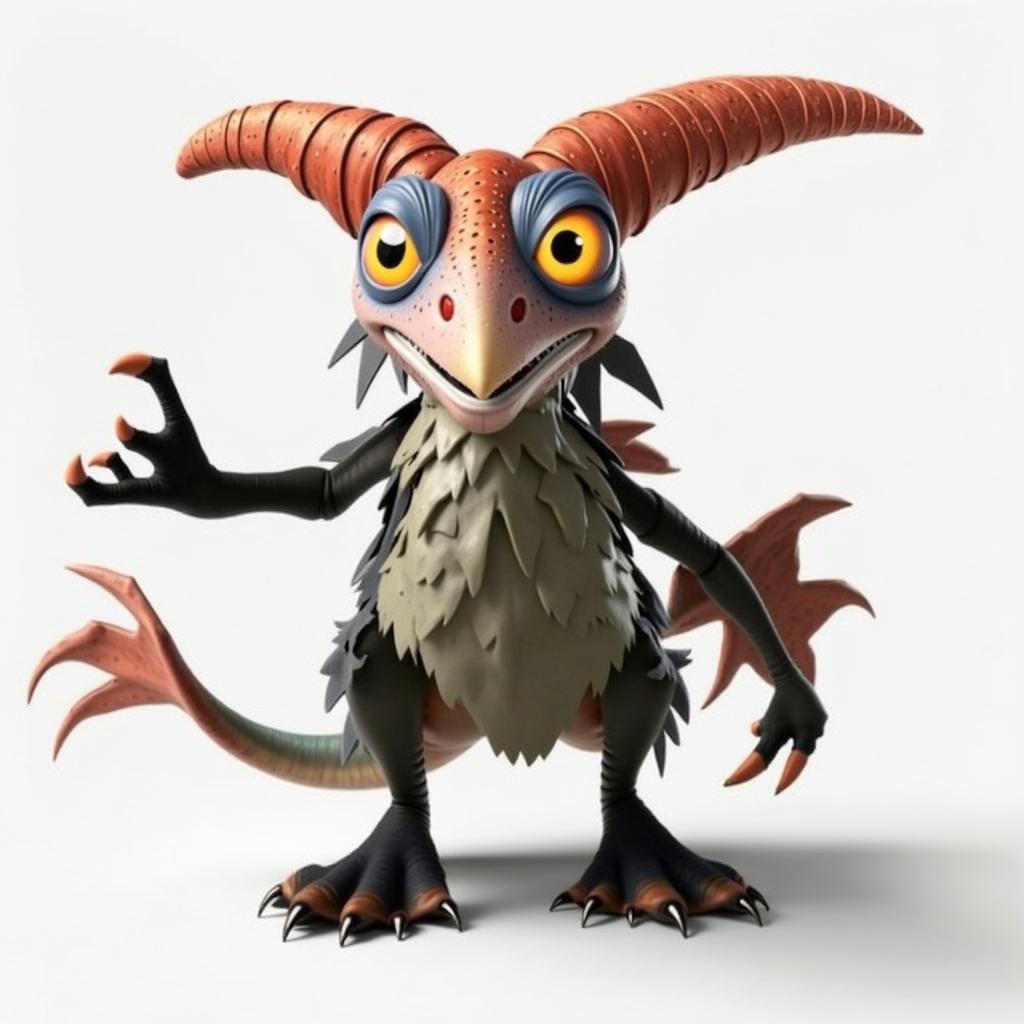}} &
        \includegraphics[width=0.19\columnwidth]{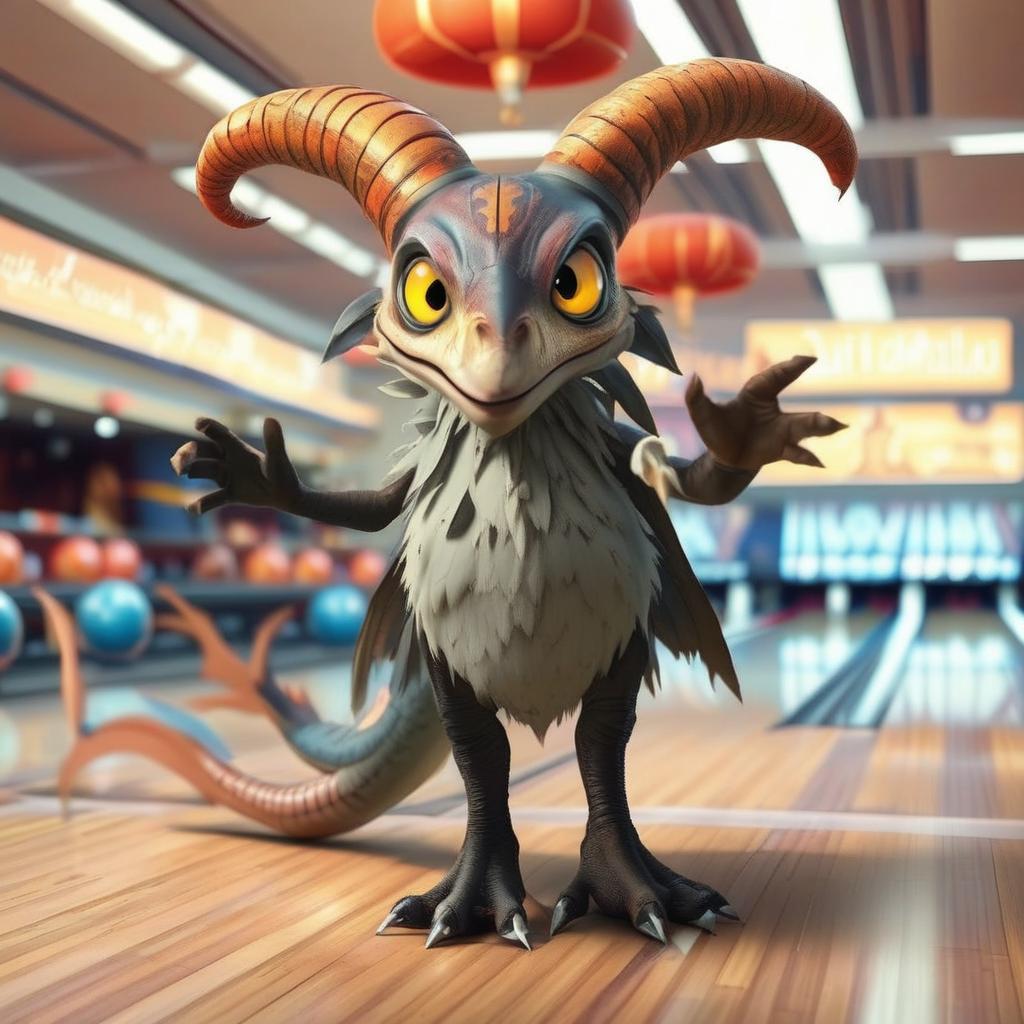} &
        \includegraphics[width=0.19\columnwidth]{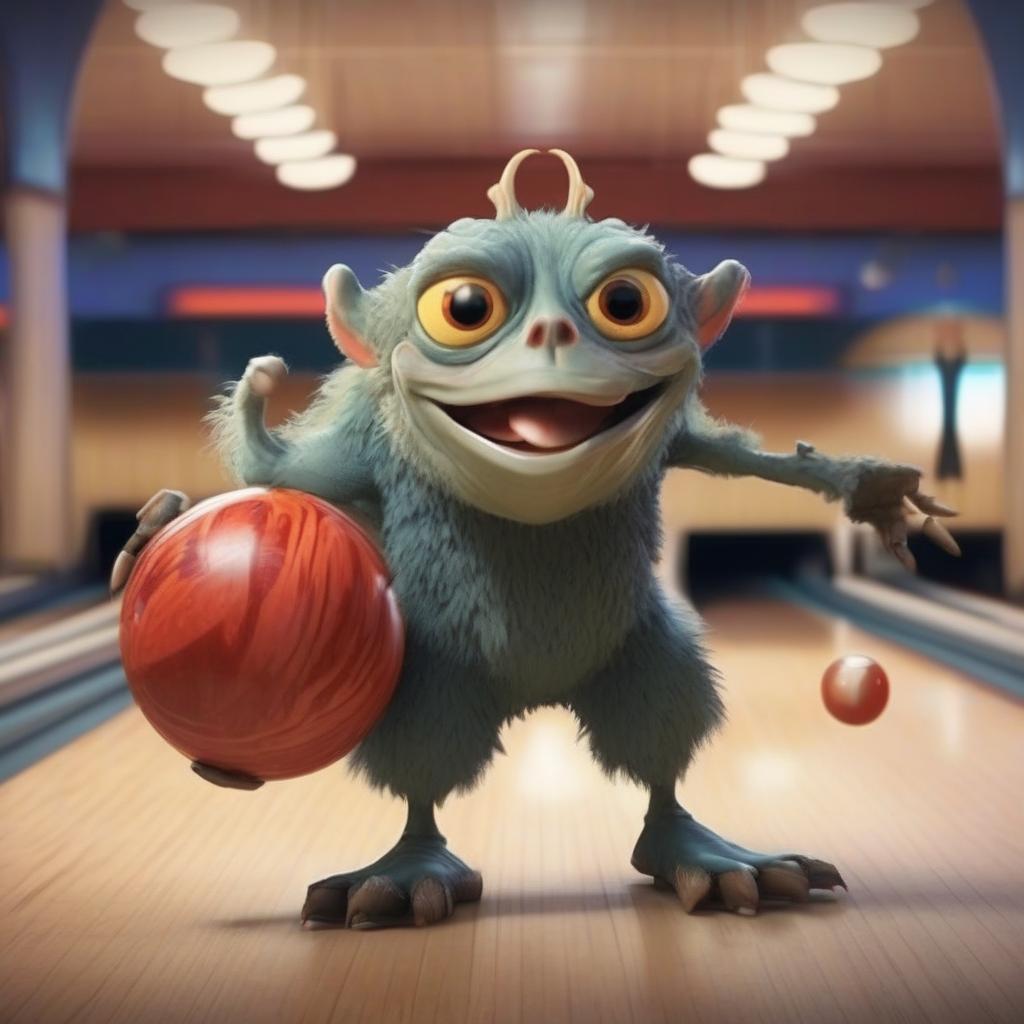} &  
        \includegraphics[width=0.19\columnwidth]{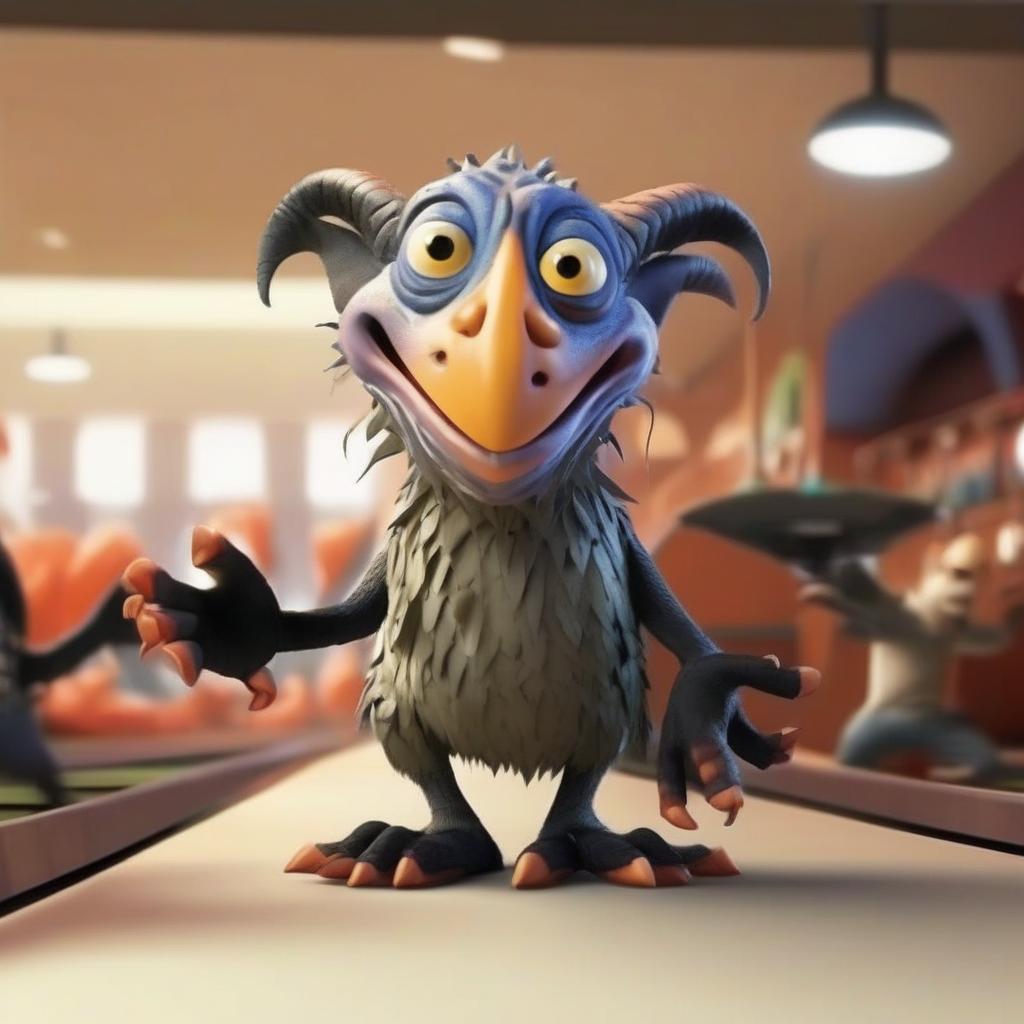} &
        \includegraphics[width=0.19\columnwidth]{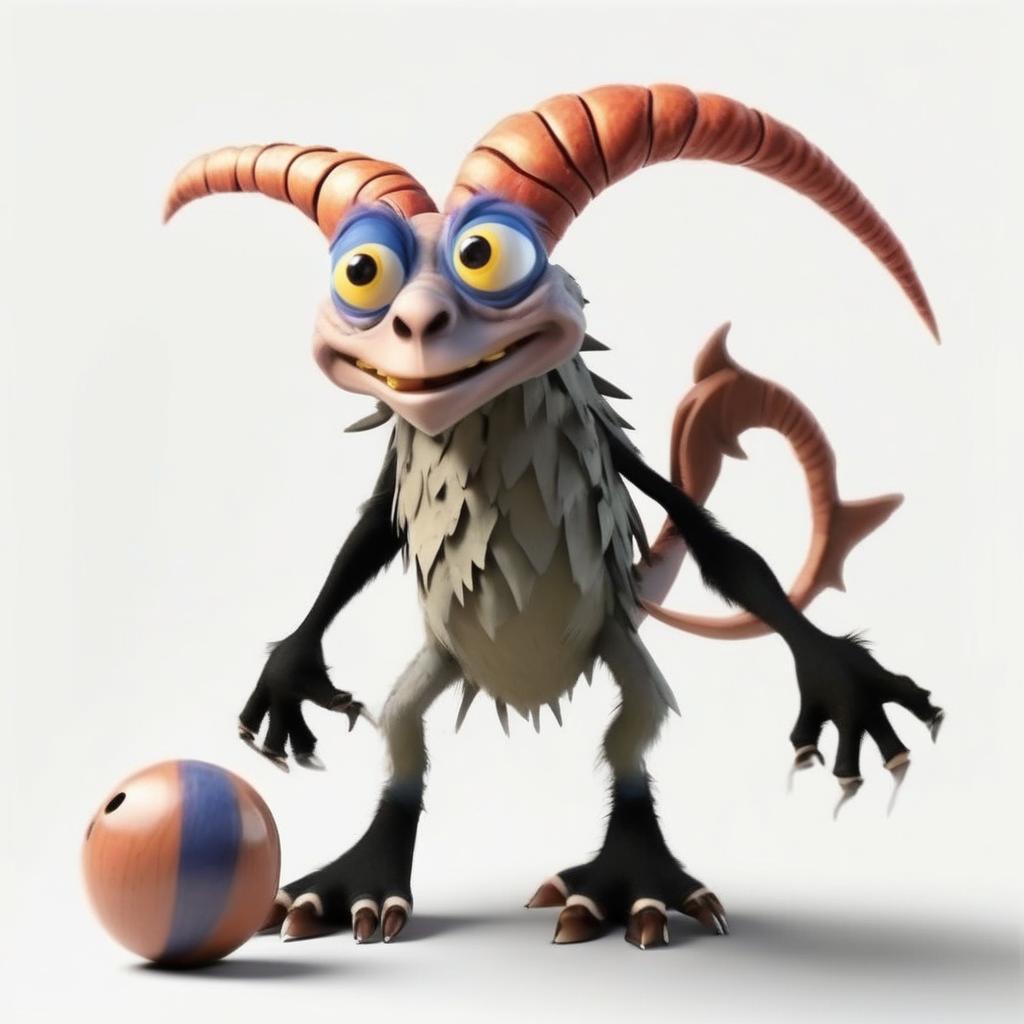} \\
        & \multicolumn{4}{c}{``...in a bowling alley''}
        
    \end{tabular}
    }
    \vspace{-0.2cm}
    \caption{\textbf{IP-Lora vs. IP-Adapter+.} Given a learned representation, we show results of rendering the concept in a new scene. 
    }
    \label{fig:lora_vs_baseline}
\end{figure}

\begin{figure*}
    \centering
    \setlength{\tabcolsep}{0.5pt}
    \renewcommand{\arraystretch}{0.5}
    \addtolength{\belowcaptionskip}{-5pt}
    {\small

    \begin{tabular}{c c c c c c c c c}

        & 
        \includegraphics[width=0.115\textwidth]{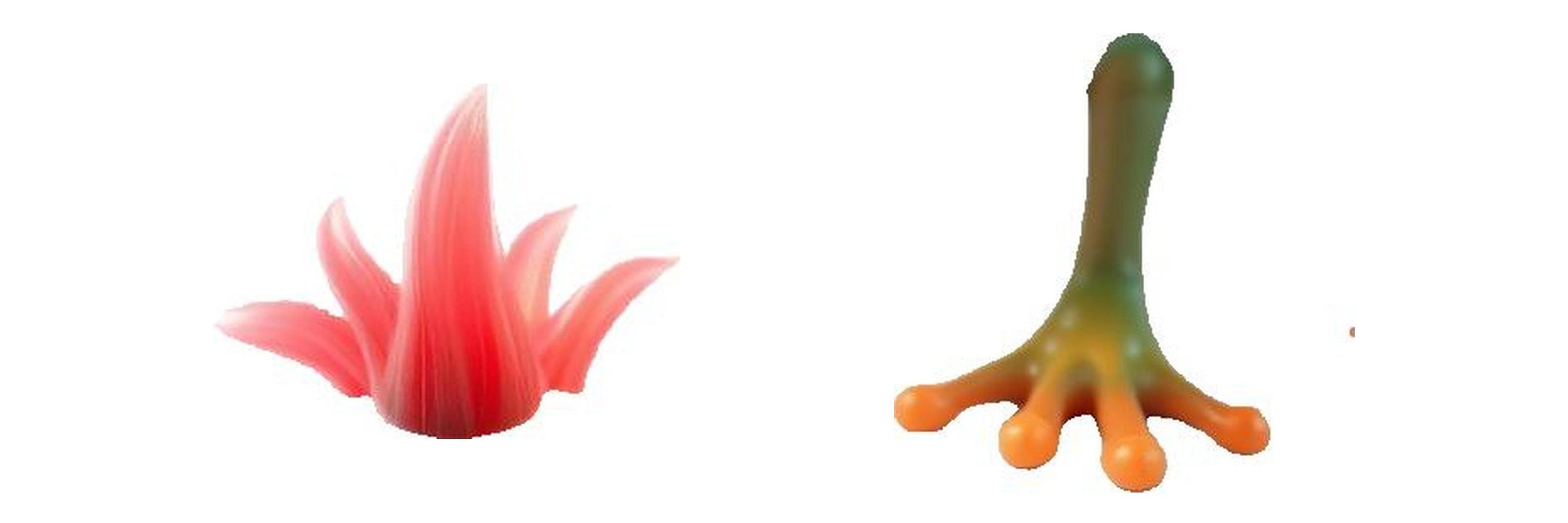} &
        \includegraphics[width=0.115\textwidth]{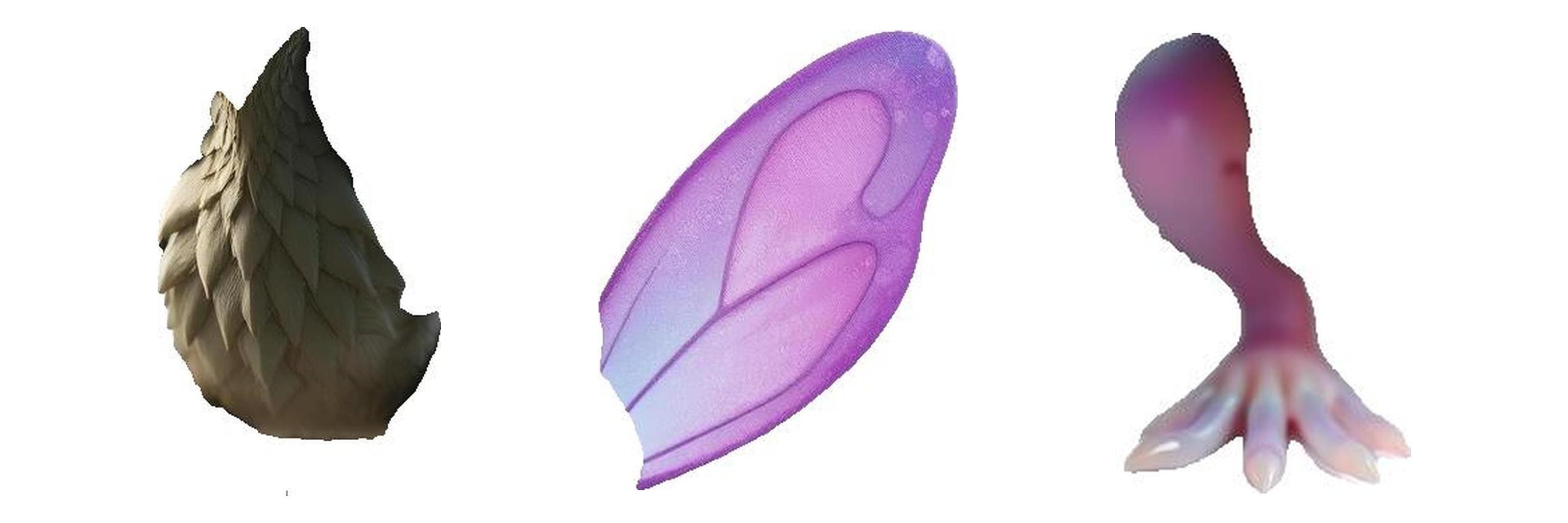} &
        \includegraphics[width=0.115\textwidth]{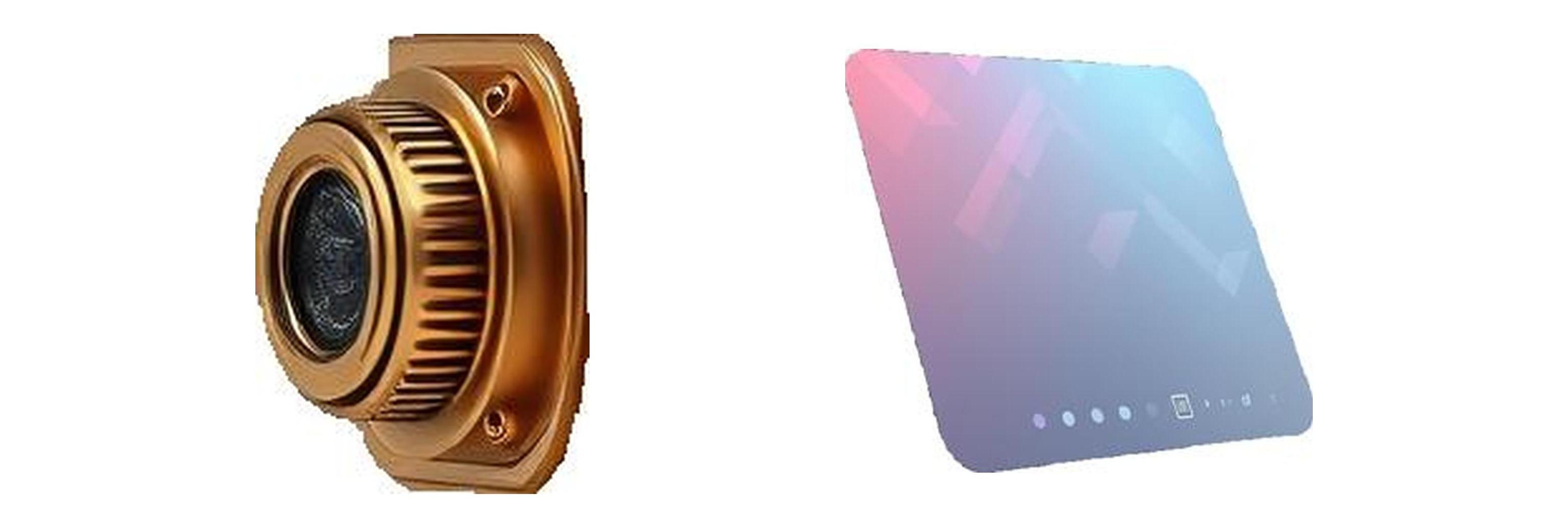} &
        \includegraphics[width=0.115\textwidth]{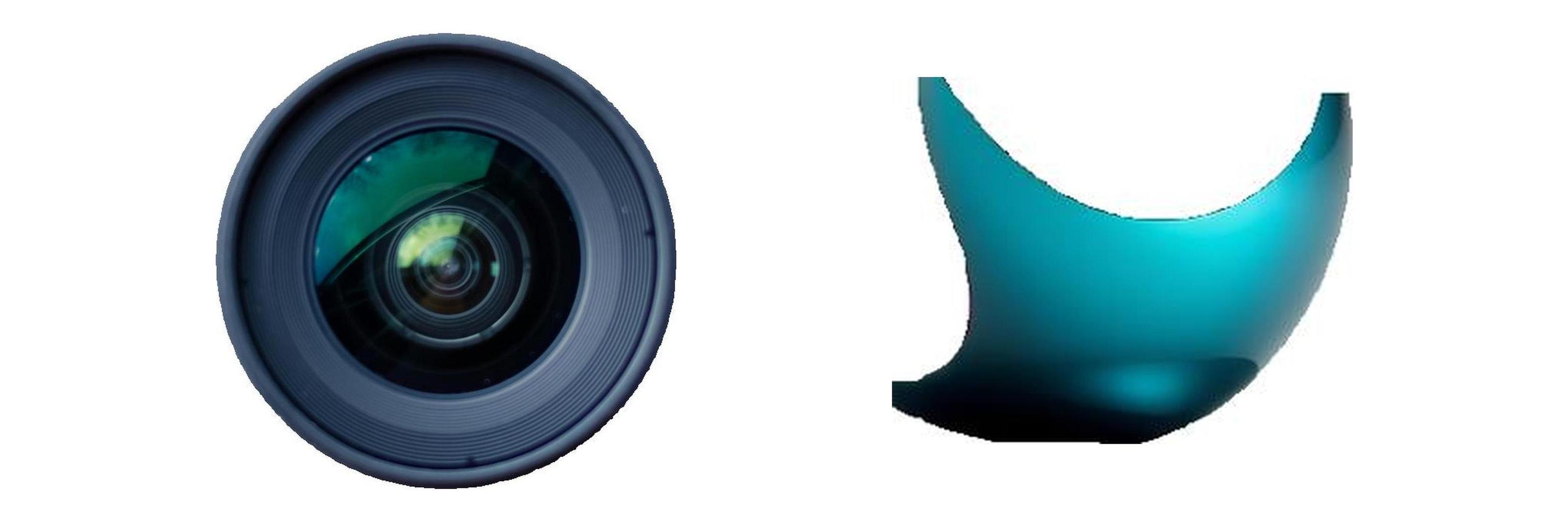} &      
        \includegraphics[width=0.115\textwidth]{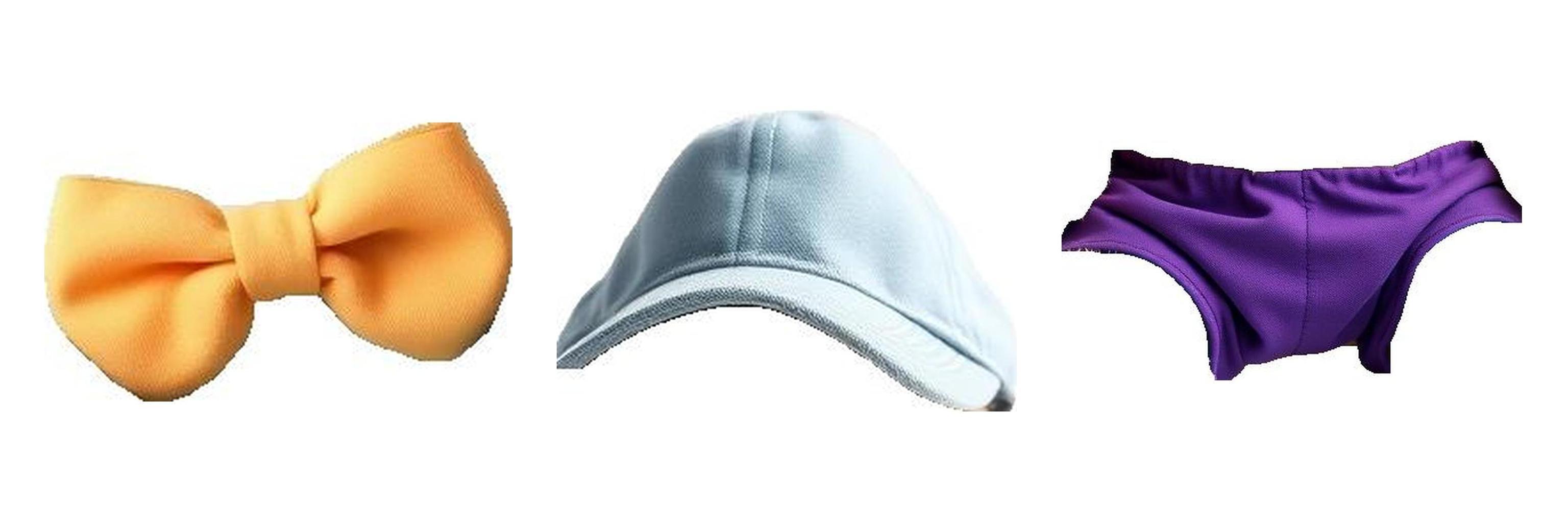} &
        
        \includegraphics[width=0.115\textwidth]{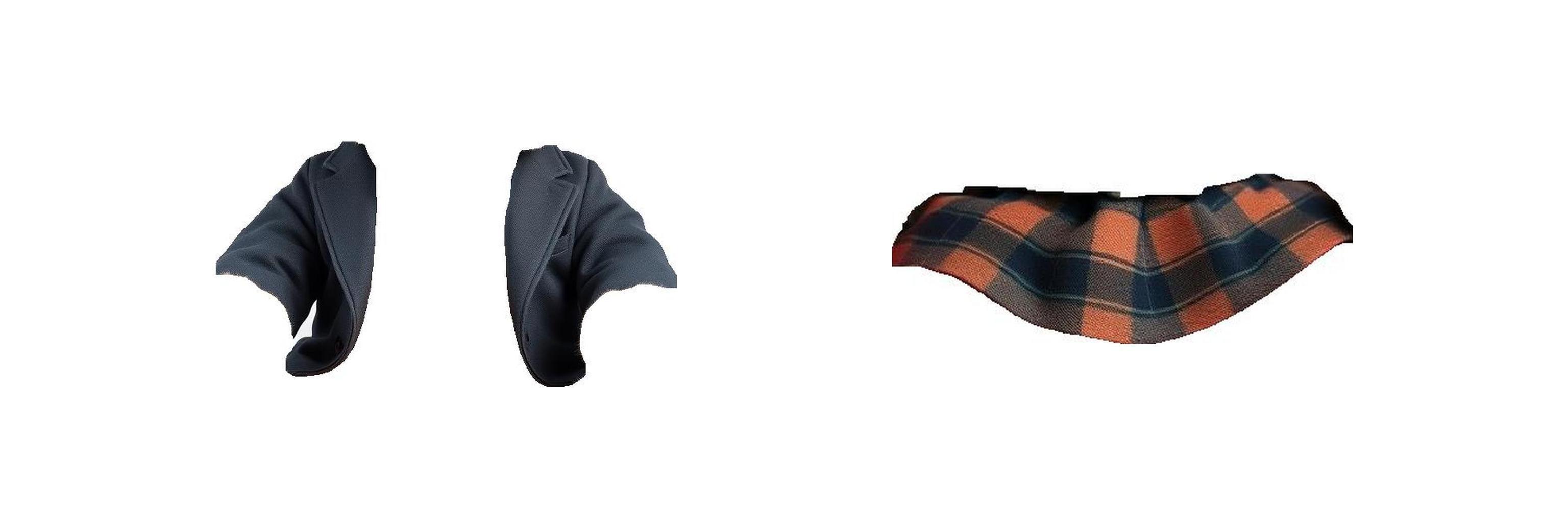} &
        
        \includegraphics[width=0.115\textwidth]{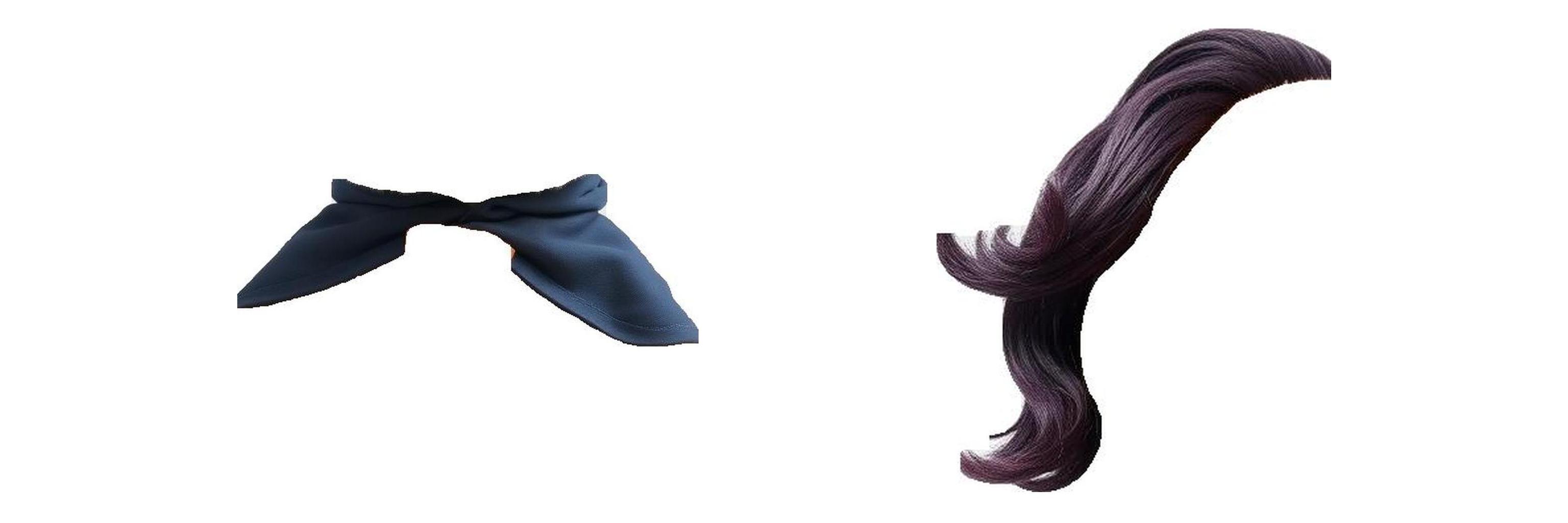} &
        
        \includegraphics[width=0.115\textwidth]{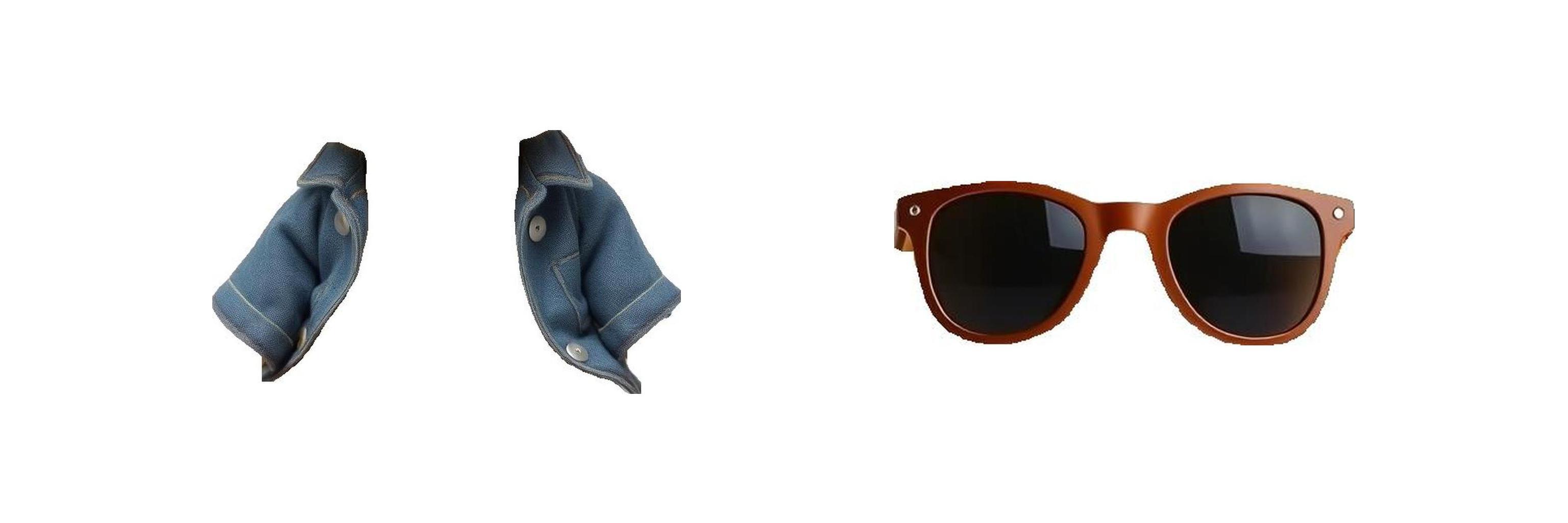} \\

        \raisebox{0.0325\linewidth}{\rotatebox[origin=t]{90}{\begin{tabular}{c@{}c@{}c@{}c@{}} PiT \end{tabular}}} &
        \includegraphics[width=0.115\textwidth]{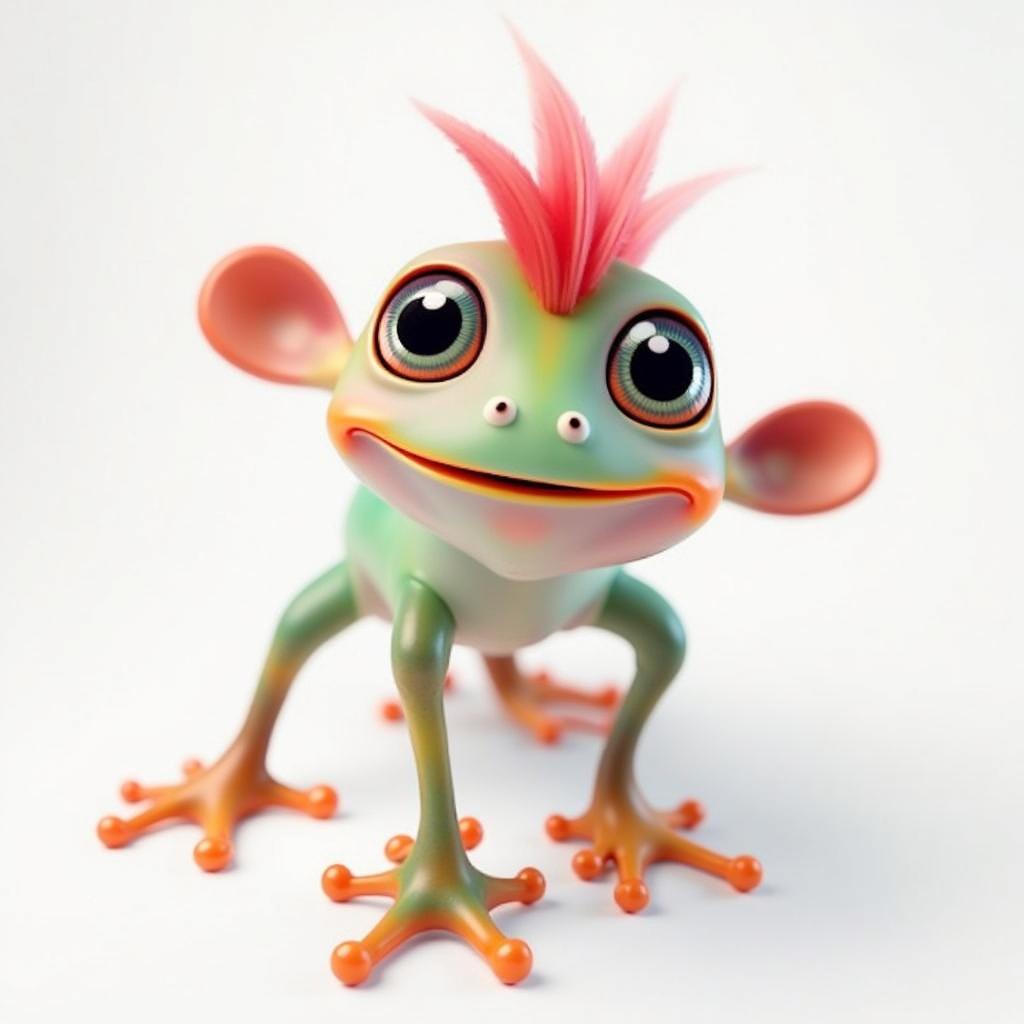} &
        \includegraphics[width=0.115\textwidth]{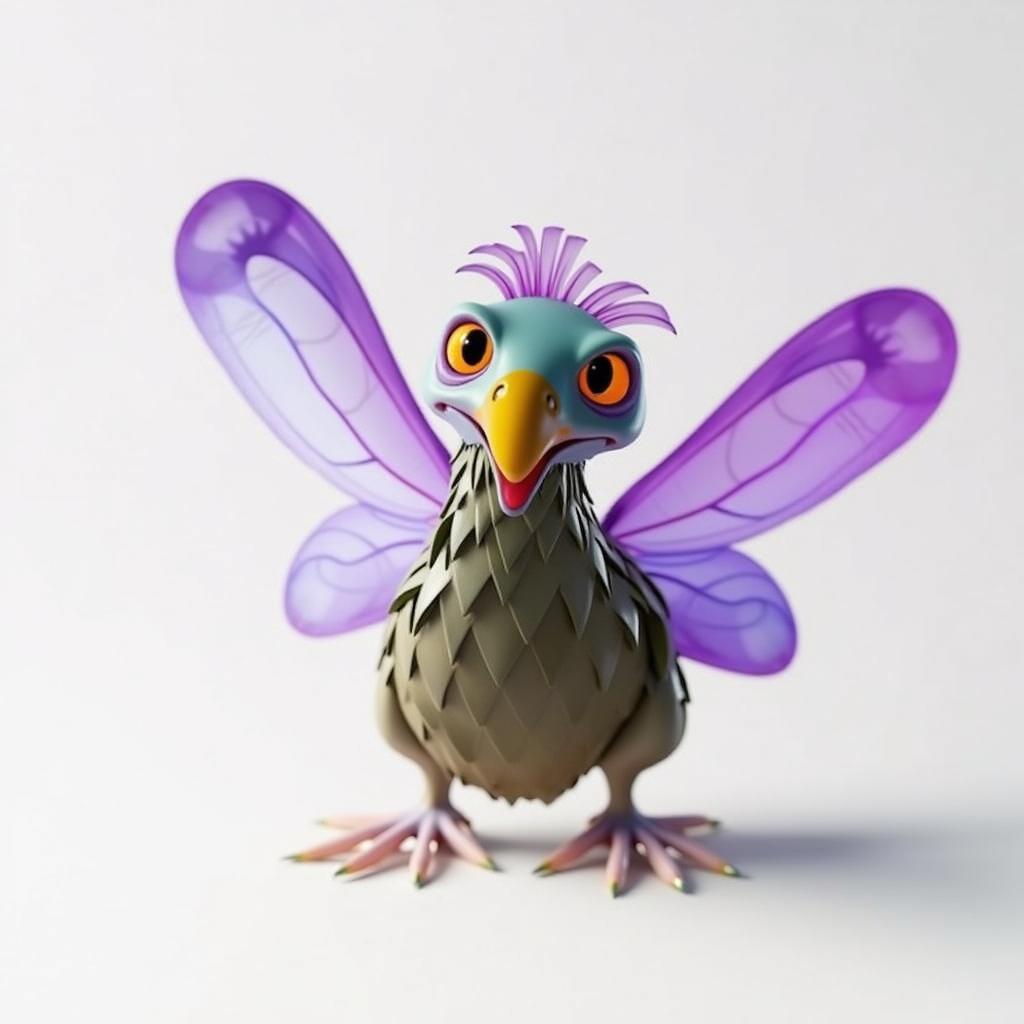} &
        \includegraphics[width=0.115\textwidth]{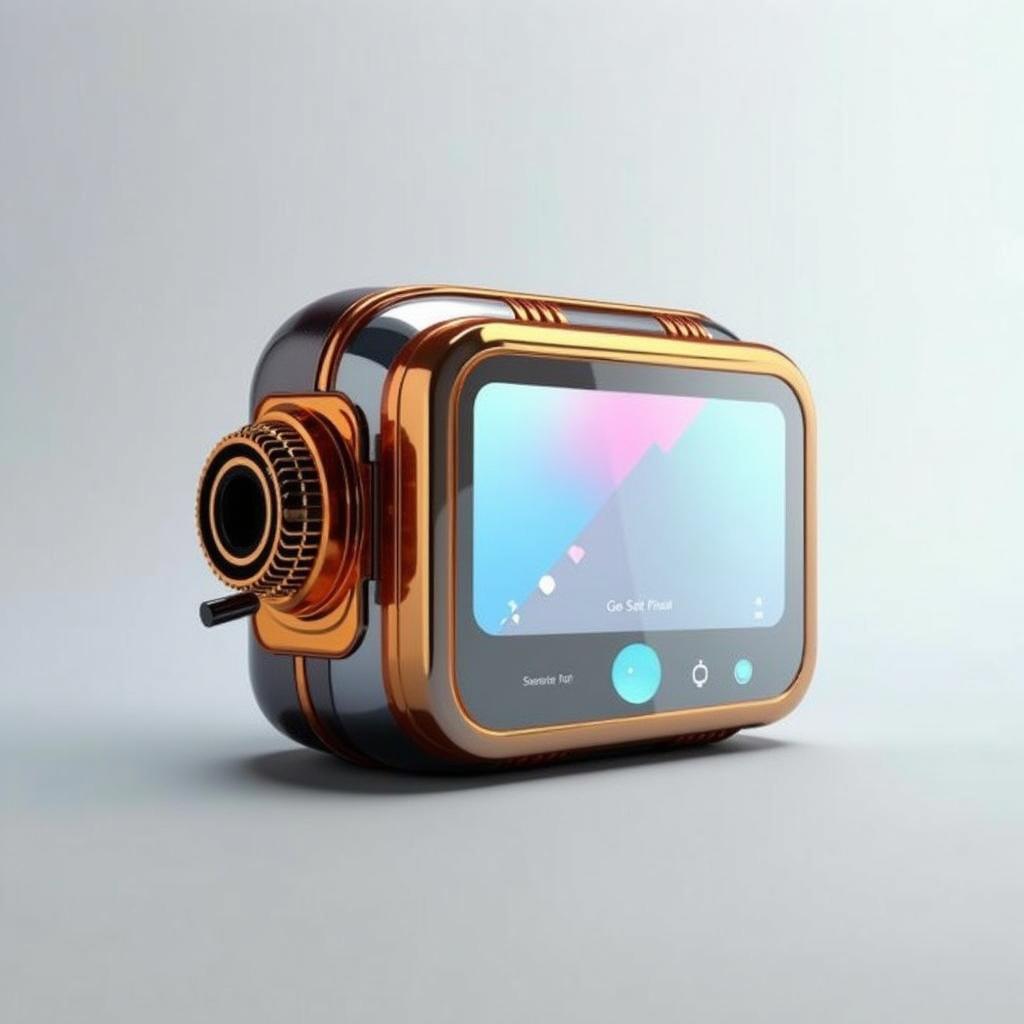} &
        \includegraphics[width=0.115\textwidth]{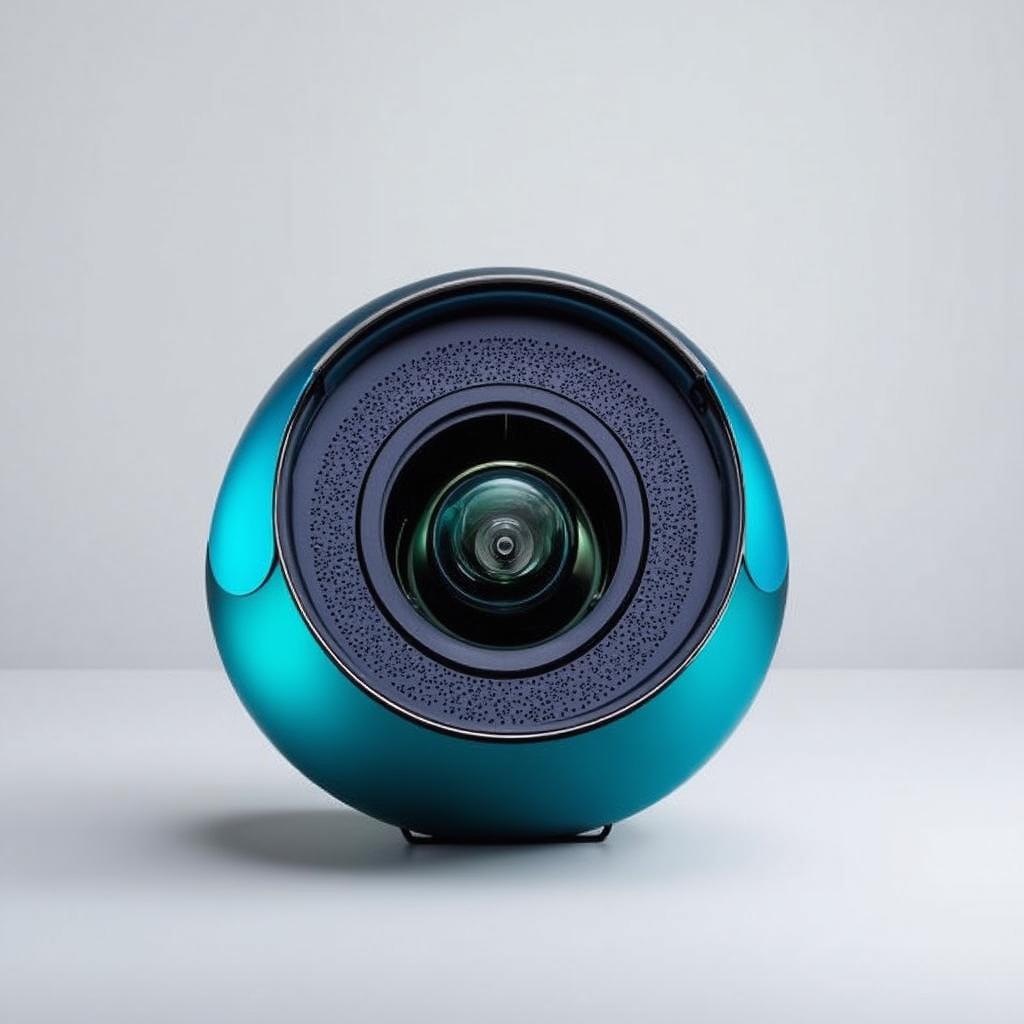} &
        \includegraphics[width=0.115\textwidth]{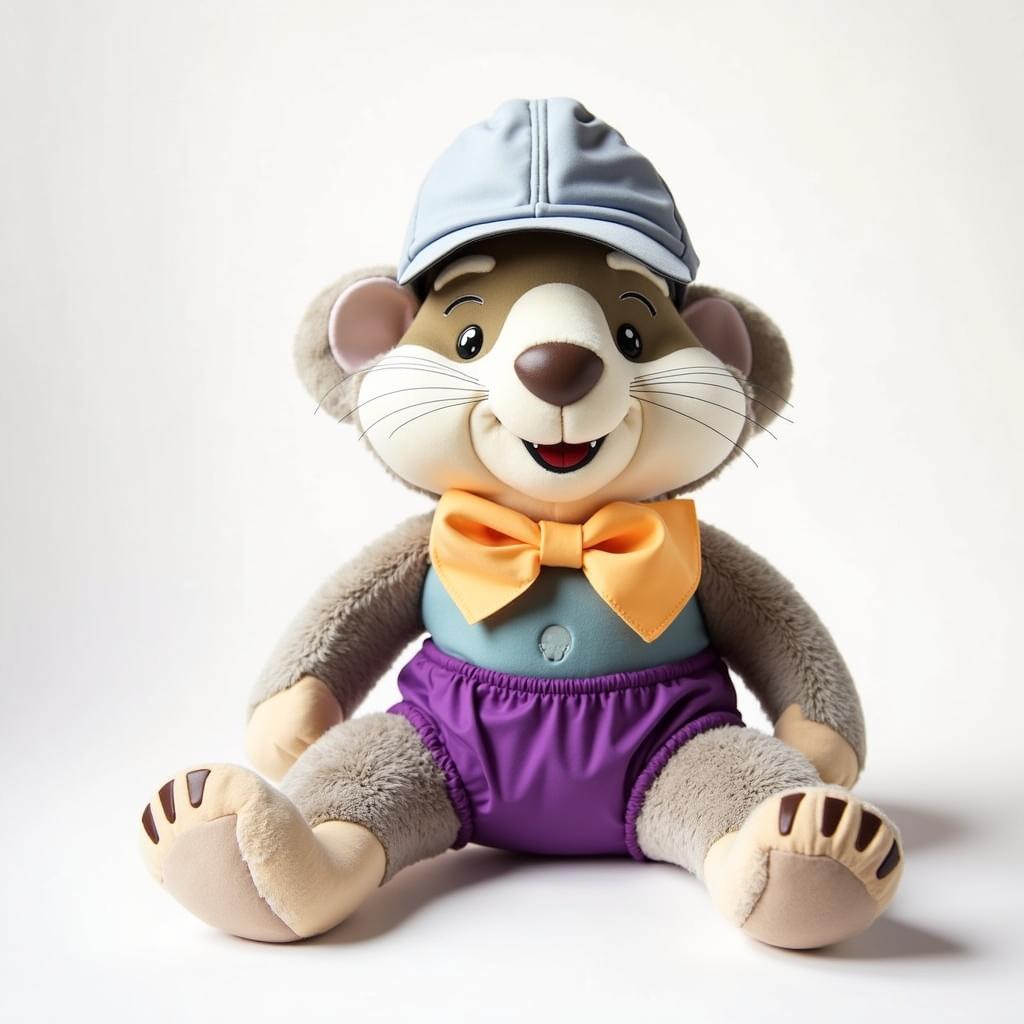} &
        \includegraphics[width=0.115\textwidth]{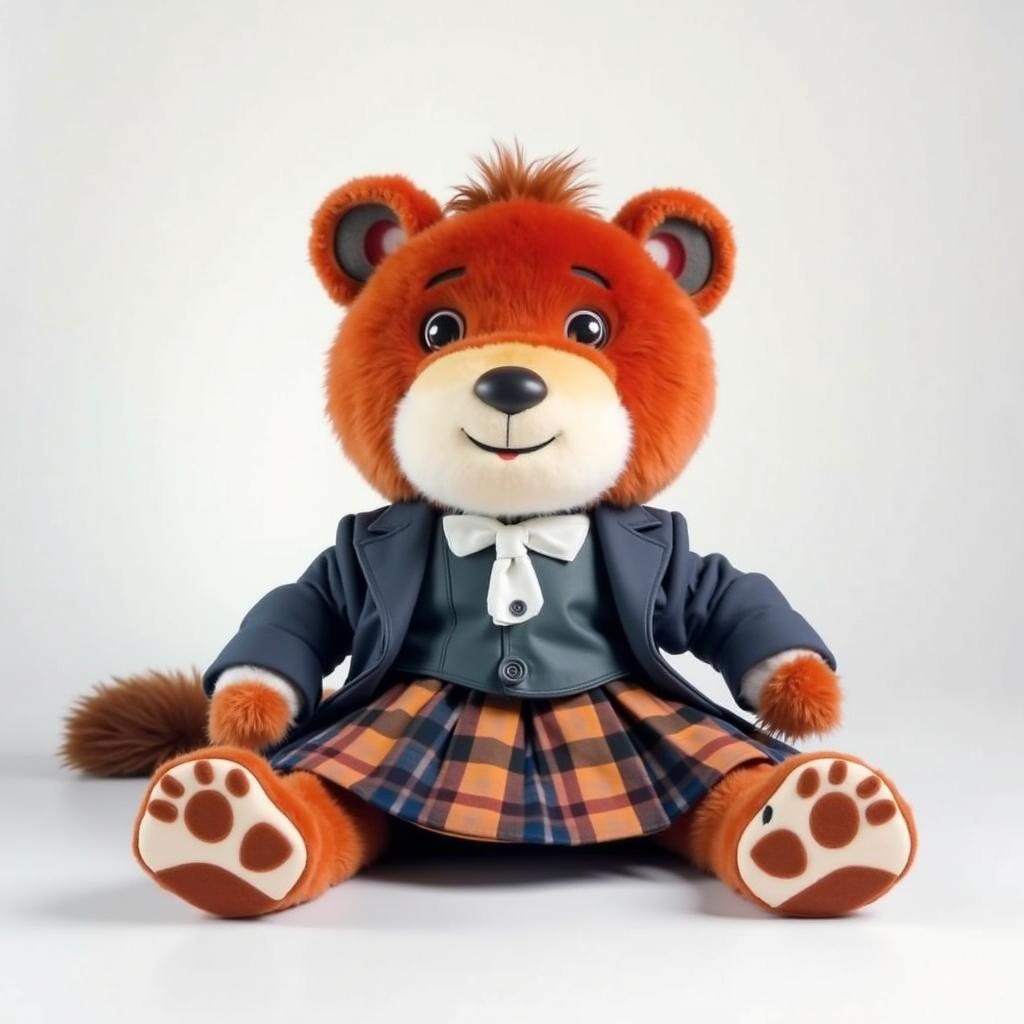} &
        \includegraphics[width=0.115\textwidth]{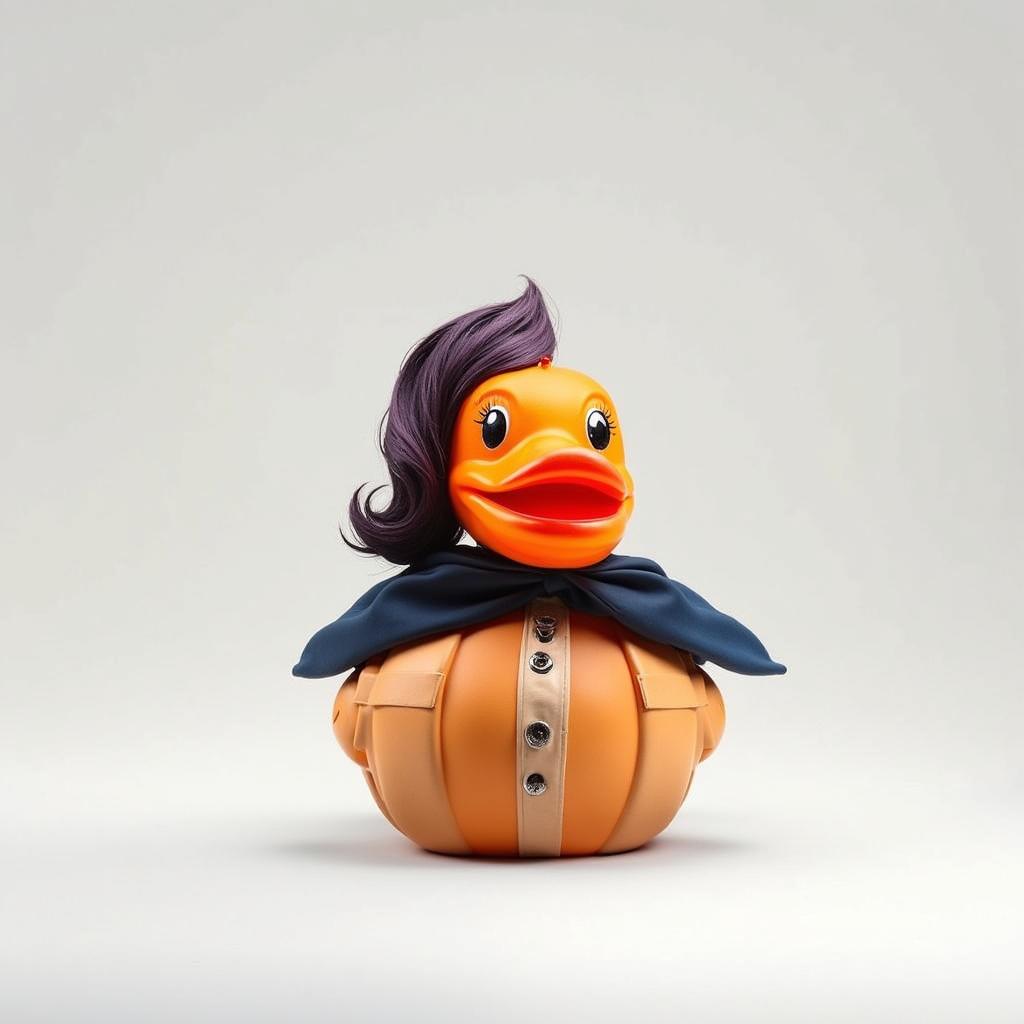 } &
        
        \includegraphics[width=0.115\textwidth]{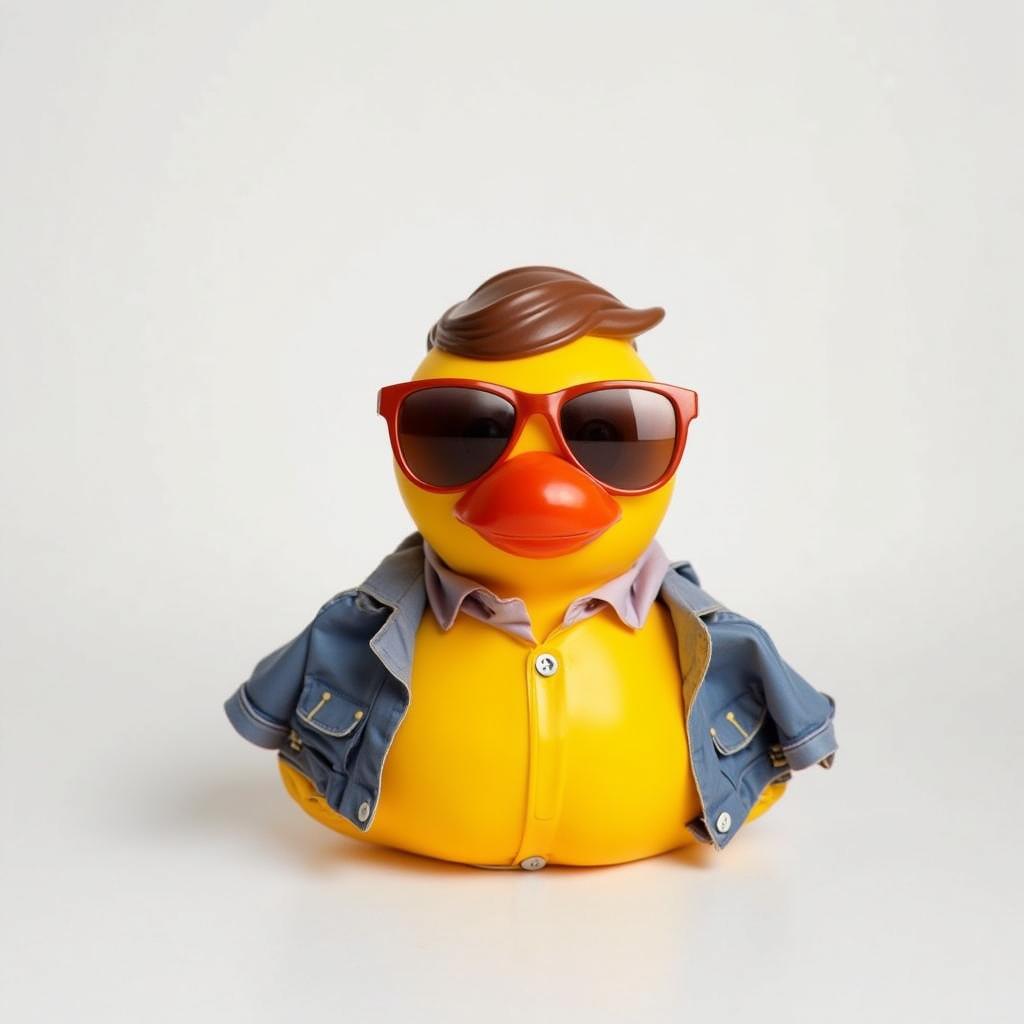} \\

        \raisebox{0.045\linewidth}{\rotatebox[origin=t]{90}{\begin{tabular}{c@{}c@{}c@{}c@{}} IP-Adapter+ \end{tabular}}} &
        \includegraphics[width=0.115\textwidth]{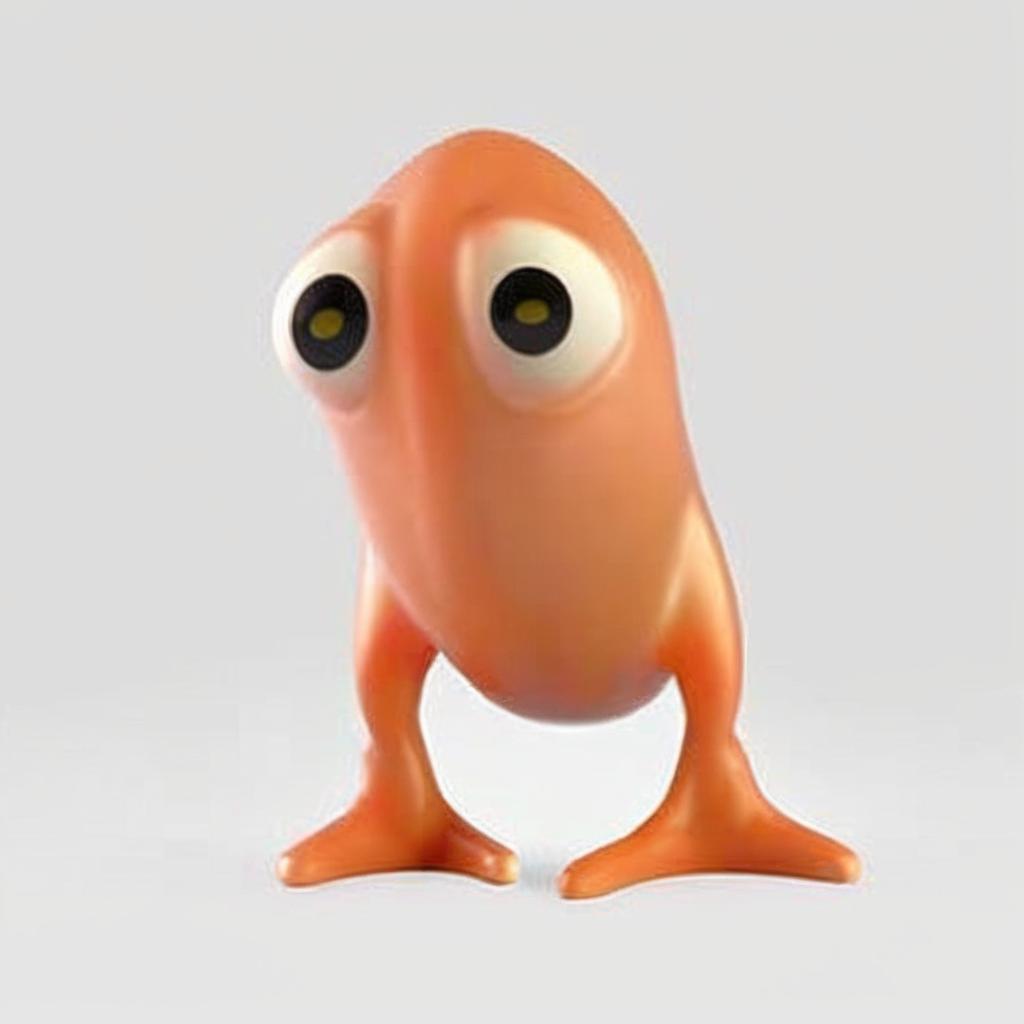} &
        \includegraphics[width=0.115\textwidth]{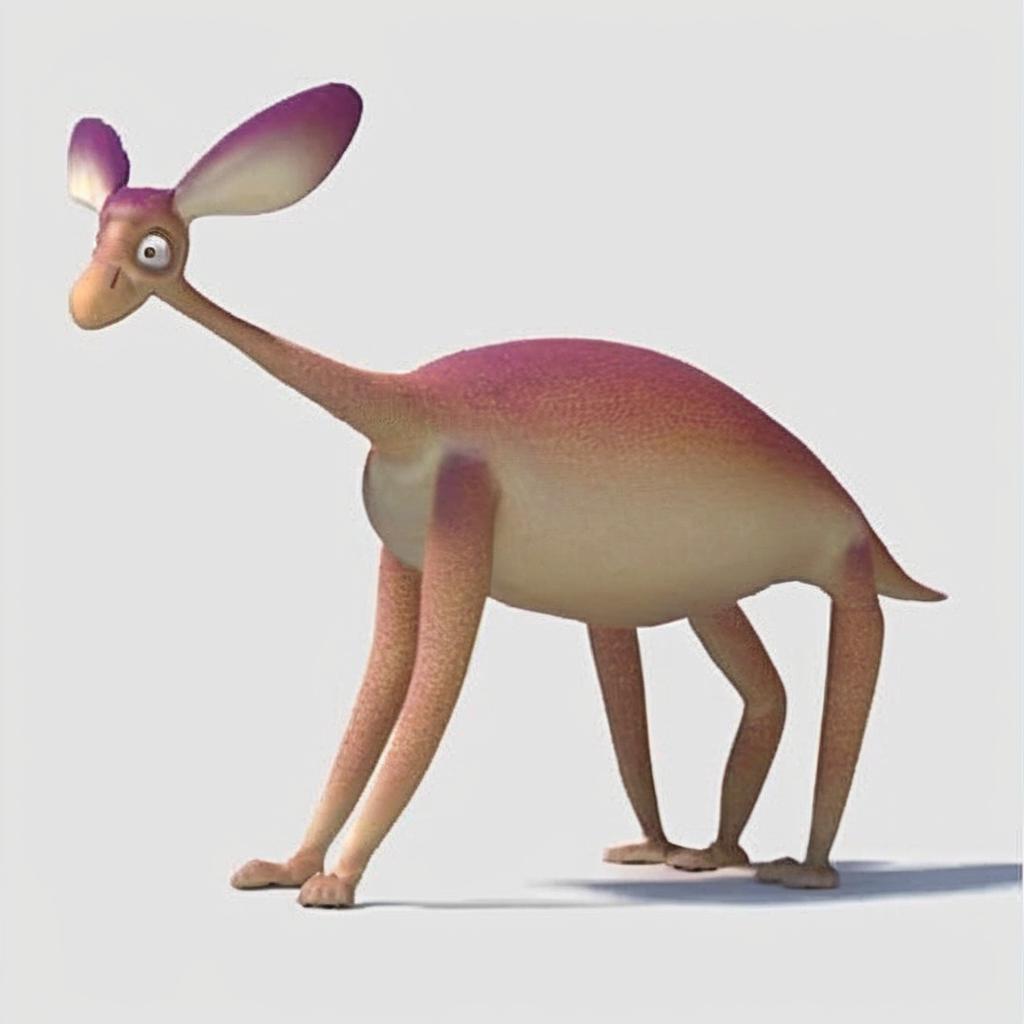} &
        \includegraphics[width=0.115\textwidth]{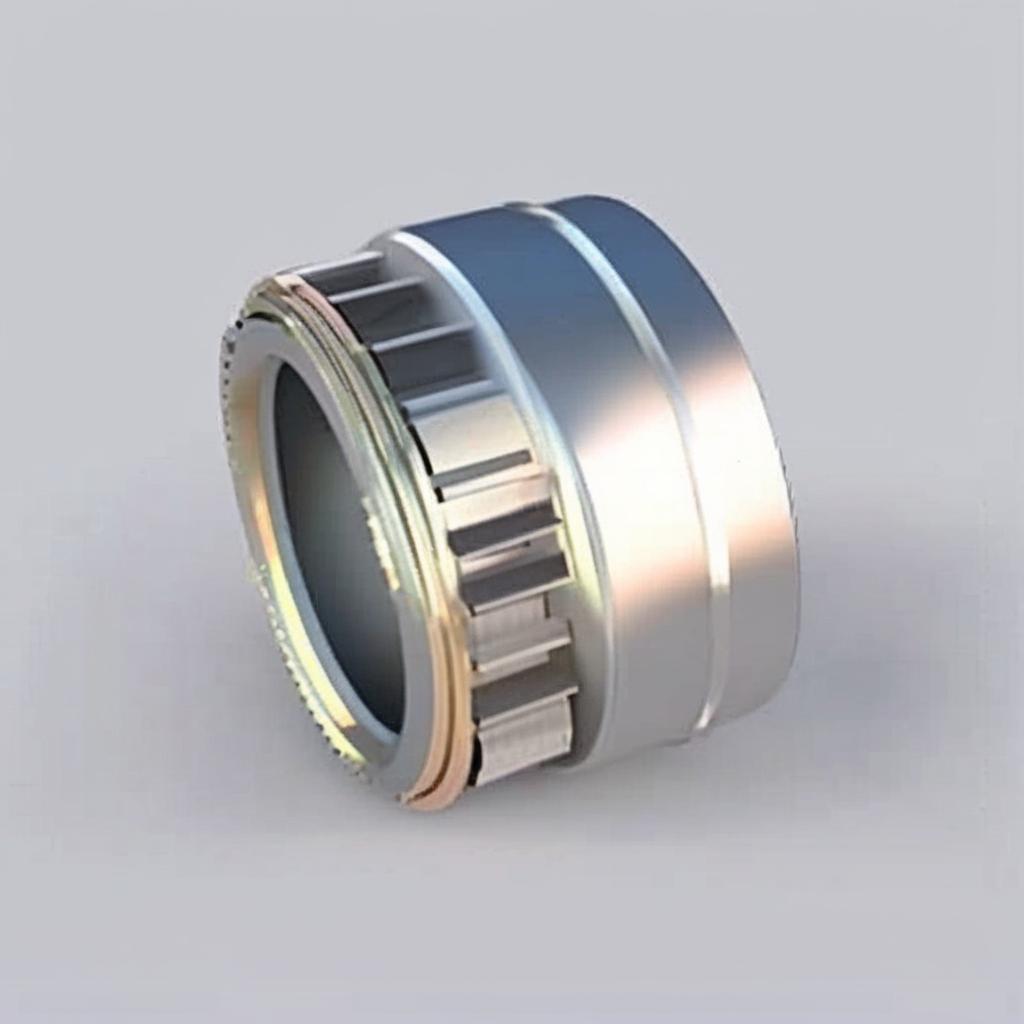} &
        \includegraphics[width=0.115\textwidth]{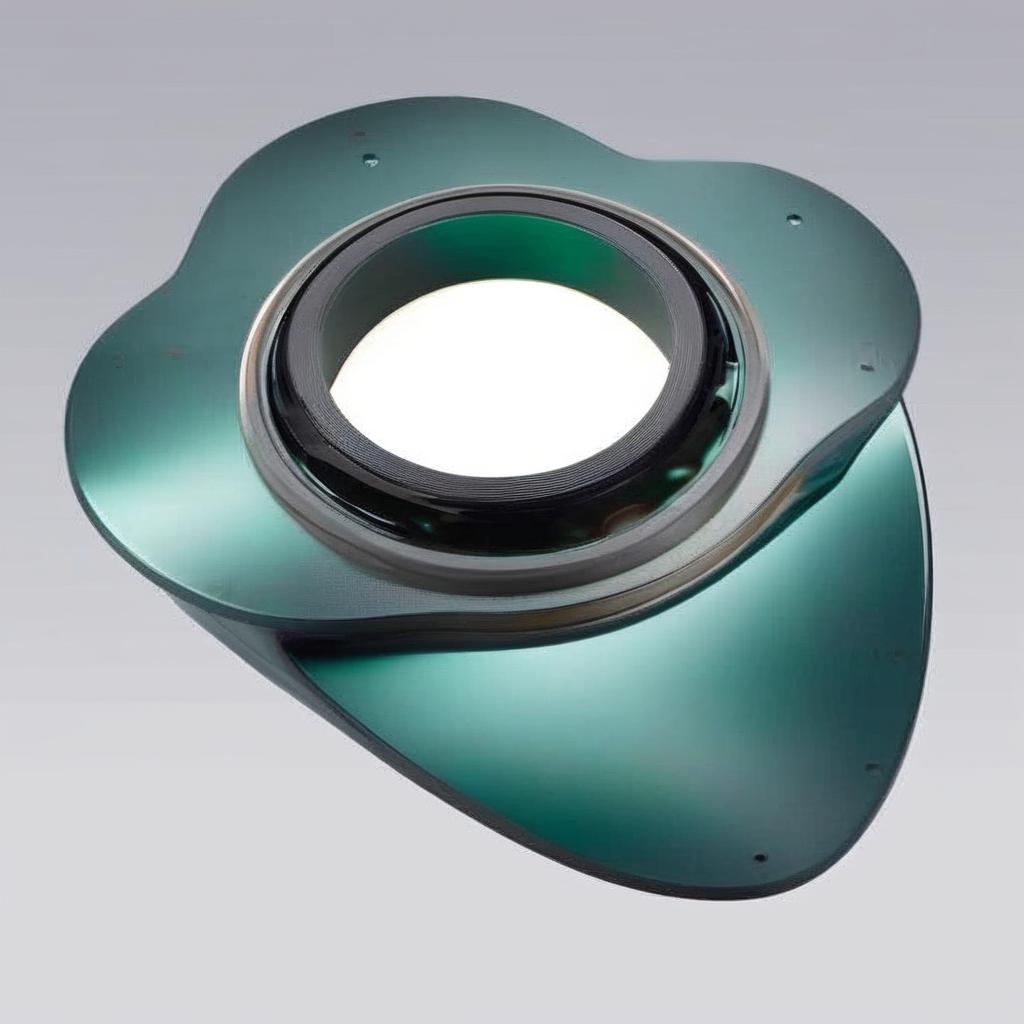} &
        \includegraphics[width=0.115\textwidth]{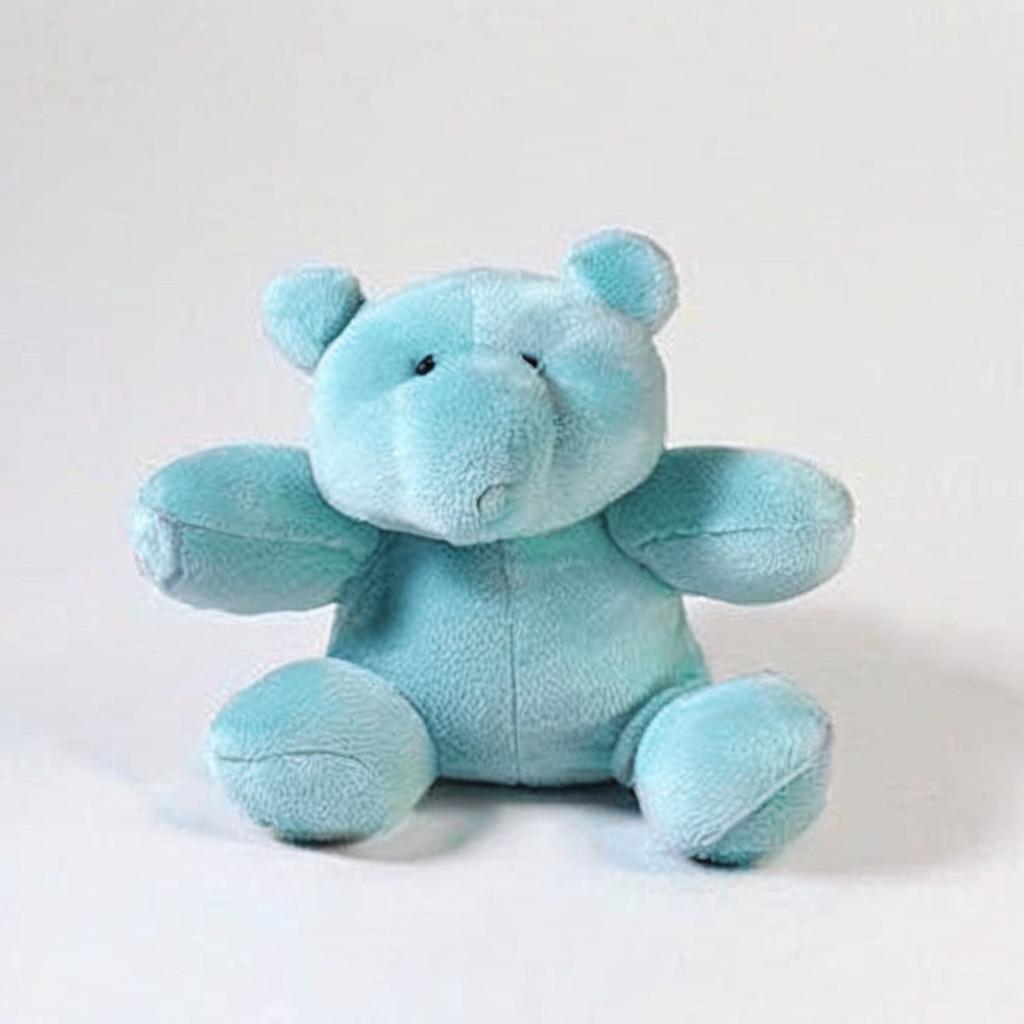} &
        \includegraphics[width=0.115\textwidth]{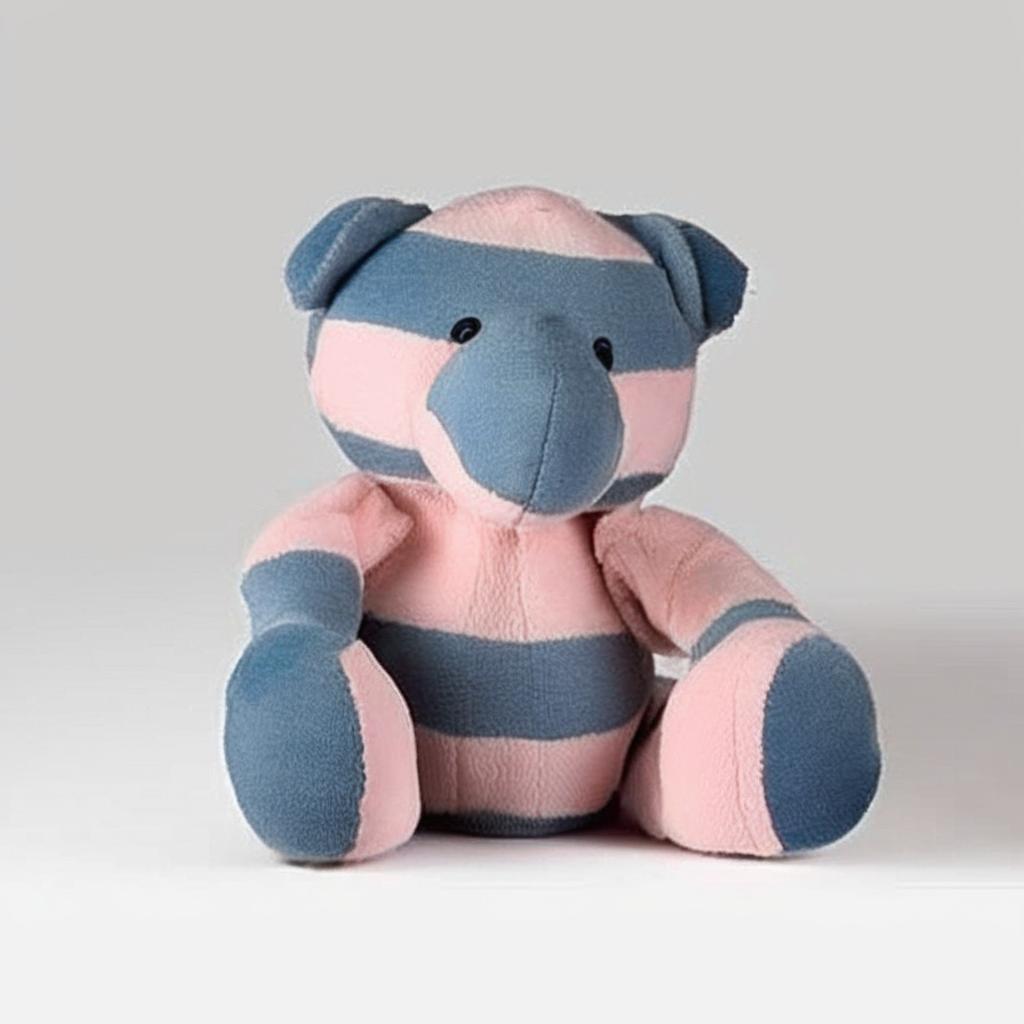} &
        \includegraphics[width=0.115\textwidth]{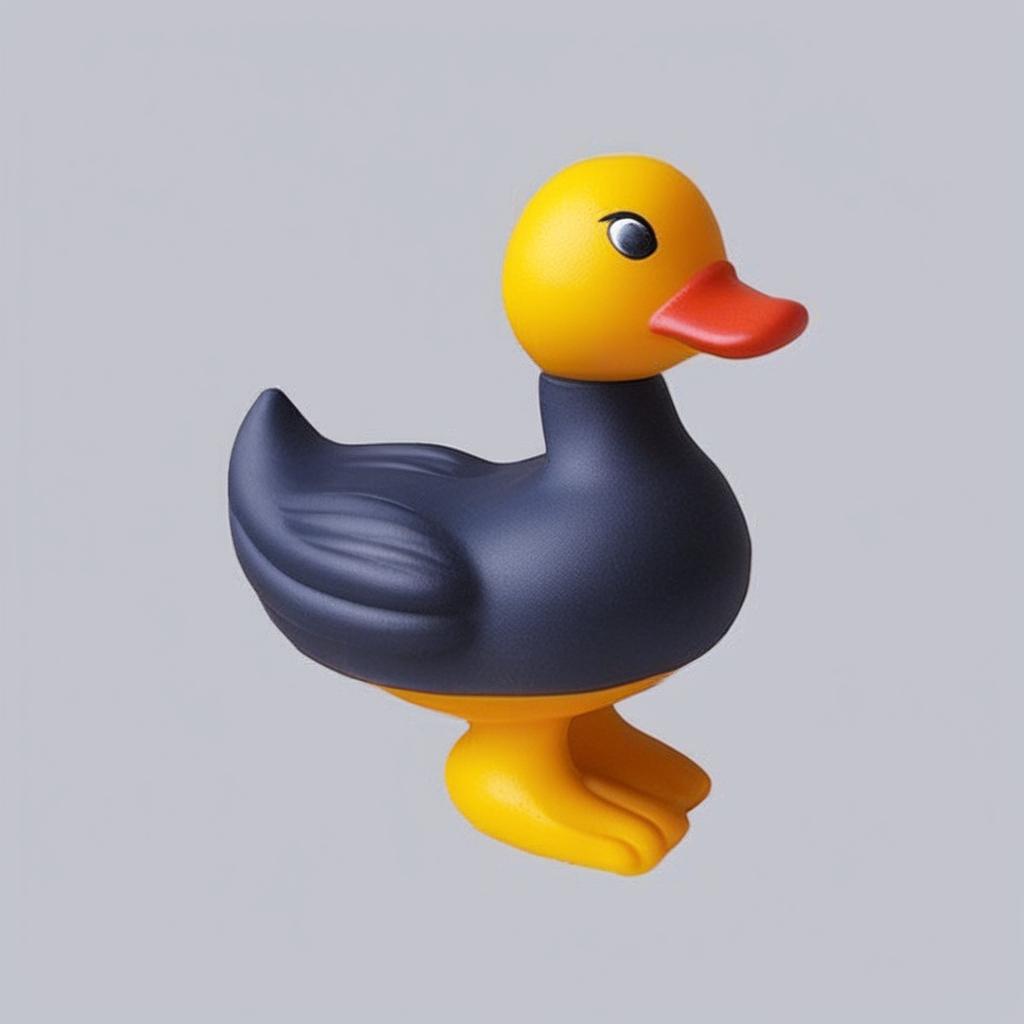} &
        \includegraphics[width=0.115\textwidth]{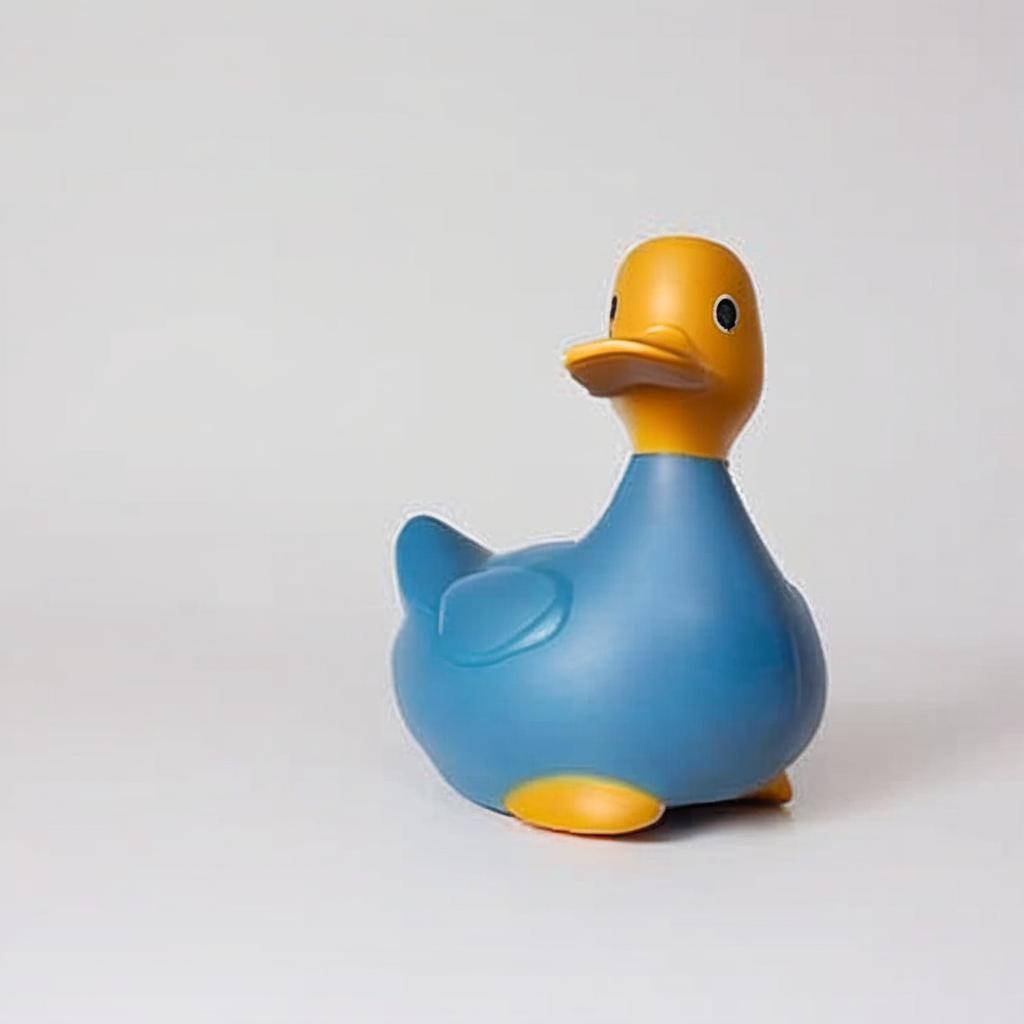} \\

        \raisebox{0.045\linewidth}{\rotatebox[origin=t]{90}{\begin{tabular}{c@{}c@{}c@{}c@{}} $\lambda$-ECLIPSE \end{tabular}}} &
        \includegraphics[width=0.115\textwidth]{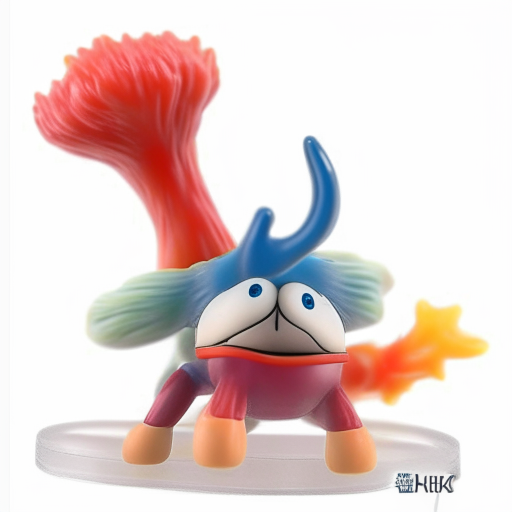} &
        \includegraphics[width=0.115\textwidth]{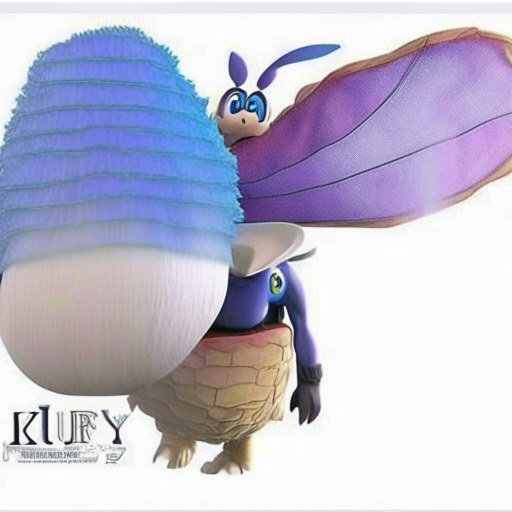} &
        \includegraphics[width=0.115\textwidth]{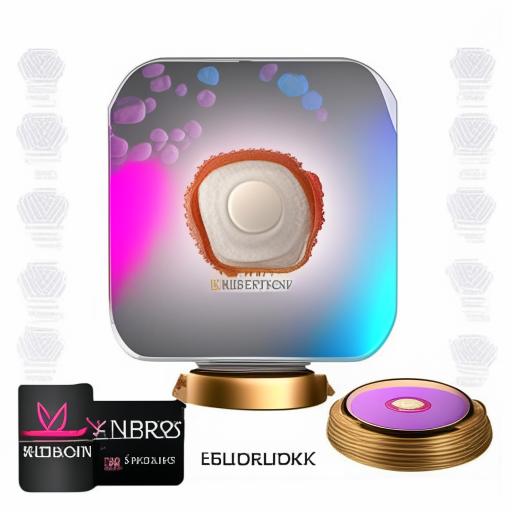} &
        \includegraphics[width=0.115\textwidth]{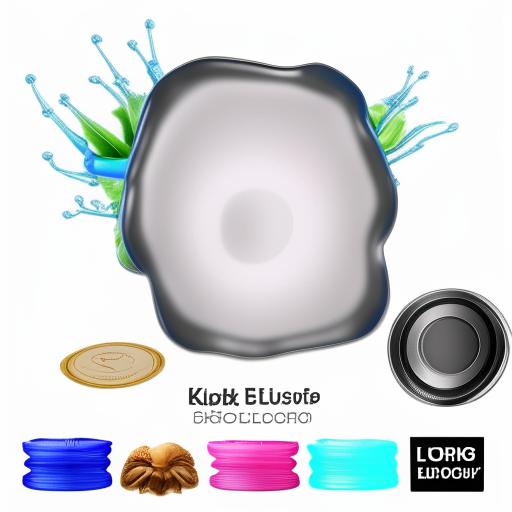} &
        \includegraphics[width=0.115\textwidth]{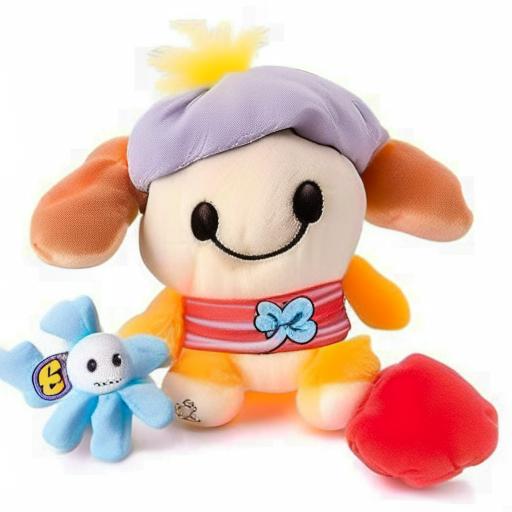} &
        \includegraphics[width=0.115\textwidth]{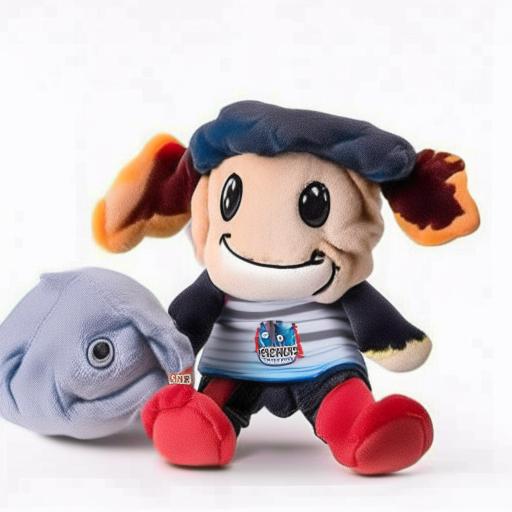} &
        \includegraphics[width=0.115\textwidth]{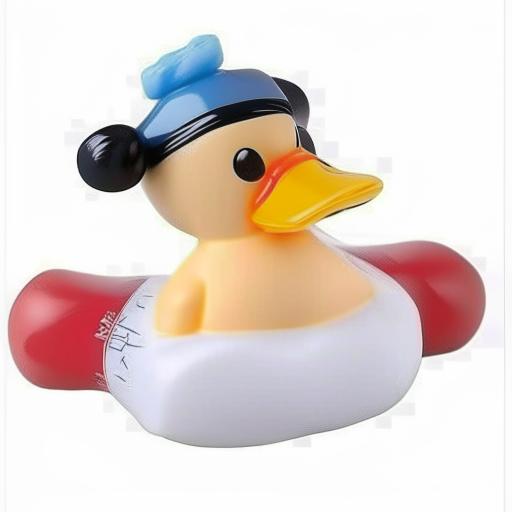} &
        \includegraphics[width=0.115\textwidth]{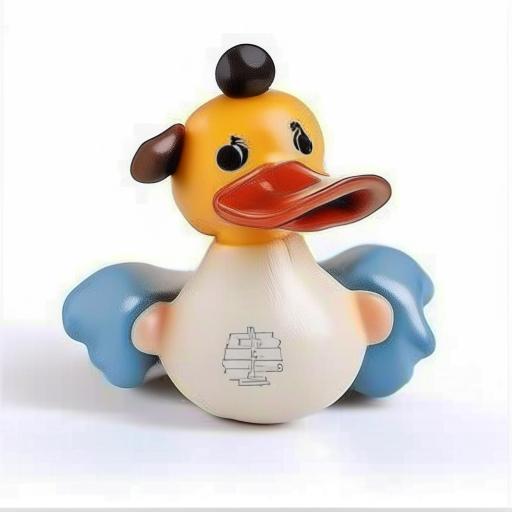} \\

        \raisebox{0.035\linewidth}{\rotatebox[origin=t]{90}{\begin{tabular}{c@{}c@{}c@{}c@{}} OmniGen \end{tabular}}} &
        \includegraphics[width=0.115\textwidth]{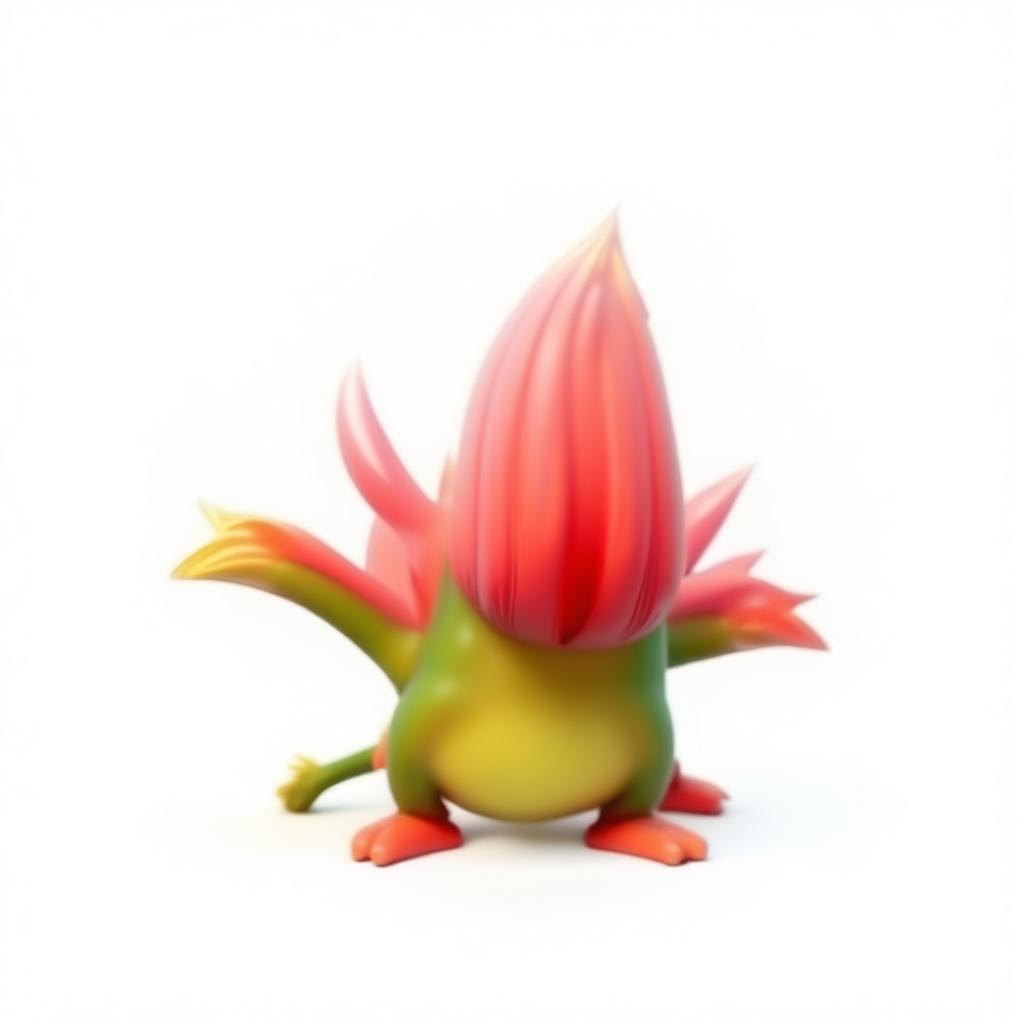} &
        \includegraphics[width=0.115\textwidth]{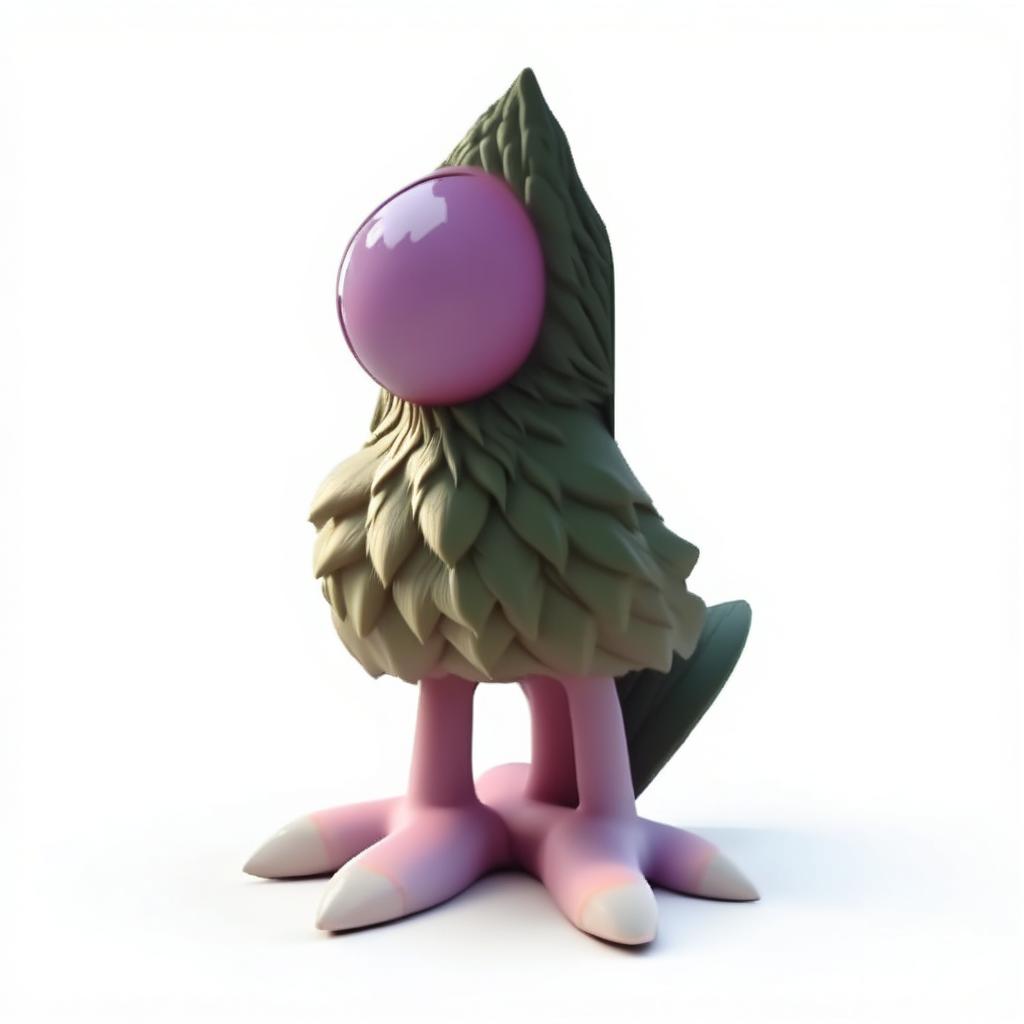} &
        \includegraphics[width=0.115\textwidth]{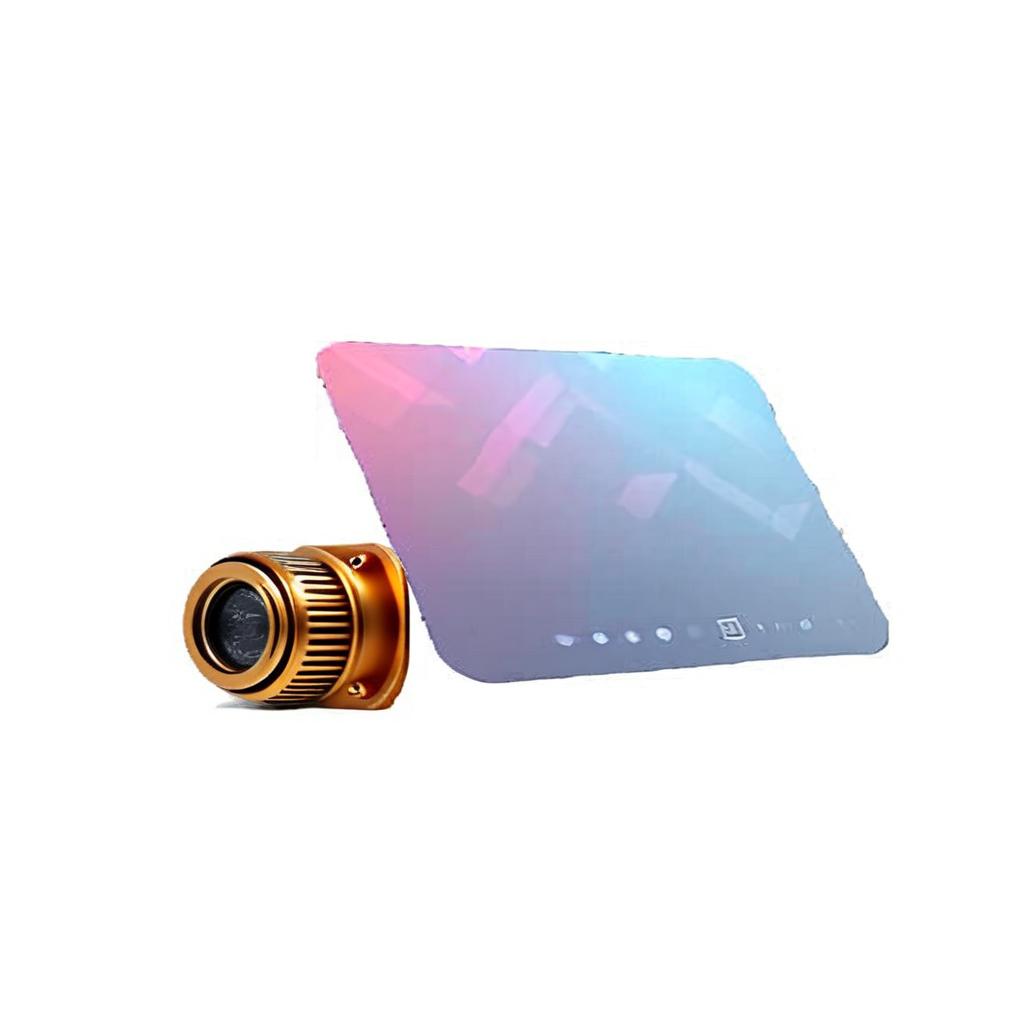} &
        \includegraphics[width=0.115\textwidth]{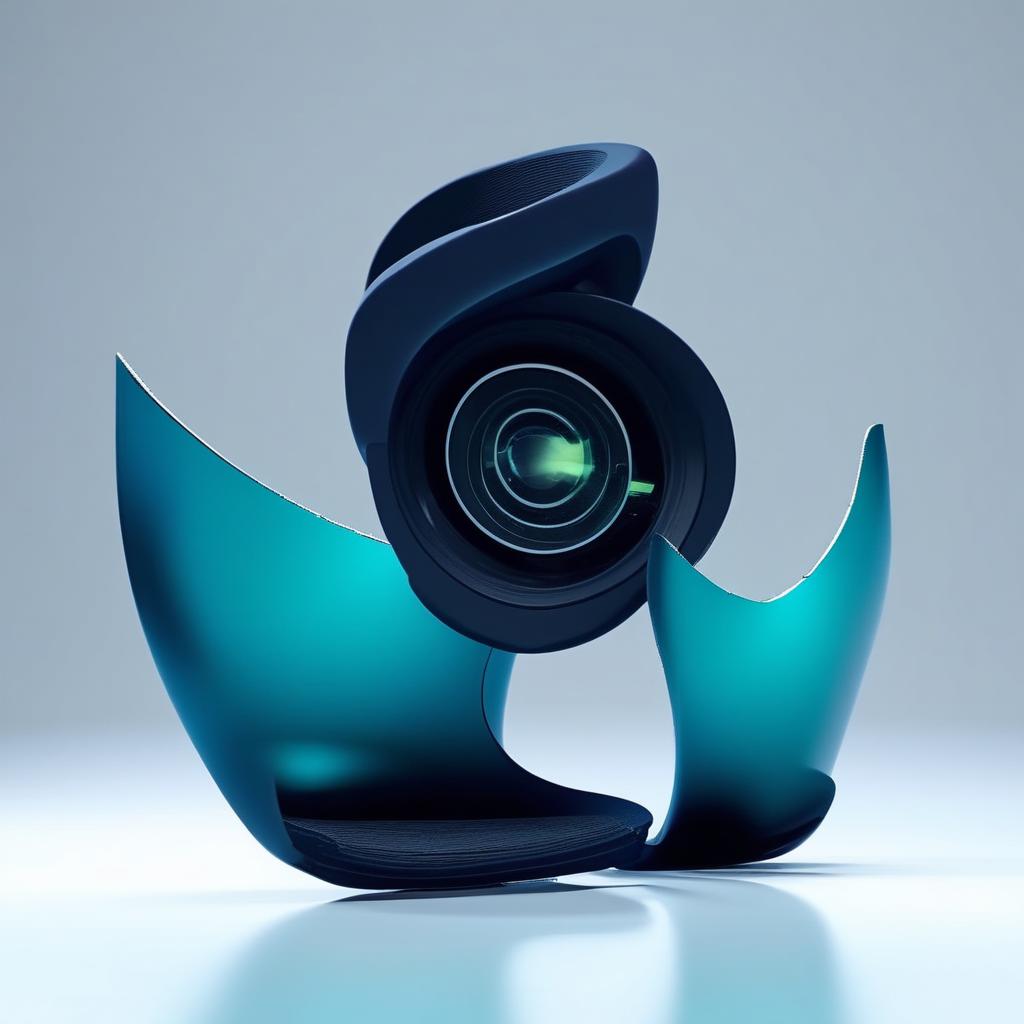} &
        \includegraphics[width=0.115\textwidth]{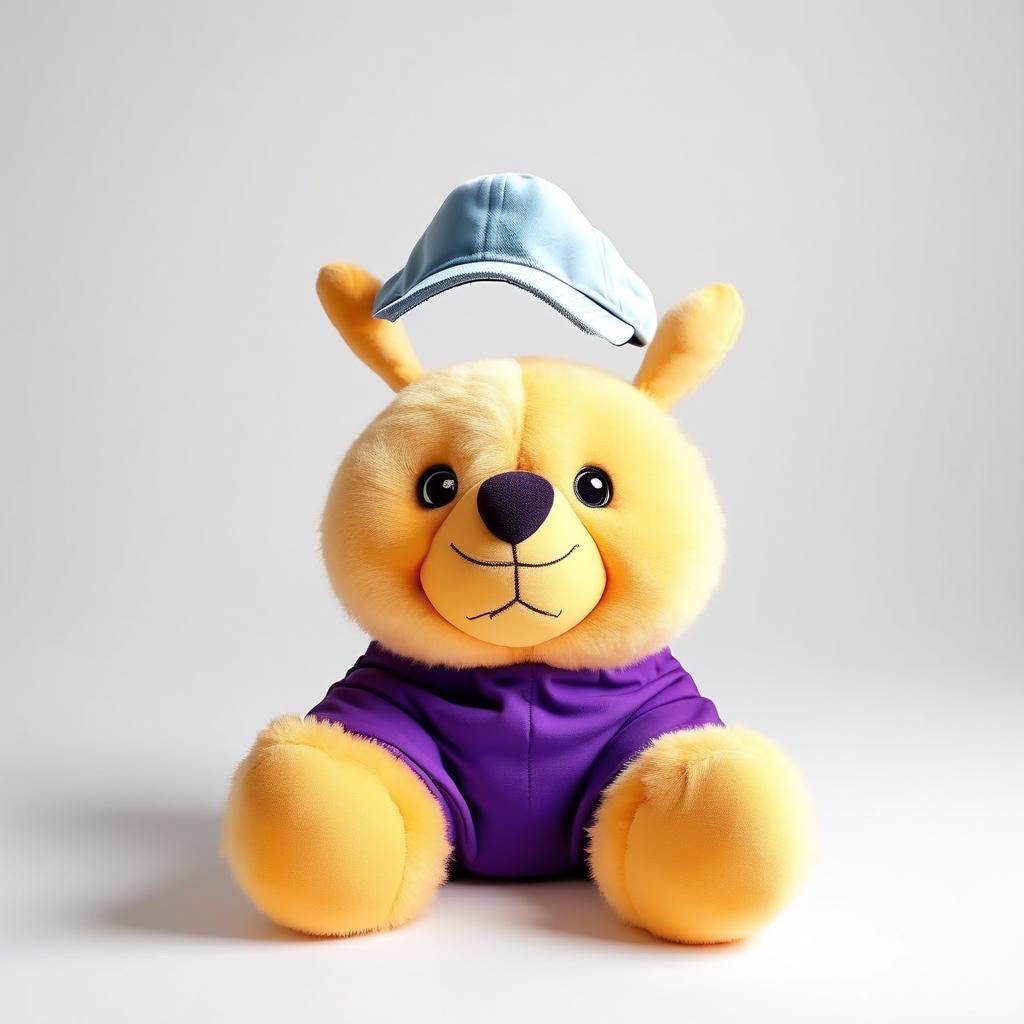} &
        \includegraphics[width=0.115\textwidth]{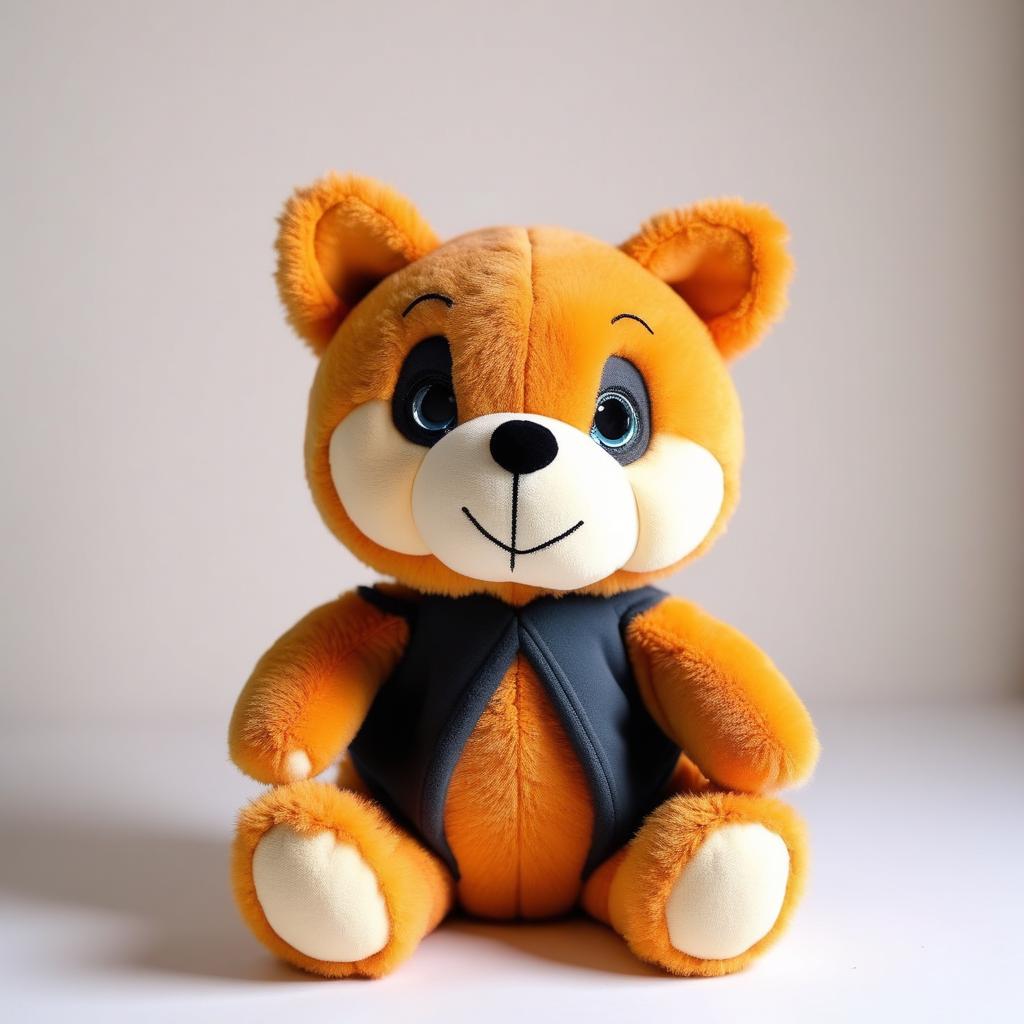} &
        \includegraphics[width=0.115\textwidth]{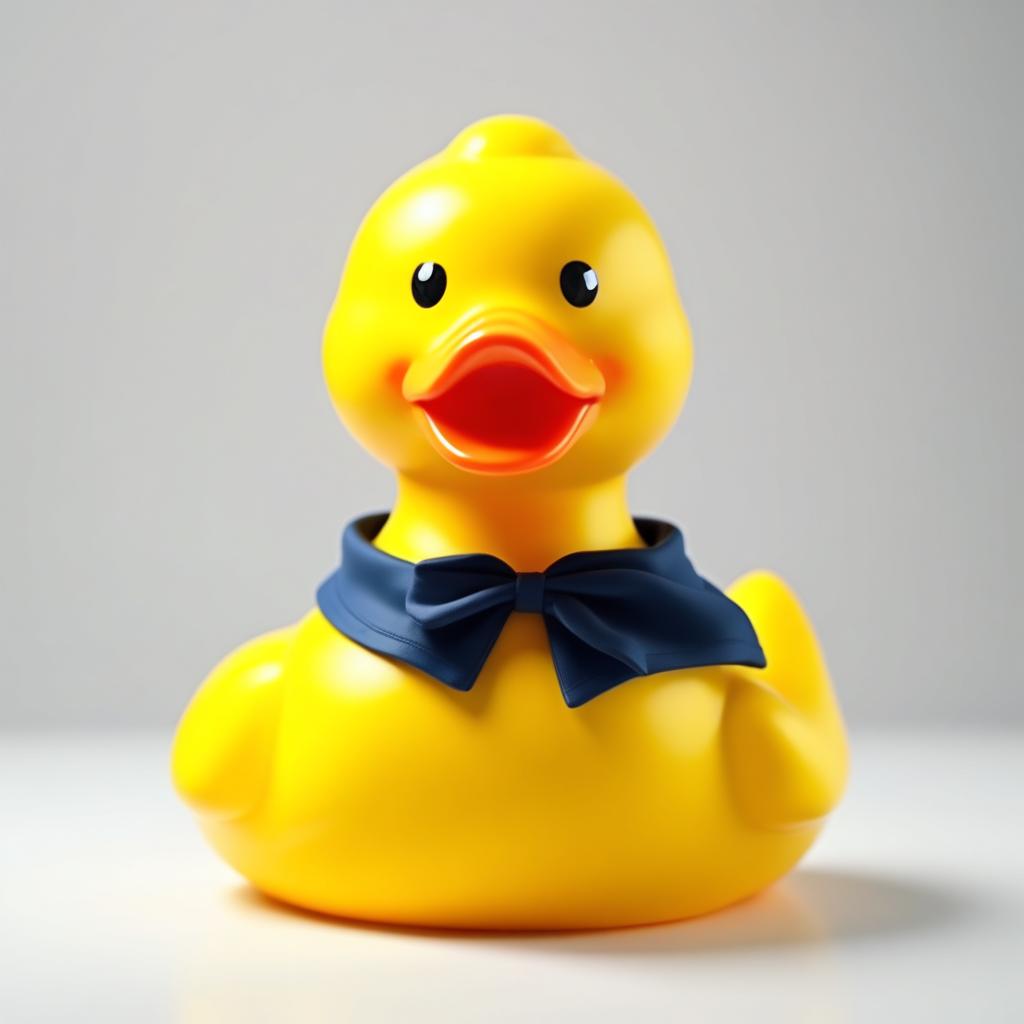} &
        \includegraphics[width=0.115\textwidth]{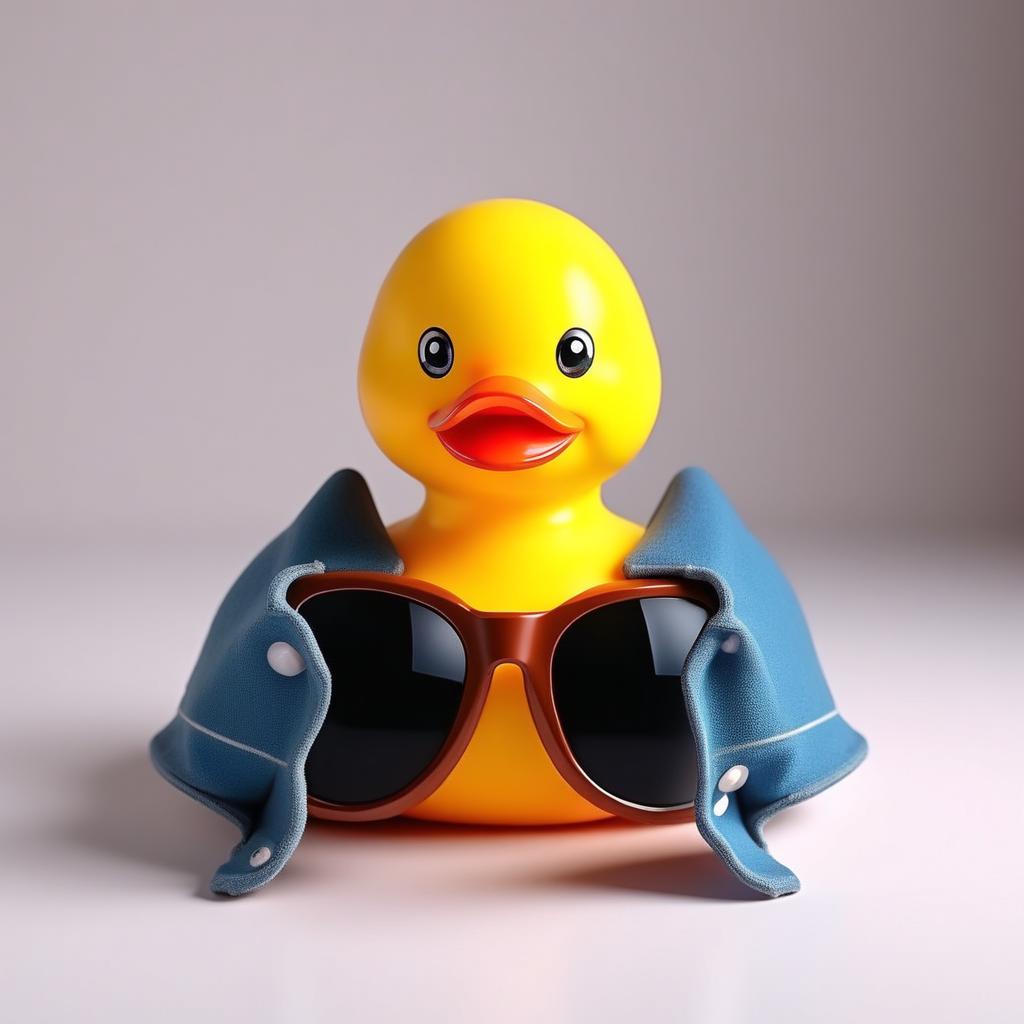}

    \end{tabular}
    }
    \vspace{-0.325cm}
    \caption{\textbf{Qualitative Comparisons}. We provide visual comparisons with alternative methods across various domains.
    }
    \vspace{-0.225cm}
    \label{fig:qualitative_comparisons}
\end{figure*}

\begin{figure}
    \centering
    \setlength{\tabcolsep}{0.5pt}
    \renewcommand{\arraystretch}{0.5}
    \addtolength{\belowcaptionskip}{-5pt}
    {\small
    \begin{tabular}{c c c c}

        \includegraphics[height=0.105\textwidth]{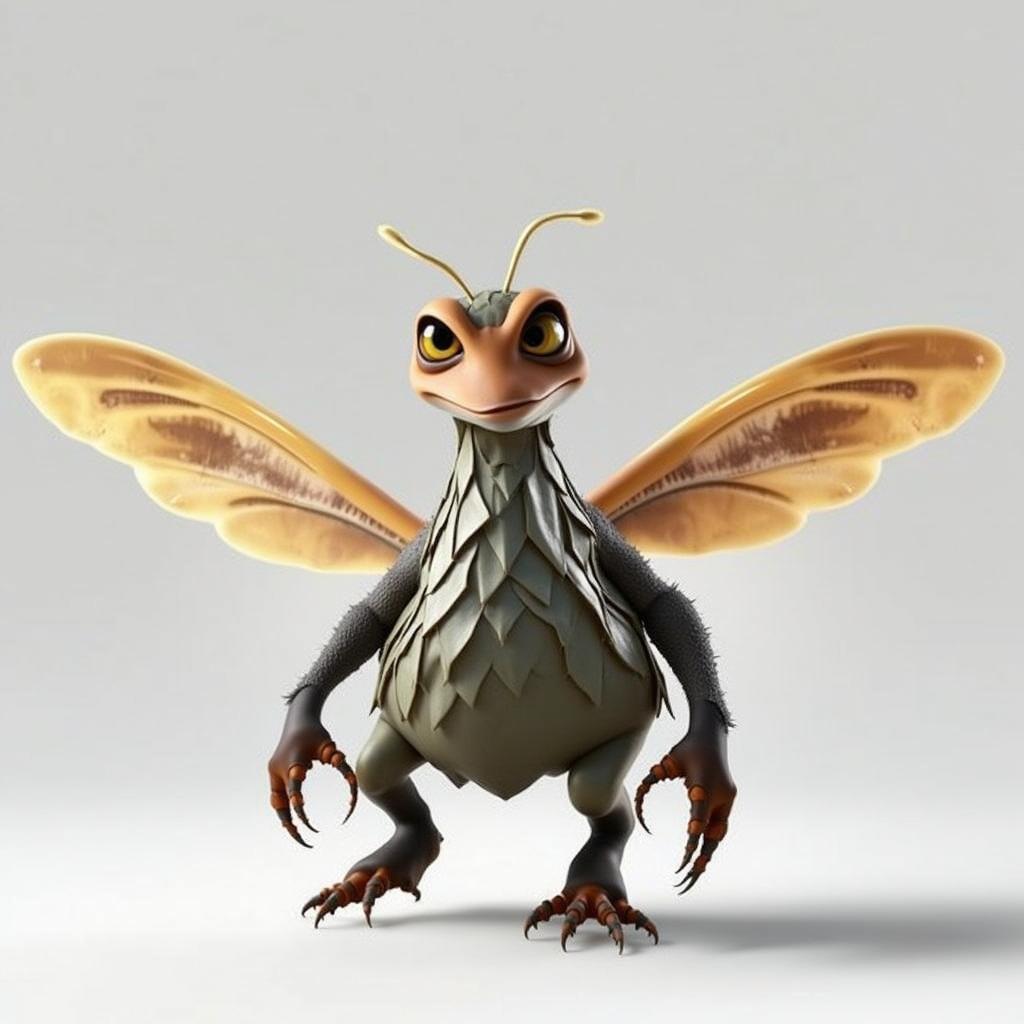} &
        \includegraphics[height=0.105\textwidth]{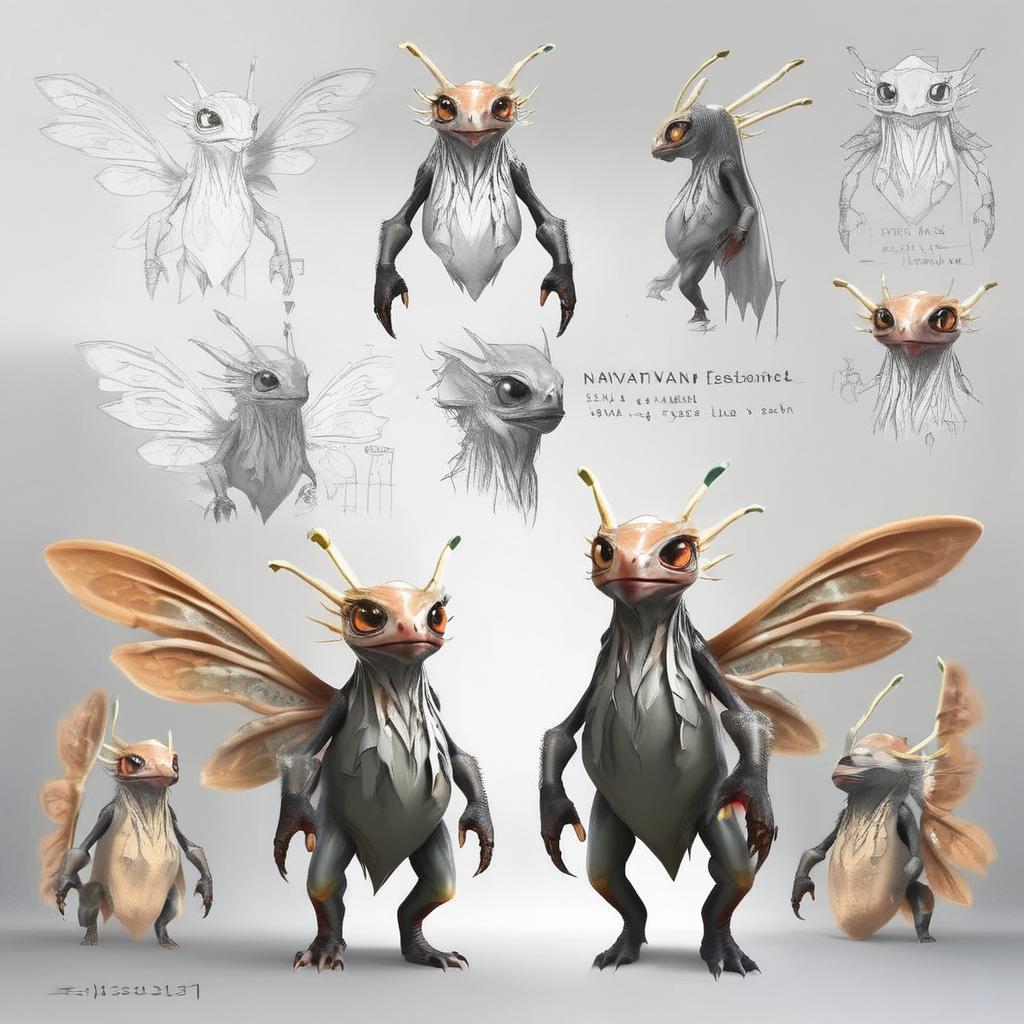} &
        \includegraphics[height=0.105\textwidth]{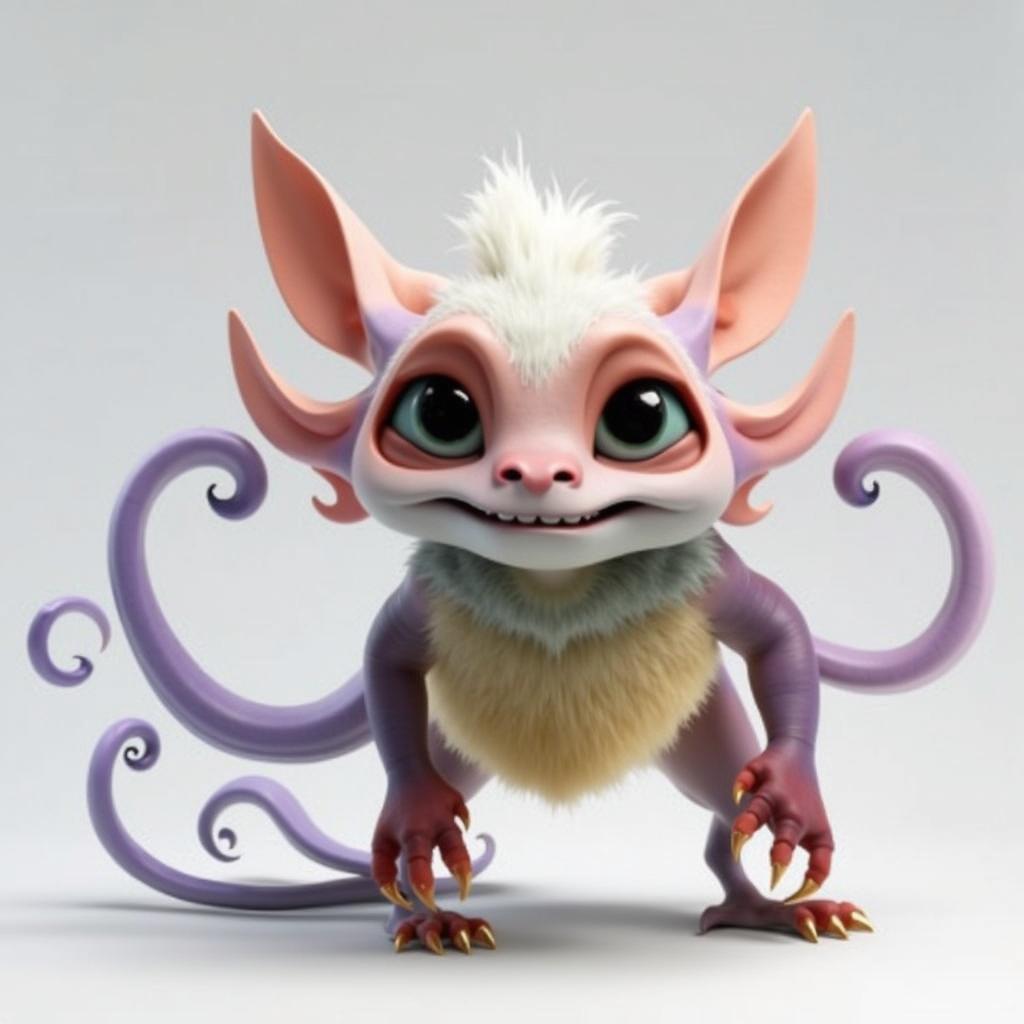} &
        \includegraphics[height=0.105\textwidth]{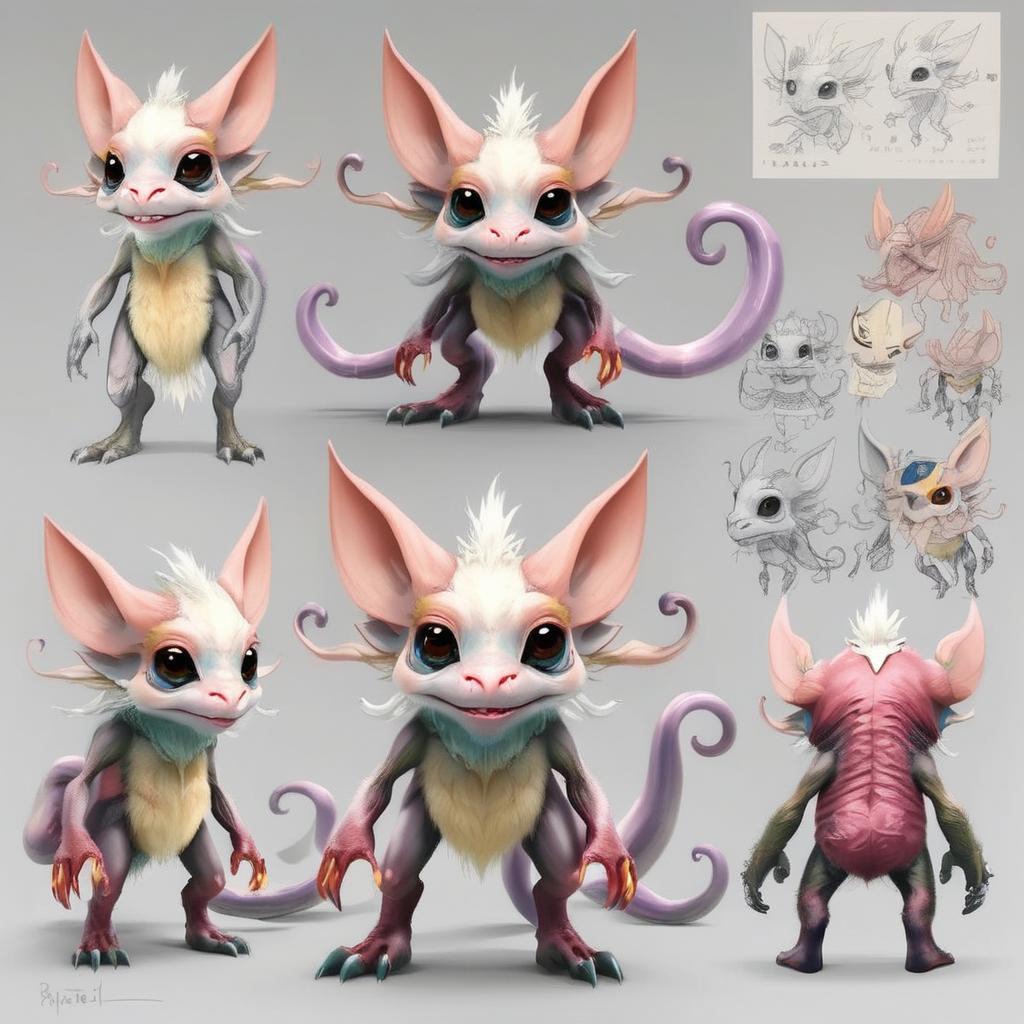} \\
        \includegraphics[height=0.105\textwidth]{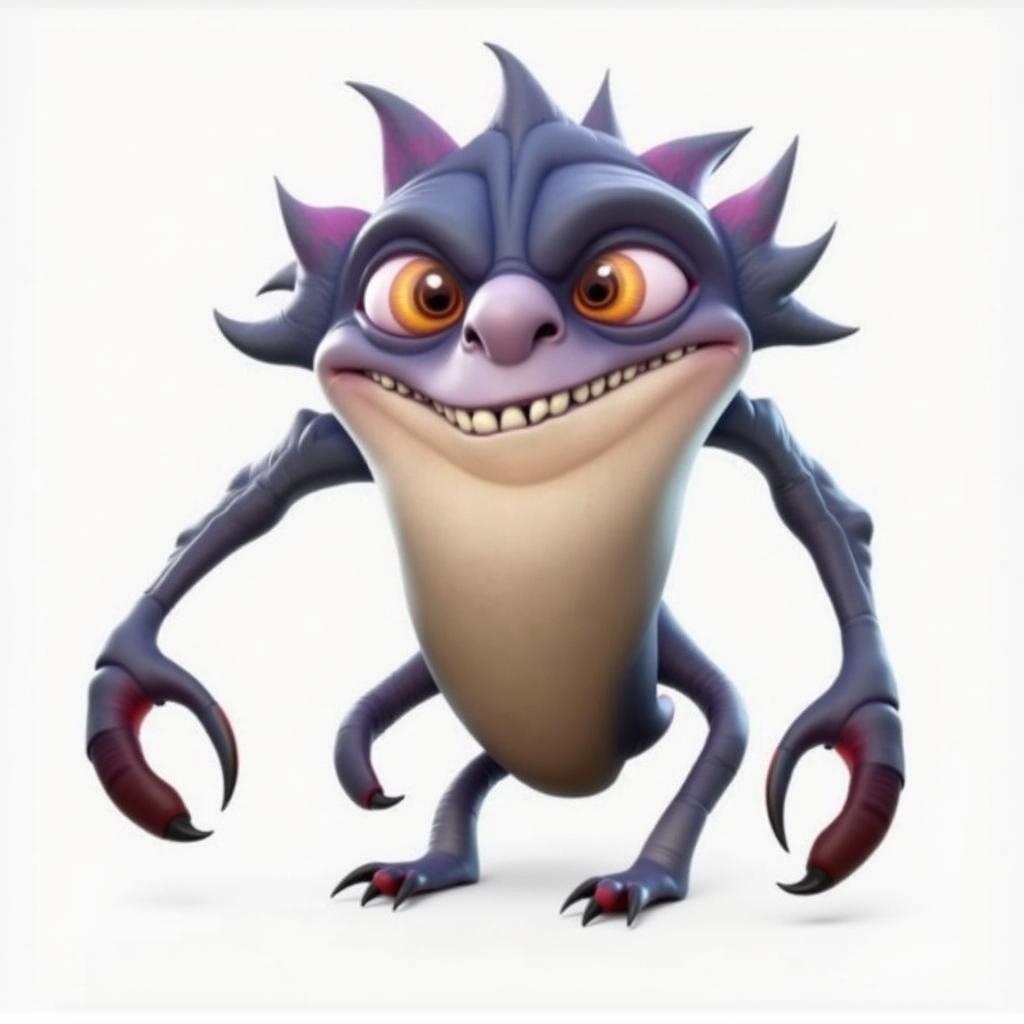} &
        \includegraphics[height=0.105\textwidth]{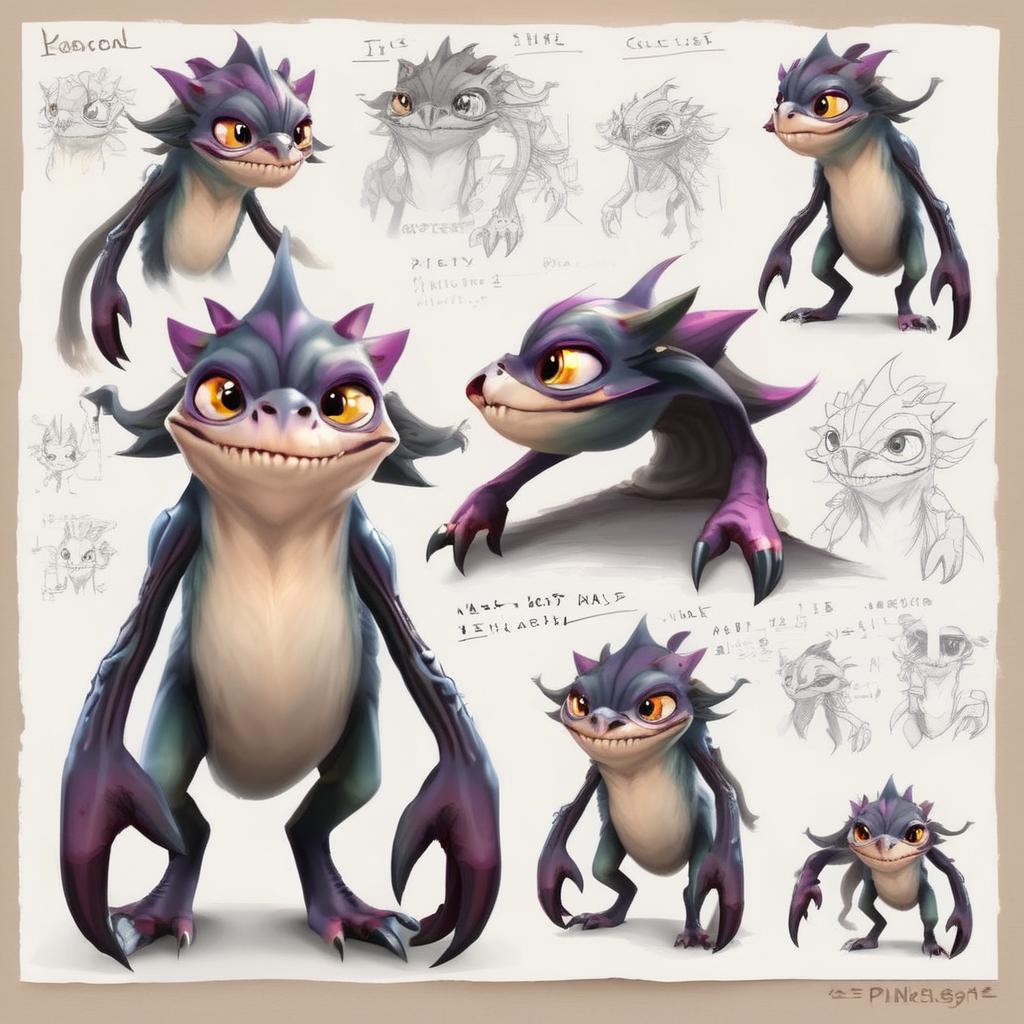} &
        \includegraphics[height=0.105\textwidth]{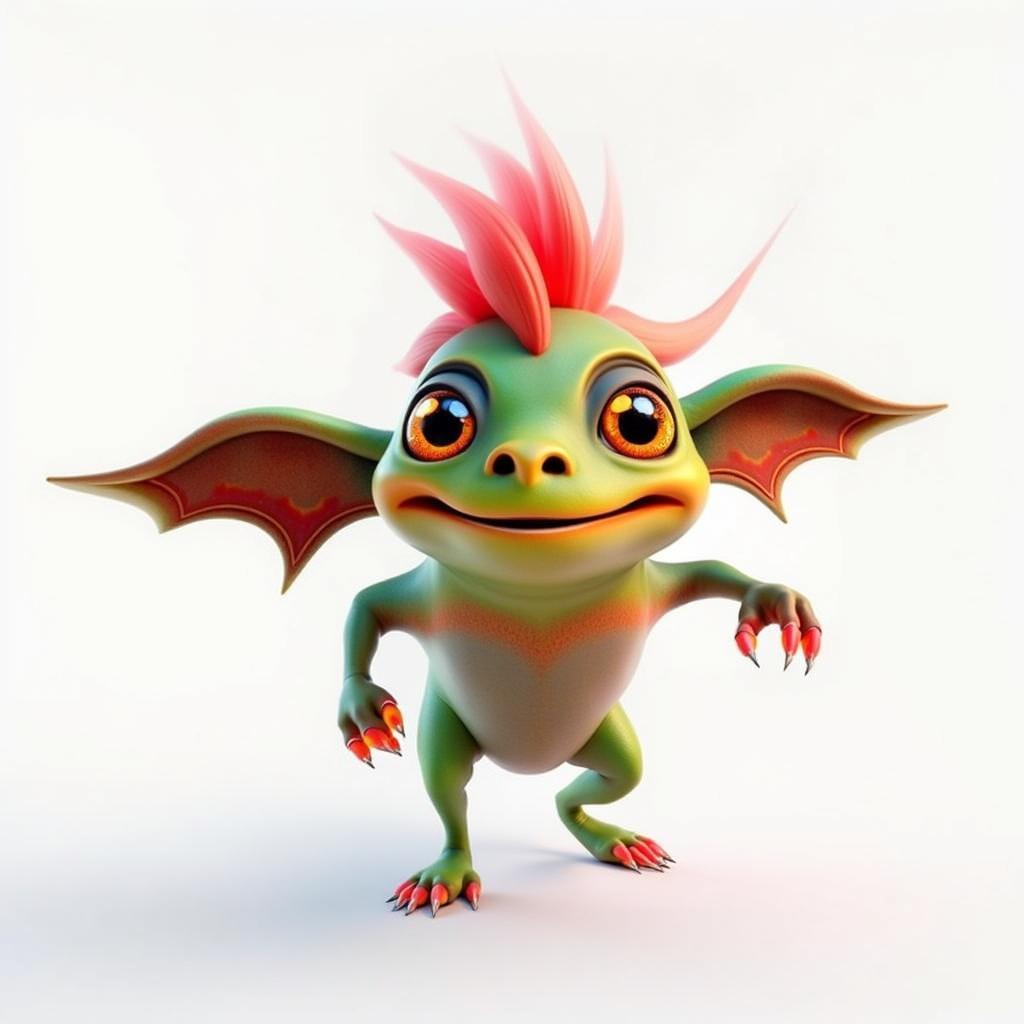} &
        \includegraphics[height=0.105\textwidth]{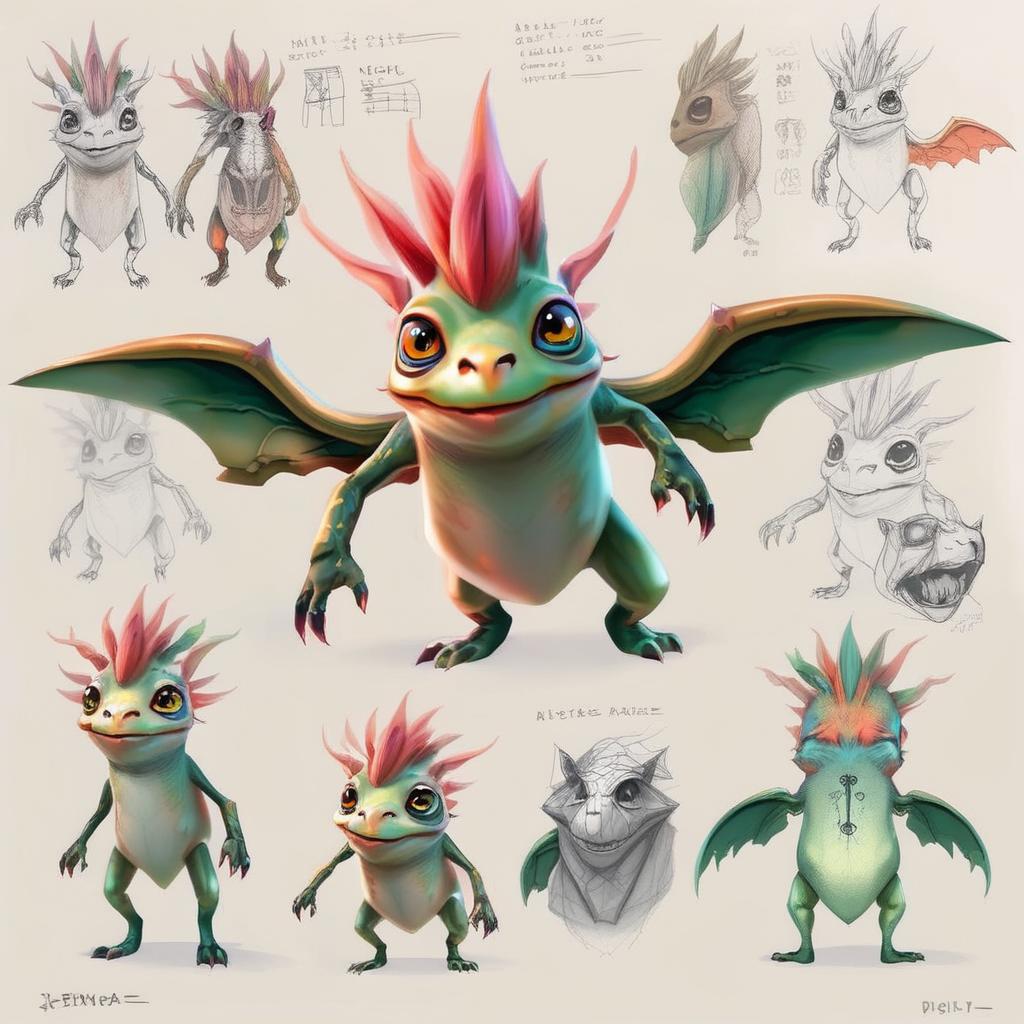}

    \end{tabular}
    }
    \vspace{-0.3cm}
    \caption{\textbf{Style Generation.} We train a LoRA to generate character reference sheets when given a concept embedding. 
    }
    \vspace{-0.2cm}
    \label{fig:ref_paired_results}
\end{figure}

\subsection{IP-LoRA Results}\label{sec:ip_lora}
In the previous section, all results were rendered as-is using the original IP-Adapter+ over SDXL. As discussed, a desirable property of our approach is the ability to take a generated concept and integrate it into different scenes or styles, enabling a continuous and flexible creative process.

\begin{table}
\small
\centering
\setlength{\tabcolsep}{2pt}
\caption{\textbf{Quantitative Comparison.} We show the average text and visual scores on a scale of $1$-$5$, computed using Qwen~\cite{yang2024qwen2}.\\[-0.65cm]} 
\begin{tabular}{l | c c c c c} 
    \toprule
    & PiT & IP-A (0.2) & IP-A (0.4) & IP-A (0.6) & IP-A (0.8) \\
    \midrule
    Text Score  $\uparrow$ & $3.60$ & $3.70$ & $2.45$ & $1.37$ & $1.06$ \\
    Visual Score $\uparrow$ & $4.55$ & $2.41$ & $3.95$ & $4.64$ & $4.84$  \\
    \bottomrule
\end{tabular}
\vspace{-0.25cm}
\label{tb:ip_lora_quant}
\end{table}

\vspace{-0.2cm}
\paragraph{Text-Conditioning.}
We first present results using an IP-LoRA trained to restore text conditioning for describing a background in which we aim to place our generated concept, as shown in~\Cref{fig:lora_vs_baseline}. 
In this setup, SDXL is conditioned both on the standard text prompt and on concepts generated using our method, which are passed as encodings to the adapter. 
As shown, our method effectively balances the text conditioning while remaining faithful to the concept.
For comparison, we include the original behavior of IP-Adapter+. When using strong adapter scales (e.g., $0.6$), the model completely disregards the text and simply reconstructs the given image embedding. Lowering the scaling value allows the model to incorporate the textual information but significantly degrades reconstruction quality. 

We now validate this claim quantitatively. A common metric for measuring text and image similarity is CLIP-space similarity~\cite{hessel2022clipscore}, but it is unsuitable for our evaluation, as we aim to preserve information beyond what CLIP embeddings can capture. Instead, we use Qwen 2~\cite{yang2024qwen2}. Specifically, we prompt the VLM to rate each image on a scale of 1 to 5 based on its adherence to the given text and its similarity to the reference image. The results, presented in~\Cref{tb:ip_lora_quant}, align with the visual observation, demonstrating the effectiveness of the IP-LoRA approach.

\vspace{-0.3cm}
\paragraph{Styled Generations.}
Another common use case for model fine-tuning is personalization to specific styles or domains~\cite{ruiz2022dreambooth, gal2022image}. 
Similarly, in~\Cref{fig:ref_paired_results}, we present a LoRA trained to generate character reference sheets when conditioned on the concept embedding of a given character, where paired data is obtained by extracting a sample from the reference sheet and using its embedding as conditioning.

\subsection{Comparisons}
Given the nature of our task, there is no direct method that tackles the same task. Still, to offer an analysis of PiT with respect to the state-of-the-art, we compare it to a representative set of multi-image baselines, testing their generalization. 
More specifically, we consider OmniGen~\cite{xiao2024omnigen}, a strong multi-modal model,
$\lambda$-ECLIPSE~\cite{patel2024lambdaeclipse} as a prior-based technique for multi-image compositions, 
and IP-Adapter+, modified to operate over multiple images by aggregating the generated embeddings.

We present results in~\Cref{fig:qualitative_comparisons} across multiple concept domains. As expected, the naive approach of averaging the generated embeddings and using IP-Adapter+ blends the input components together, producing a hybrid of their visual characteristics rather than preserving distinct parts. Similarly, $\lambda$-ECLIPSE may capture the target domain (e.g., ducks) and the color palette of the input components, but it fails to integrate the specified parts into the output. Finally, OmniGen, exhibits inconsistent behavior across domains: in some cases, it omits certain parts, while in others, it strictly preserves their spatial structure, preventing it from semantically assembling the parts together, as shown in the third column. In contrast, our approach effectively incorporates the provided parts while naturally completing the missing information based on the target domain.

\subsection{Additional Inputs}
\begin{figure}
    \centering
    \setlength{\tabcolsep}{0.5pt}
    \renewcommand{\arraystretch}{0.5}
    {\small
    \begin{tabular}{c c @{\hspace{0.3cm}} c c}

        \includegraphics[height=0.1\textheight]{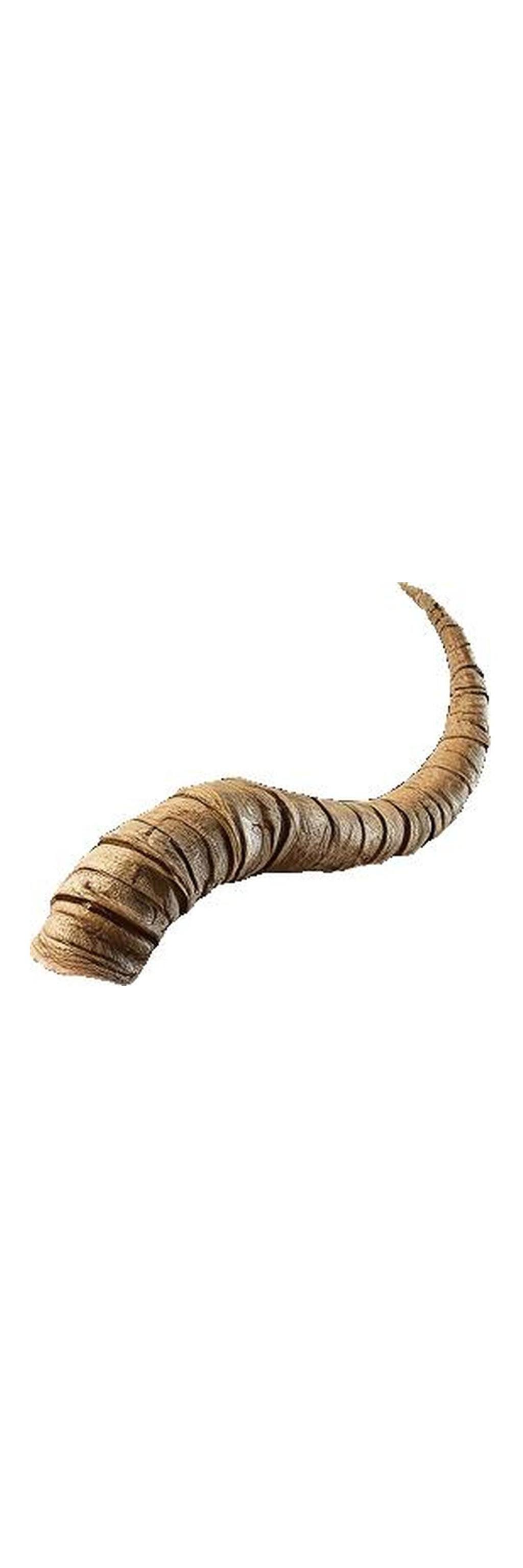} &
        \includegraphics[height=0.1\textheight]{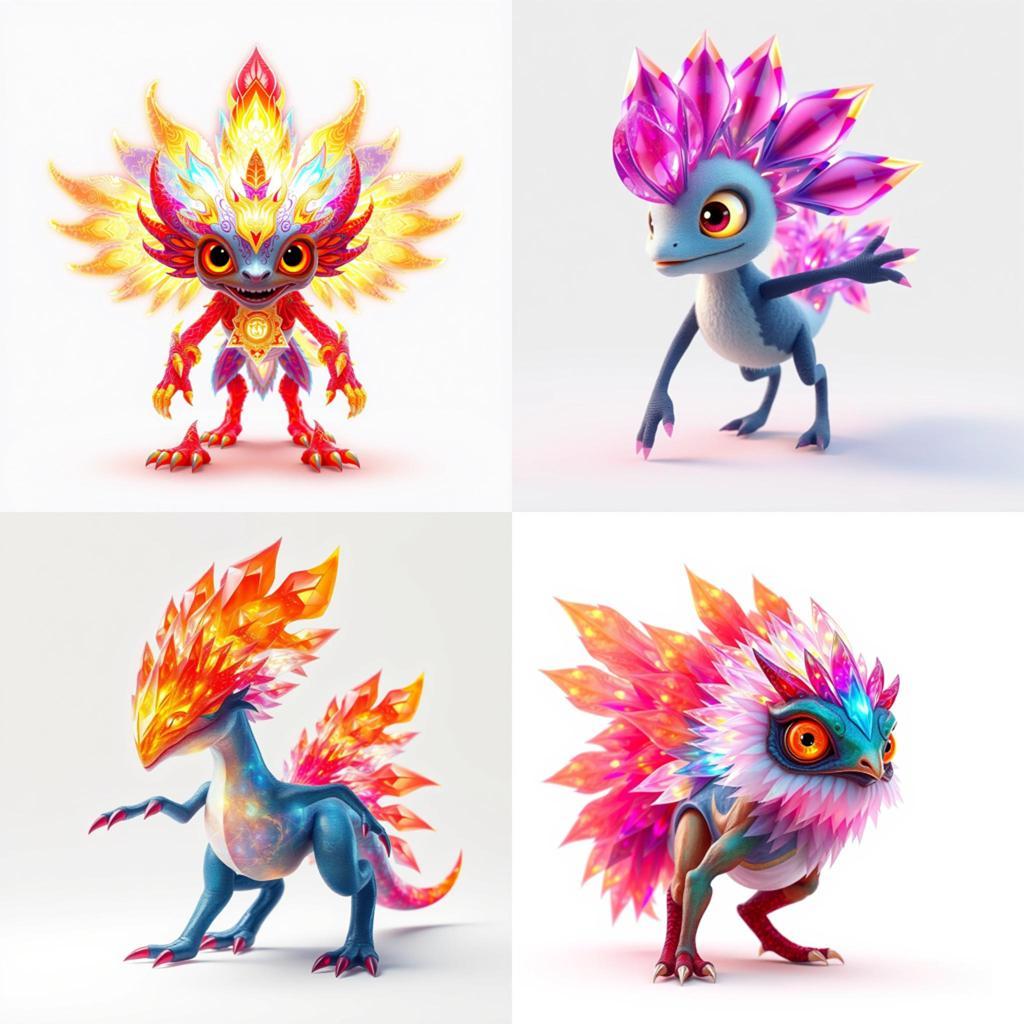} &        
        \includegraphics[height=0.1\textheight]{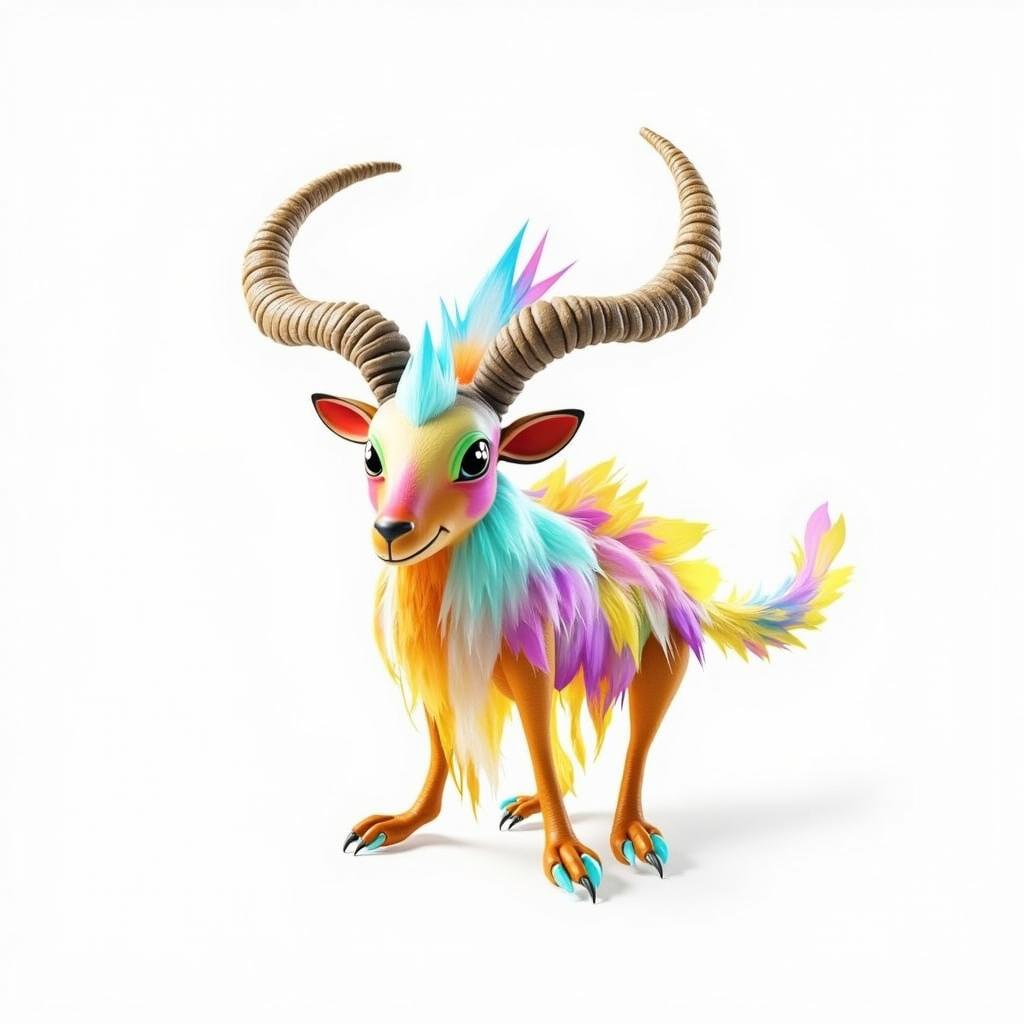} &
        \includegraphics[height=0.1\textheight]{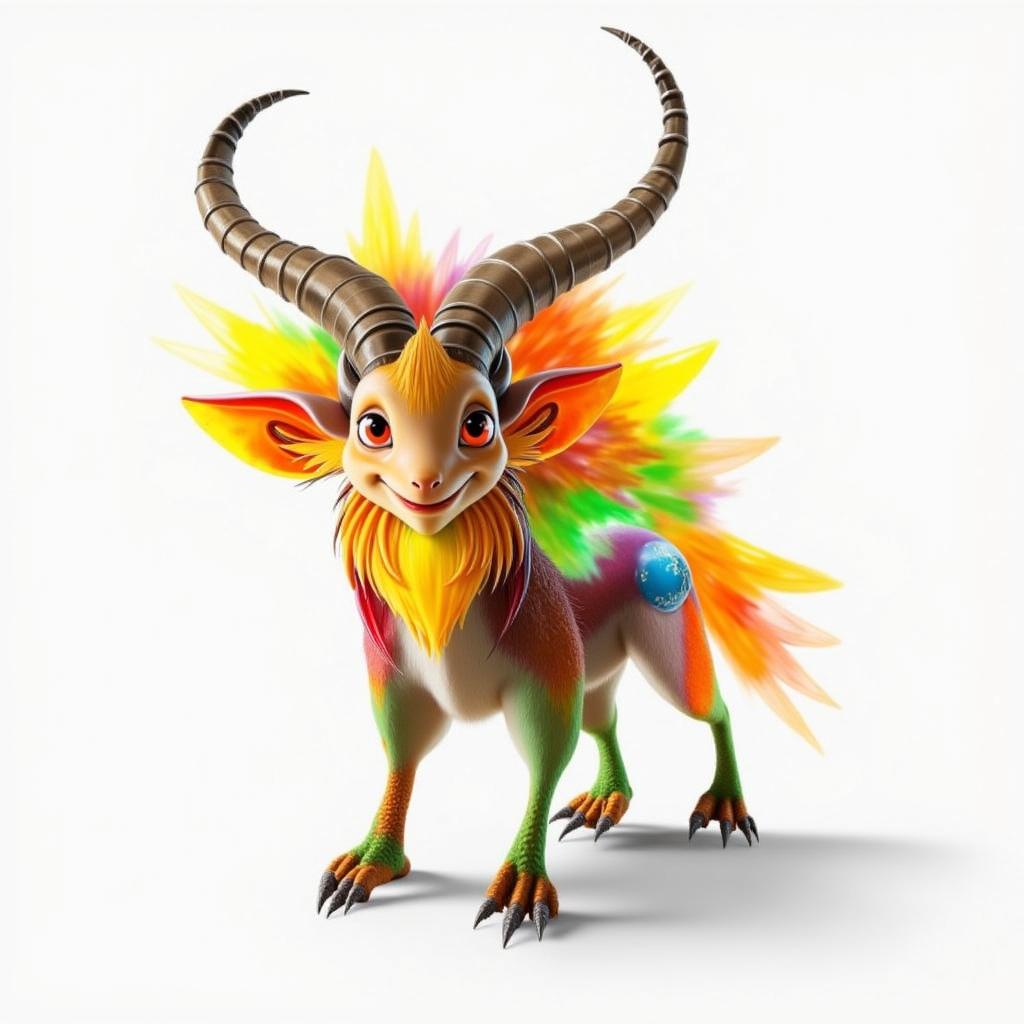} 
        \\        
        \includegraphics[height=0.1\textheight]{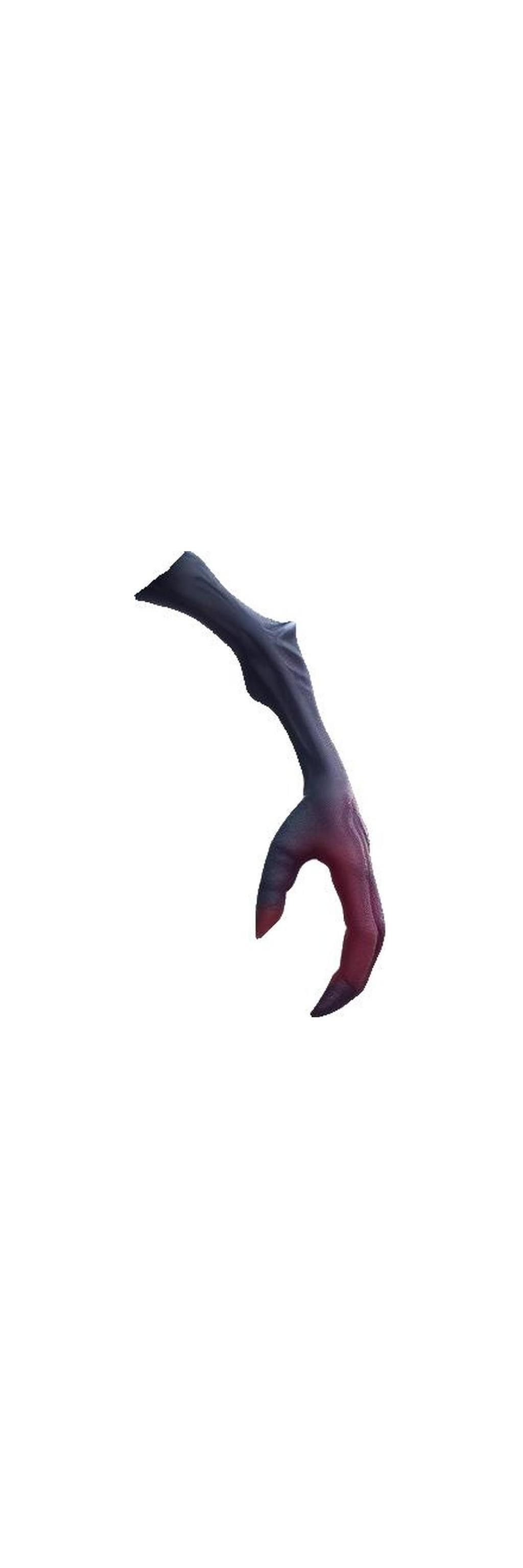} &
        
        \includegraphics[height=0.1\textheight]{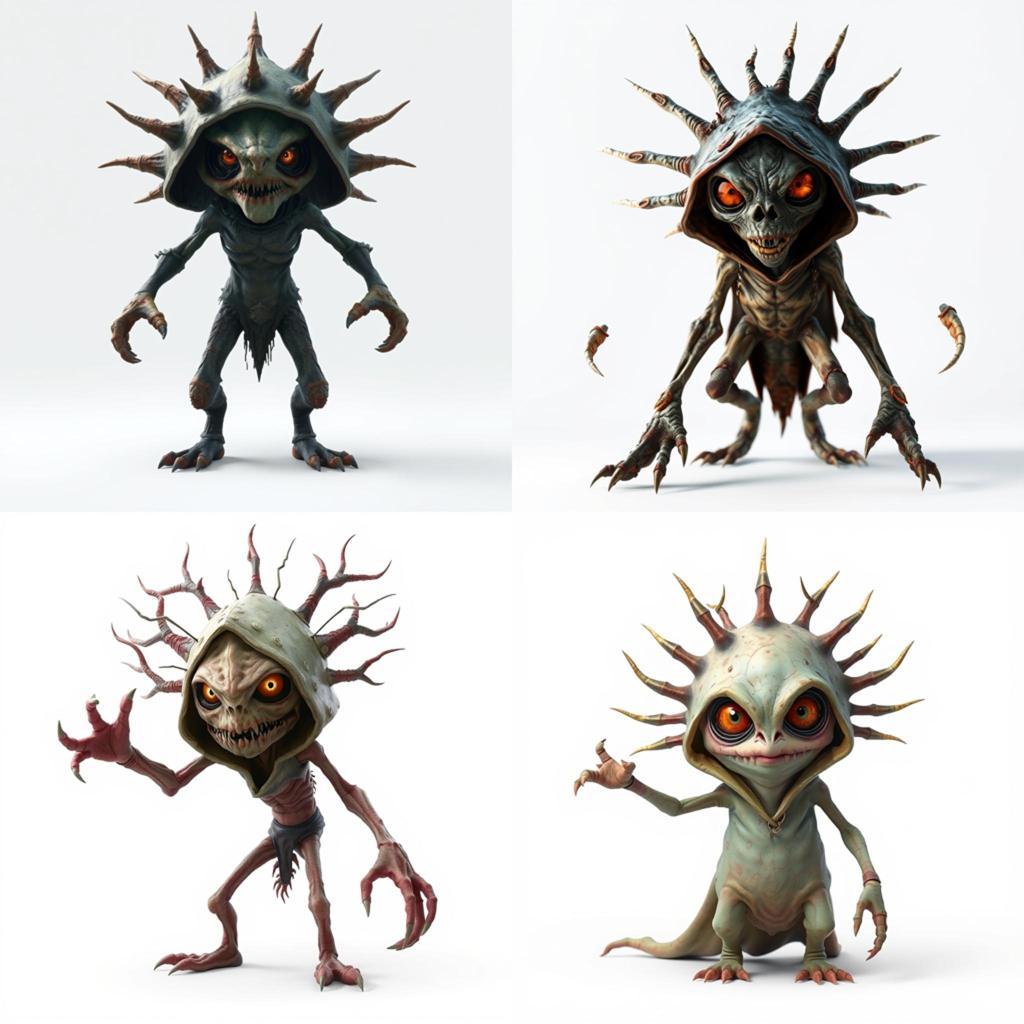} &        
        \includegraphics[height=0.1\textheight]{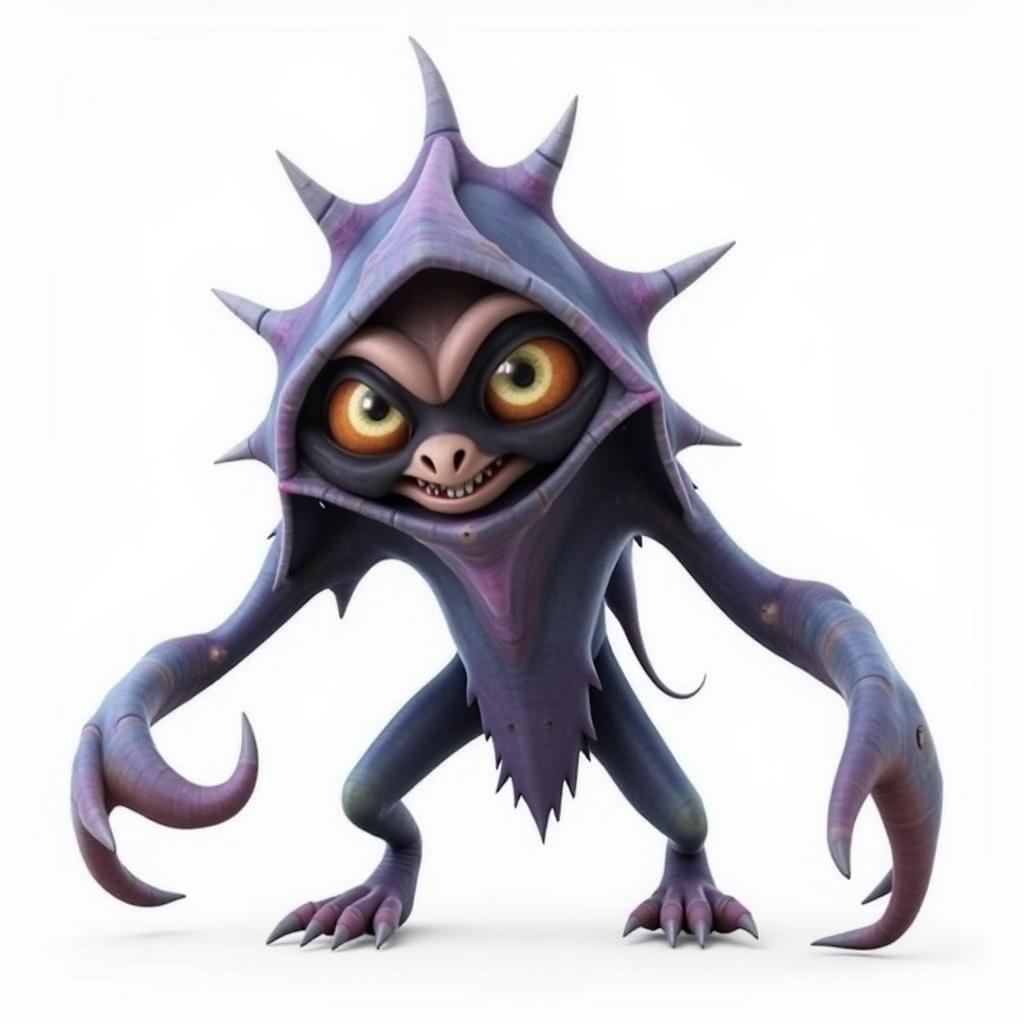} &
        \includegraphics[height=0.1\textheight]{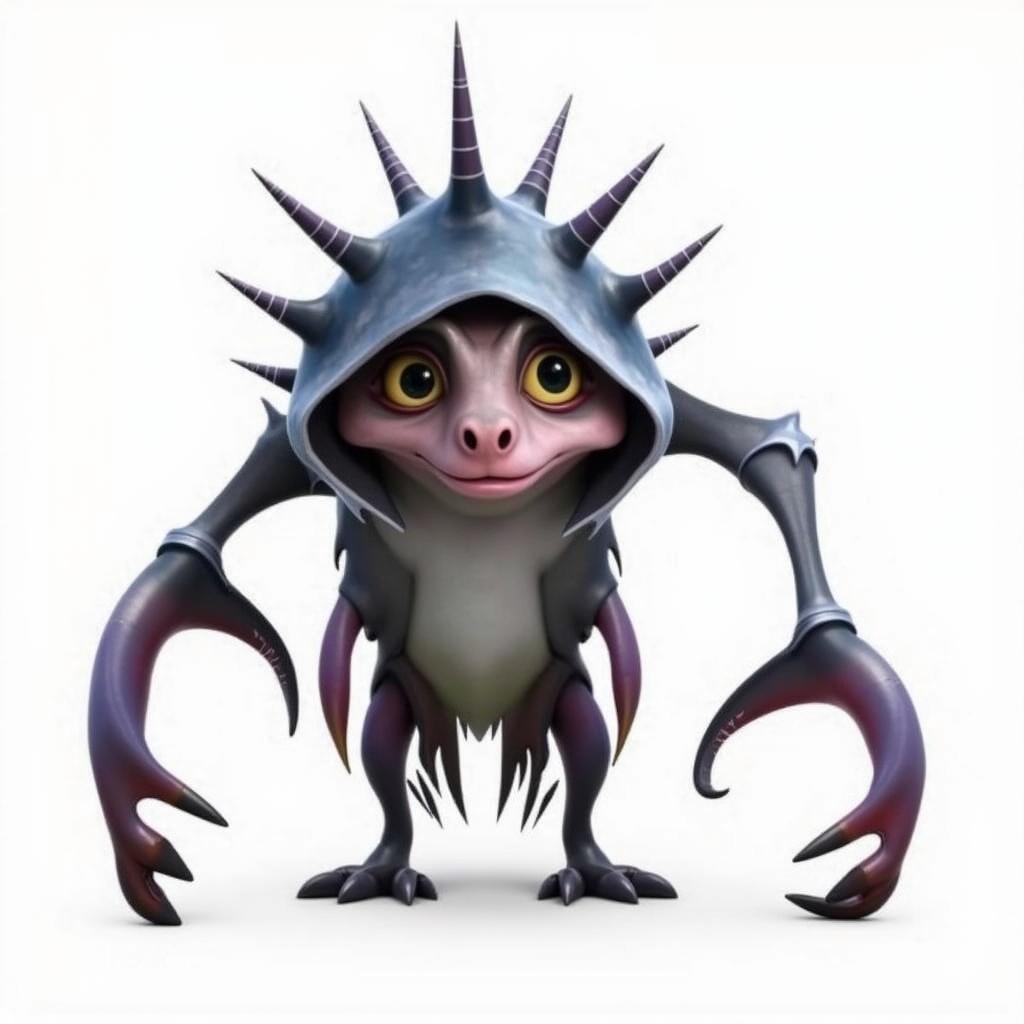} 
        \\    
        
        \includegraphics[height=0.1\textheight]{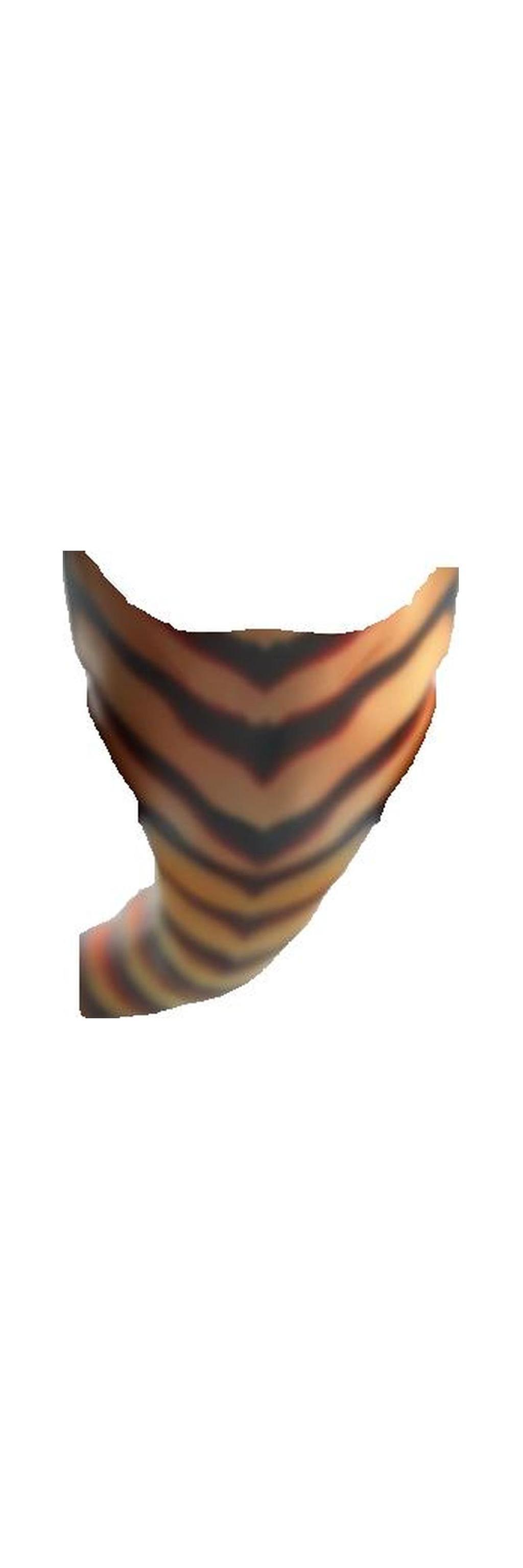} &
        \includegraphics[height=0.1\textheight]{figures/results_ref/spiky_hand/original_0.jpg} &
        \includegraphics[height=0.1\textheight]{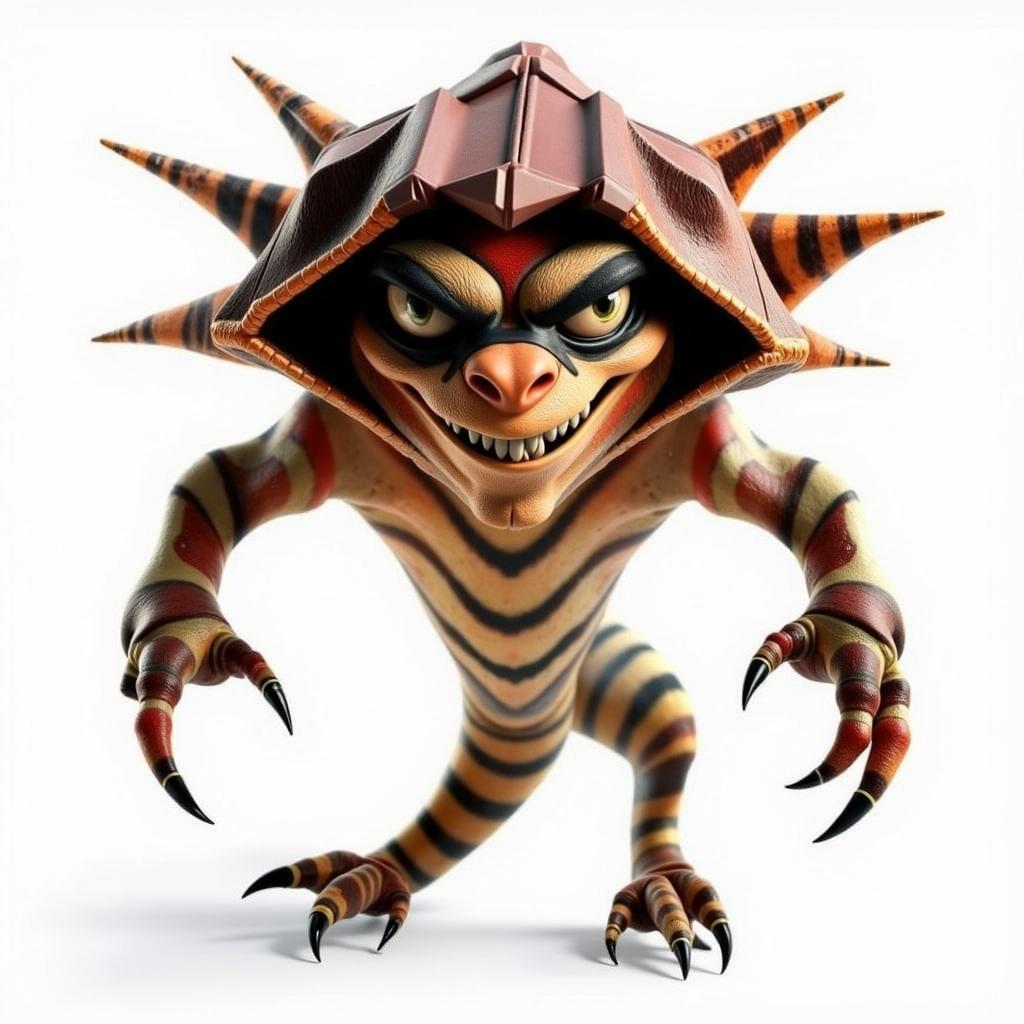} &
        \includegraphics[height=0.1\textheight]{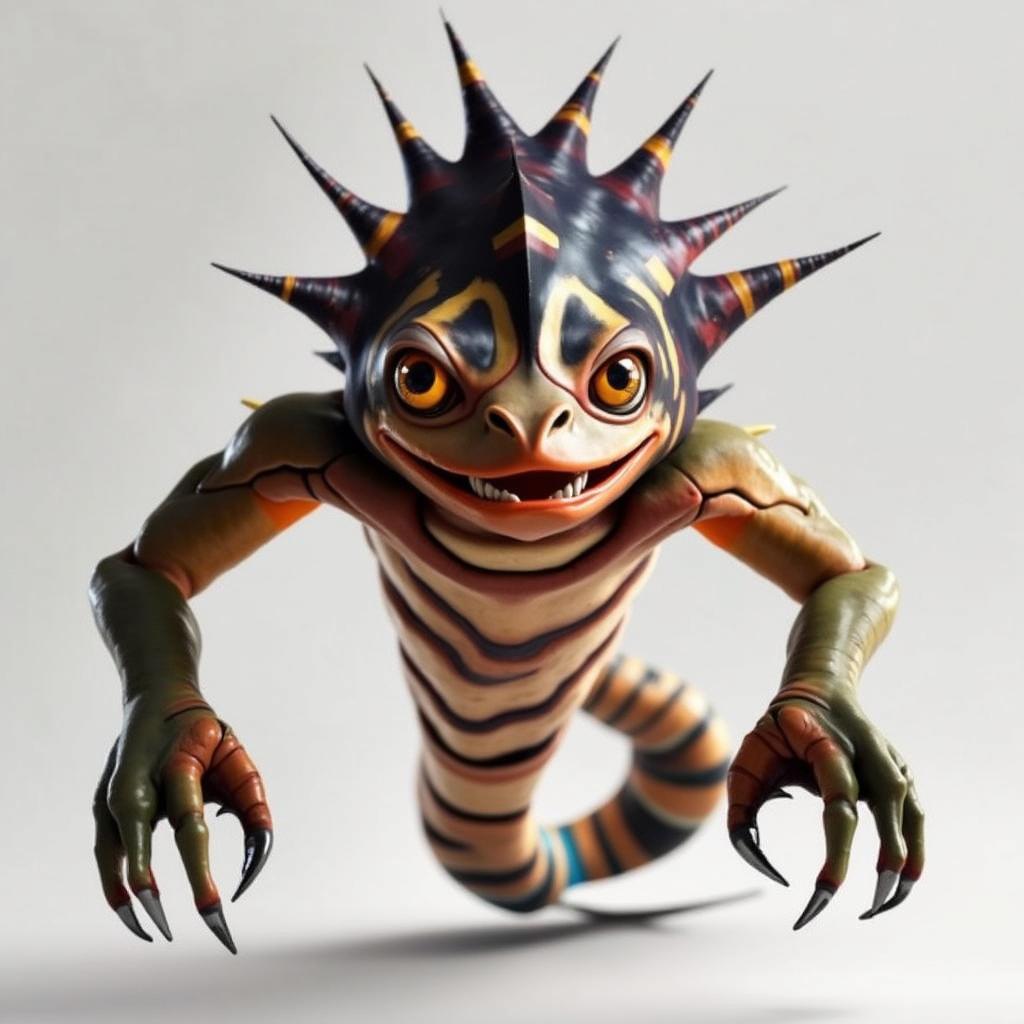} 
        \\    
        \includegraphics[height=0.1\textheight]{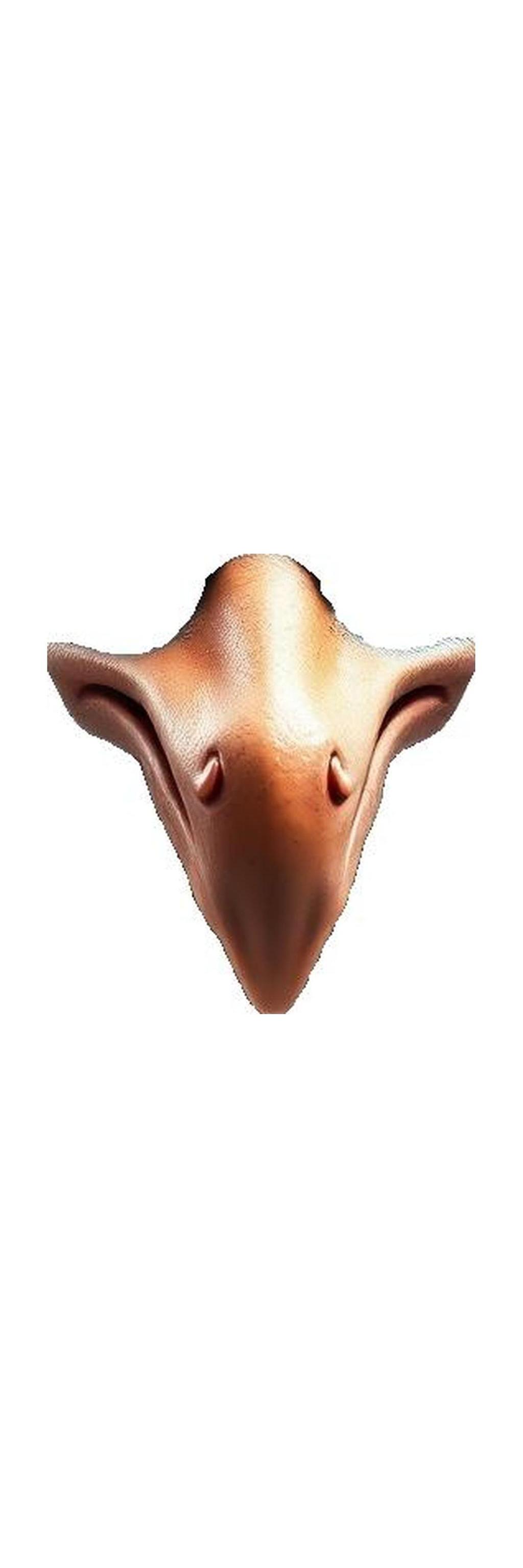} &
        \includegraphics[height=0.1\textheight]{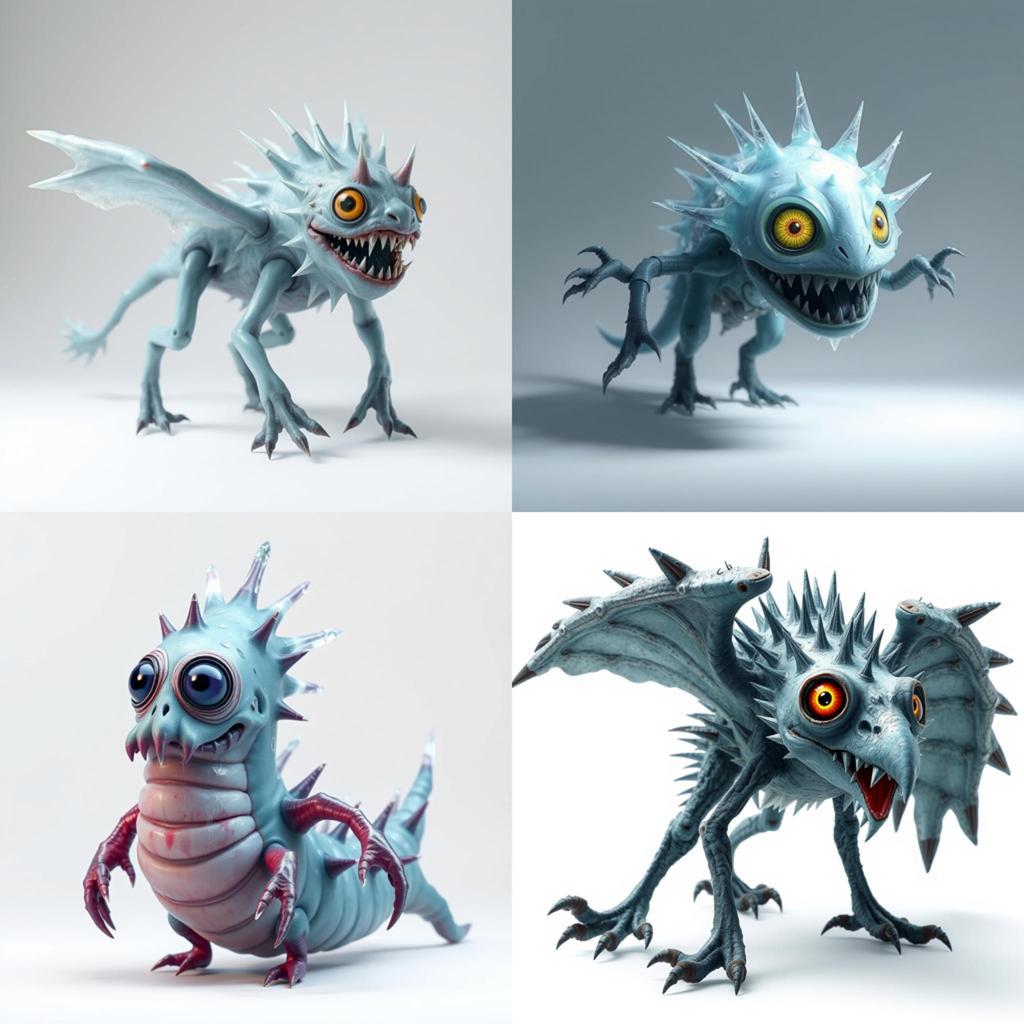} &        
        \includegraphics[height=0.1\textheight]{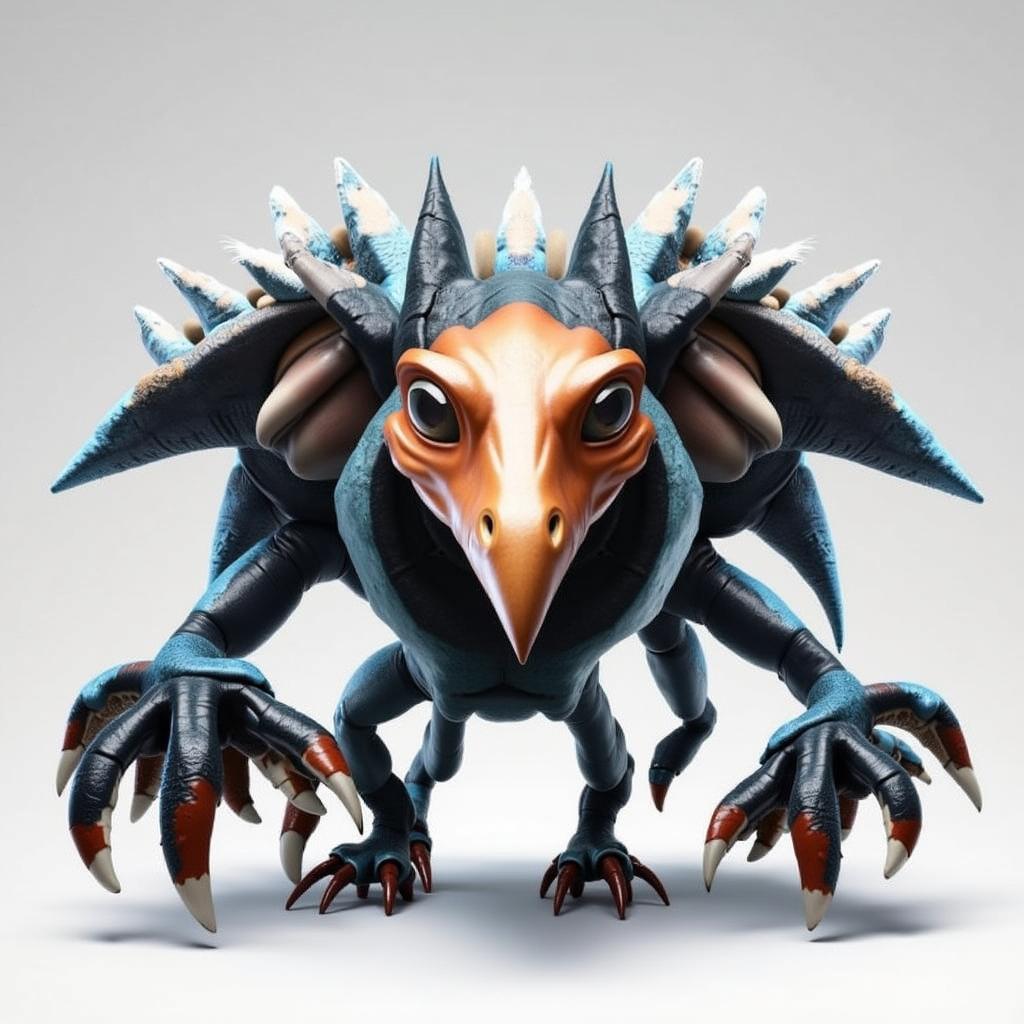} &
        \includegraphics[height=0.1\textheight]{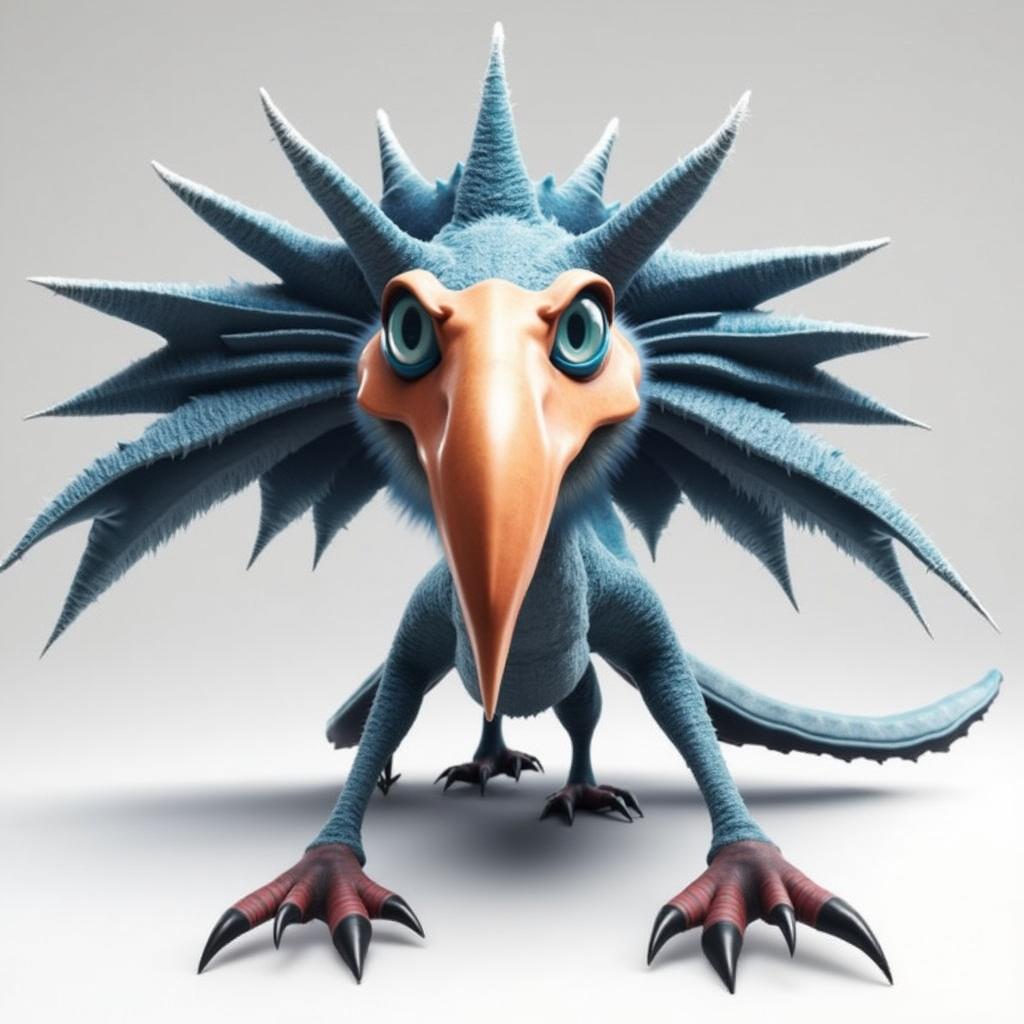} 
        \\           
        Part & Style  & \multicolumn{2}{c}{Sampled Results} 

    \end{tabular}
    }
    \caption{\textbf{PiT results with reference conditioning.} The model is conditioned on both image parts as well as a grid representing a target look.}
    \label{fig:supp_ref}
\end{figure}

While our primary focus is on object parts as input visual concepts, our approach is not inherently limited to this specific semantic meaning. As an example, in~\Cref{fig:supp_ref}, we show that our model can also be conditioned on a grid-like arrangement of reference images, describing some form of visual style. This allows one to visually describe a target look for the generated concept alongside the target parts.

\begin{figure}
    \centering
    \setlength{\tabcolsep}{0.5pt}
    \renewcommand{\arraystretch}{0.5}
    {\small
    \begin{tabular}{c @{\hspace{0.2cm}} c c c}

        \includegraphics[height=0.1\textheight]{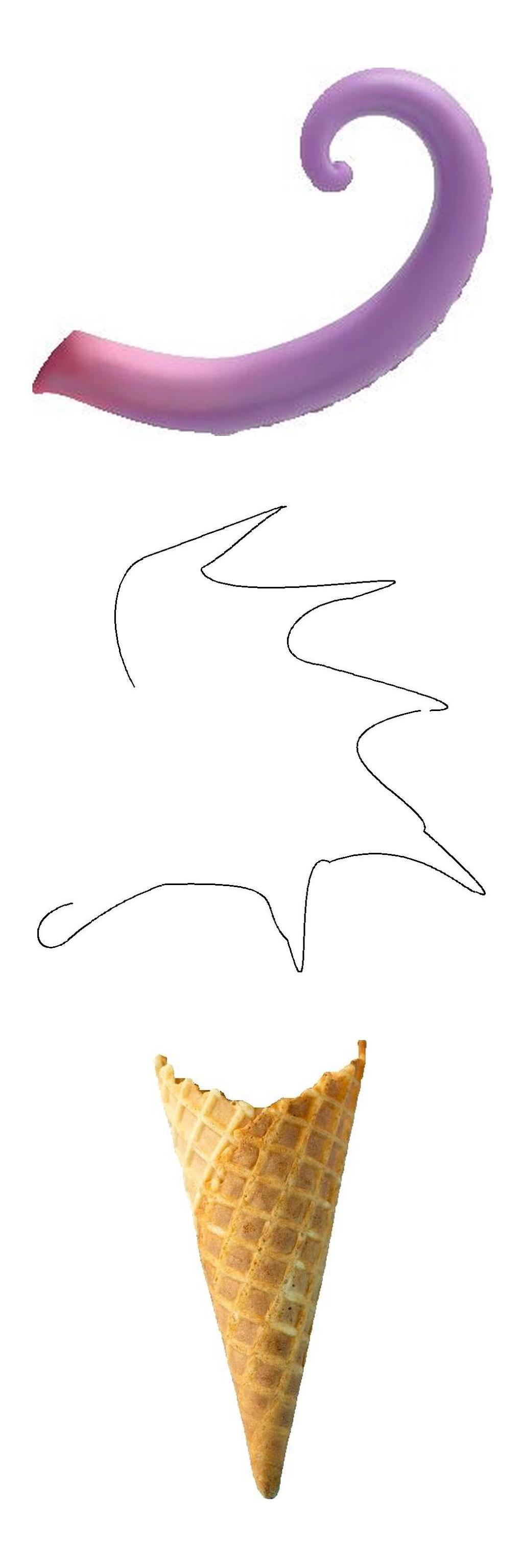} &
        \includegraphics[height=0.1\textheight]{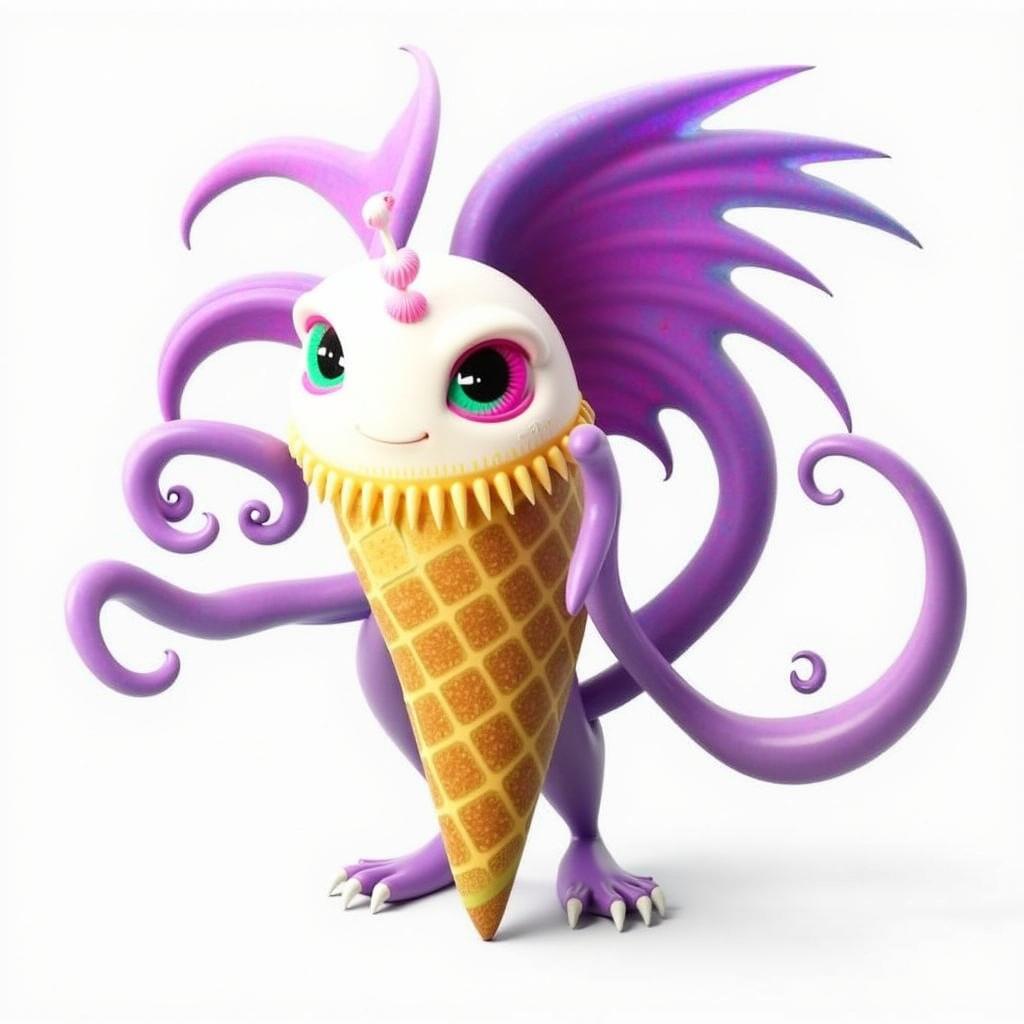} &
        \includegraphics[height=0.1\textheight]{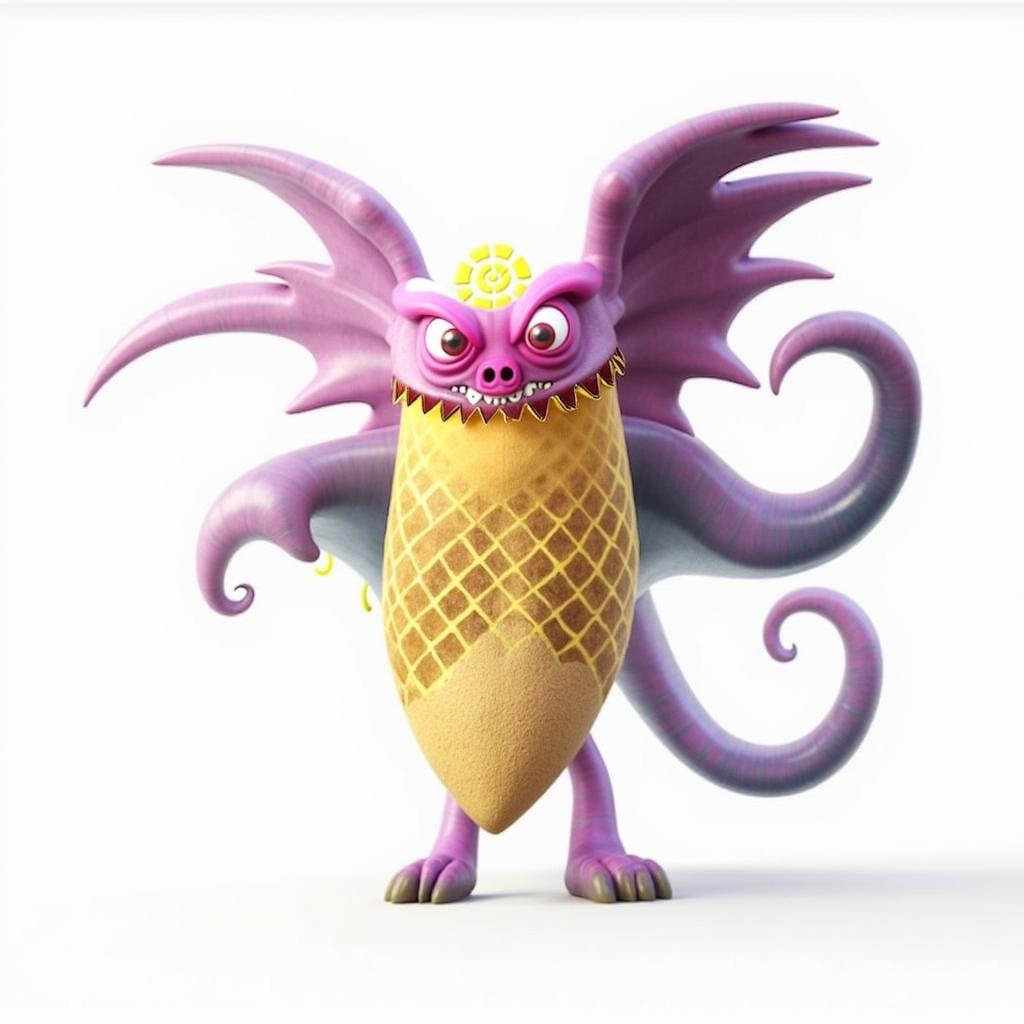} &
        \includegraphics[height=0.1\textheight]{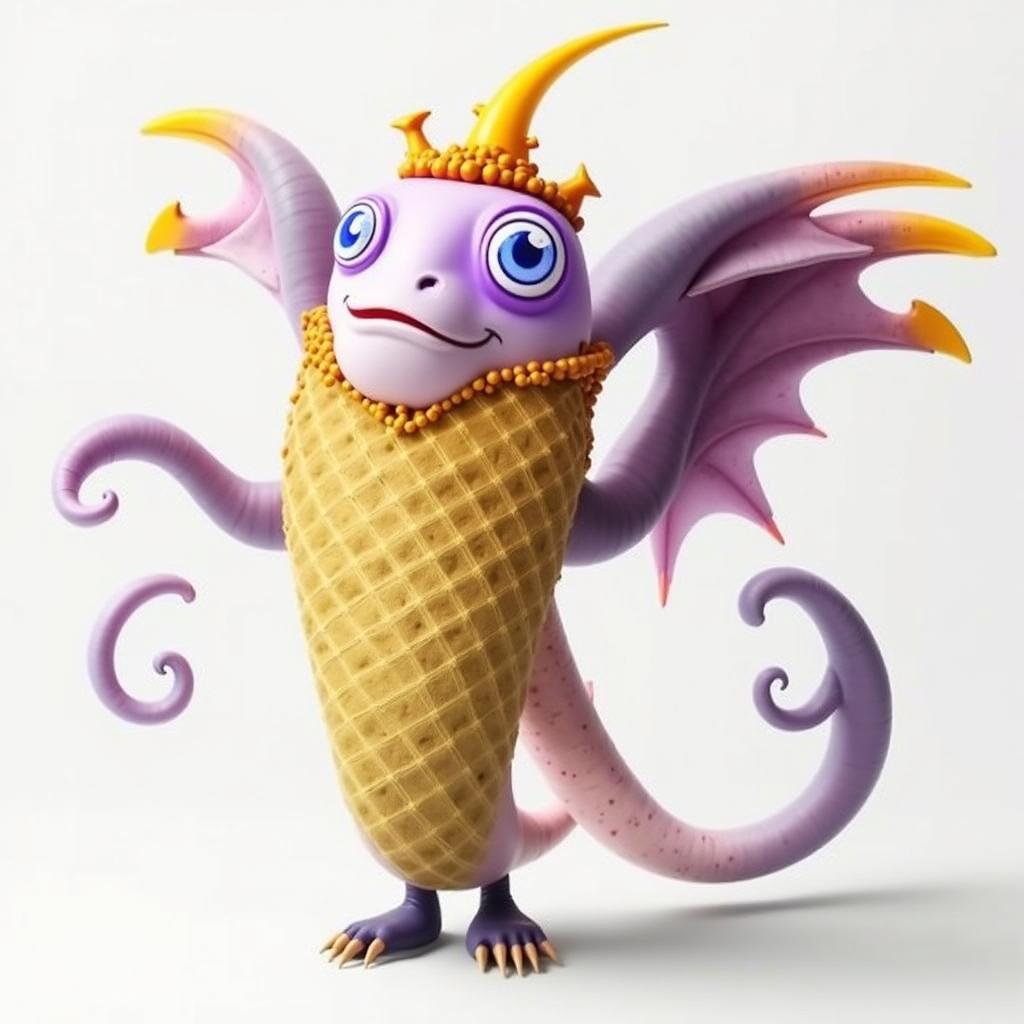} 
        \\     

        \includegraphics[height=0.1\textheight]{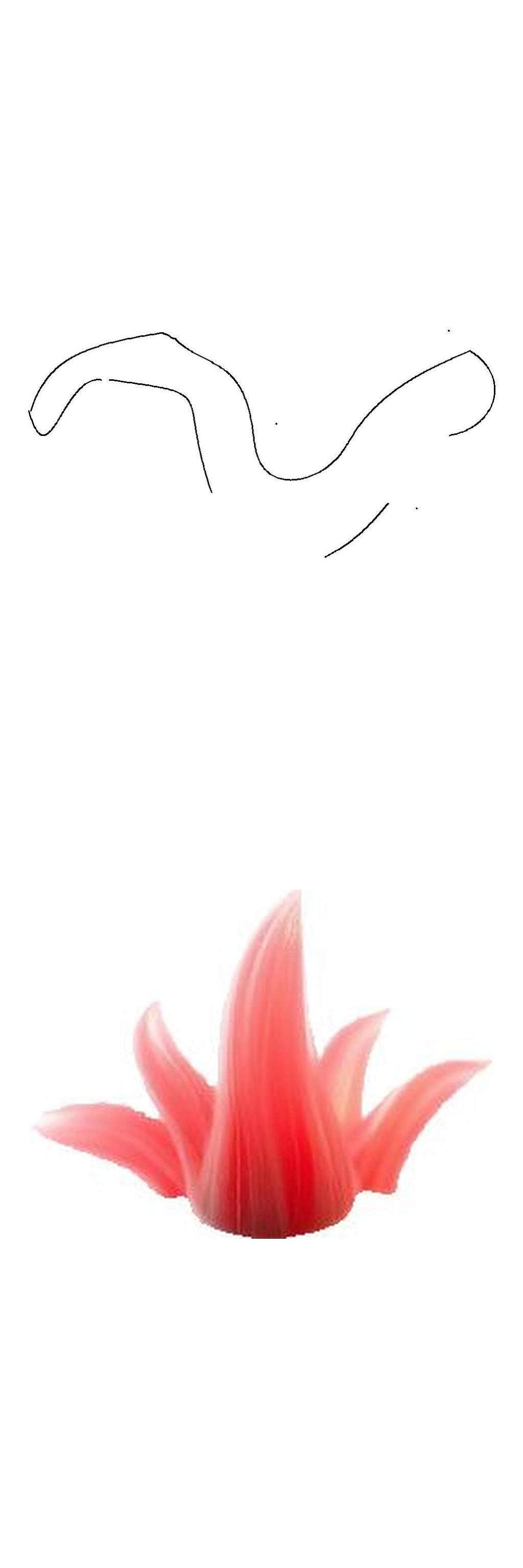} &
        \includegraphics[height=0.1\textheight]{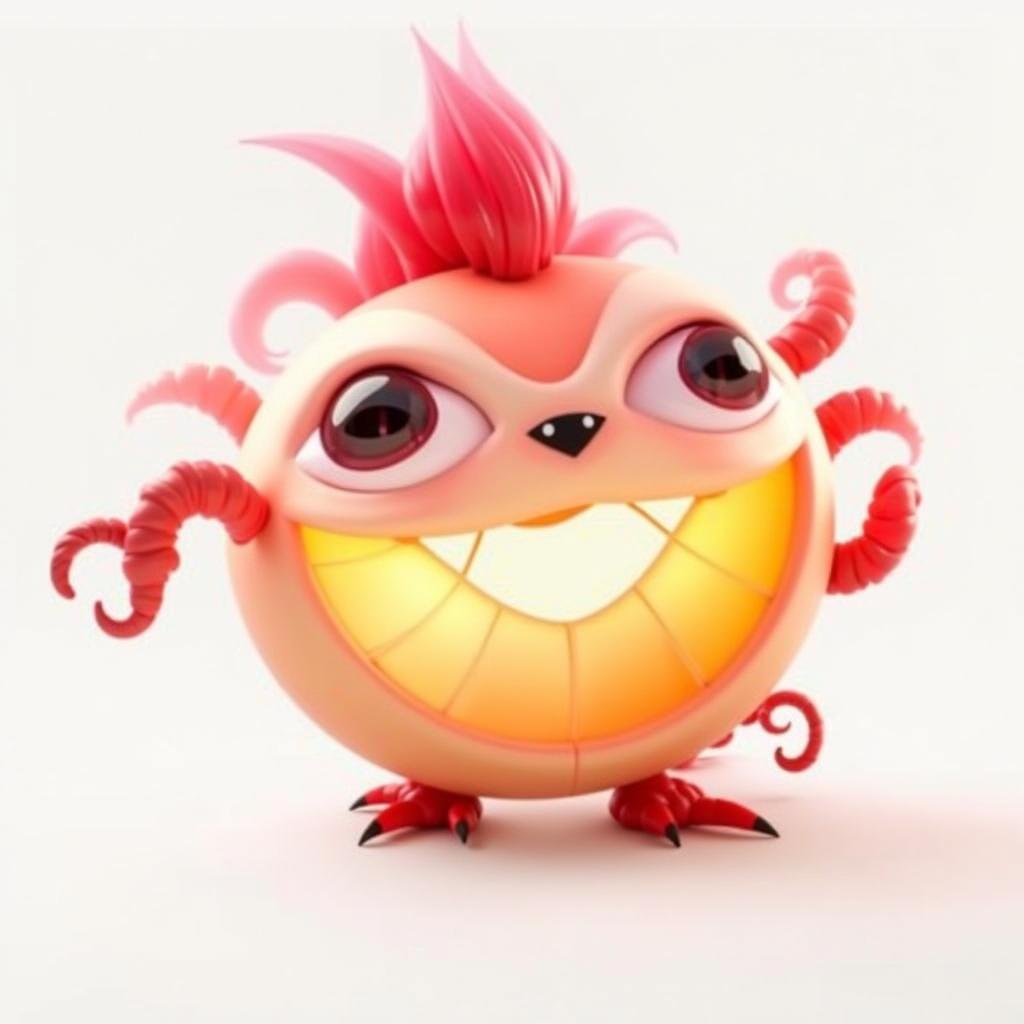} &
        \includegraphics[height=0.1\textheight]{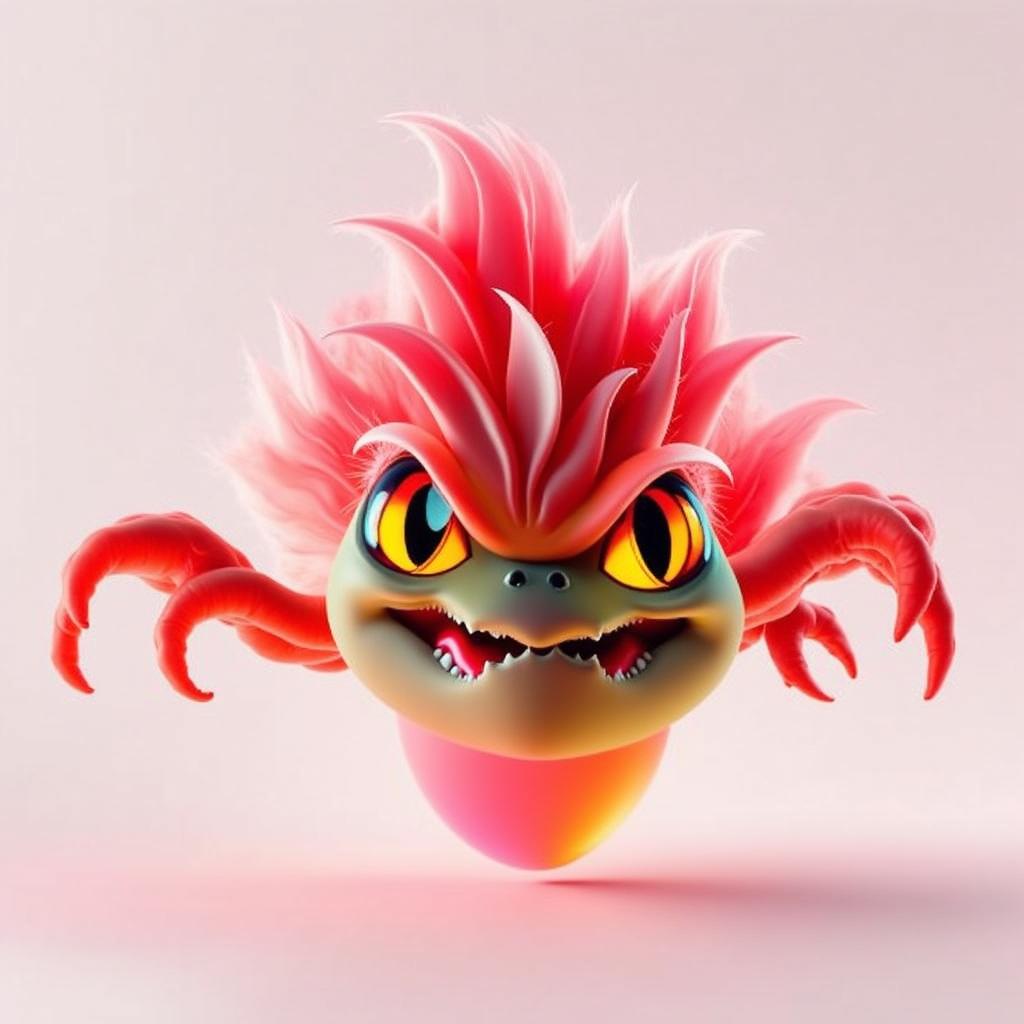} &
        \includegraphics[height=0.1\textheight]{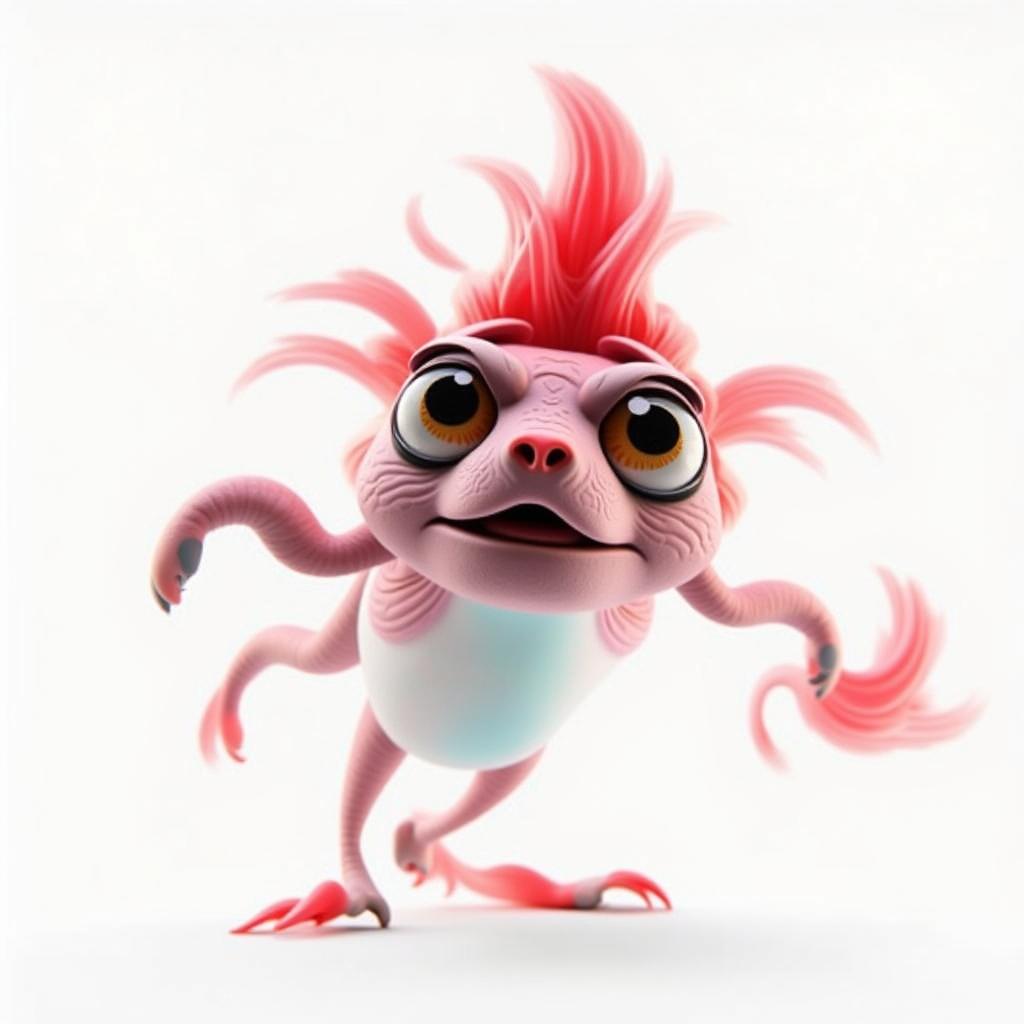} 
        \\        

        \includegraphics[height=0.1\textheight]{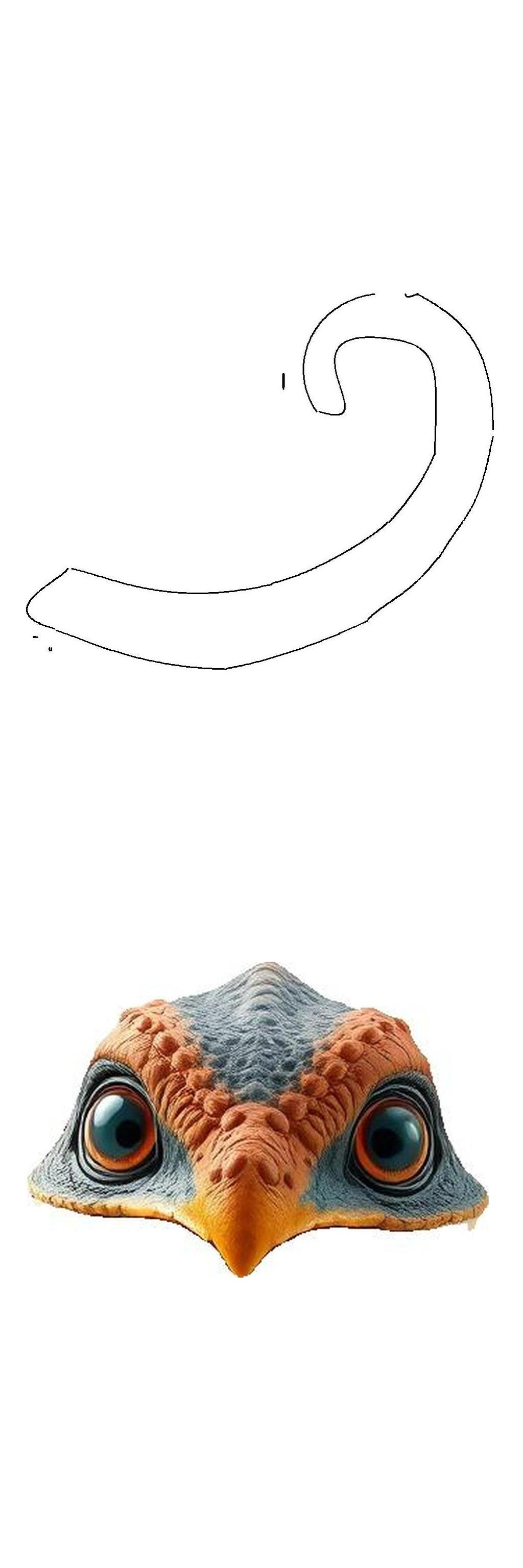} &
        \includegraphics[height=0.1\textheight]{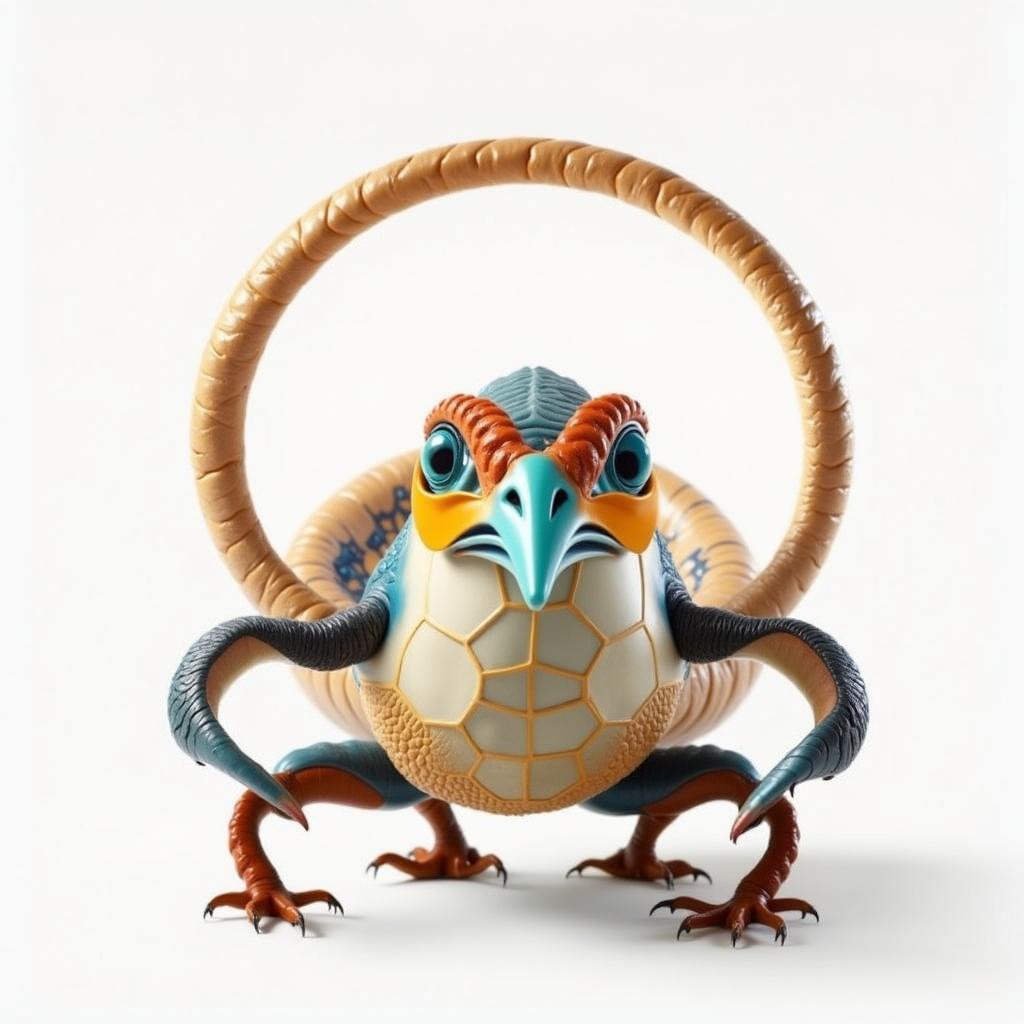} &
        \includegraphics[height=0.1\textheight]{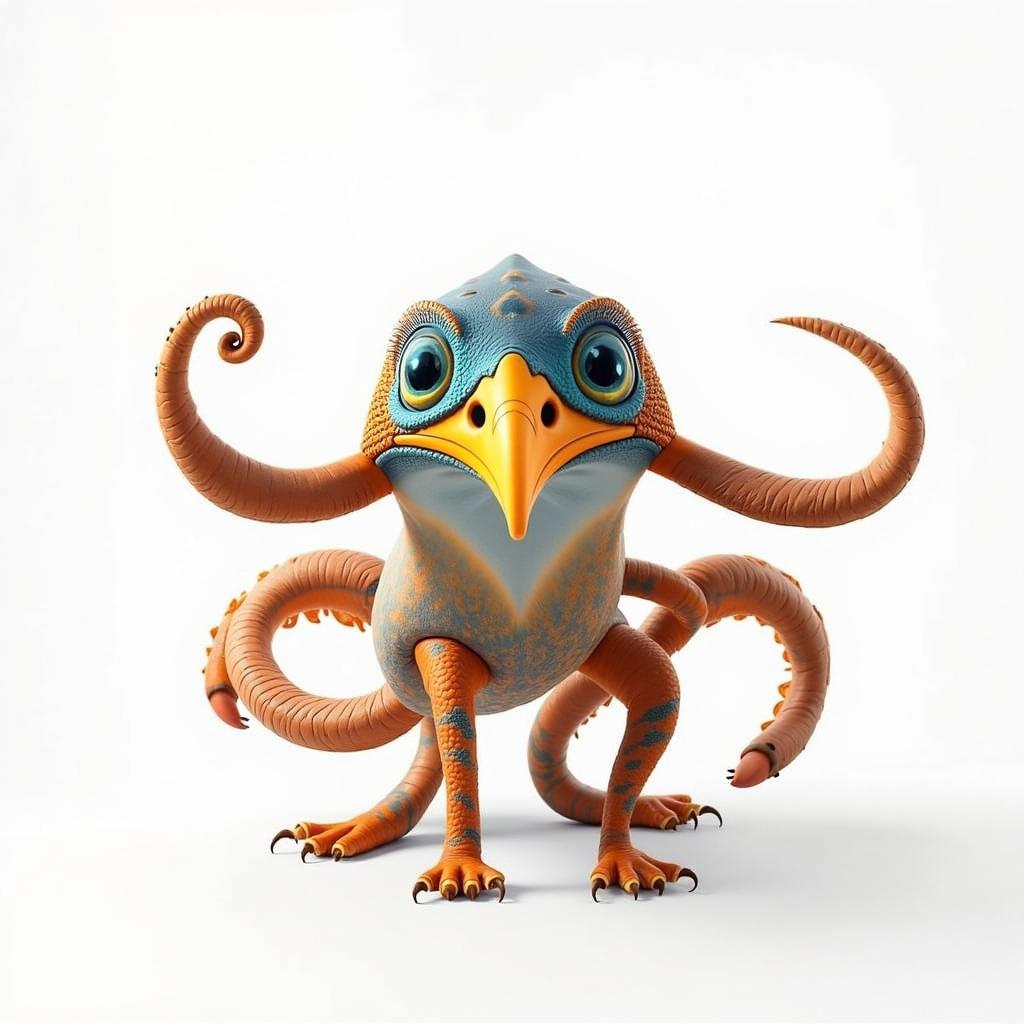} &
        \includegraphics[height=0.1\textheight]{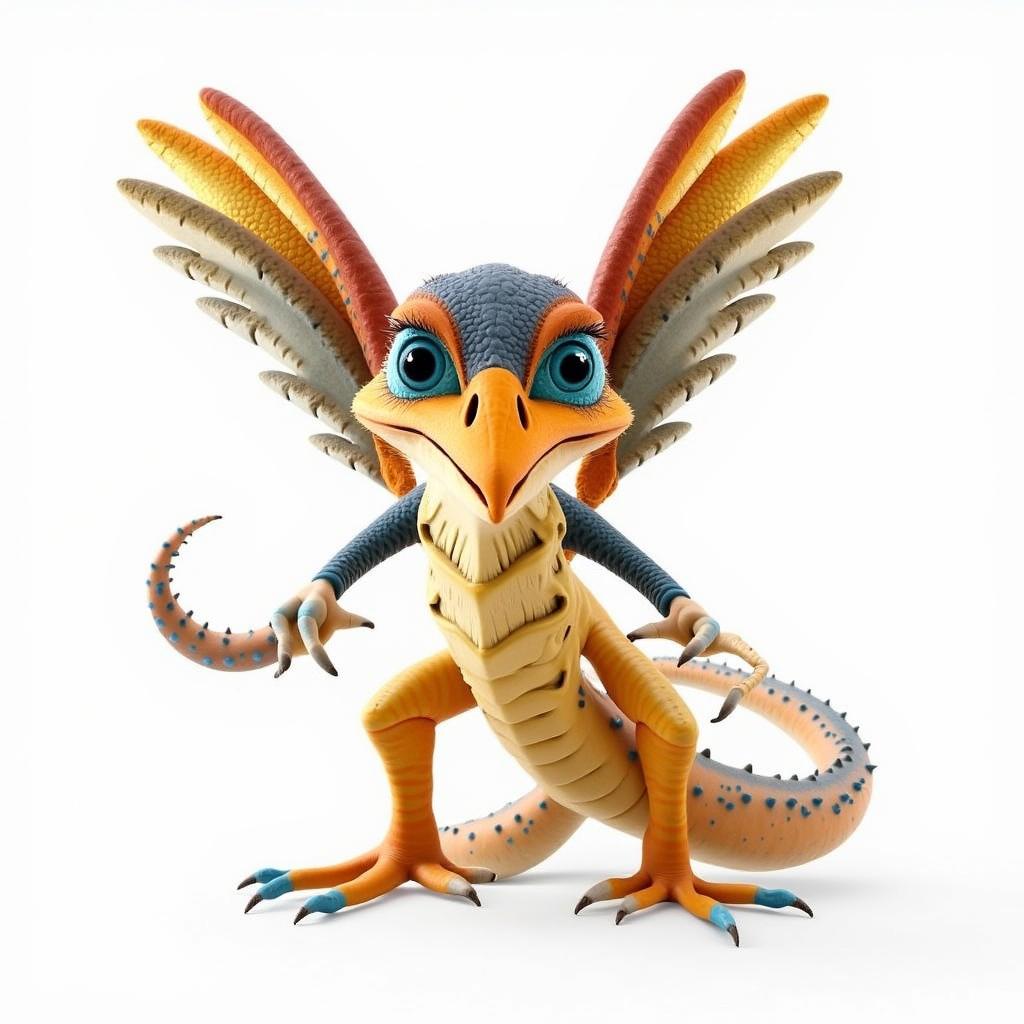} 
        \\            

        \includegraphics[height=0.1\textheight]{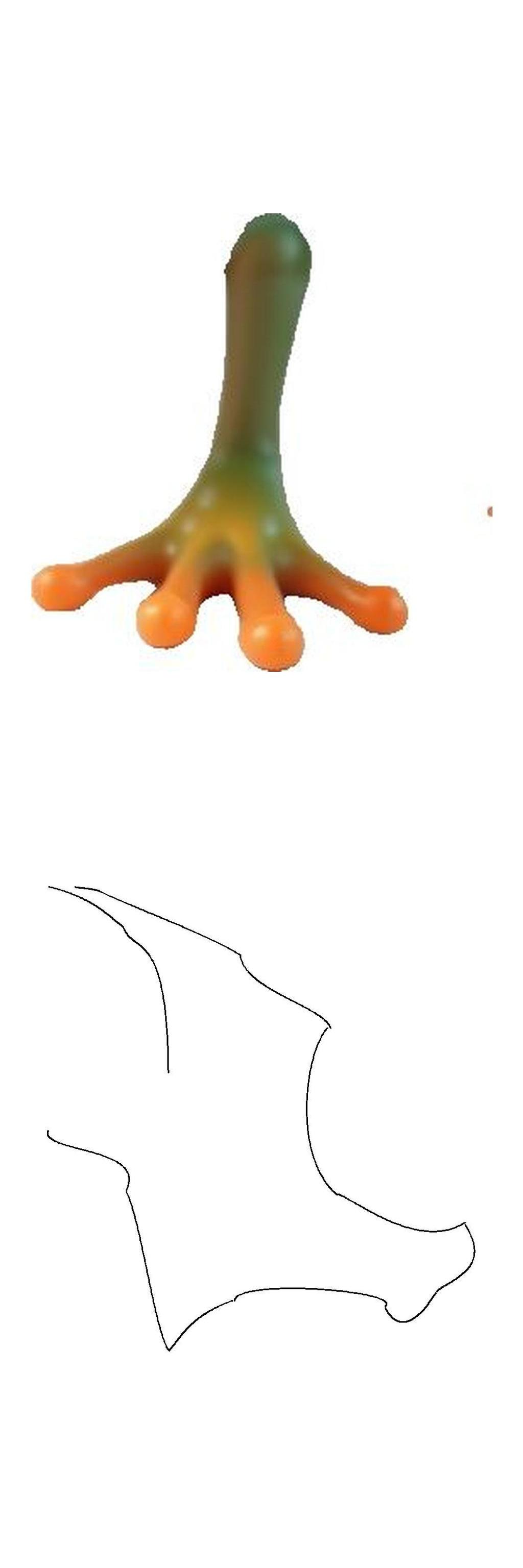} &
        \includegraphics[height=0.1\textheight]{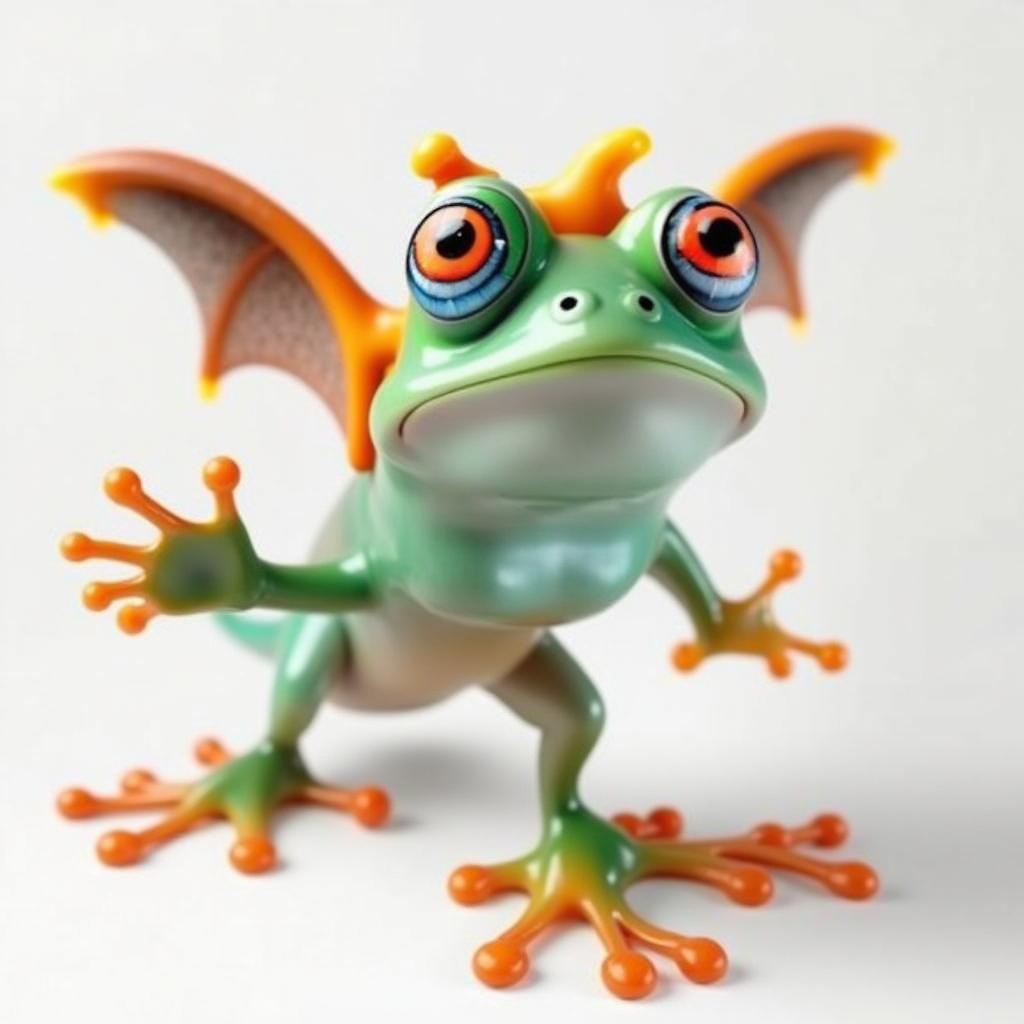} &
        \includegraphics[height=0.1\textheight]{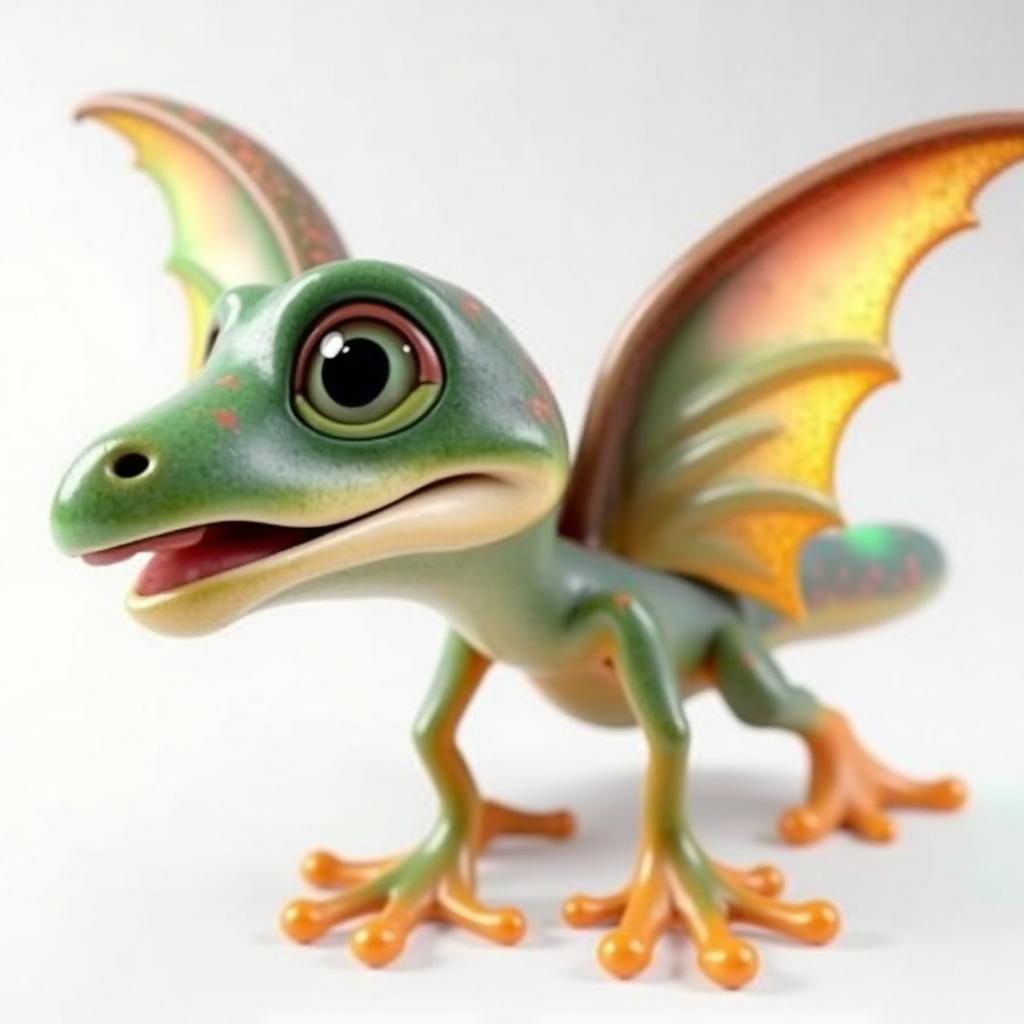} &
        \includegraphics[height=0.1\textheight]{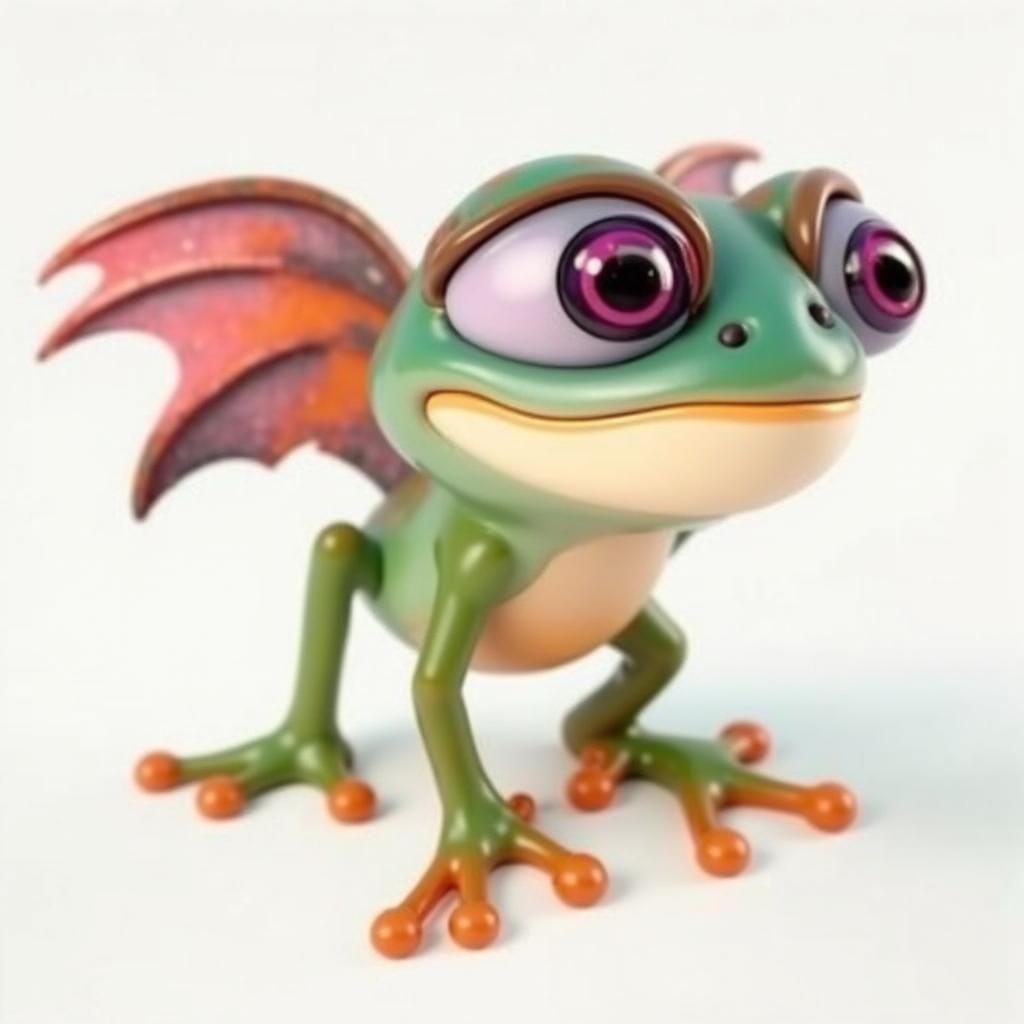} 
        \\             
        
        Input & \multicolumn{3}{c}{Sampled Results} 

    \end{tabular}
    }
    \caption{\textbf{PiT results for sketch conditioning.} The model is conditioned on sketch images, allowing one to specify rough shapes without color.}
    \label{fig:supp_sketch}
\end{figure}

We additionally demonstrate in~\Cref{fig:supp_sketch} that PiT can also be conditioned on sketches depicting parts of an object, providing users with greater flexibility when specific visual parts are unavailable. In this case, the model interprets each sketch based on the learned prior.

\subsection{Limitations}
It is important to note that, as with any embedding-based method, the amount of information encoded in the compact embedding space is inherently limited. While our use of the $\mathcal{IP}+$ space significantly improves the tradeoff between reconstruction and editability --- mitigating prominent issues such as attribute mixing discussed in~\cite{richardson2024pops,ramesh2022hierarchical} --- not all information can be successfully preserved.
For example, the model struggles to encode small or high-frequency details, limiting our ability to condition on fine-grained regions of the target concept or small text. Nevertheless, as we demonstrate, embedding-based mechanisms enable simple manipulations that would otherwise be difficult to achieve. Thus, for our ideation task, we find $\mathcal{IP}+$ to be a well-balanced choice.

\vspace*{-0.1cm}
\section{Conclusions}
In this work, we introduced PiT, a method for ideating new concepts from a sparse set of input parts. Our approach functions as a generative operator that receives a set of part embeddings and produces an embedding representing a plausible and complete concept within a given target domain.
PiT addresses core limitations of current CLIP-based methods by operating in the more expressive $\mathcal{IP}^+$ space, enabling improved concept reconstructions while allowing for meaningful semantic manipulations. 
We hope that PiT not only serves as a strong model for photo-inspired generation and ideation but also provides a foundation for solving additional tasks that would benefit from operating within the same representation space.

\section*{Acknowledgements}
We thank Or Lichter and Yael Vinker for their feedback and insightful comments. This research was supported in part by the Israel Science Foundation (grants no. 2492/20 and 1473/24), Len Blavatnik and the Blavatnik family foundation.

{
    \small
    \bibliographystyle{ieeenat_fullname}
    \bibliography{main}
}

\clearpage
\appendix
\appendixpage
\section{Additional Details}

\subsection{Training Data for IP-Prior}

As explained in the main text, the data for each domain shown in the paper was generated using Flux-Schnell~\cite{blackforest2024flux} with a dedicated prompt. We next detail the specific prompt used for each such domain
\begin{itemize}
    \item \textbf{Characters} - ``studio photo pixar style concept art, An imaginary fantasy (adjectives) (character) creature with eyes arms legs mouth, white background studio photo pixar asset''.
    
    Where ``adjectives'' is a set of 4-6 adjectives randomly sampled from a pool of about 200 adjectives, collected using an LLM, and ``character'' is a set 1-3 characters sampled from a list of characters generated by an LLM concatenated with a list of 18K classes from OpenImages~\cite{kuznetsova2020open}. 
    \item \textbf{Products} - ``A product design photo of a (attributes) product with (character-like) attributes, integrated together to create one seamless product.  It is set against a light gray background with a soft gradient, creating a neutral and elegant backdrop that emphasizes the contemporary design. The soft, even lighting highlights the contours and textures, lending a professional, polished quality to the composition''
    
    Where ``attributes'' is a similarly generated list of $\sim$300 product attributes including materials and features and ``character-like'' is an optional description on an object or animal.
    \item \textbf{Toys} - ``Professional studio photo of an extremely cute and friendly smiling (animal) plush toy is sitting in a natural frontal position, facing the camera. He is wearing a (outfit) (item). It is set against a white background with soft, even lighting, lending a professional quality to the composition''
    
    Where, as before, ``animal'' is a set of texts of different animals, and ``outfit'' is a set of attributes covering clothing and hairstyle.
    \item \textbf{Portraits} - ``An artistic face collage crafted from a sparse and minimal set of large everyday objects, such as (object). The assemblage forms expressive features, such as lips, textured eyes, and a sculpted nose, set against a pristine white background that highlights the intricate details and creative use of materials.'' 

    Where ``object'' is chosen from a set of $60$ relevant objects, such as ``shells'' and ``fruits''.

    \item \textbf{Rubber Ducks} - ``Professional studio photo of a rubber duck. He is wearing a (outfit). It is set against a white background with soft, even lighting, lending a professional quality to the composition''

    Where ``outfit'' is the same list as the one used for toys.
    
\end{itemize}

In terms of quantity, the data itself was generated on the fly, alongside the training process. We use four steps for generating the image with FLUX-Schnell model~\cite{blackforest2024flux}, followed by segmentation using SAM~\cite{kirillov2023segment}.

\subsection{Training Data for IP-LoRA}
We follow the following process to train the background generation LoRA, where the model is conditioned on an image embedding representing the concept alongside the text prompt. First, we generate images of different creatures using FLUX, and add to the prompt a description of a background chosen from a set of $50$ prompts. Then, we use the Bria RMBG2.0 model~\cite{BiRefNet, bria2024rmbg20} to separate the creature from the background. This image is used as the input to the IP-Adapter+ during training alongside the text prompt describing the target background.

Similarly, for the reference sheet LoRA, we first generate a set of images using the prompt ``A character sheet displaying an imaginary fantasy (adjectives) (creatures) creature with eyes mouth, from several angles with 1 large front view in the middle, clean white background. In the background, we can see half-completed, partially colored sketches of different parts of the object''. Here, the adjectives and creatures are sampled from the same list as above. Next, we use grounded SAM~\cite{ren2024groundedsam} to find the largest single view of the creature in the generated image. Its segmented and cropped instance then serves as the visual conditioning for the model. In this scenario, the text prompt remains fixed during tuning, as the only control comes from the given image embedding.

\section{Additional Results and Comparisons}
\begin{figure*}
    \centering
    \setlength{\tabcolsep}{0.5pt}
    \addtolength{\belowcaptionskip}{-5pt}
    {\small

    \begin{tabular}{c c c c @{\hspace{0.2cm}} c c c c}
       \includegraphics[height=0.135\textwidth]{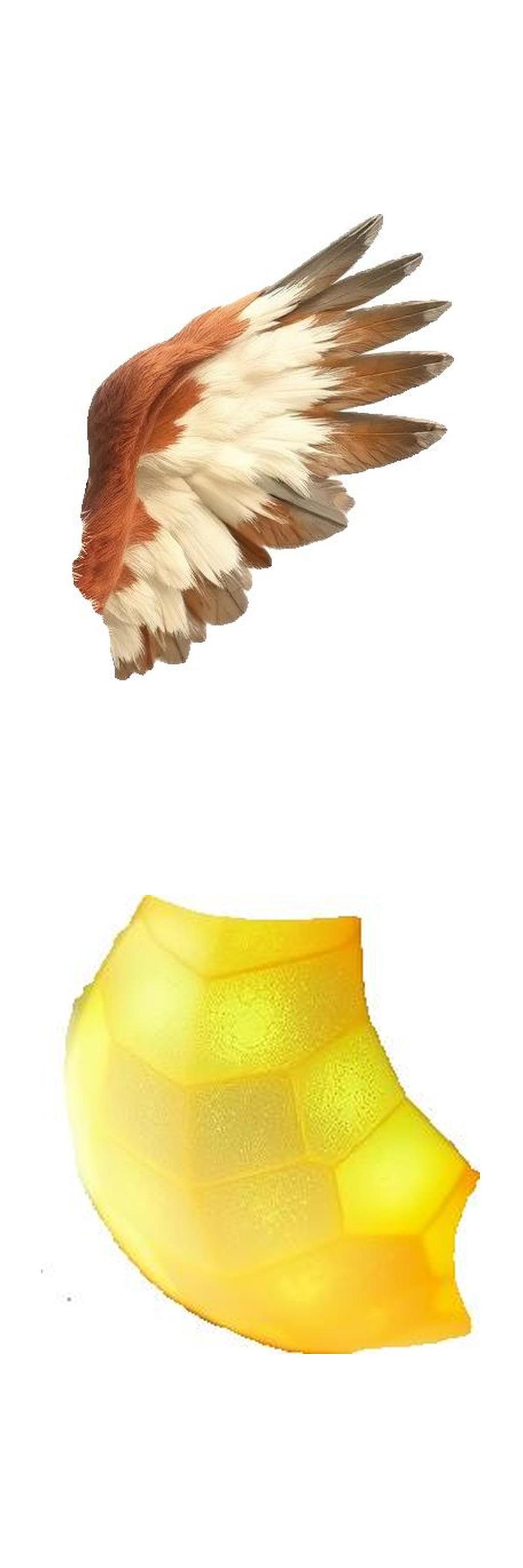} &
        \includegraphics[height=0.135\textwidth]{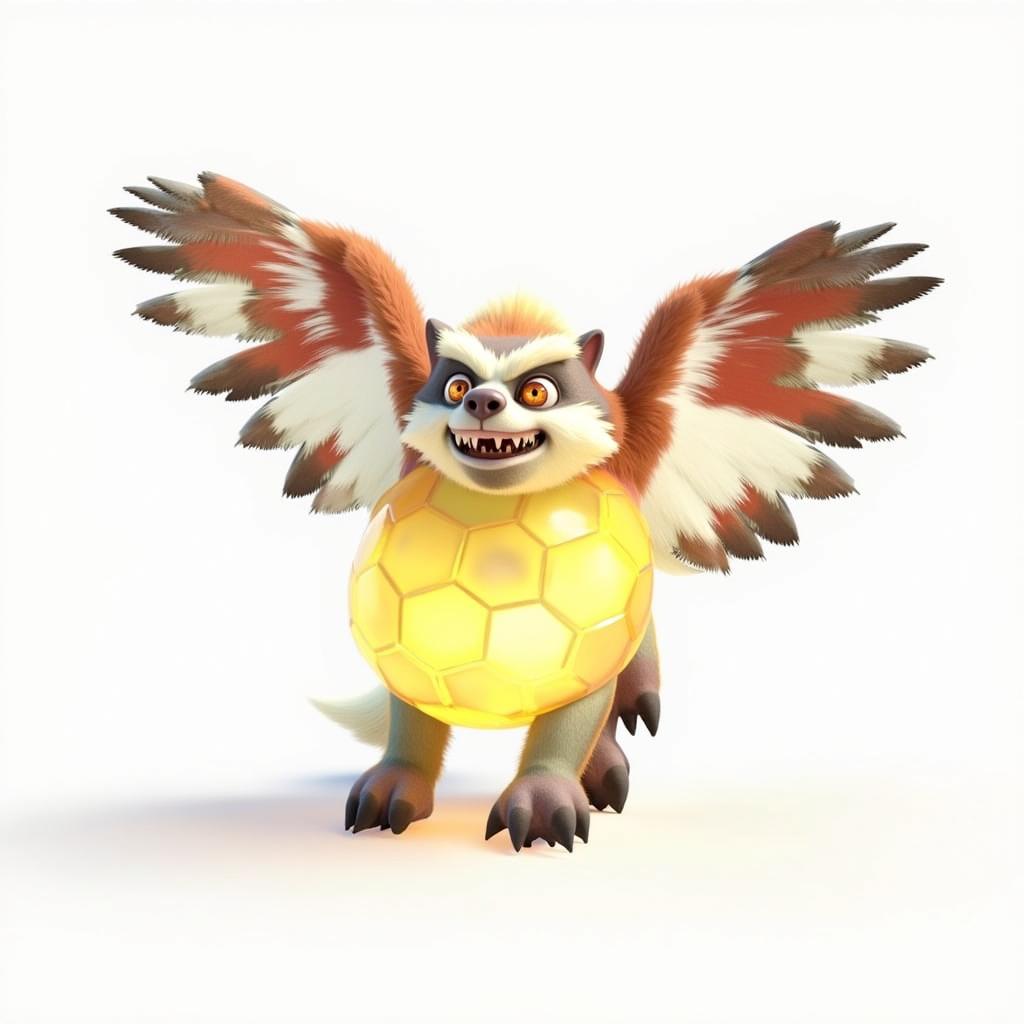} &
        \includegraphics[height=0.135\textwidth]{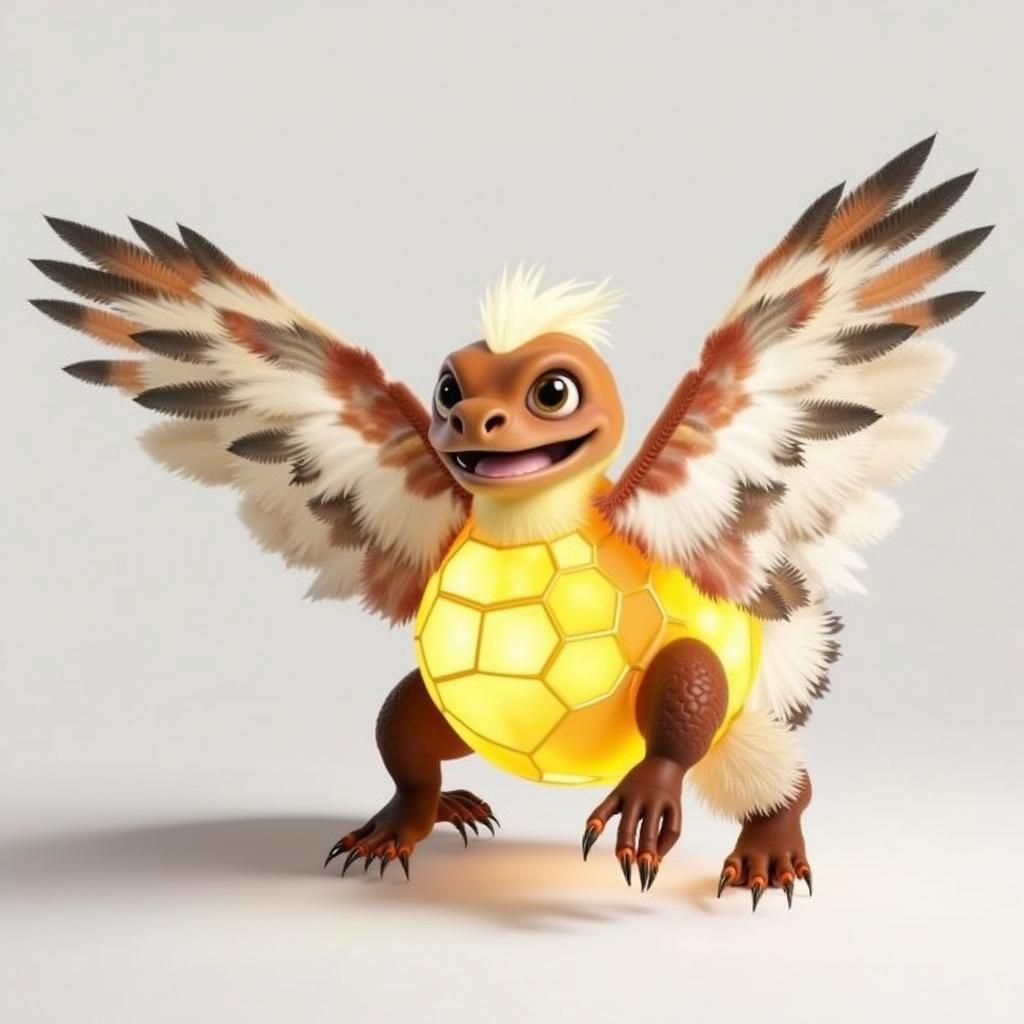} &

        \includegraphics[height=0.135\textwidth]{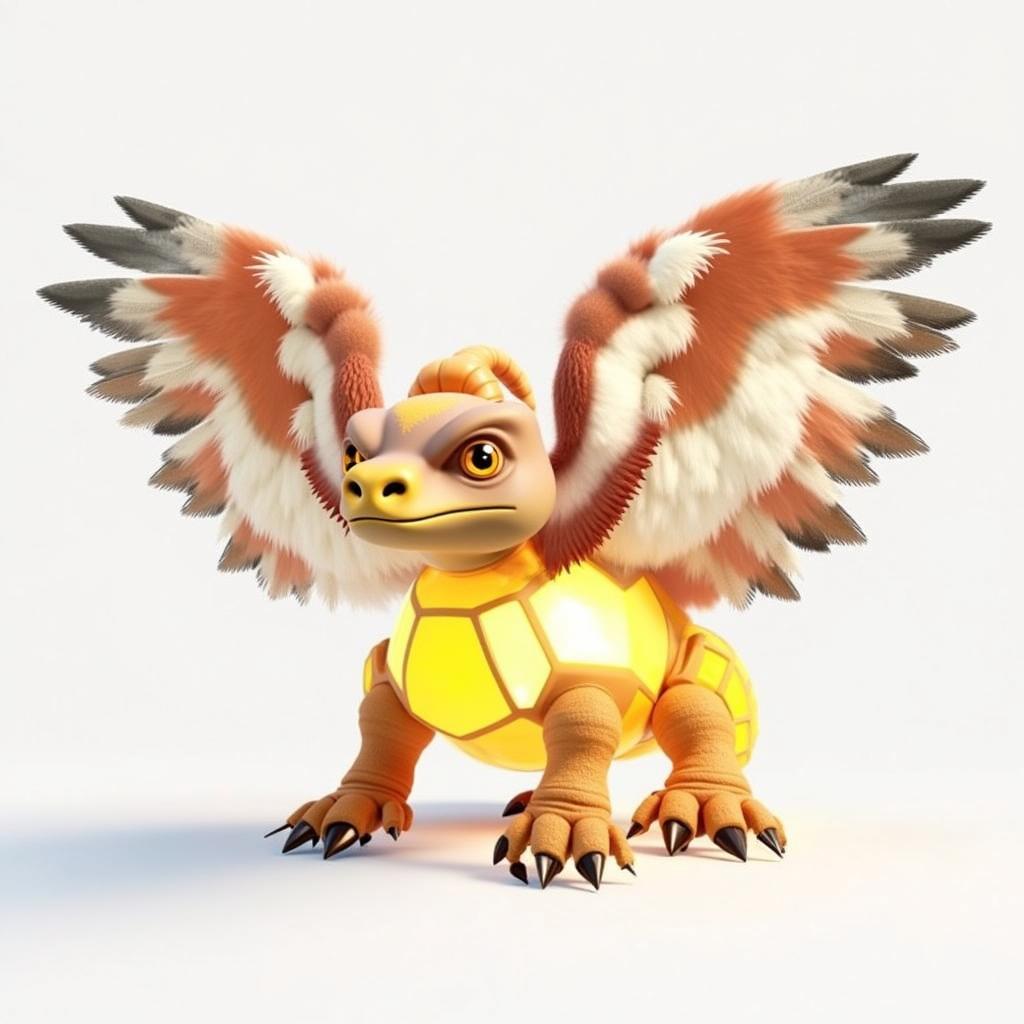} &
        
        \includegraphics[height=0.135\textwidth]{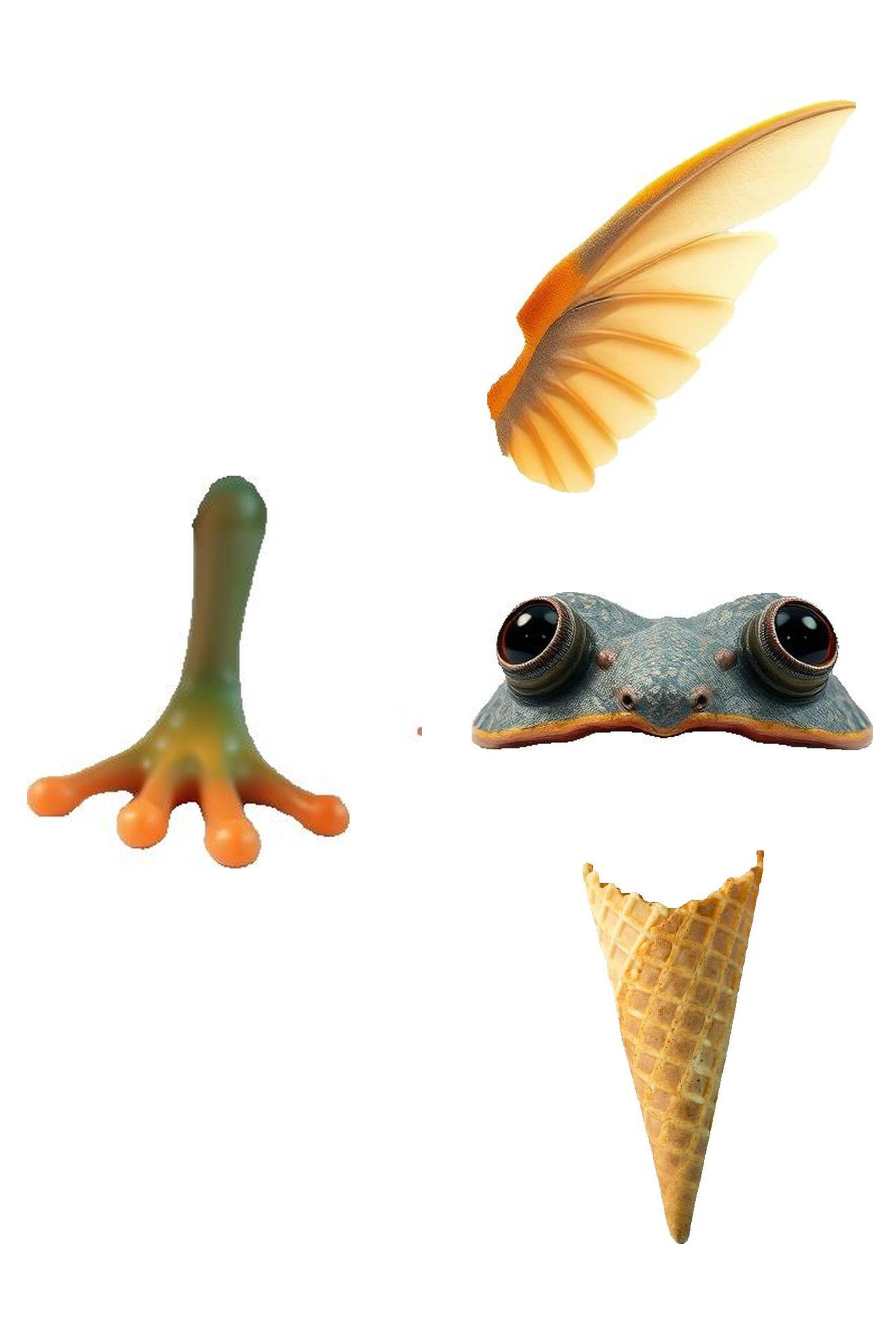} &
        \includegraphics[height=0.135\textwidth]{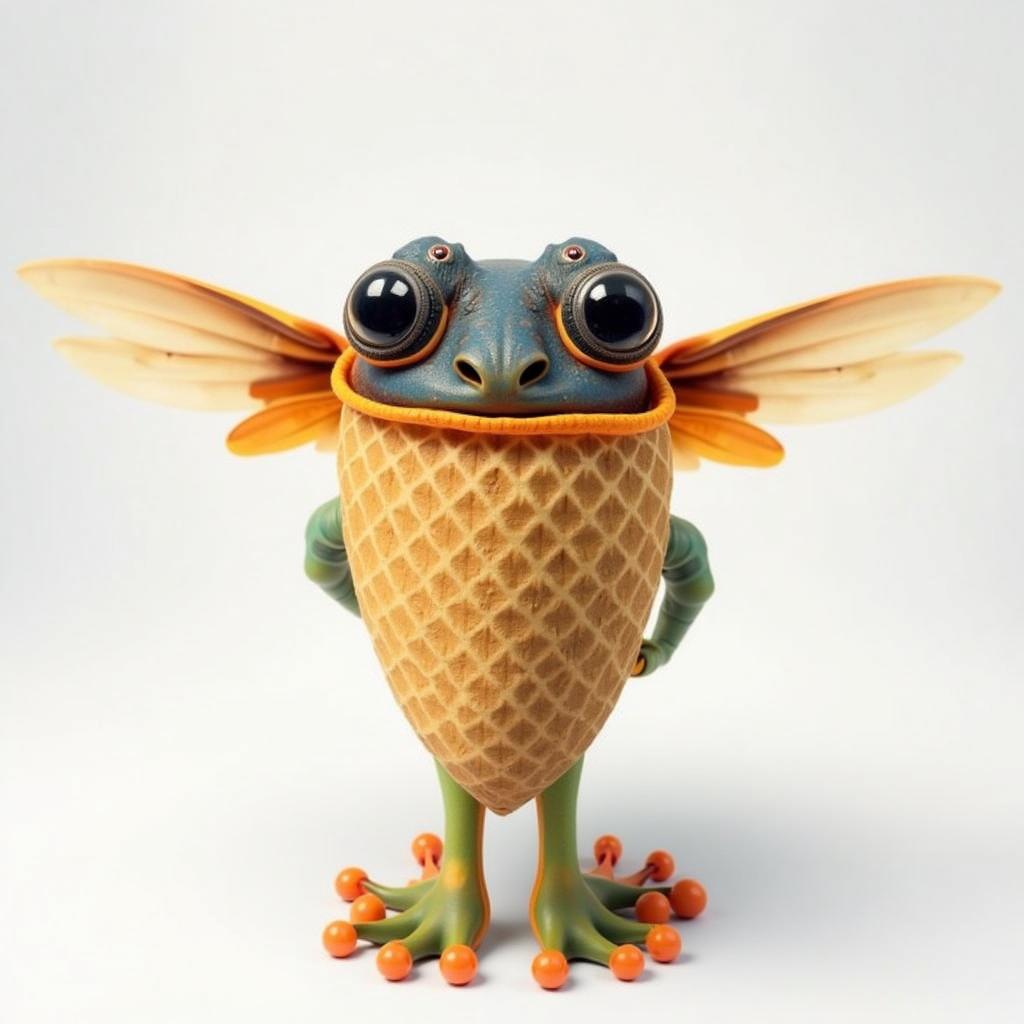} &
        \includegraphics[height=0.135\textwidth]{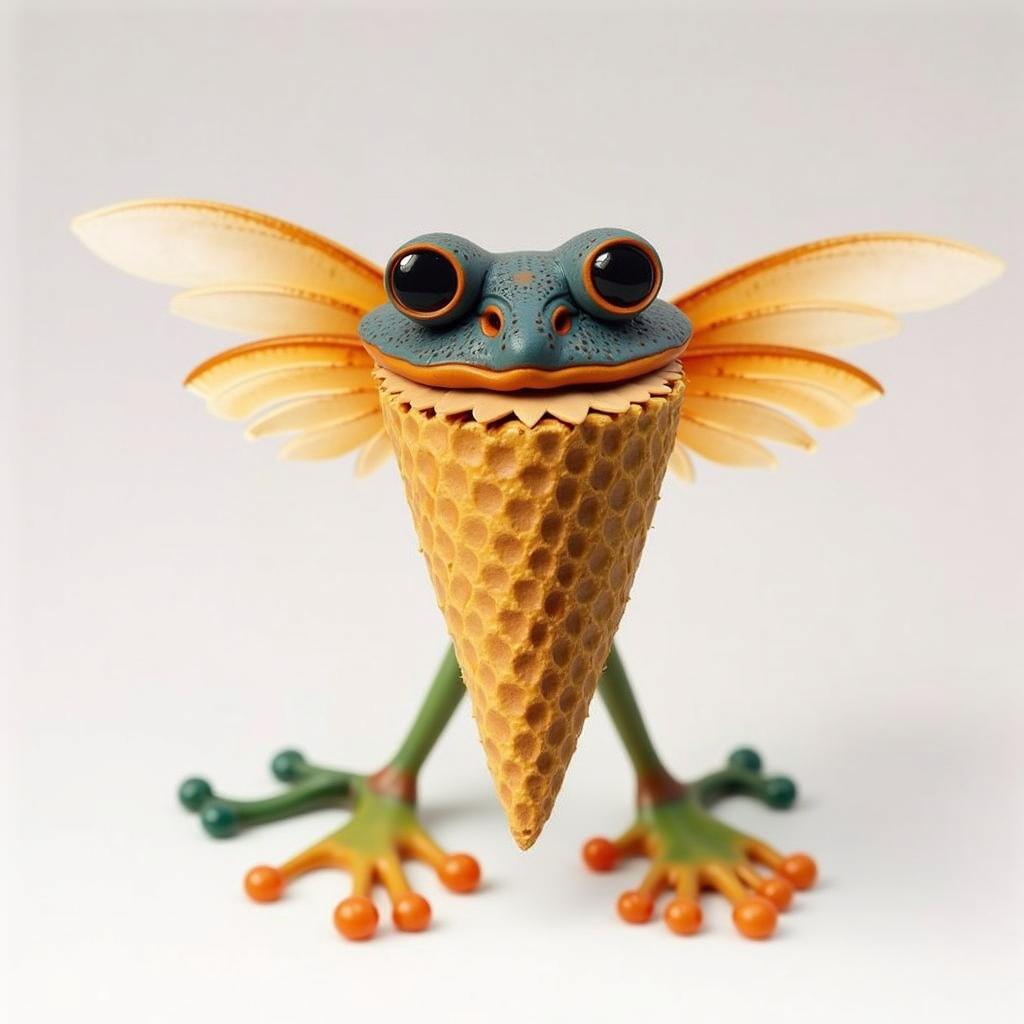} &

        \includegraphics[height=0.135\textwidth]{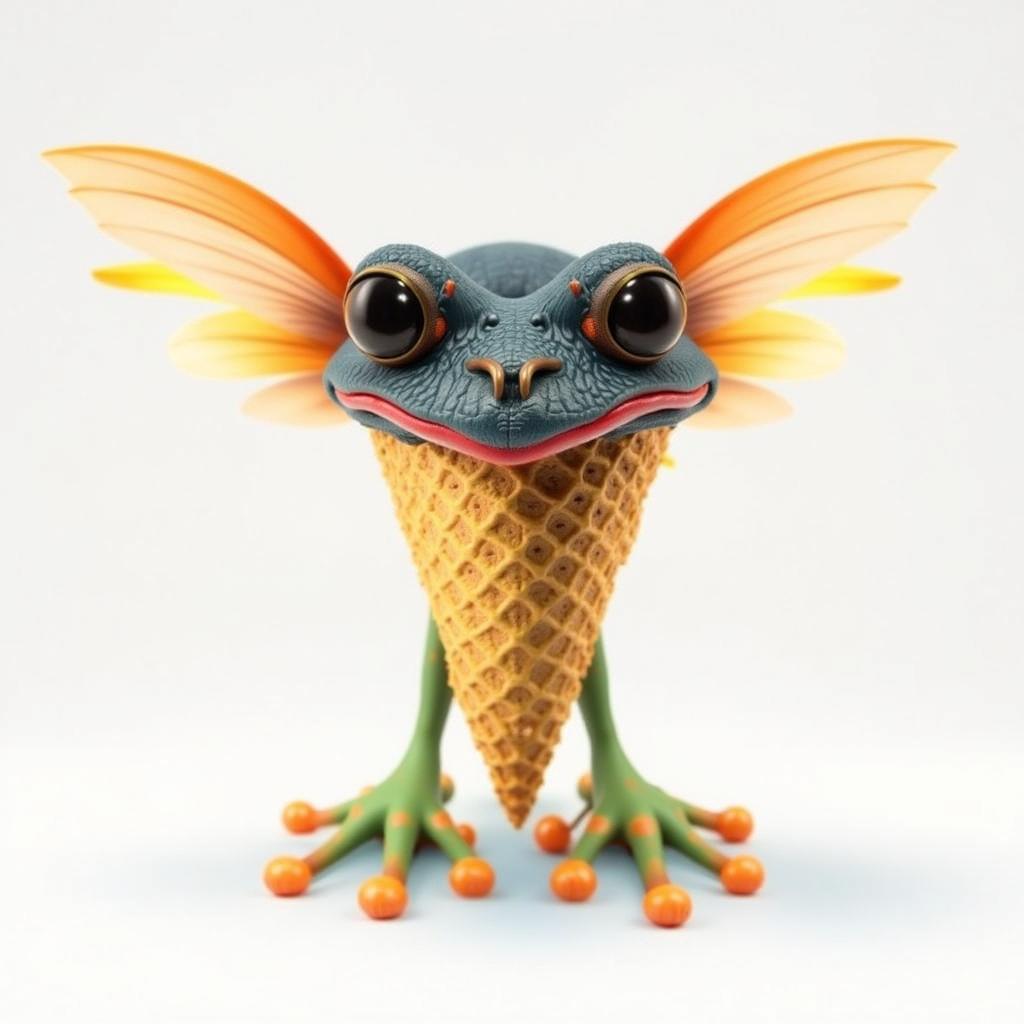}         
        \\

        \includegraphics[height=0.135\textwidth]{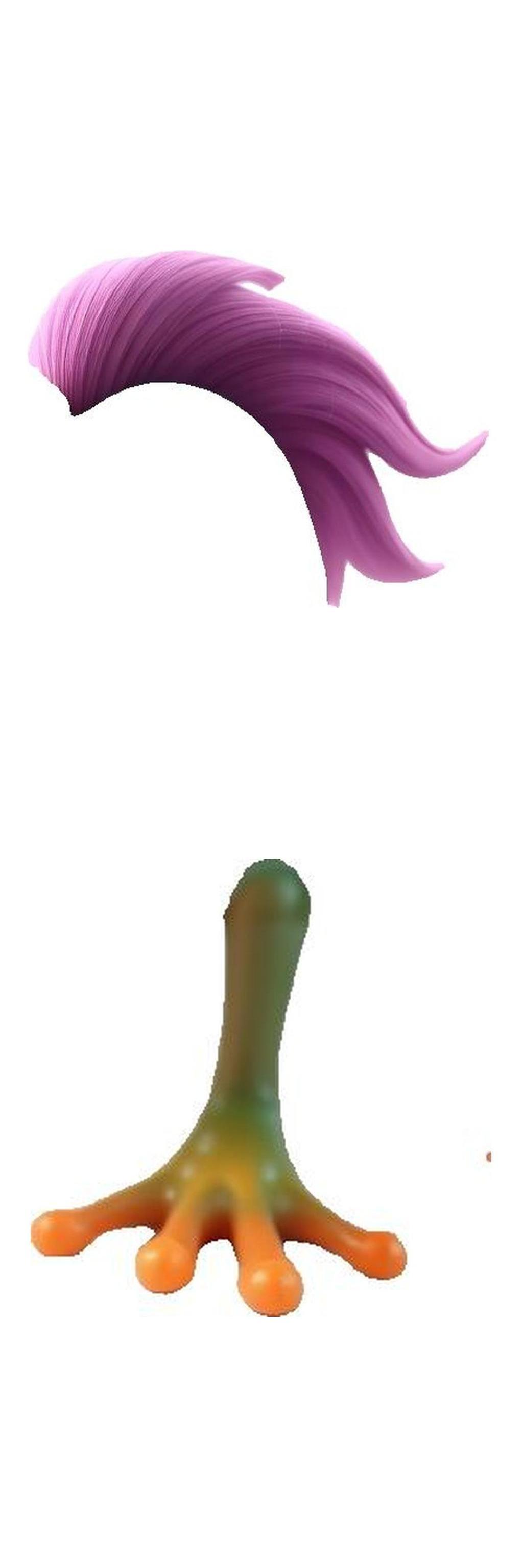} &
        \includegraphics[height=0.135\textwidth]{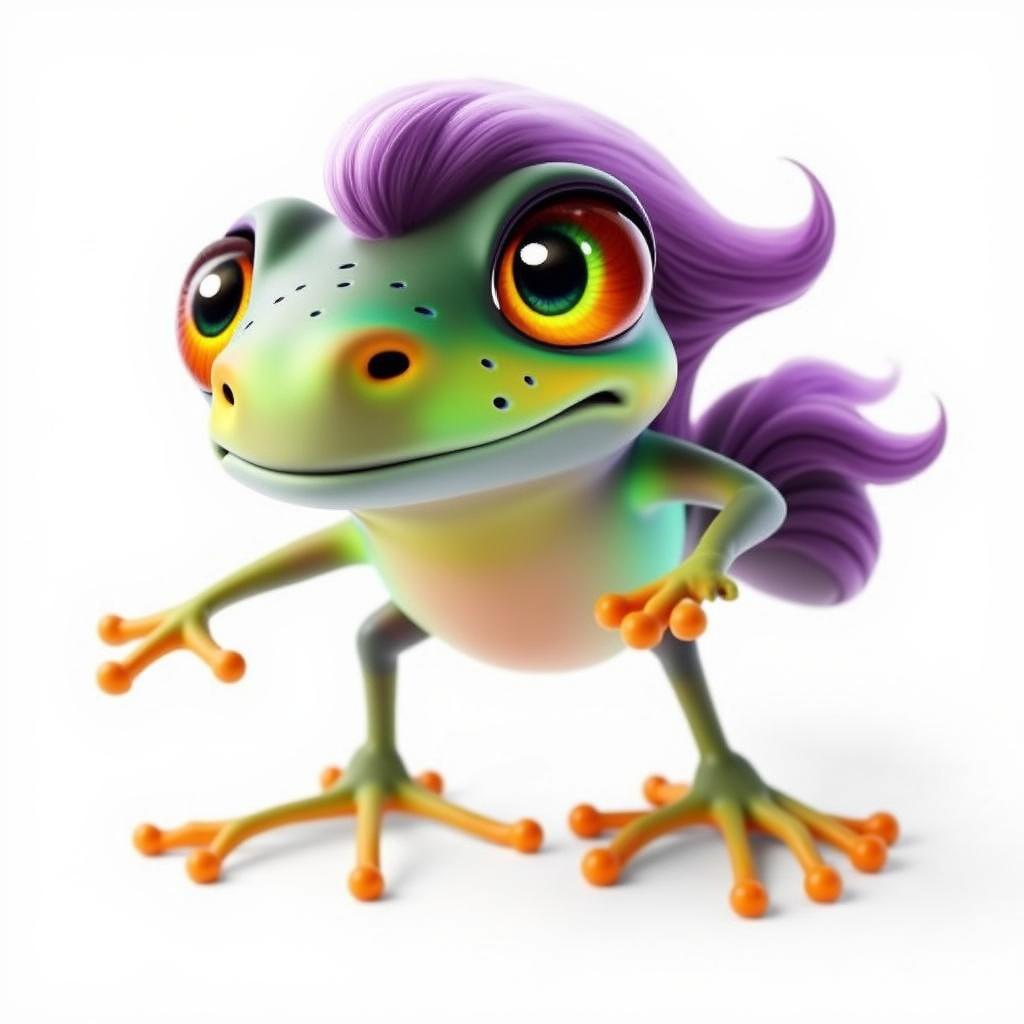} &
        \includegraphics[height=0.135\textwidth]{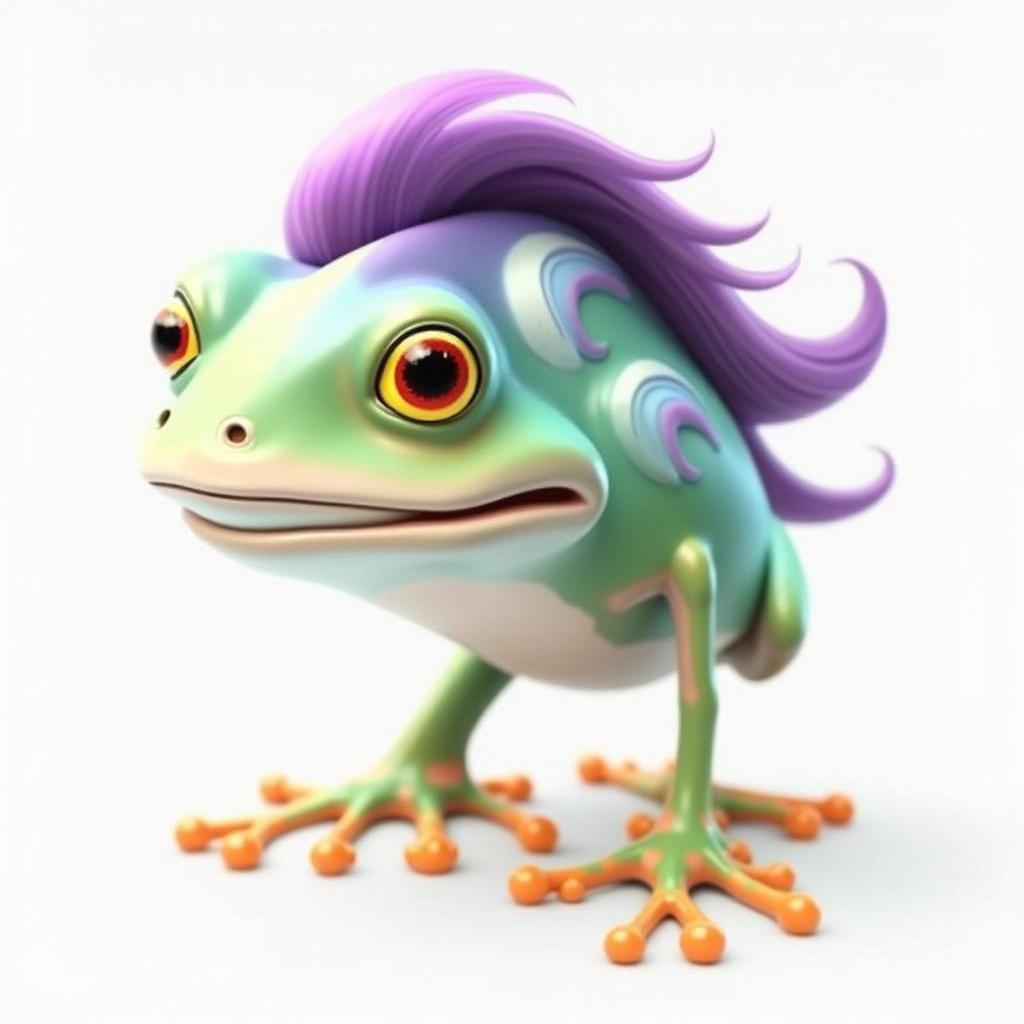} &

        \includegraphics[height=0.135\textwidth]{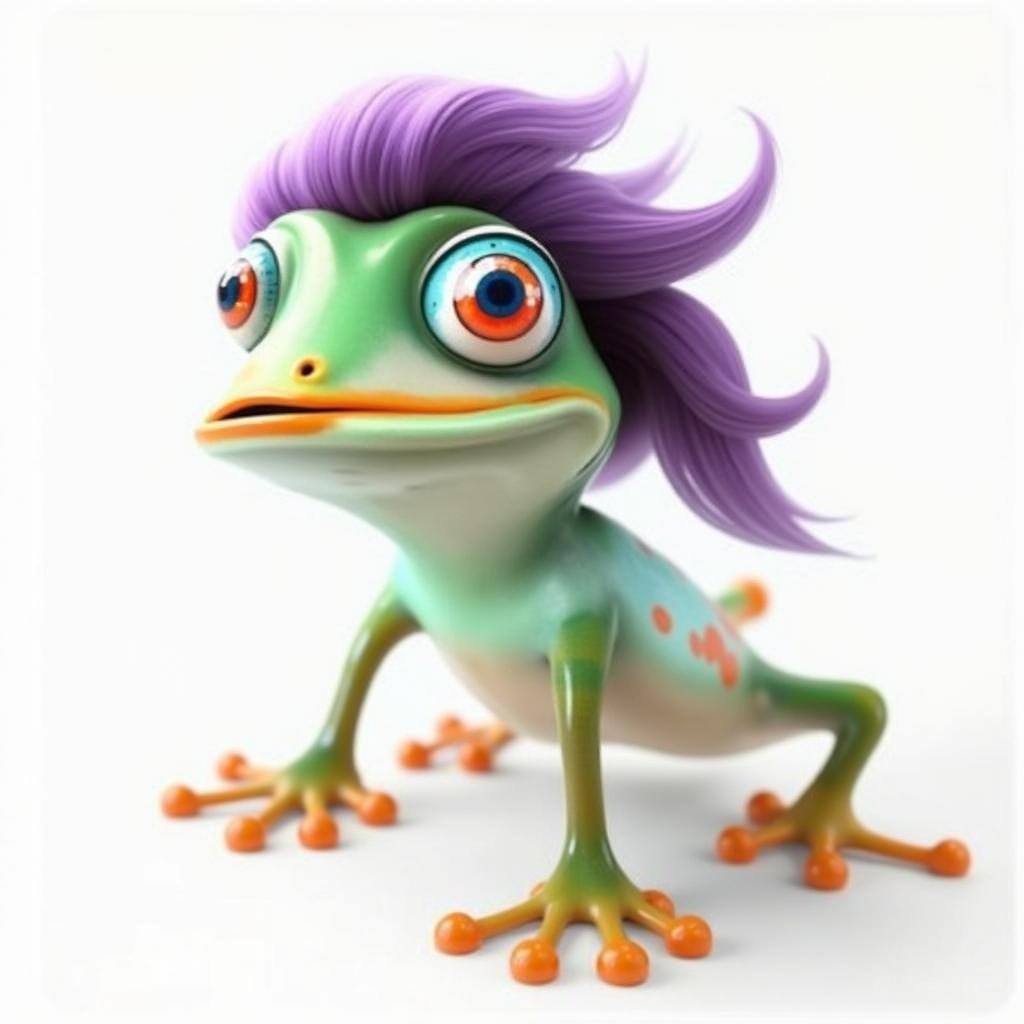} 
        &        

        \includegraphics[height=0.135\textwidth]{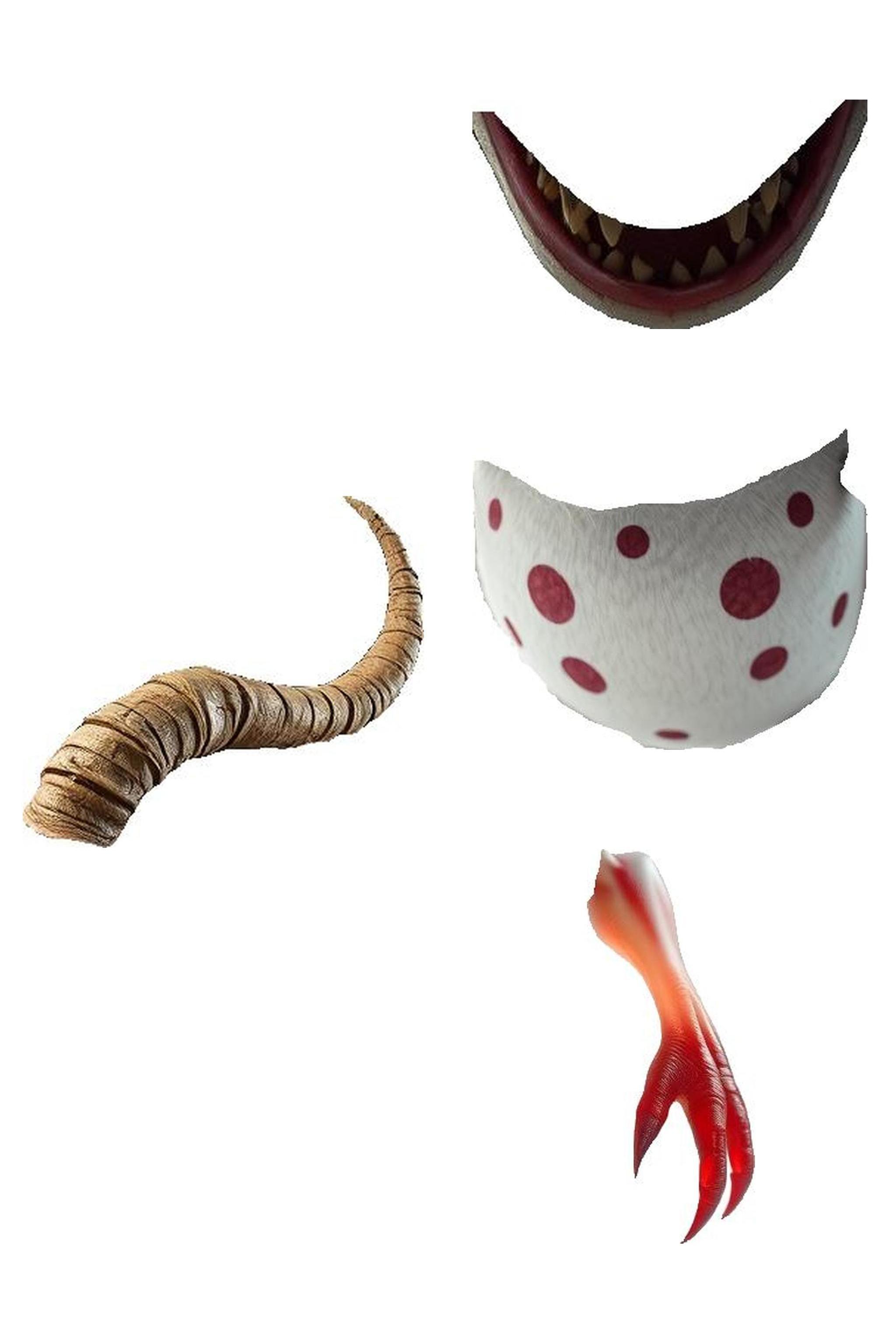} &
        \includegraphics[height=0.135\textwidth]{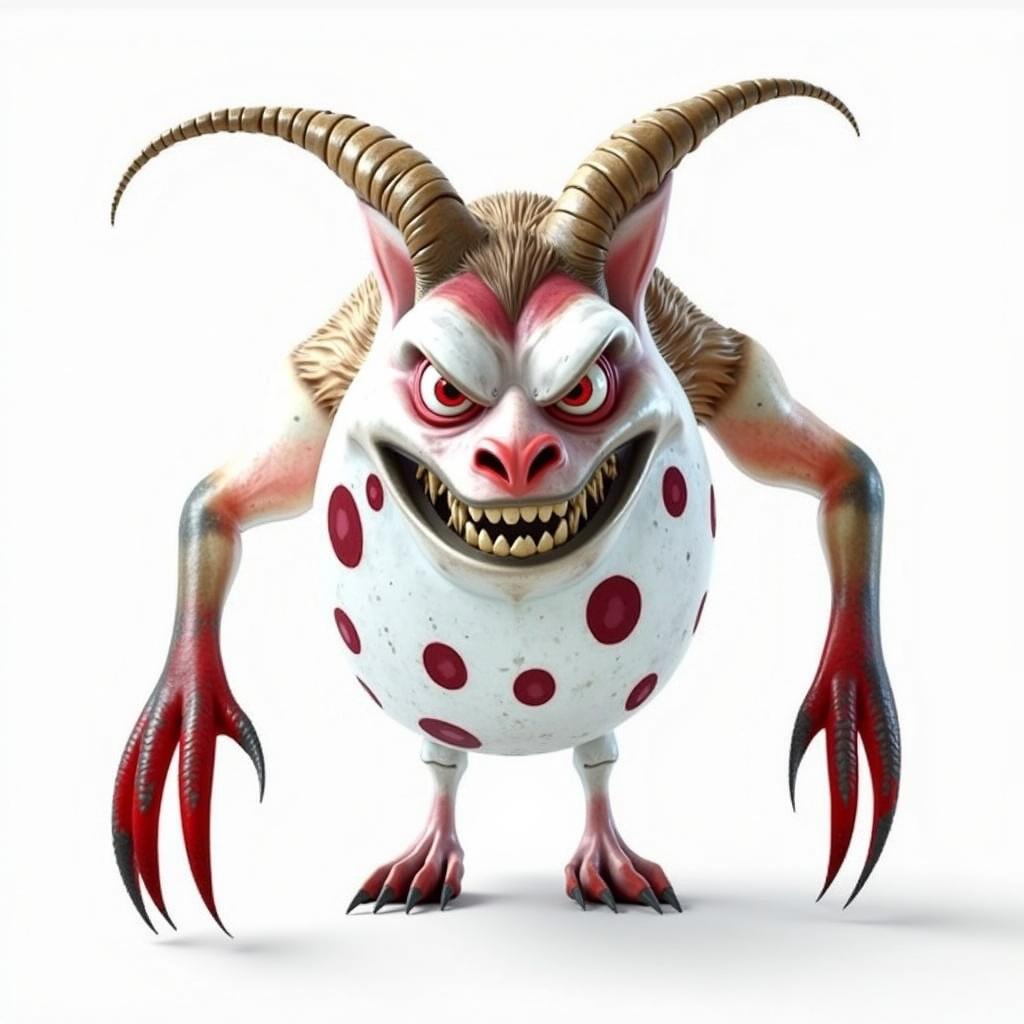} &
        \includegraphics[height=0.135\textwidth]{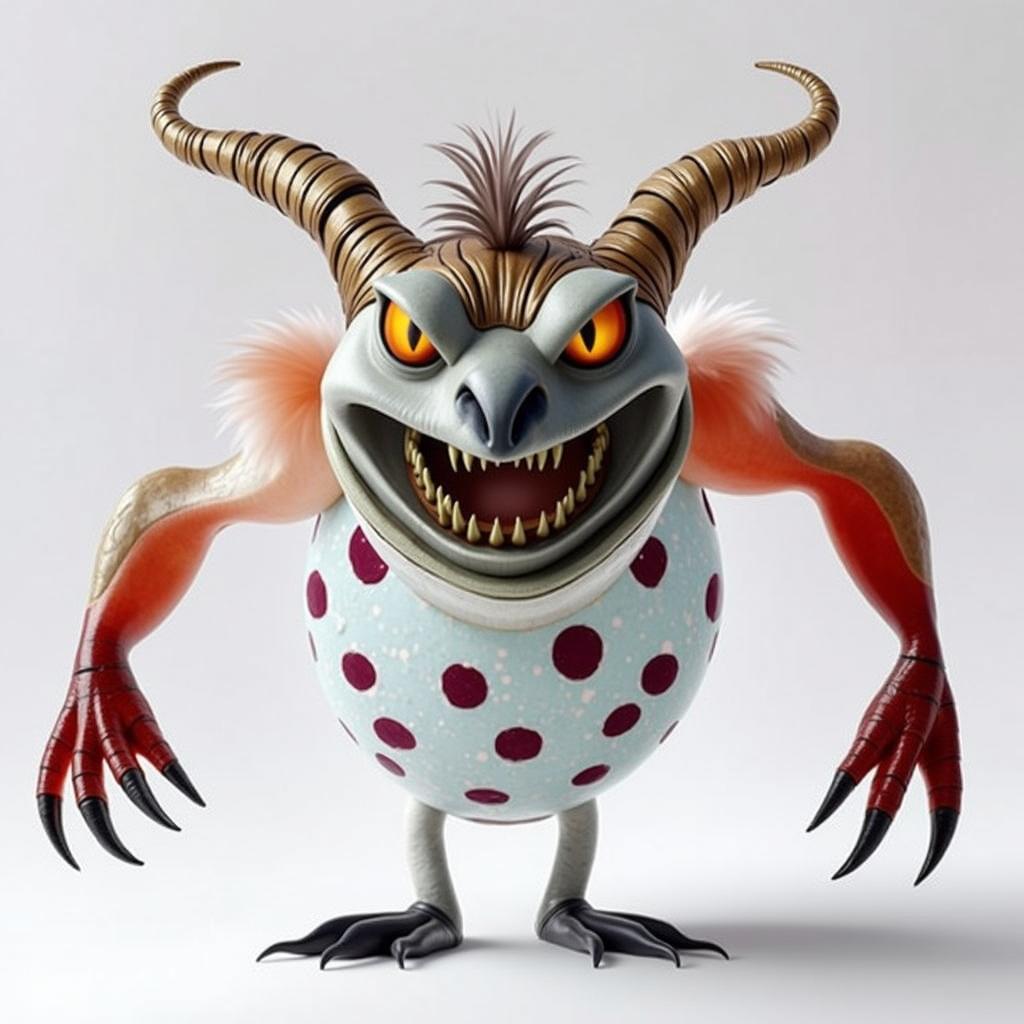} &

        \includegraphics[height=0.135\textwidth]{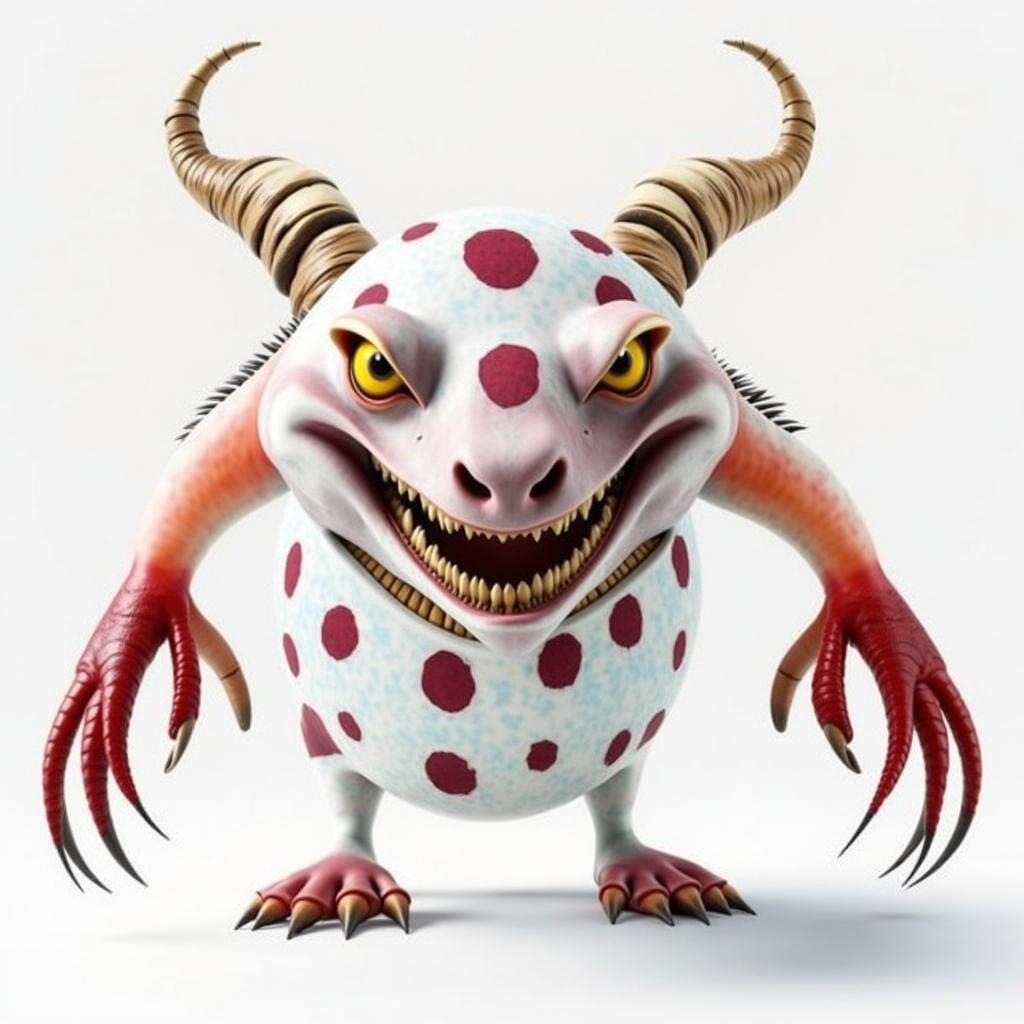} 
        \\   

        \includegraphics[height=0.135\textwidth]{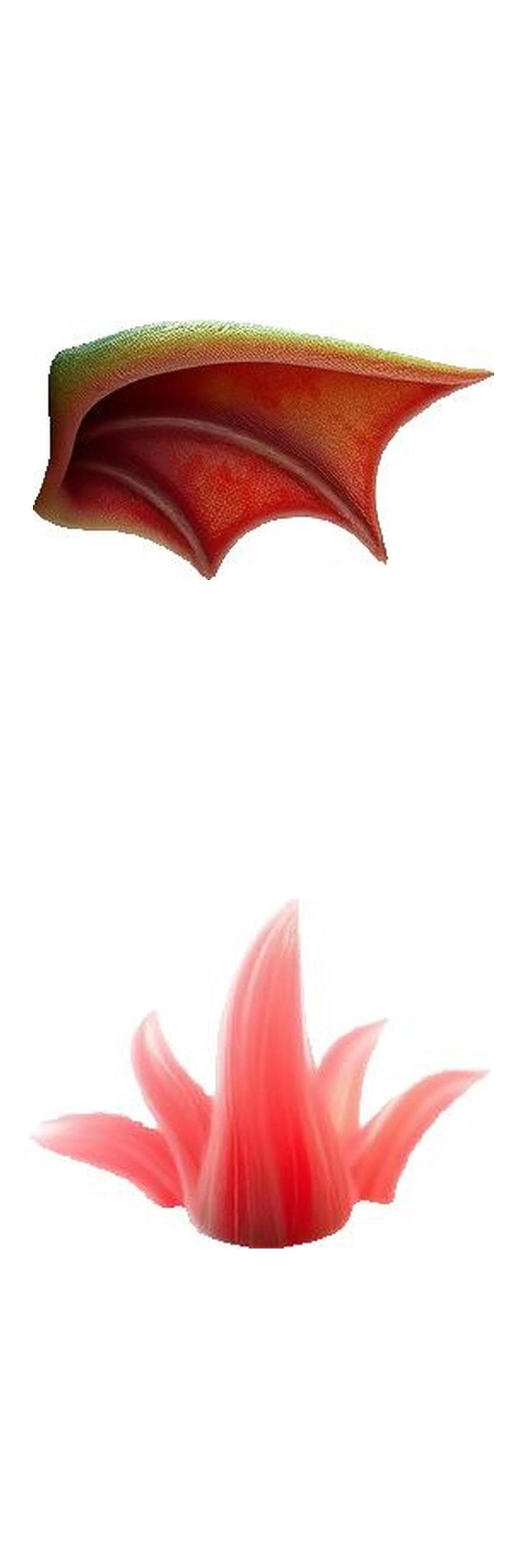} &
        \includegraphics[height=0.135\textwidth]{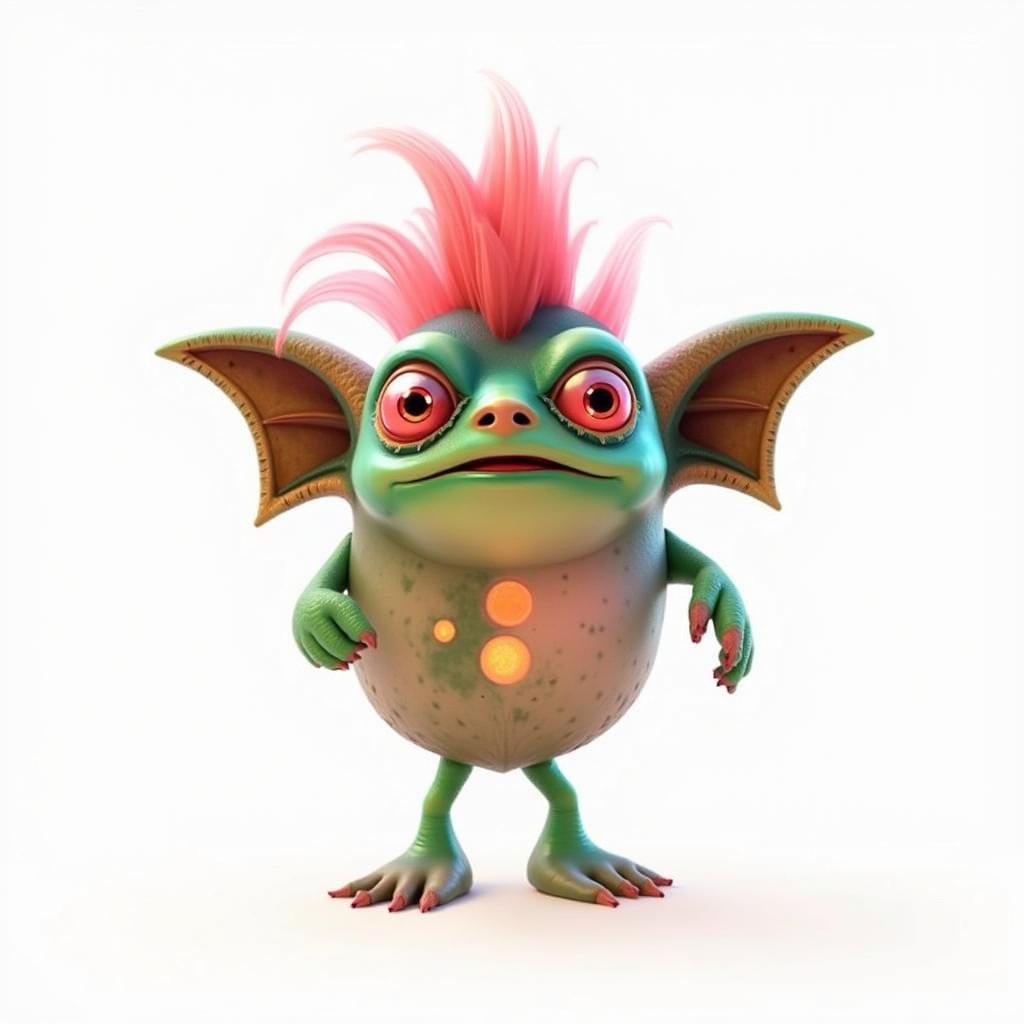} &
        \includegraphics[height=0.135\textwidth]{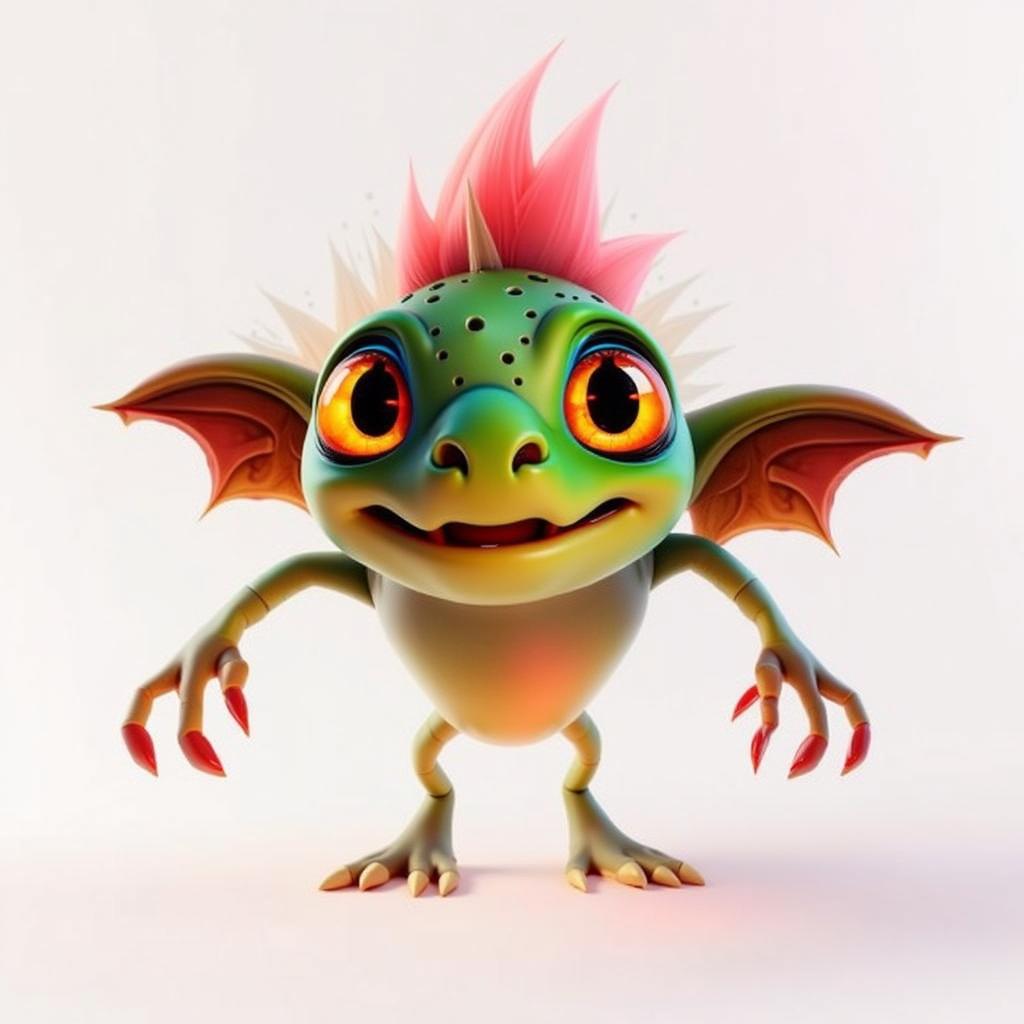} &

        \includegraphics[height=0.135\textwidth]{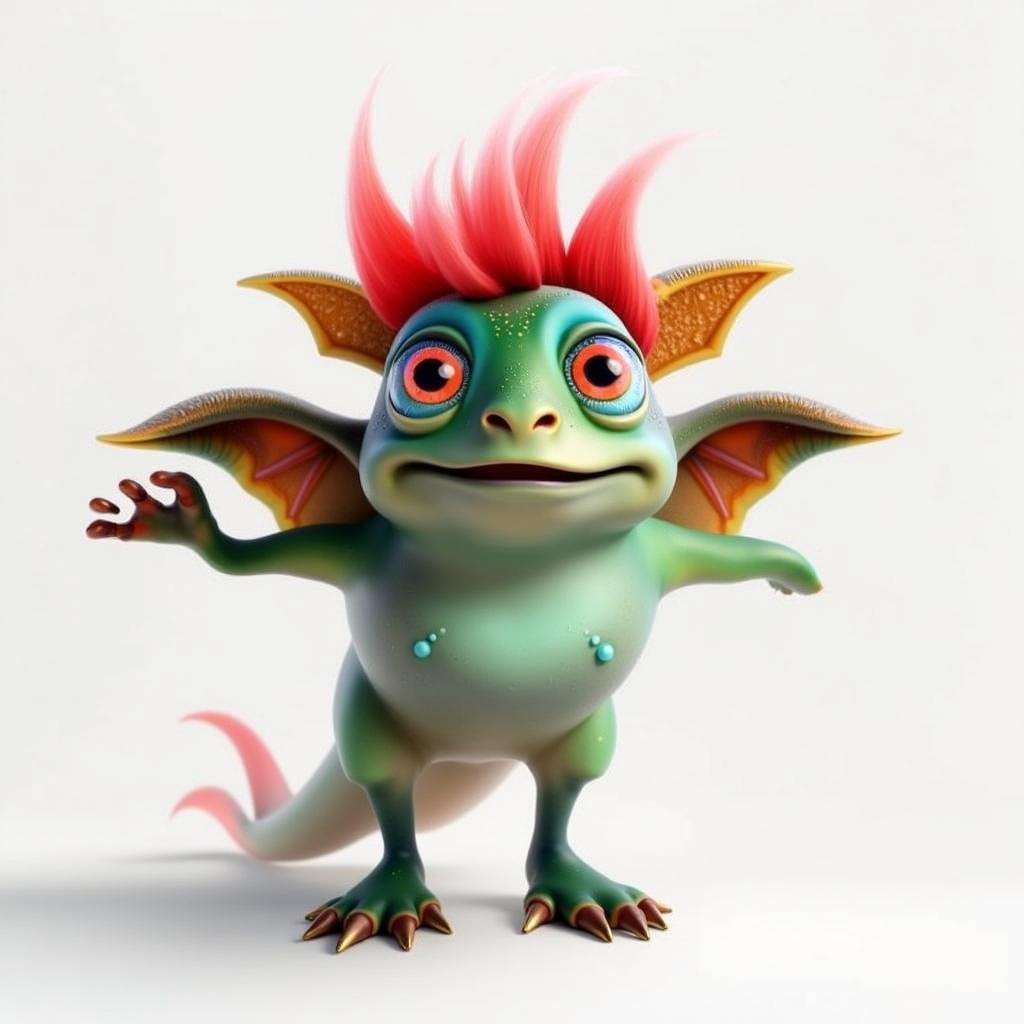} 
        &

        \includegraphics[height=0.135\textwidth]{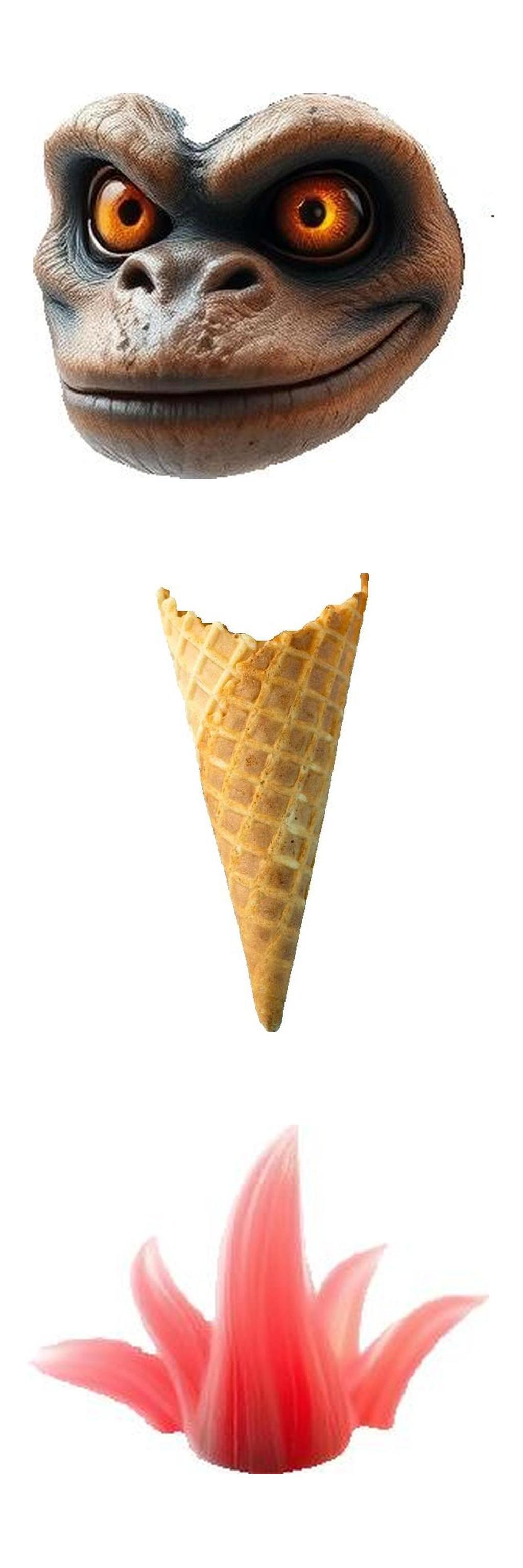} &
        \includegraphics[height=0.135\textwidth]{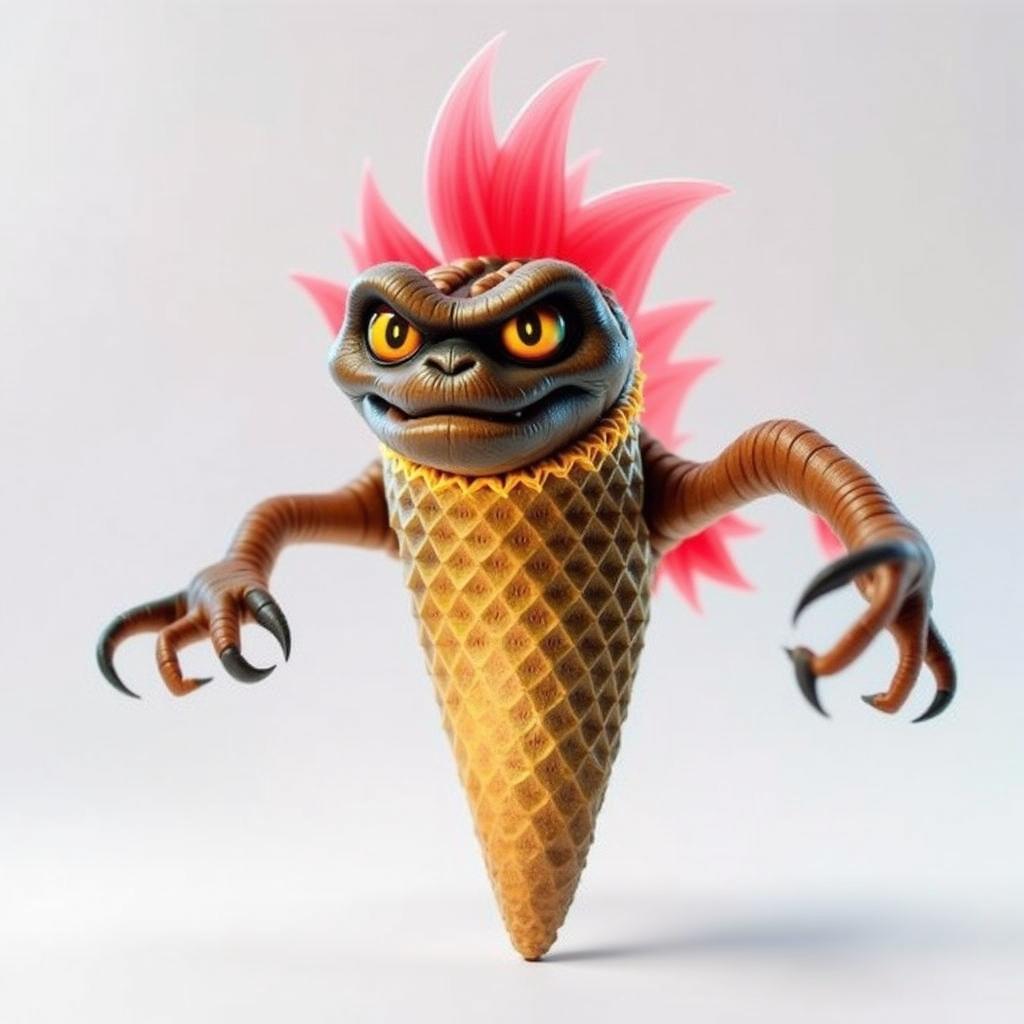} &
        \includegraphics[height=0.135\textwidth]{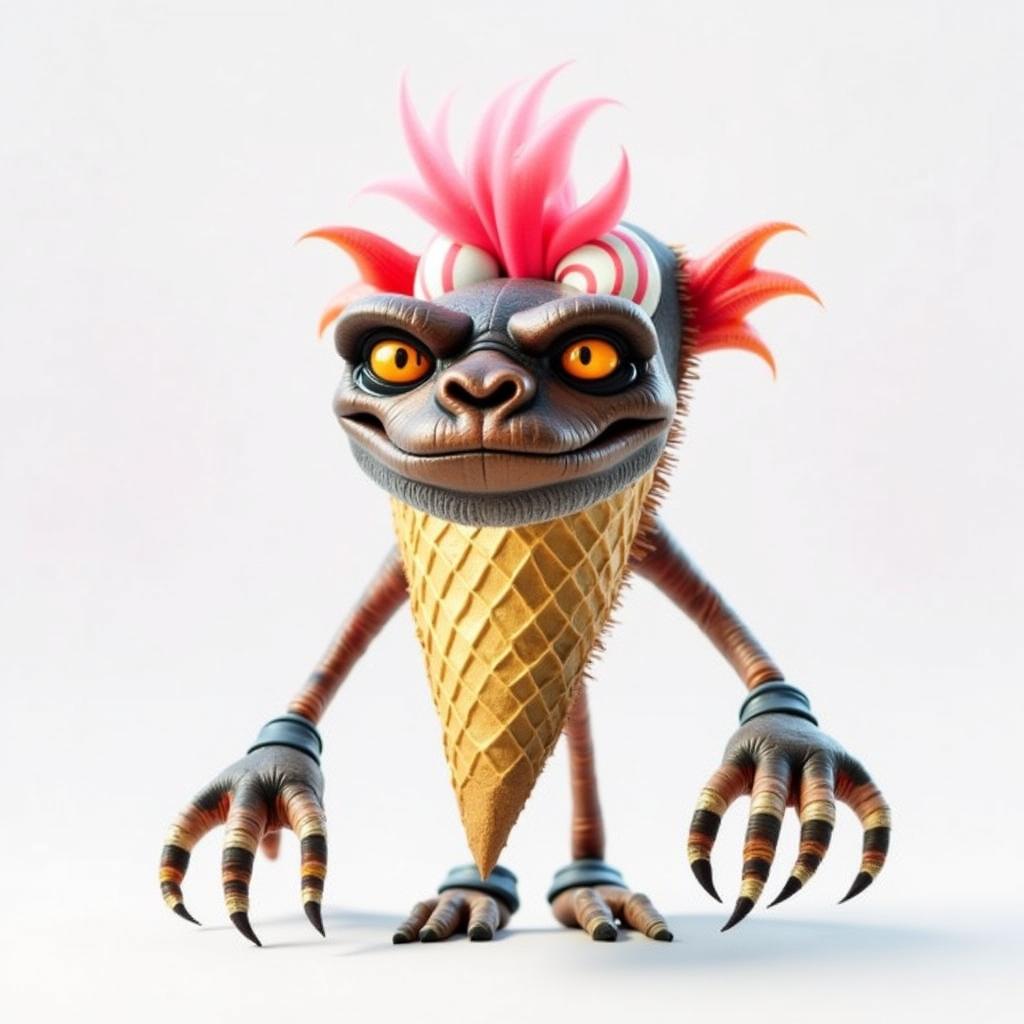} &

        \includegraphics[height=0.135\textwidth]{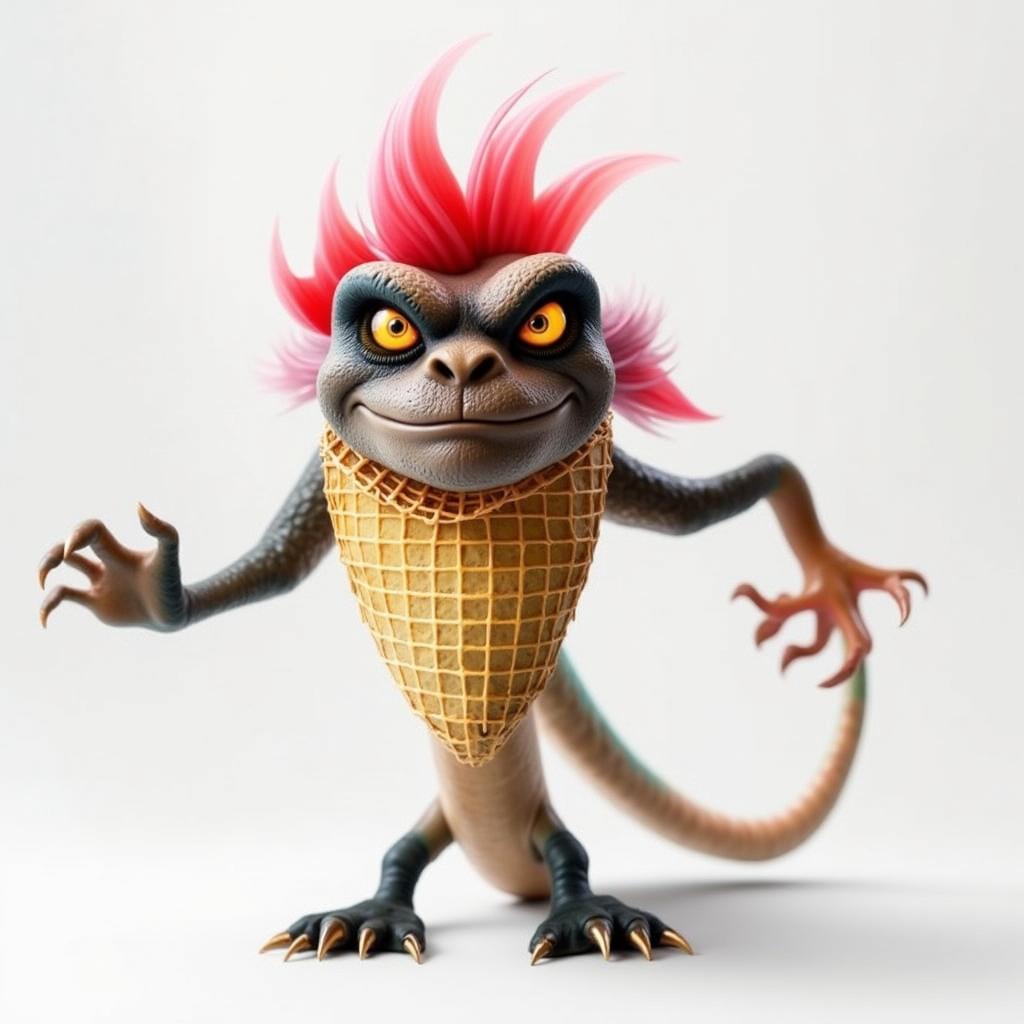} 
        \\

        \includegraphics[height=0.135\textwidth]{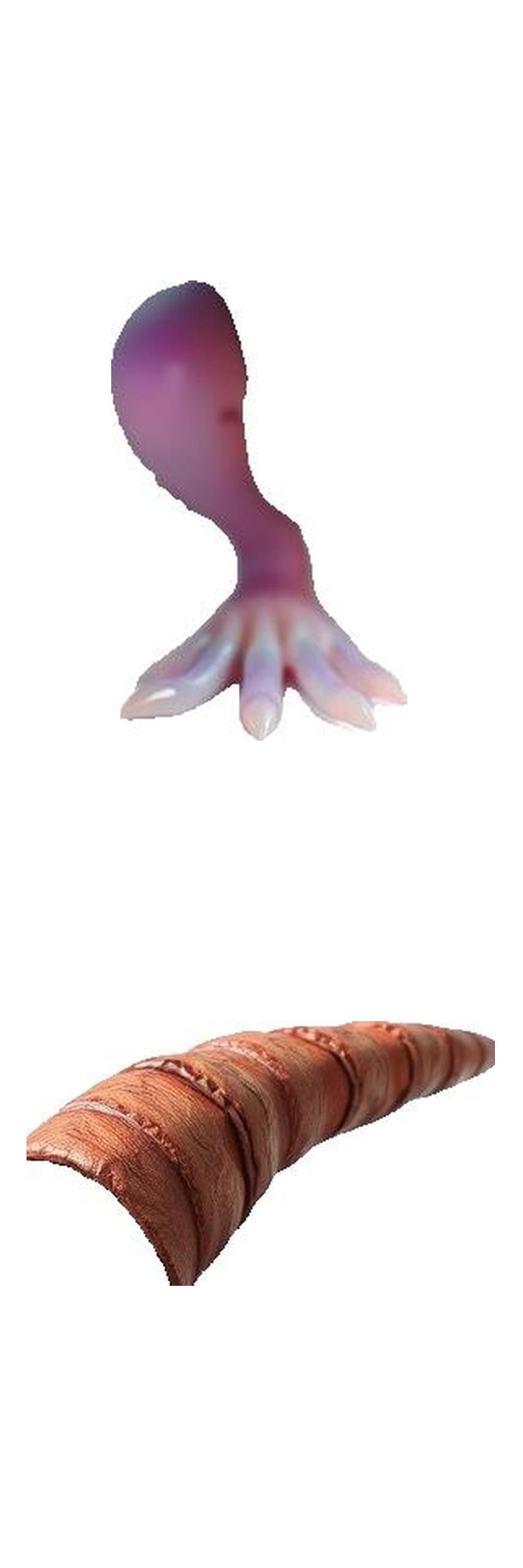} &

        \includegraphics[height=0.135\textwidth]{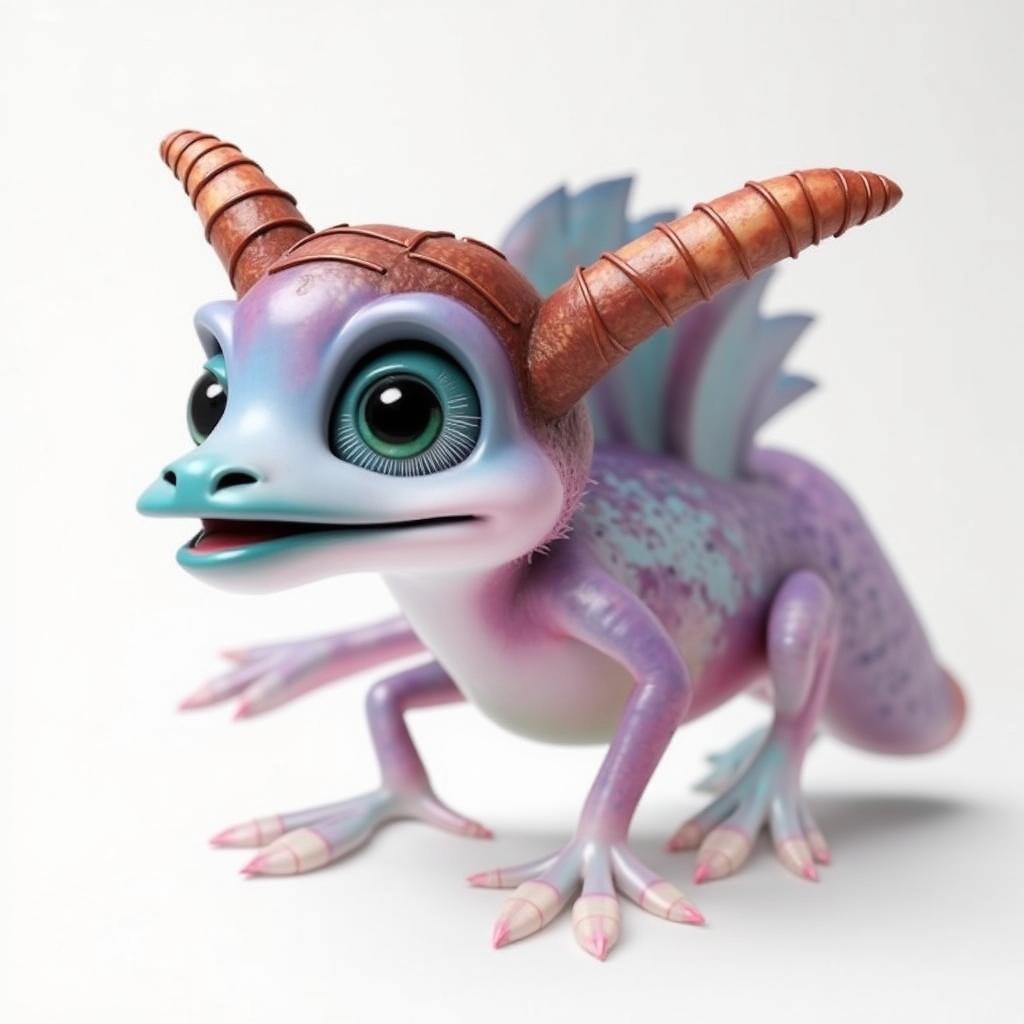} &

        \includegraphics[height=0.135\textwidth]{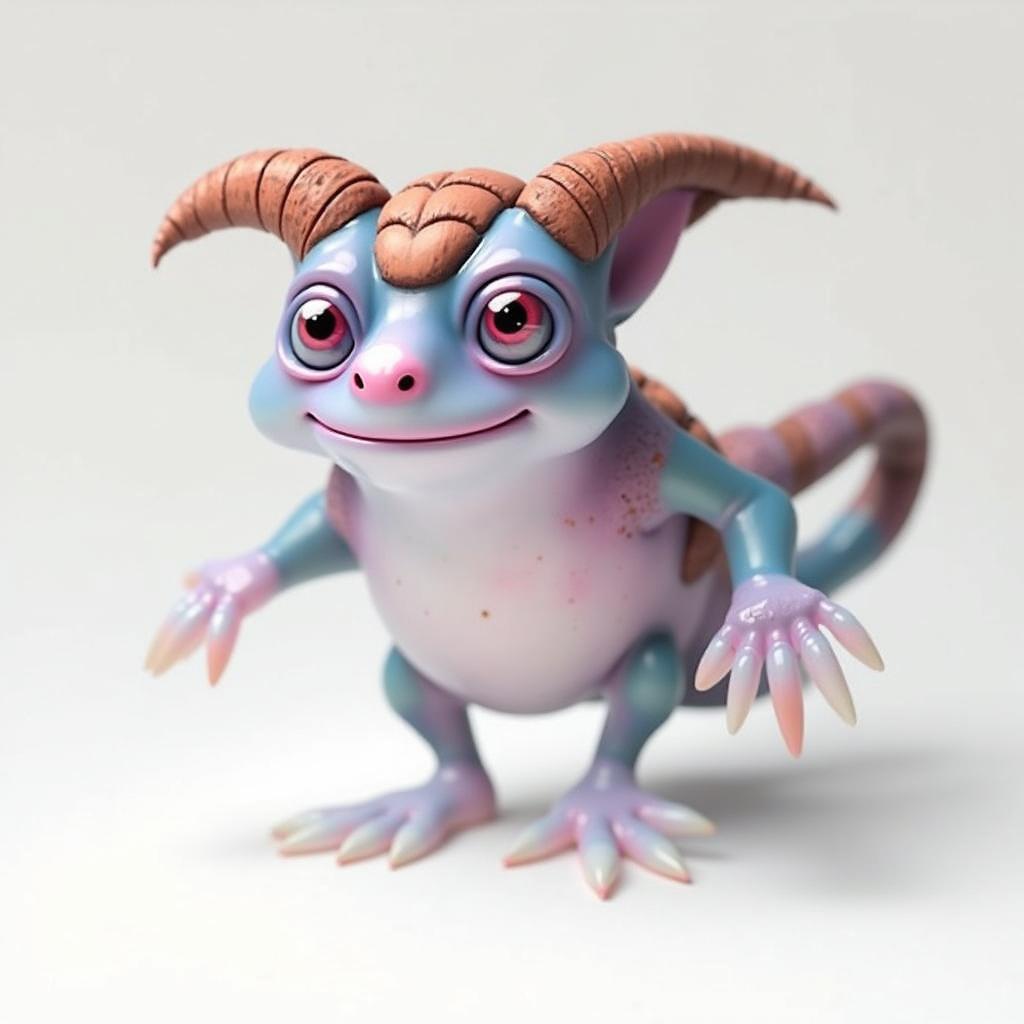} &

        \includegraphics[height=0.135\textwidth]{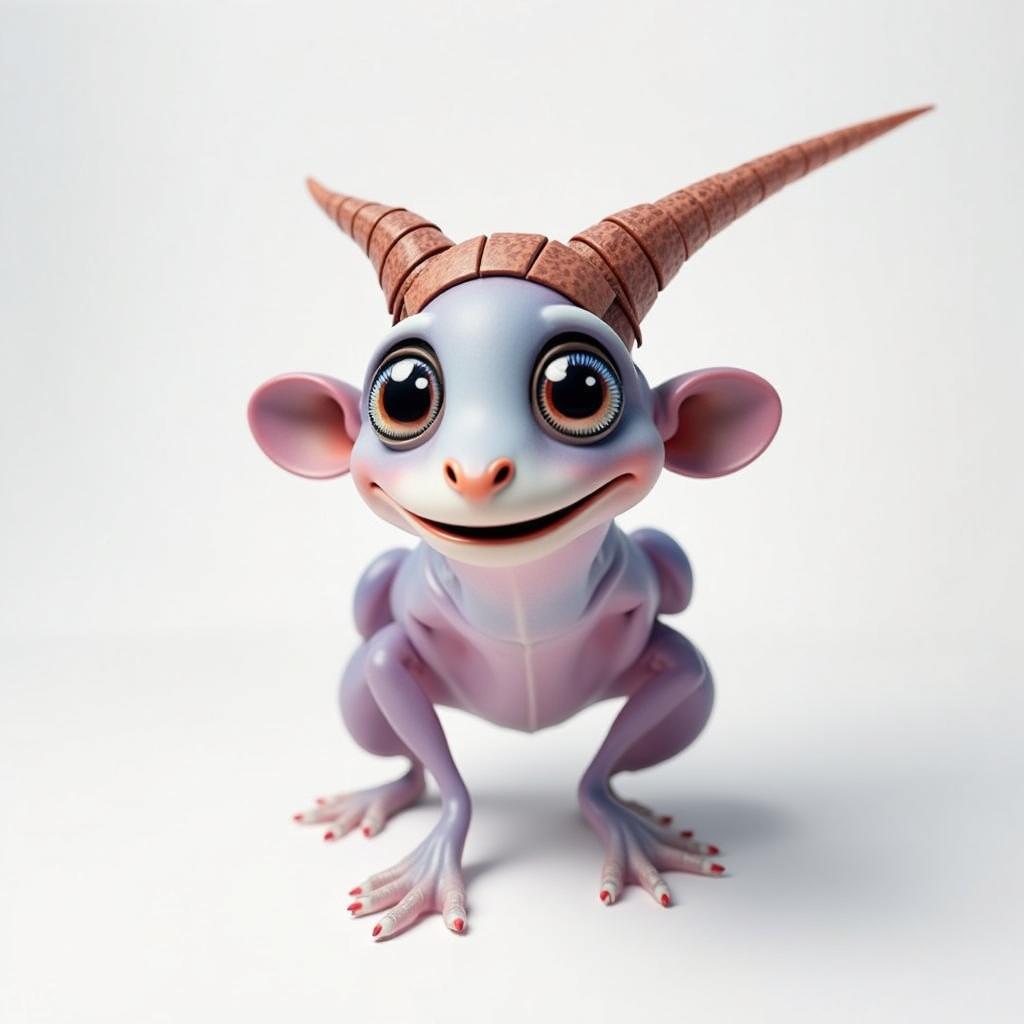} &

        \includegraphics[height=0.135\textwidth]{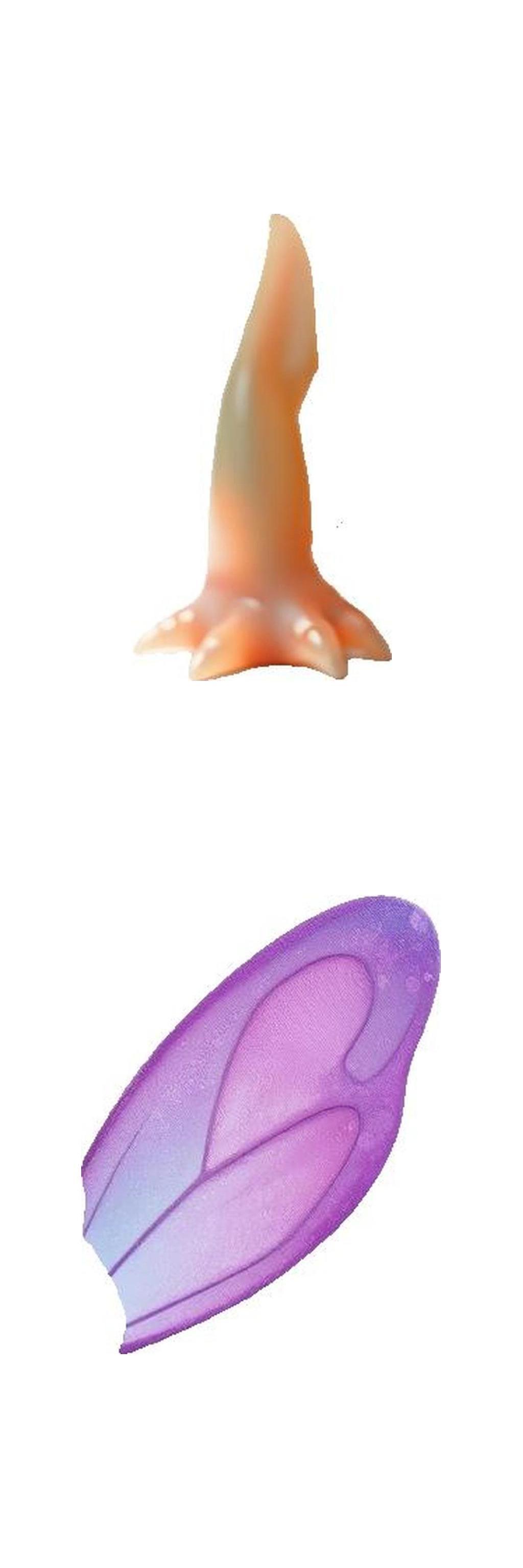} &

        \includegraphics[height=0.135\textwidth]{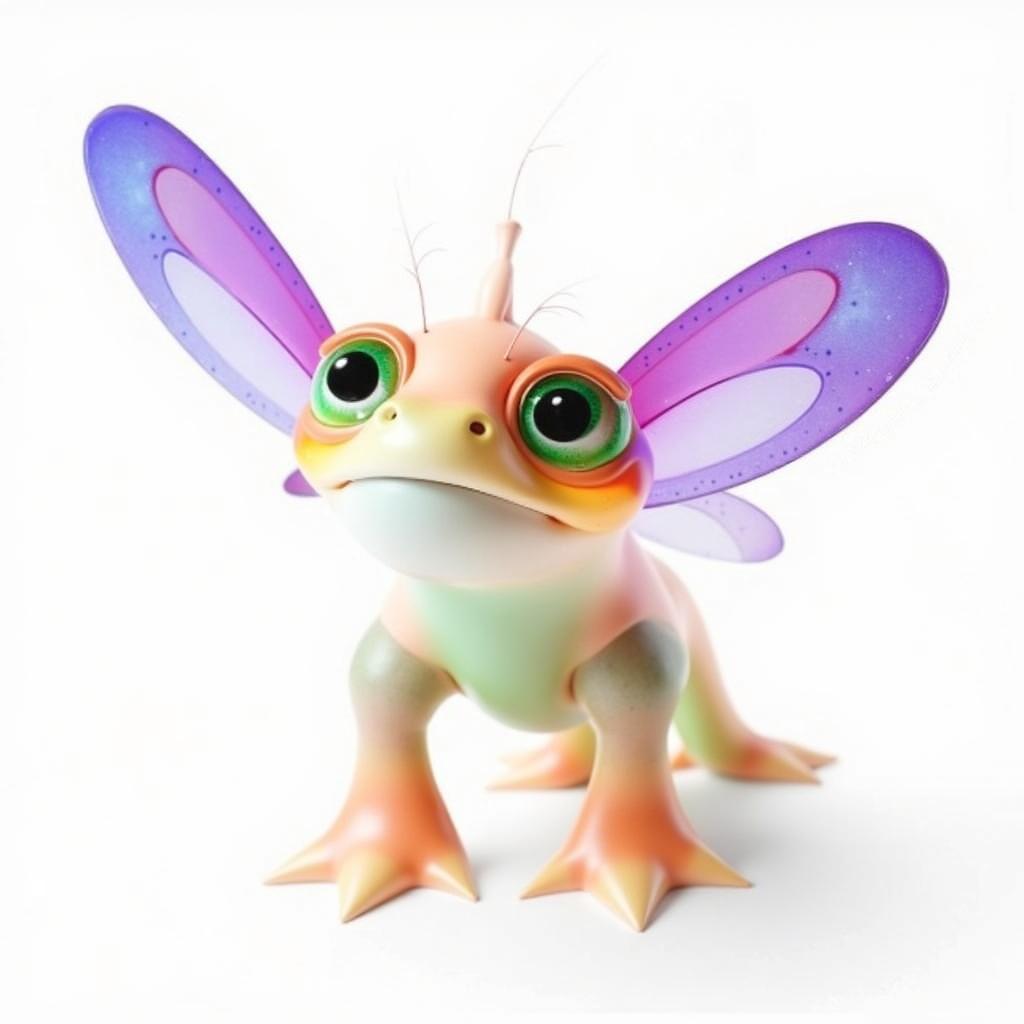} &

        \includegraphics[height=0.135\textwidth]{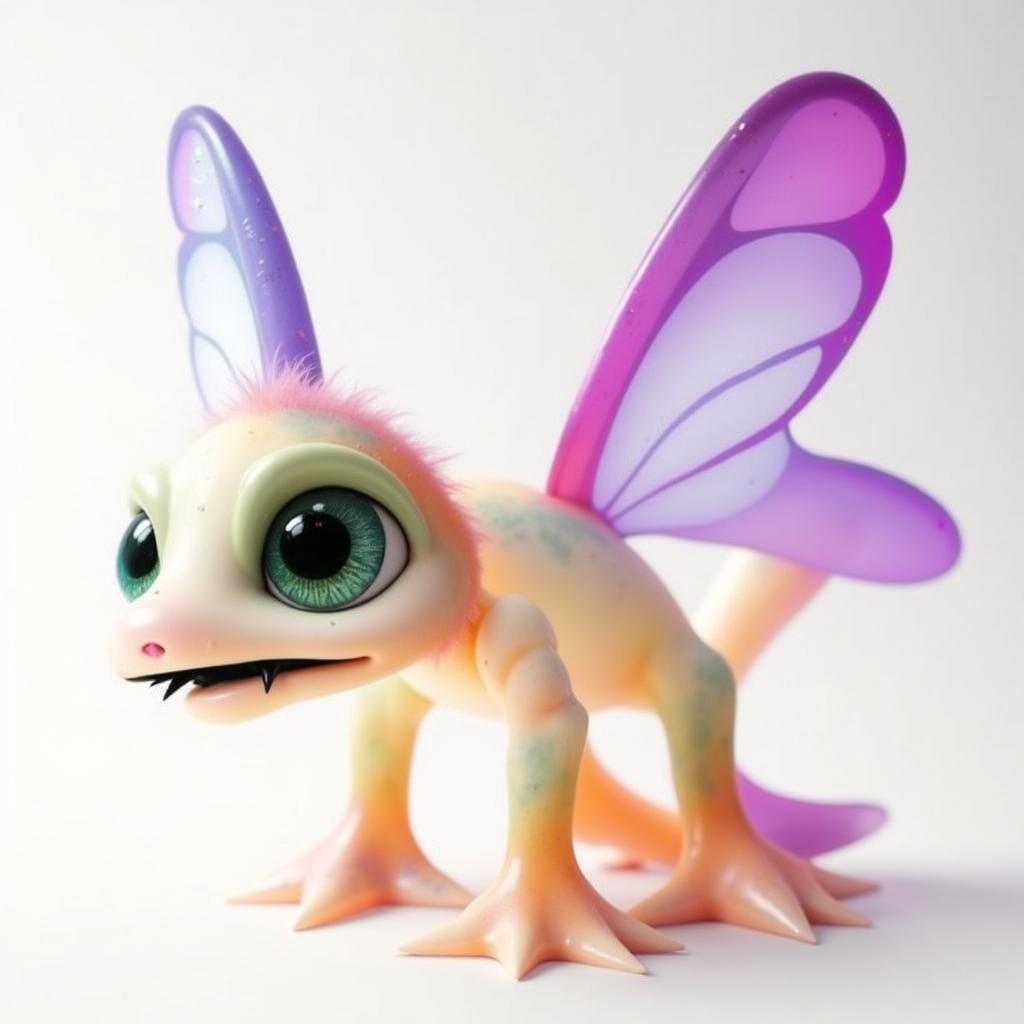} &

        \includegraphics[height=0.135\textwidth]{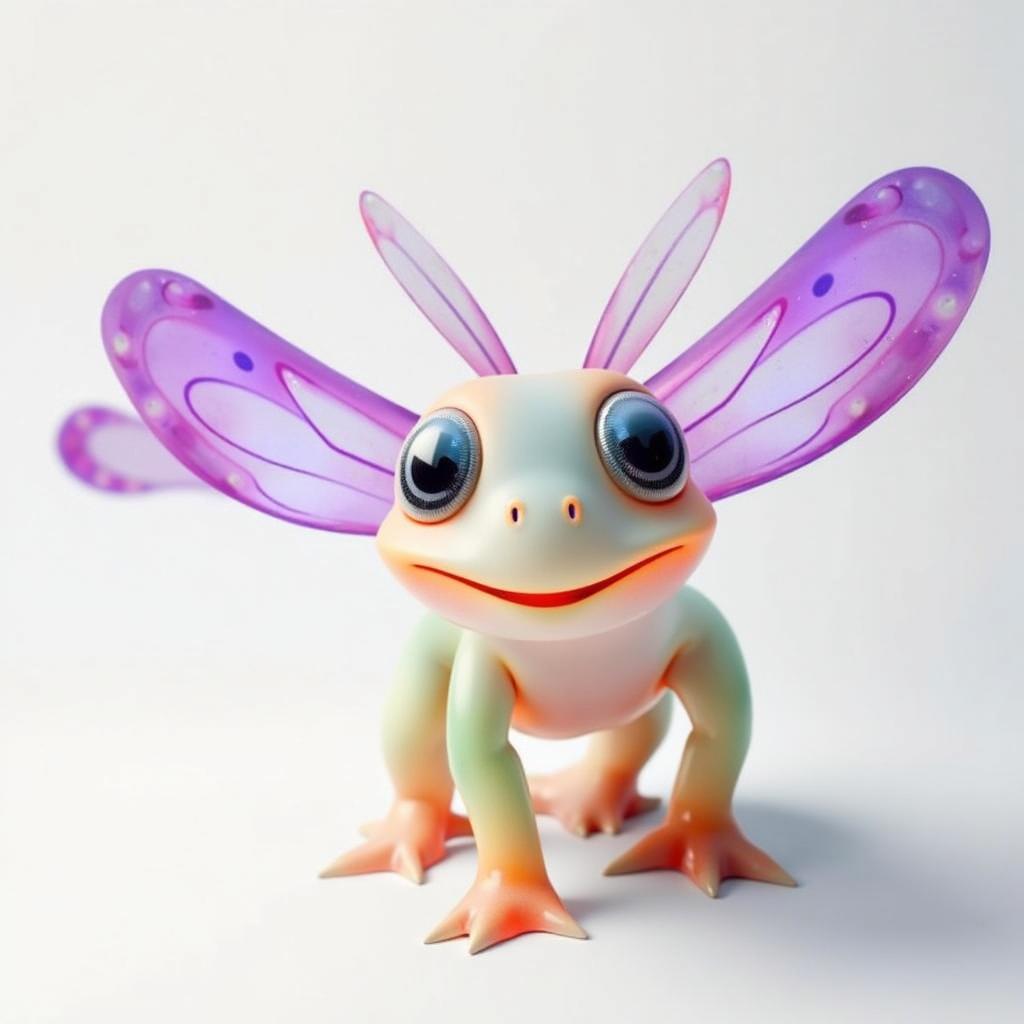} 
        
        \\

        \includegraphics[height=0.135\textwidth]{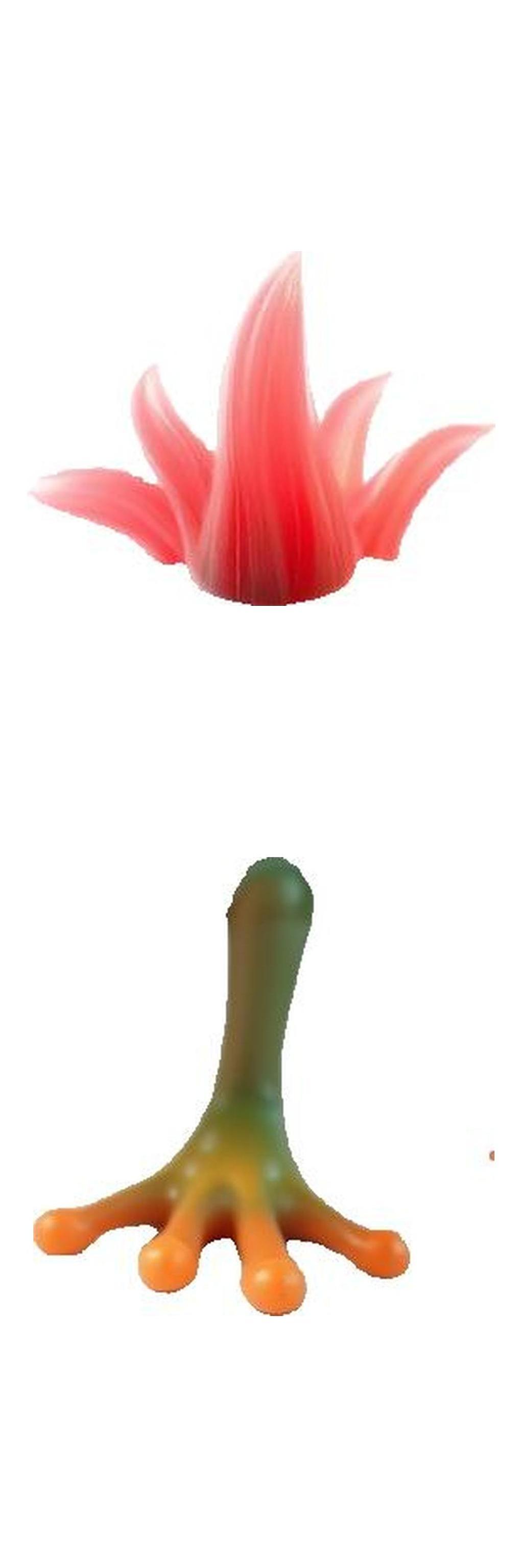} &

        \includegraphics[height=0.135\textwidth]{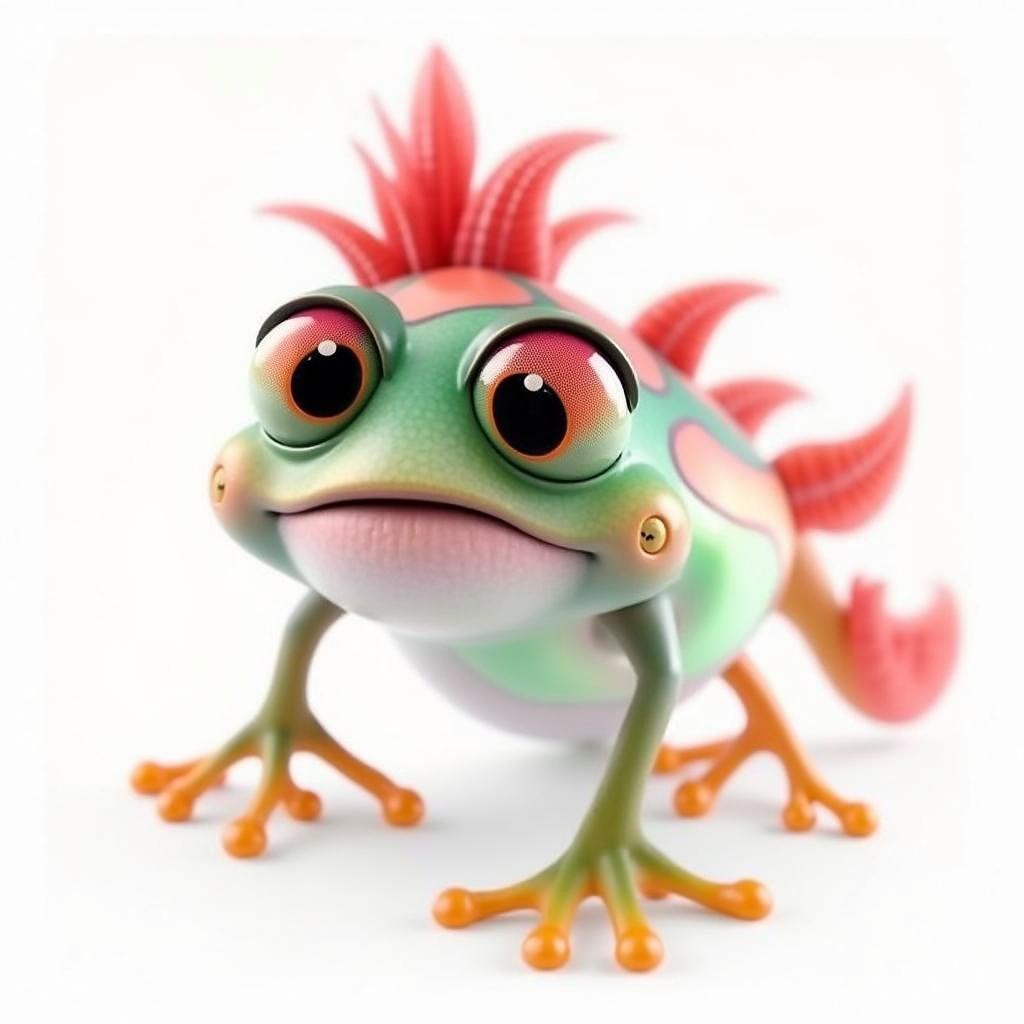} &

        \includegraphics[height=0.135\textwidth]{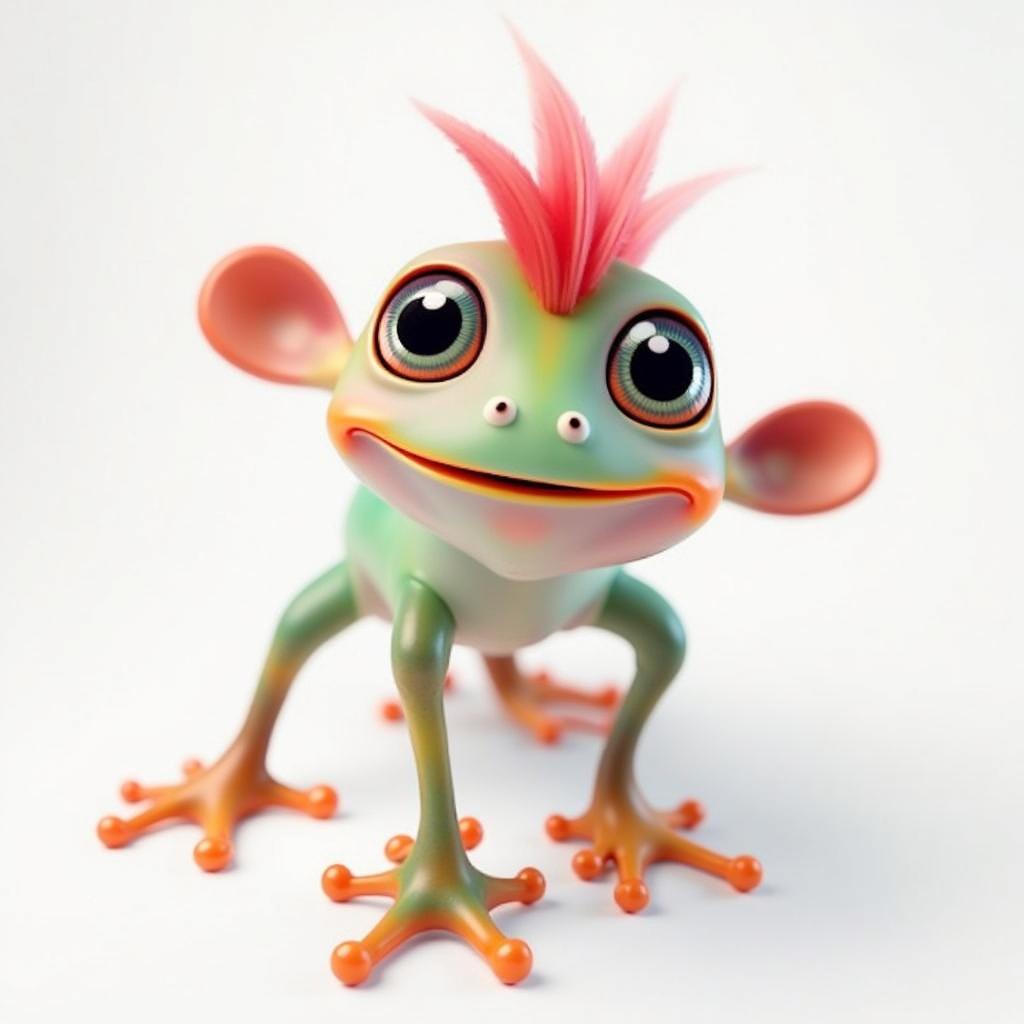} &

        \includegraphics[height=0.135\textwidth]{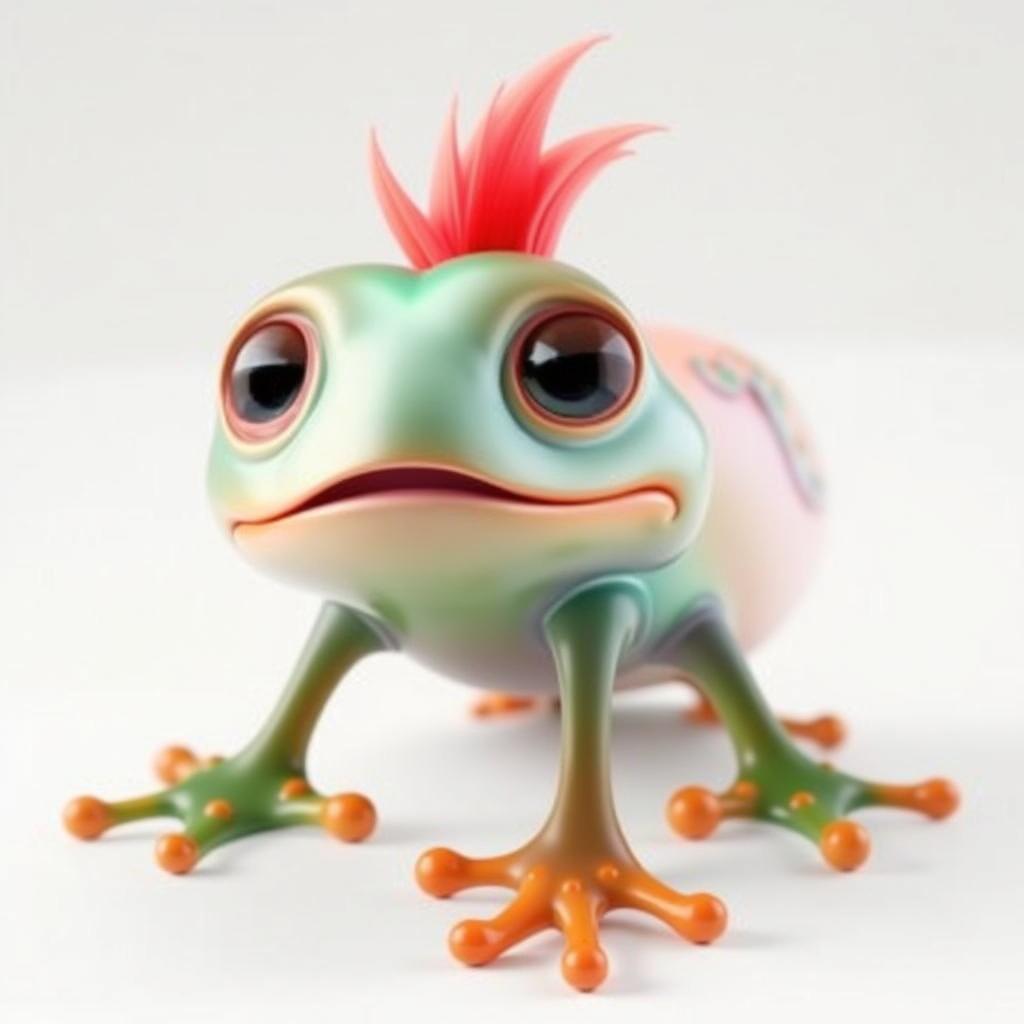} &

        \includegraphics[height=0.135\textwidth]{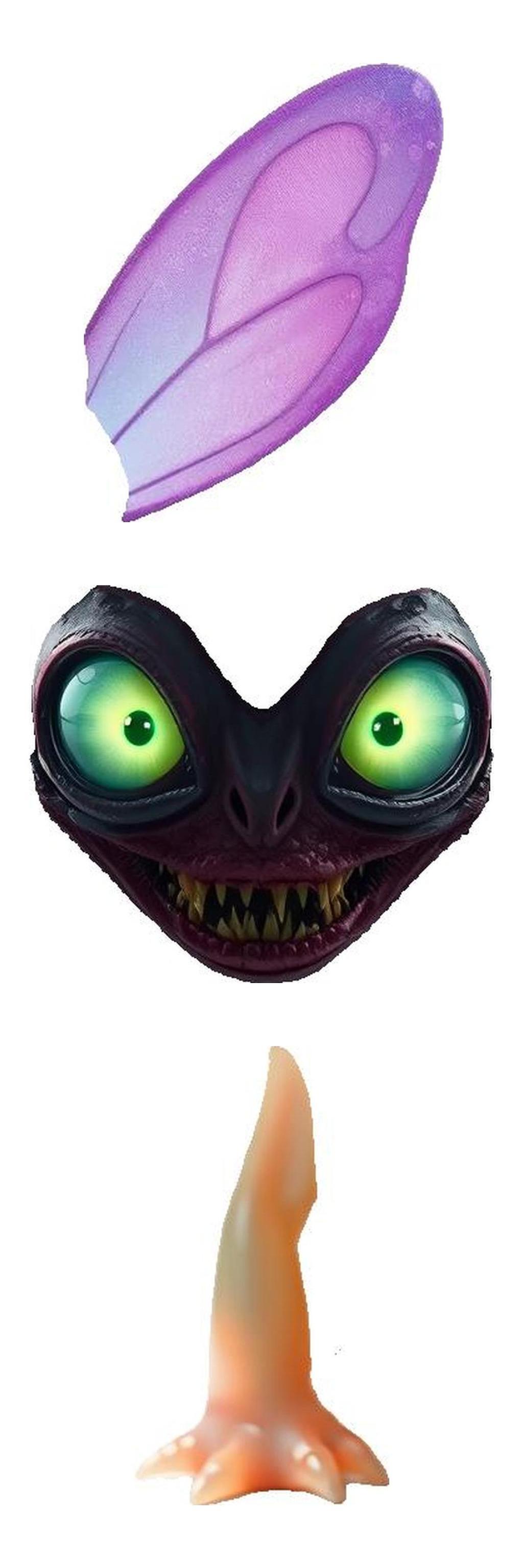} &

        \includegraphics[height=0.135\textwidth]{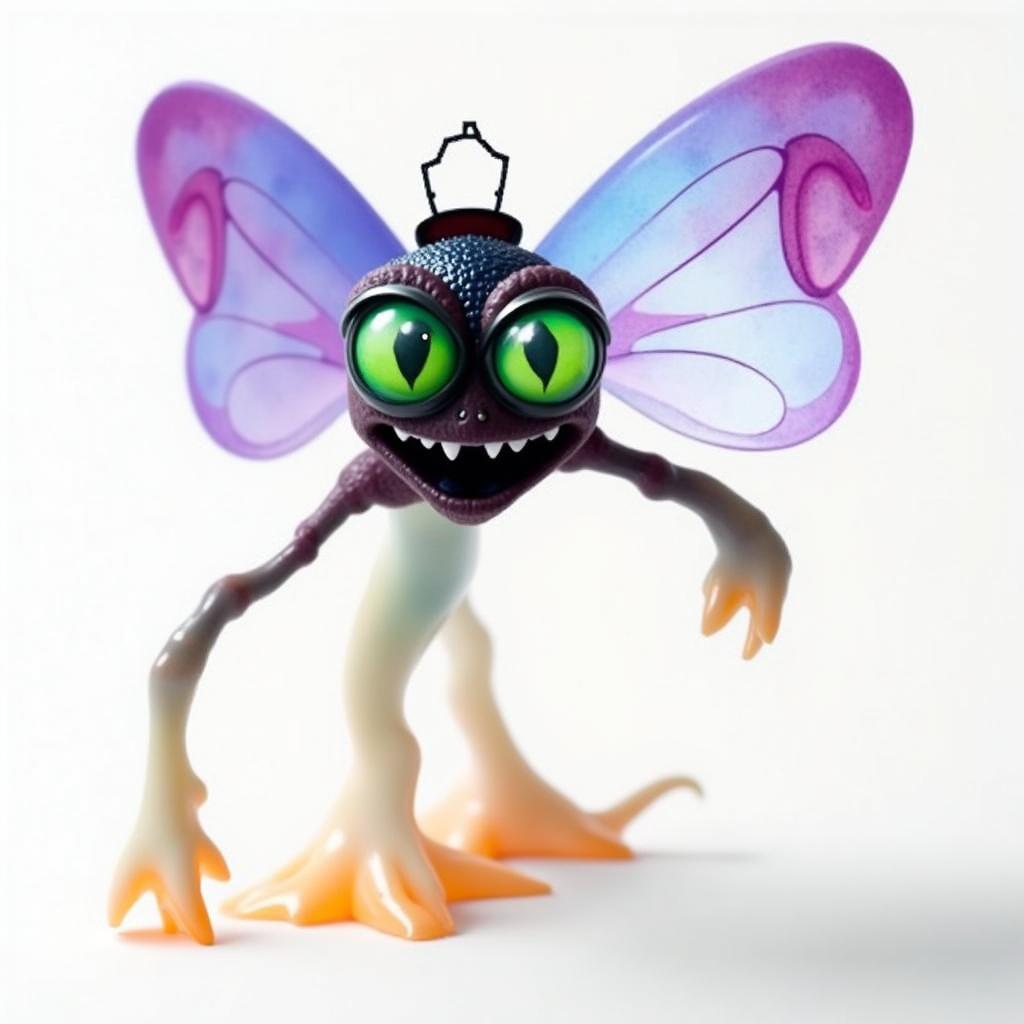} &

        \includegraphics[height=0.135\textwidth]{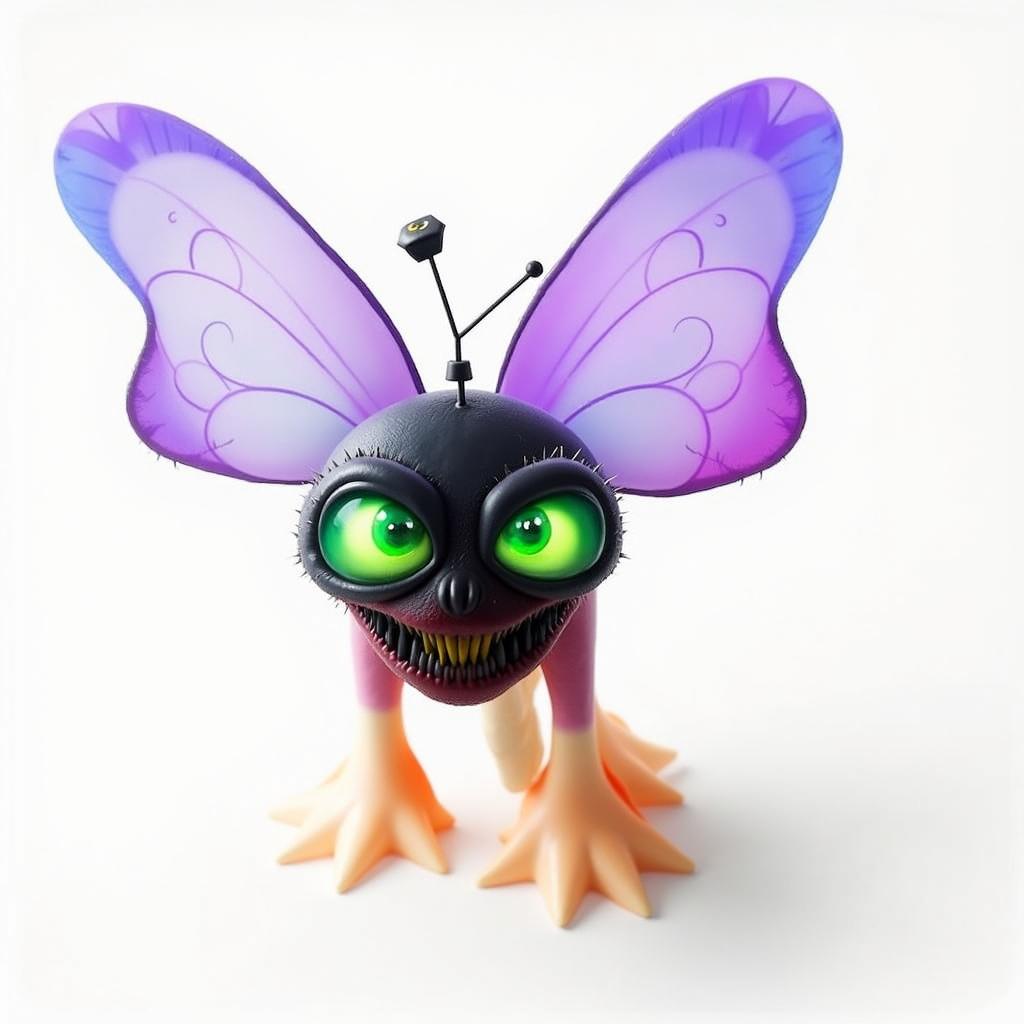} &

        \includegraphics[height=0.135\textwidth]{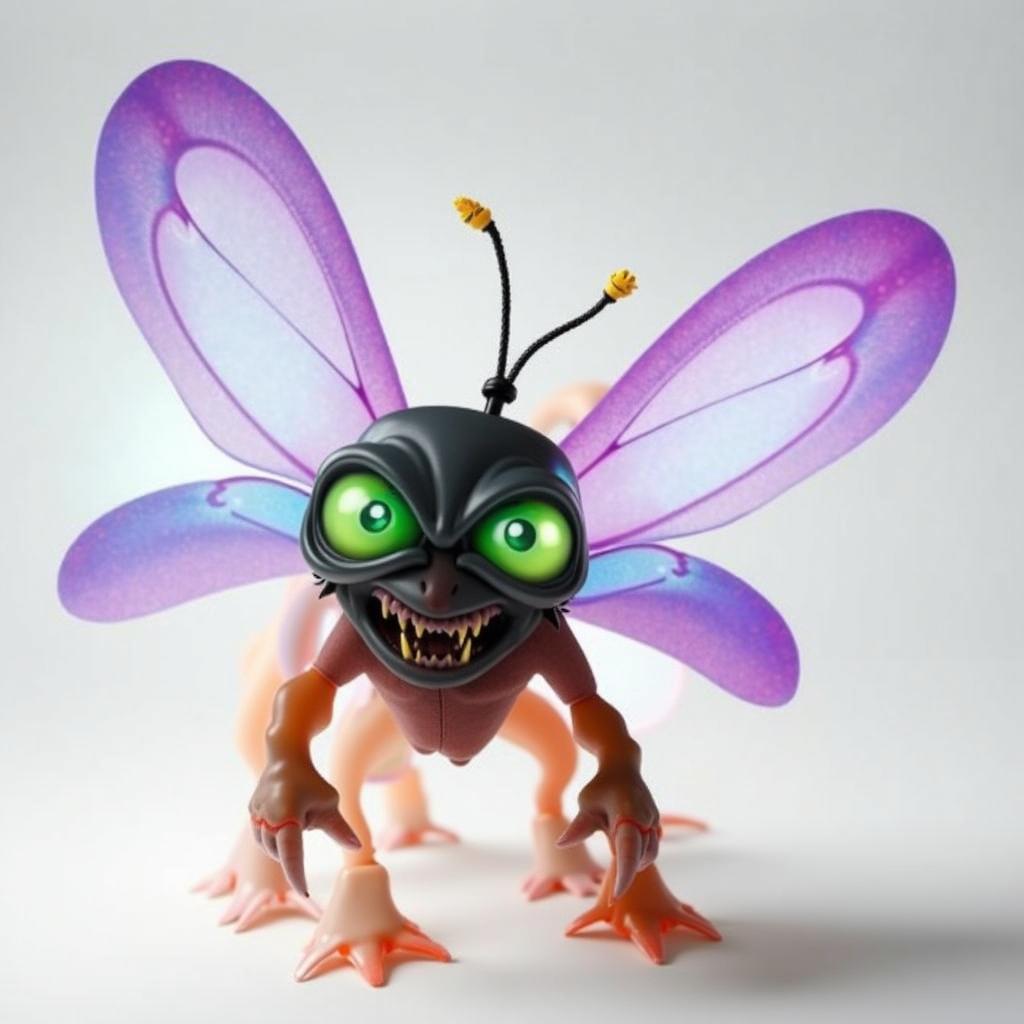} 
        
        \\
        
        \includegraphics[height=0.135\textwidth]{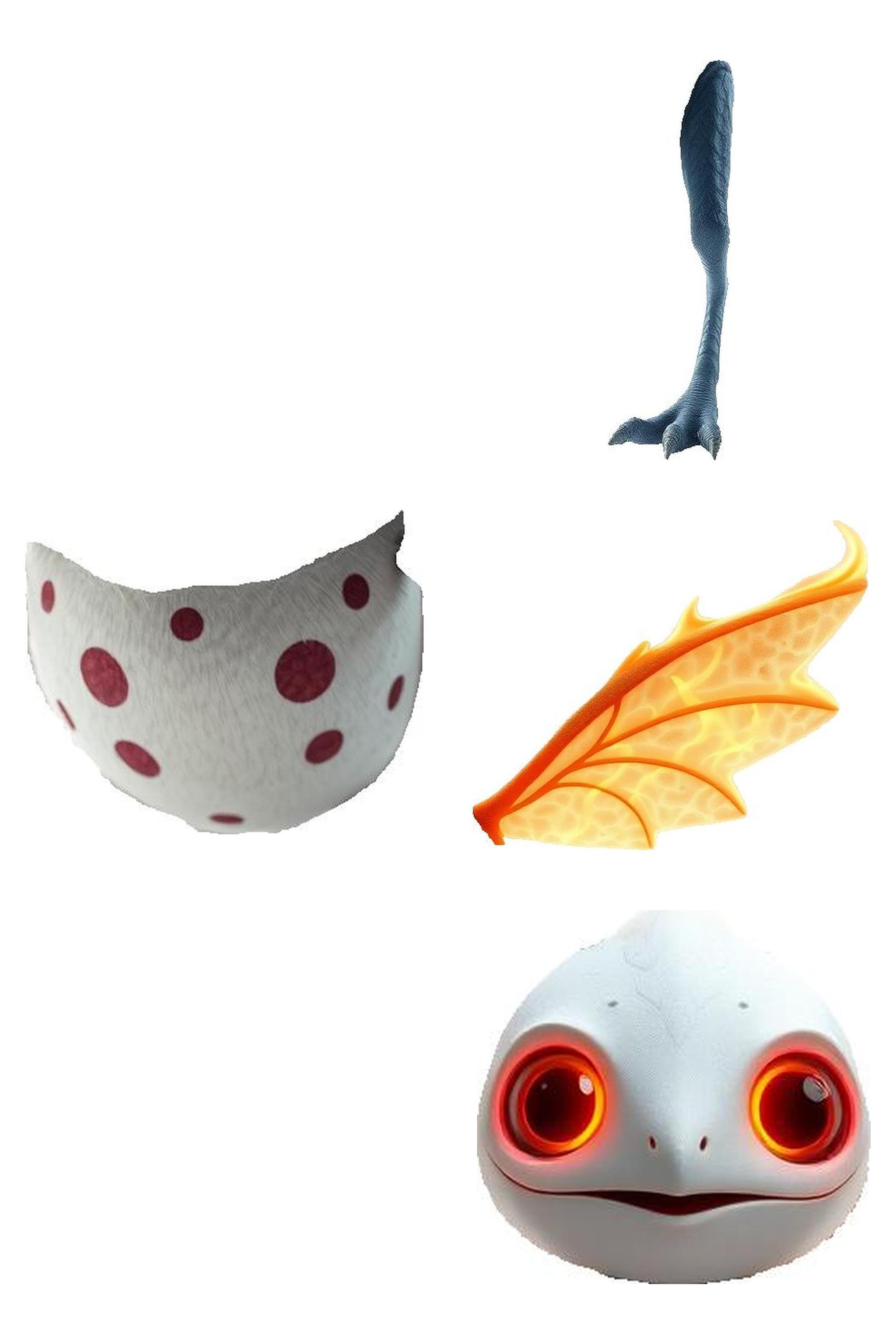} &

        \includegraphics[height=0.135\textwidth]{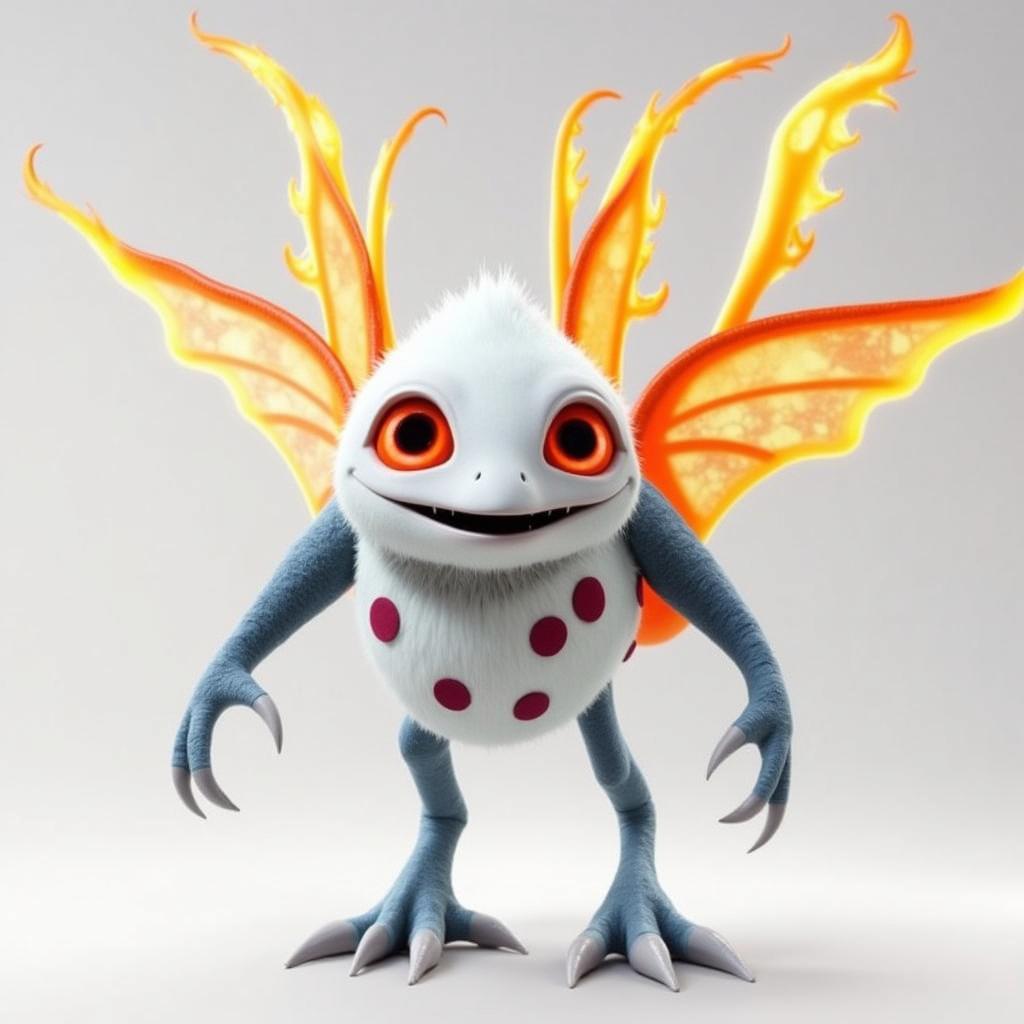} &

        \includegraphics[height=0.135\textwidth]{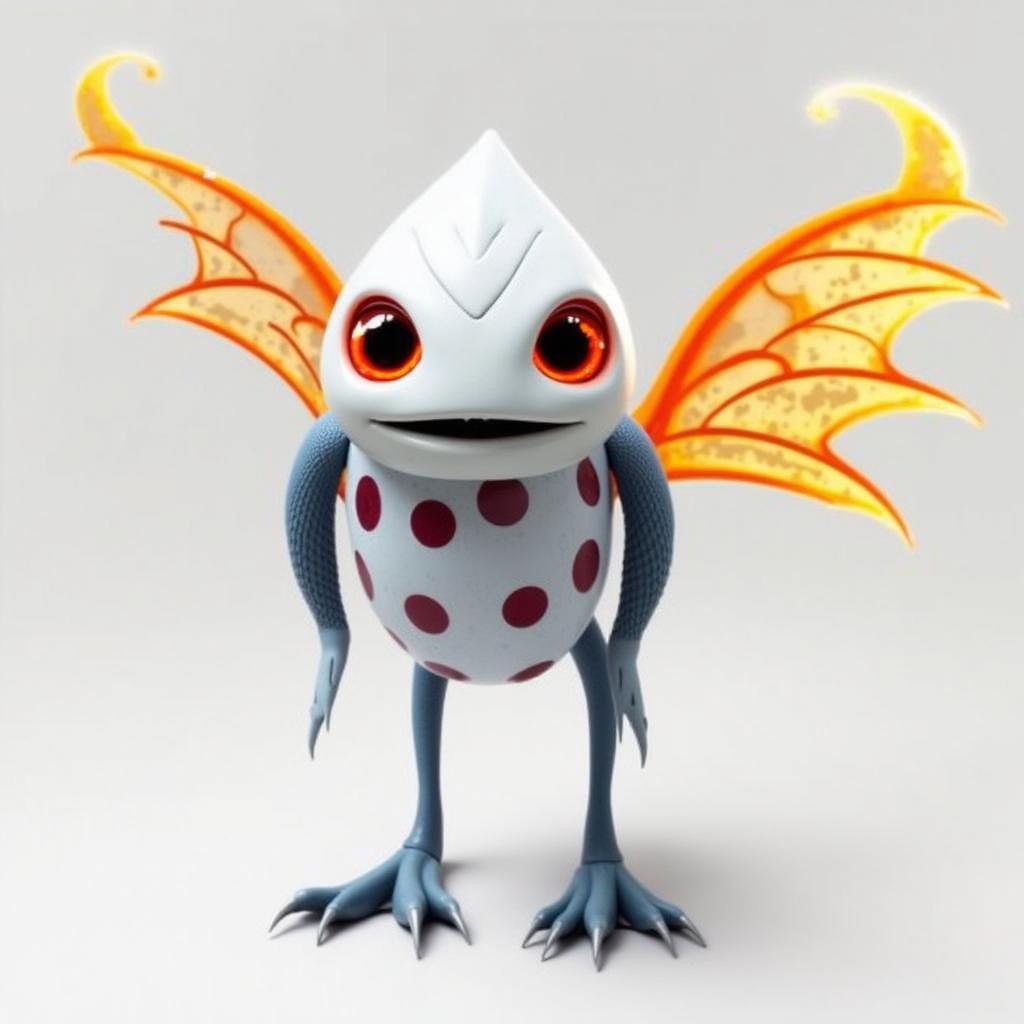} &

        \includegraphics[height=0.135\textwidth]{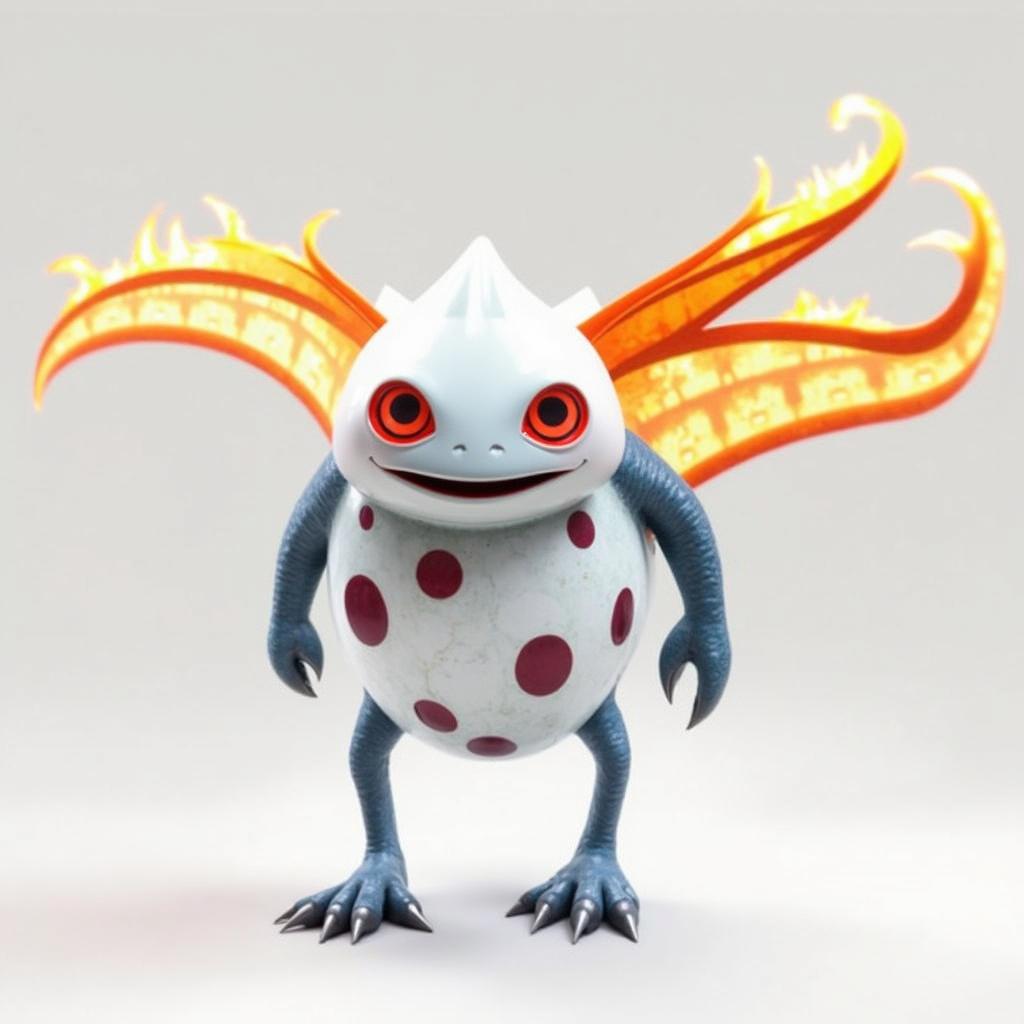} &

        \includegraphics[height=0.135\textwidth]{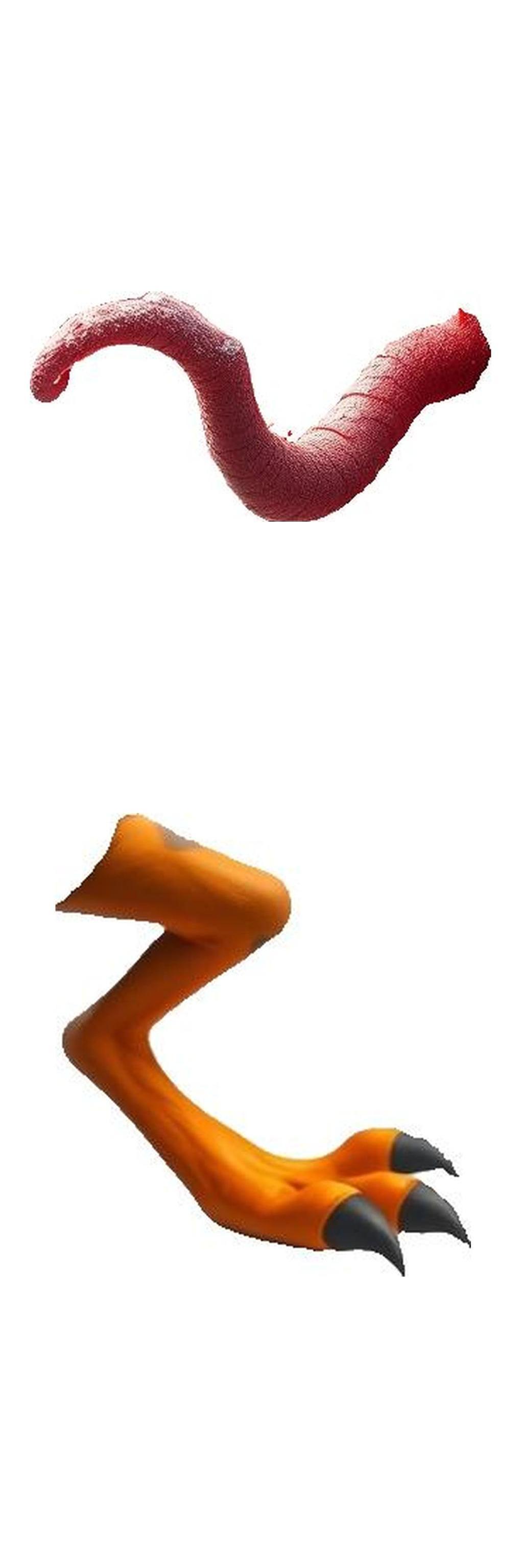} &

        \includegraphics[height=0.135\textwidth]{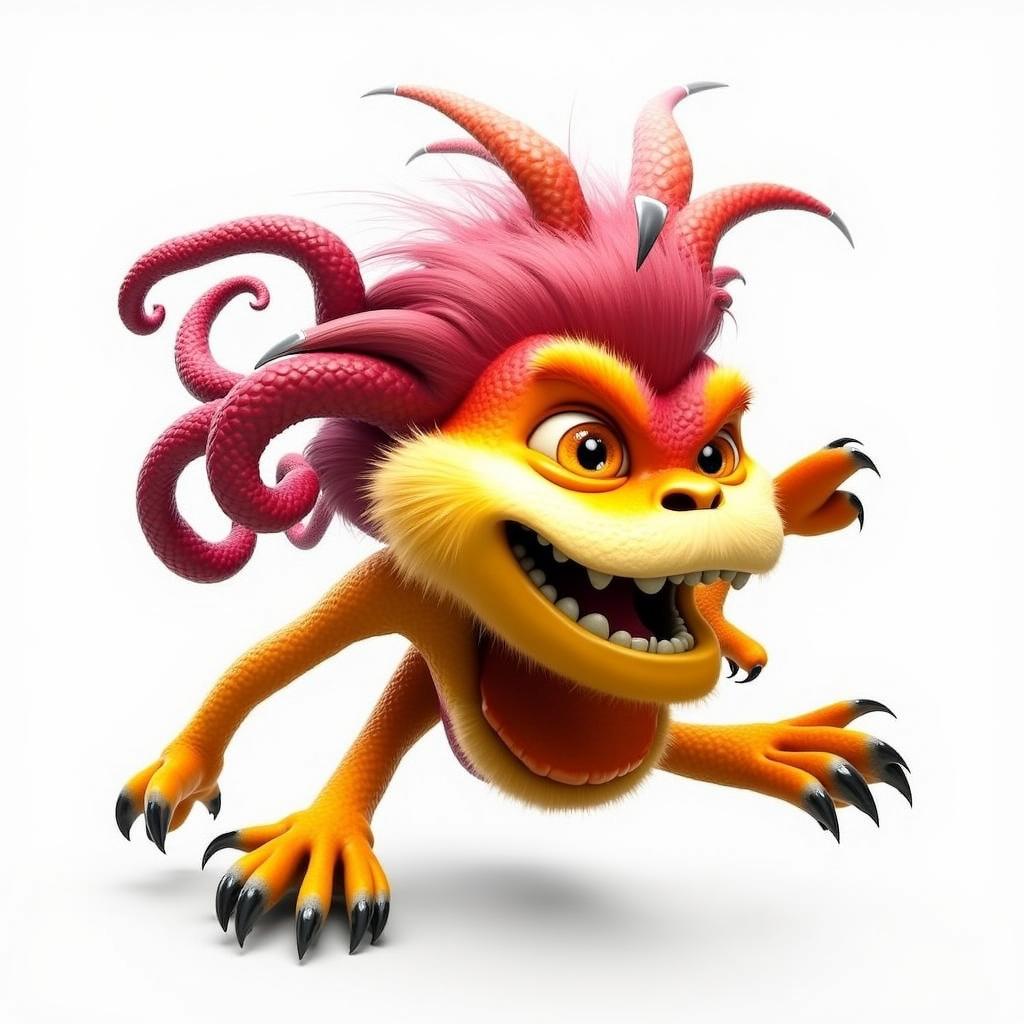} &

        \includegraphics[height=0.135\textwidth]{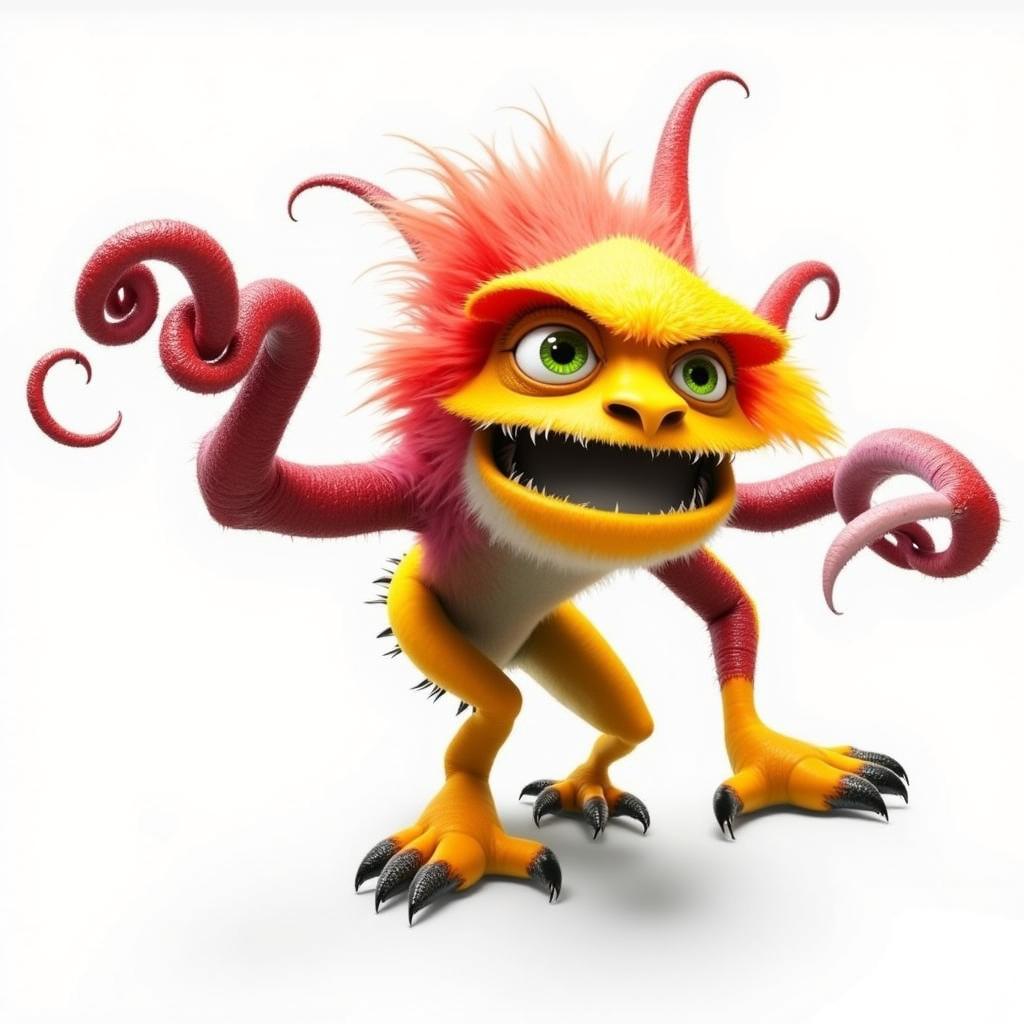} &

        \includegraphics[height=0.135\textwidth]{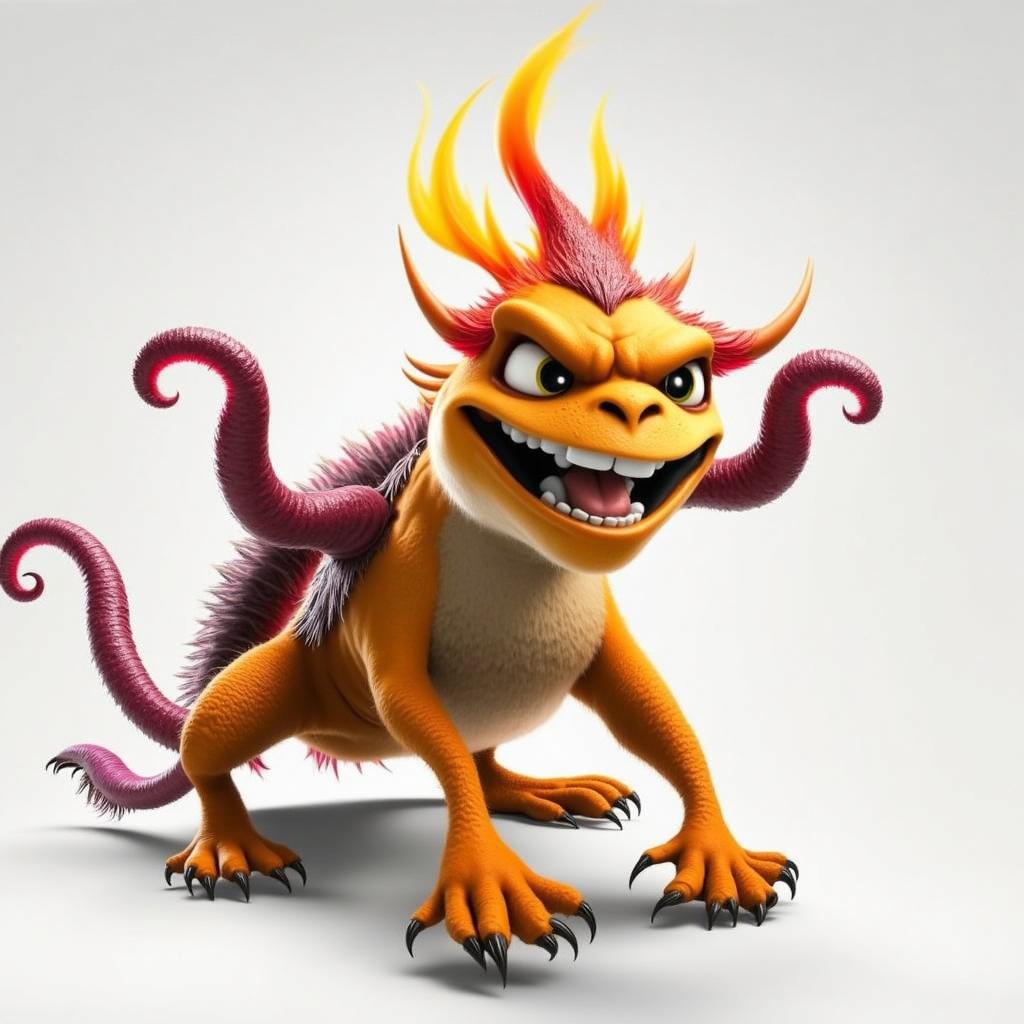} 
        
        \\
        
        \includegraphics[height=0.135\textwidth]{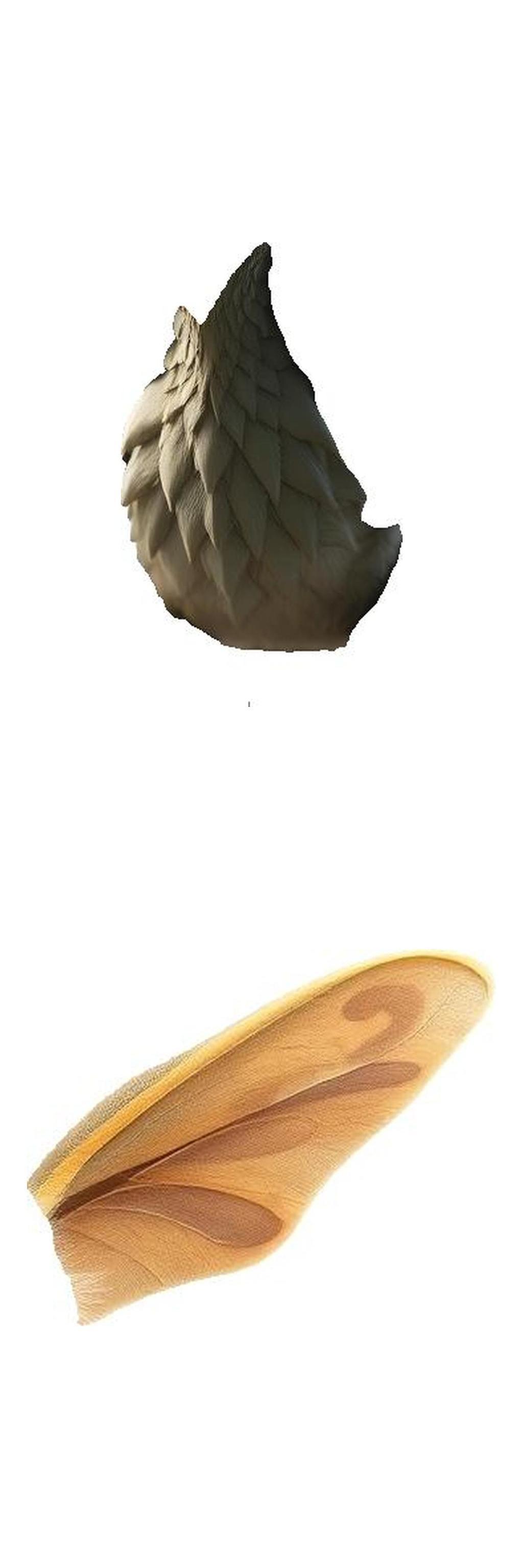} &

        \includegraphics[height=0.135\textwidth]{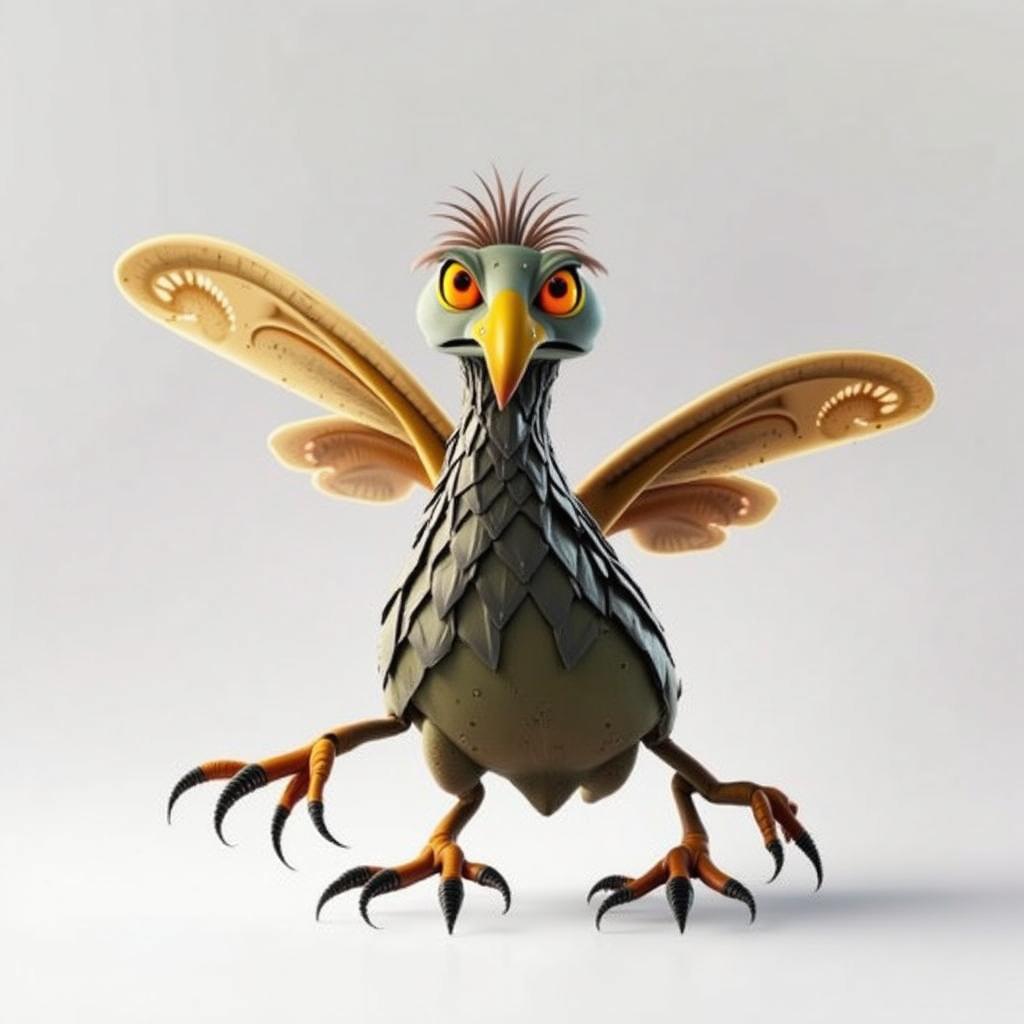} &

        \includegraphics[height=0.135\textwidth]{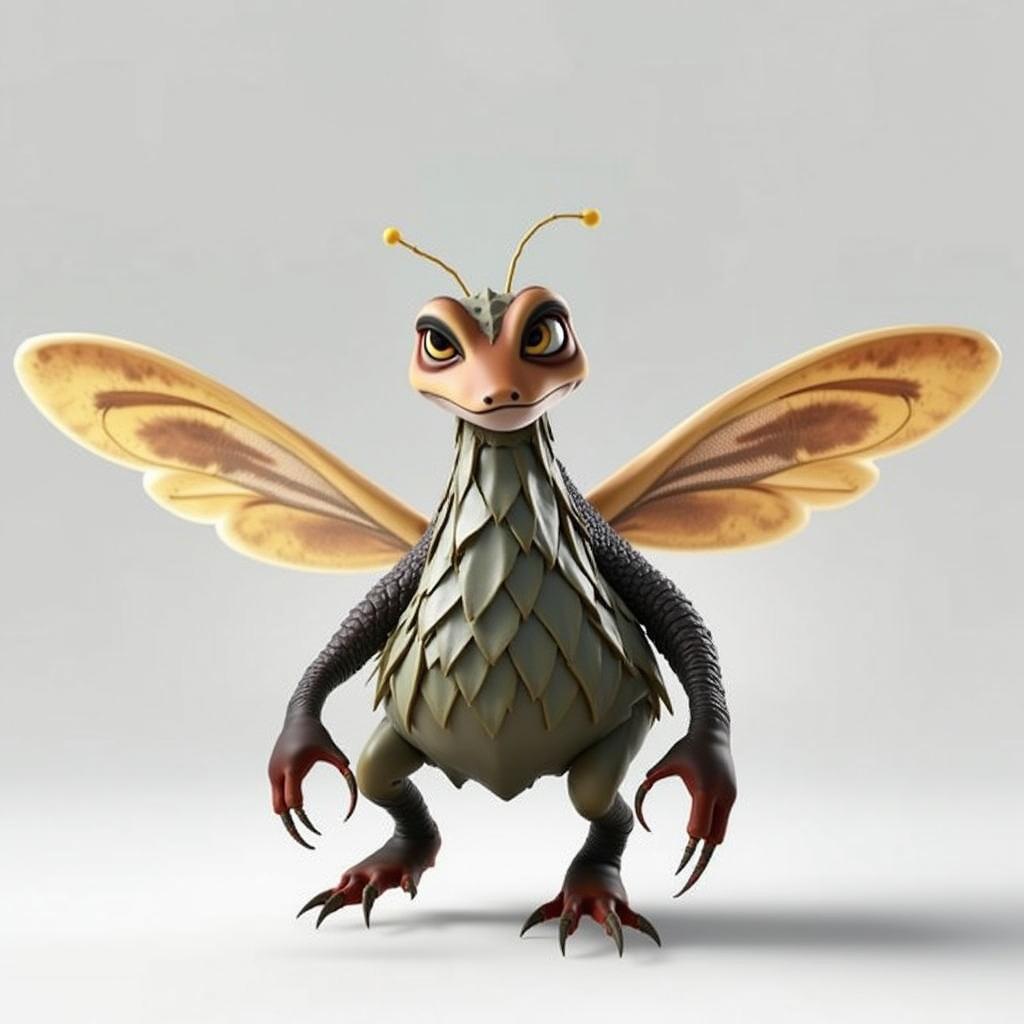} &

        \includegraphics[height=0.135\textwidth]{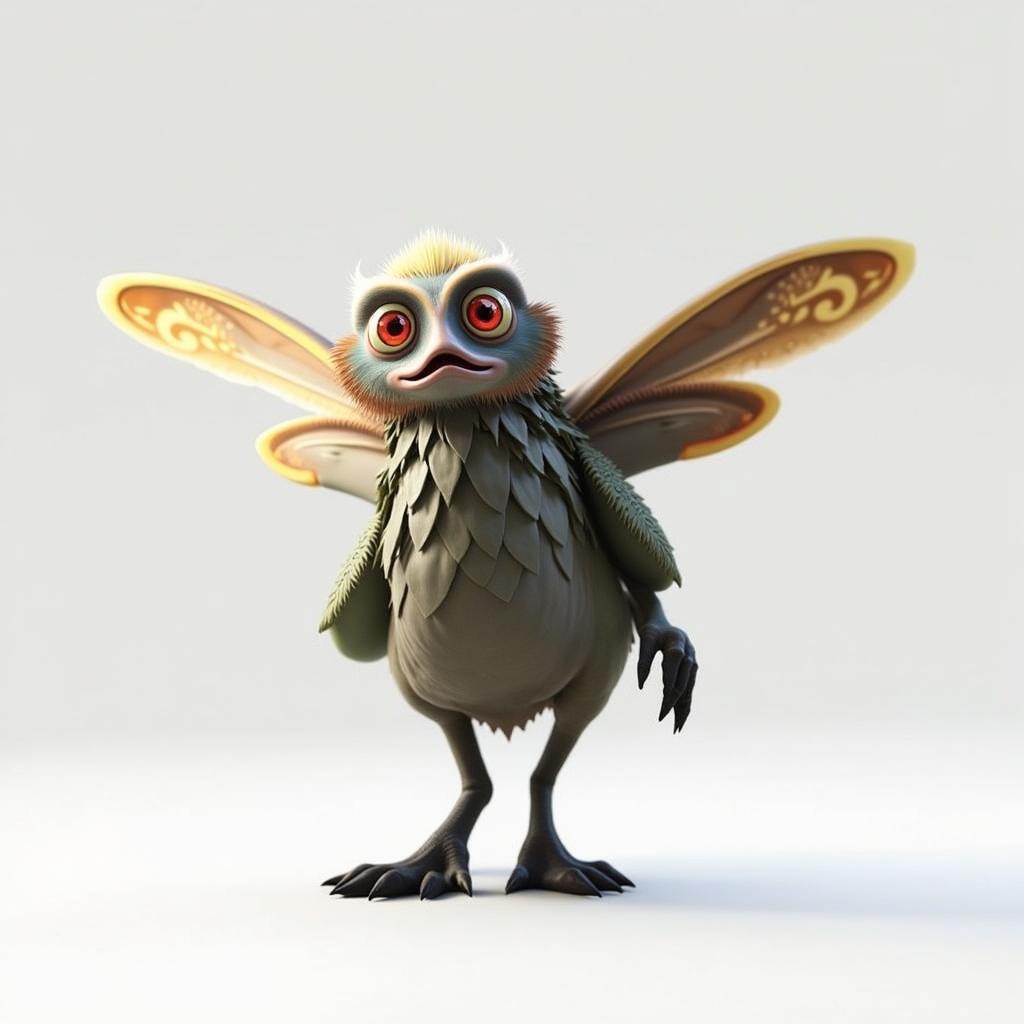} &

        \includegraphics[height=0.135\textwidth]{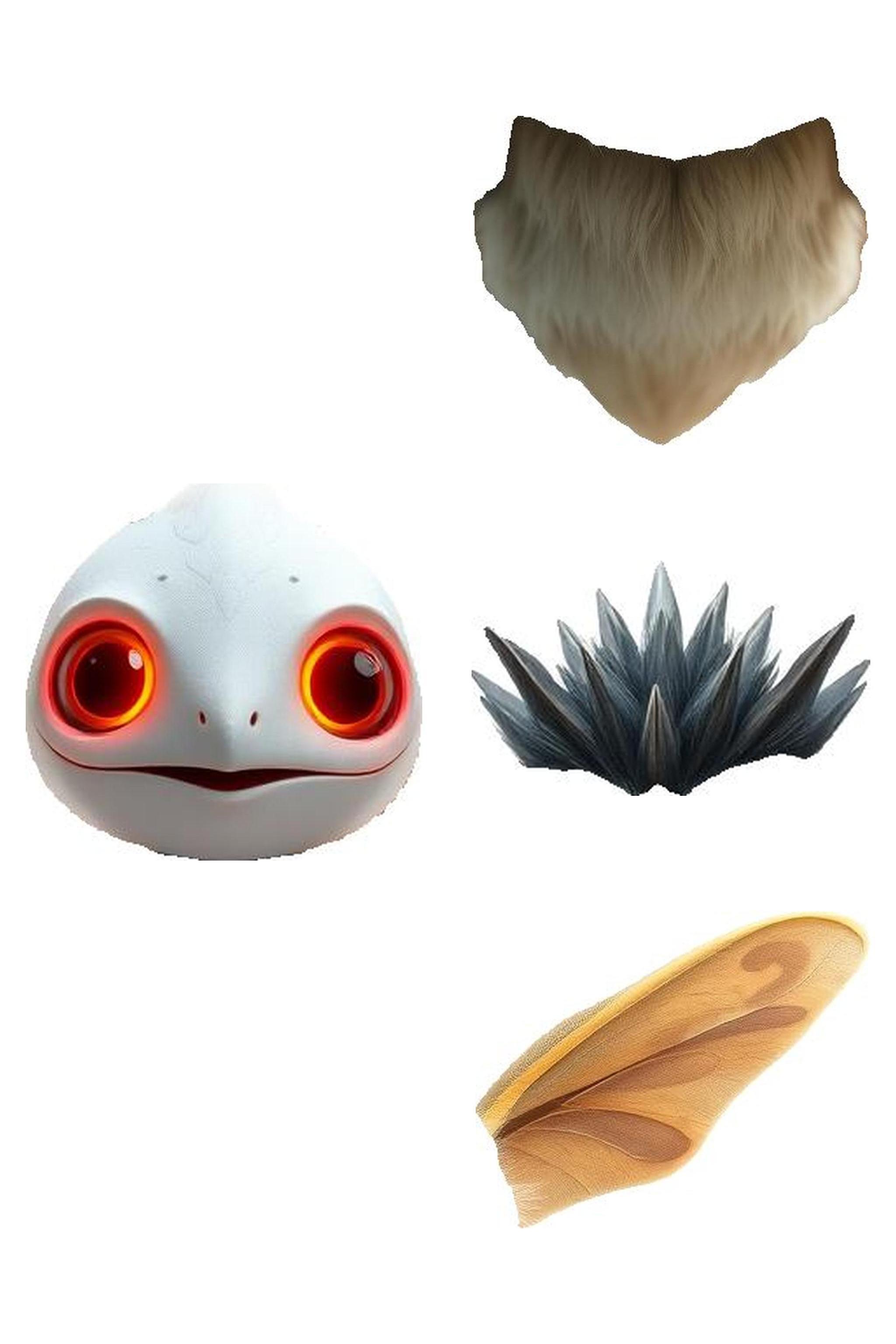} &

        \includegraphics[height=0.135\textwidth]{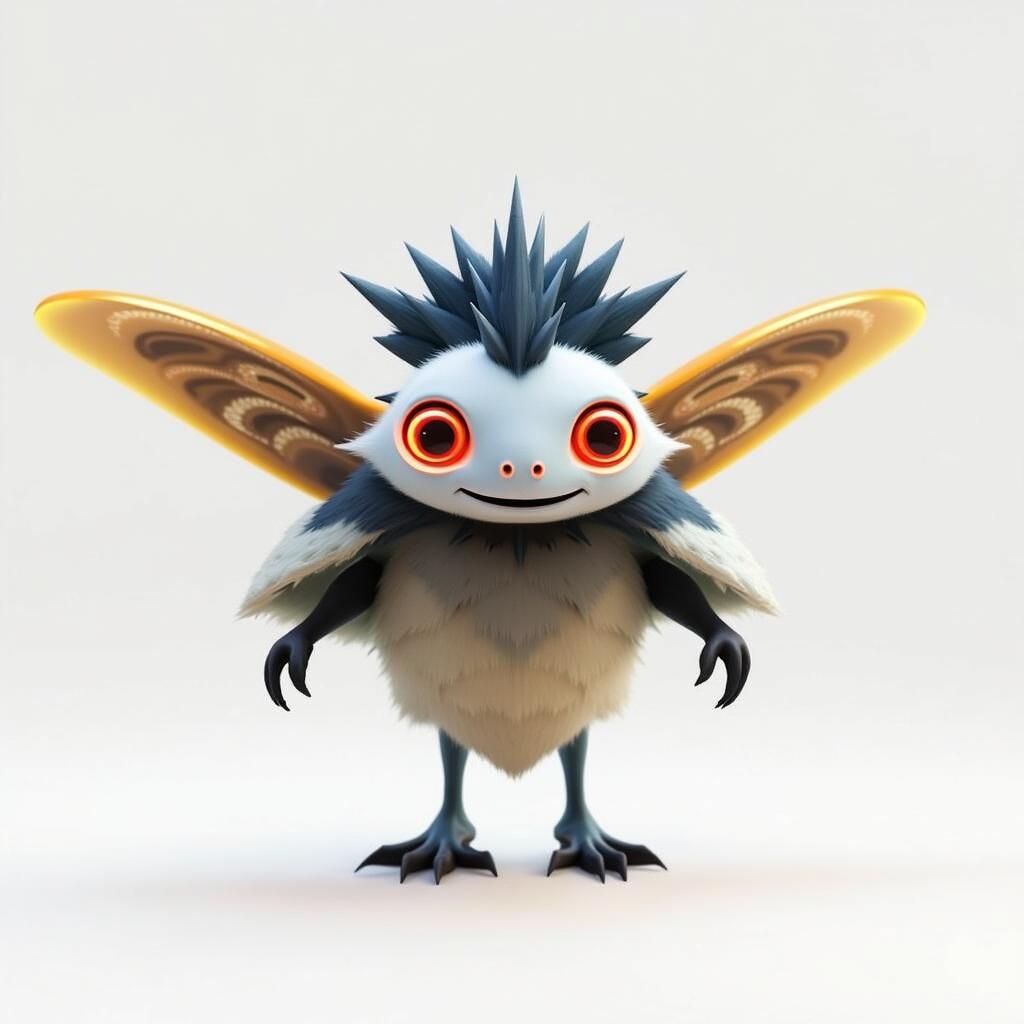} &

        \includegraphics[height=0.135\textwidth]{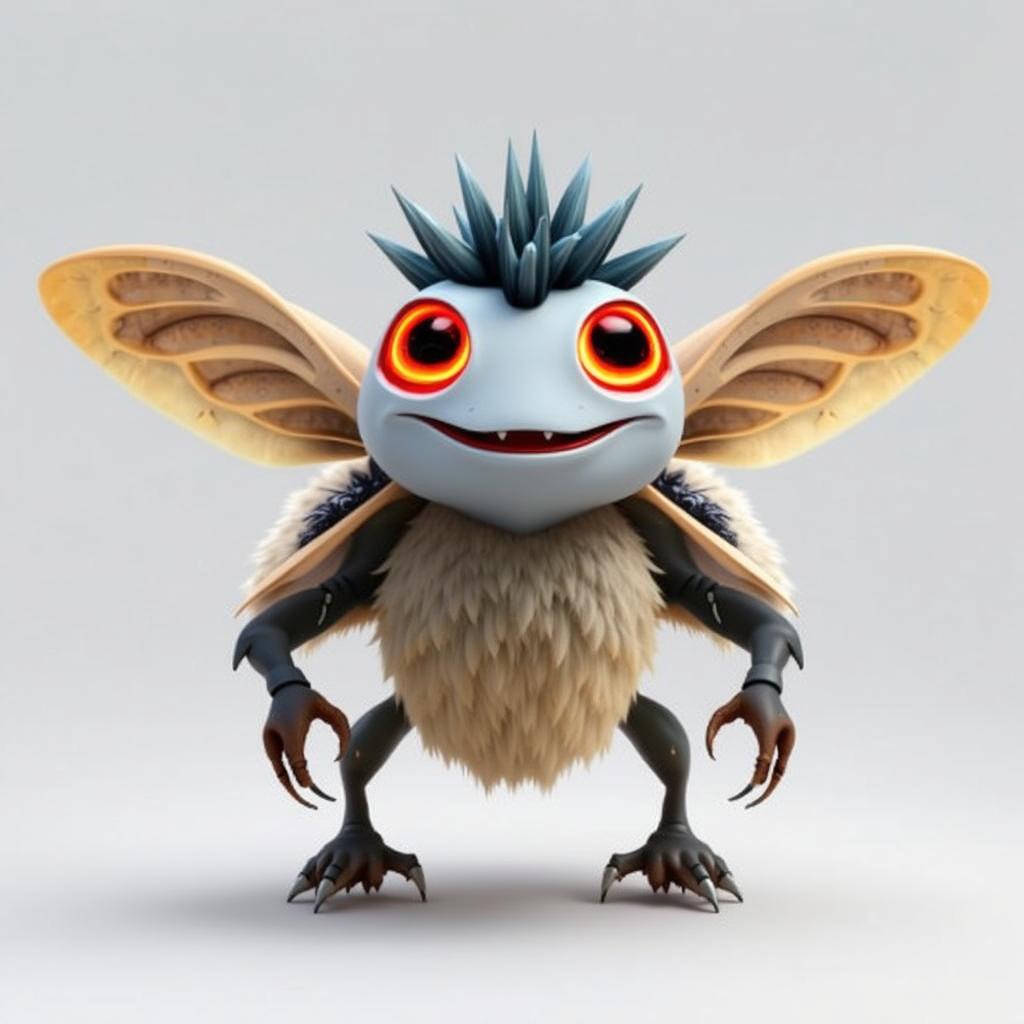} &

        \includegraphics[height=0.135\textwidth]{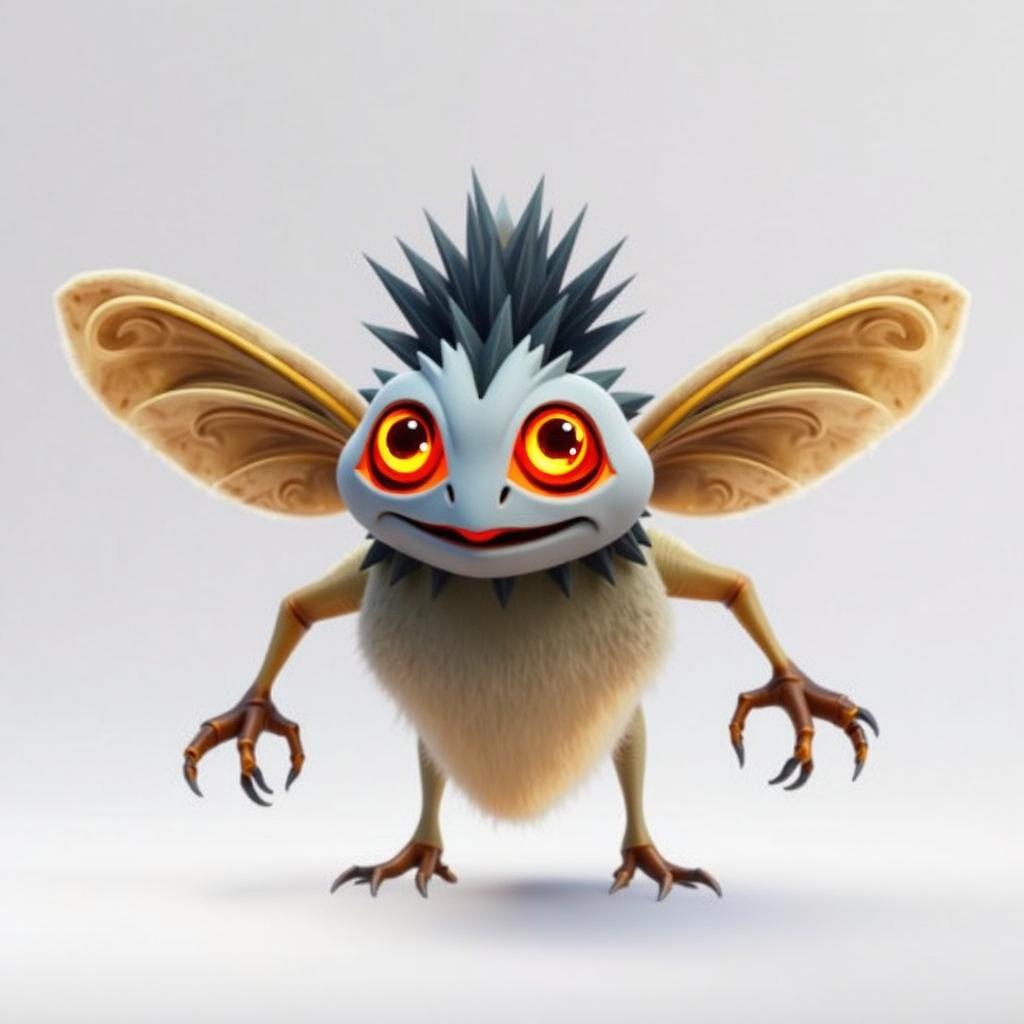} 
        
        \\

        \includegraphics[height=0.135\textwidth]{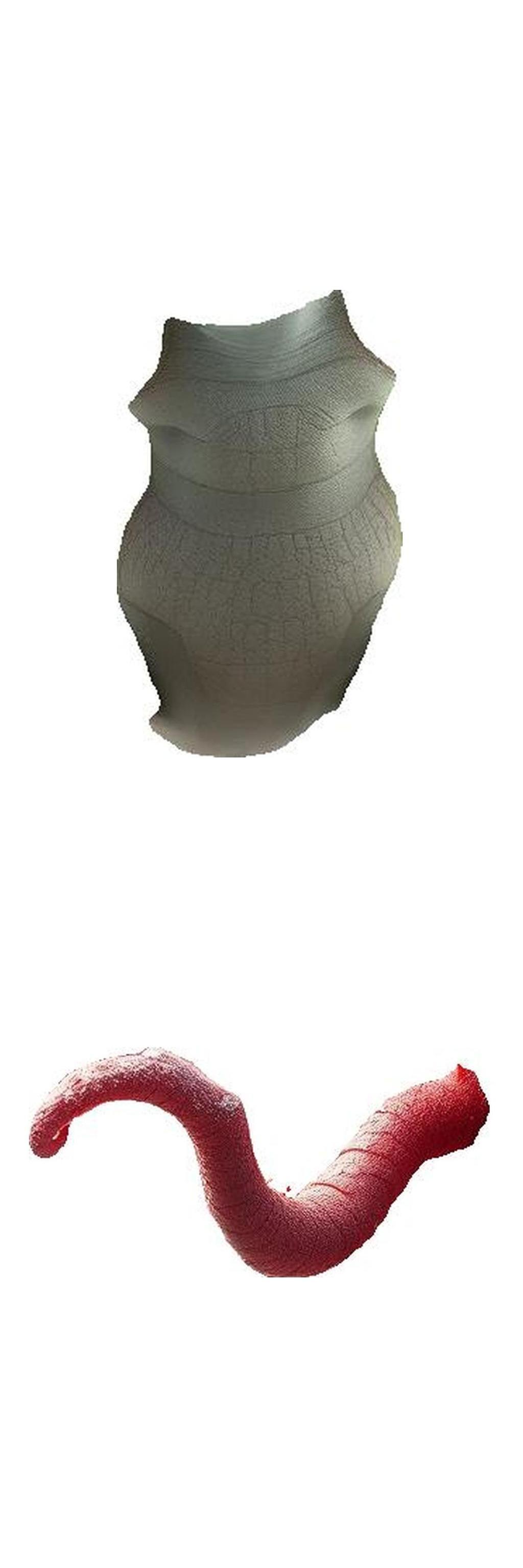} &

        \includegraphics[height=0.135\textwidth]{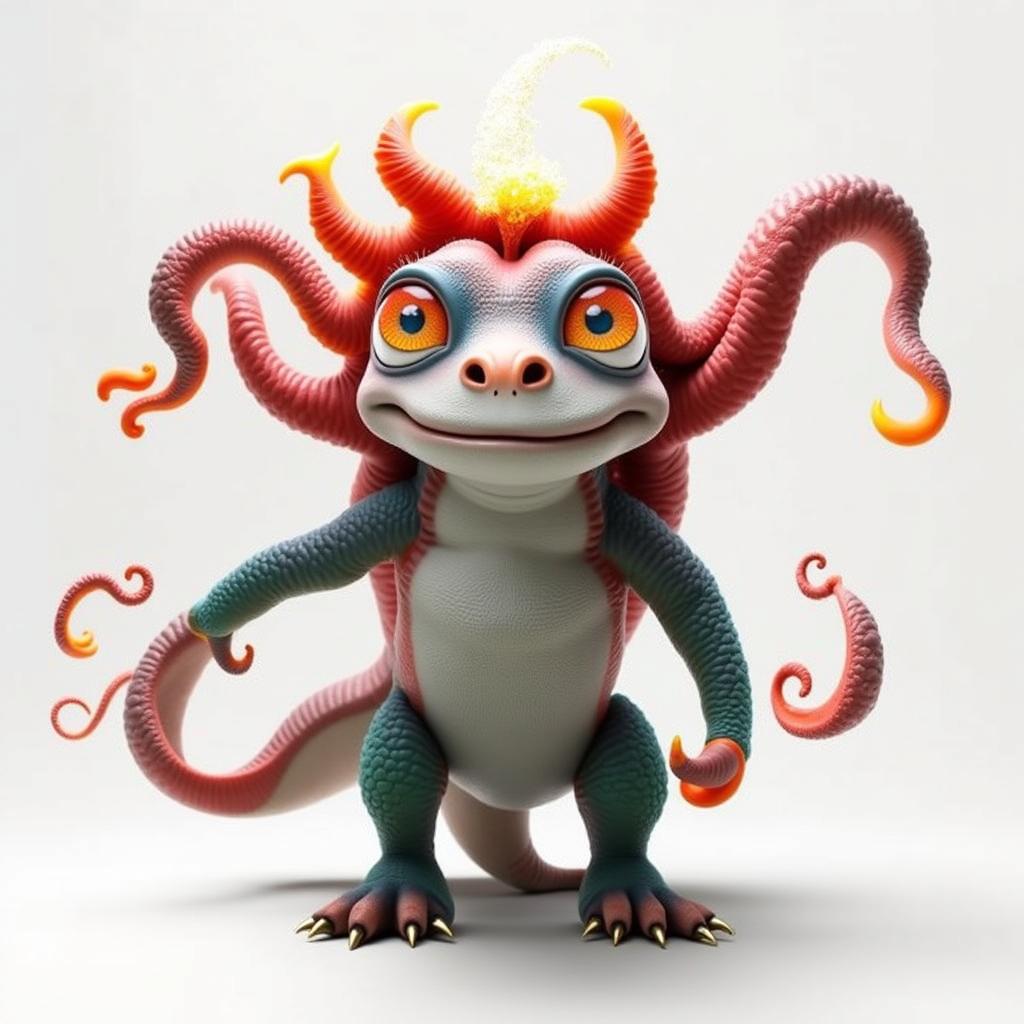} &

        \includegraphics[height=0.135\textwidth]{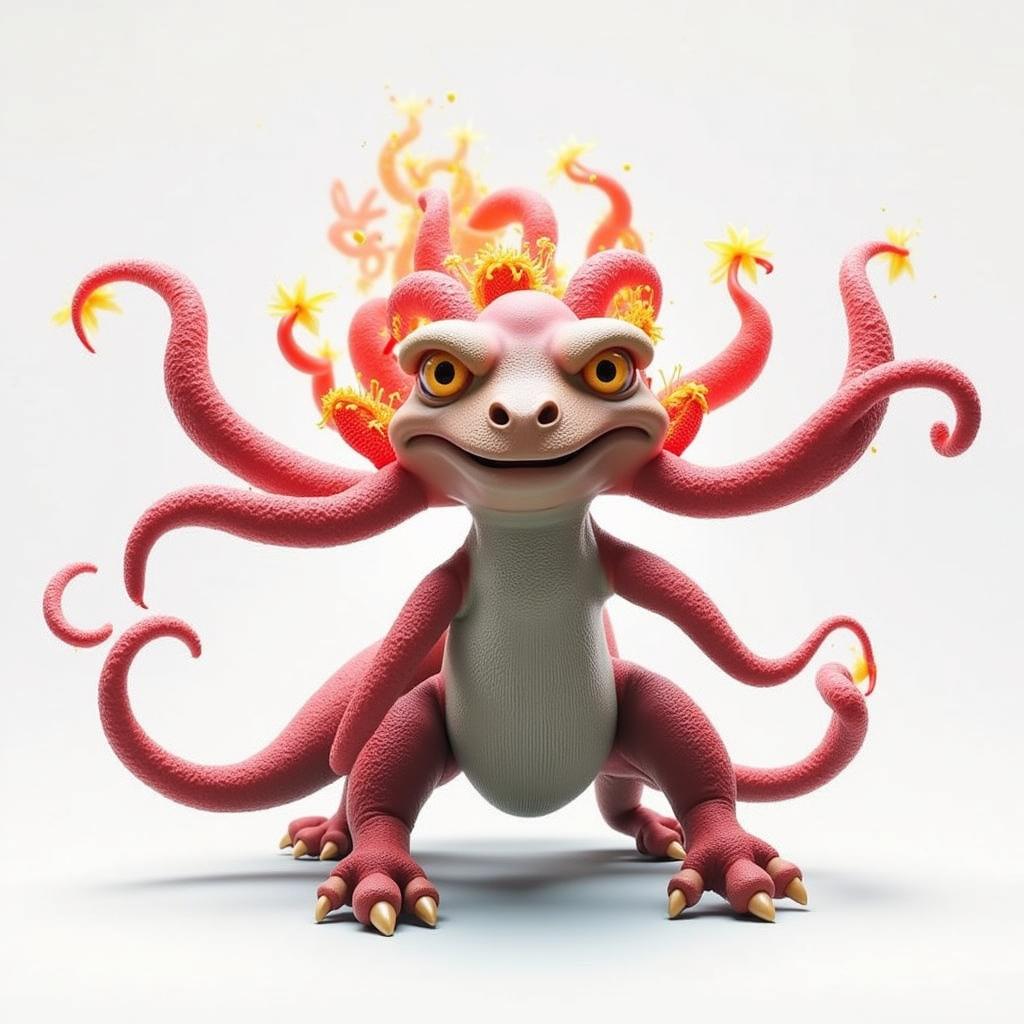} &

        \includegraphics[height=0.135\textwidth]{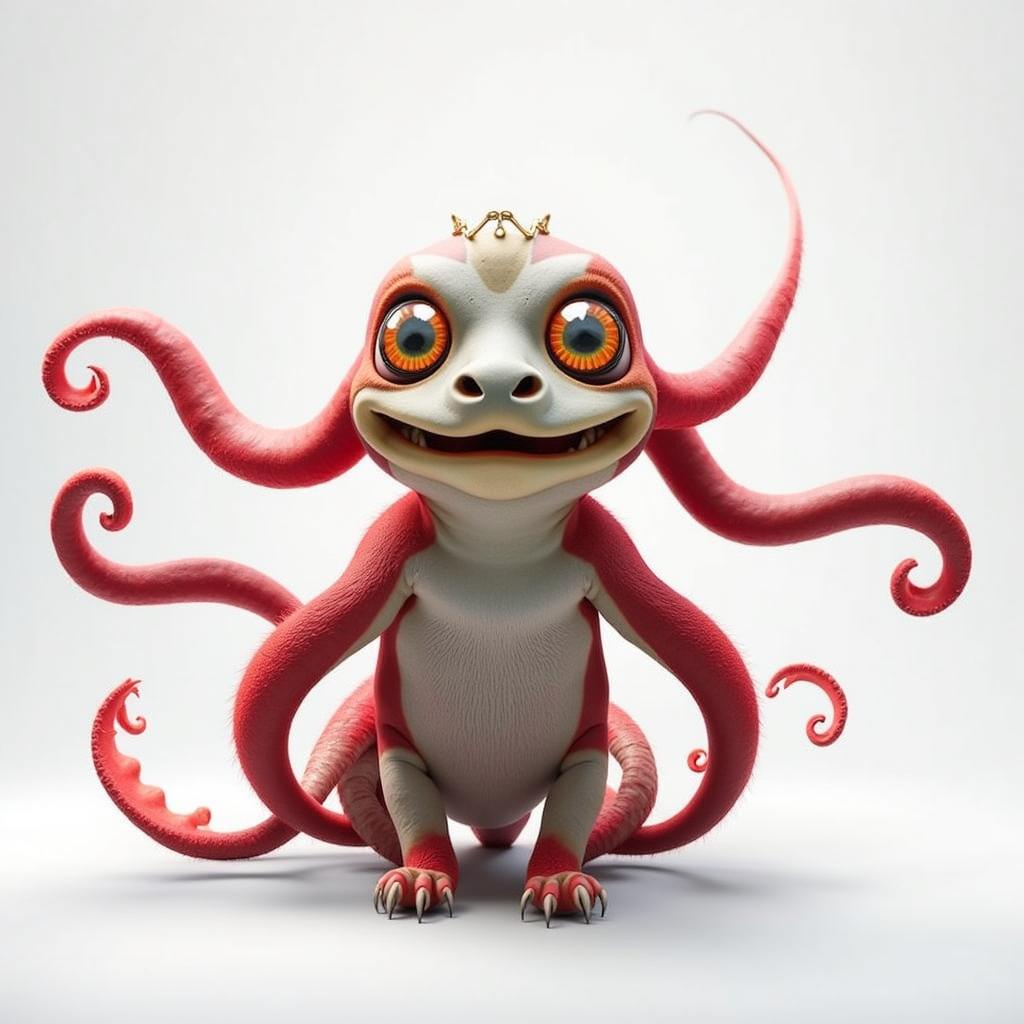} &

        \includegraphics[height=0.135\textwidth]{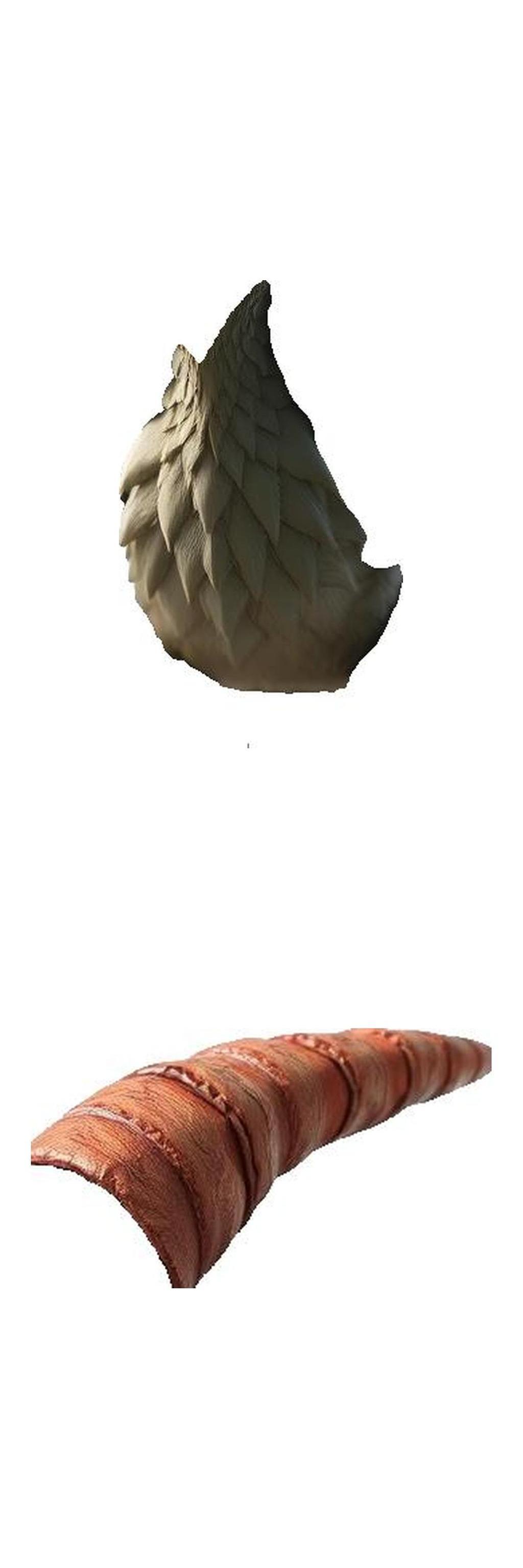} &

        \includegraphics[height=0.135\textwidth]{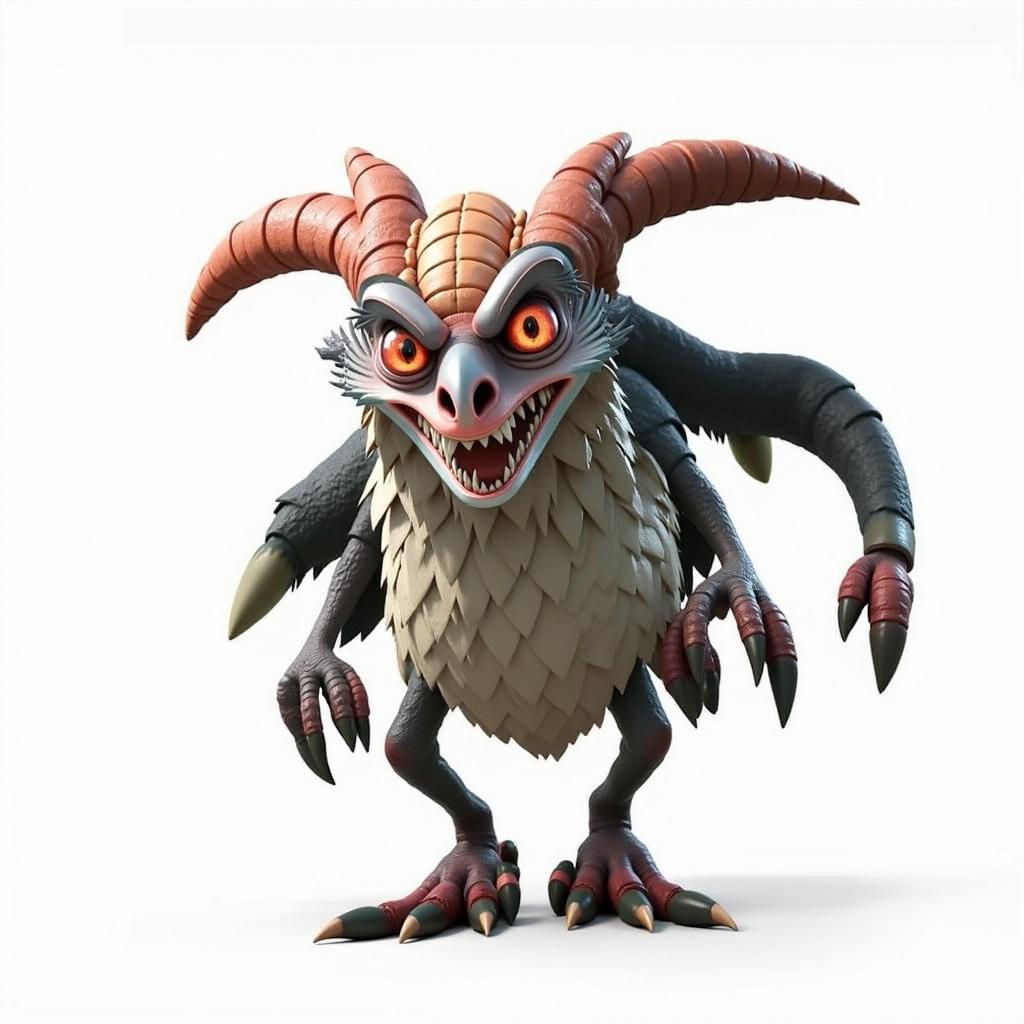} &

        \includegraphics[height=0.135\textwidth]{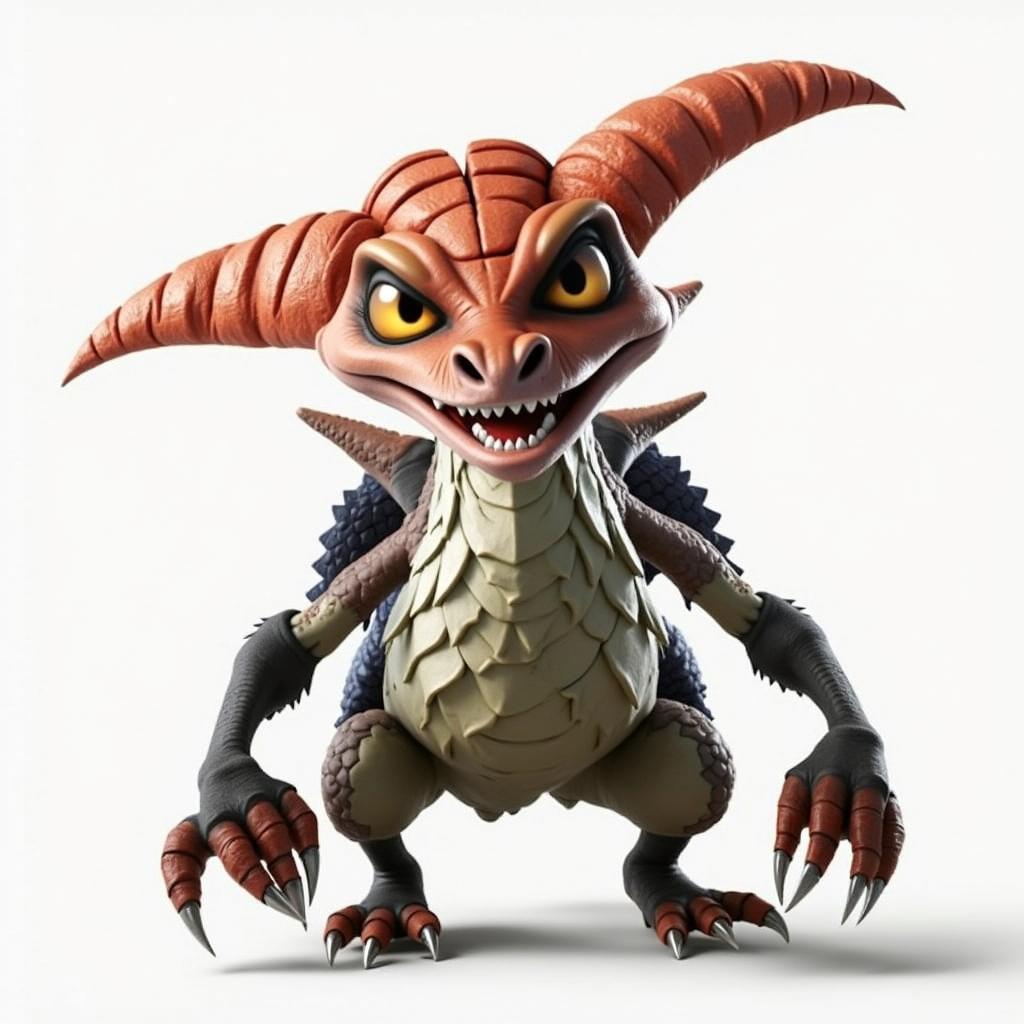} &

        \includegraphics[height=0.135\textwidth]{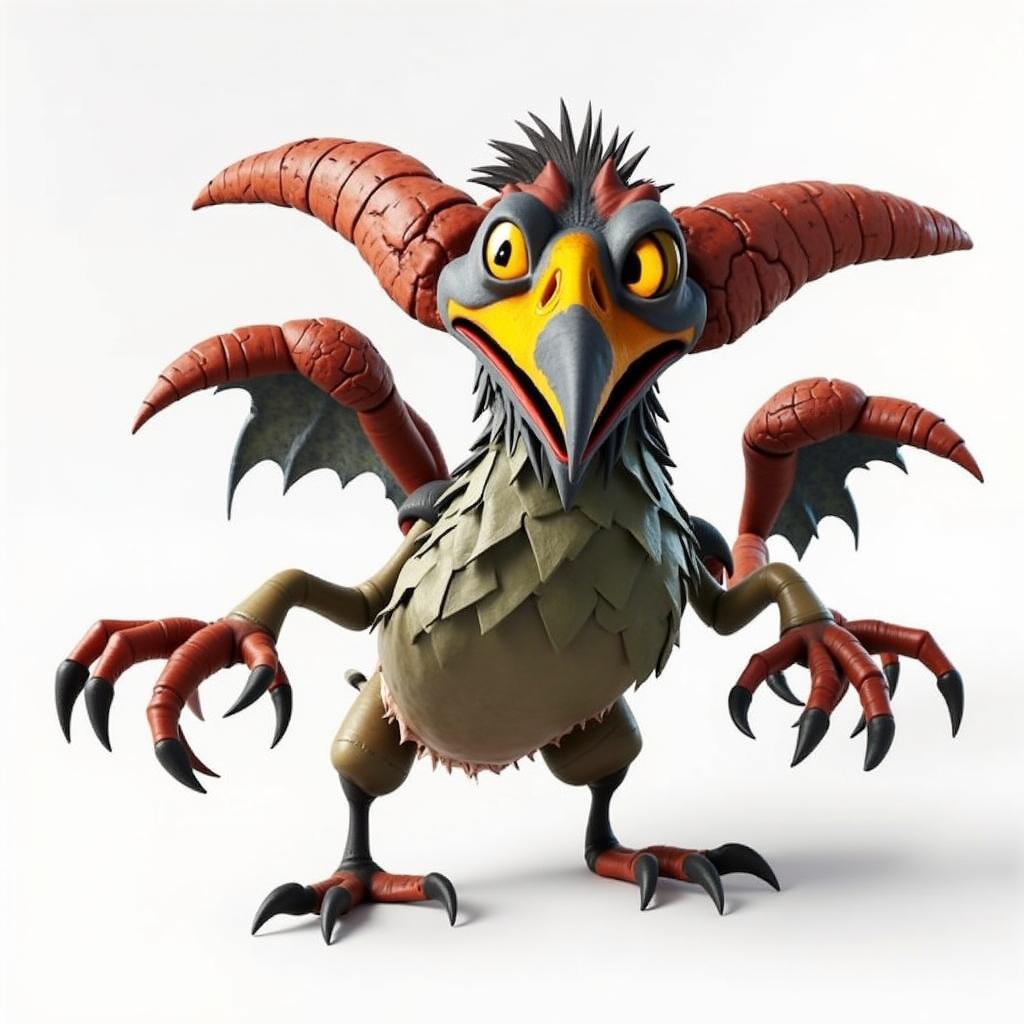} 
        
        \\
               
        \includegraphics[height=0.135\textwidth]{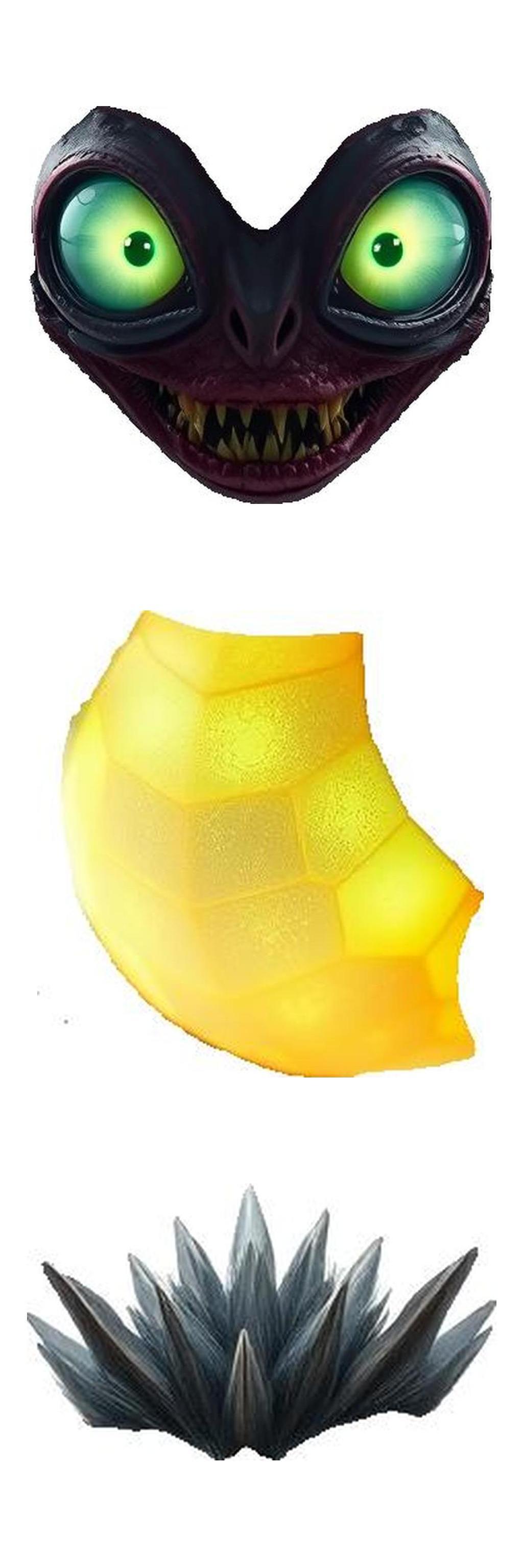} &

        \includegraphics[height=0.135\textwidth]{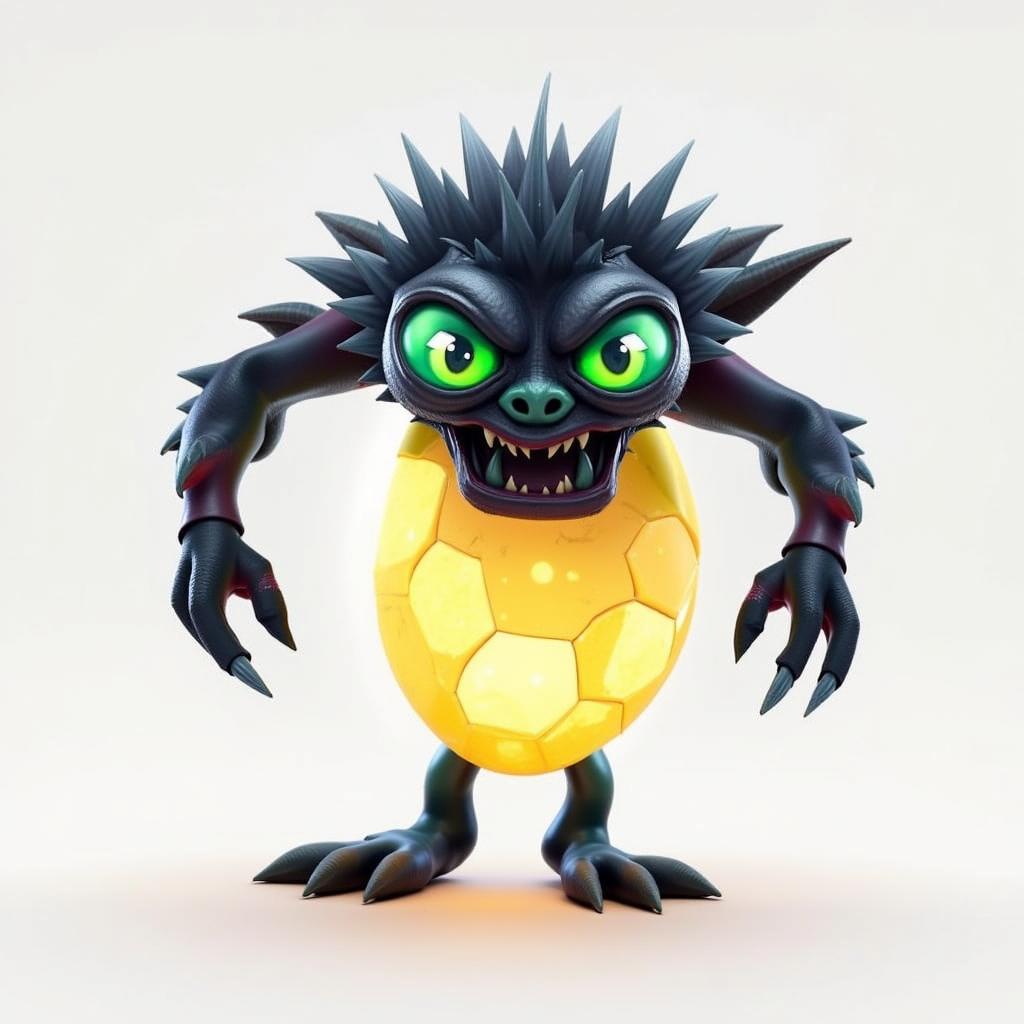} &

        \includegraphics[height=0.135\textwidth]{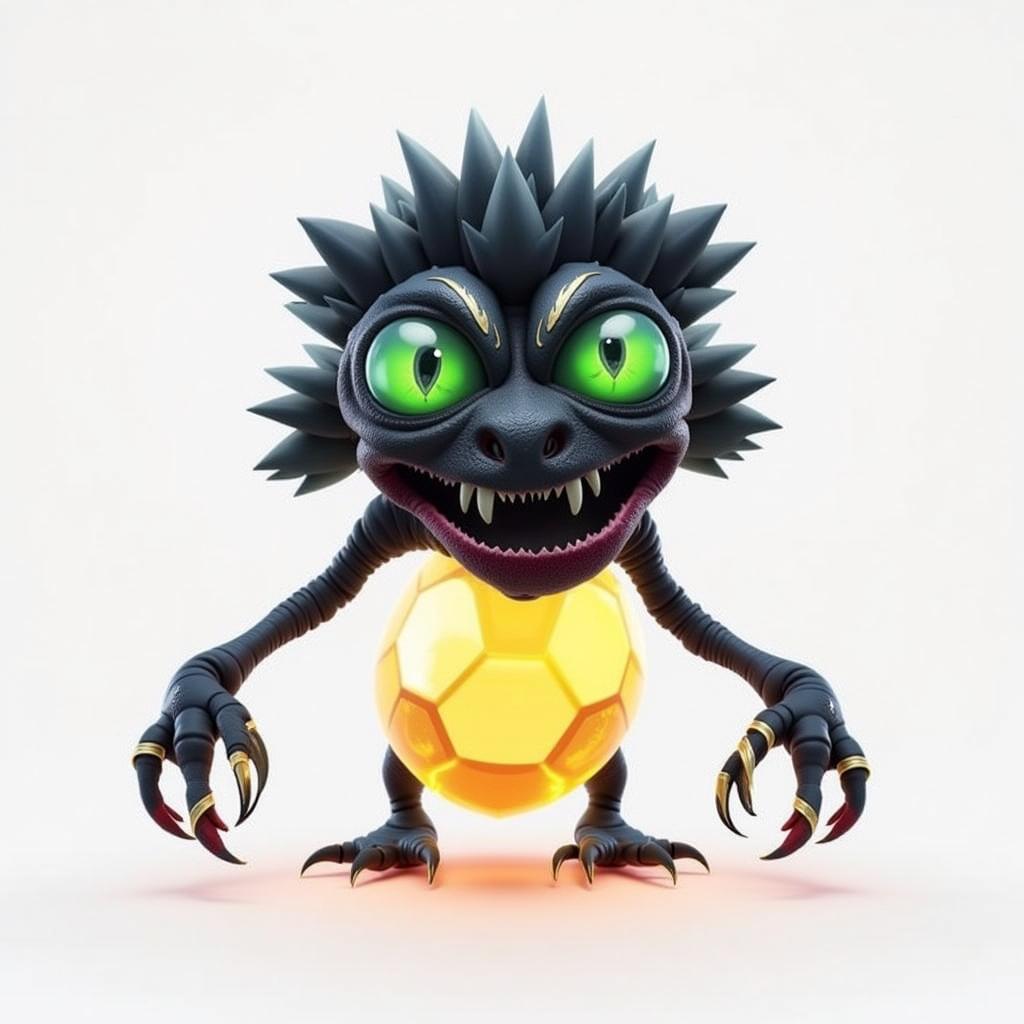} &

        \includegraphics[height=0.135\textwidth]{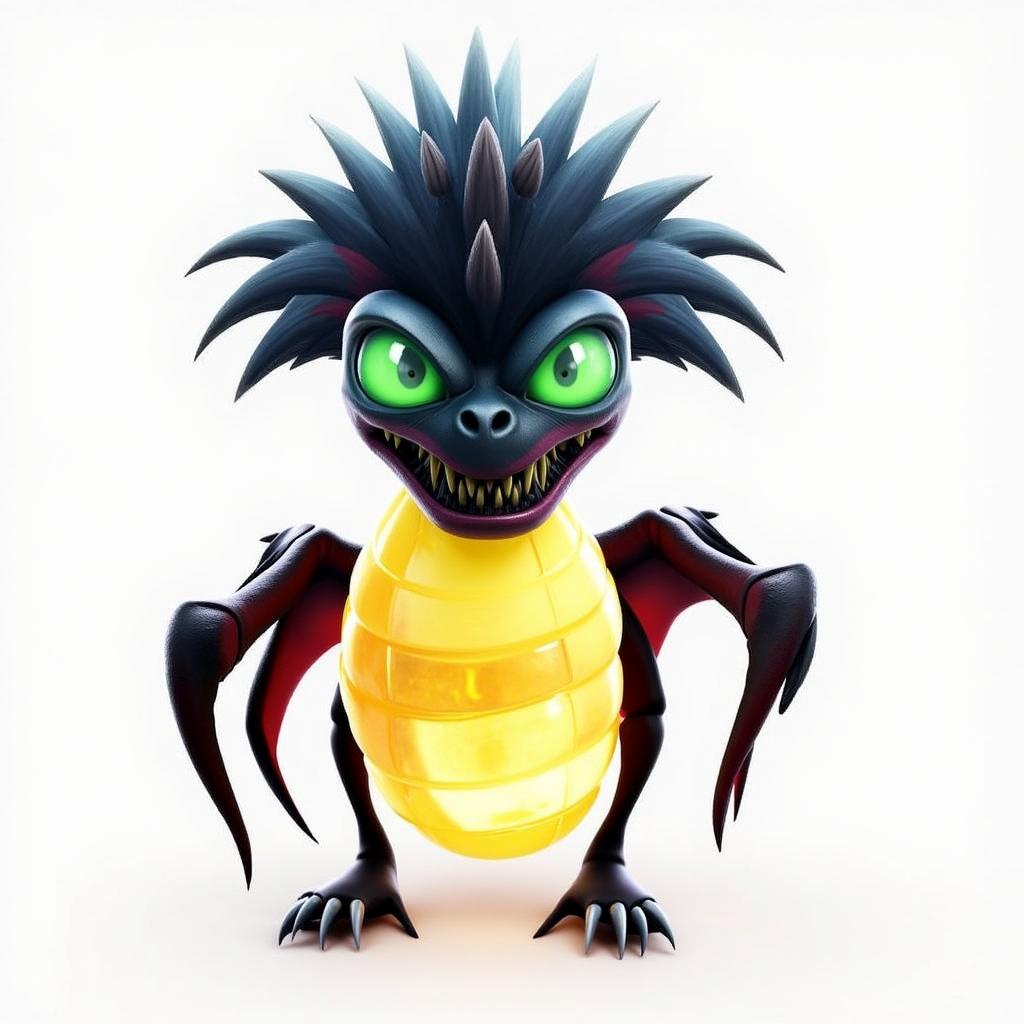} &

        \includegraphics[height=0.135\textwidth]{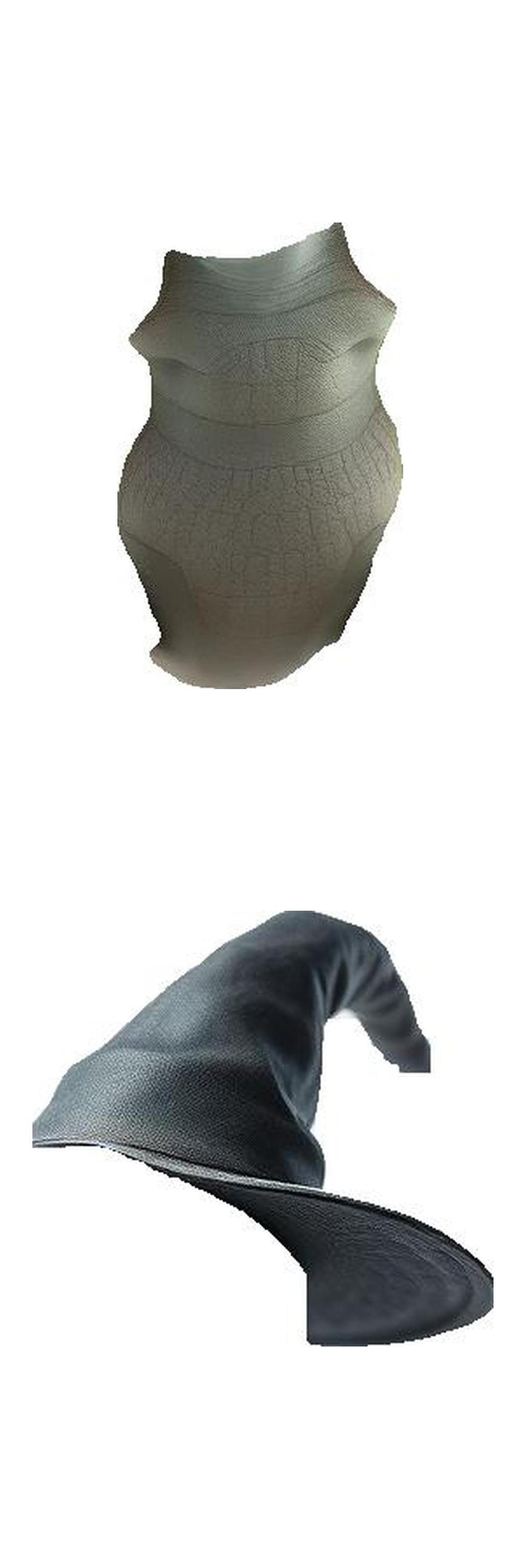} &

        \includegraphics[height=0.135\textwidth]{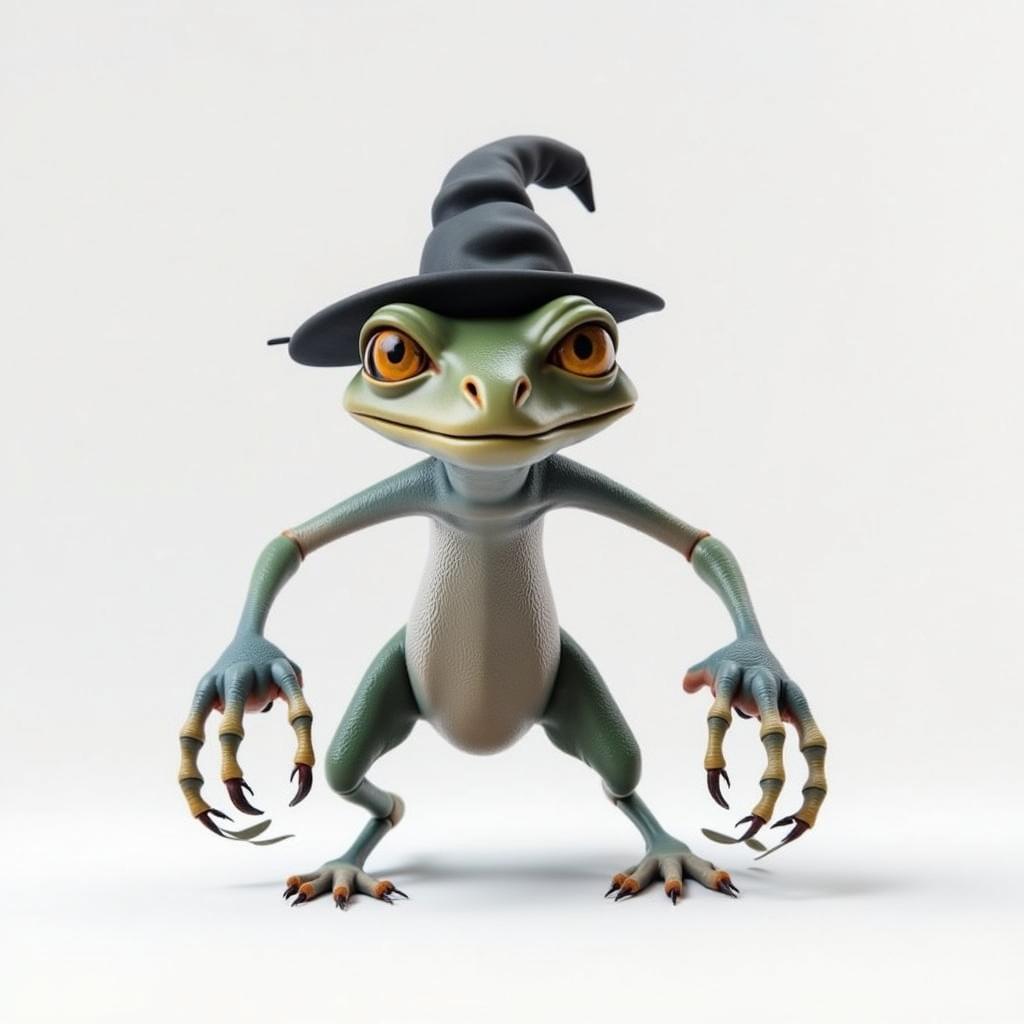} &

        \includegraphics[height=0.135\textwidth]{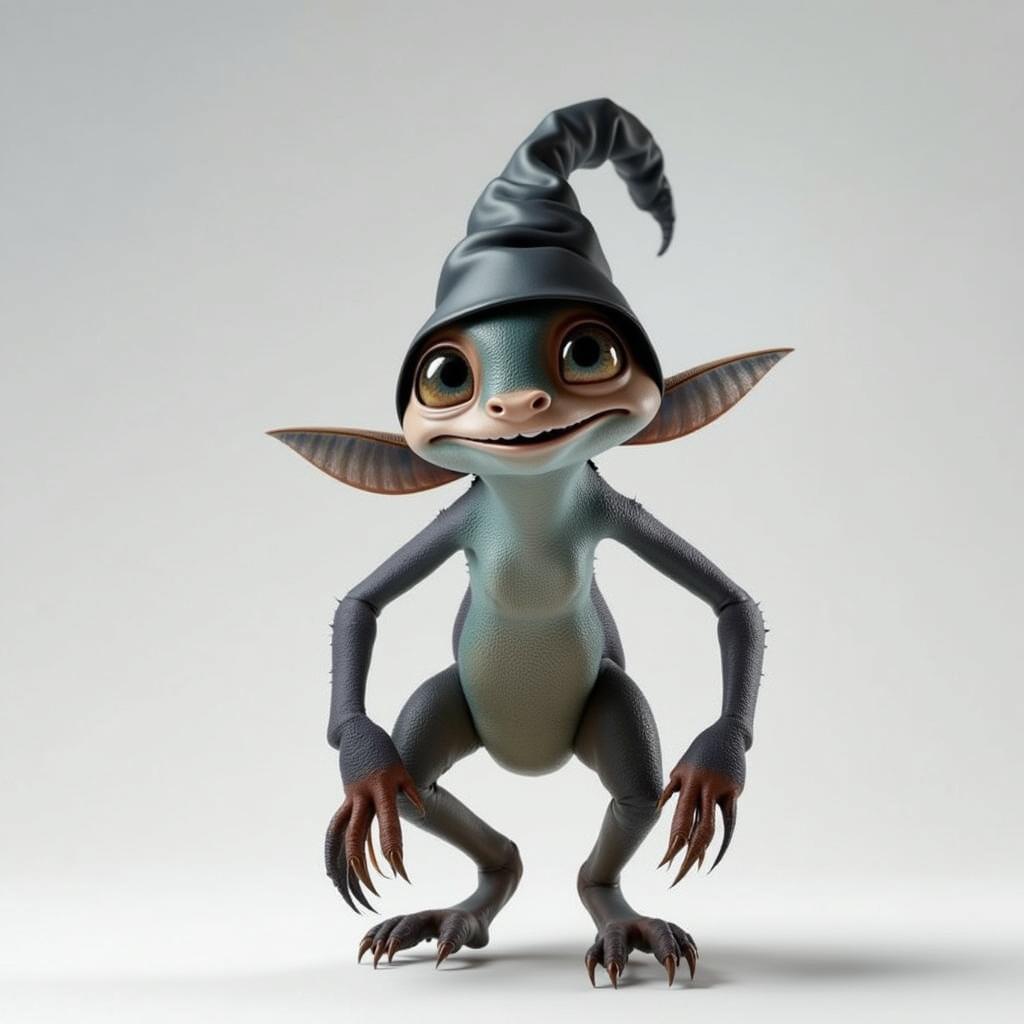} &

        \includegraphics[height=0.135\textwidth]{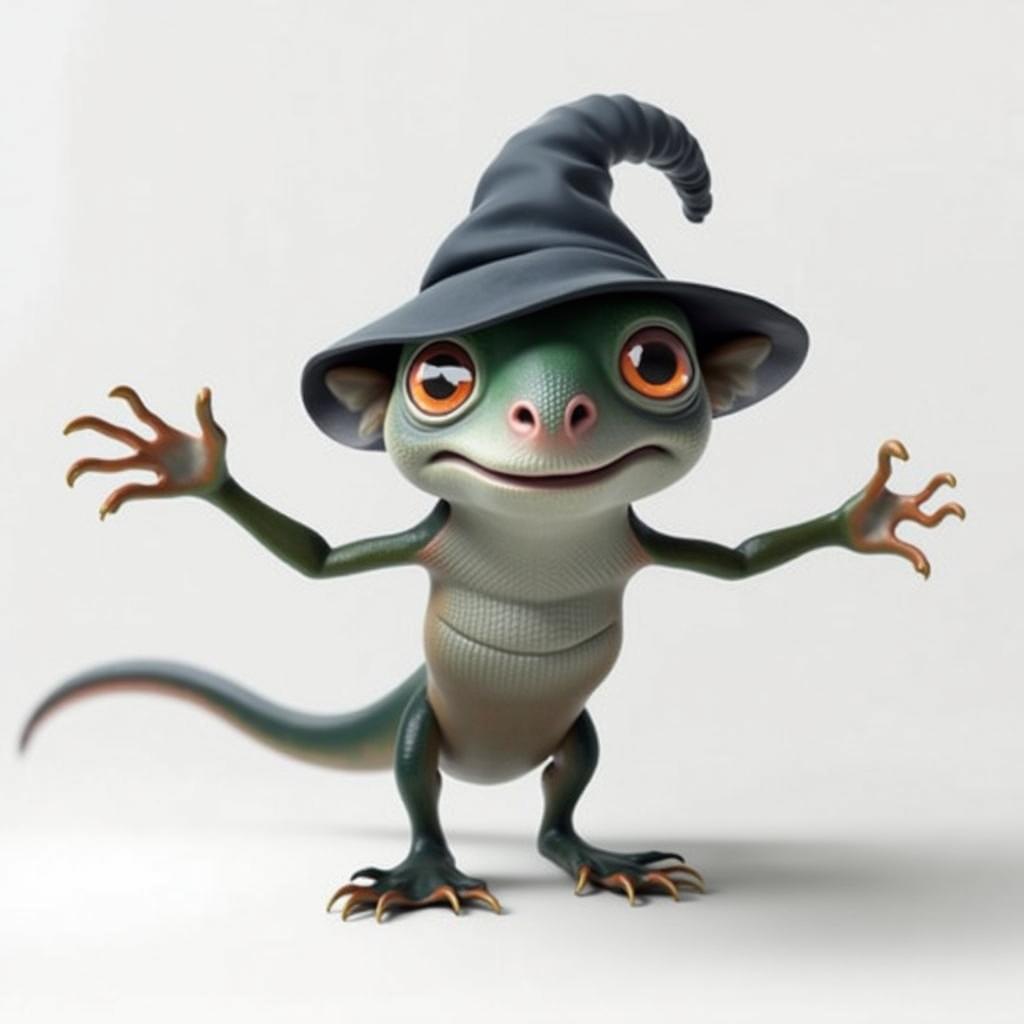} 

        \\
           Input & \multicolumn{3}{c}{Sampled Results} &    Input & \multicolumn{3}{c}{Sampled Results}

    \end{tabular}
    }
    \vspace{-0.2cm}
    \caption{Additional PiT results for character ideation
    }
    \label{fig:supp_charactes_1}
\end{figure*}

\begin{figure}
    \centering
    \setlength{\tabcolsep}{0.5pt}
    \renewcommand{\arraystretch}{0.5}
    {
    \begin{tabular}{c @{\hspace{0.2cm}} c c c}

        \includegraphics[height=0.1\textheight]{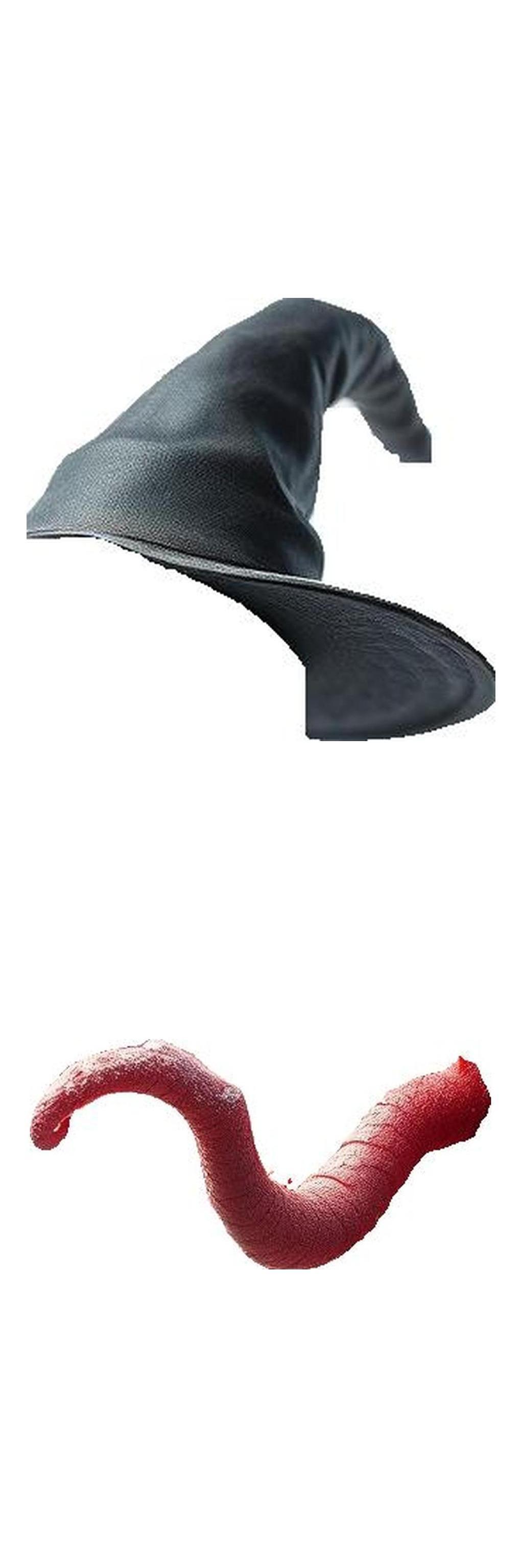} &
        \includegraphics[height=0.1\textheight]{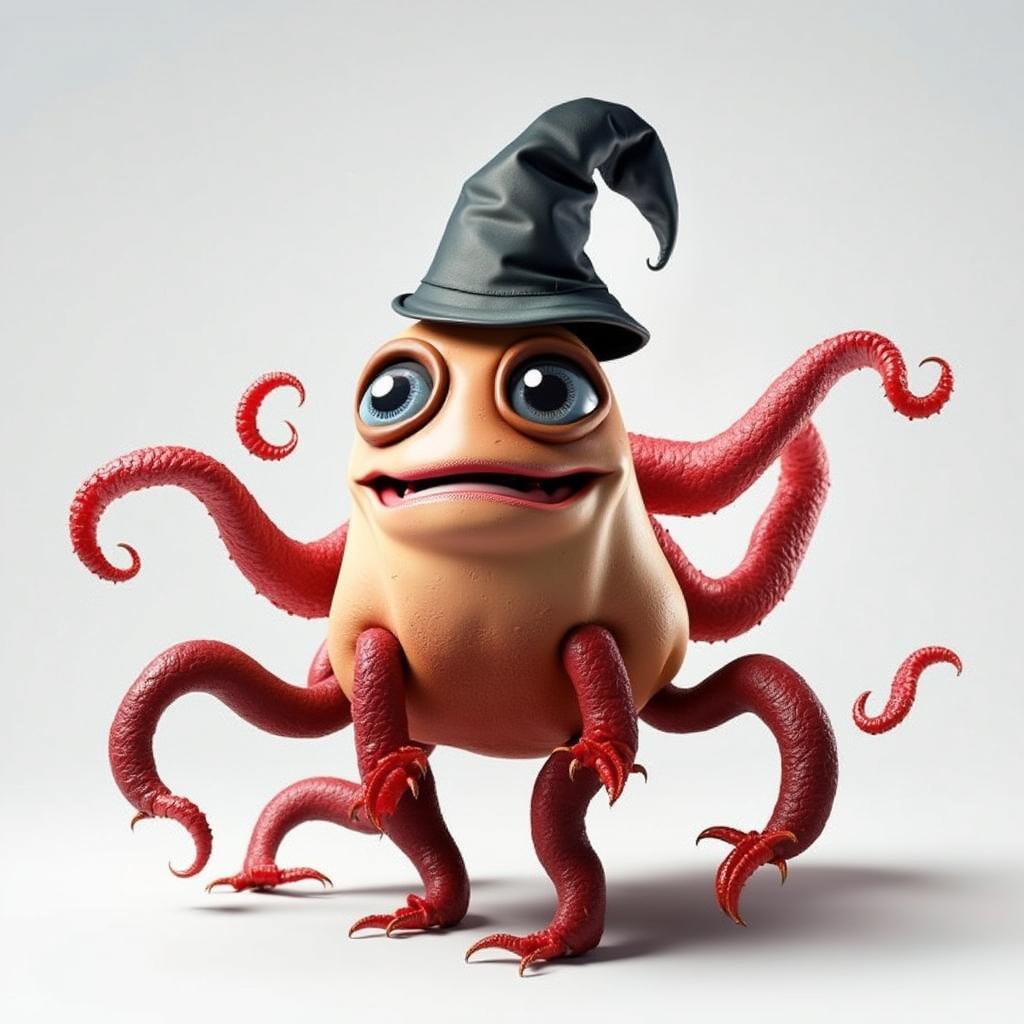} &
        \includegraphics[height=0.1\textheight]{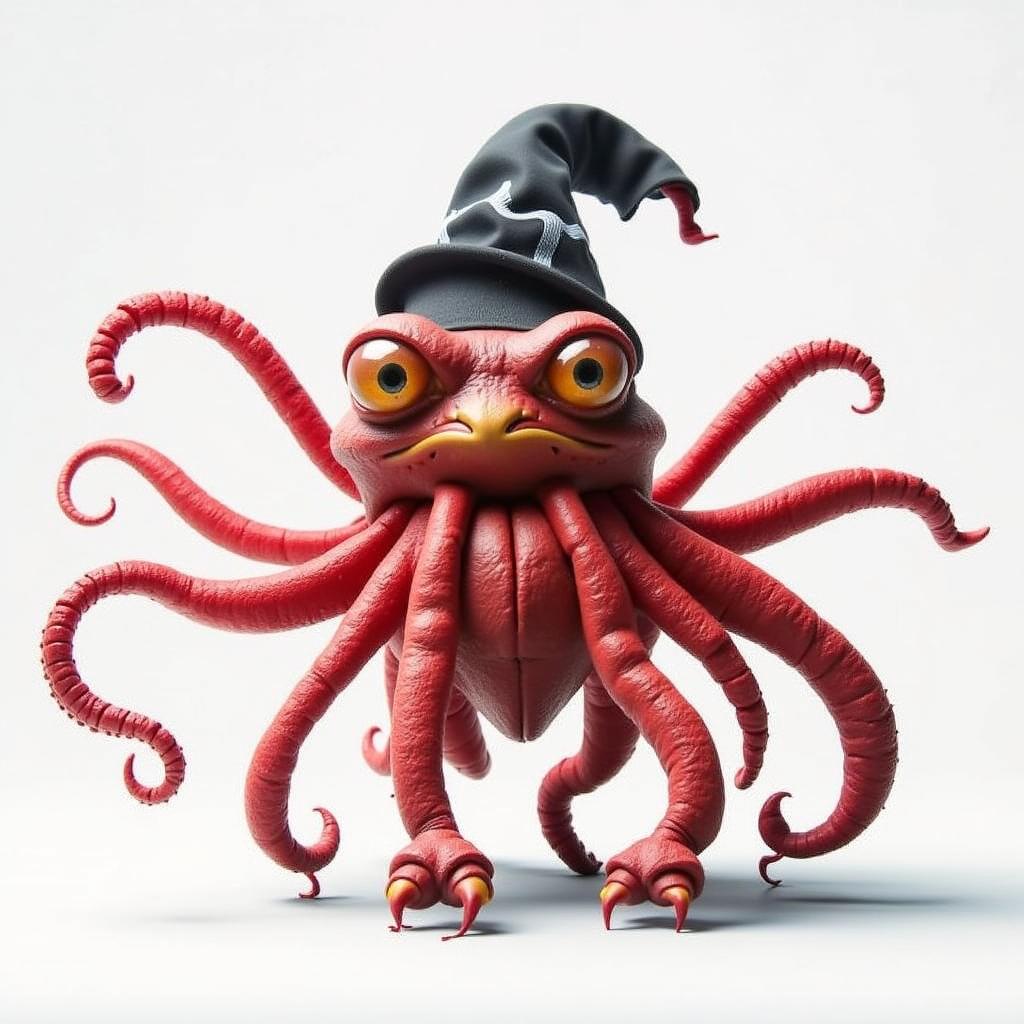} &

        \includegraphics[height=0.1\textheight]{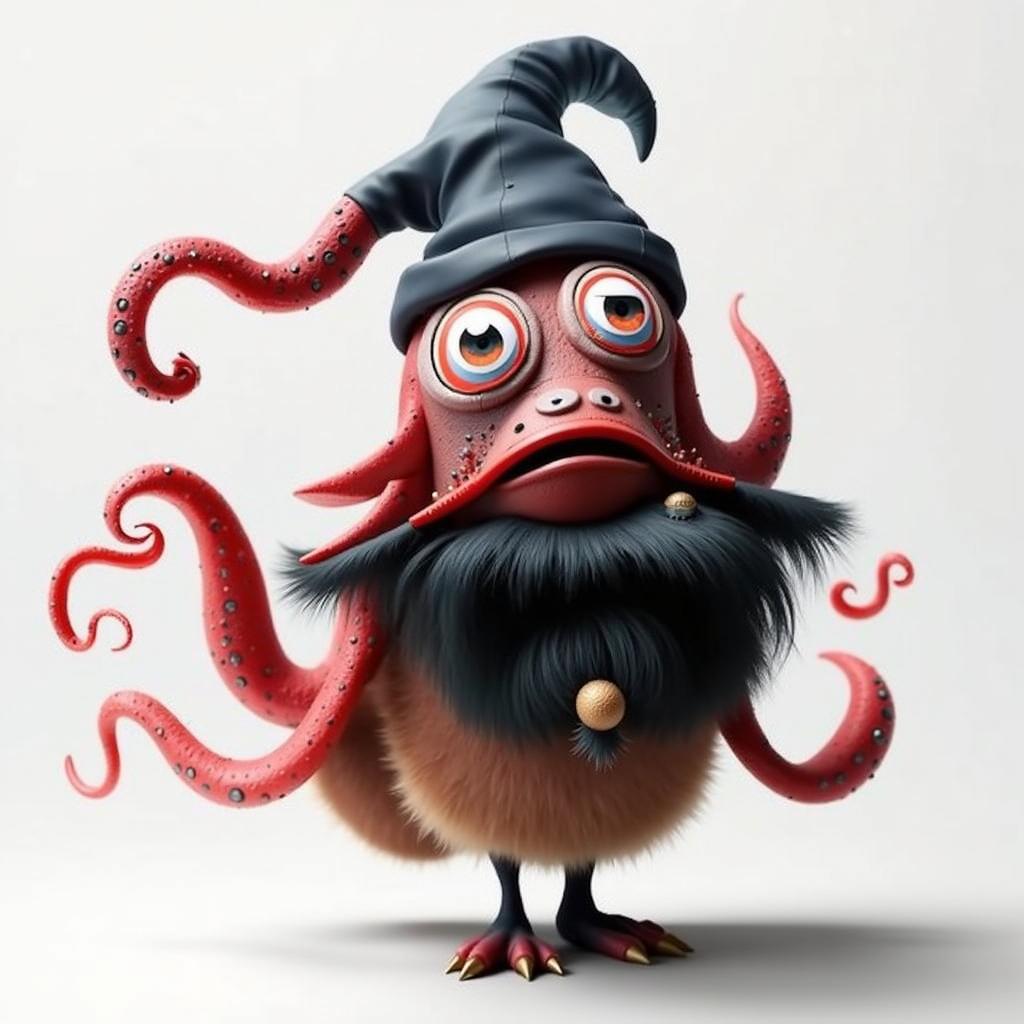} 
        \\

        \includegraphics[height=0.1\textheight]{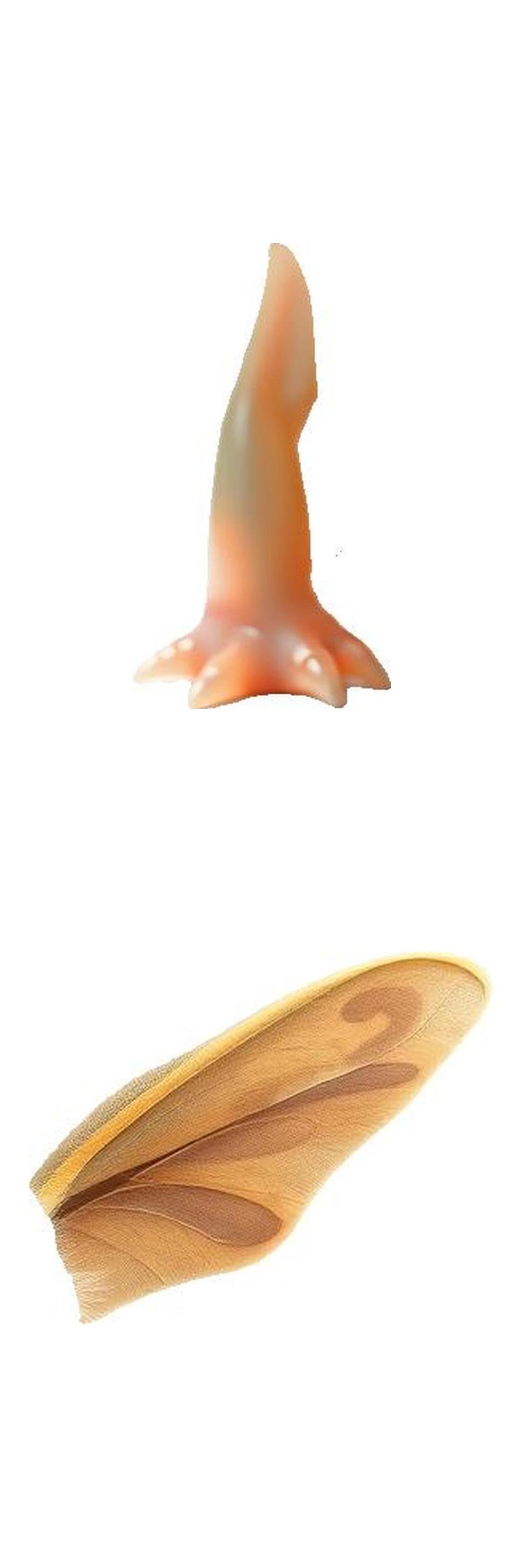} &
        \includegraphics[height=0.1\textheight]{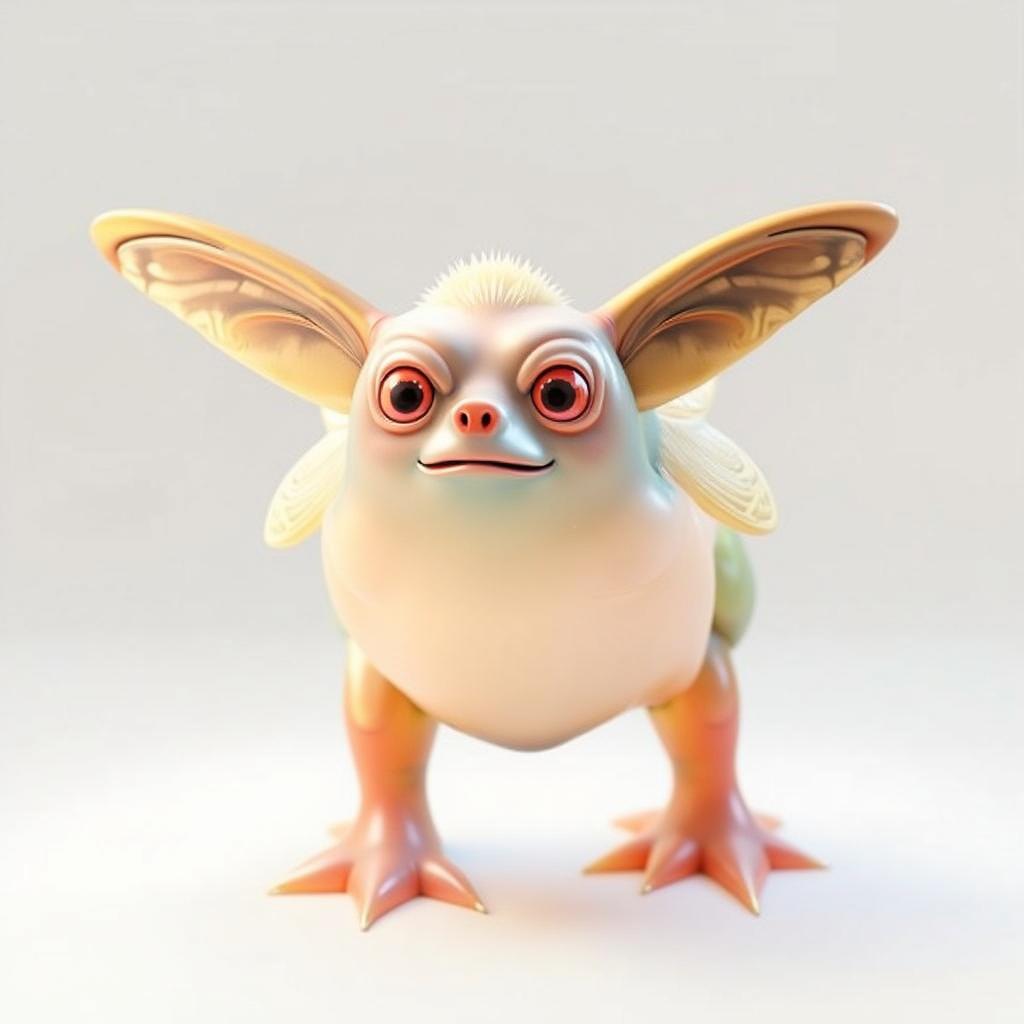} &
        \includegraphics[height=0.1\textheight]{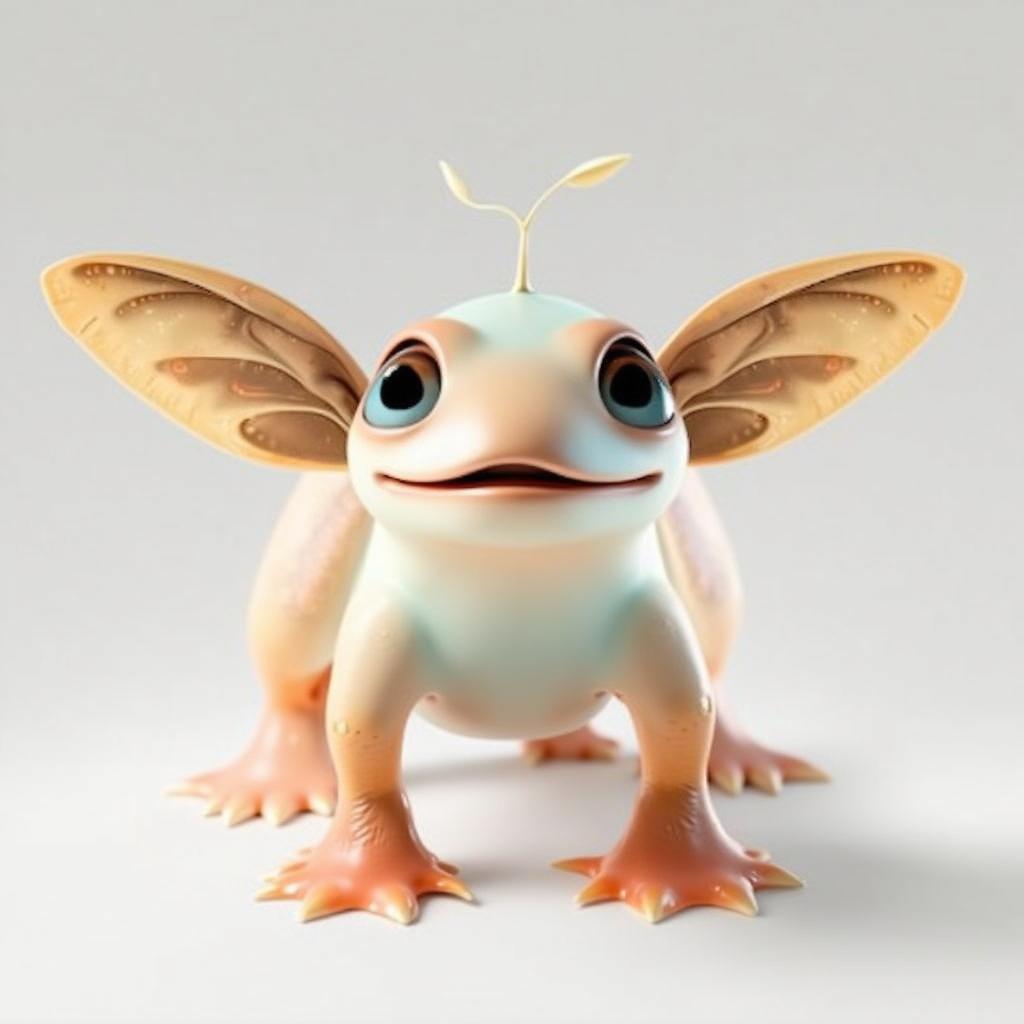} &

        \includegraphics[height=0.1\textheight]{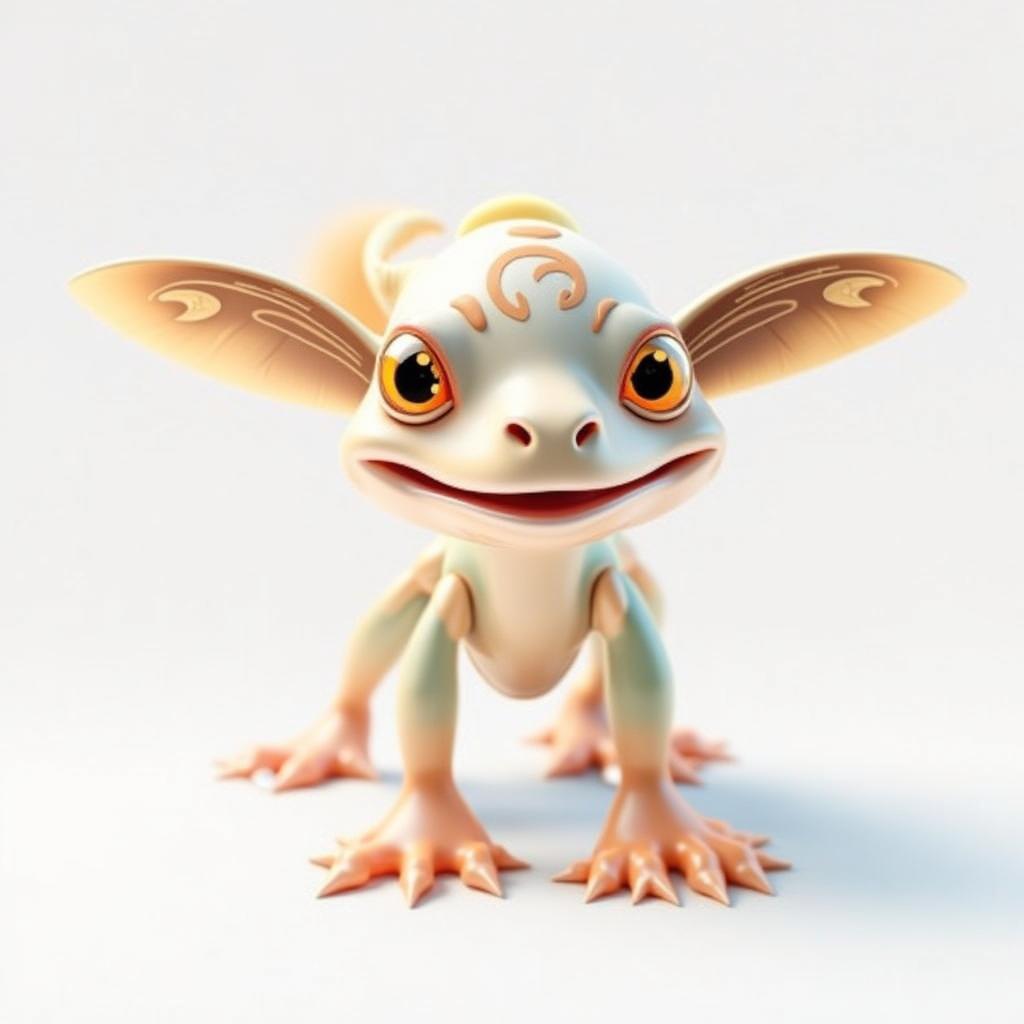} 
        \\

        \includegraphics[height=0.1\textheight]{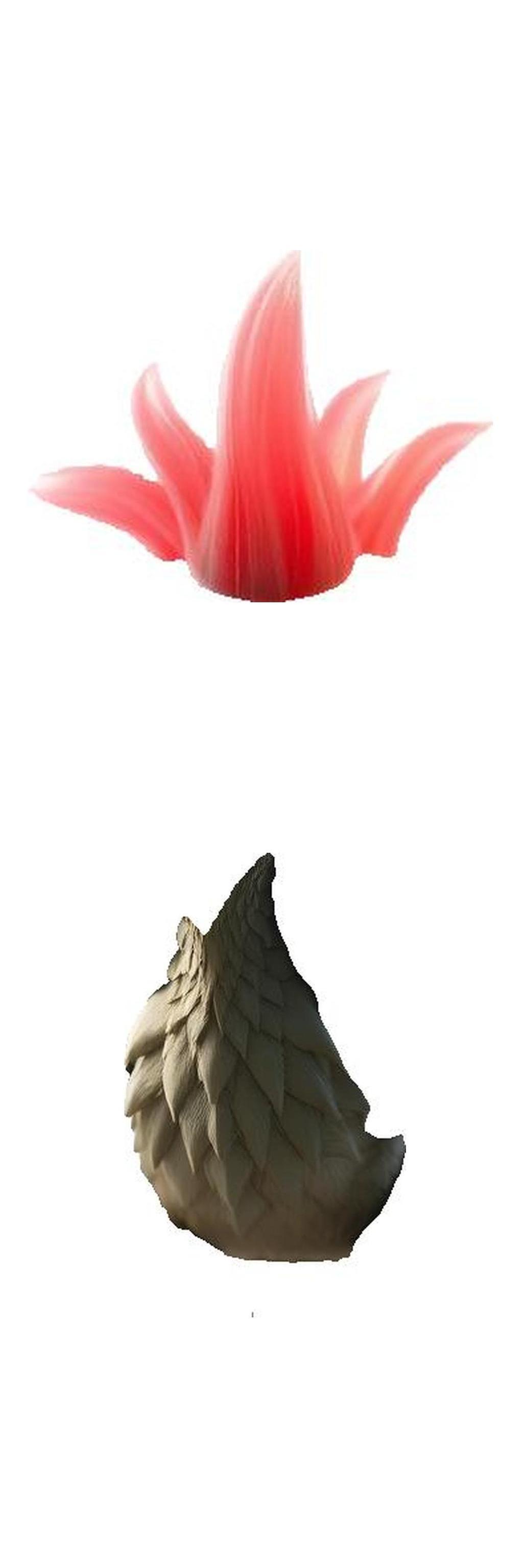} &
        \includegraphics[height=0.1\textheight]{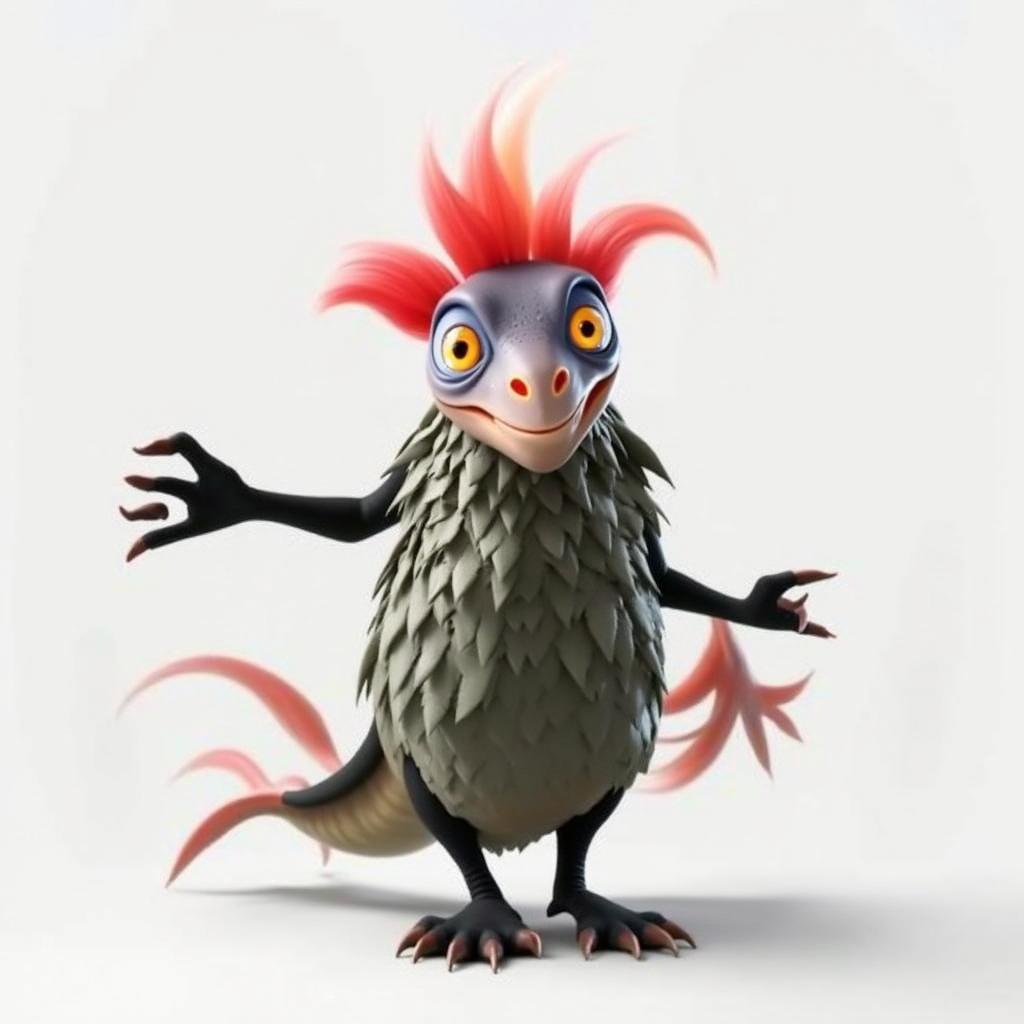} &
        \includegraphics[height=0.1\textheight]{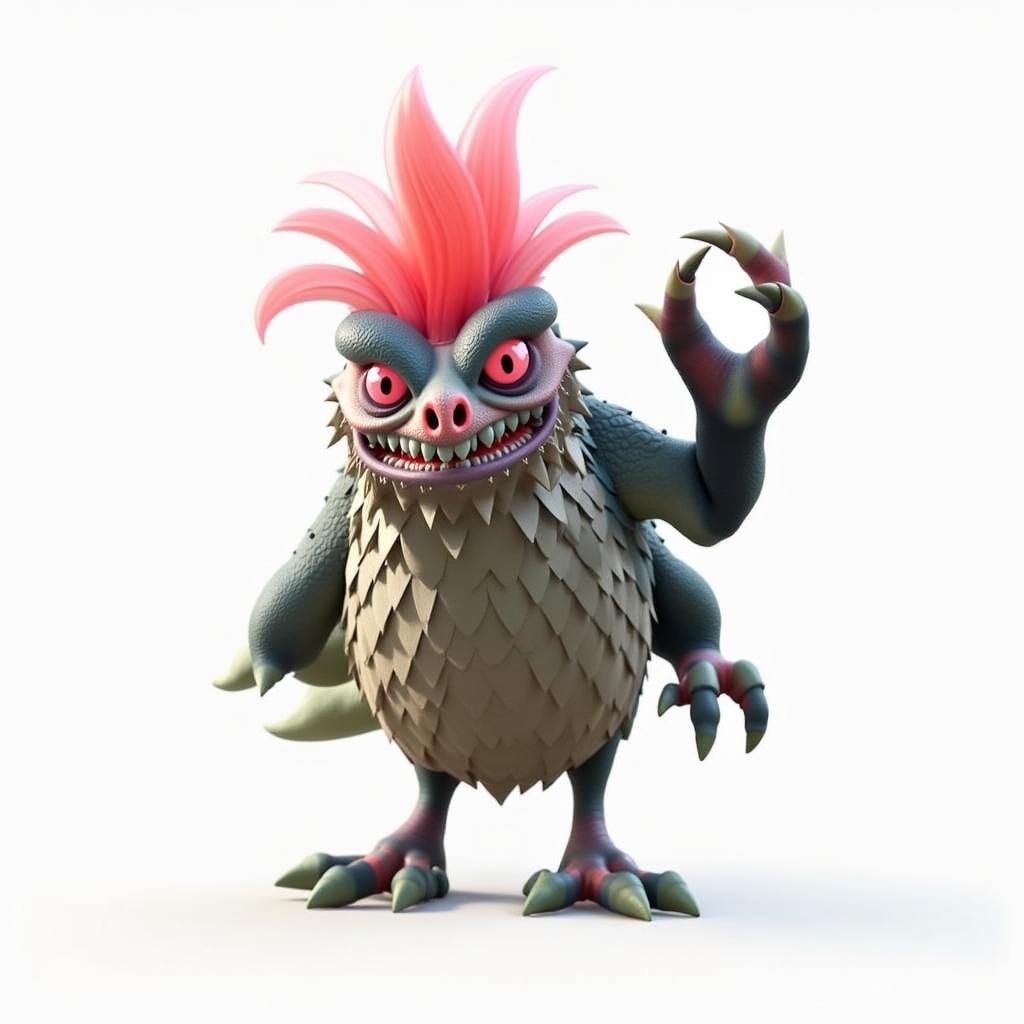} &

        \includegraphics[height=0.1\textheight]{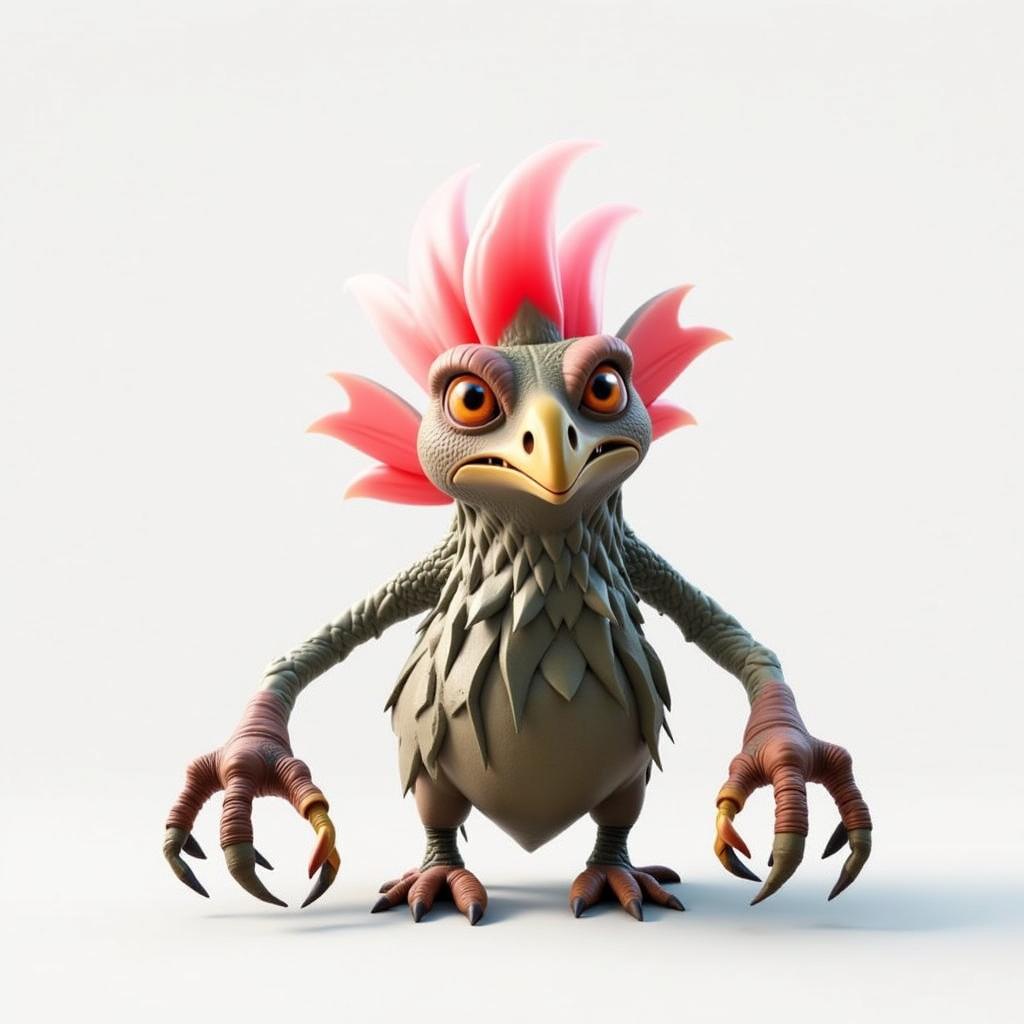} 
        \\

                \includegraphics[height=0.1\textheight]{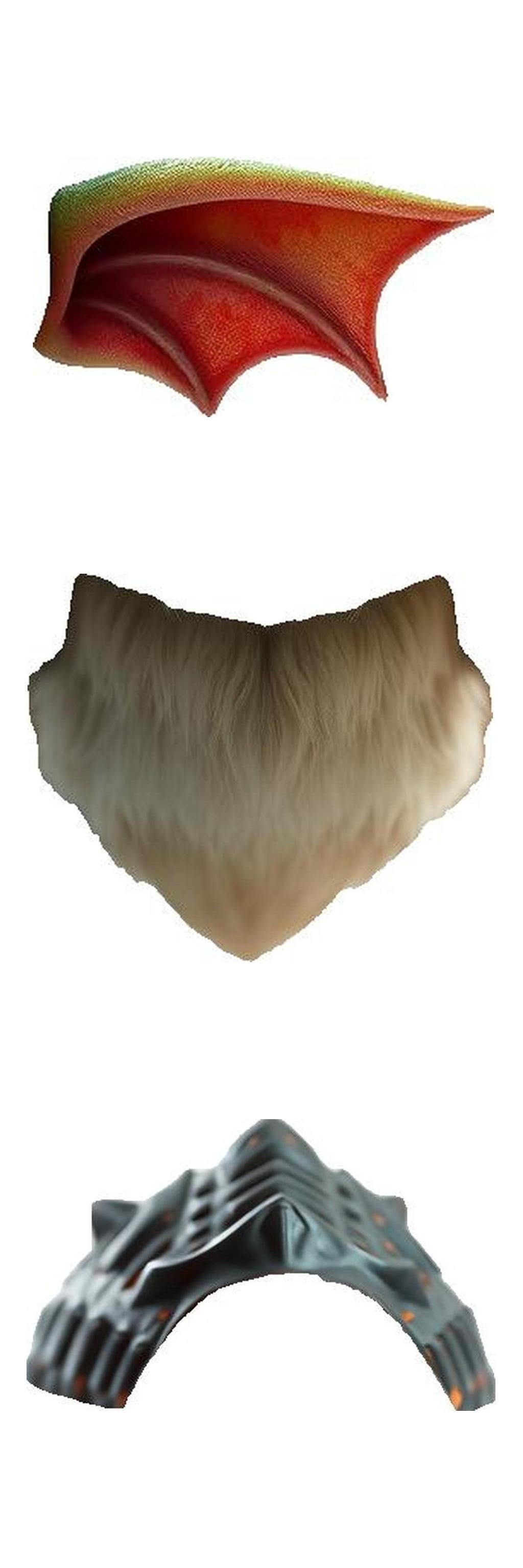} &
        \includegraphics[height=0.1\textheight]{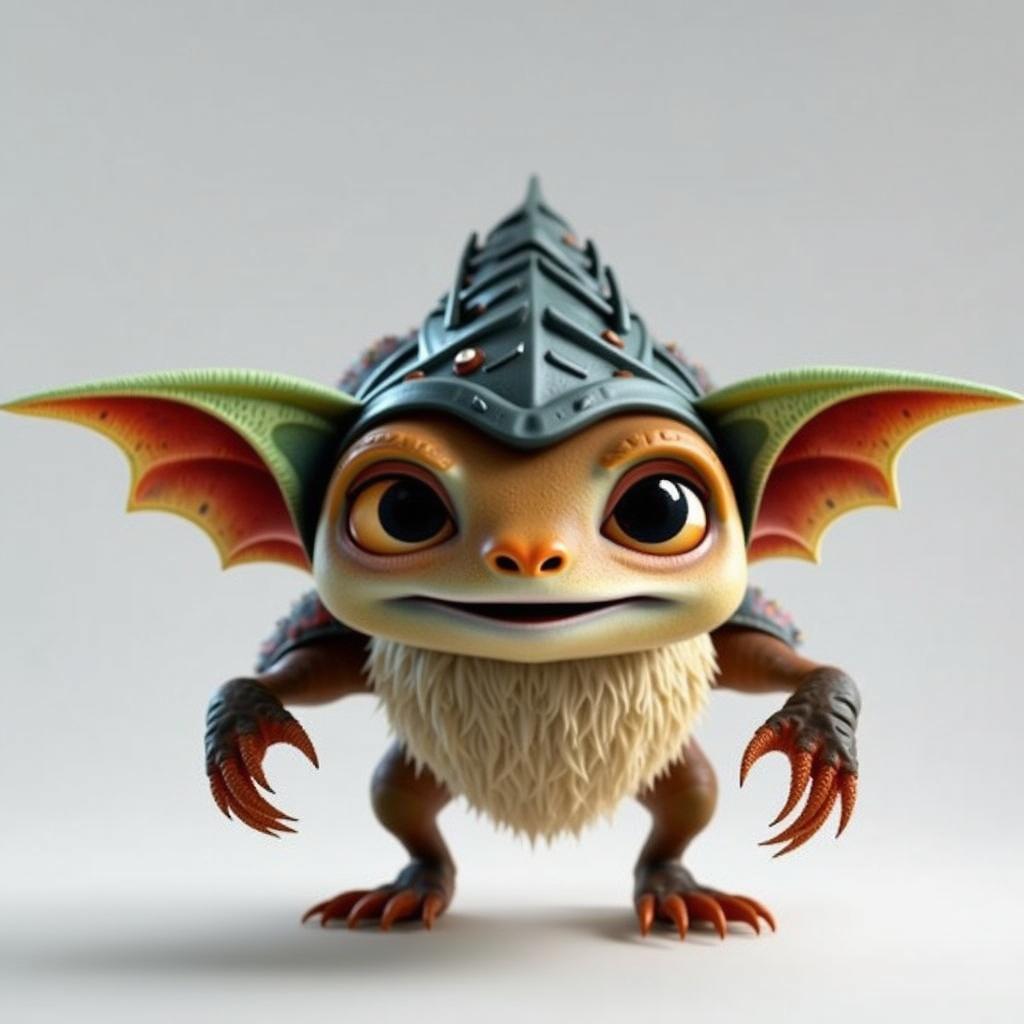} &
        \includegraphics[height=0.1\textheight]{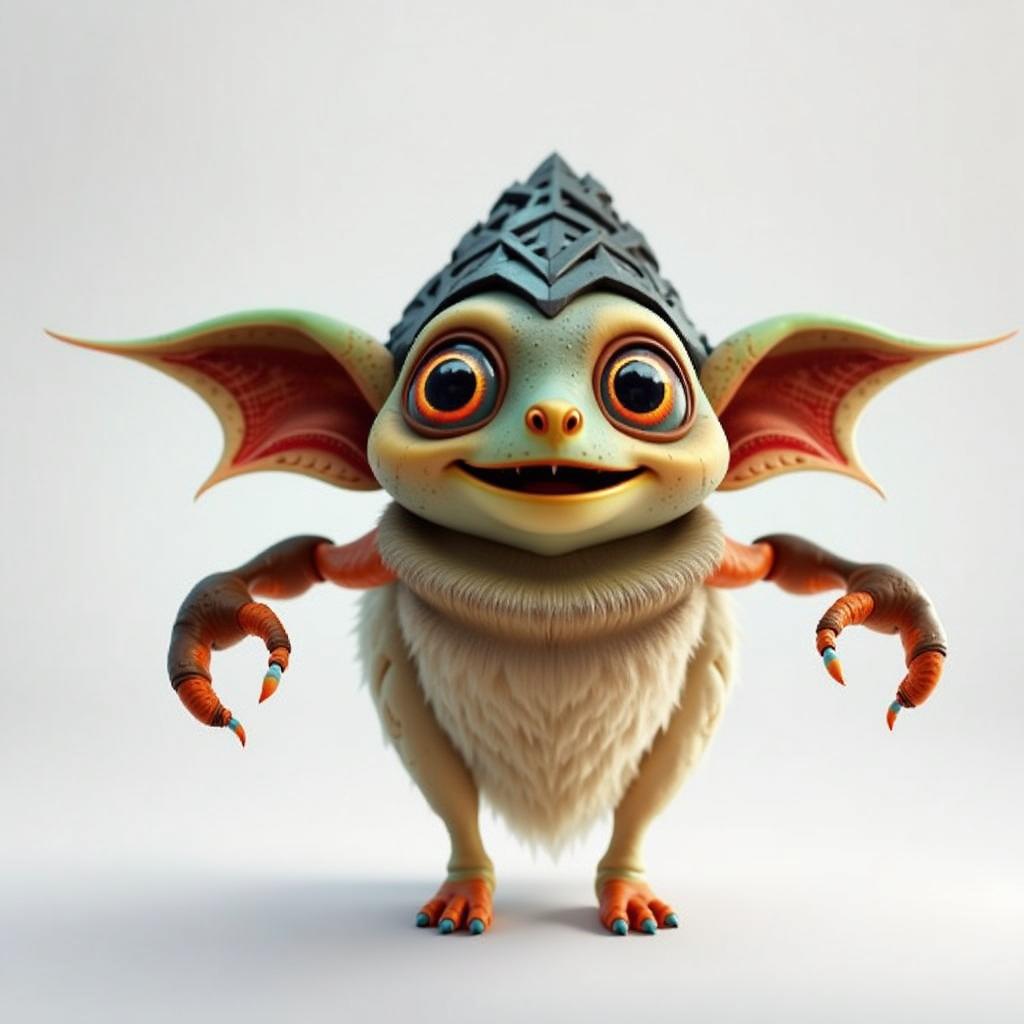} &

        \includegraphics[height=0.1\textheight]{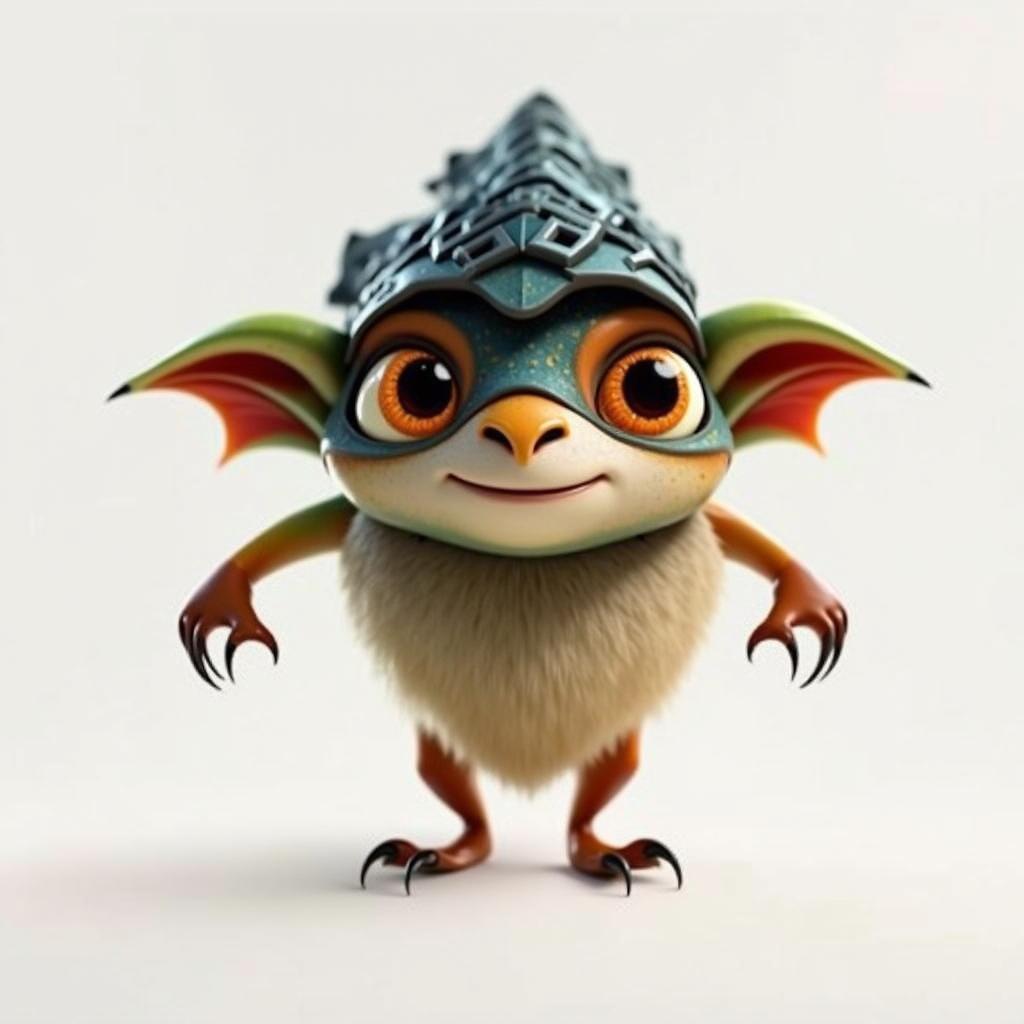} 
        \\

        \includegraphics[height=0.1\textheight]{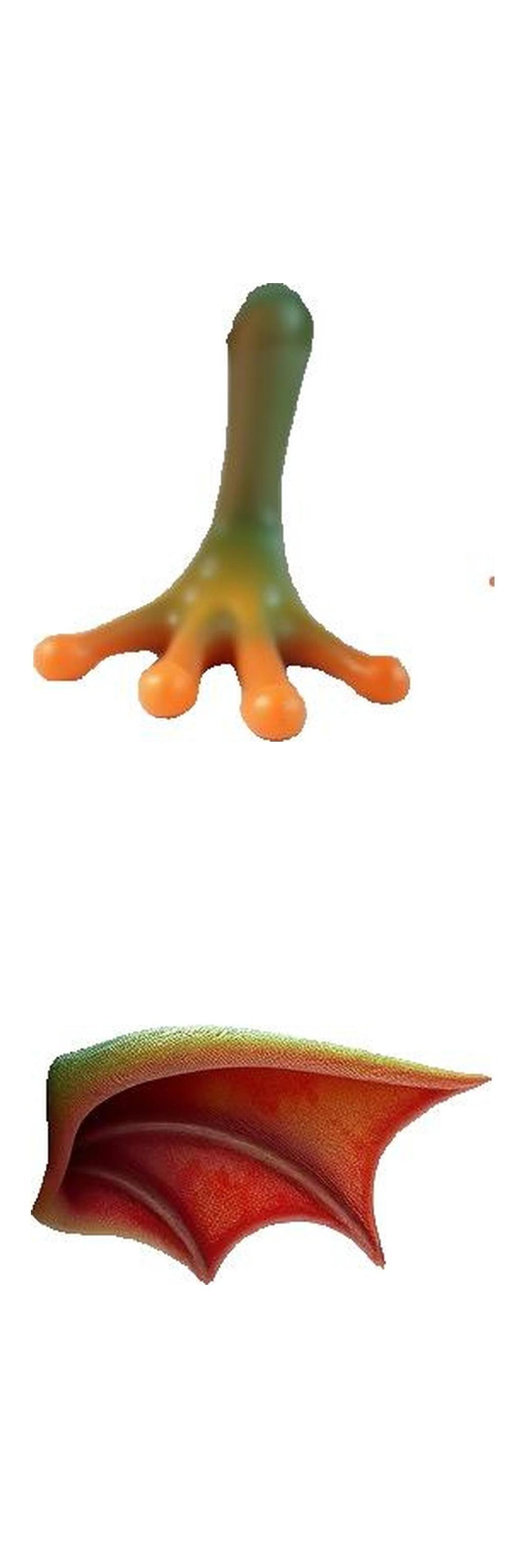} &
        \includegraphics[height=0.1\textheight]{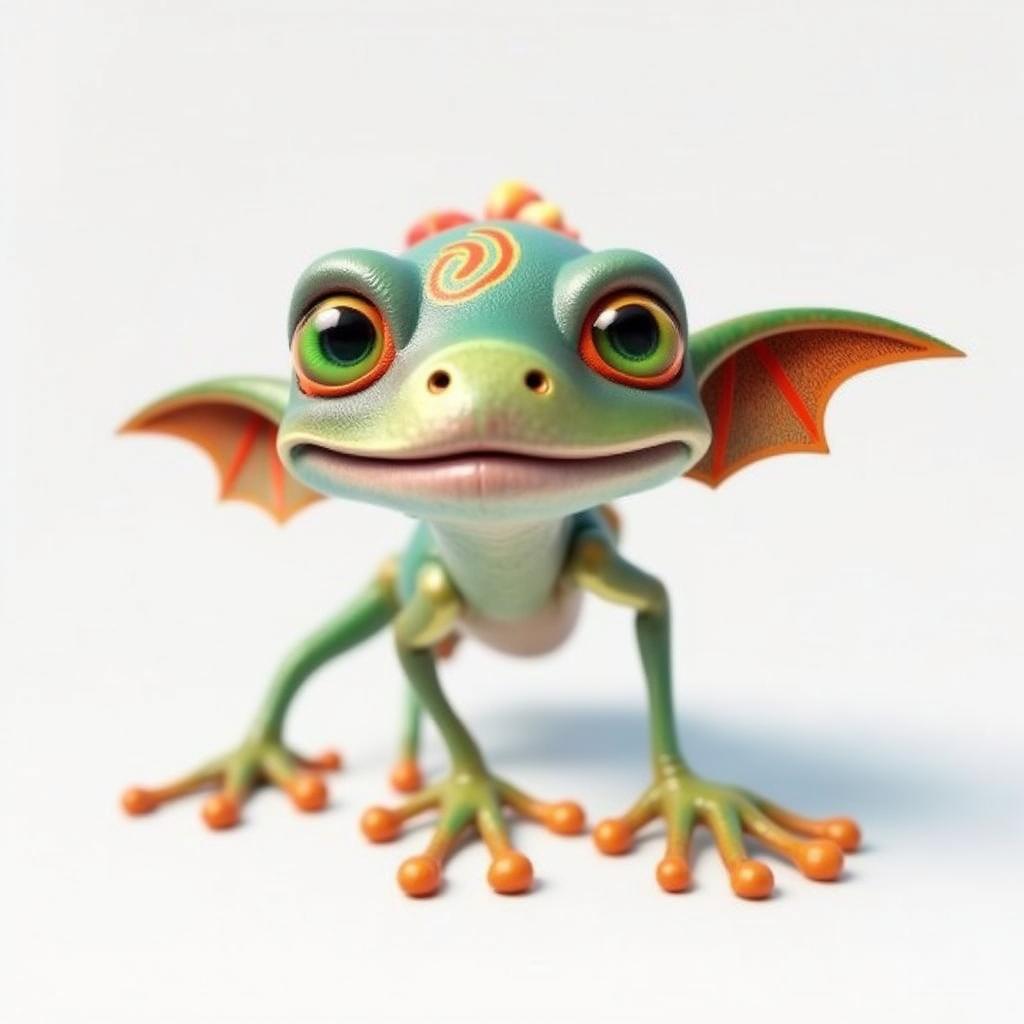} &
        \includegraphics[height=0.1\textheight]{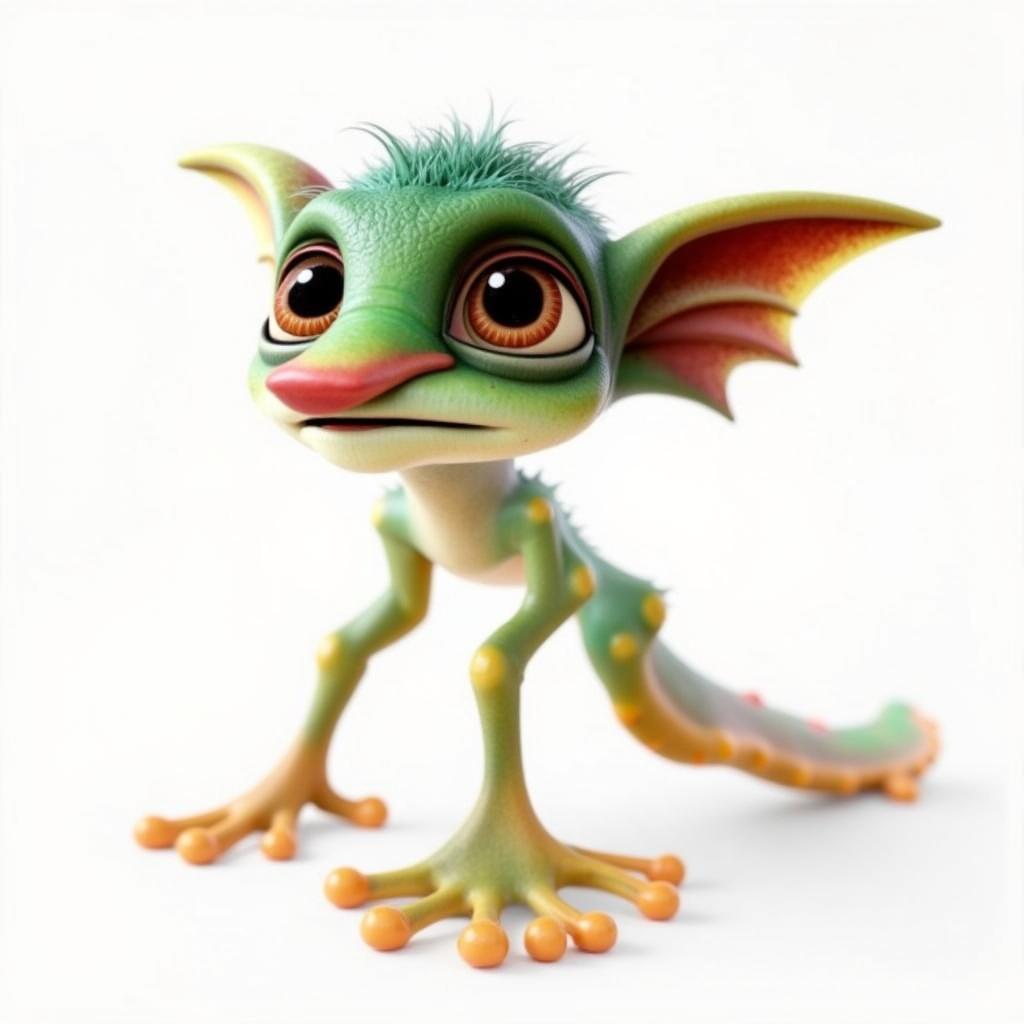} &

        \includegraphics[height=0.1\textheight]{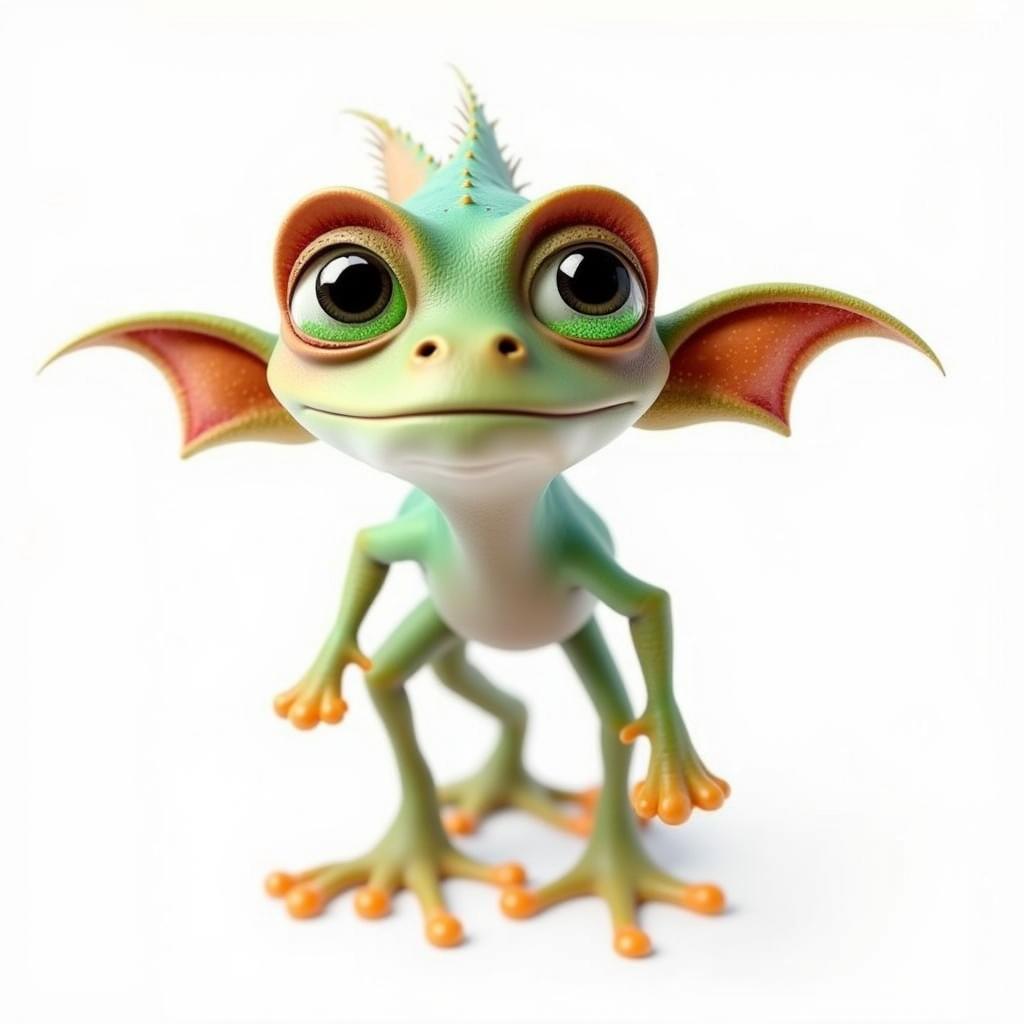} 
        \\

        \includegraphics[height=0.1\textheight]{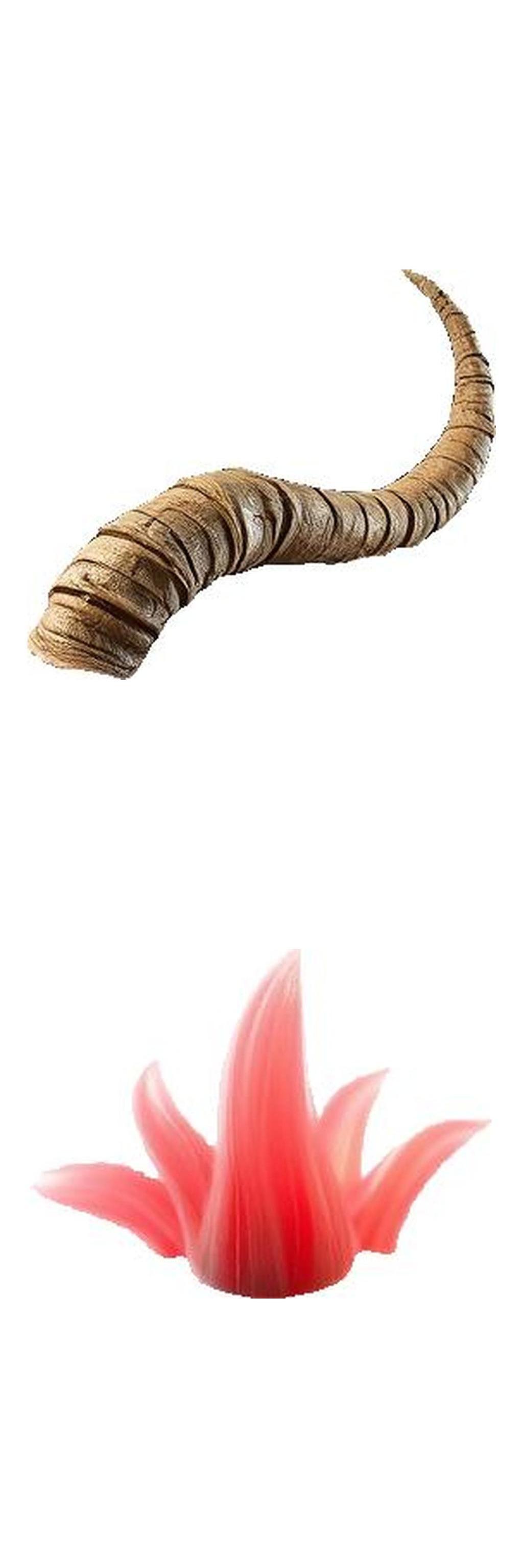} &
        \includegraphics[height=0.1\textheight]{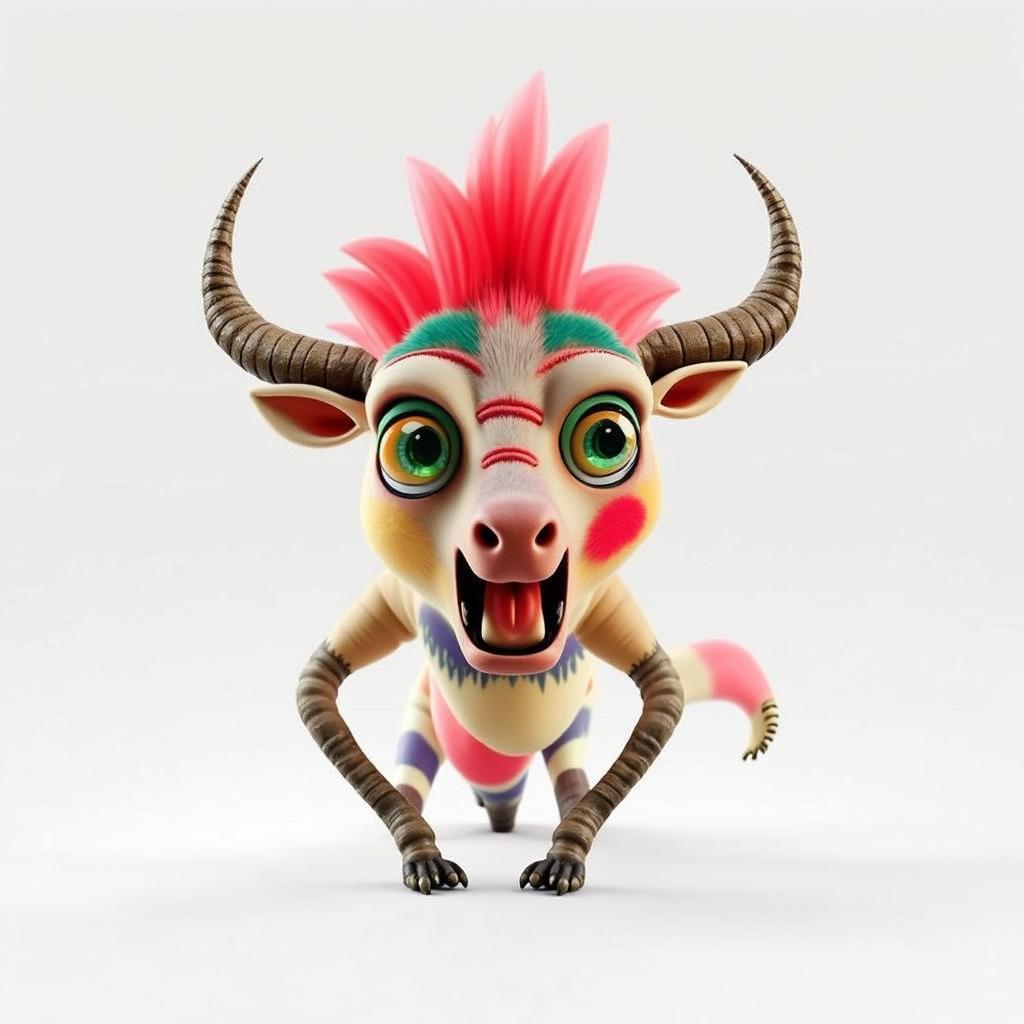} &
        \includegraphics[height=0.1\textheight]{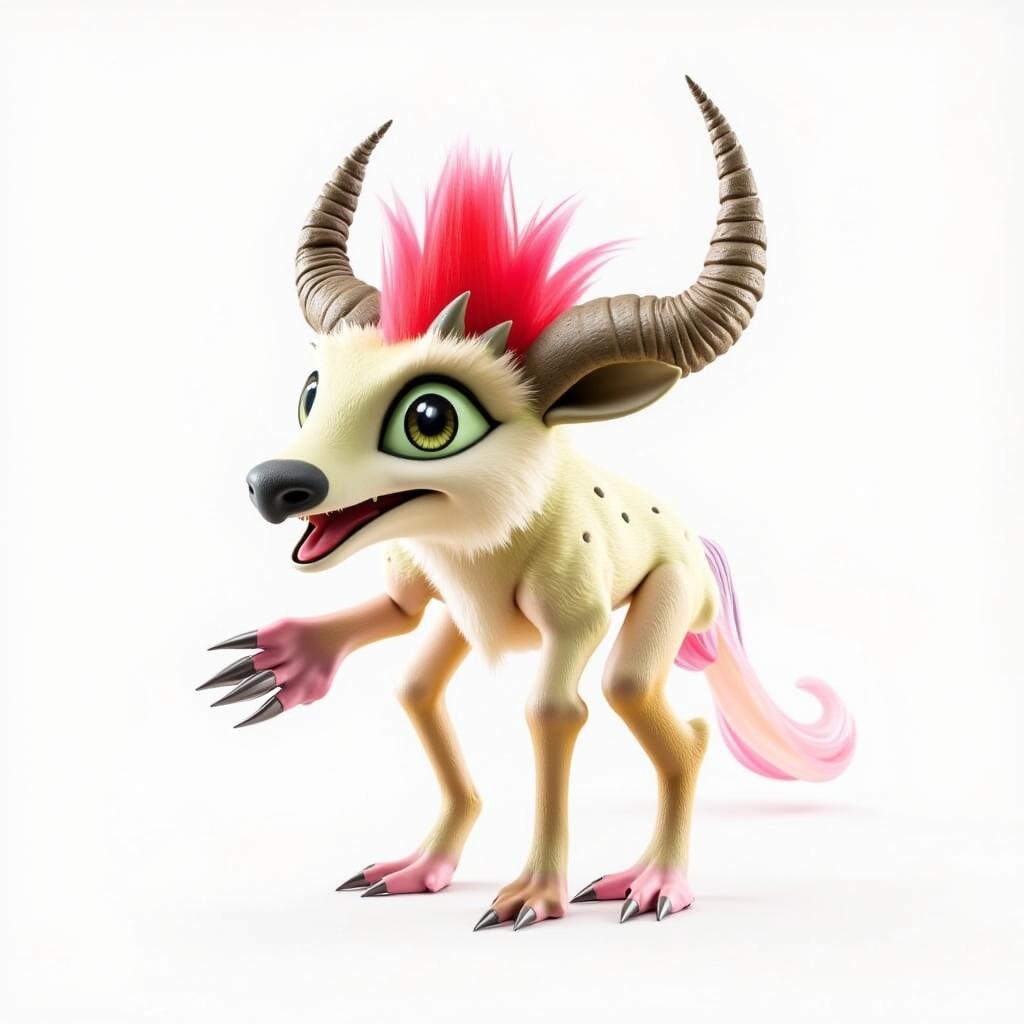} &

        \includegraphics[height=0.1\textheight]{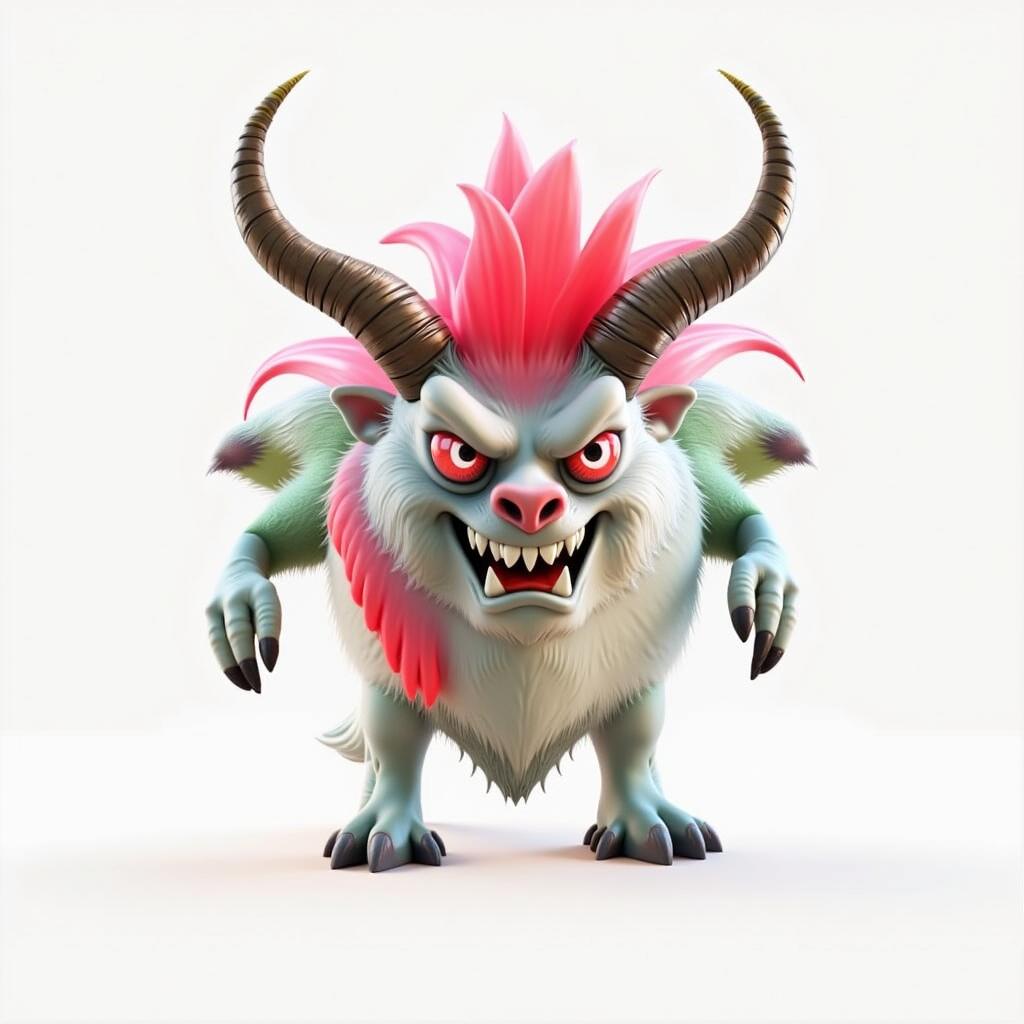} 
        \\

        \\      

        Input & \multicolumn{3}{c}{Sampled Results} 

    \end{tabular}
    }
    \caption{Additional PiT results for character ideation}
    \label{fig:supp_characters_more}
\end{figure}

\begin{figure}
    \centering
    \setlength{\tabcolsep}{0.5pt}
    \renewcommand{\arraystretch}{0.5}
    {
    \begin{tabular}{c @{\hspace{0.2cm}} c c c}

    \includegraphics[height=0.1\textheight]{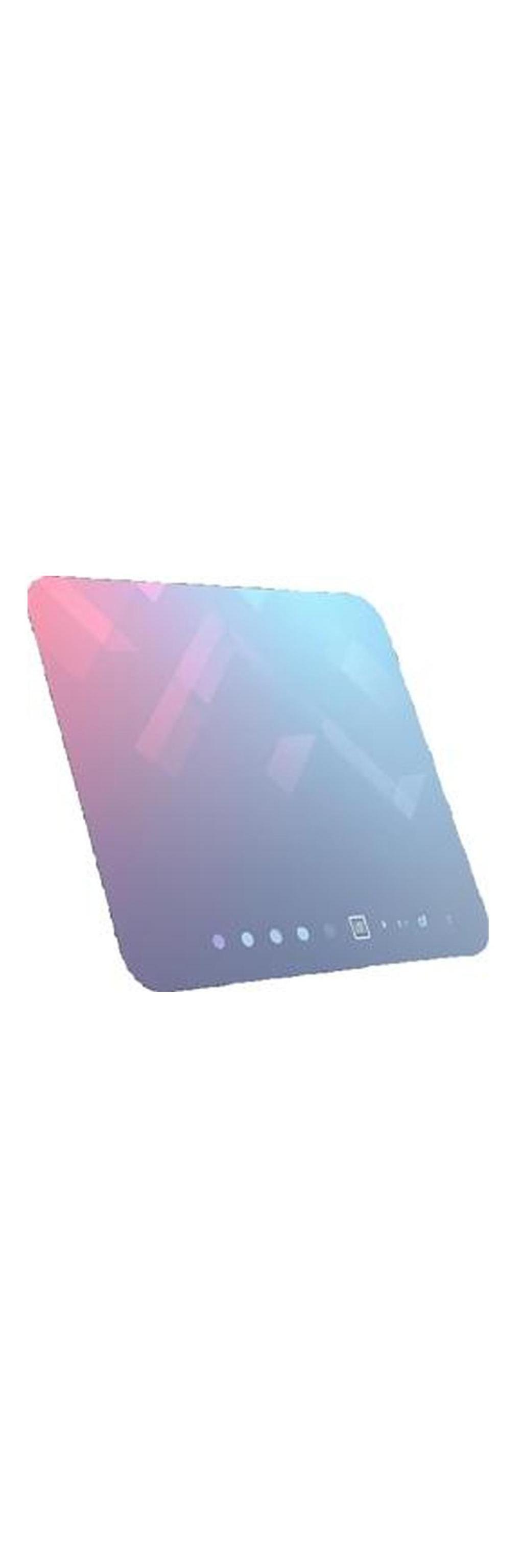} &
        \includegraphics[height=0.1\textheight]{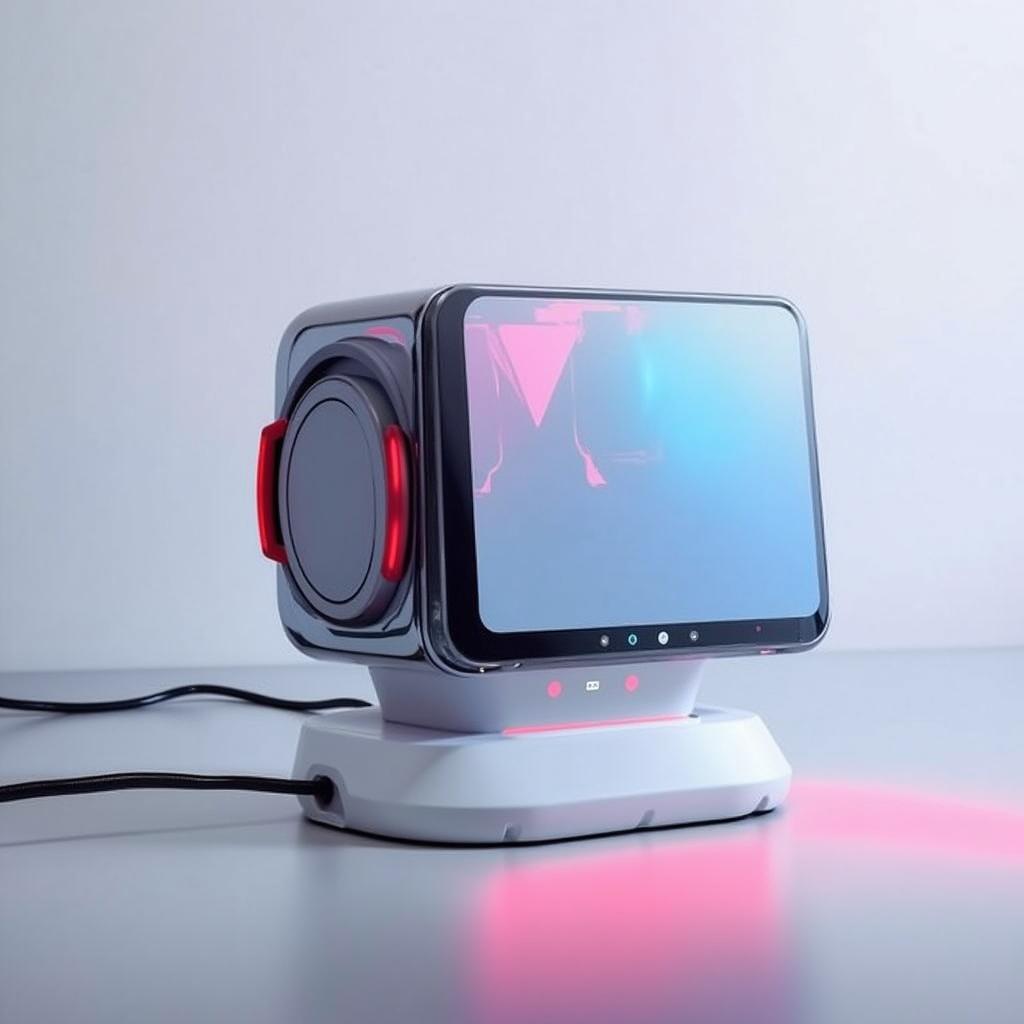} &
        \includegraphics[height=0.1\textheight]{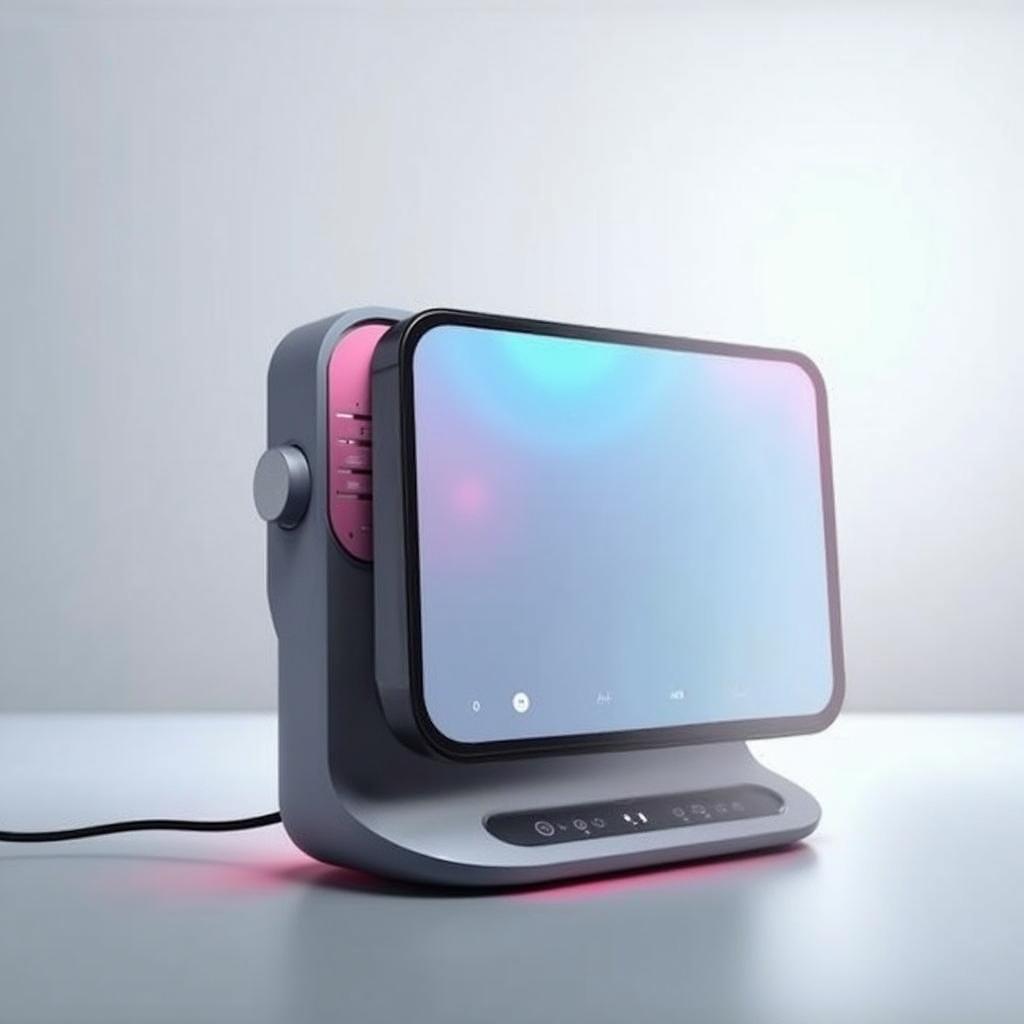} &

        \includegraphics[height=0.1\textheight]{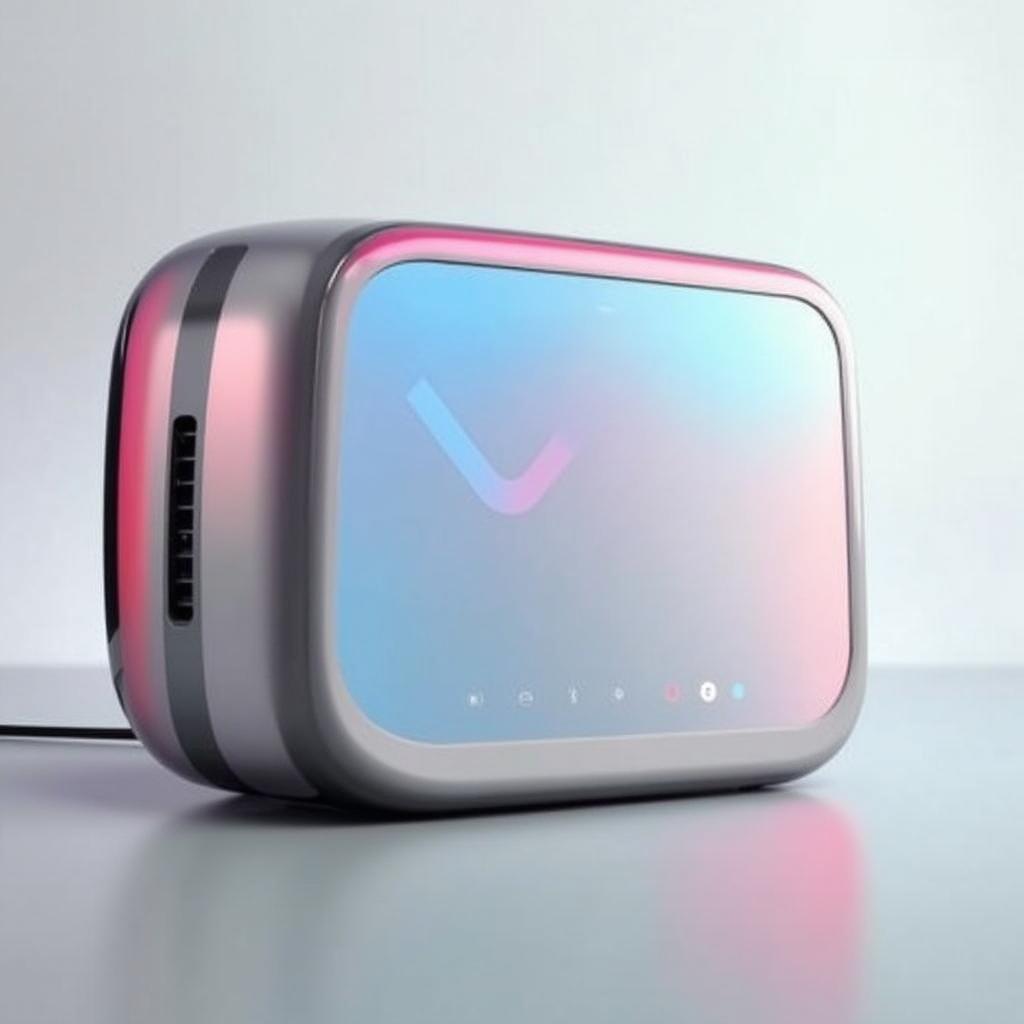} 
        \\

        \includegraphics[height=0.1\textheight]{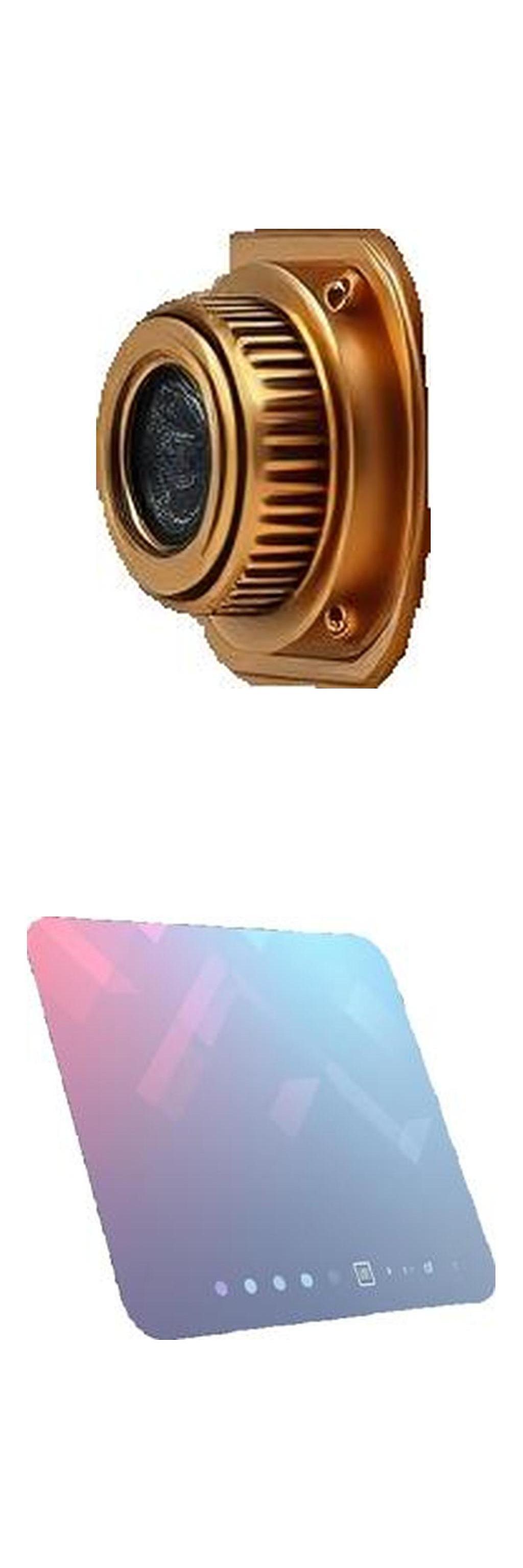} &
        \includegraphics[height=0.1\textheight]{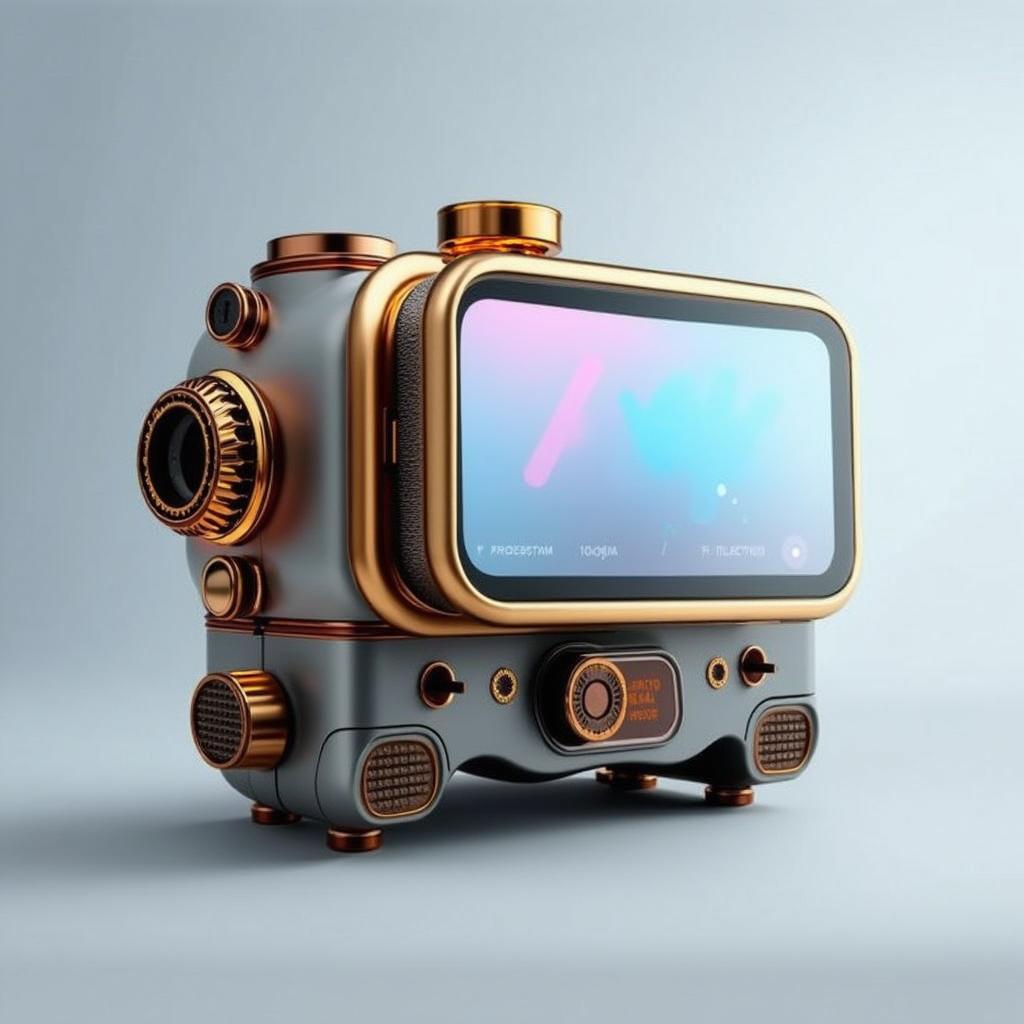} &
        \includegraphics[height=0.1\textheight]{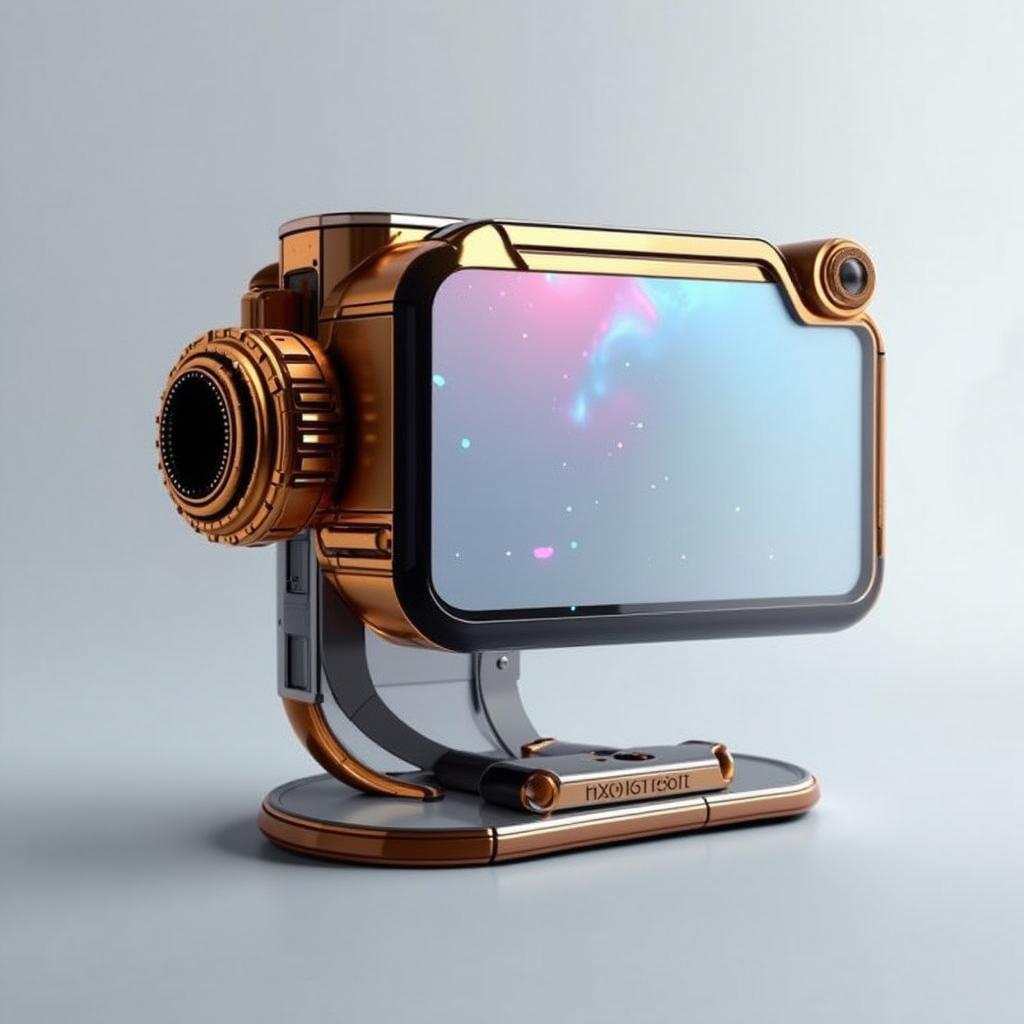} &

        \includegraphics[height=0.1\textheight]{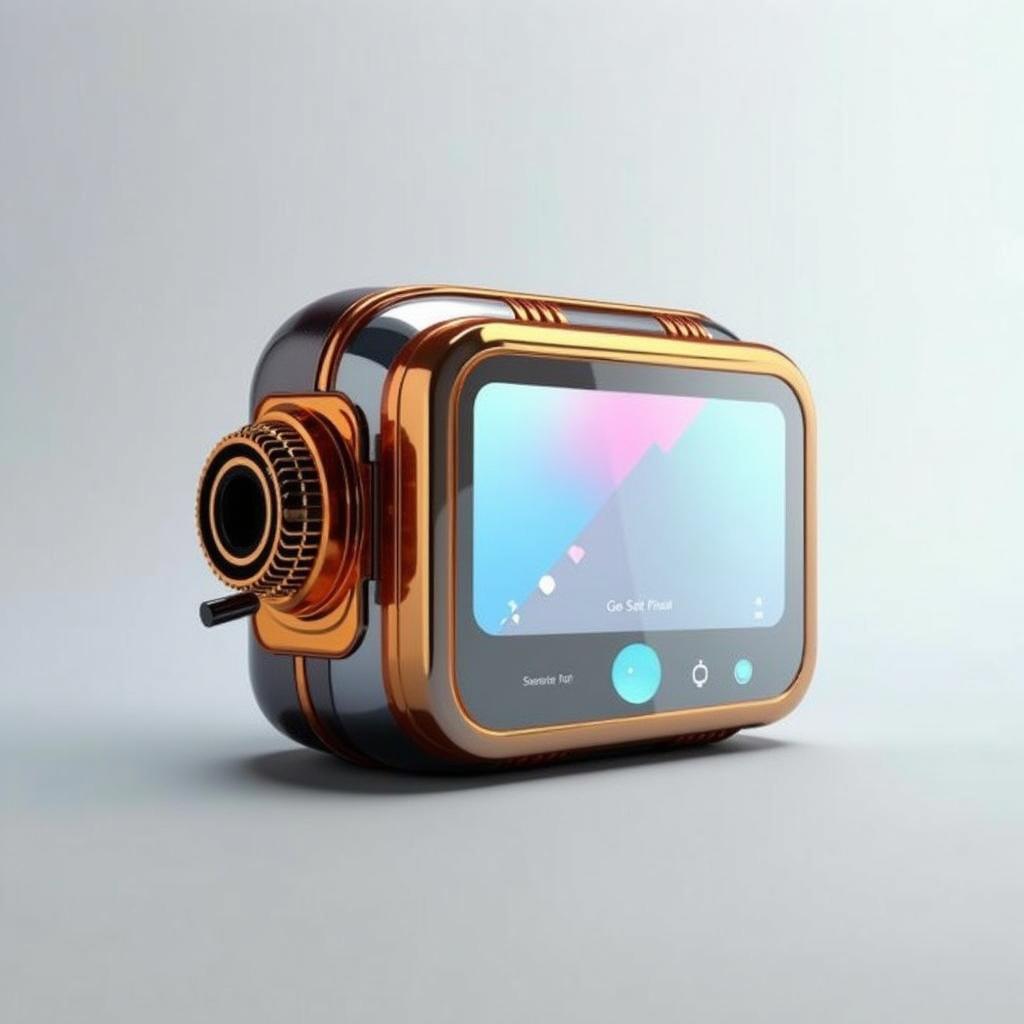} 
        \\

        \includegraphics[height=0.1\textheight]{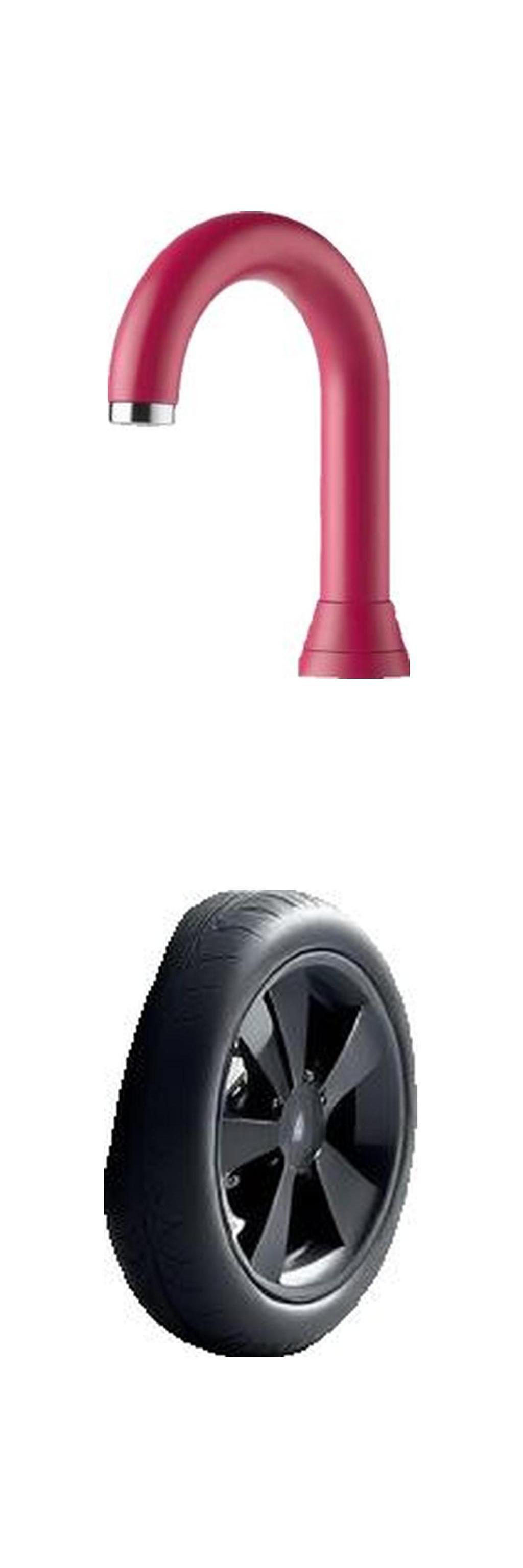} &
        \includegraphics[height=0.1\textheight]{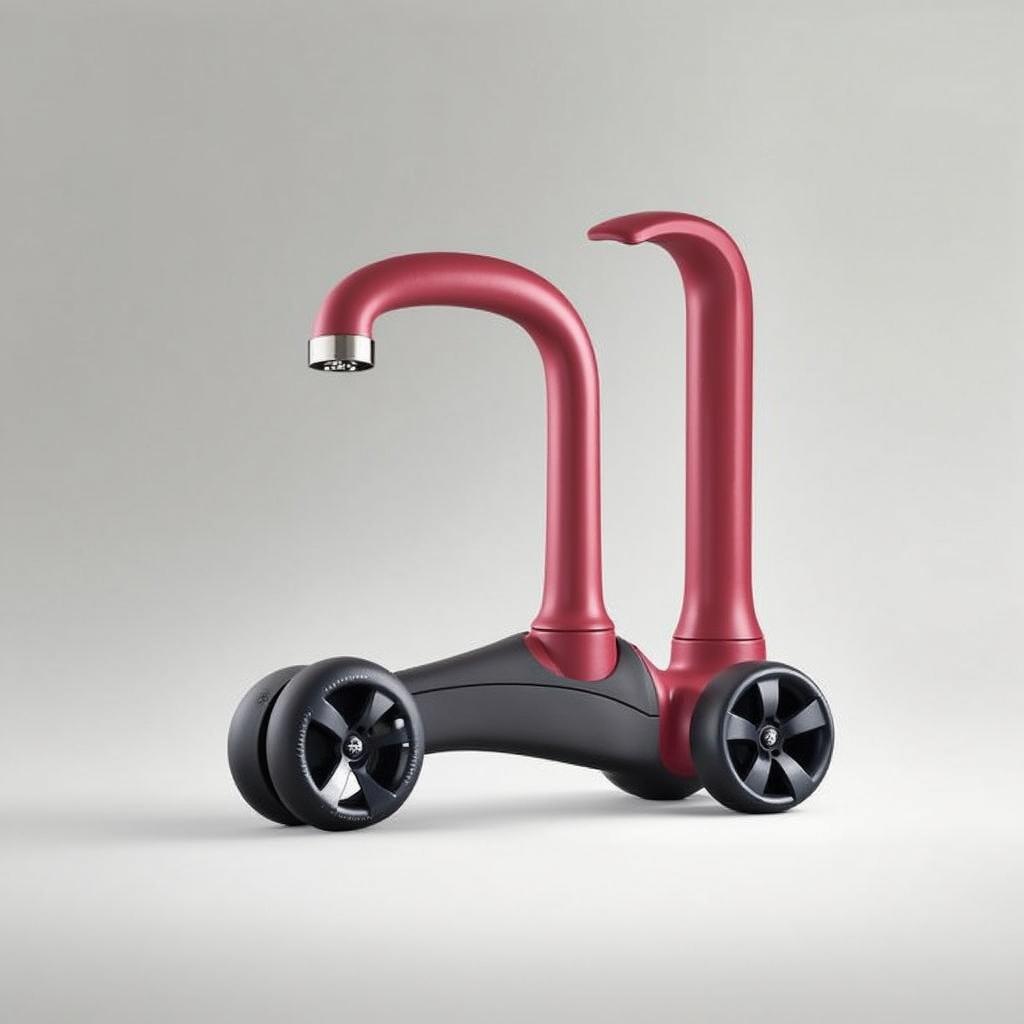} &
        \includegraphics[height=0.1\textheight]{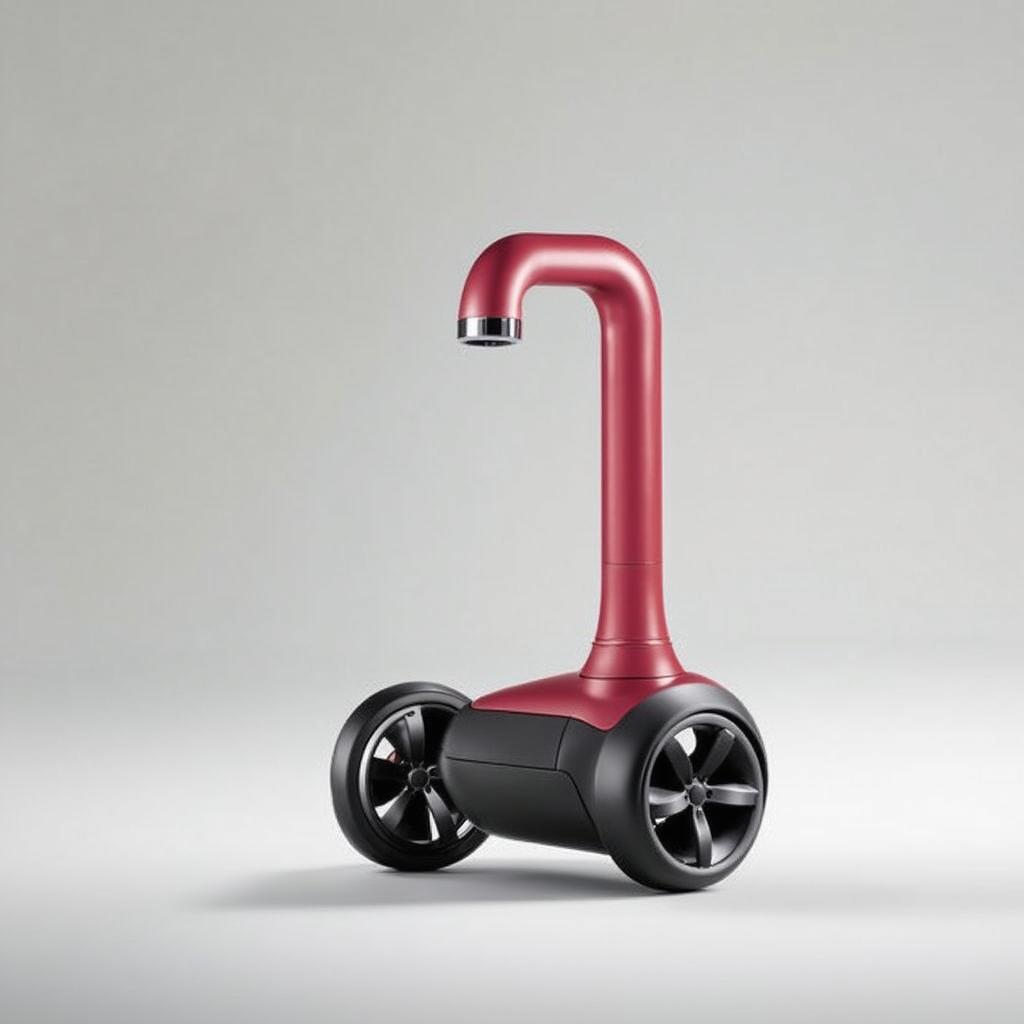} &

        \includegraphics[height=0.1\textheight]{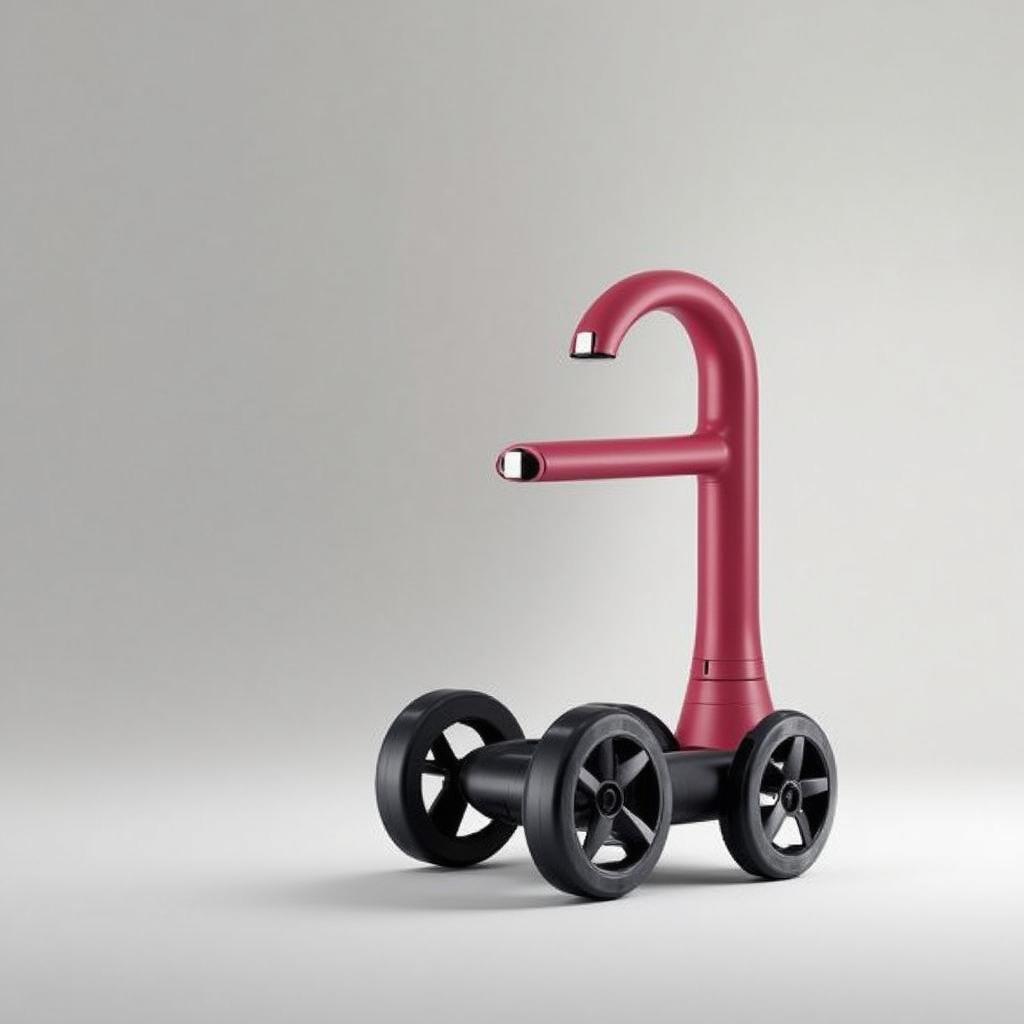} 
        \\

        \includegraphics[height=0.1\textheight]{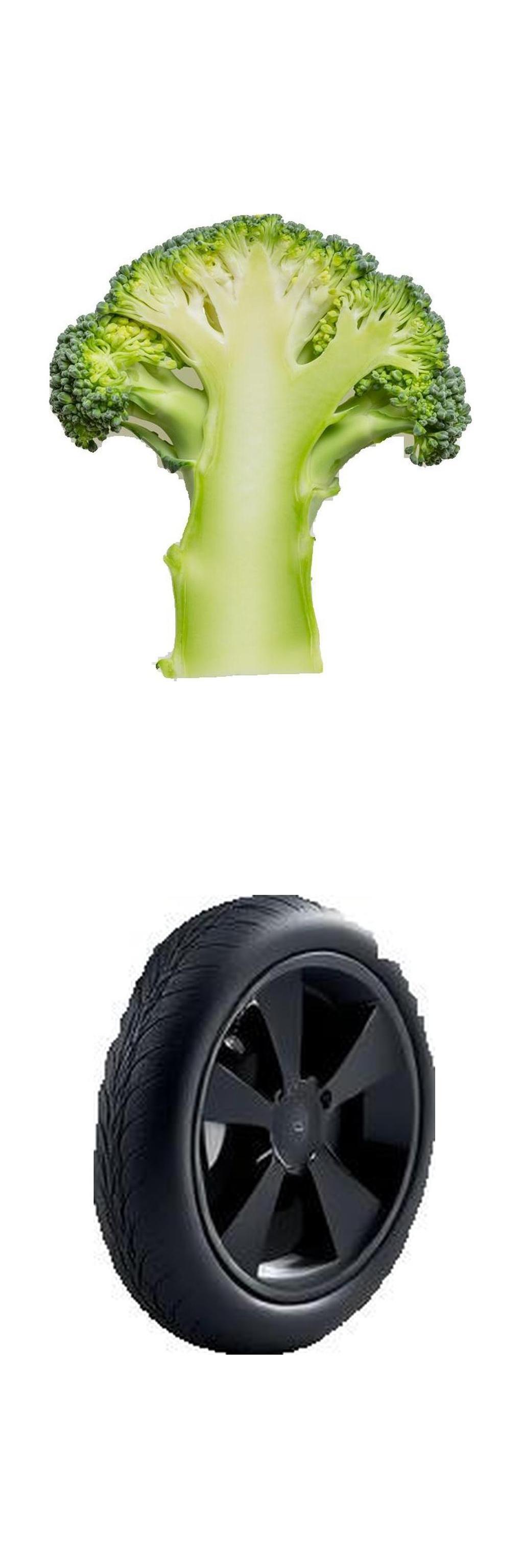} &
        \includegraphics[height=0.1\textheight]{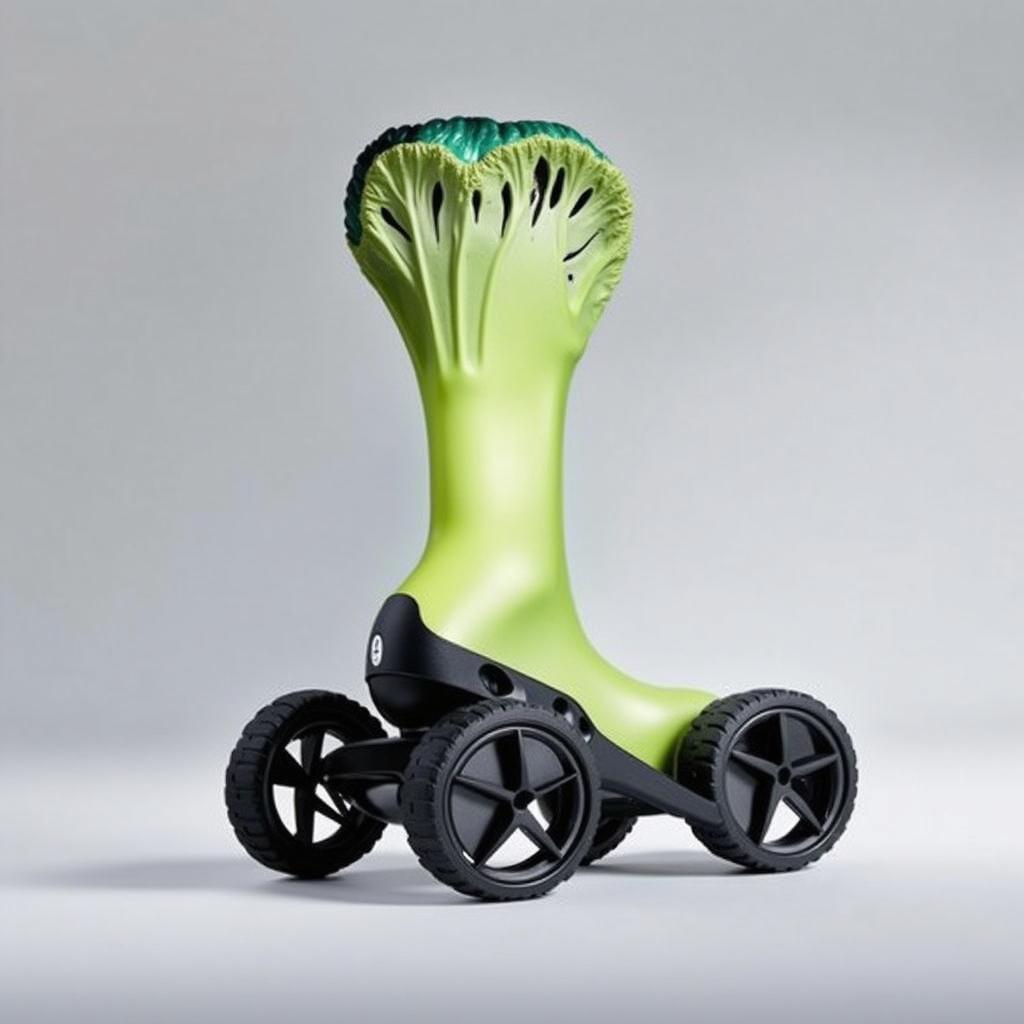} &
        \includegraphics[height=0.1\textheight]{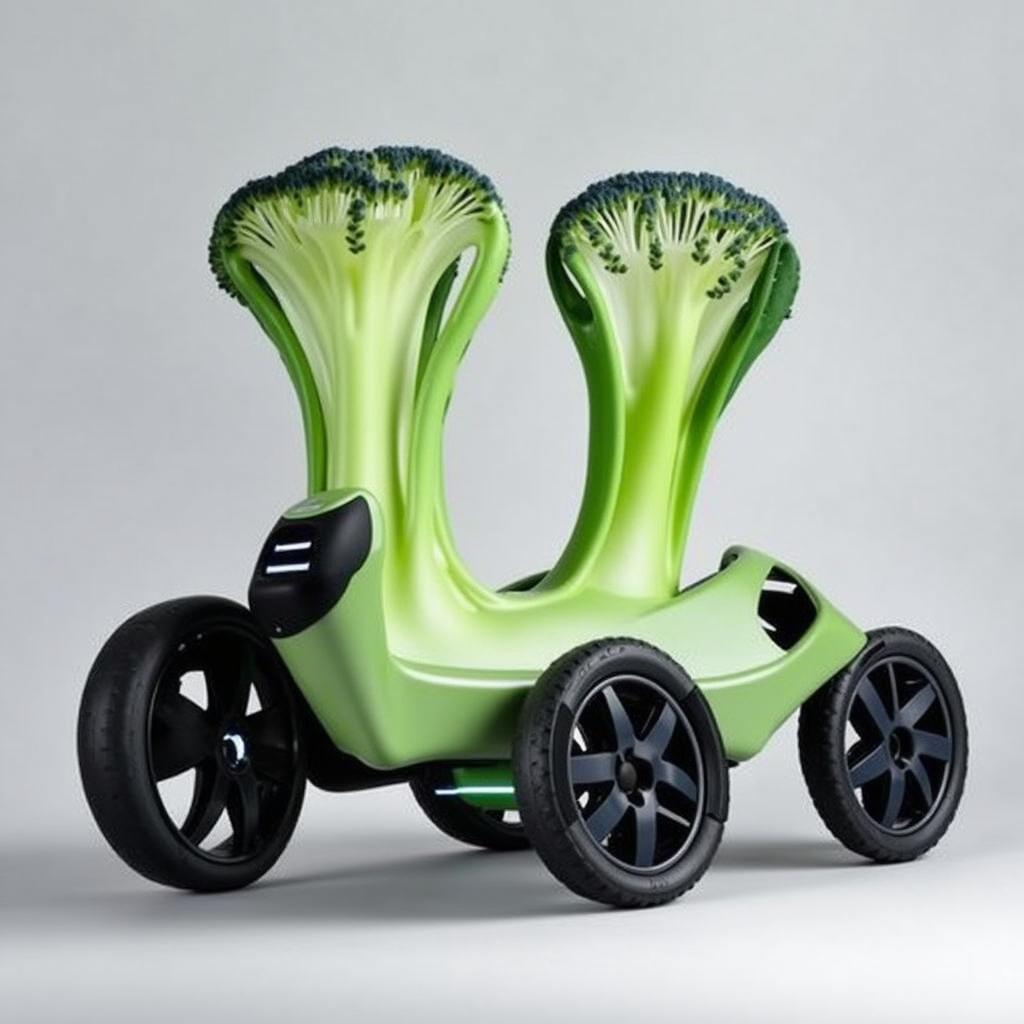} &

        \includegraphics[height=0.1\textheight]{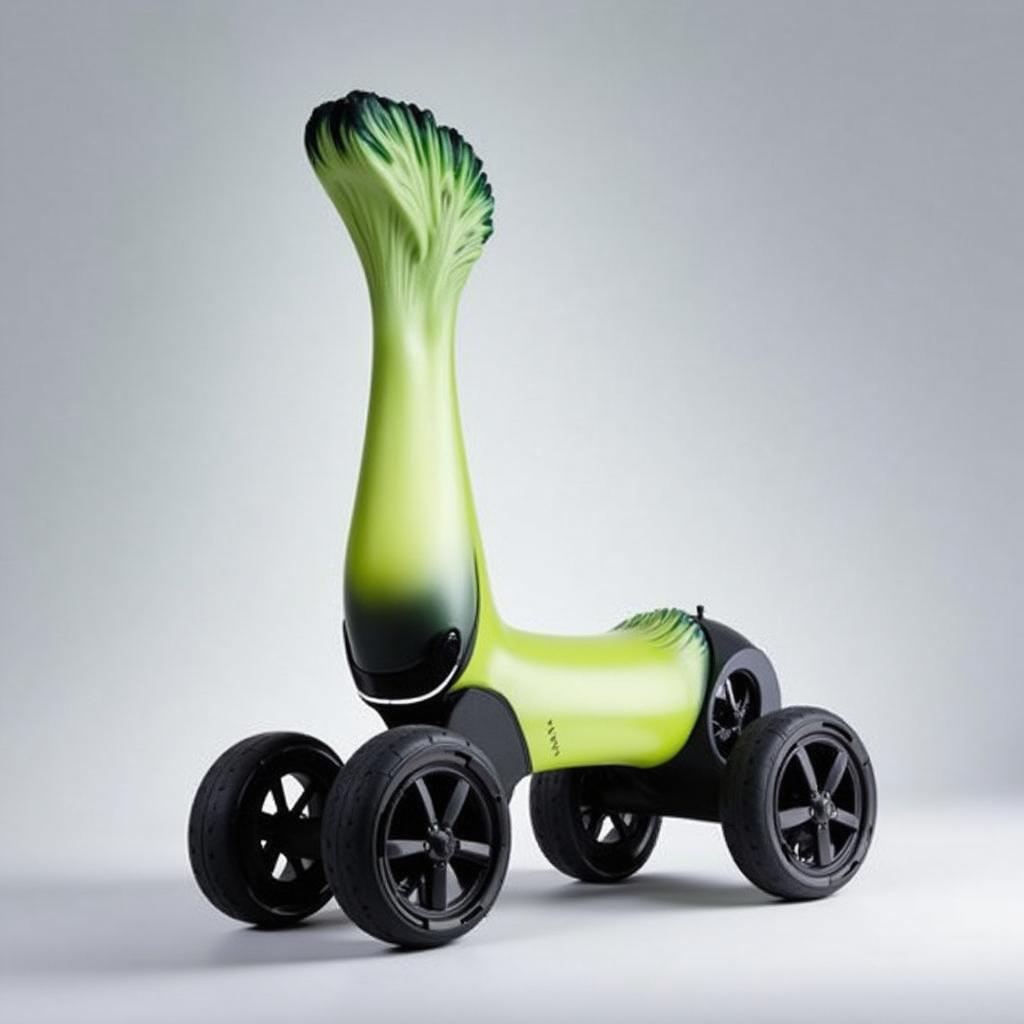} 
        \\

        \includegraphics[height=0.1\textheight]{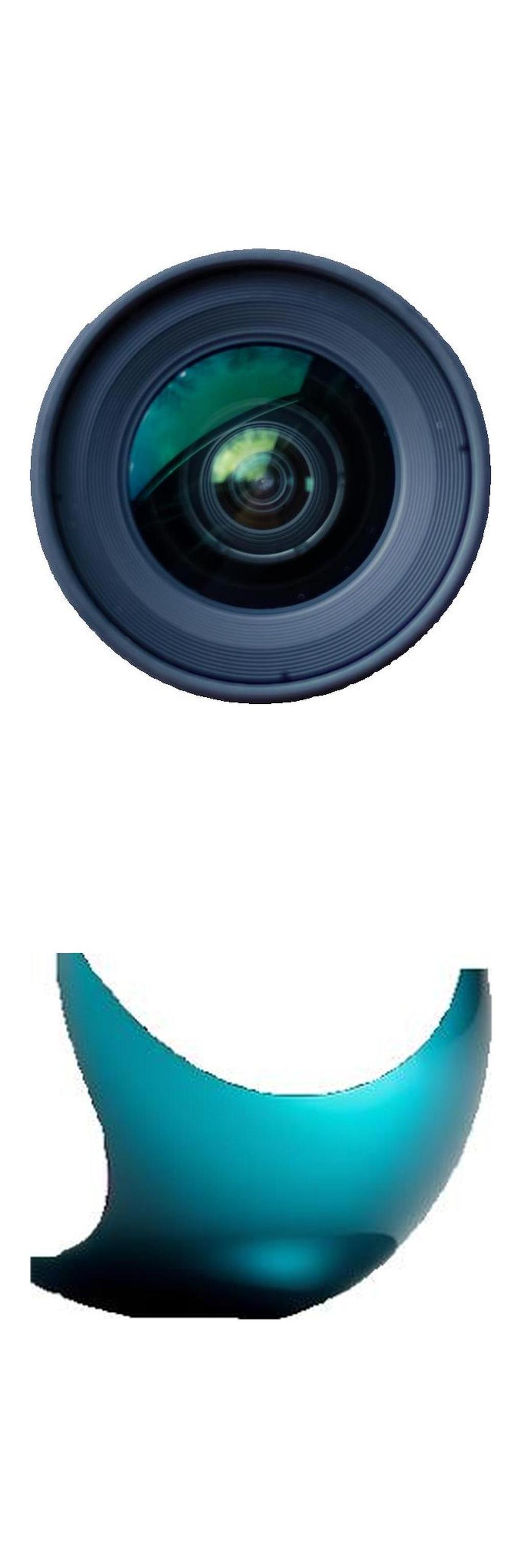} &
        \includegraphics[height=0.1\textheight]{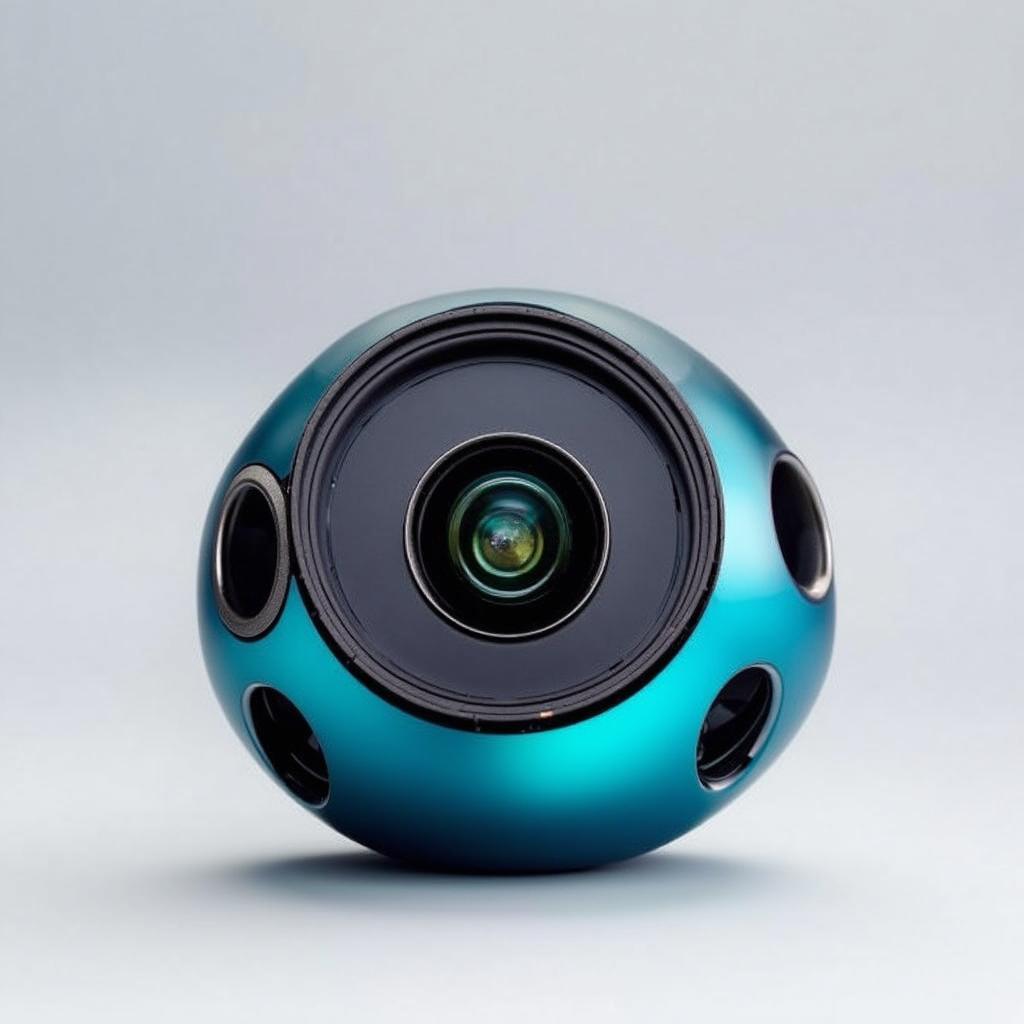} &
        \includegraphics[height=0.1\textheight]{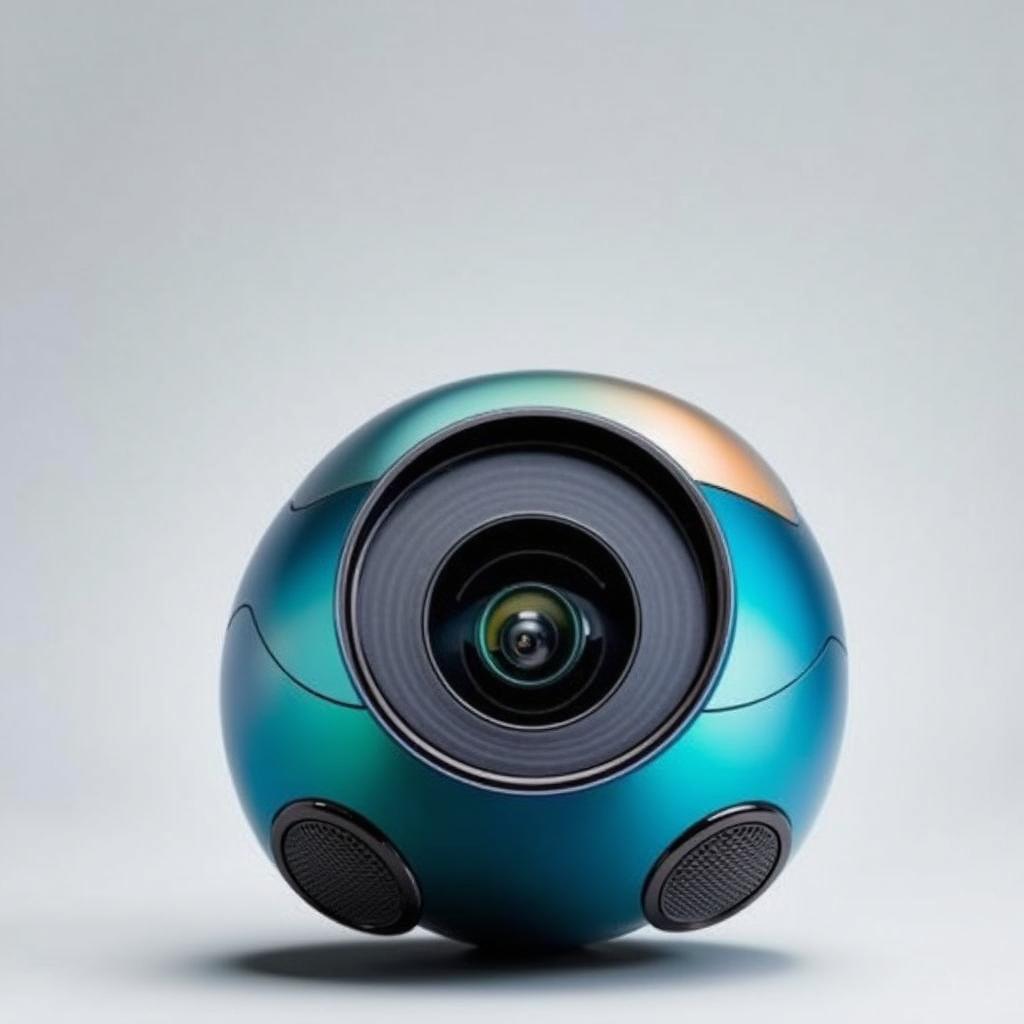} &

        \includegraphics[height=0.1\textheight]{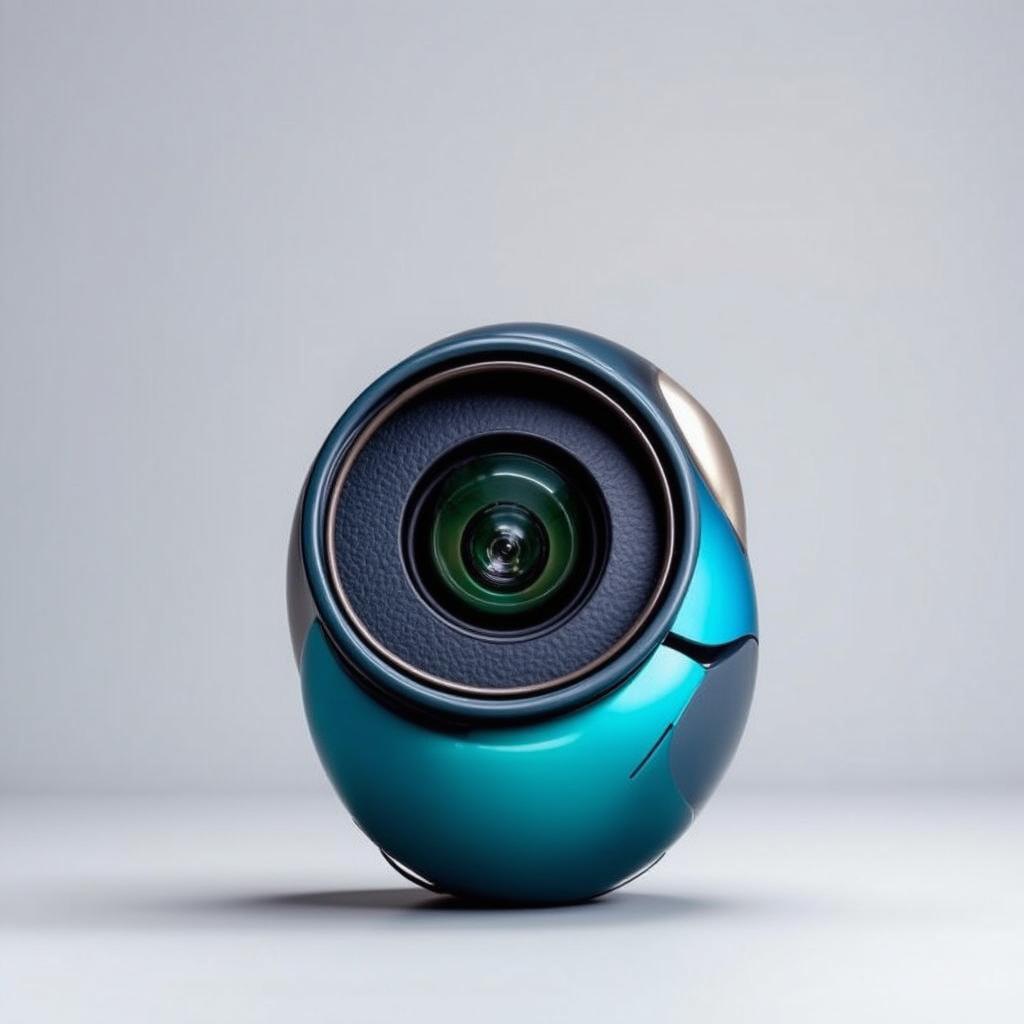} 
        \\

        \includegraphics[height=0.1\textheight]{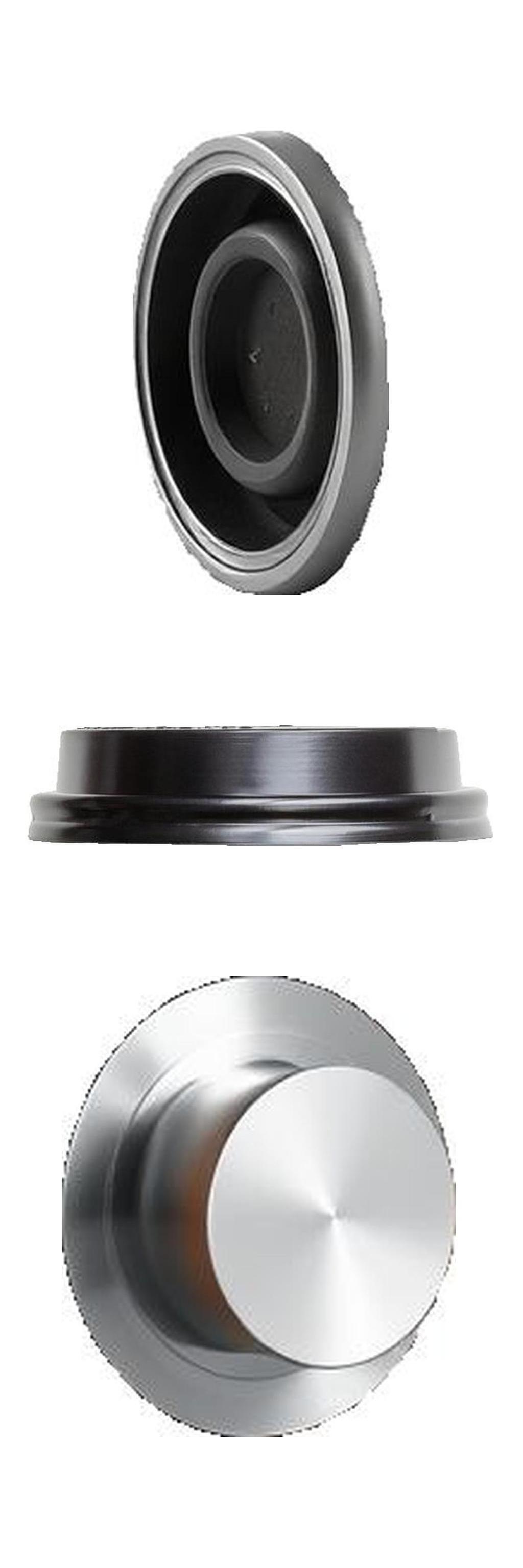} &
        \includegraphics[height=0.1\textheight]{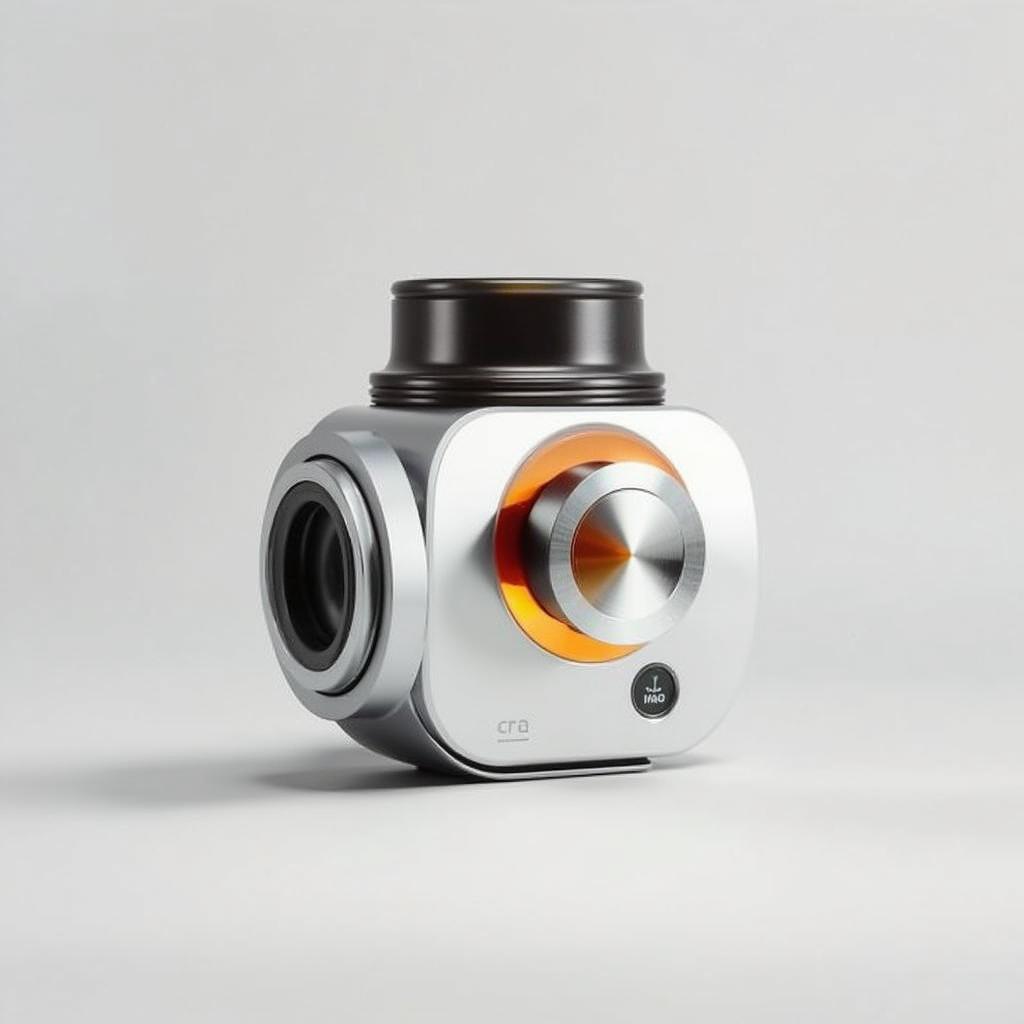} &
        \includegraphics[height=0.1\textheight]{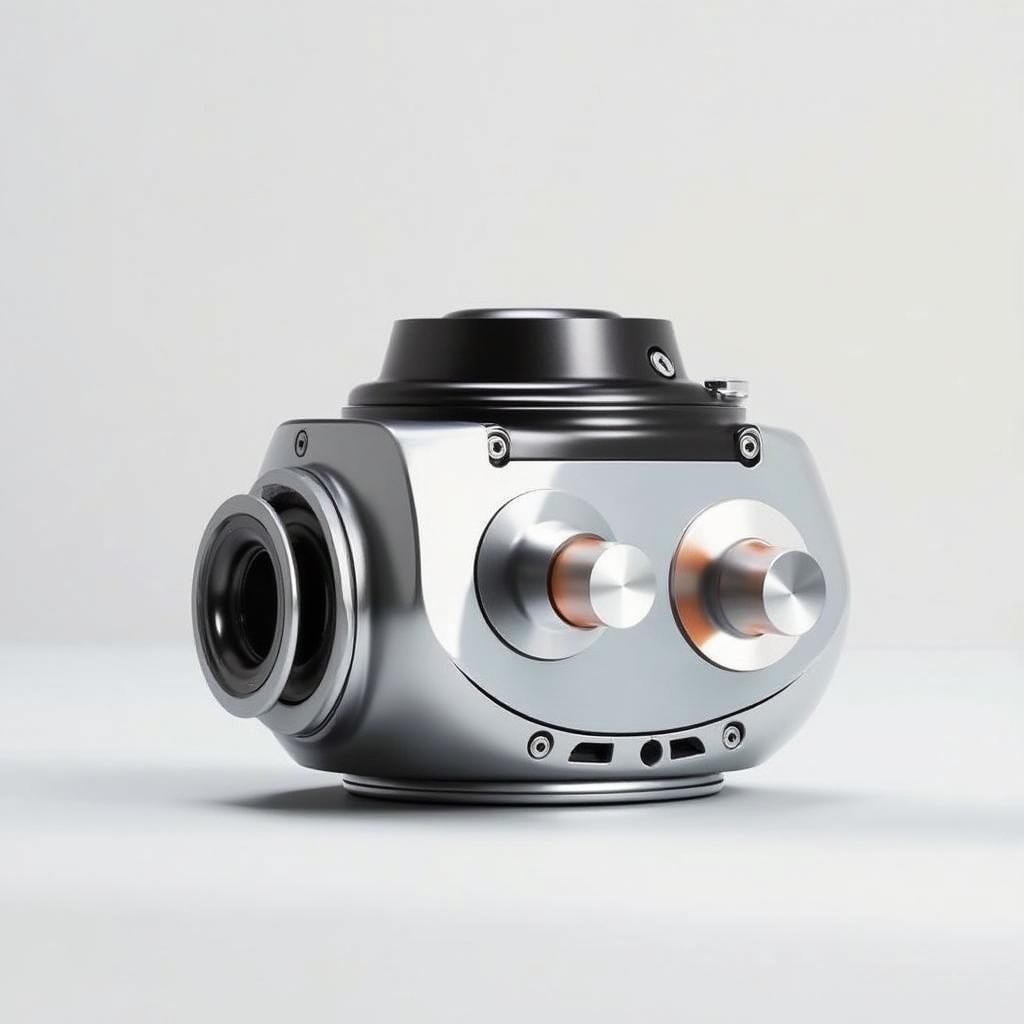} &

        \includegraphics[height=0.1\textheight]{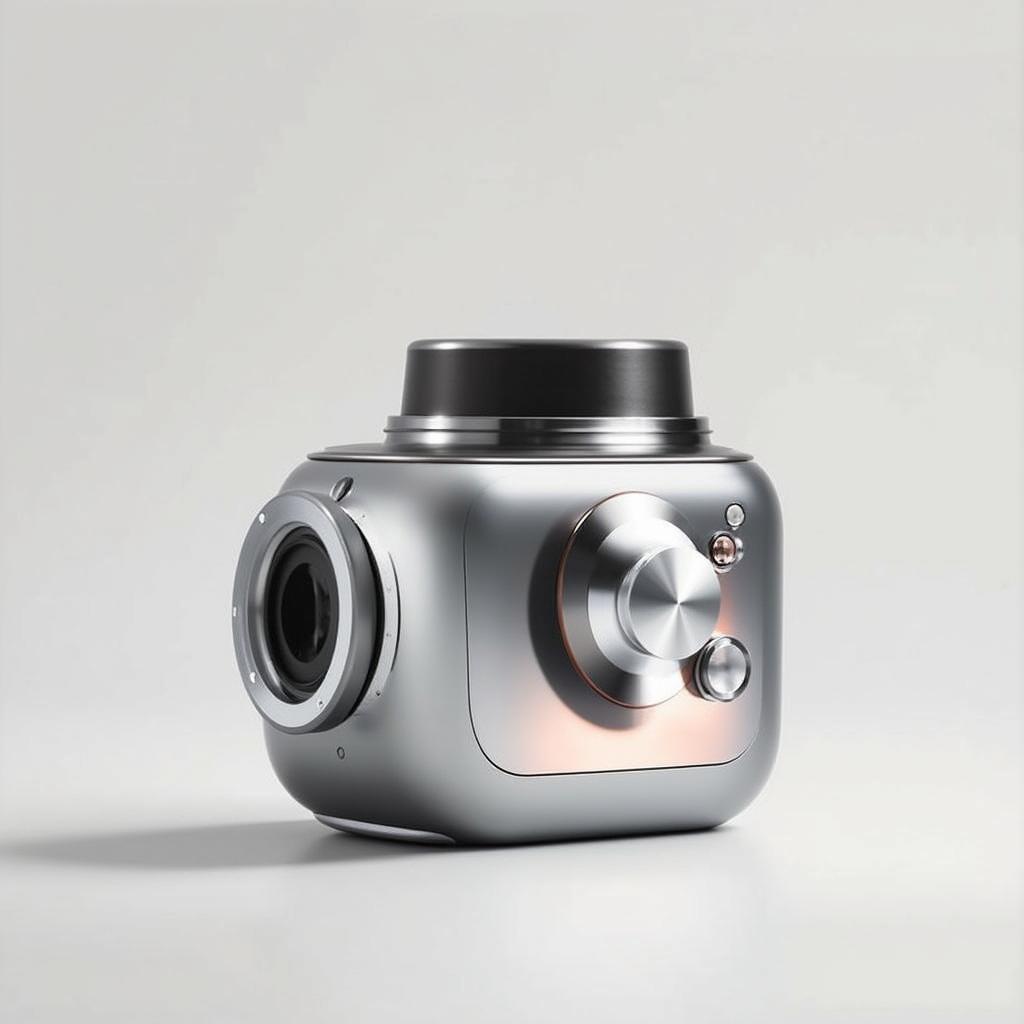} 
        \\

        \includegraphics[height=0.1\textheight]{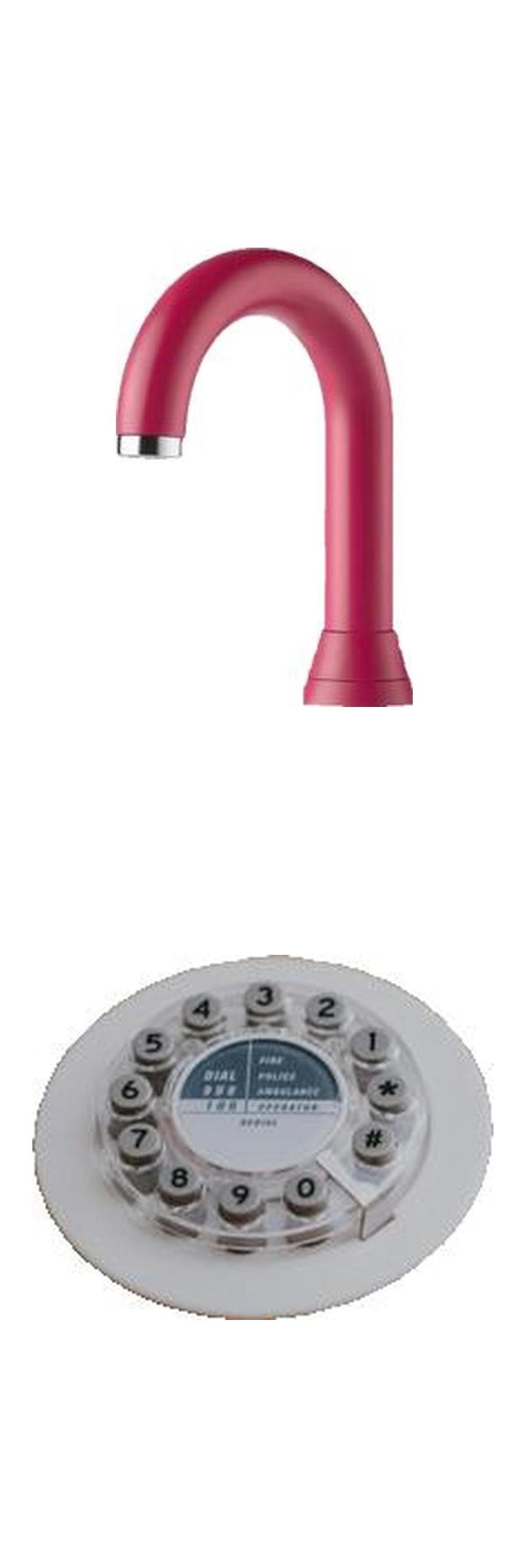} &
        \includegraphics[height=0.1\textheight]{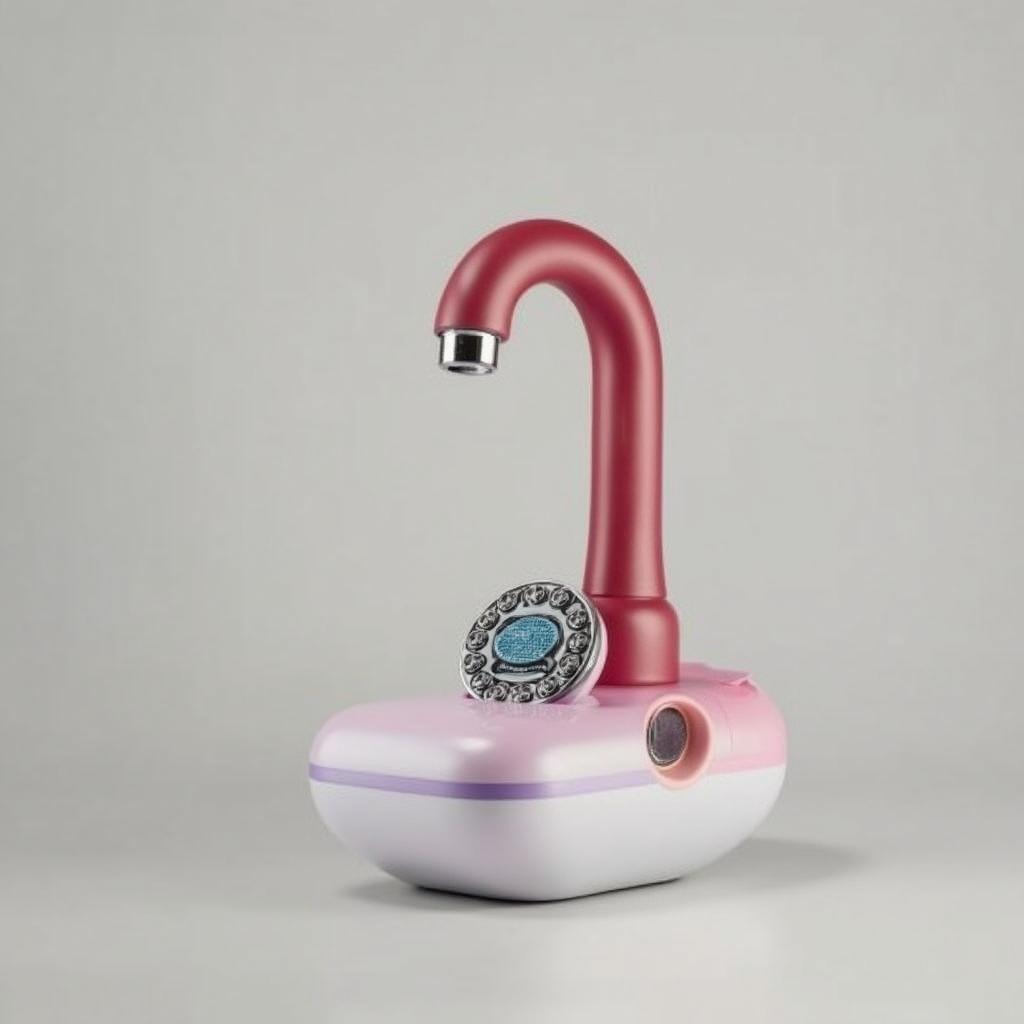} &
        \includegraphics[height=0.1\textheight]{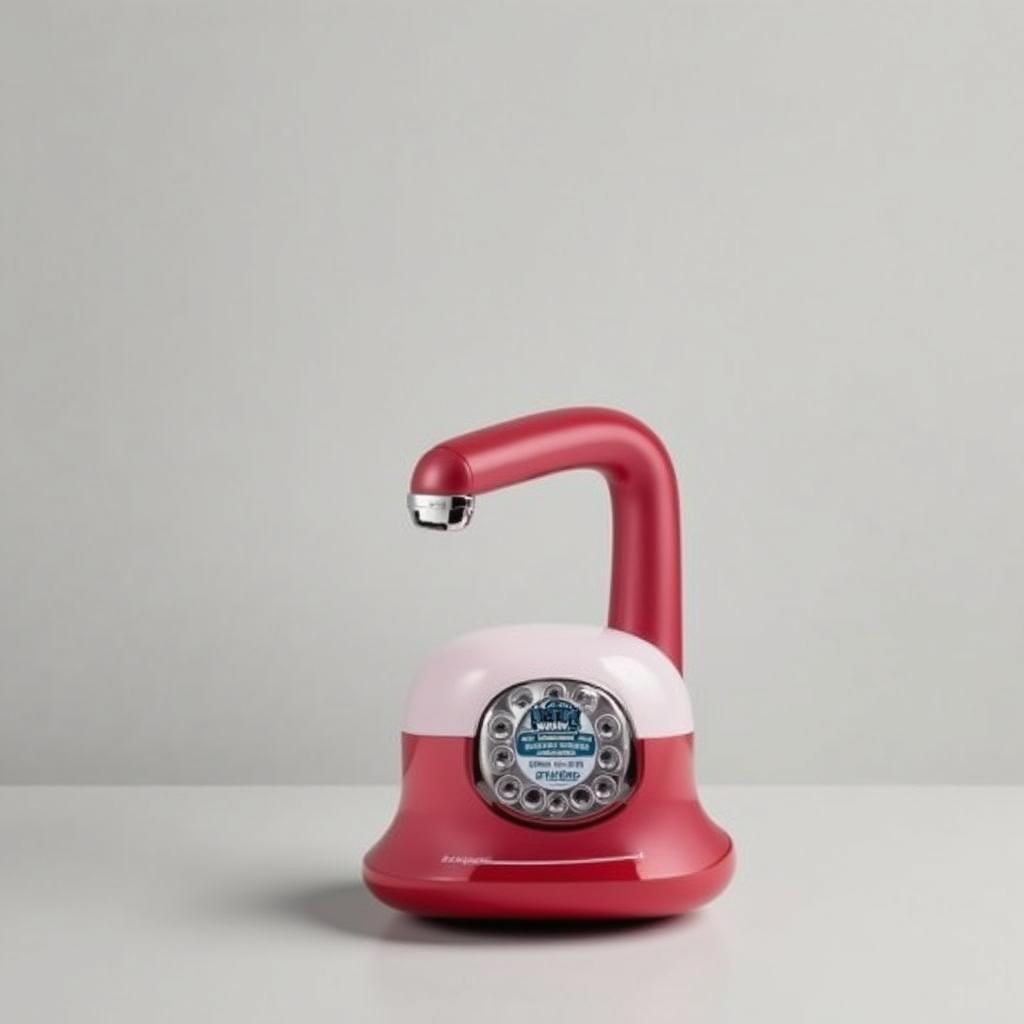} &

        \includegraphics[height=0.1\textheight]{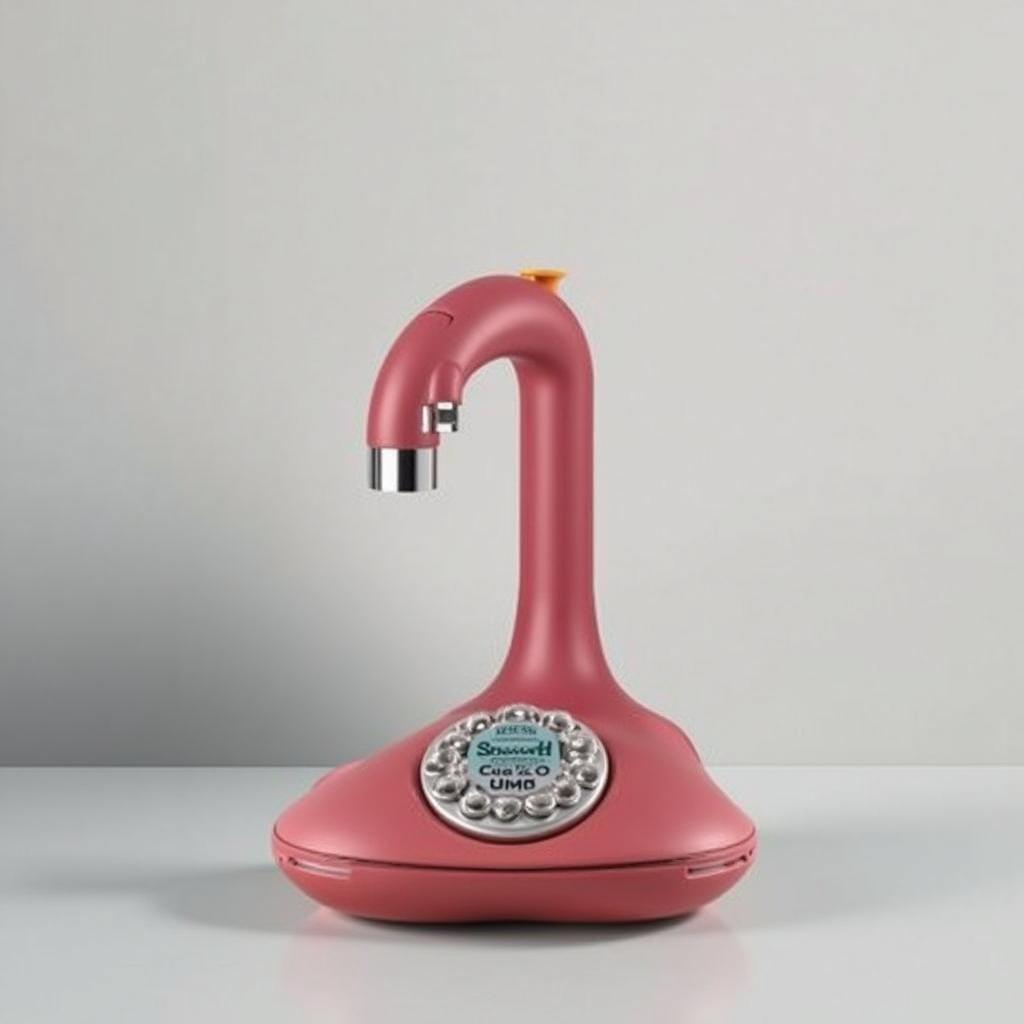} 
        \\

        \includegraphics[height=0.1\textheight]{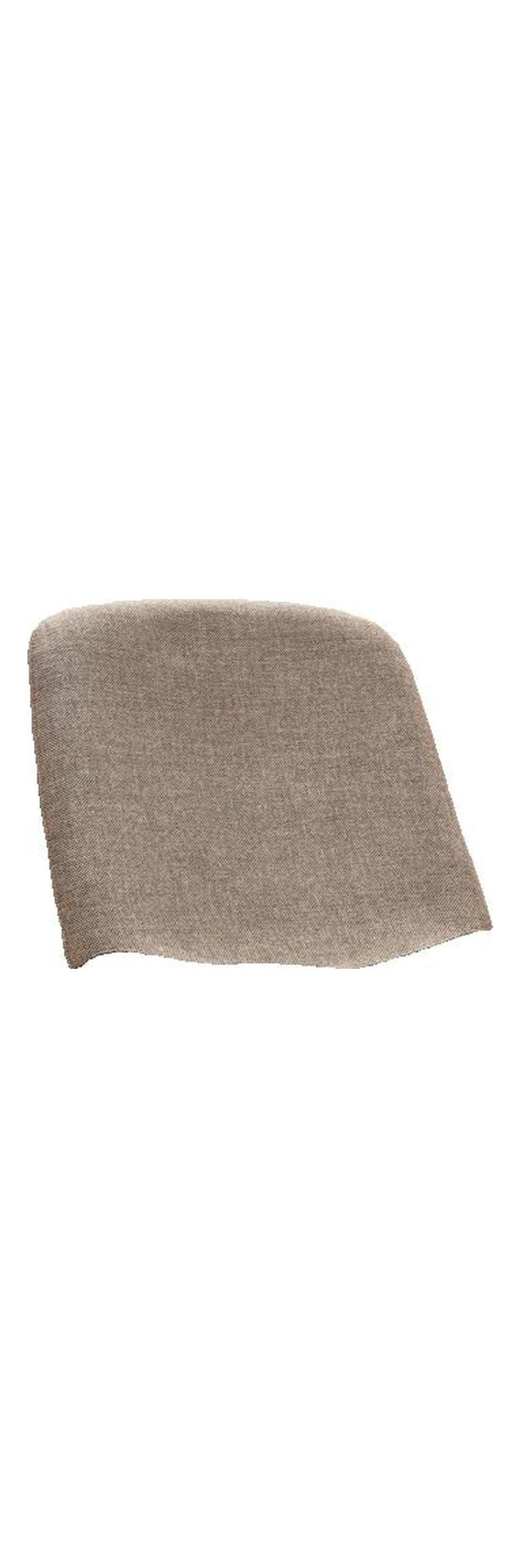} &
        \includegraphics[height=0.1\textheight]{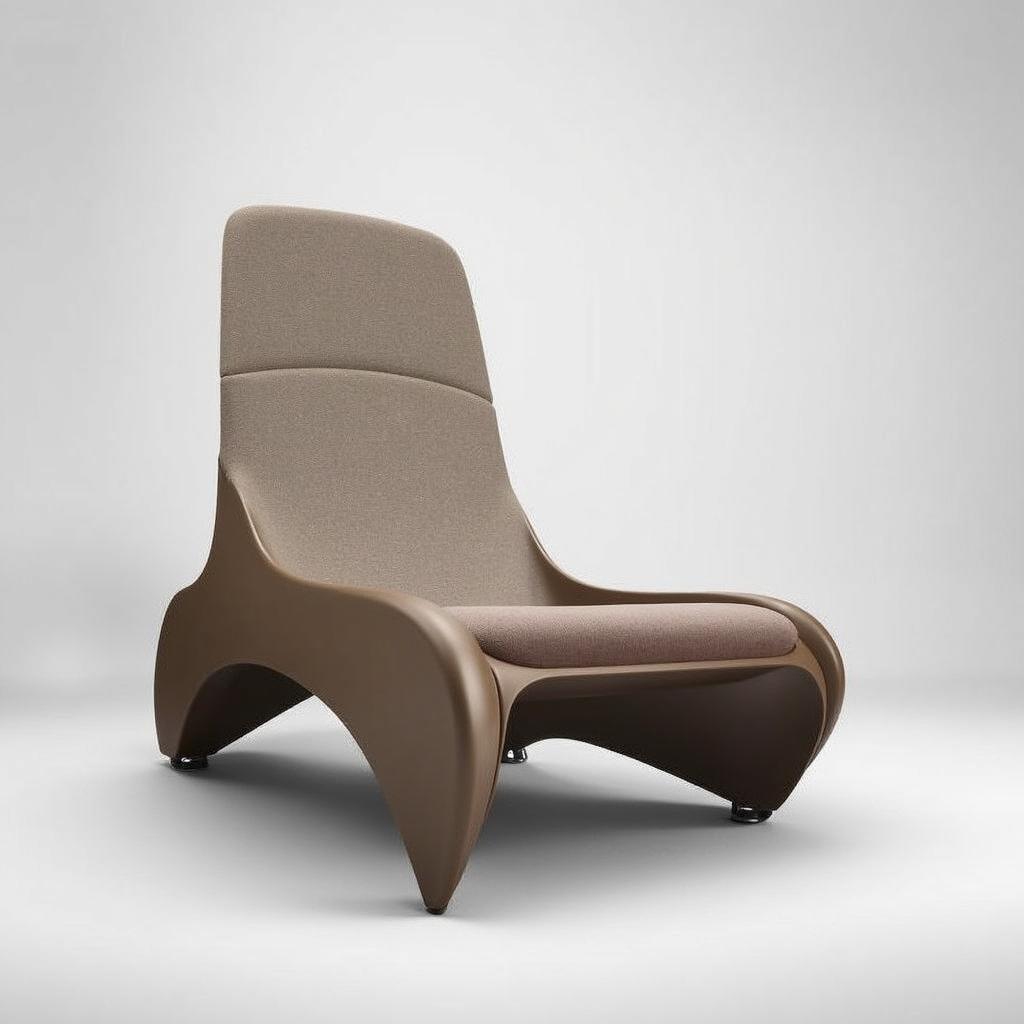} &
        \includegraphics[height=0.1\textheight]{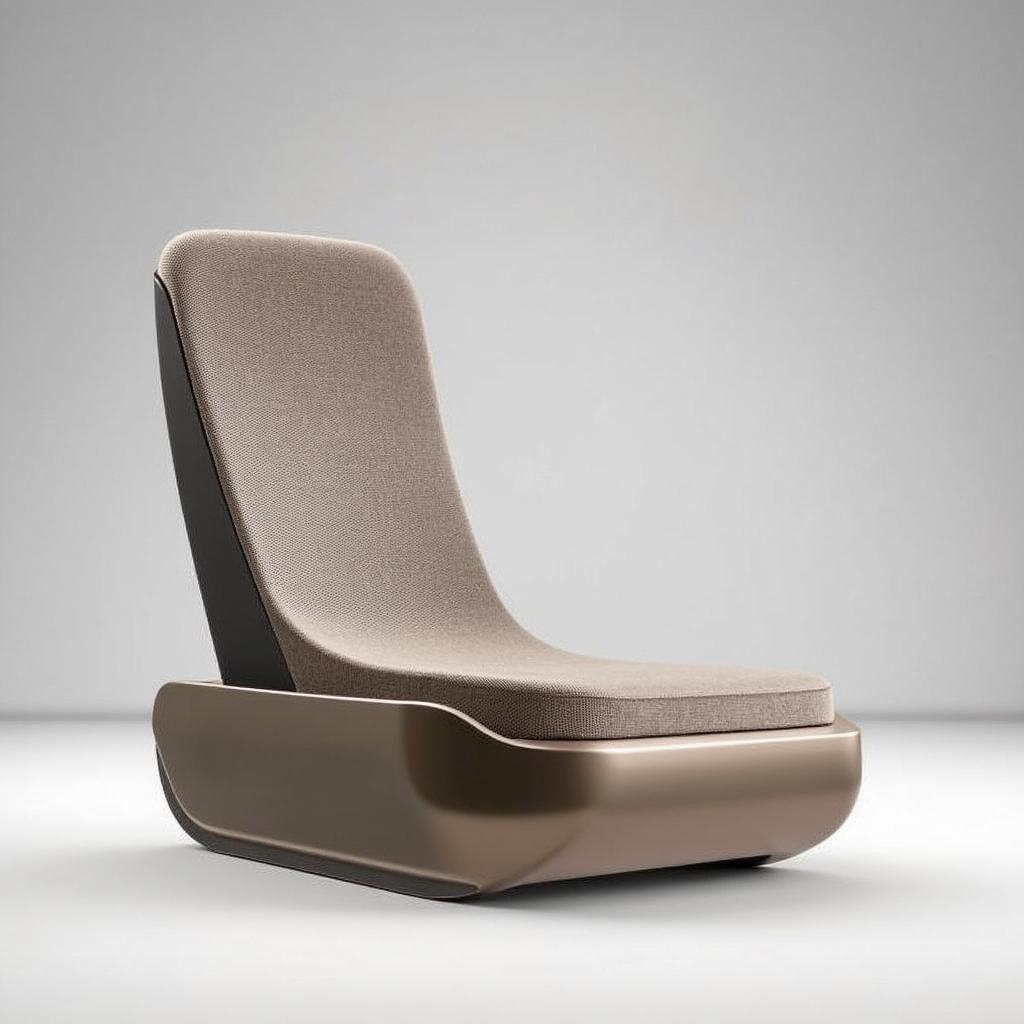} &

        \includegraphics[height=0.1\textheight]{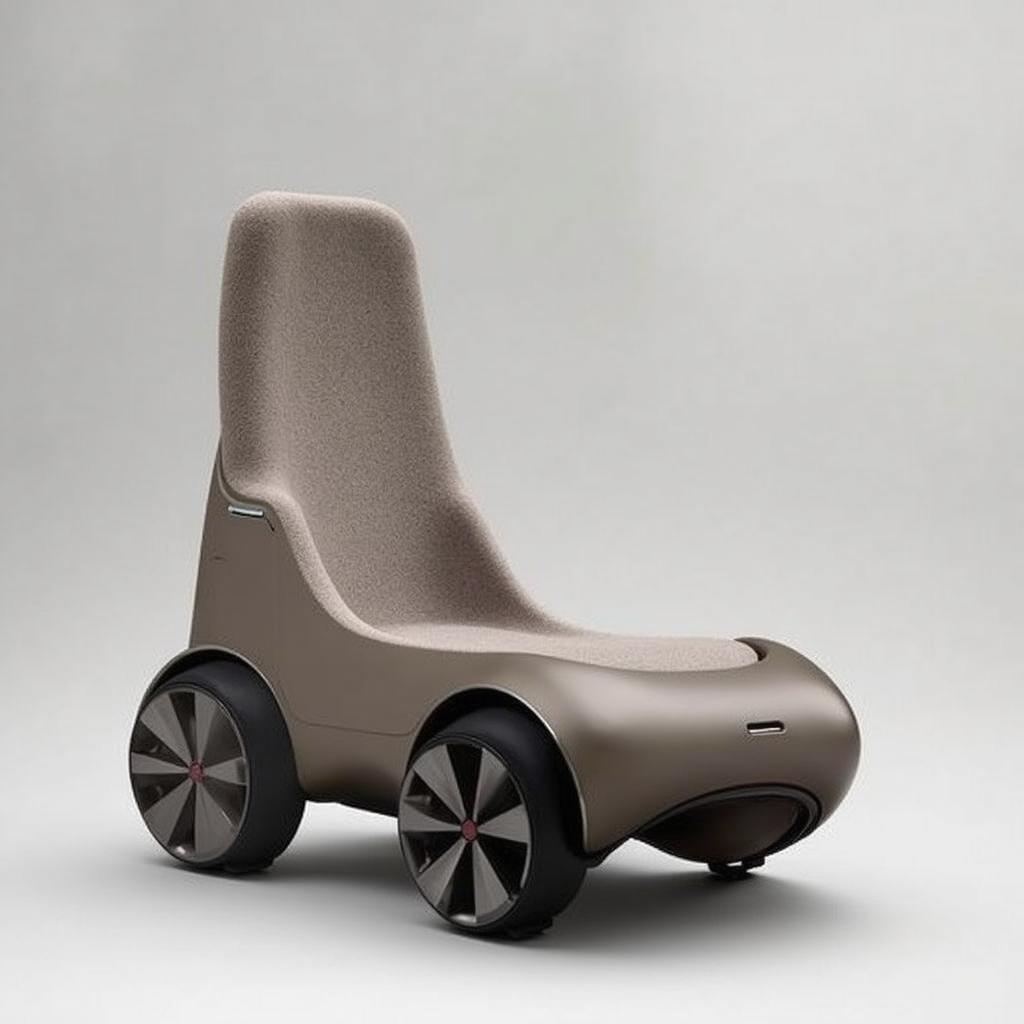} 
        \\ \\[-0.1cm]

        Input & \multicolumn{3}{c}{Sampled Results} 

    \end{tabular}
    }
    \caption{Additional PiT results for product design
    }
    \label{fig:supp_products_1}
\end{figure}

\begin{figure}
    \centering
    \setlength{\tabcolsep}{0.5pt}
    \renewcommand{\arraystretch}{0.5}
    {
    \begin{tabular}{c @{\hspace{0.2cm}} c c c}

        \includegraphics[height=0.1\textheight]{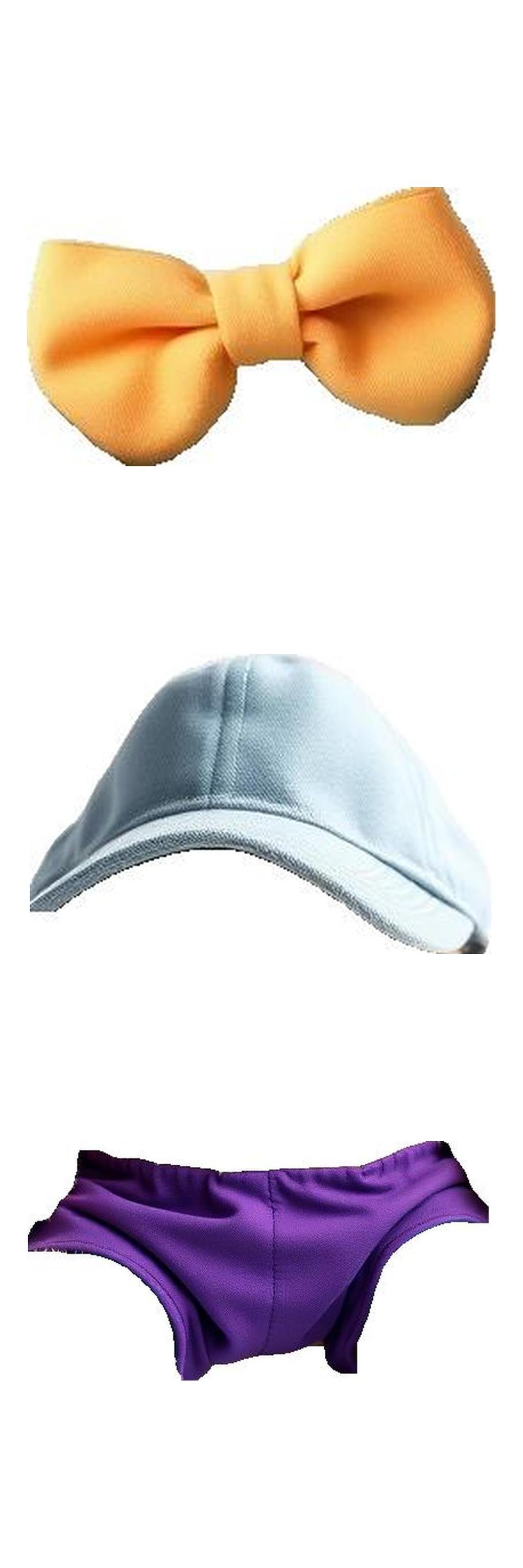} &
        \includegraphics[height=0.1\textheight]{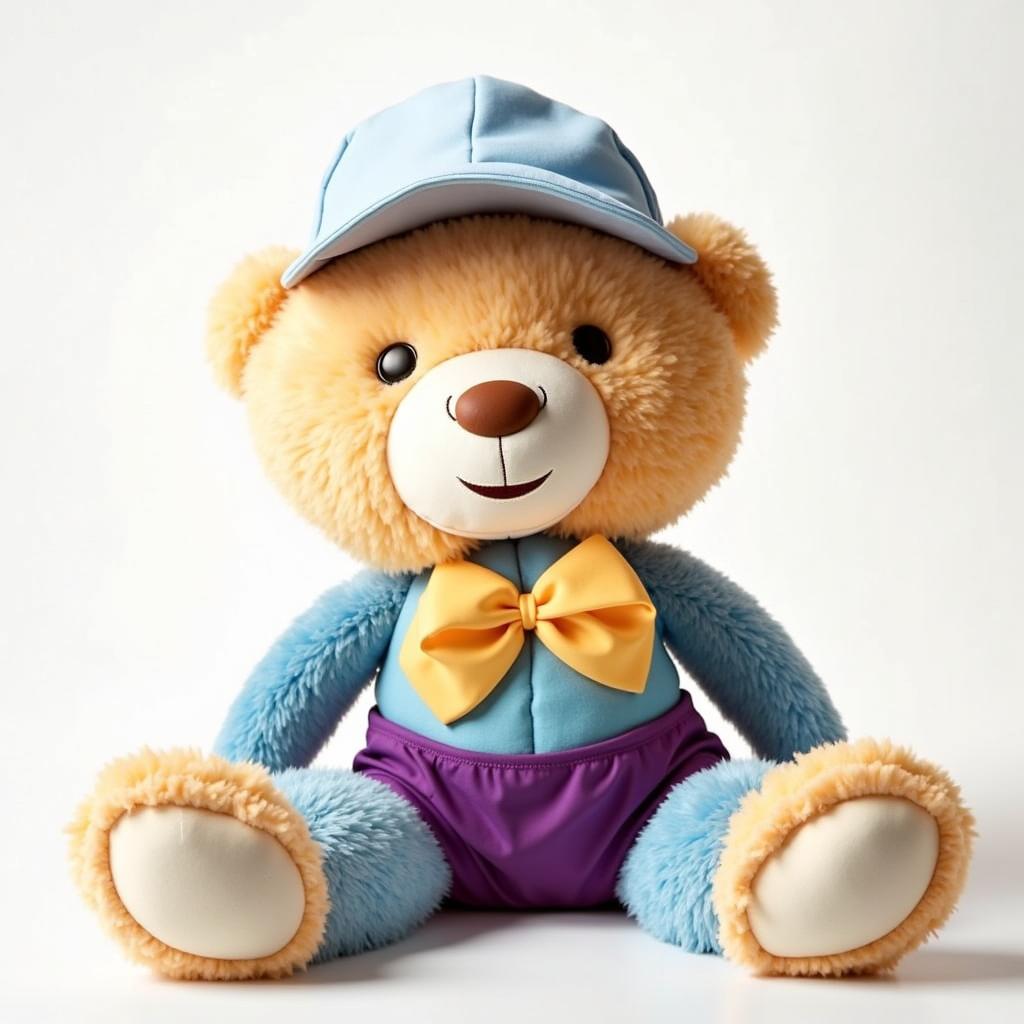} &
        \includegraphics[height=0.1\textheight]{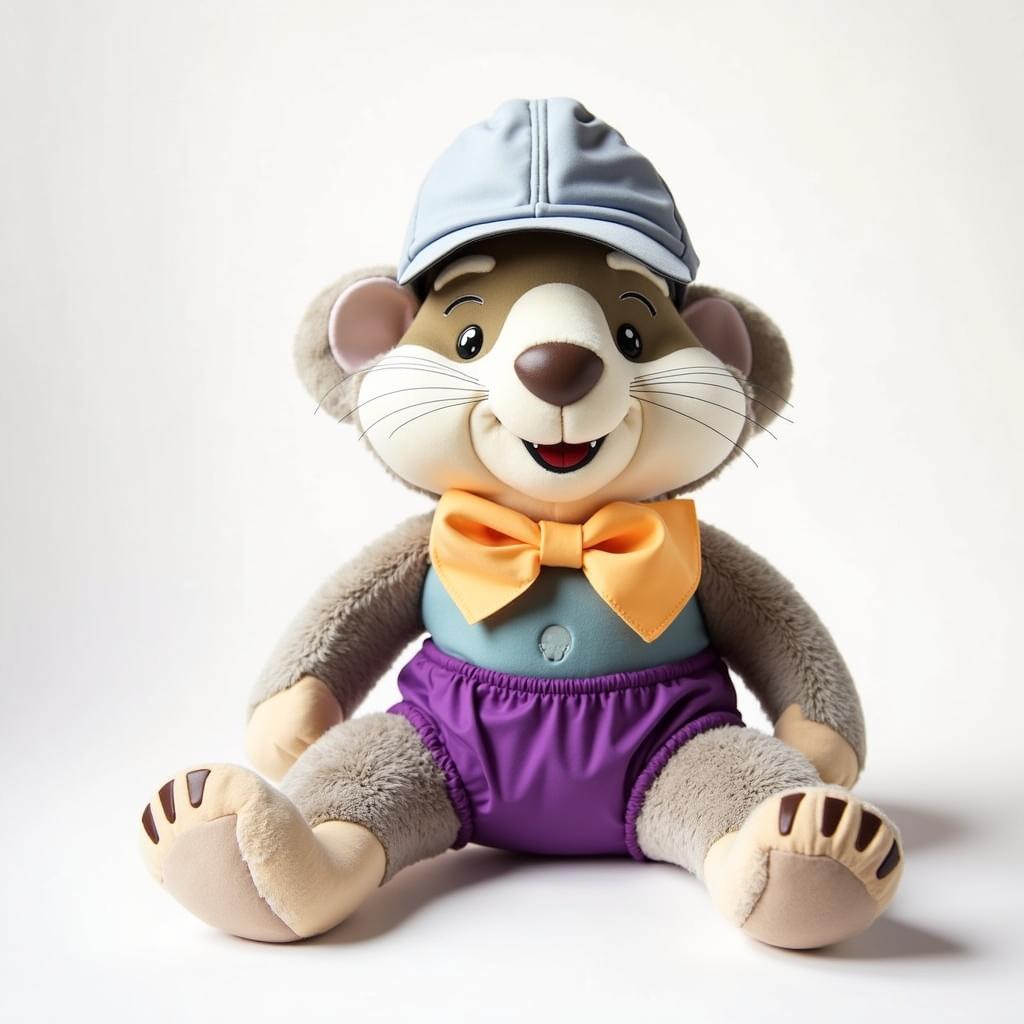} &

        \includegraphics[height=0.1\textheight]{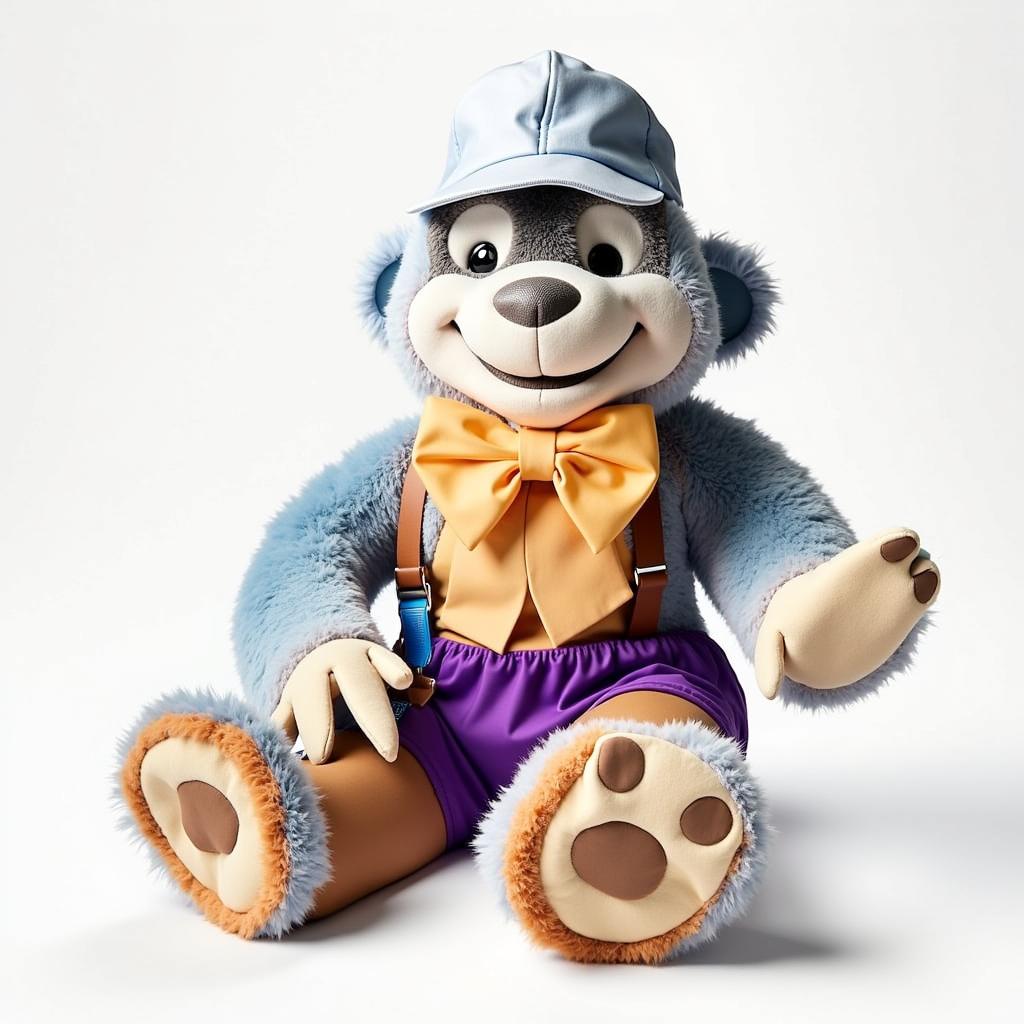} 
        \\

        \includegraphics[height=0.1\textheight]{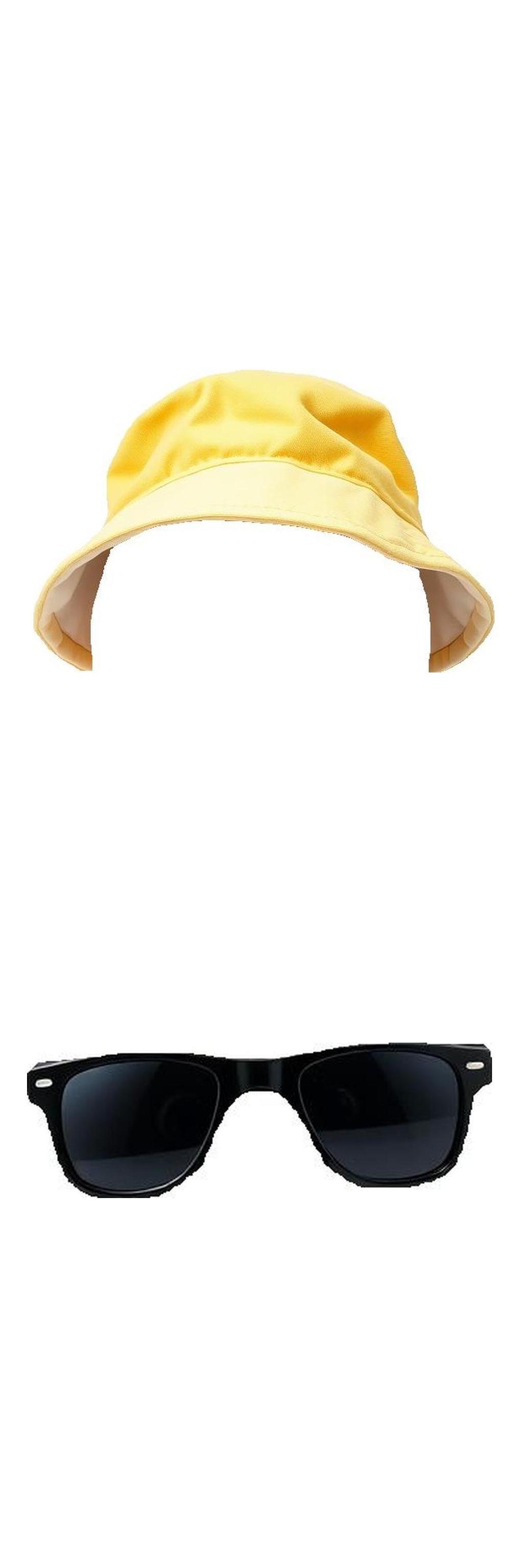} &
        \includegraphics[height=0.1\textheight]{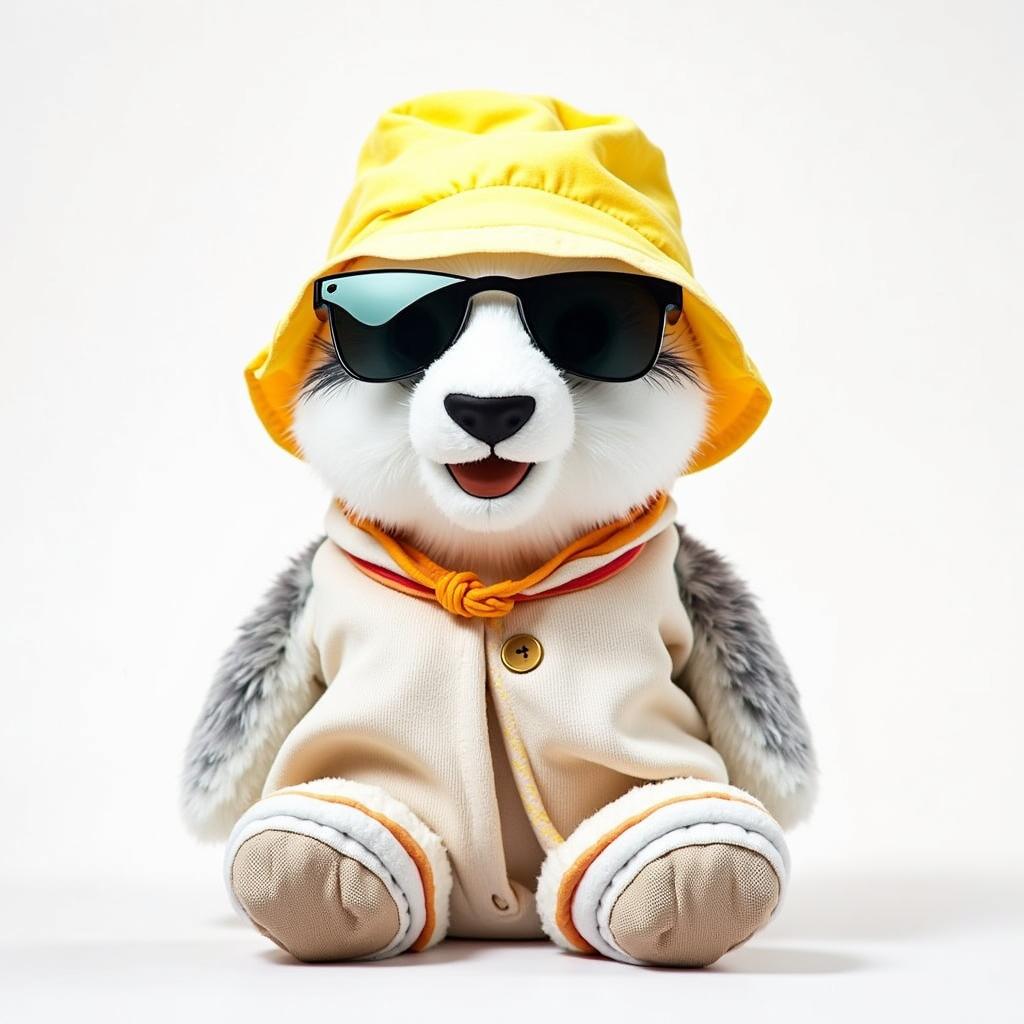} &
        \includegraphics[height=0.1\textheight]{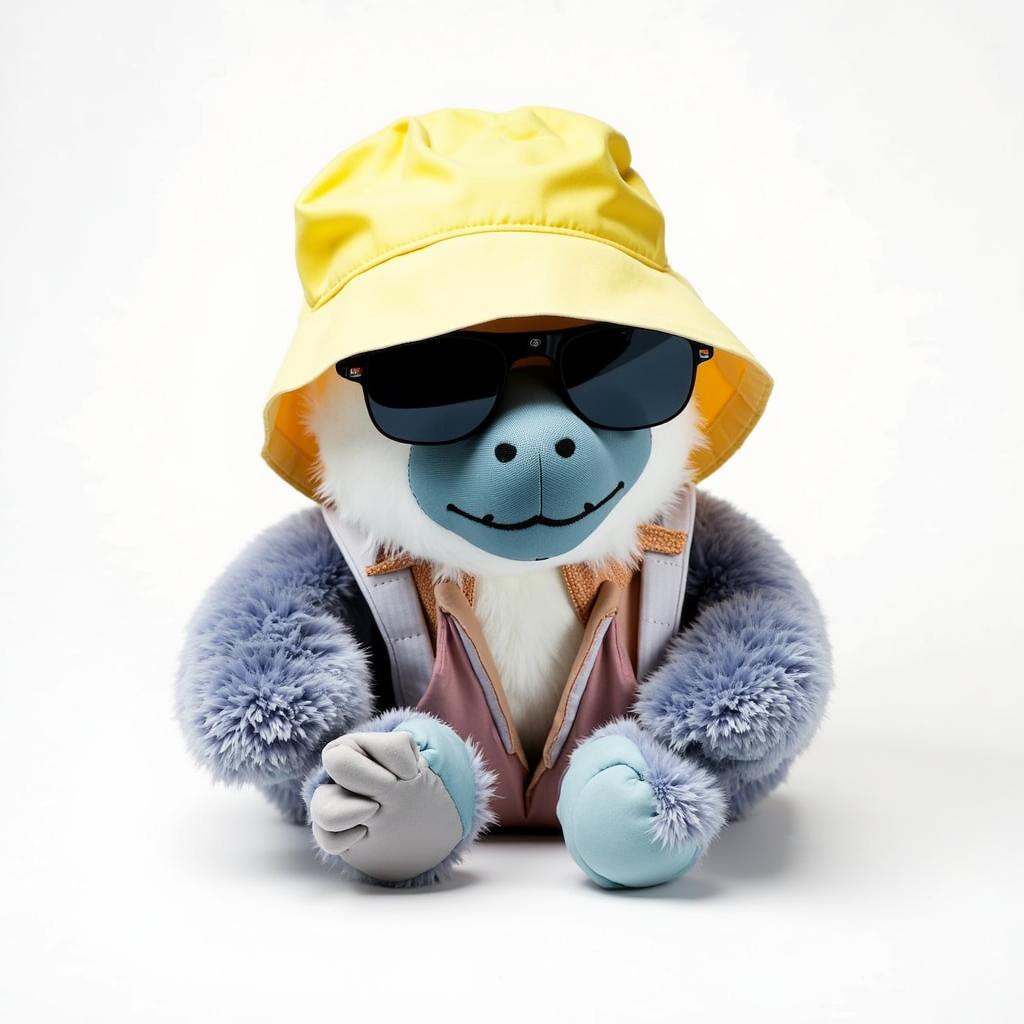} &

        \includegraphics[height=0.1\textheight]{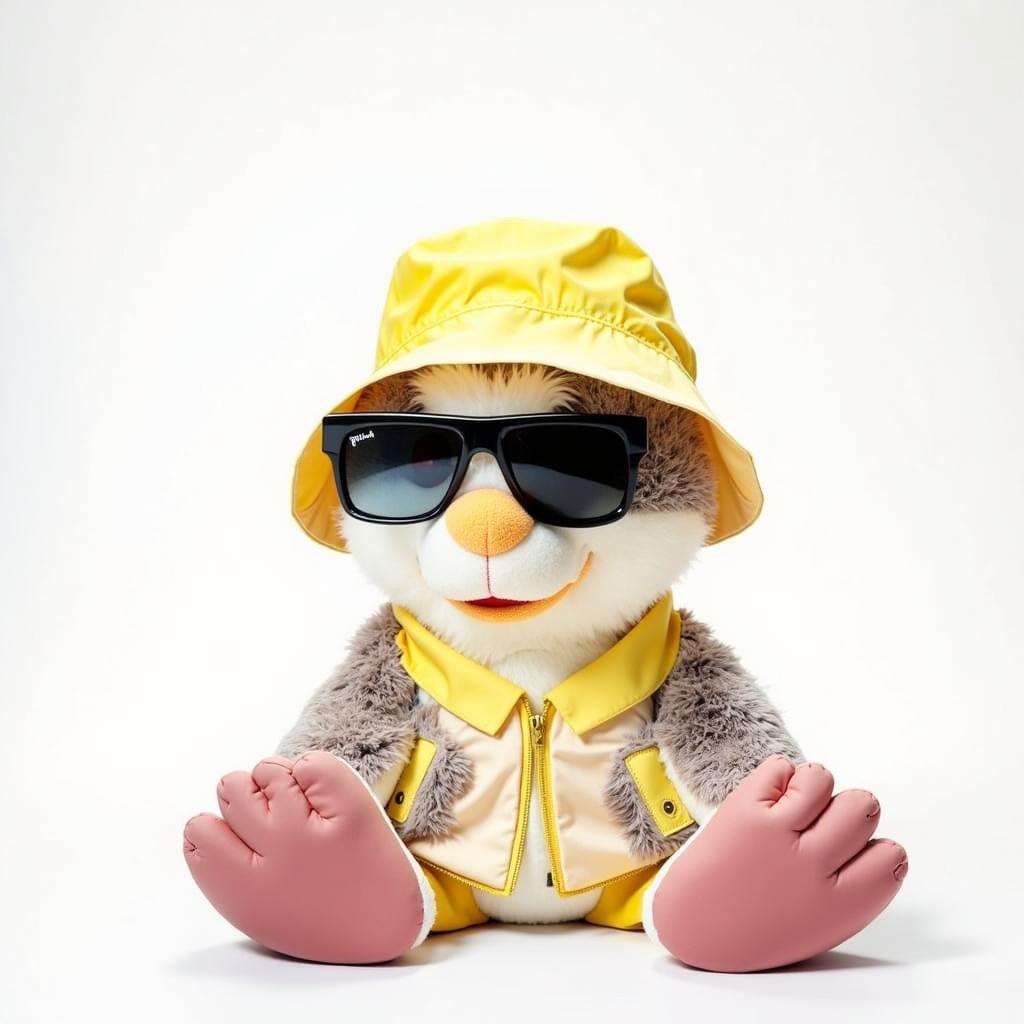} 
        \\

        \includegraphics[height=0.1\textheight]{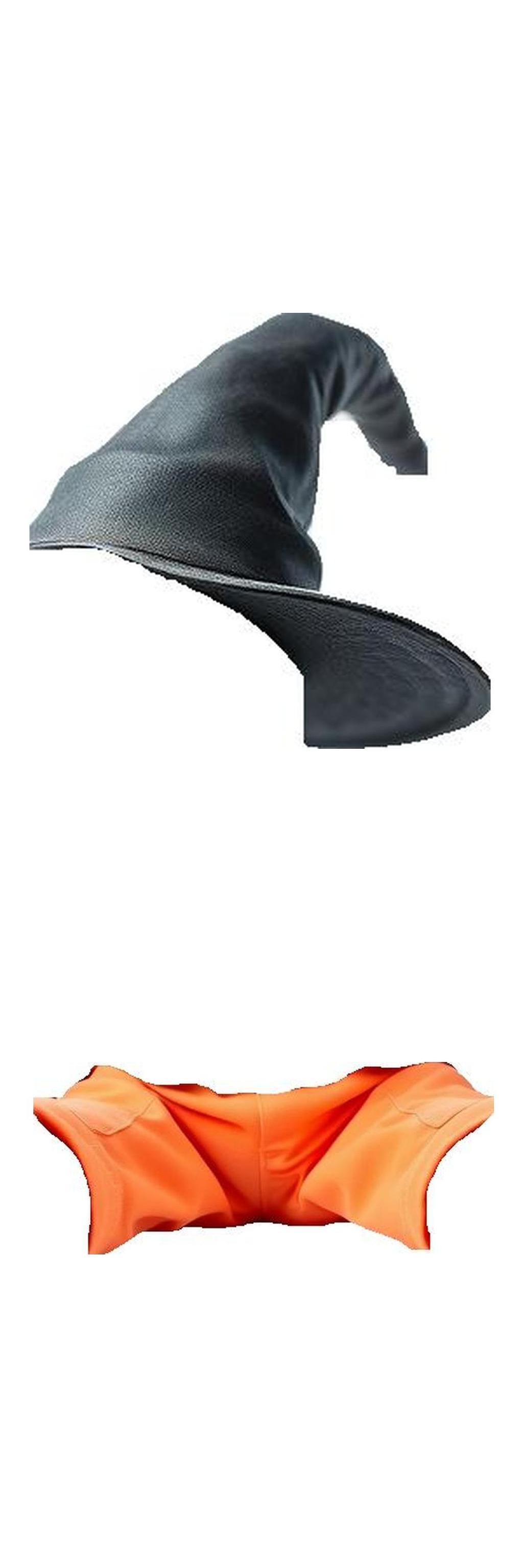} &
        \includegraphics[height=0.1\textheight]{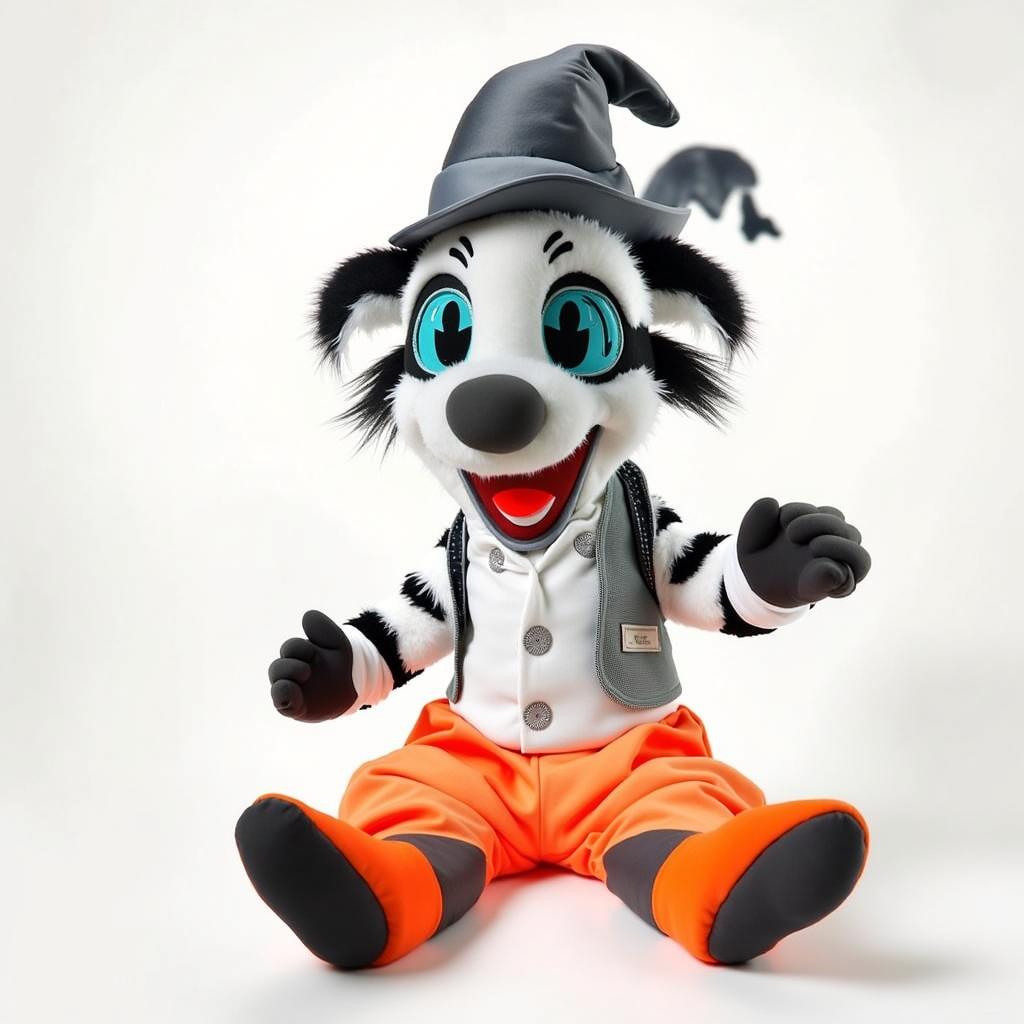} &
        \includegraphics[height=0.1\textheight]{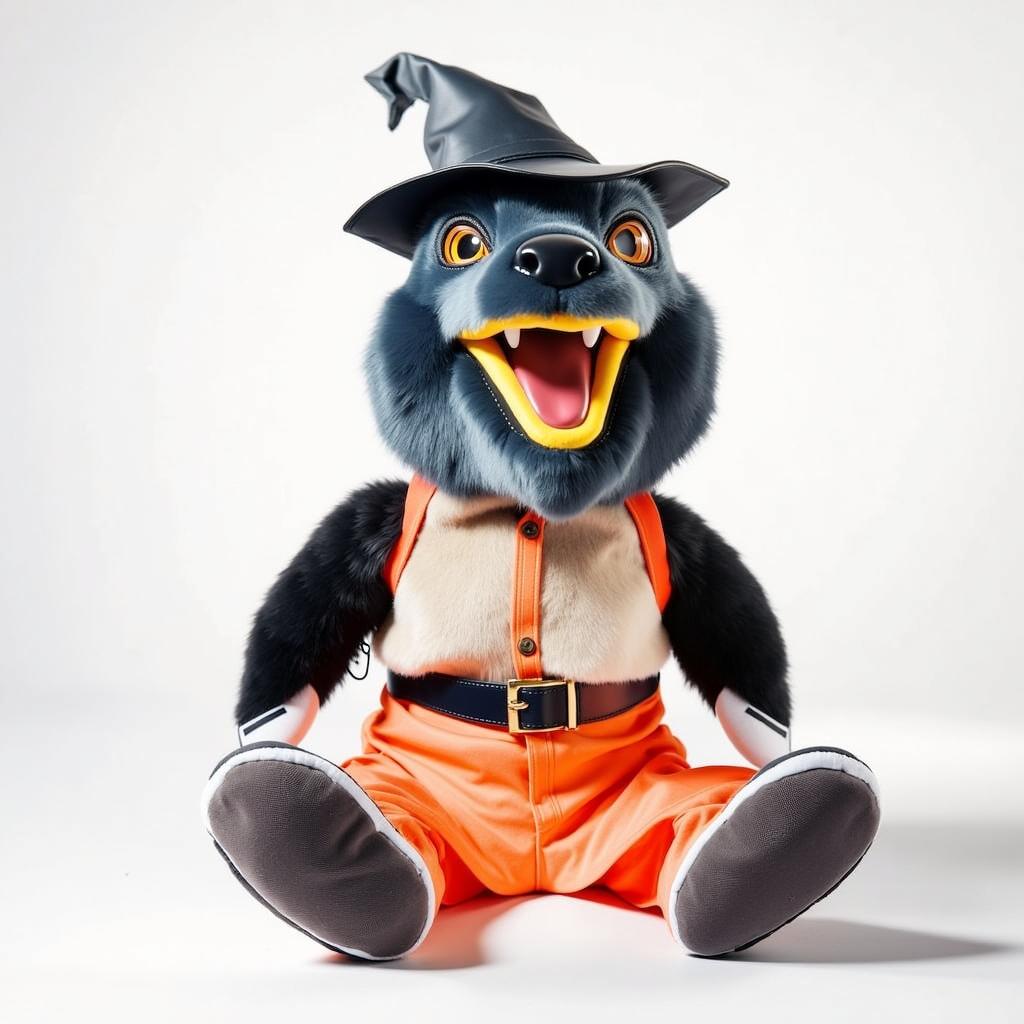} &

        \includegraphics[height=0.1\textheight]{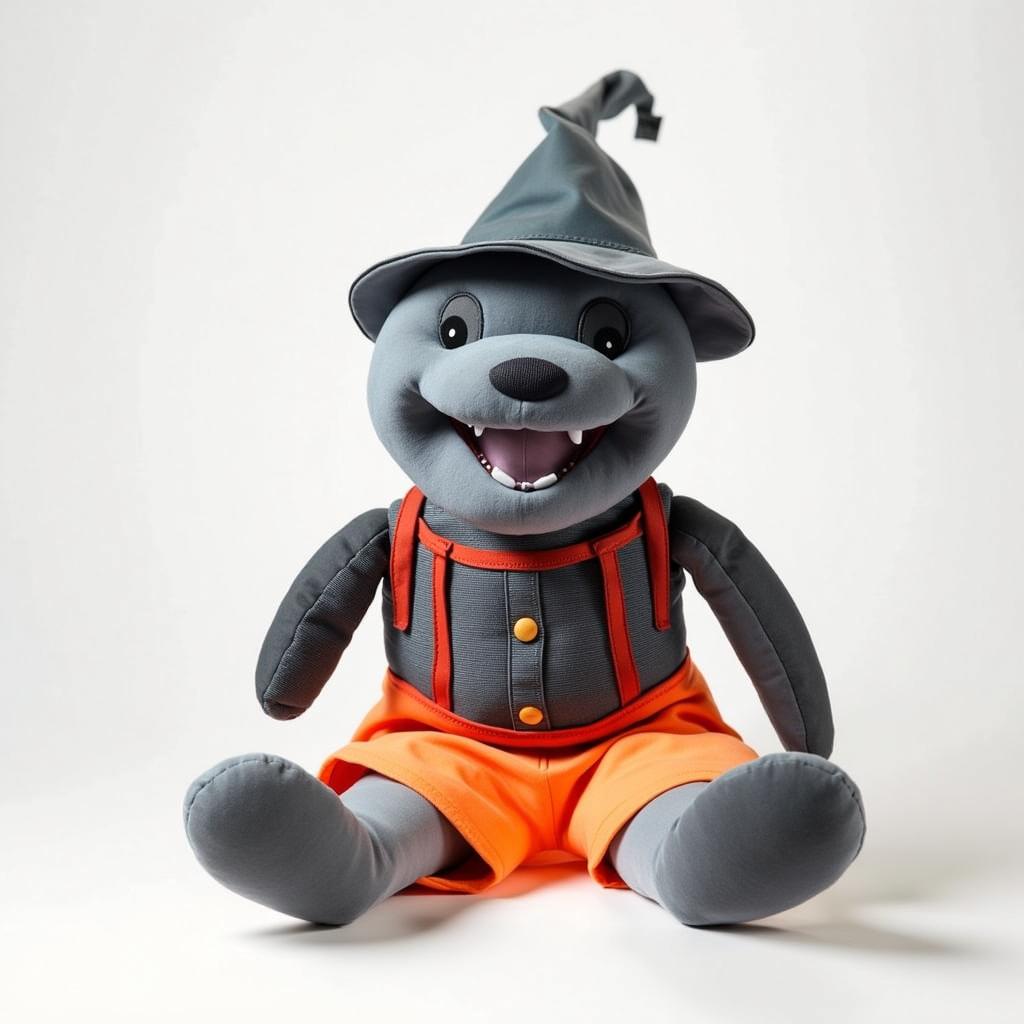} 
        \\

        \includegraphics[height=0.1\textheight]{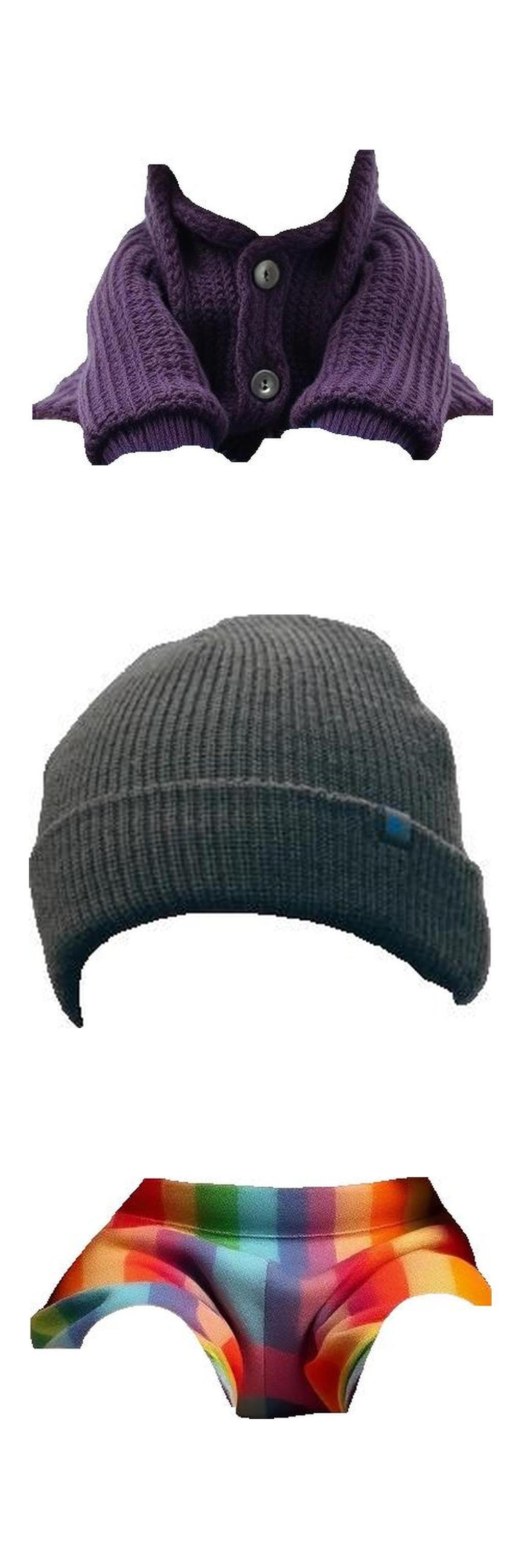} &
        \includegraphics[height=0.1\textheight]{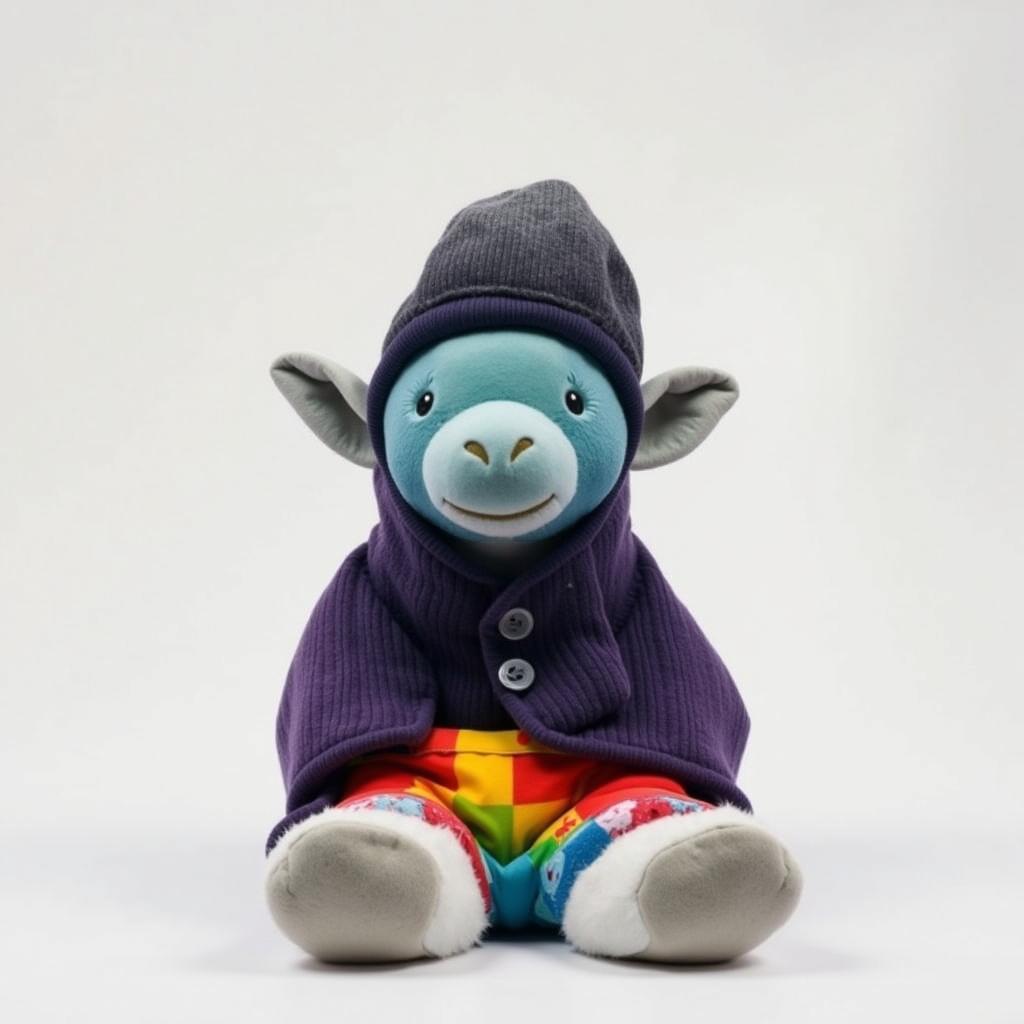} &
        \includegraphics[height=0.1\textheight]{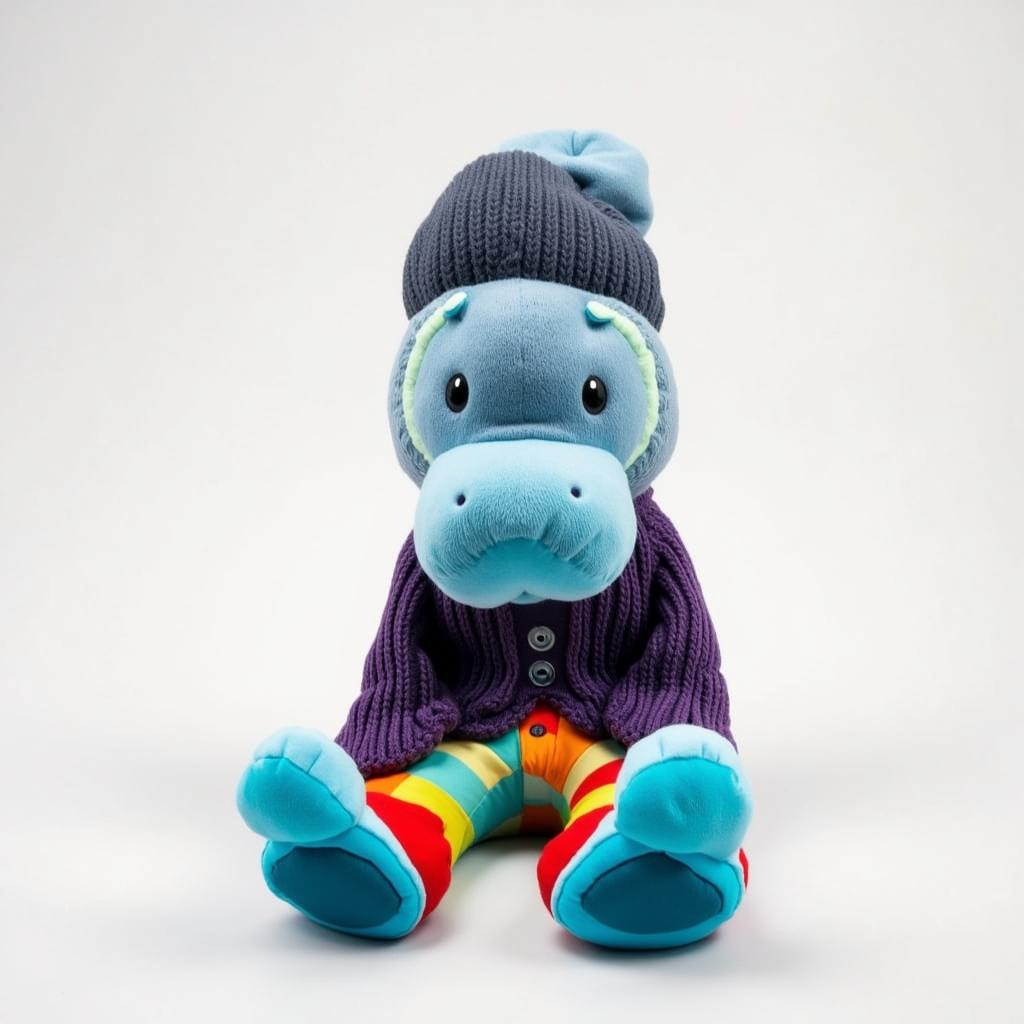} &

        \includegraphics[height=0.1\textheight]{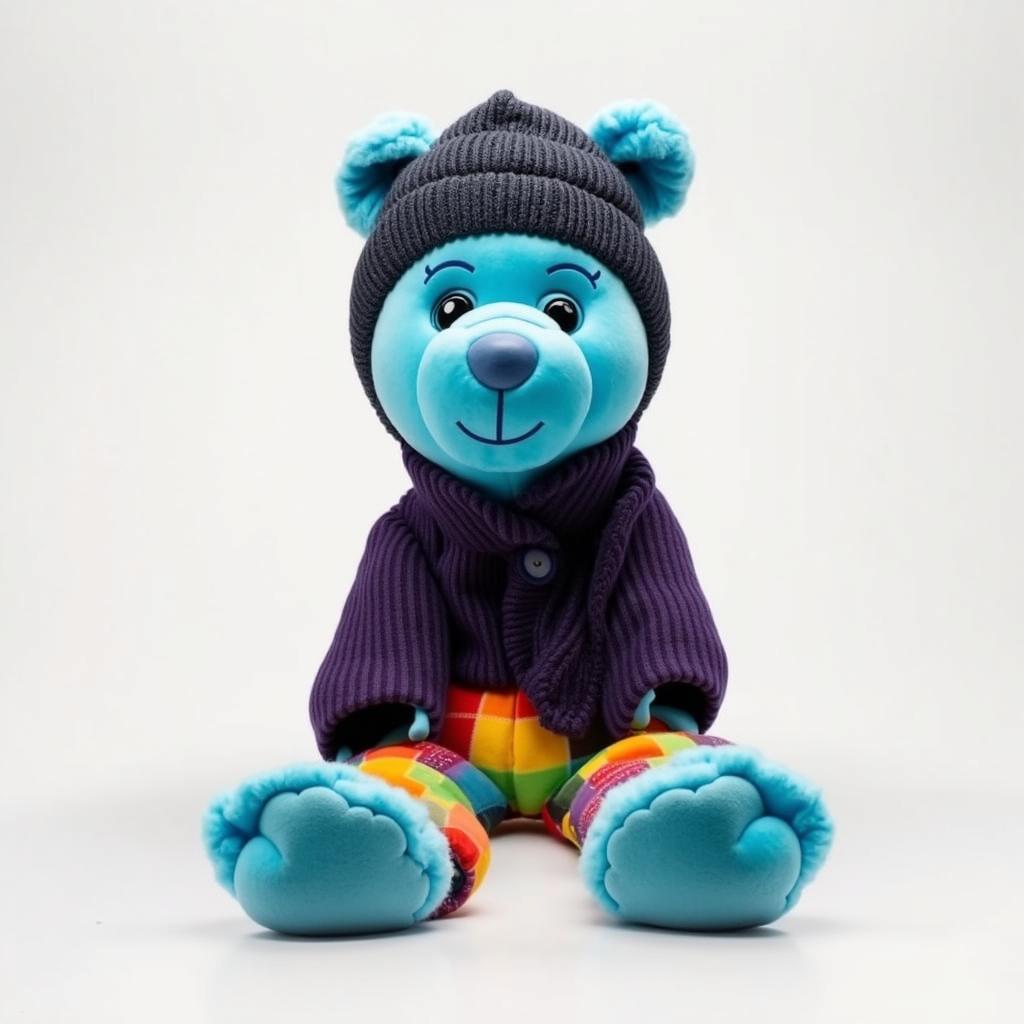} 
        \\

        \includegraphics[height=0.1\textheight]{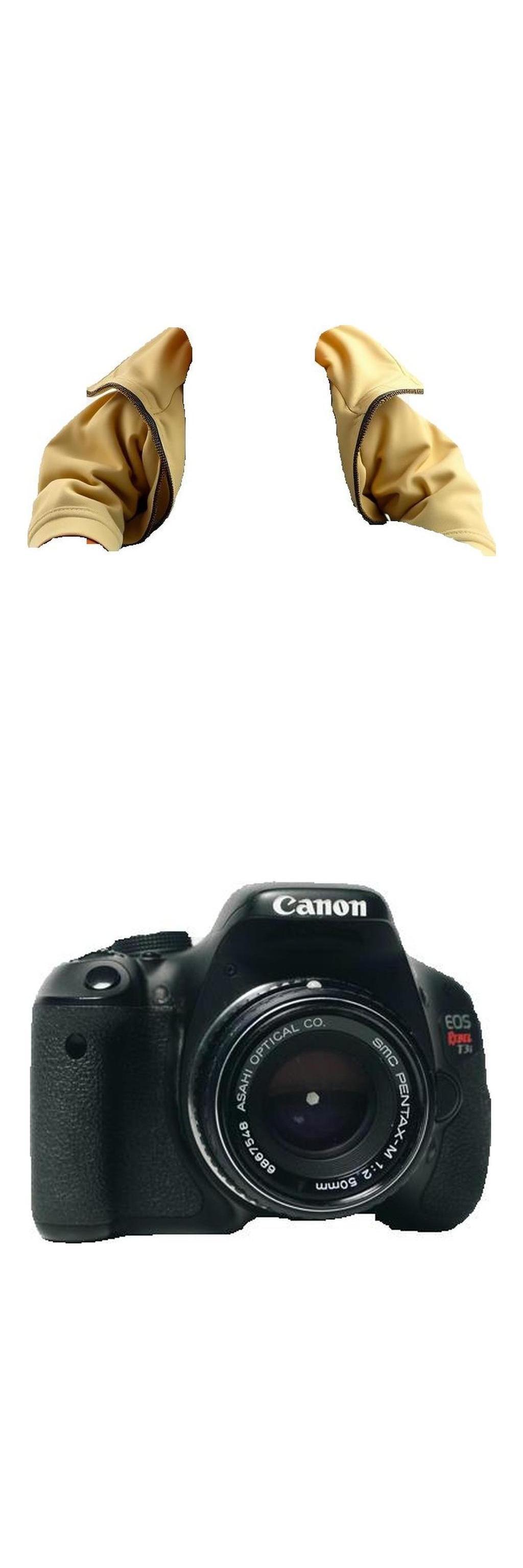} &
        \includegraphics[height=0.1\textheight]{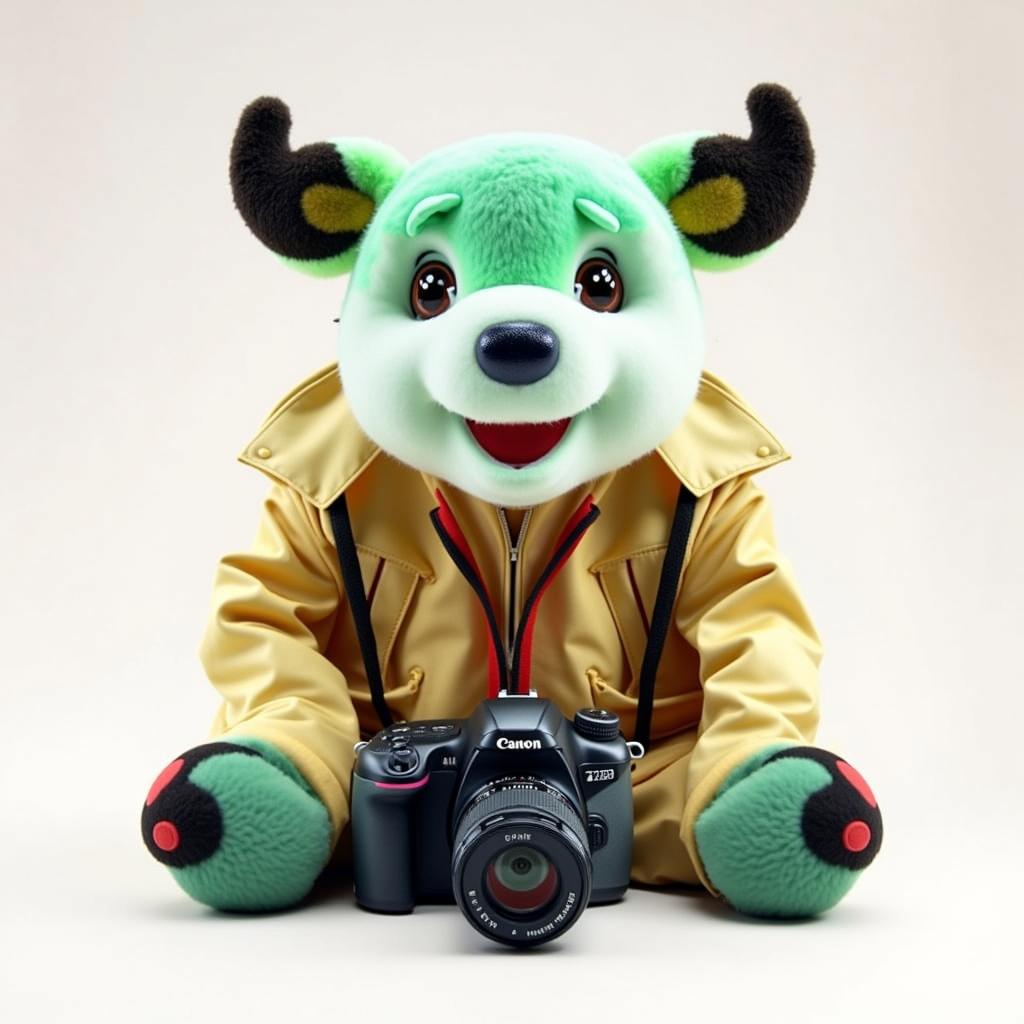} &
        \includegraphics[height=0.1\textheight]{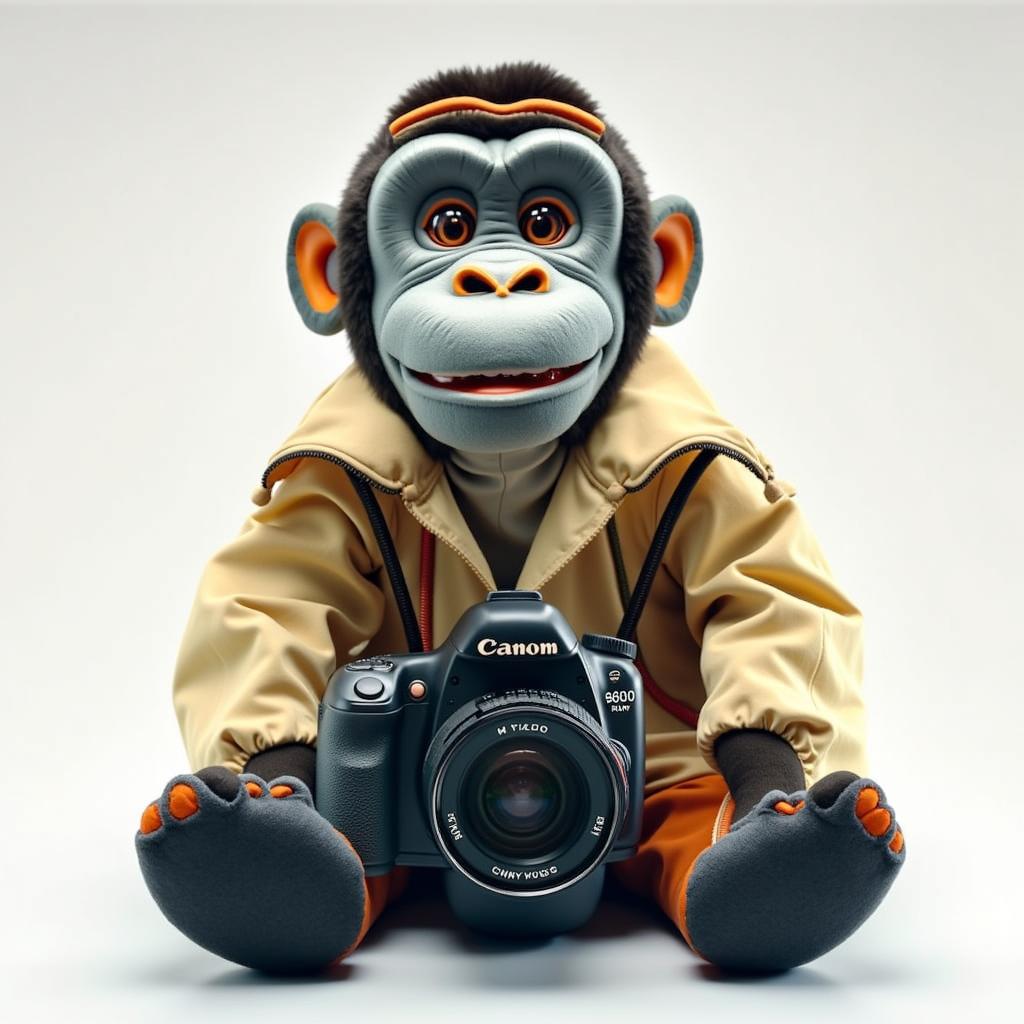} &

        \includegraphics[height=0.1\textheight]{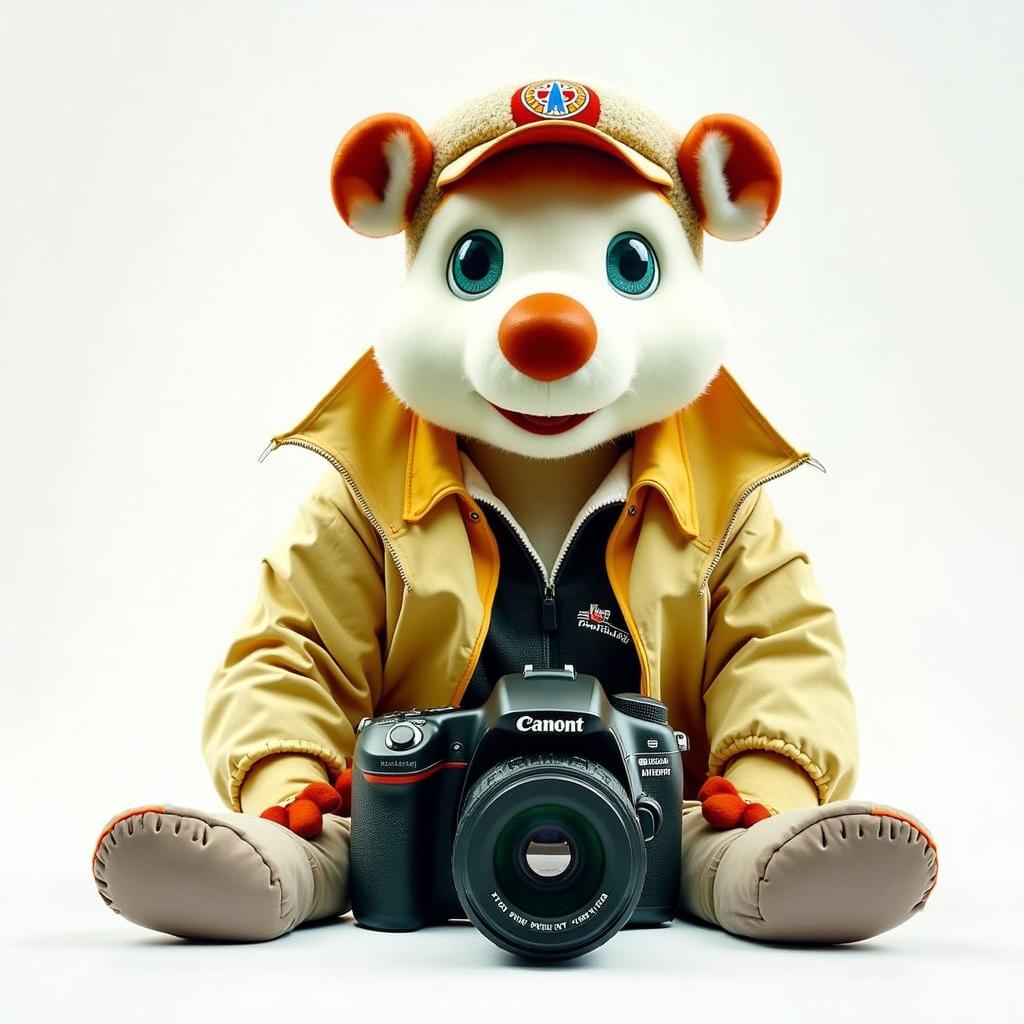} 
        \\

        \includegraphics[height=0.1\textheight]{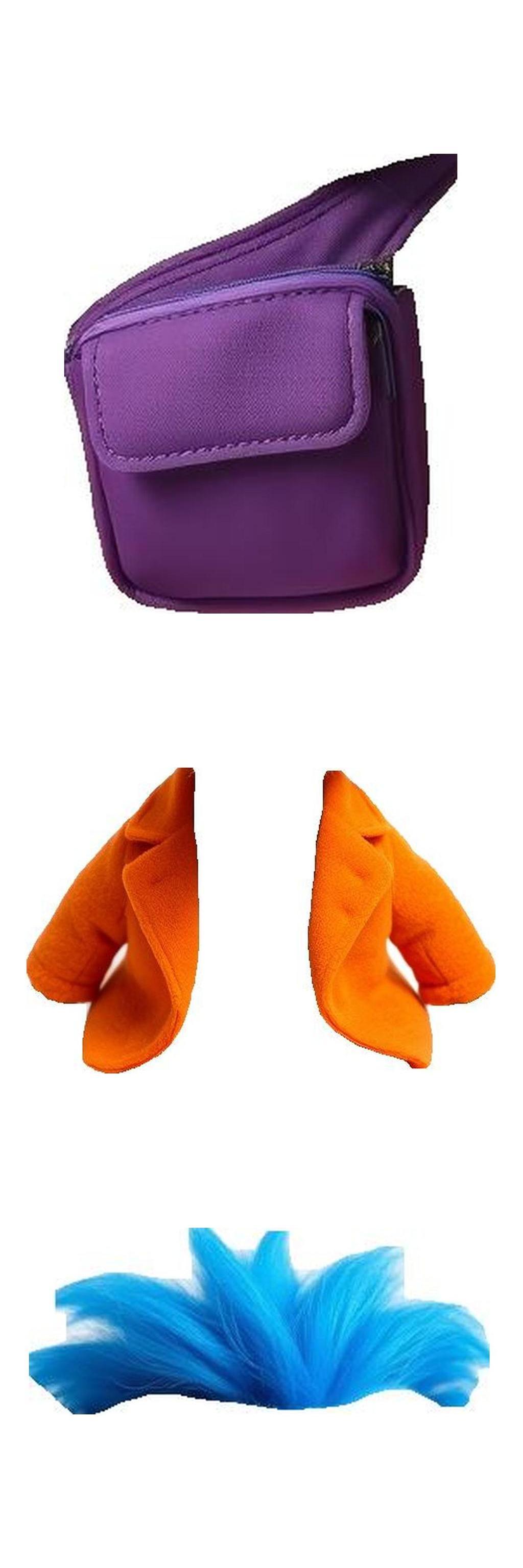} &
        \includegraphics[height=0.1\textheight]{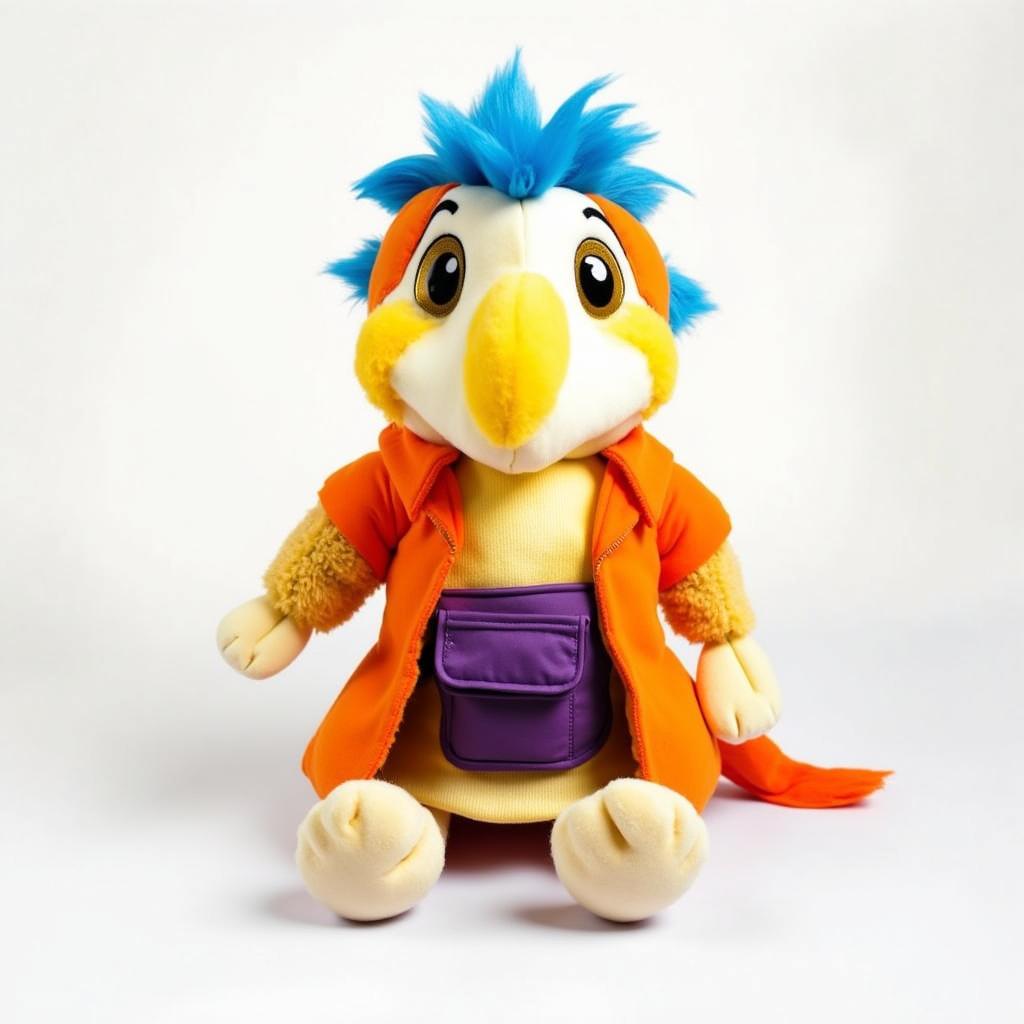} &
        \includegraphics[height=0.1\textheight]{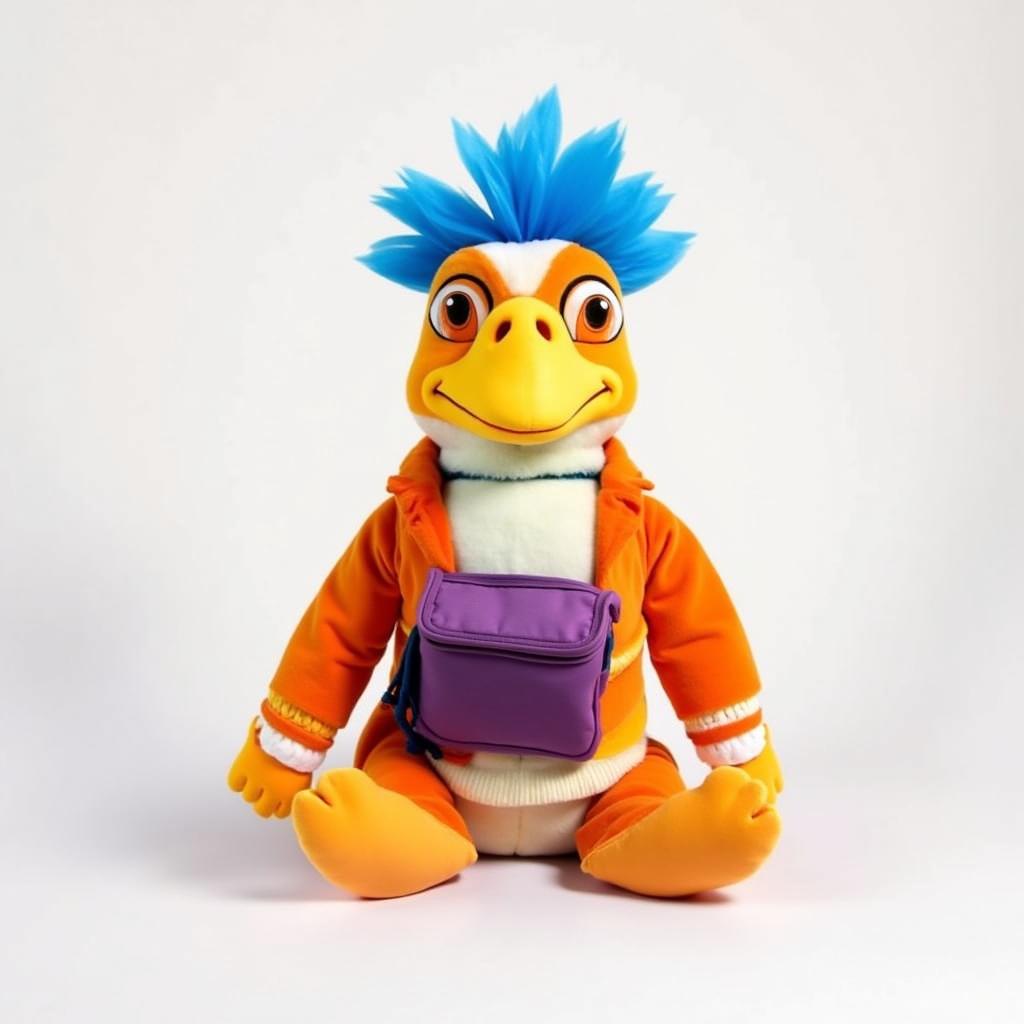} &

        \includegraphics[height=0.1\textheight]{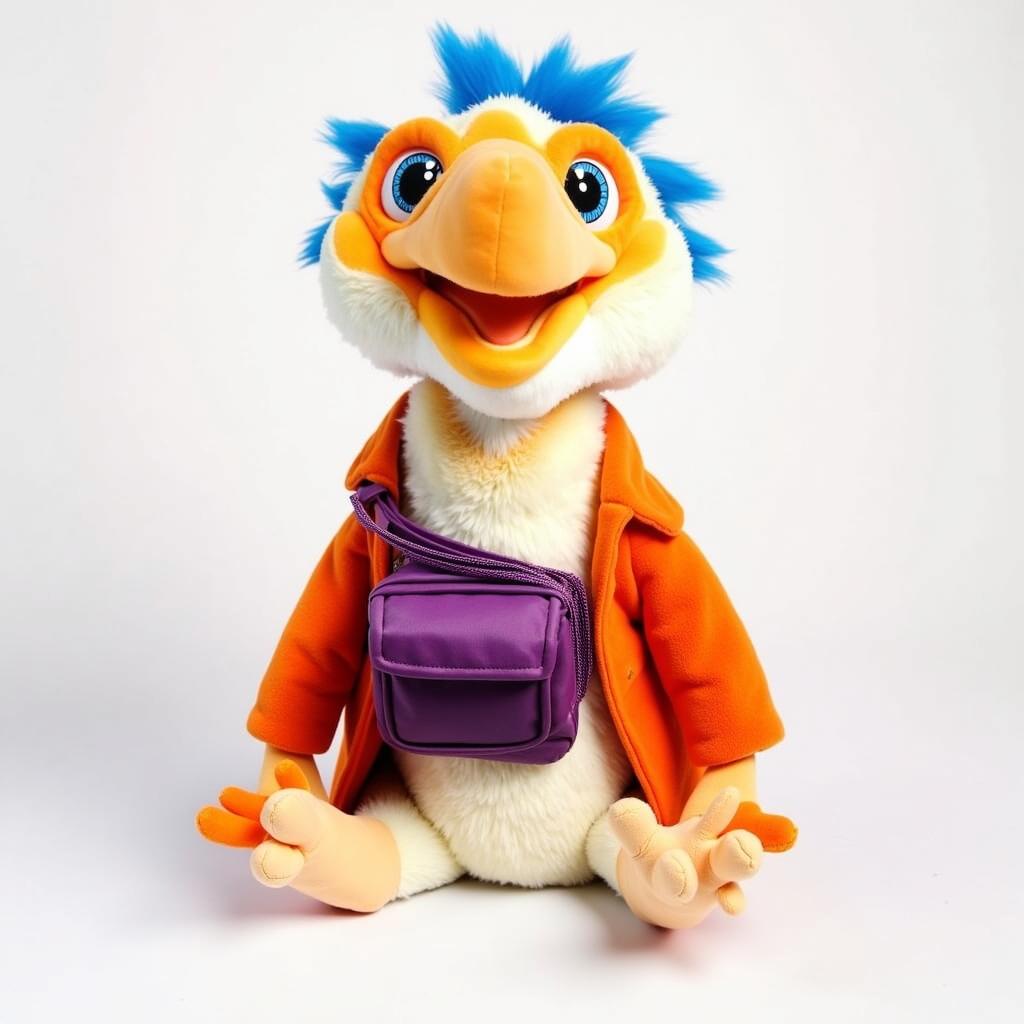} 
        \\ \\[-0.1cm]

        Input & \multicolumn{3}{c}{Sampled Results} 

    \end{tabular}
    }
    \caption{Additional PiT results for toy concepting}
    \label{fig:supp_toys_1}
\end{figure}

\begin{figure}
    \centering
    \setlength{\tabcolsep}{0.5pt}
    \renewcommand{\arraystretch}{0.5}
    {
    \begin{tabular}{c @{\hspace{0.2cm}} c c c}

        \includegraphics[height=0.1\textheight]{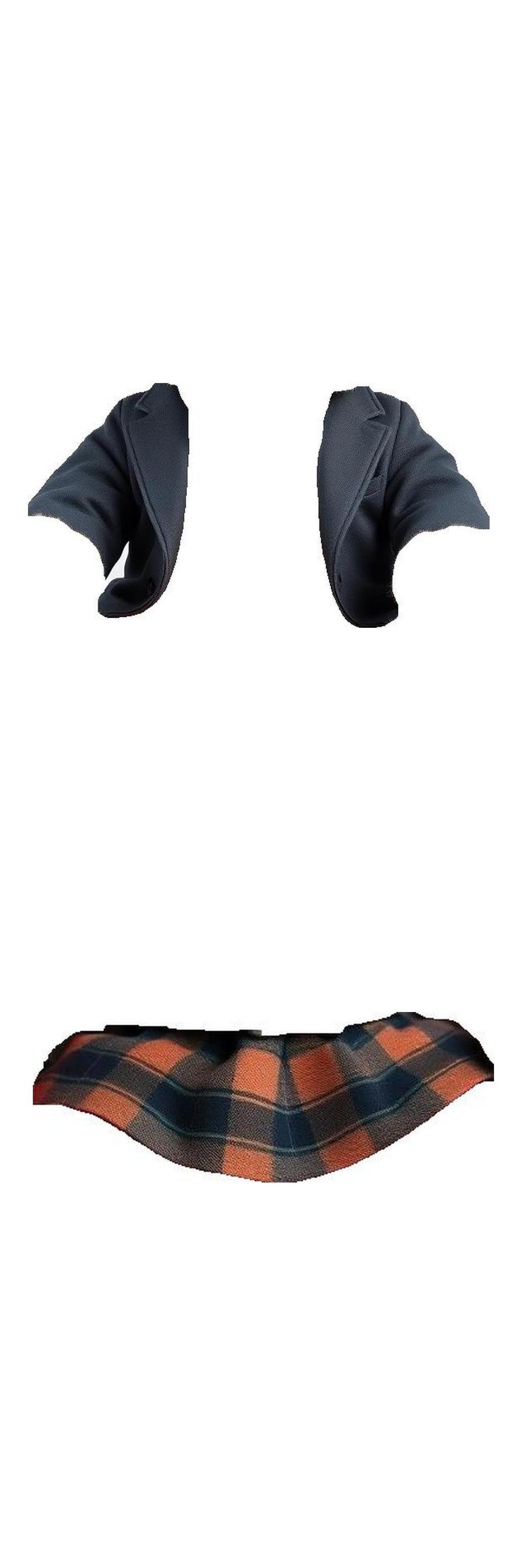} &
        \includegraphics[height=0.1\textheight]{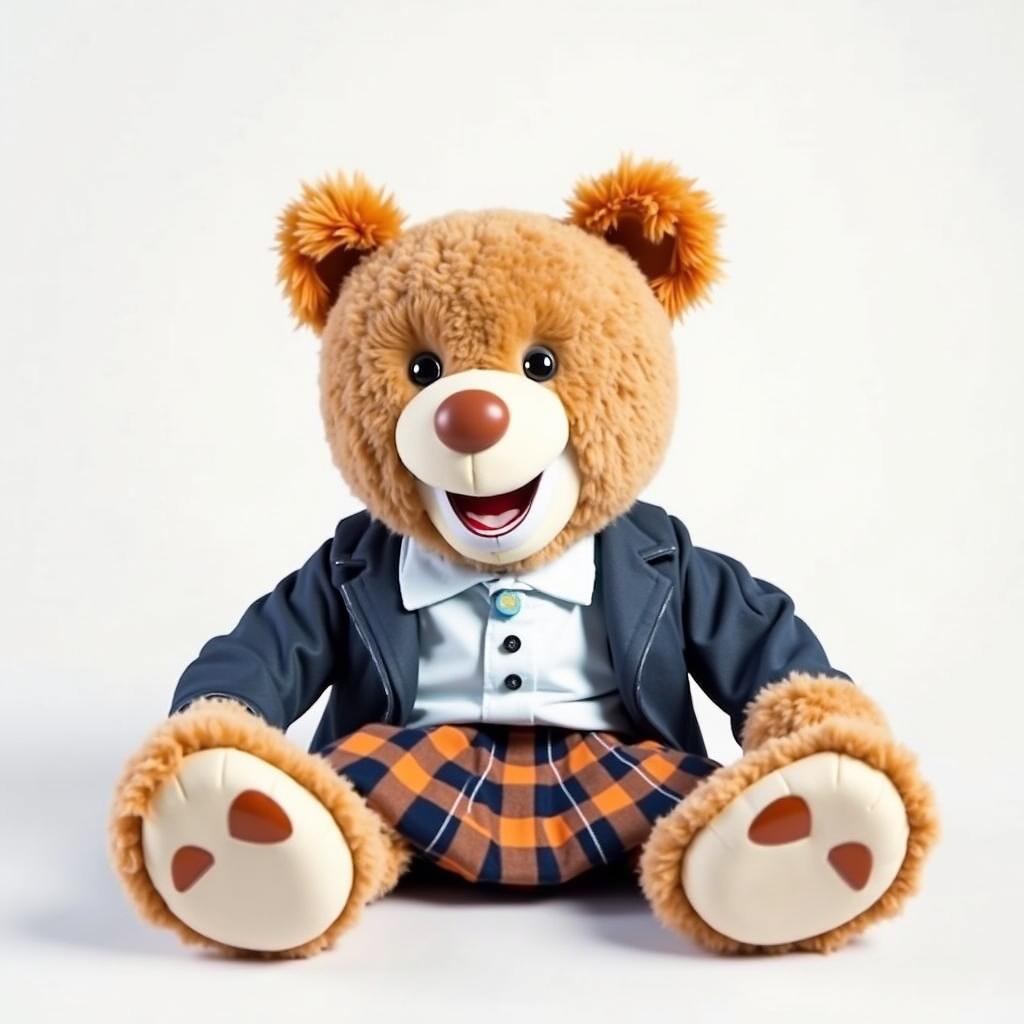} &
        \includegraphics[height=0.1\textheight]{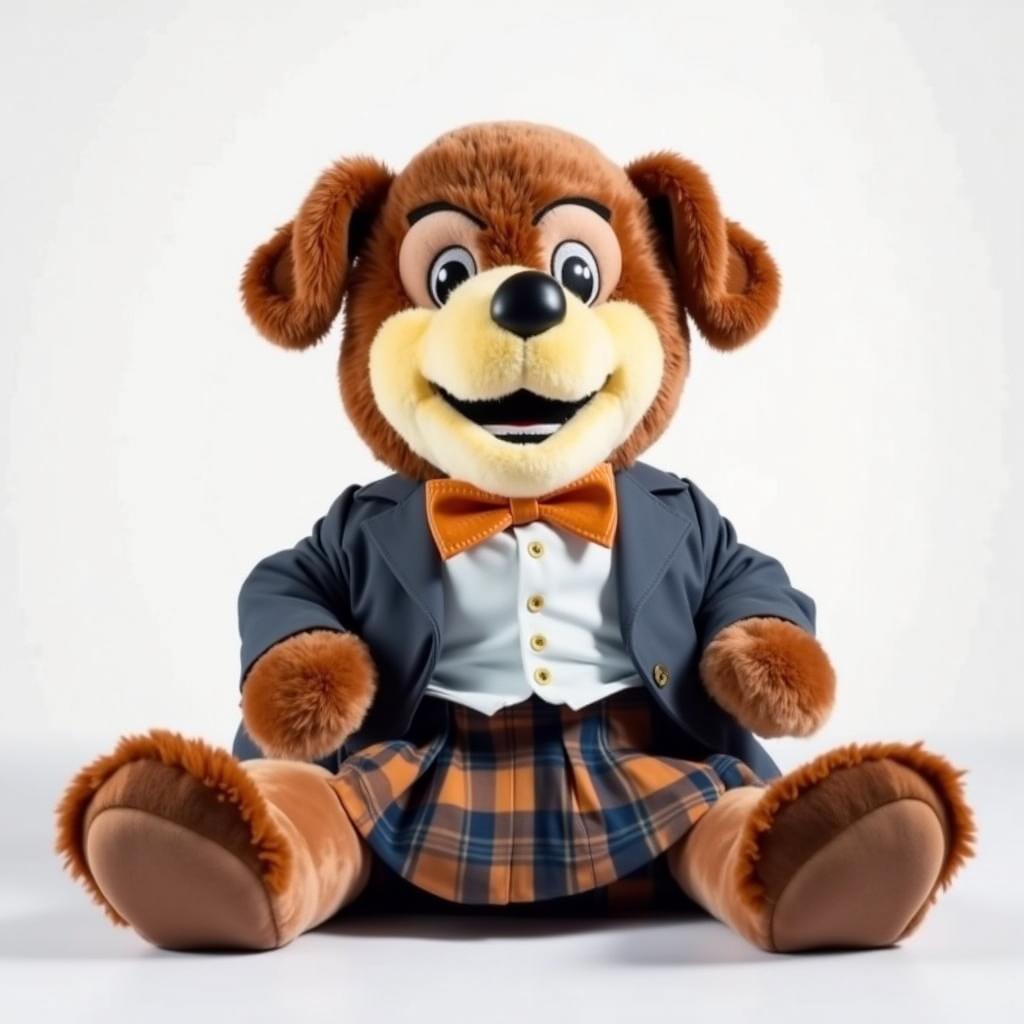} &

        \includegraphics[height=0.1\textheight]{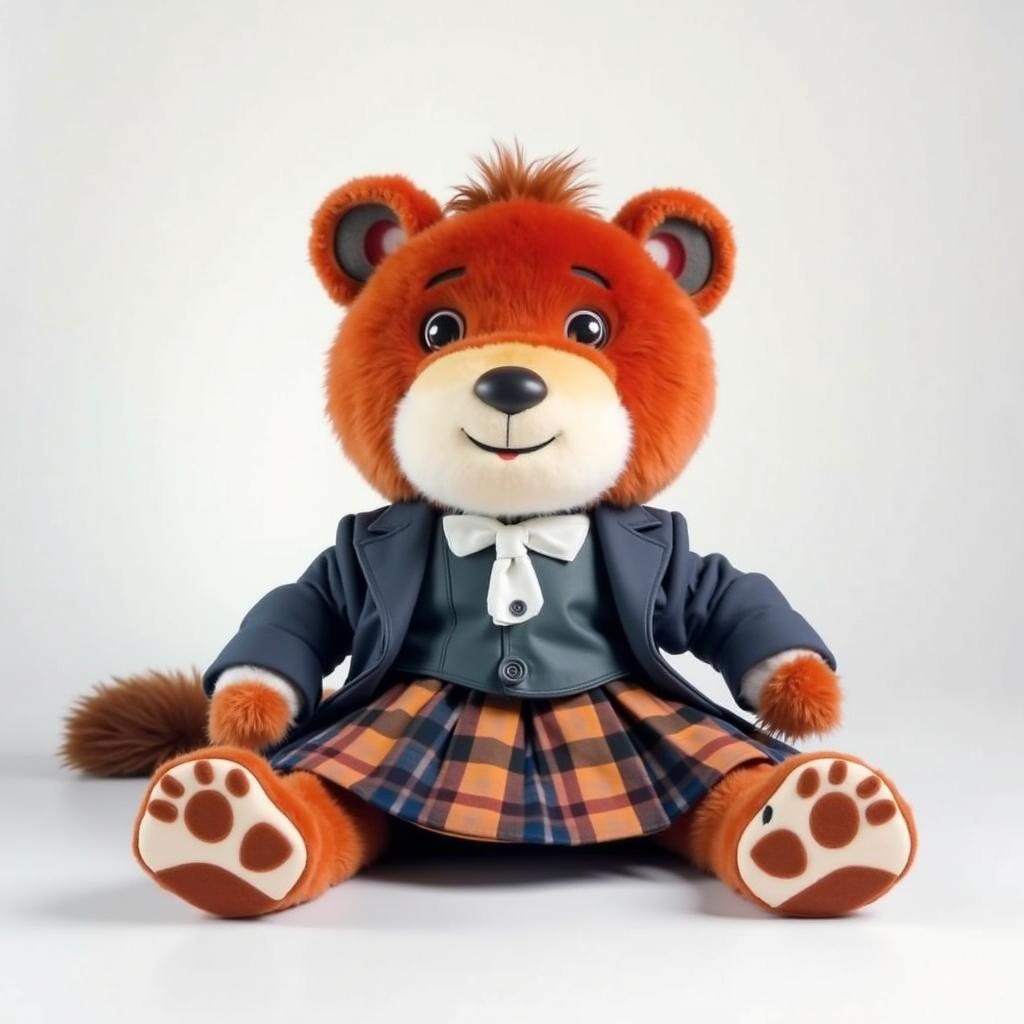} 
        \\

        \includegraphics[height=0.1\textheight]{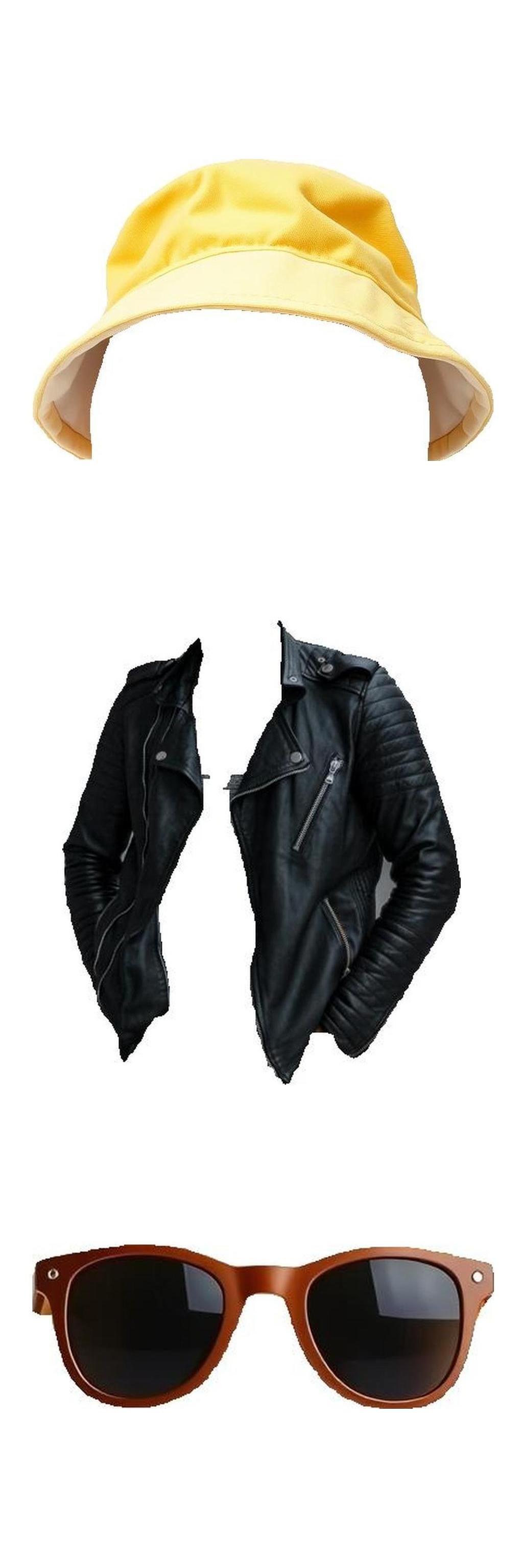} &
        \includegraphics[height=0.1\textheight]{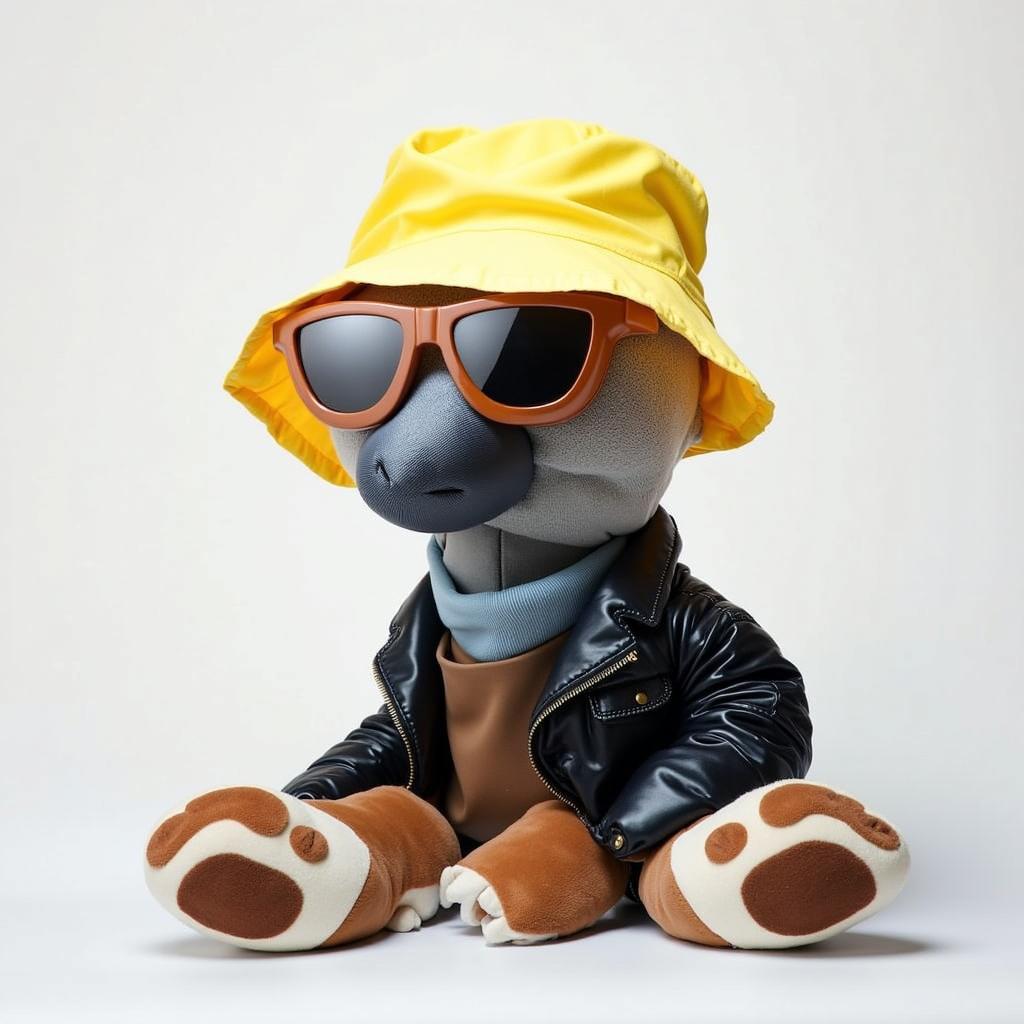} &
        \includegraphics[height=0.1\textheight]{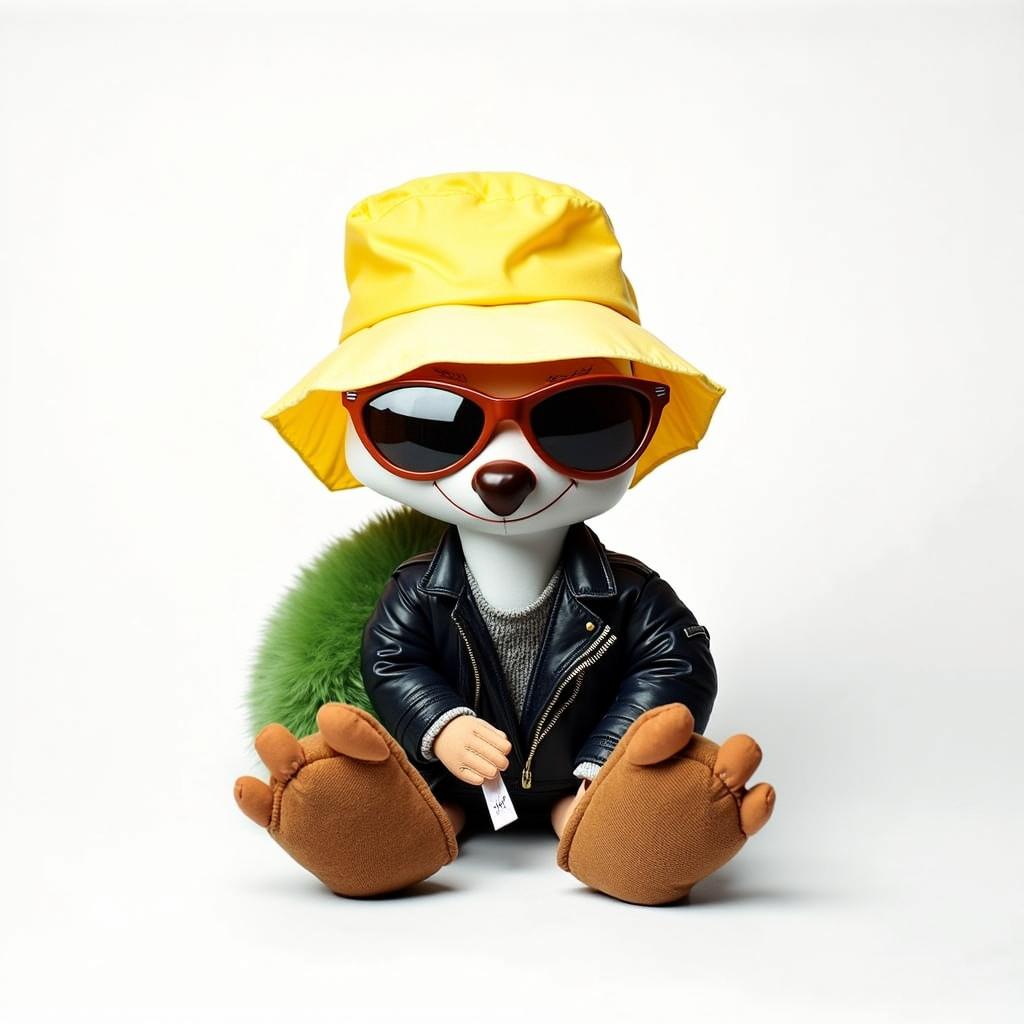} &

        \includegraphics[height=0.1\textheight]{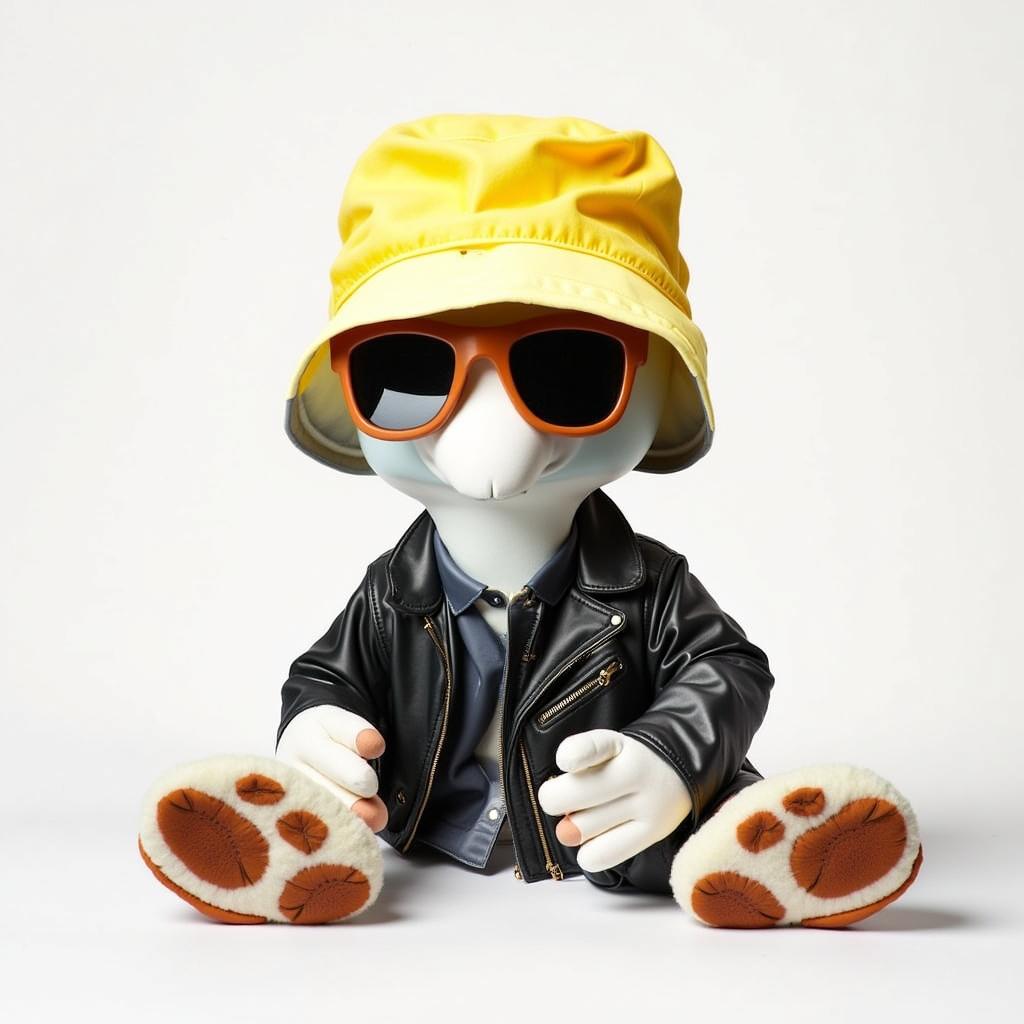} 
        \\

        \includegraphics[height=0.1\textheight]{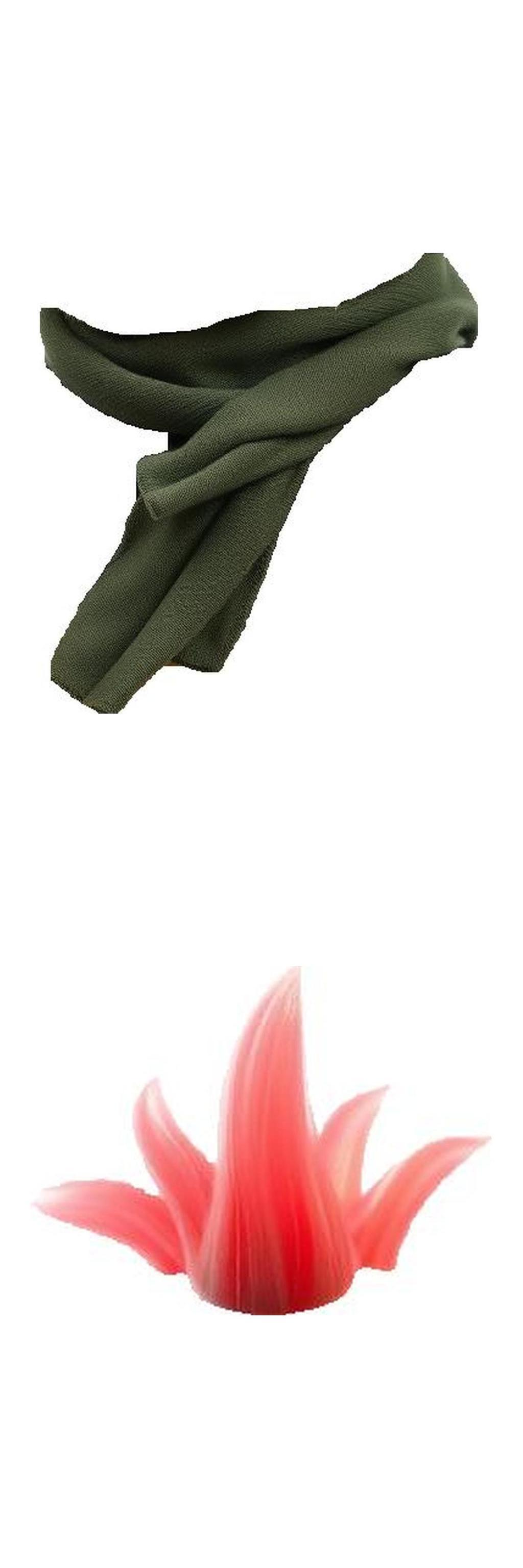} &
        \includegraphics[height=0.1\textheight]{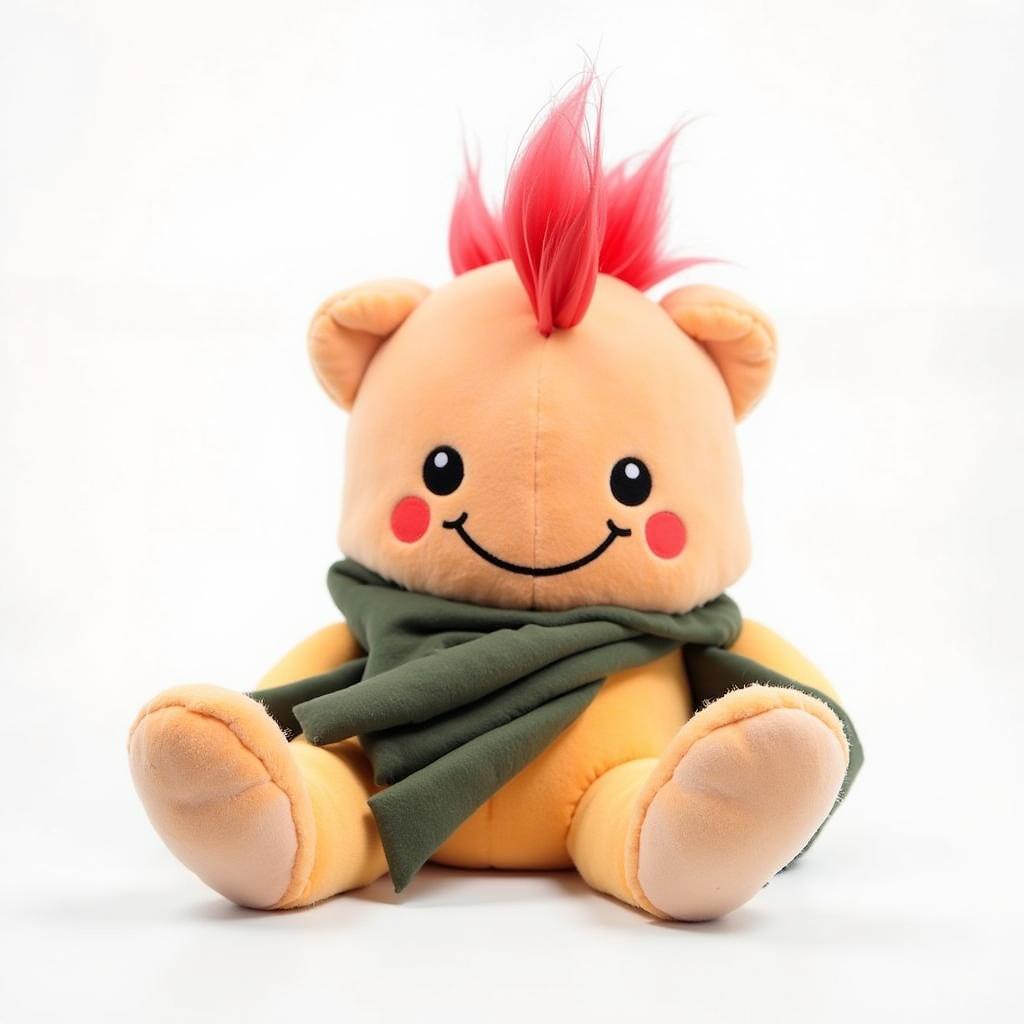} &
        \includegraphics[height=0.1\textheight]{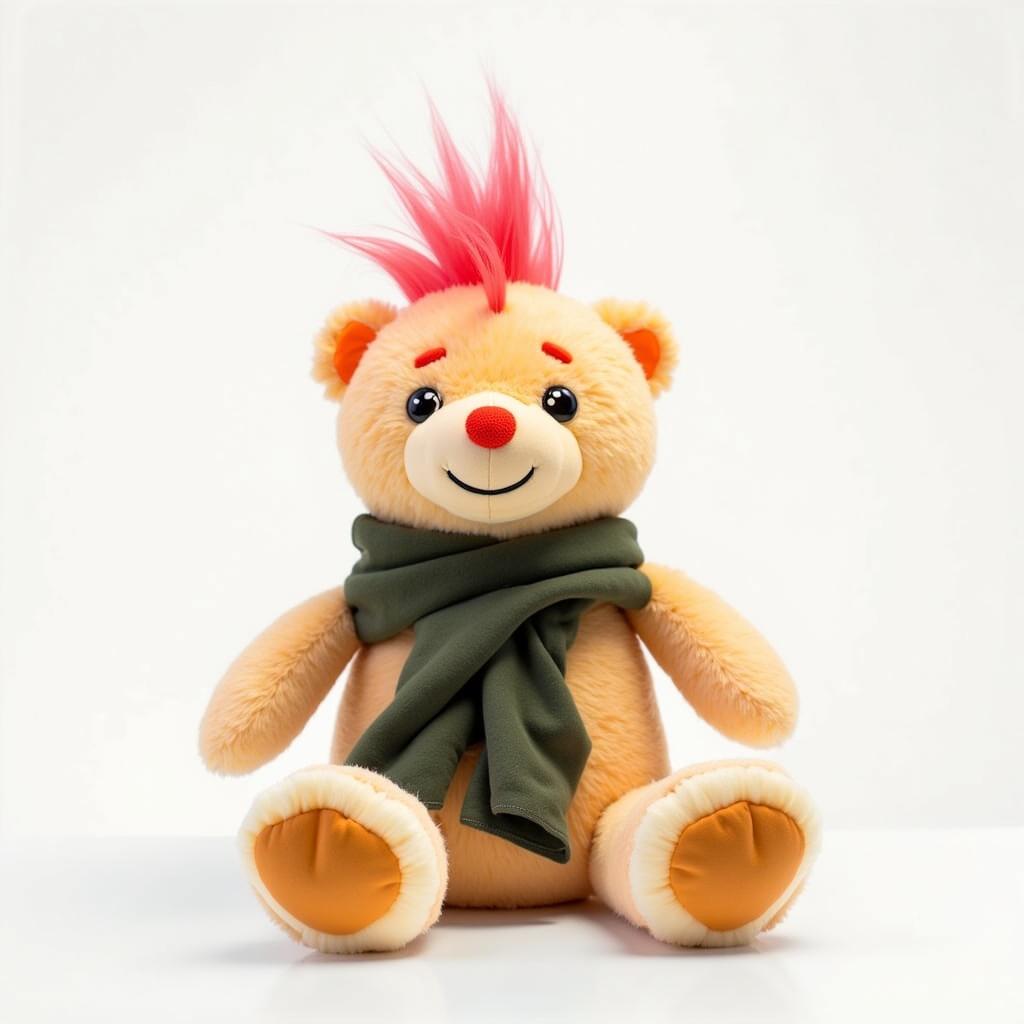} &

        \includegraphics[height=0.1\textheight]{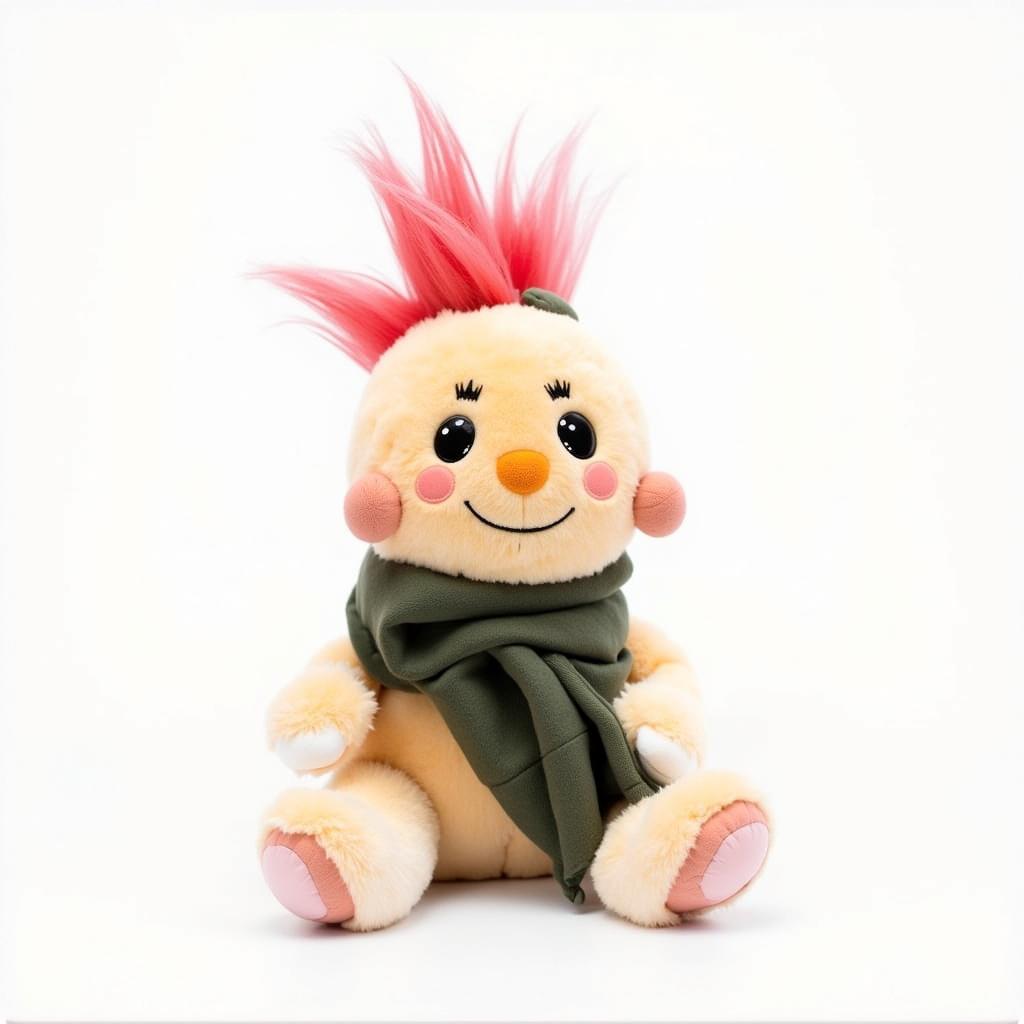} 
        \\

        \includegraphics[height=0.1\textheight]{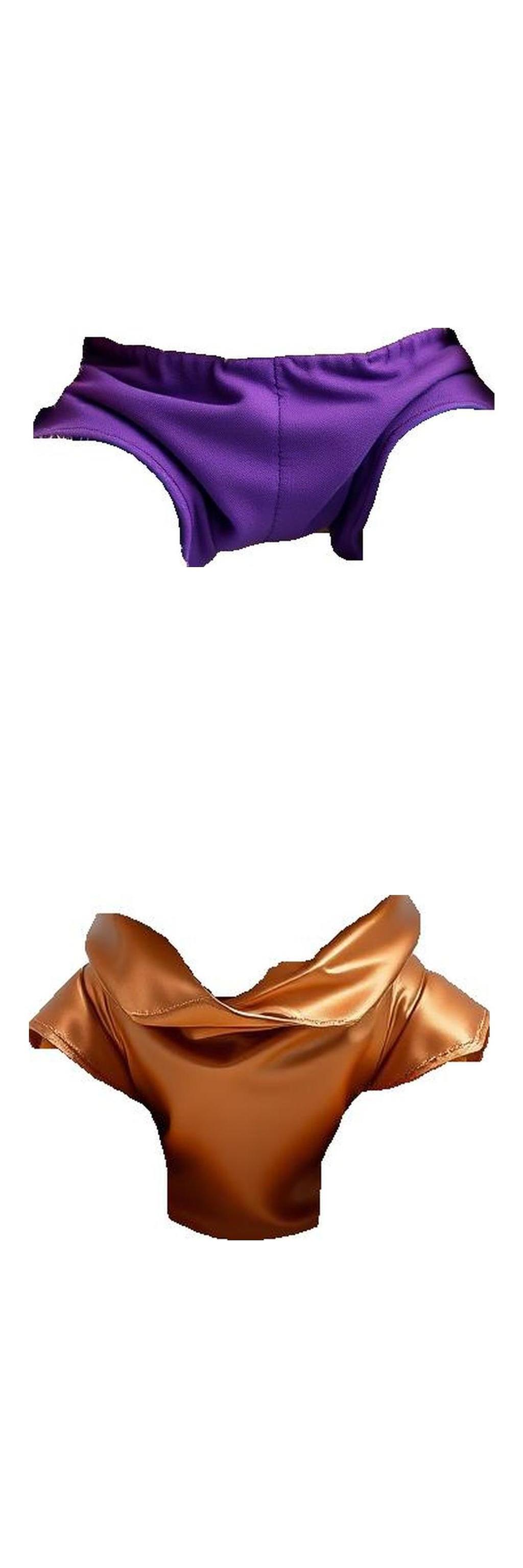} &
        \includegraphics[height=0.1\textheight]{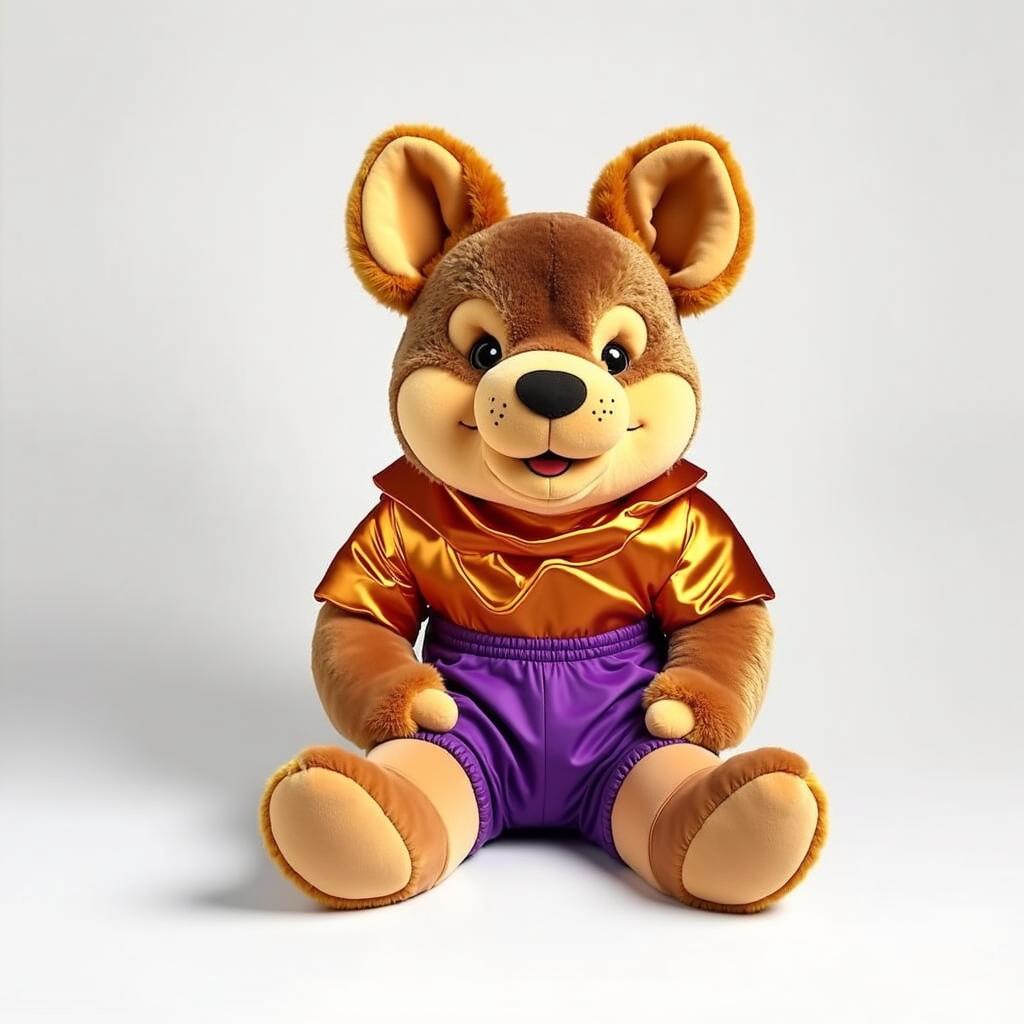} &
        \includegraphics[height=0.1\textheight]{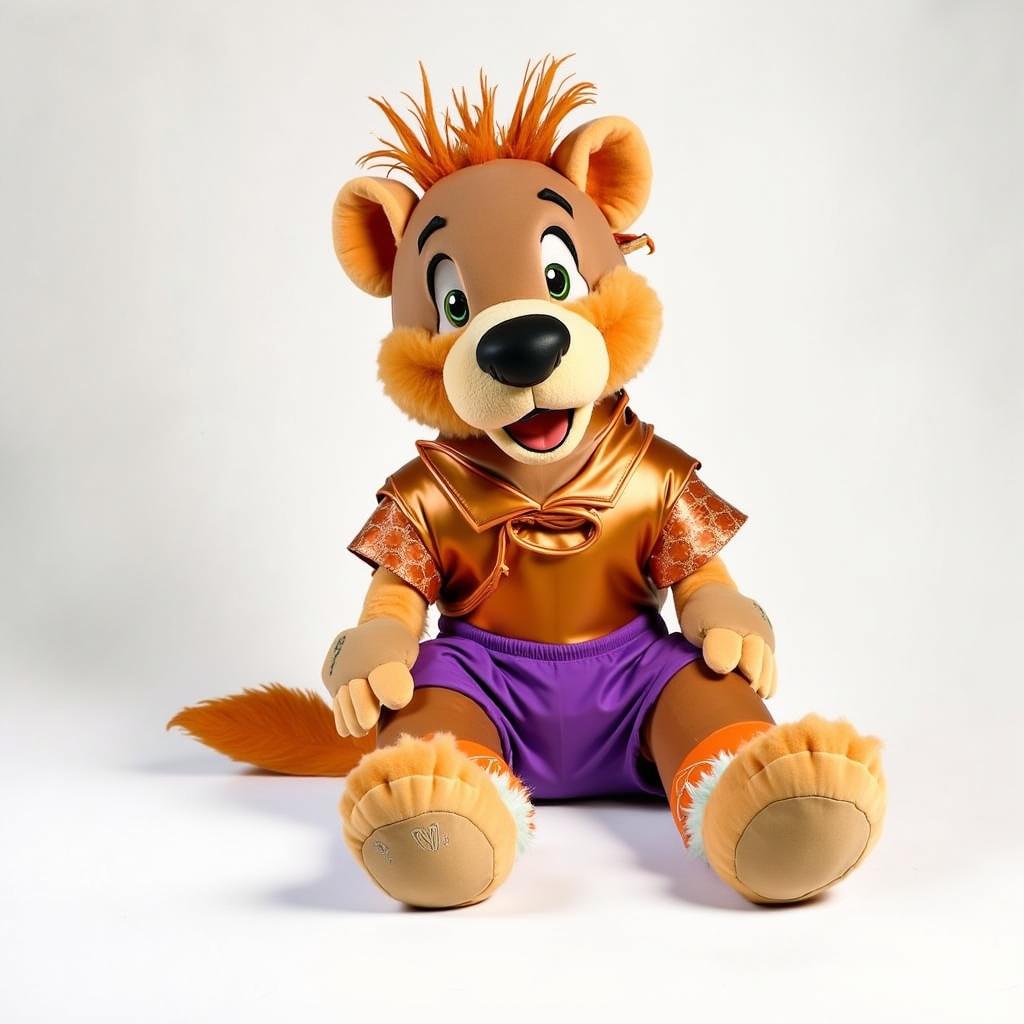} &

        \includegraphics[height=0.1\textheight]{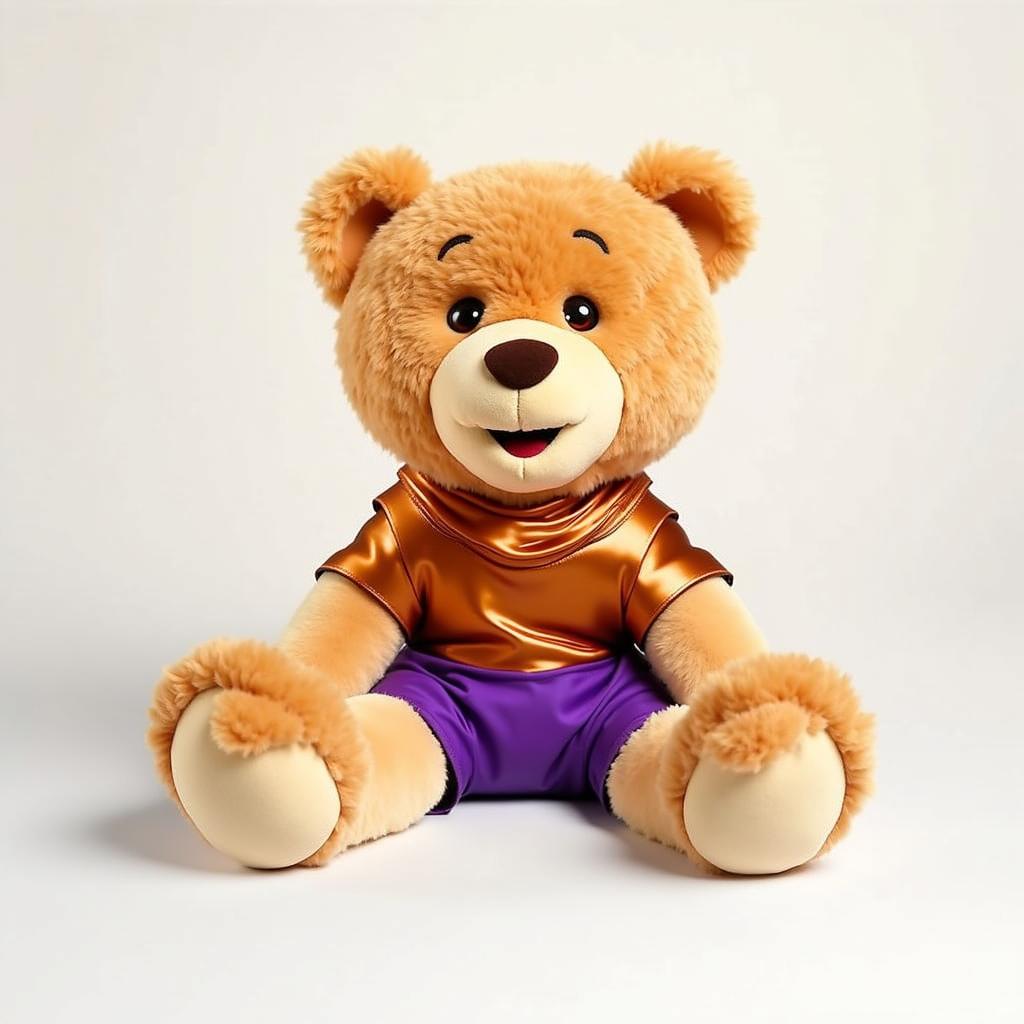} 
        \\

        \includegraphics[height=0.1\textheight]{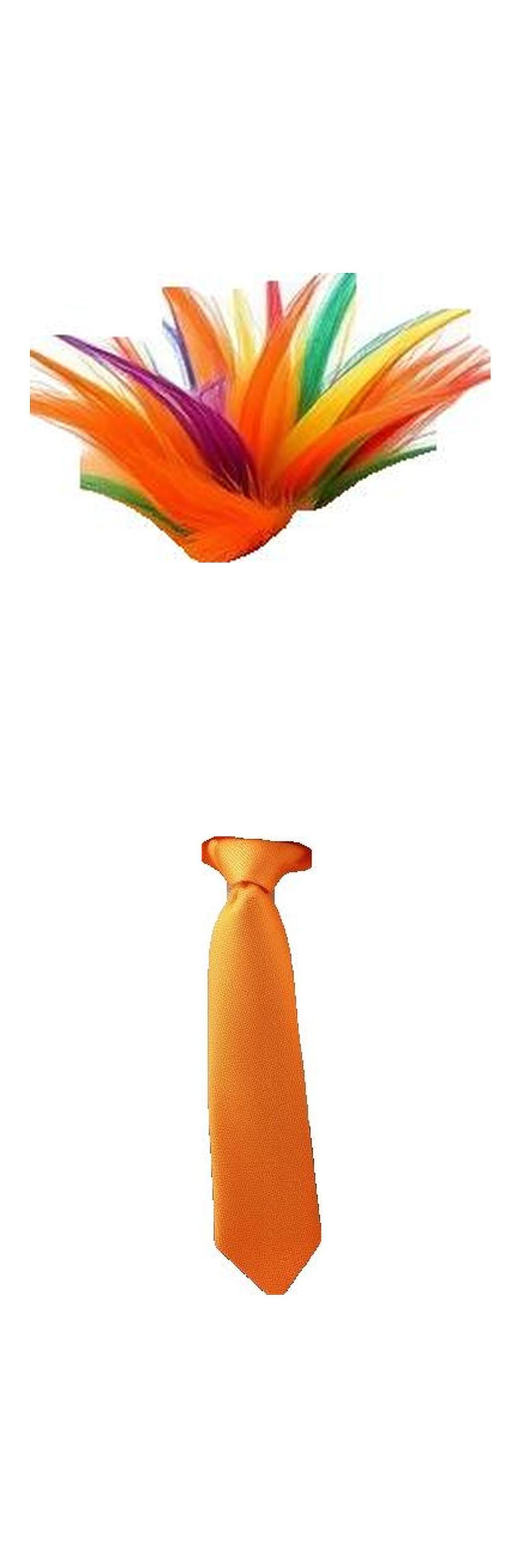} &
        \includegraphics[height=0.1\textheight]{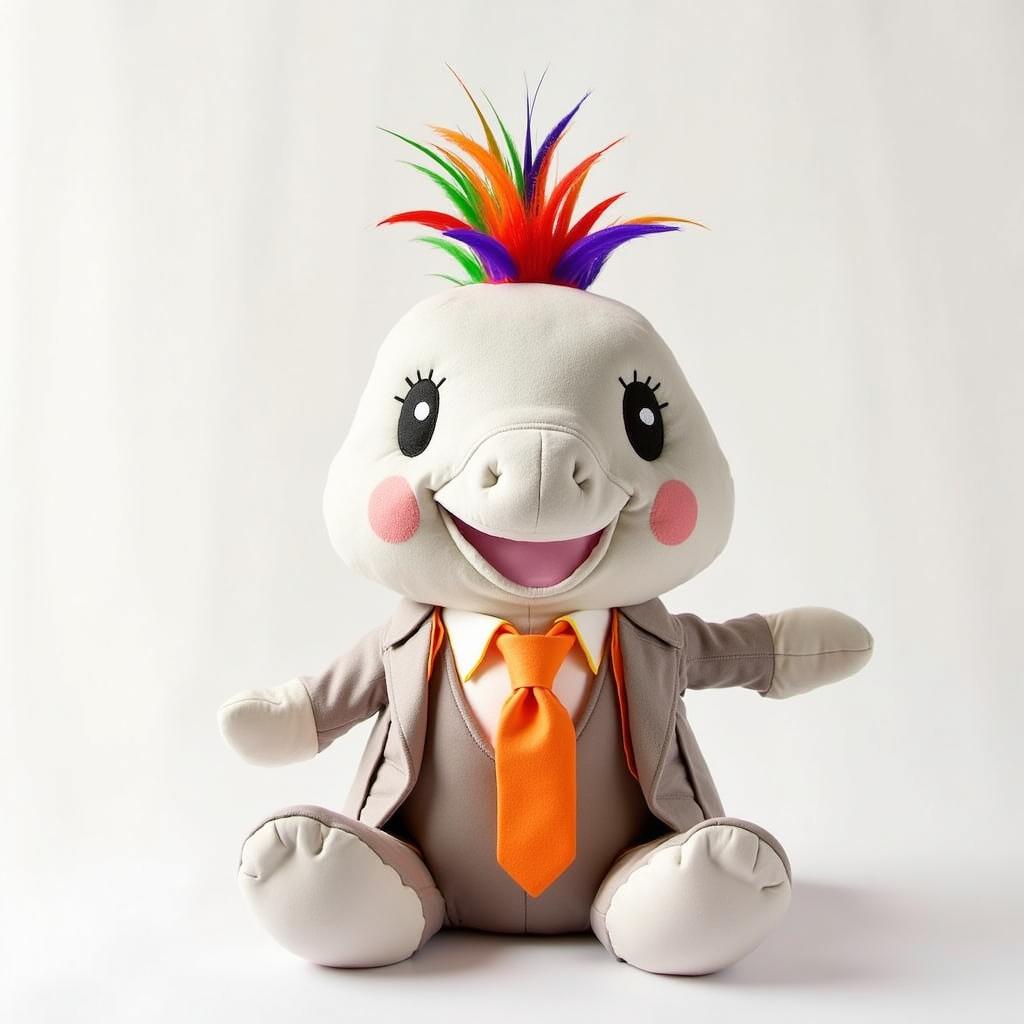} &
        \includegraphics[height=0.1\textheight]{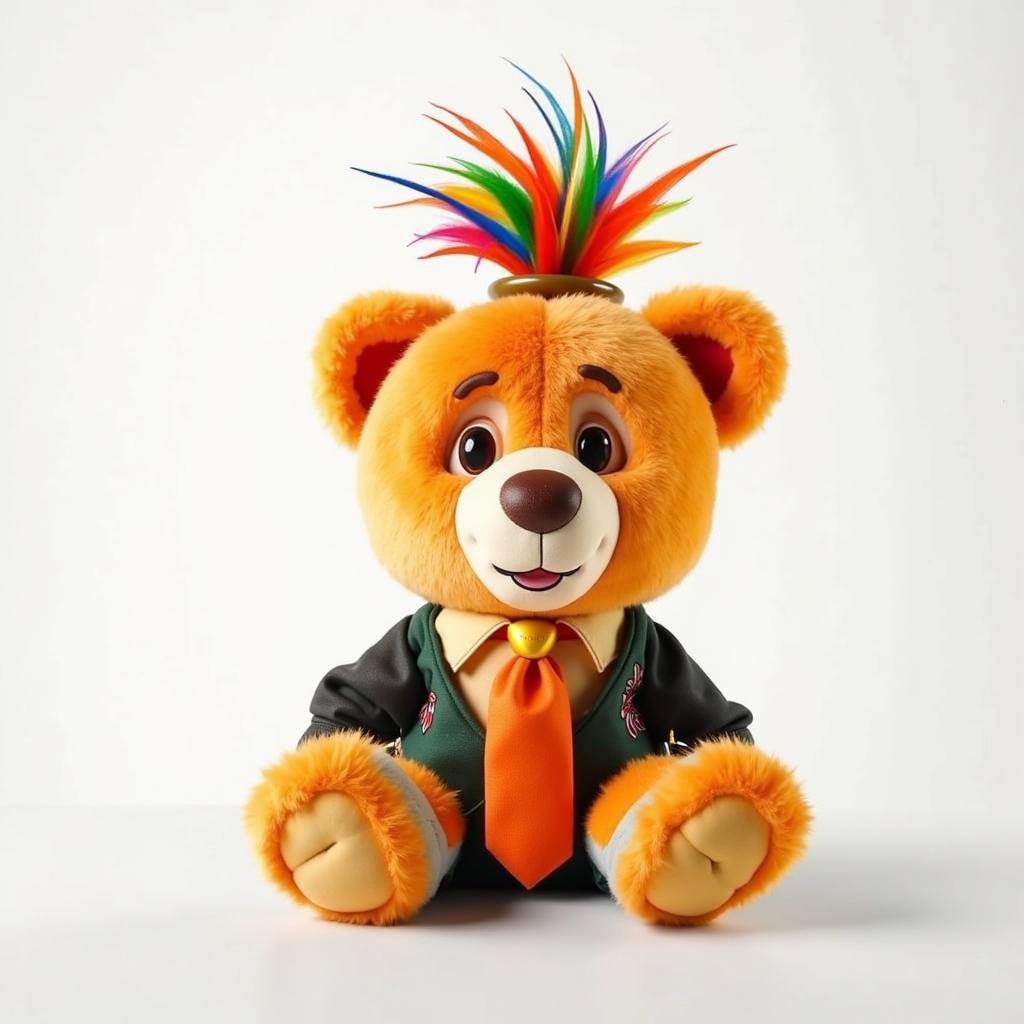} &

        \includegraphics[height=0.1\textheight]{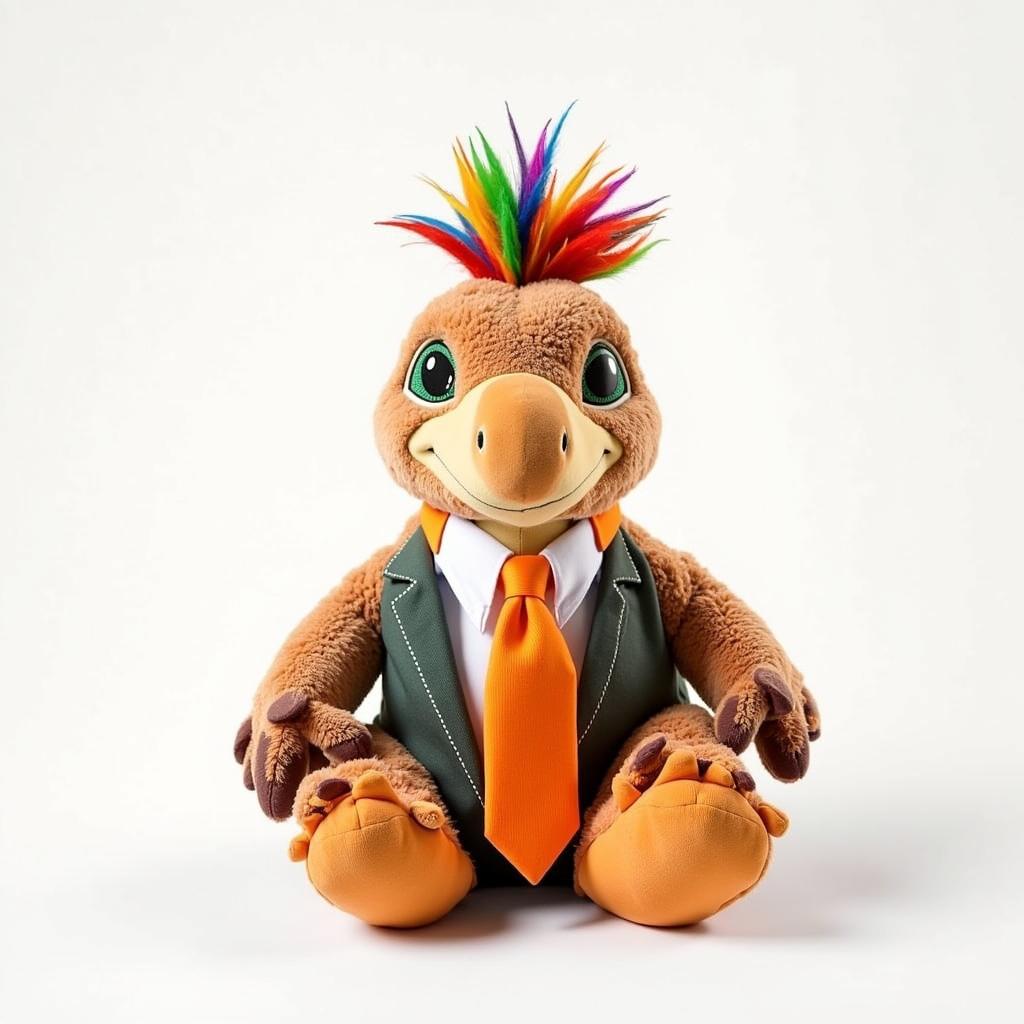} 
        \\ \\[-0.1cm]

        Input & \multicolumn{3}{c}{Sampled Results} 

    \end{tabular}
    }
    \caption{Additional PiT results for toy concepting}
    \label{fig:supp_toys_2}
\end{figure}

\begin{figure*}
    \centering
    \setlength{\tabcolsep}{0.5pt}
    \renewcommand{\arraystretch}{0.5}
    \addtolength{\belowcaptionskip}{-5pt}
    {\small
    \begin{tabular}{c c c c}
        \includegraphics[height=0.18\textwidth]{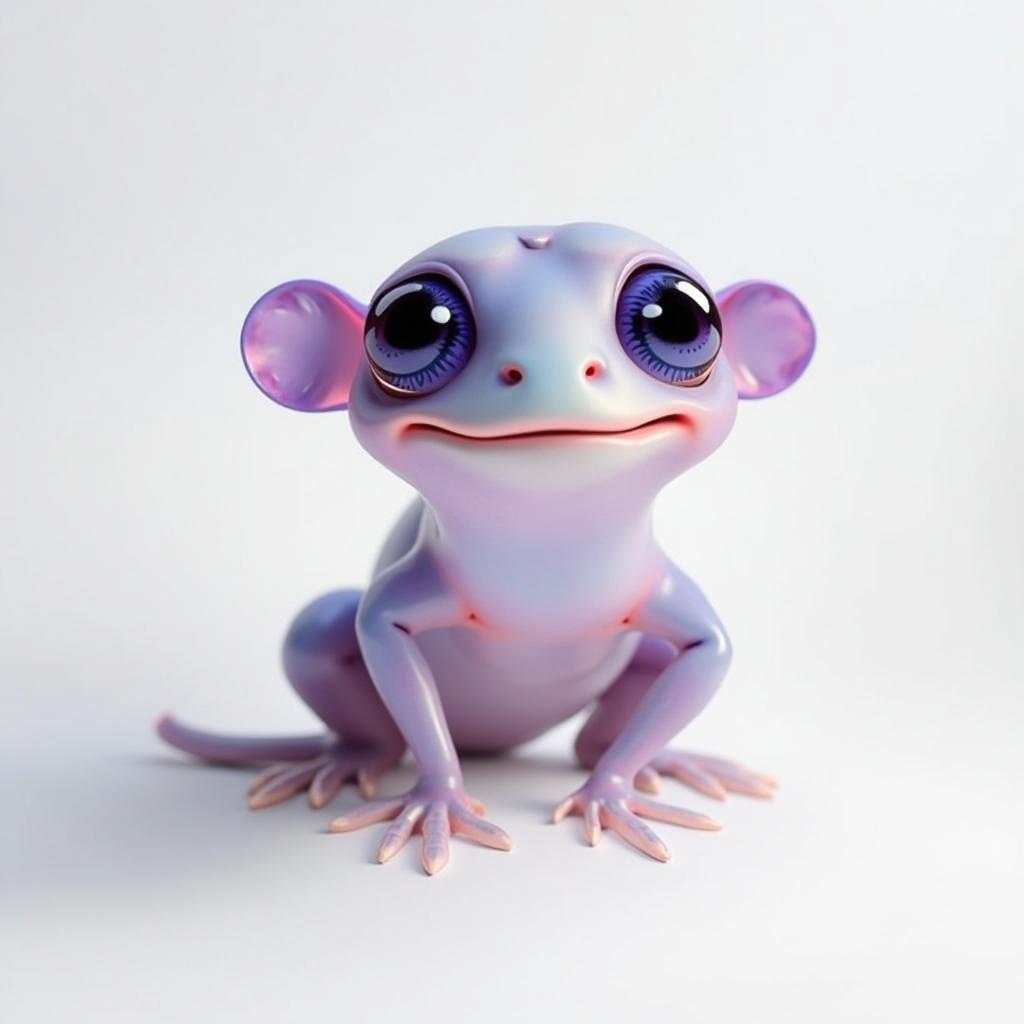} &
        \includegraphics[height=0.18\textwidth]{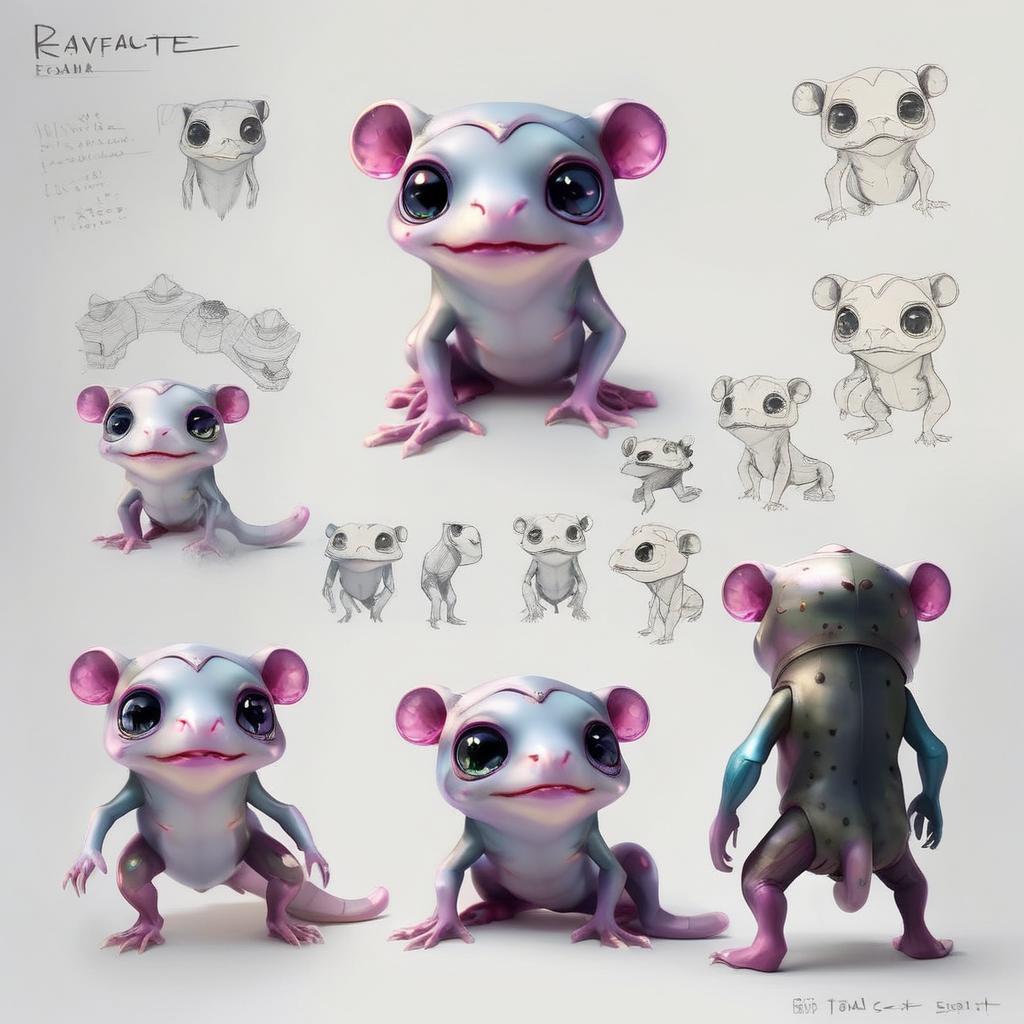} &
        \includegraphics[height=0.18\textwidth]{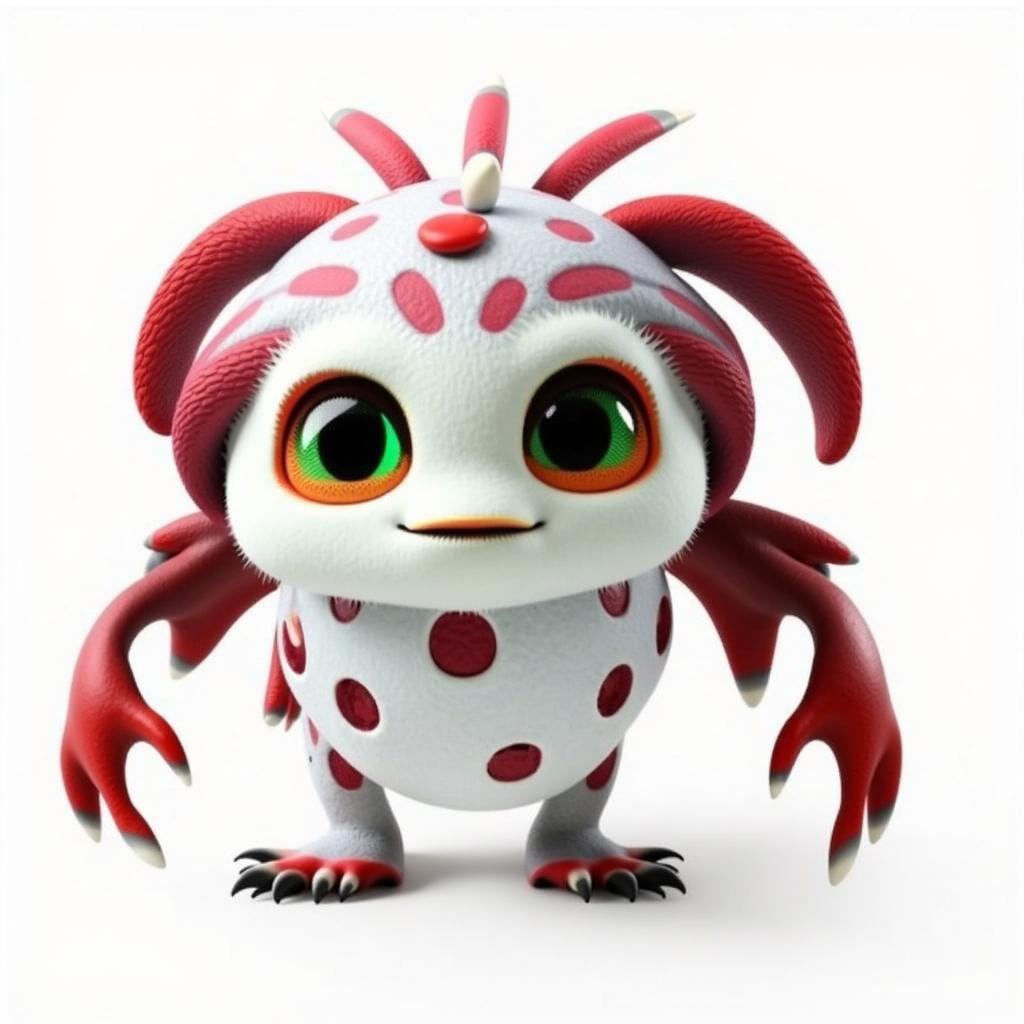} &
        \includegraphics[height=0.18\textwidth]{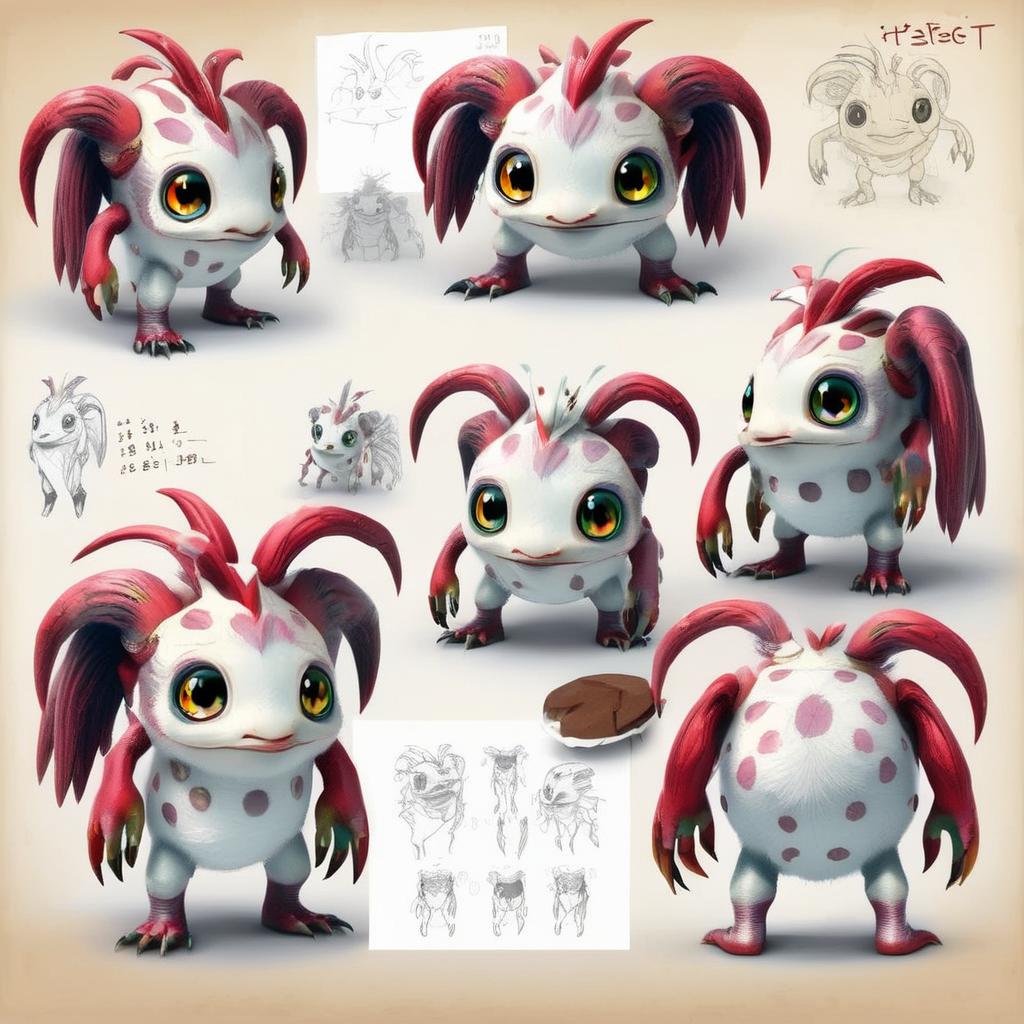} \\

        \includegraphics[height=0.18\textwidth]{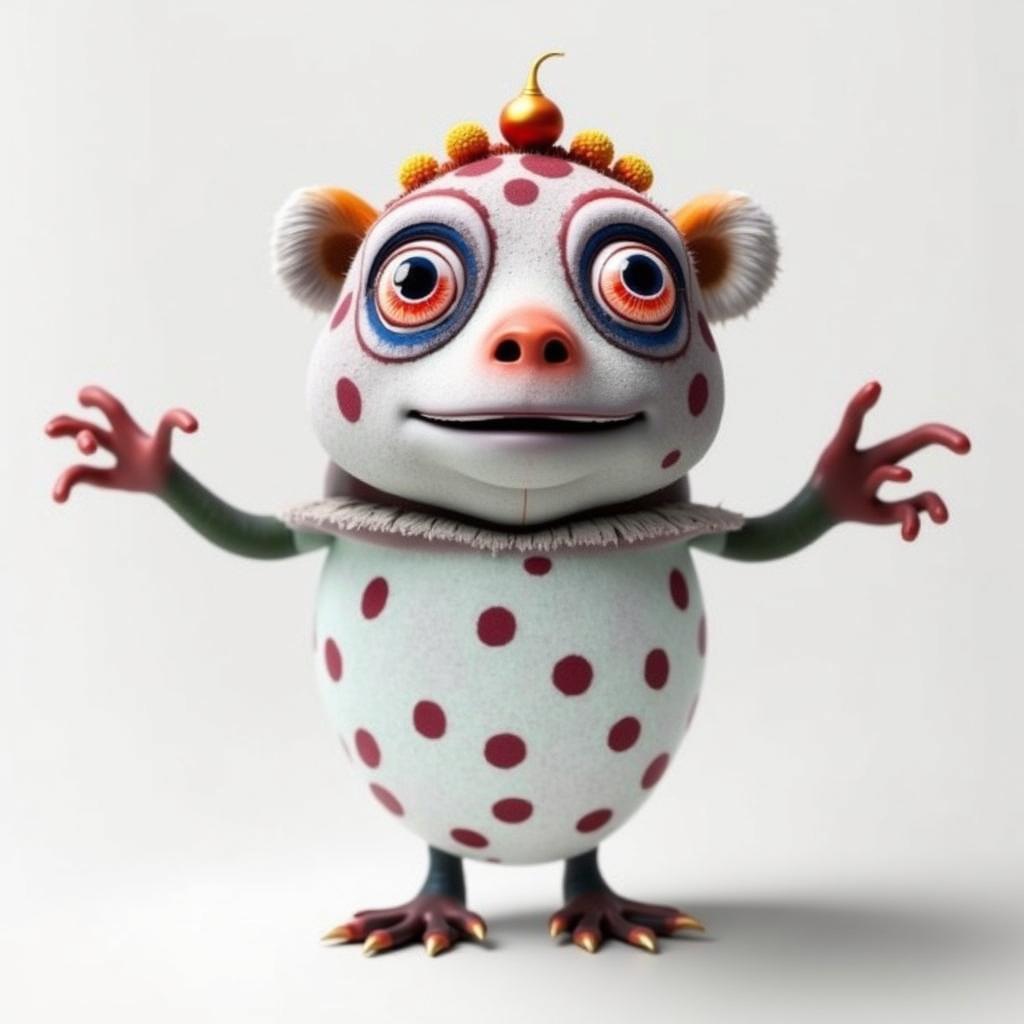} &
        \includegraphics[height=0.18\textwidth]{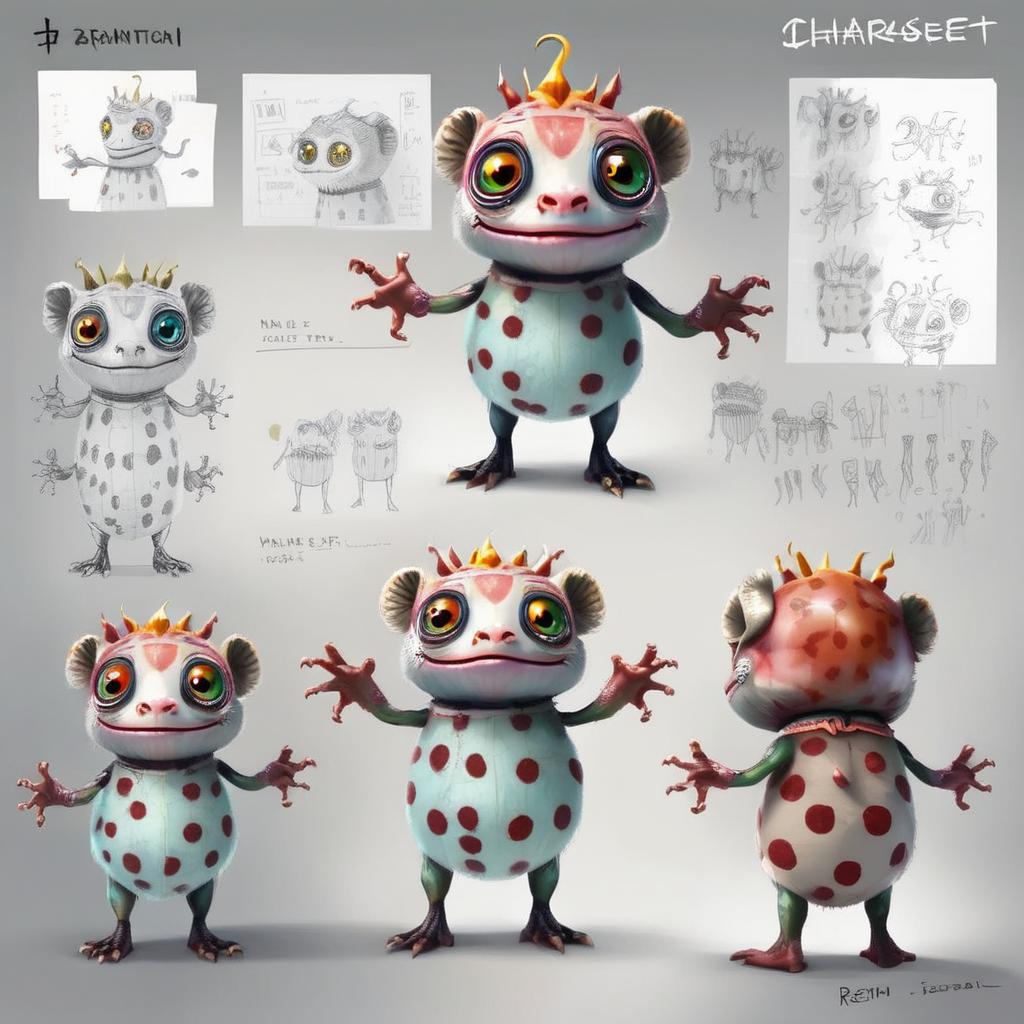} &
        \includegraphics[height=0.18\textwidth]{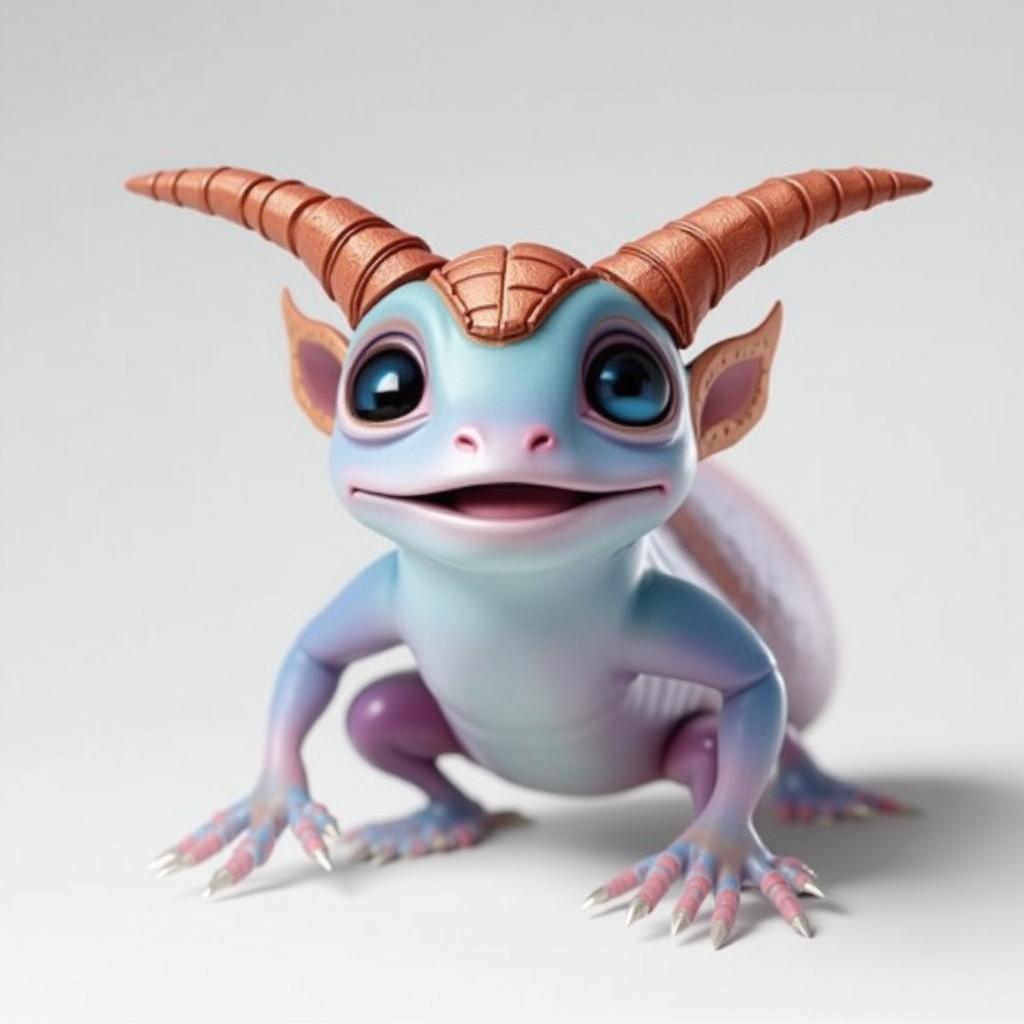} &
        \includegraphics[height=0.18\textwidth]{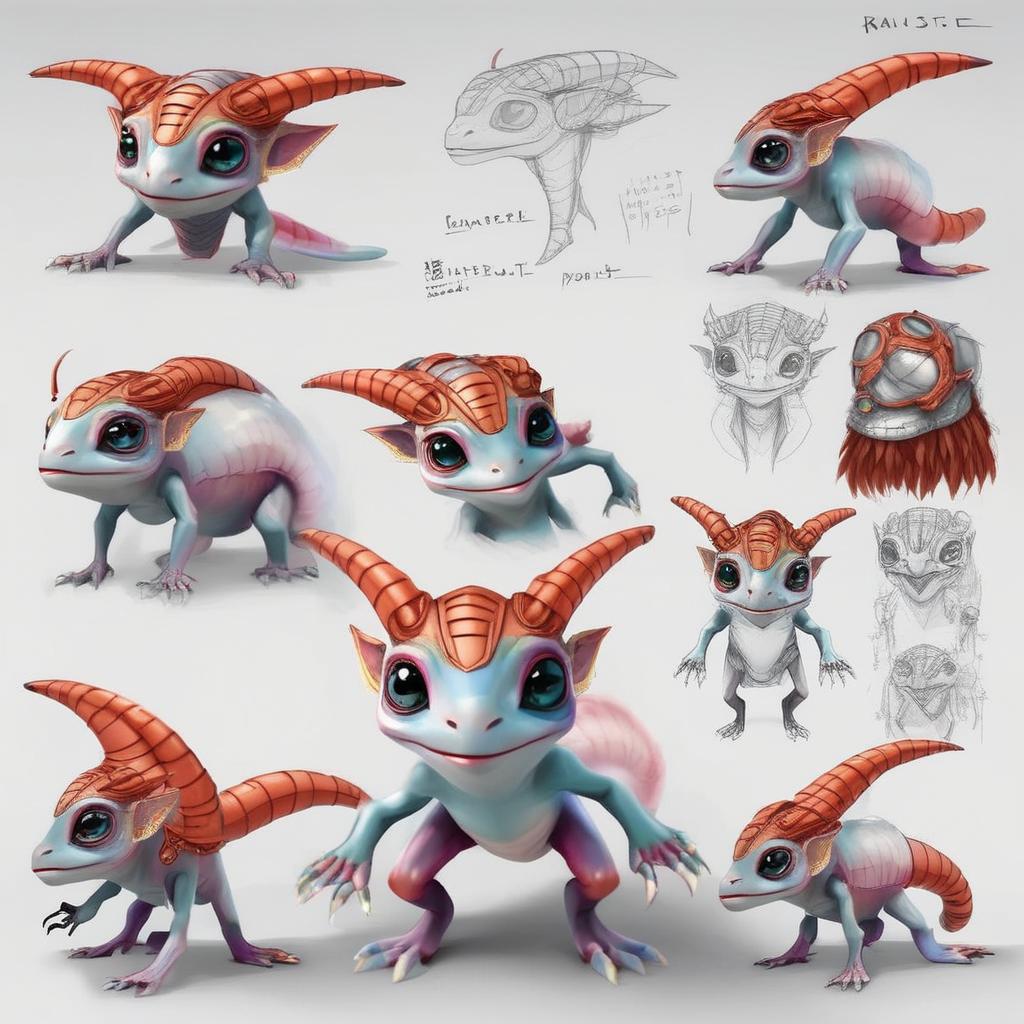} \\

        \includegraphics[height=0.18\textwidth]{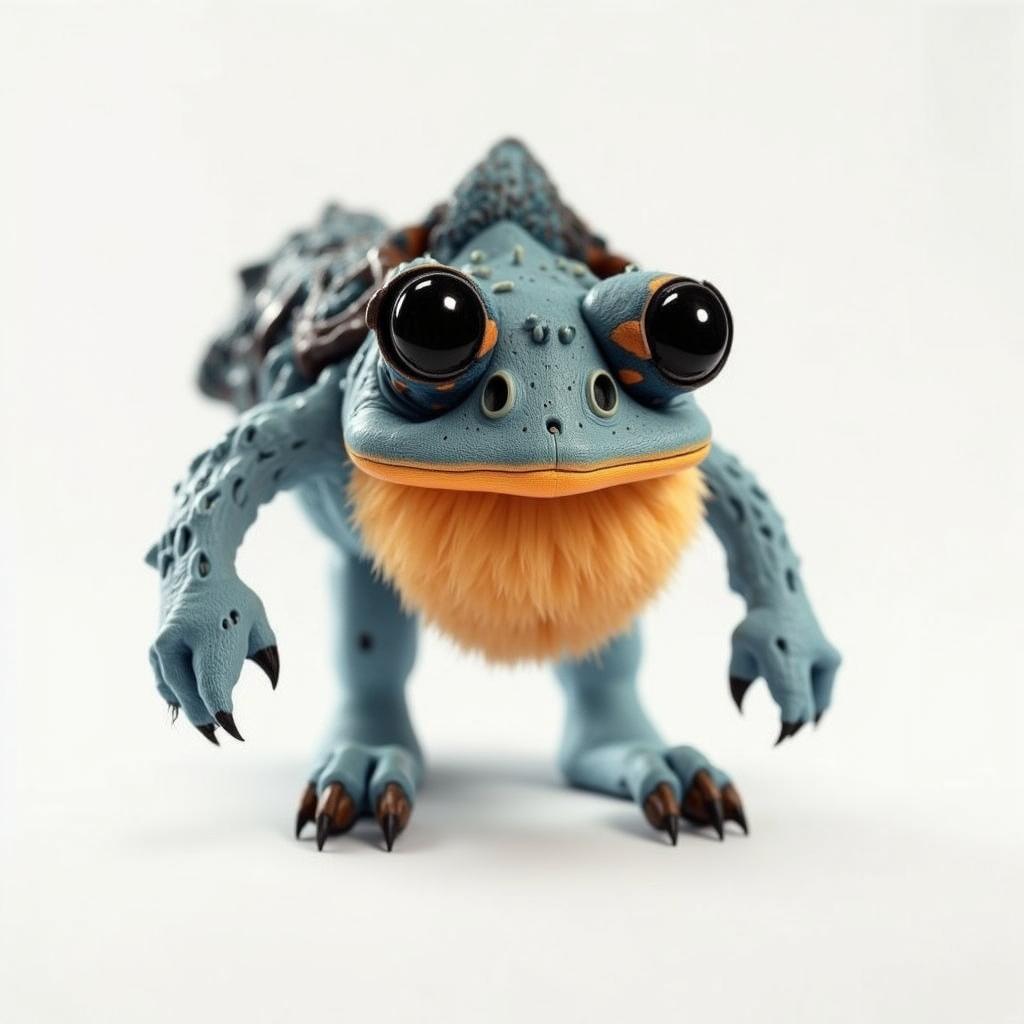} &
        \includegraphics[height=0.18\textwidth]{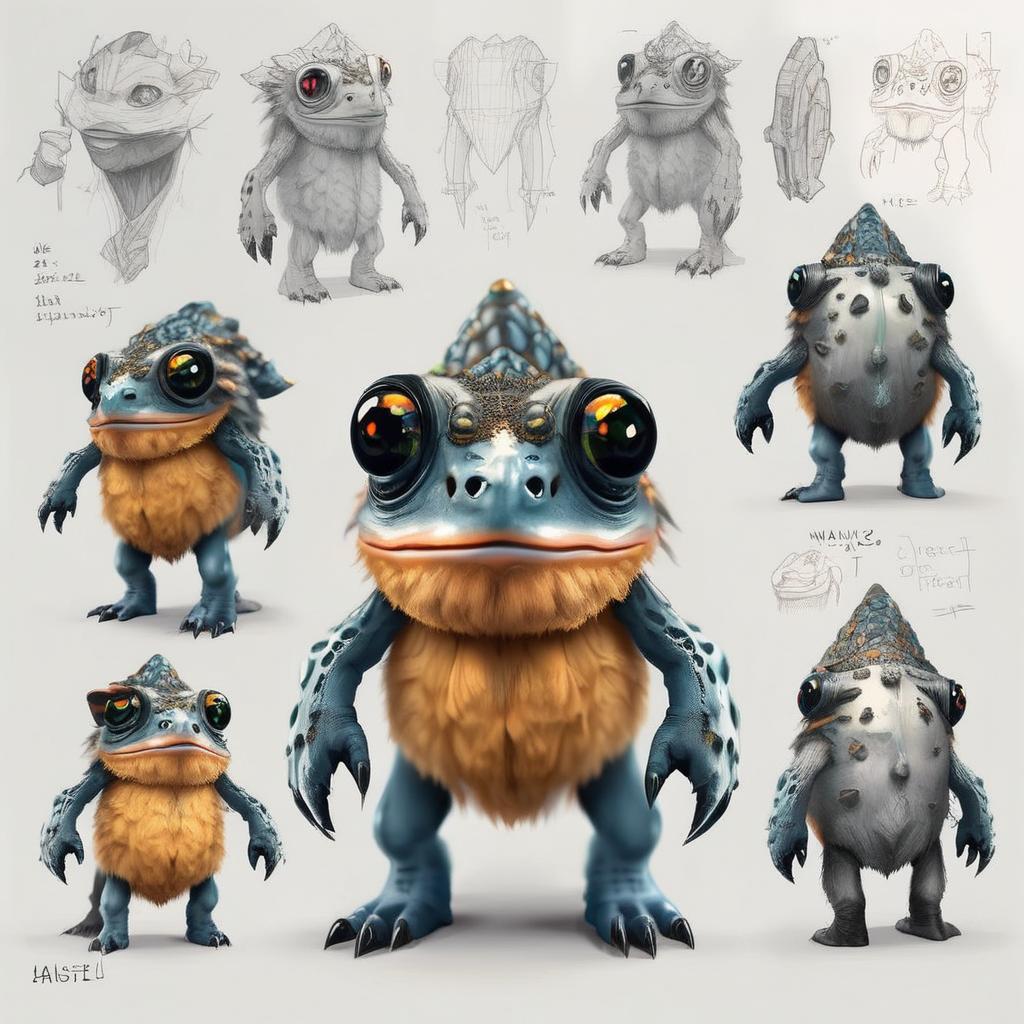} &
        \includegraphics[height=0.18\textwidth]{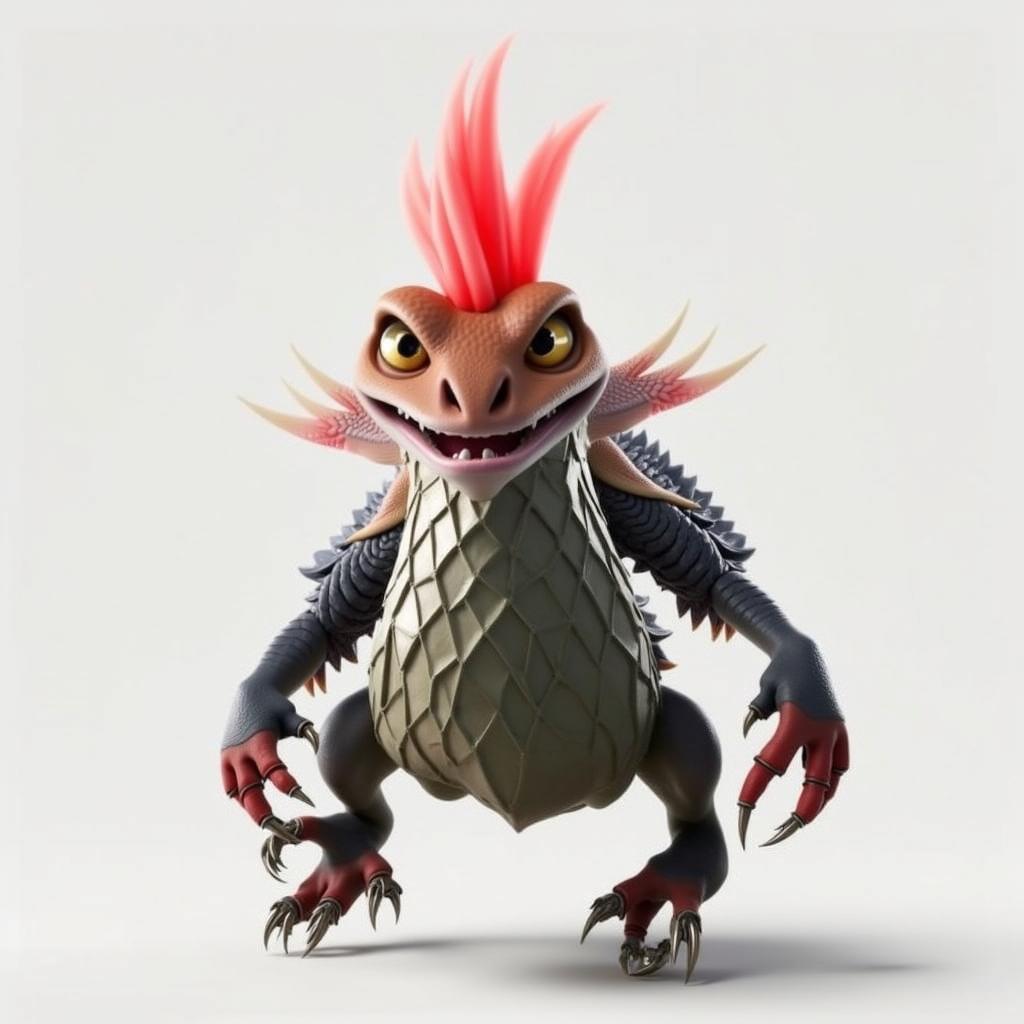} &
        \includegraphics[height=0.18\textwidth]{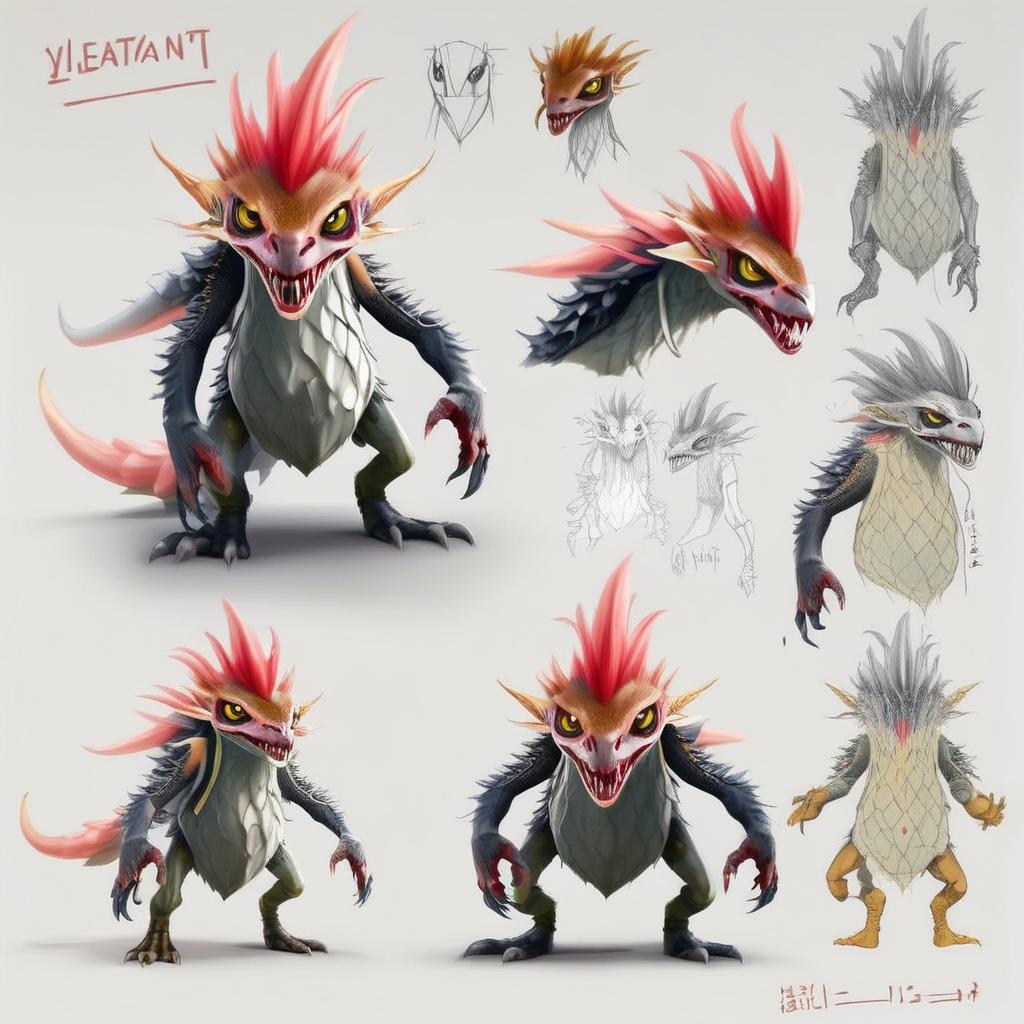} \\

        \includegraphics[height=0.18\textwidth]{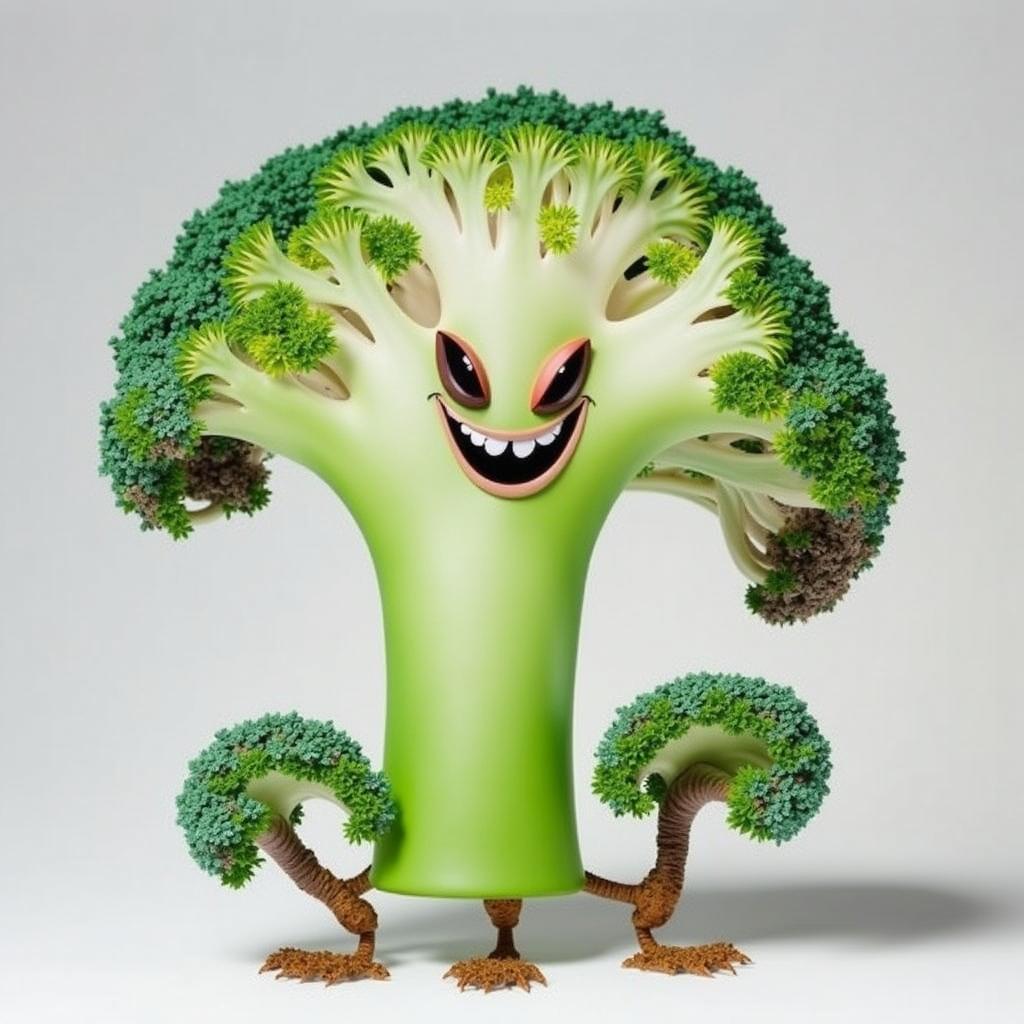} &
        \includegraphics[height=0.18\textwidth]{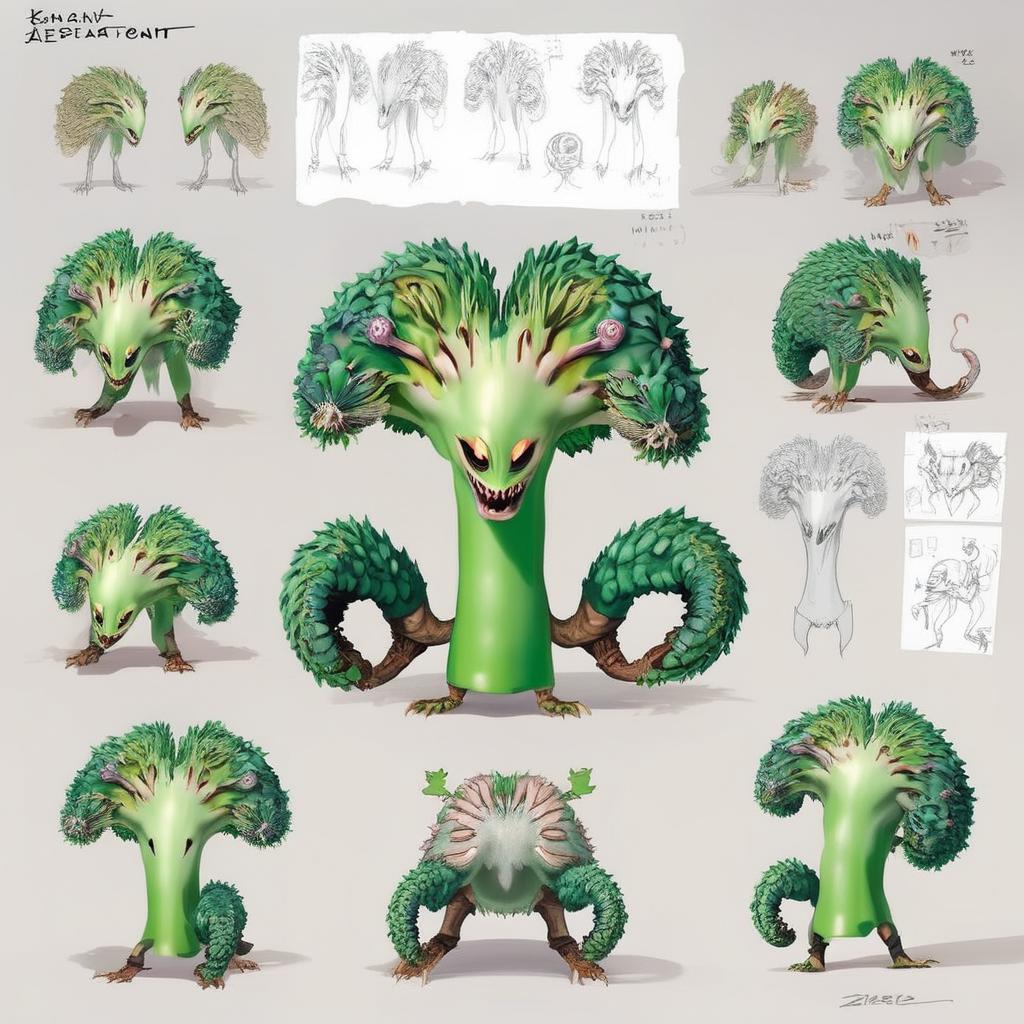} &
        \includegraphics[height=0.18\textwidth]{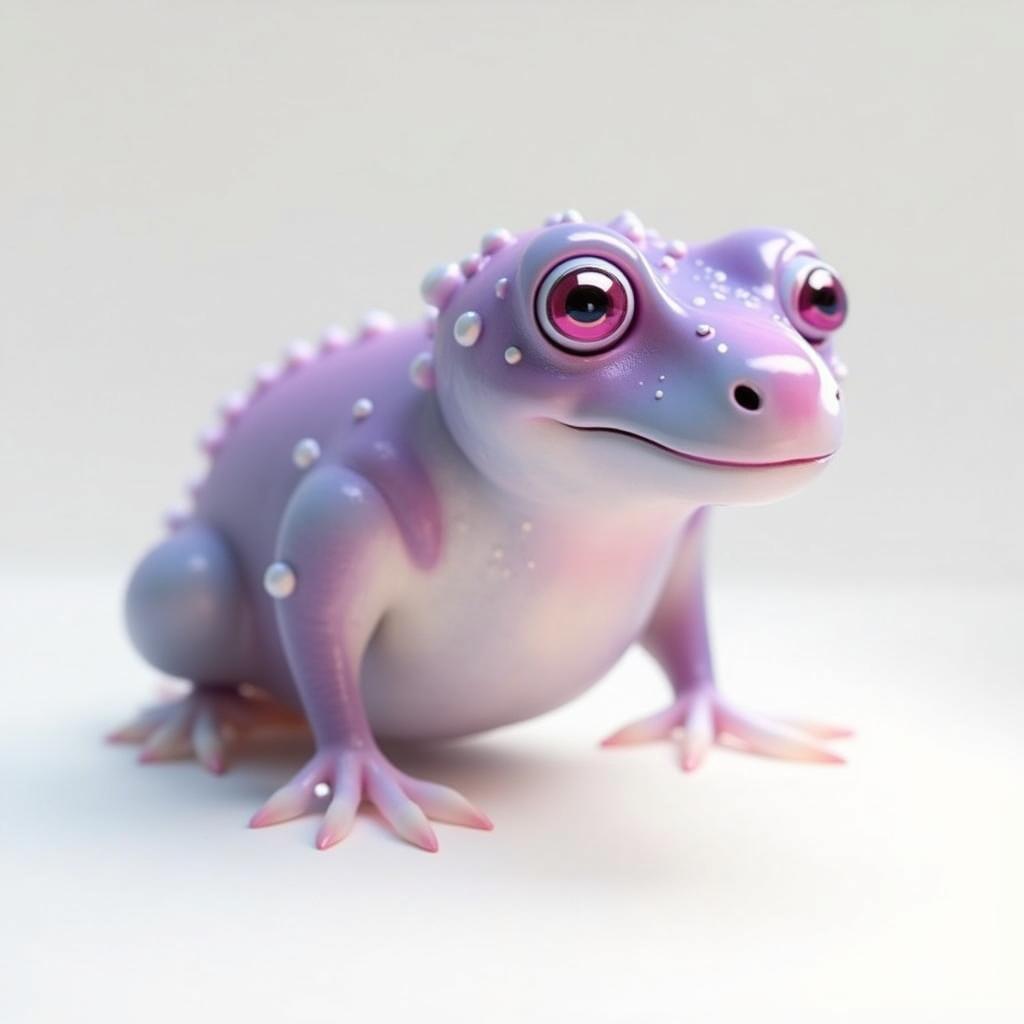} &
        \includegraphics[height=0.18\textwidth]{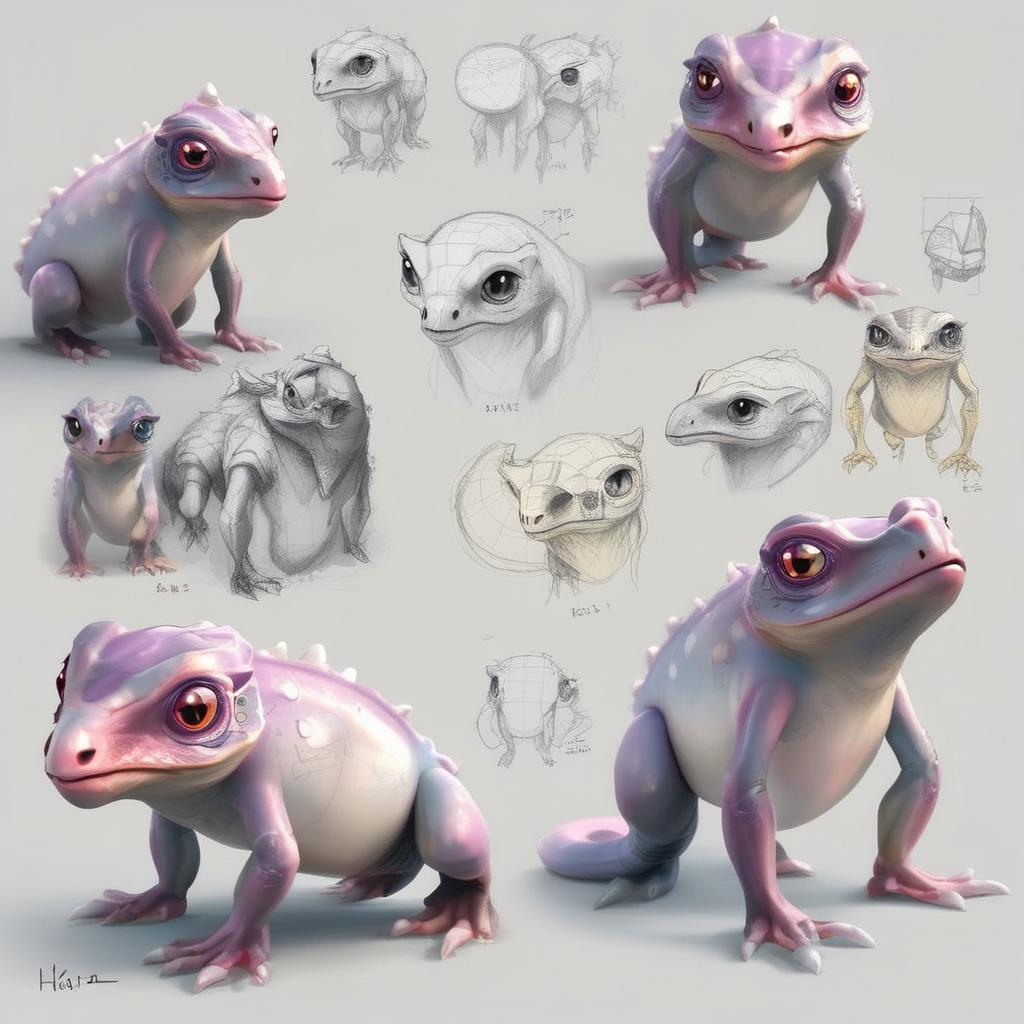} \\

        \includegraphics[height=0.18\textwidth]{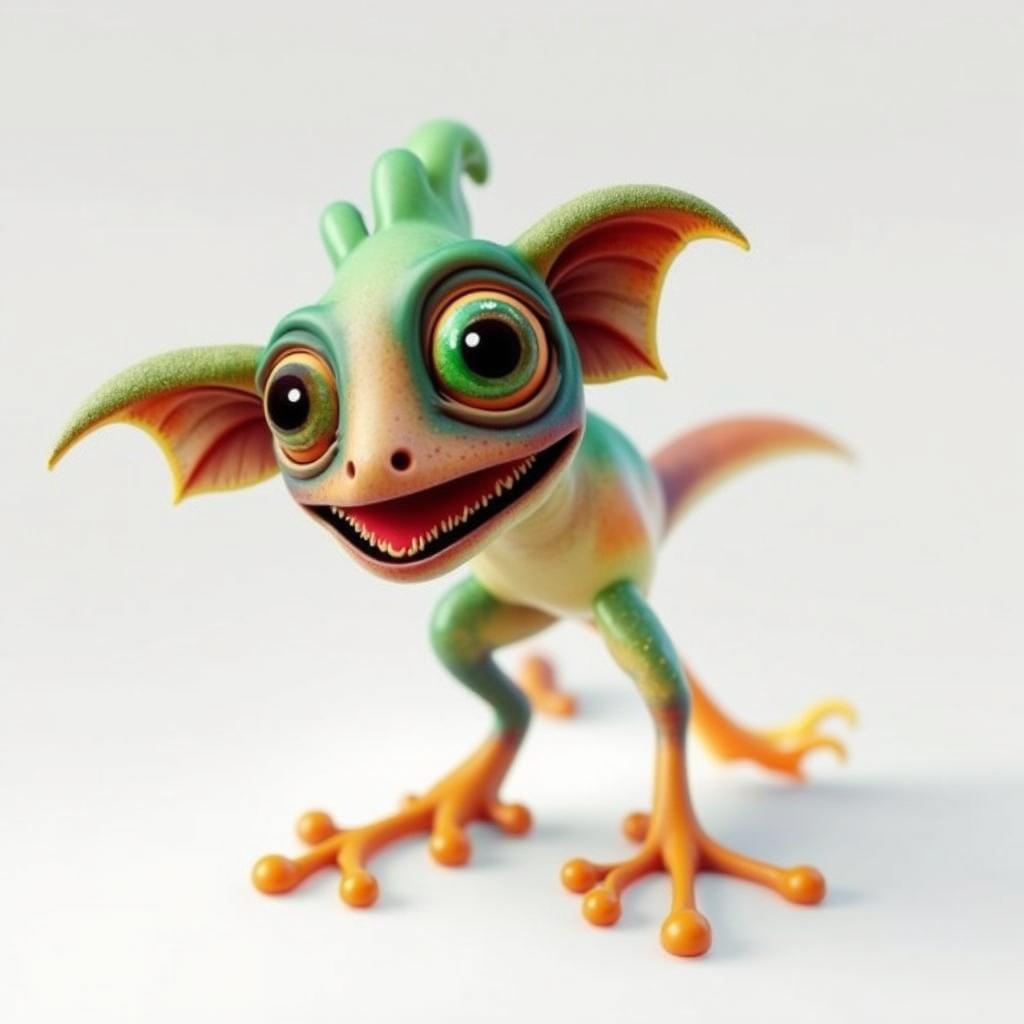} &
        \includegraphics[height=0.18\textwidth]{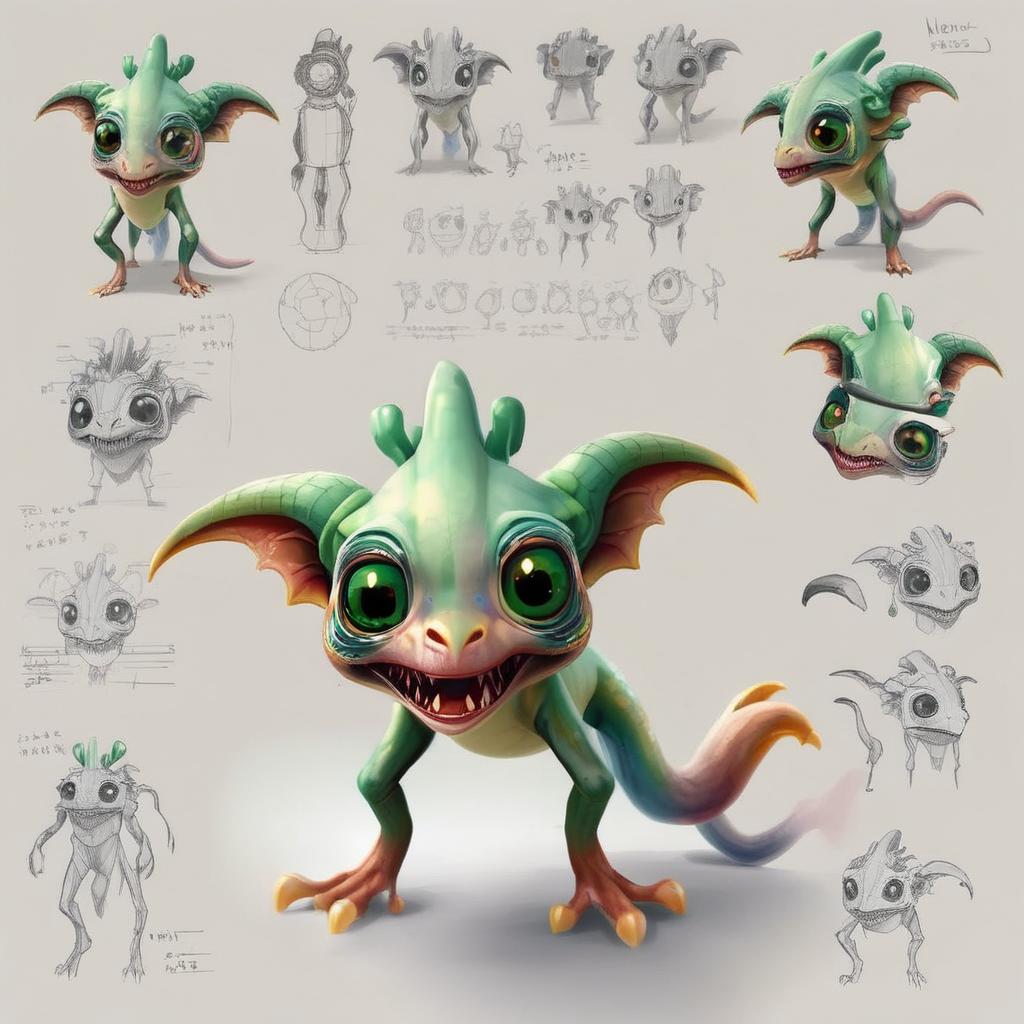} &
        \includegraphics[height=0.18\textwidth]{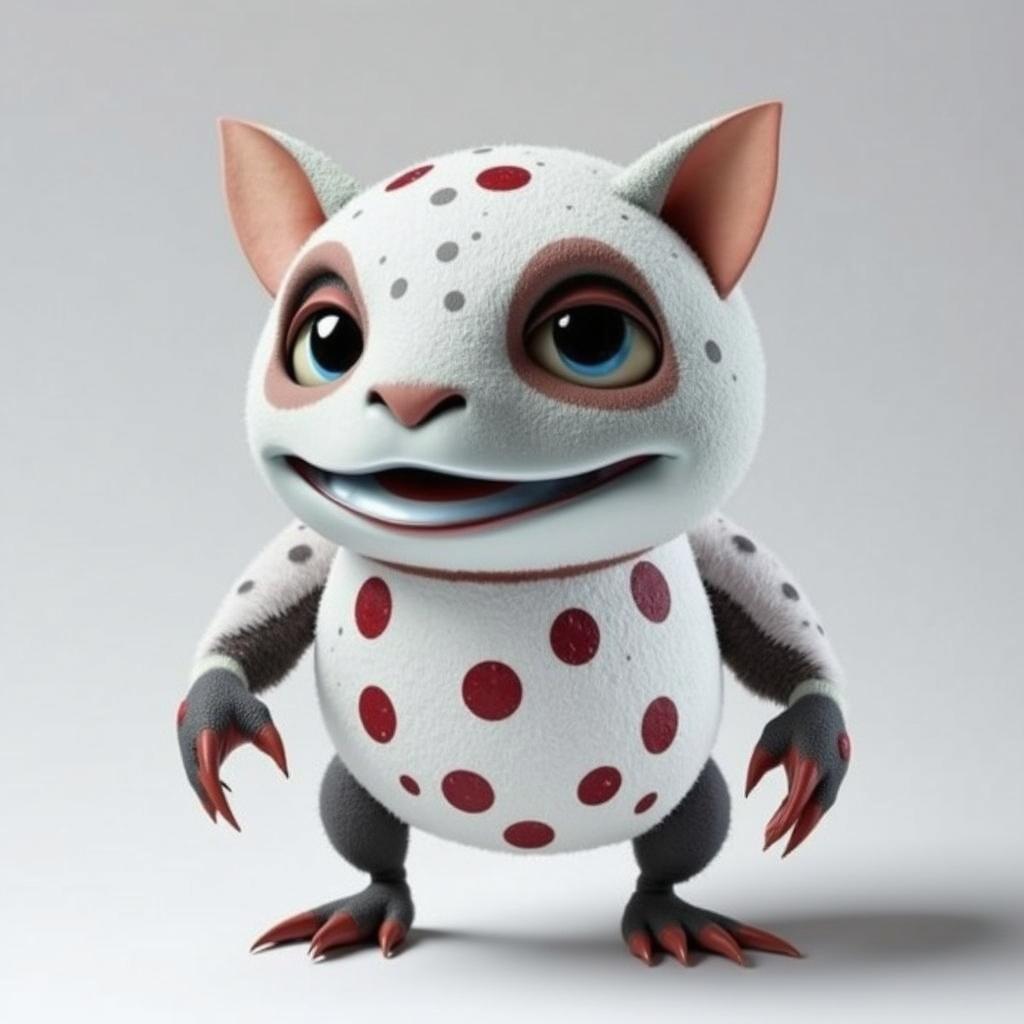} &
        \includegraphics[height=0.18\textwidth]{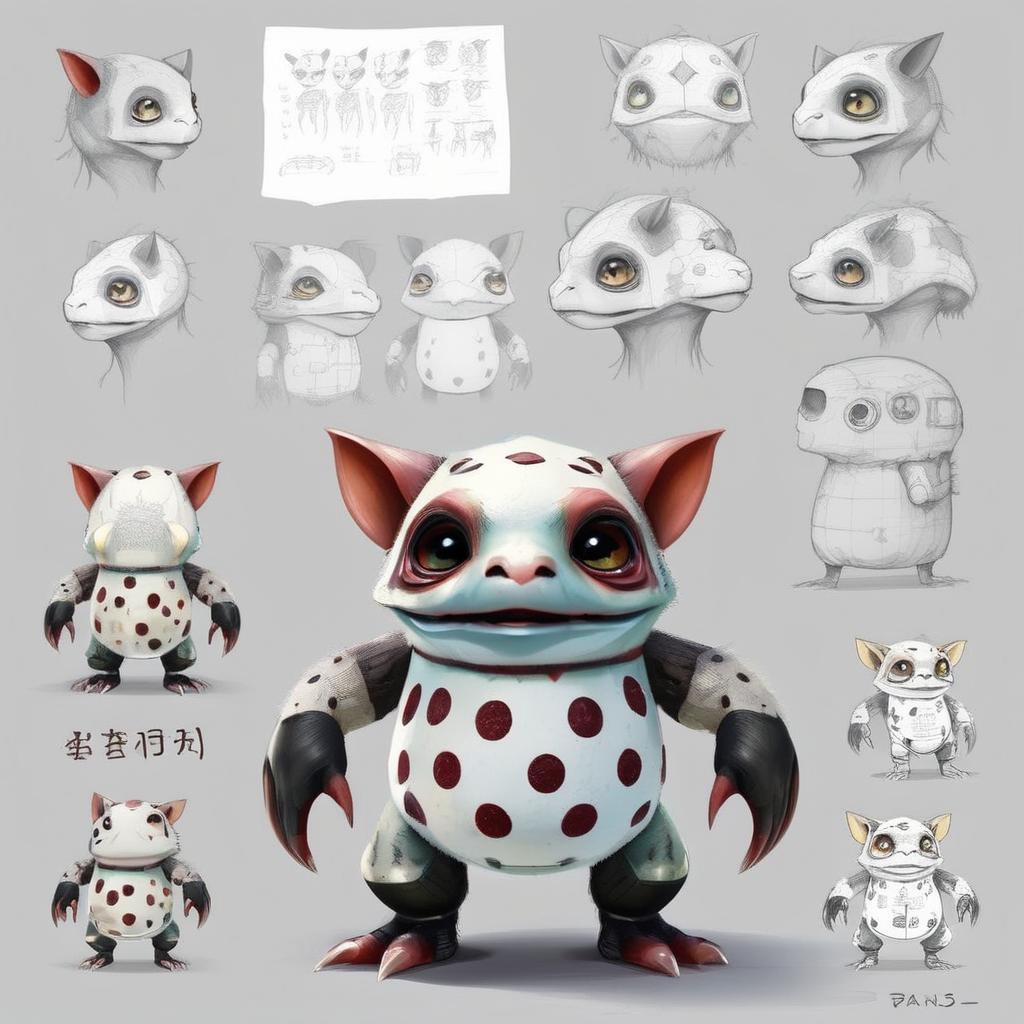} \\

        \includegraphics[height=0.18\textwidth]{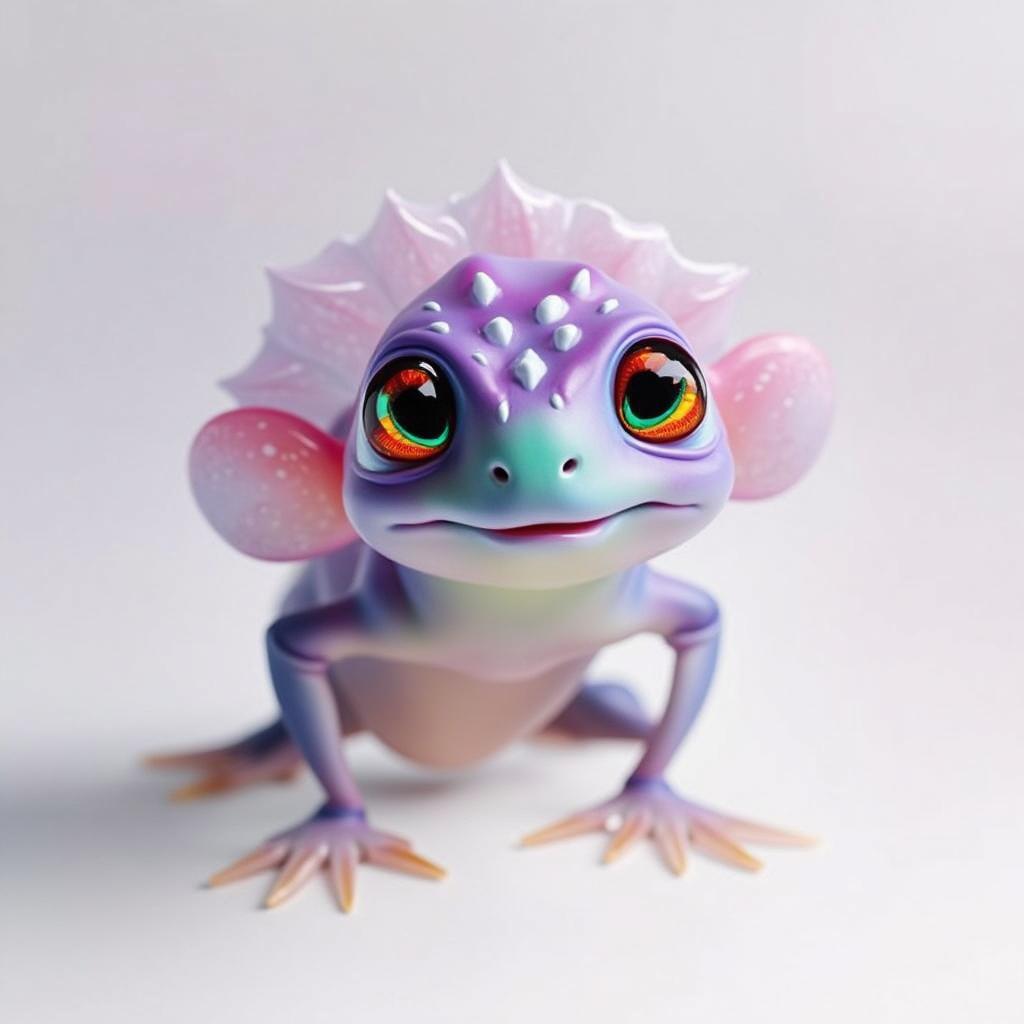} &
        \includegraphics[height=0.18\textwidth]{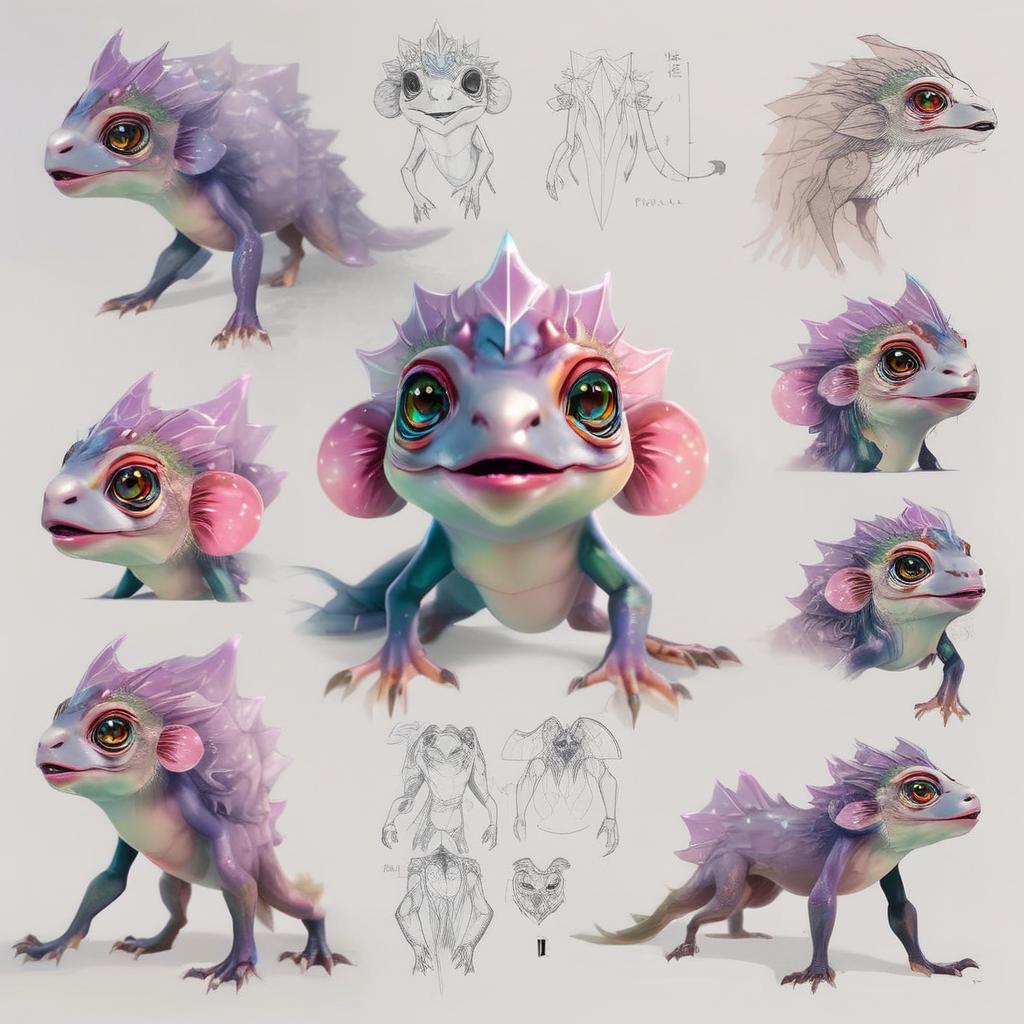} &
        \includegraphics[height=0.18\textwidth]{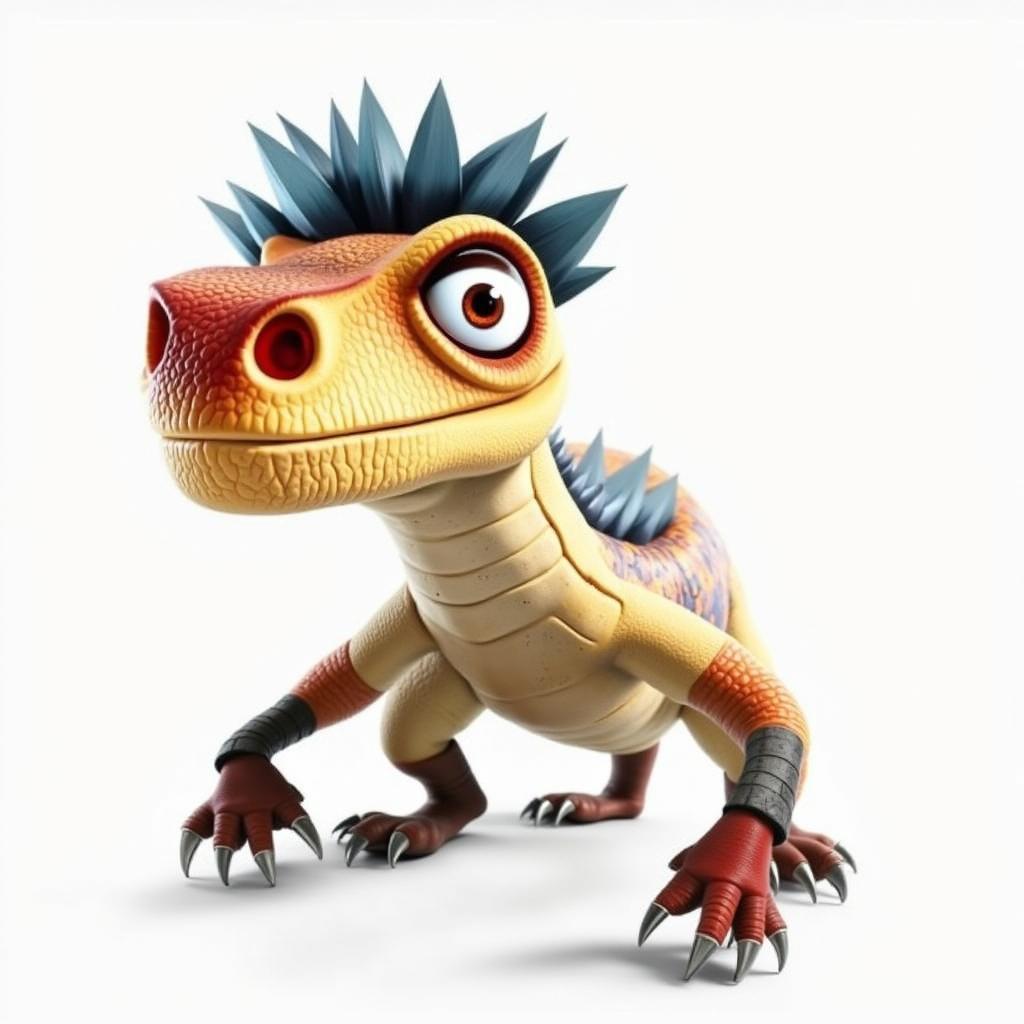} &
        \includegraphics[height=0.18\textwidth]{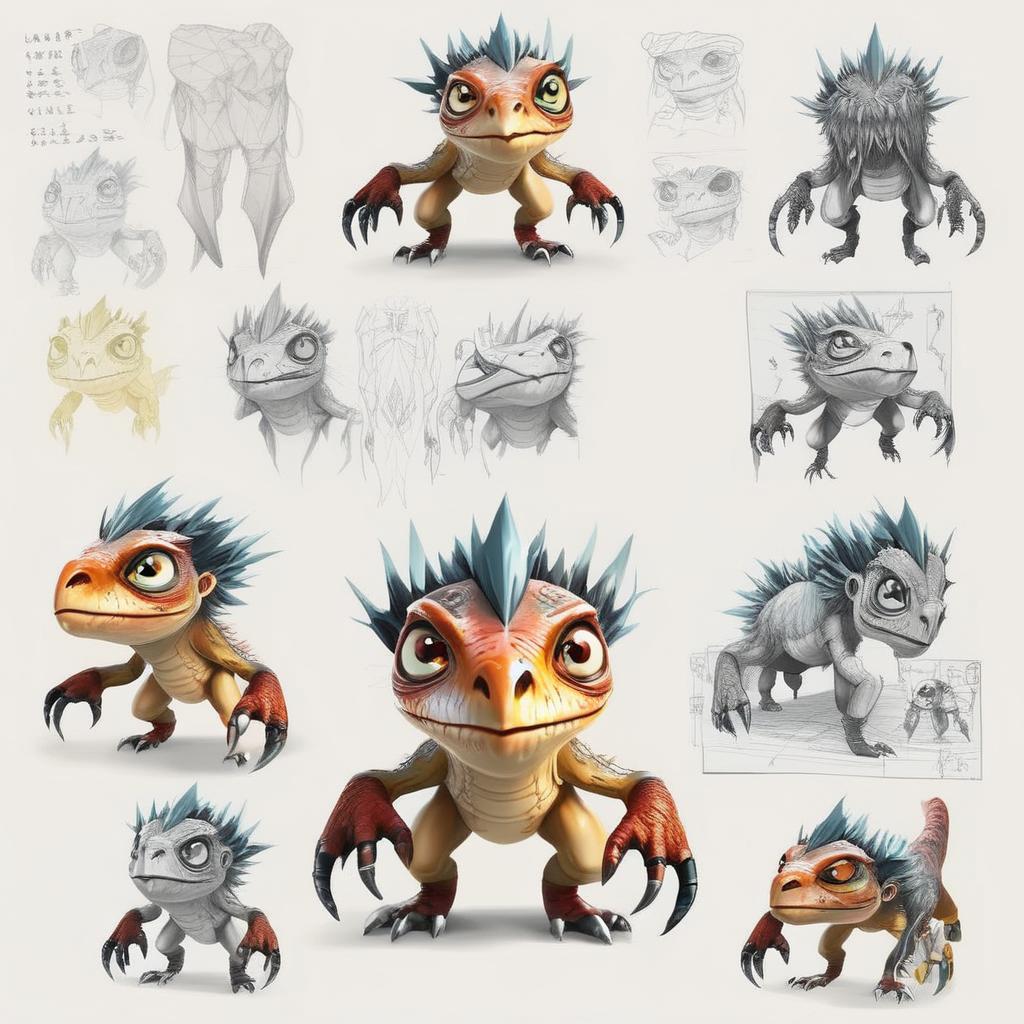} \\

        \includegraphics[height=0.18\textwidth]{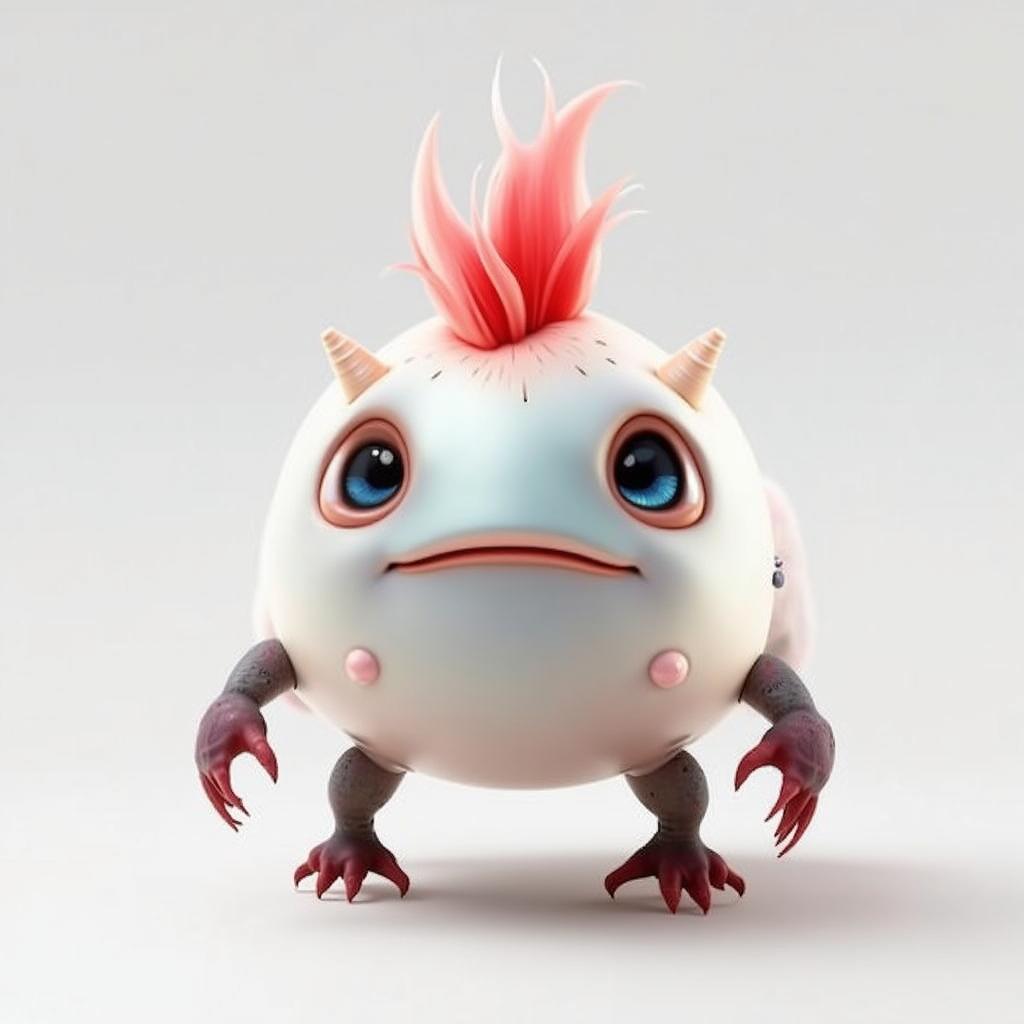} &
        \includegraphics[height=0.18\textwidth]{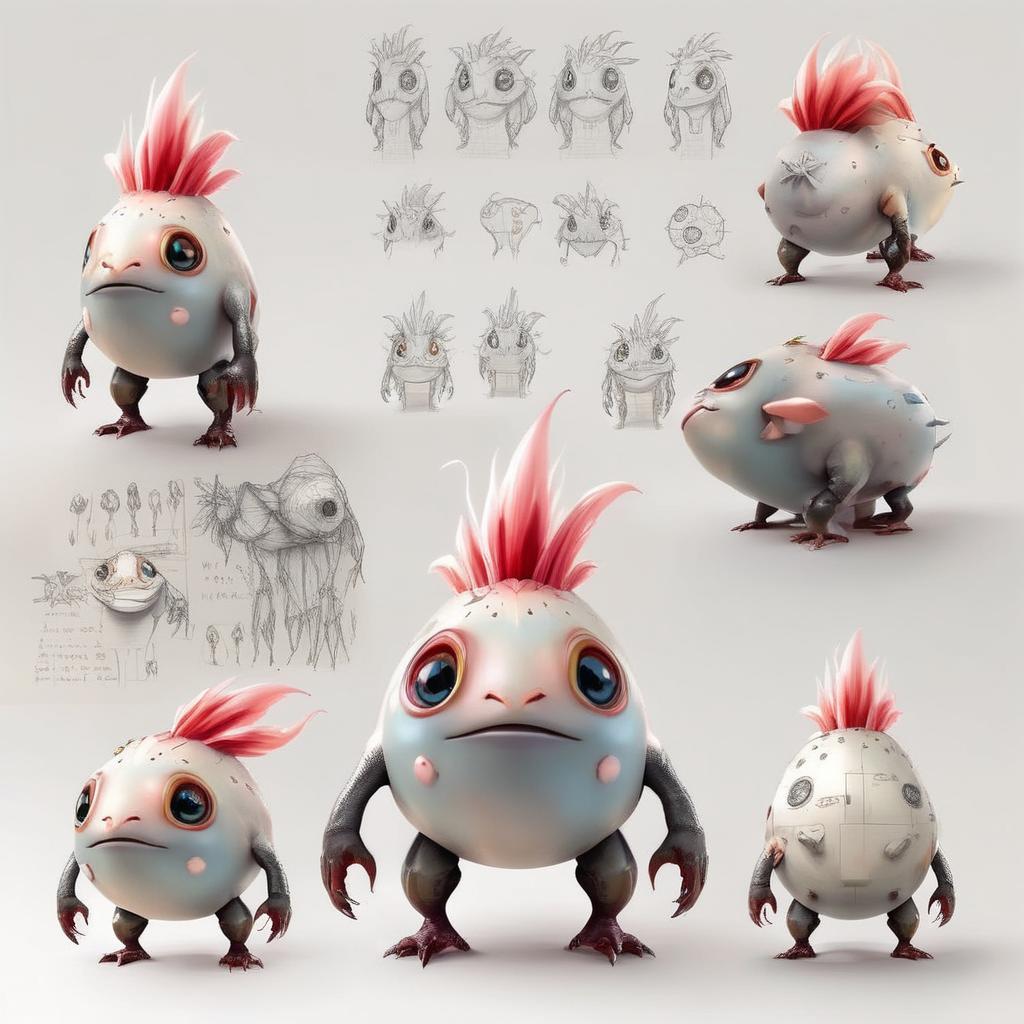} &
        \includegraphics[height=0.18\textwidth]{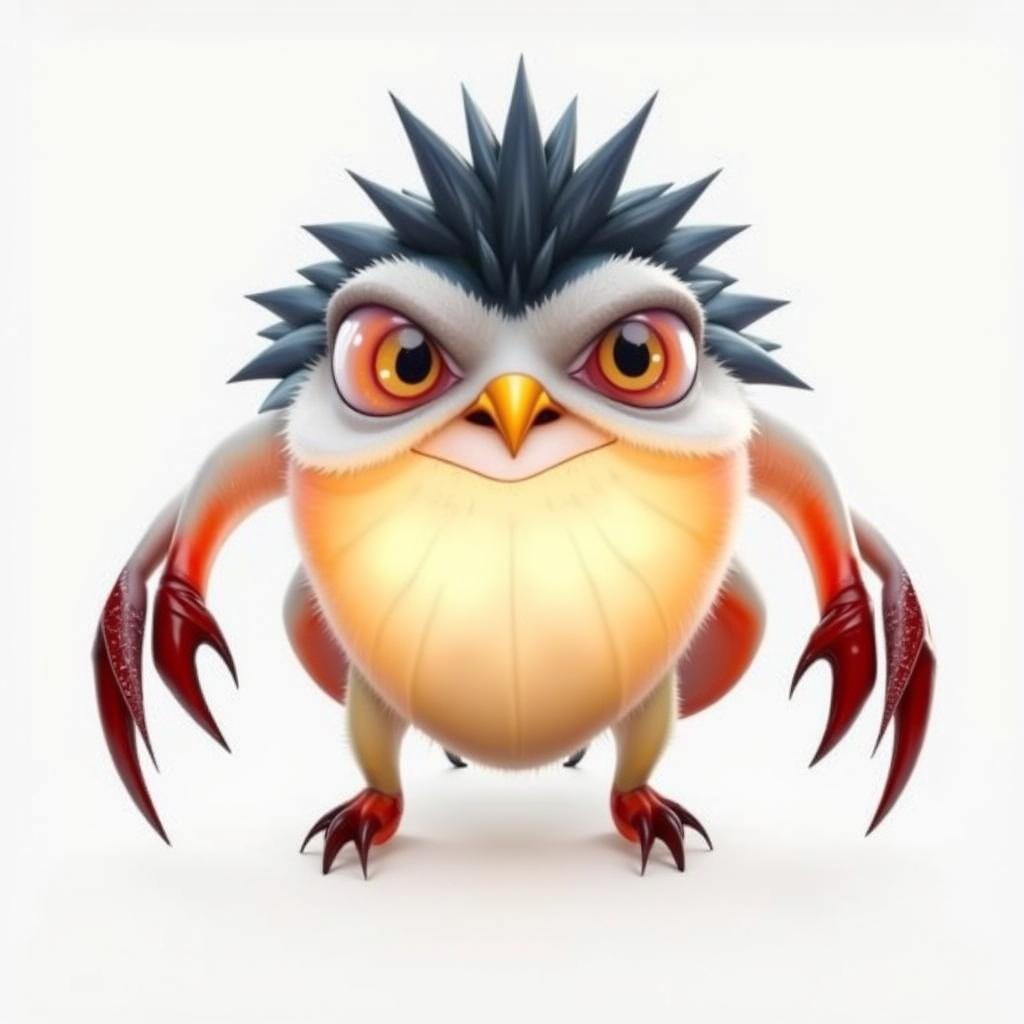} &
        \includegraphics[height=0.18\textwidth]{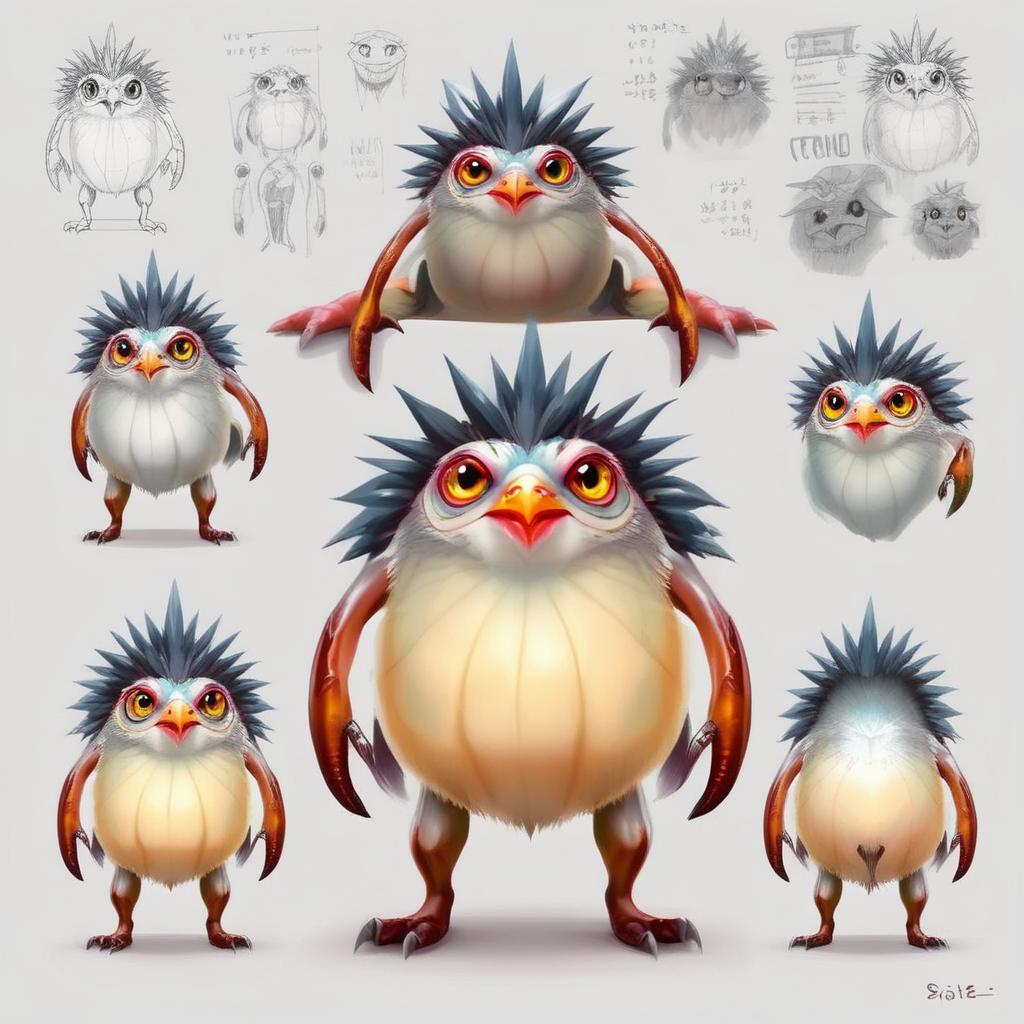} \\

    \end{tabular}
    }
    \vspace{-0.3cm}
    \caption{\textbf{Style Generation.} We train a LoRA to generate character reference sheets when given a concept embedding. 
    }
    \vspace{-0.19cm}
    \label{fig:supp_sheets_big}
\end{figure*}

Below, we provide additional qualitative results and comparisons, as follows: 
\begin{itemize}
    \item In~\Cref{fig:supp_charactes_1} and~\Cref{fig:supp_characters_more} we present additional results for the domain of character ideation, showing our model ability to successfully a wide set of inputs.
    \item In~\Cref{fig:supp_products_1} we present additional results for the product design domain.
    \item In~\Cref{fig:supp_toys_1} and~\Cref{fig:supp_toys_2} we present additional results for the toy concepting domain.
    \item In~\Cref{fig:supp_sheets_big} we present additional results for our reference sheet LoRA, showing its ability to successfully condition on a given image embedding and generating an output that resides in the target style while aligning with the given input.
\end{itemize}

\clearpage

\end{document}